\documentclass[letterpaper]{article}
\usepackage{xr}
\usepackage[table]{xcolor}
\usepackage{tikz}
\usepackage{pgf}
\usepackage{amsmath,mathtools, booktabs, makecell, tabularx}
\usepackage{float}
\floatstyle{plaintop}
\usepackage{caption}
\restylefloat{table}
\usepackage{bbm}
\usepackage{enumerate}
\usepackage{natbib}
\usetikzlibrary{arrows,positioning,automata,calc,fit,shapes.geometric,arrows.meta}
\usepackage{geometry}
\usepackage{abstract}
\usepackage{multicol}
\usepackage{multirow}   
\usepackage{color, colortbl}
\usepackage{caption}
\usepackage{subcaption}
\usepackage{longtable}
\usepackage[T1]{fontenc}
\usepackage{chngcntr}
\usepackage{apptools}
\usepackage{cancel}
\usepackage{rotating}
\usepackage{verbatim}
\usepackage{float}

\usepackage{mathtools}

\definecolor{darkred}{RGB}{100,0,0}
\definecolor{darkgreen}{RGB}{0,100,0}
\definecolor{darkblue}{RGB}{0,0,150}
\usepackage{xr}
\usepackage{dsfont}
\usepackage{xr-hyper}
\usepackage[pdftex,colorlinks]{hyperref}
\hypersetup{colorlinks=true, linkcolor=darkred, citecolor=darkgreen, urlcolor=darkblue}

\usepackage{subfiles}
\usepackage{scalerel,stackengine}
\usepackage{lipsum}

\usepackage[english]{babel}
\usepackage{amsthm}
\usepackage{bbm}
\usepackage{blindtext}
\usepackage{dsfont}
\usepackage{algorithm}
\usepackage{algorithmic}
\usepackage[utf8]{inputenc}

\usepackage{graphicx}
\usepackage{ulem}

\makeatletter
\def\BState{\State\hskip-\ALG@thistlm}
\makeatother
\usepackage{amssymb}
\usepackage[inline]{enumitem}
\usepackage{url} 

\DeclareFontFamily{OT1}{pzc}{}
\DeclareFontShape{OT1}{pzc}{m}{it}{<-> s * [1.10] pzcmi7t}{}
\DeclareMathAlphabet{\mathpzc}{OT1}{pzc}{m}{it}
\usepackage{geometry}
\usepackage[pdftex,colorlinks]{hyperref}

  \hypersetup{
colorlinks =true,
citecolor    = blue,
citebordercolor = violet,
filebordercolor=blue,
linkbordercolor=blue
}

\makeatletter
\newcommand*{\addFileDependency}[1]{
  \typeout{(#1)}
  \@addtofilelist{#1}
  \IfFileExists{#1}{}{\typeout{No file #1.}}
}
\makeatother

\theoremstyle{plain}
\newtheorem{theorem}{Theorem}
\newtheorem{lemma}{Lemma}
\newtheorem{remark}{Remark}
\newtheorem{corollary}{Corollary}
\newtheorem{definition}{Definition}

\newtheorem{cond}{Condition}
\newtheorem{prop}{Proposition}

\newcommand{\ee}{\end{eqnarray}}

\newcommand{\eqnn}{\begin{eqnarray*}}
\newcommand{\een}{\end{eqnarray*}}
\newcommand{\ea}{\end{align}}
\newcommand{\be}{\begin{eqnarray}}

\definecolor{orange}{RGB}{255,127,0}
\definecolor{LightCyan}{rgb}{0.88,1,1}
\definecolor{bured}{rgb}{0.8, 0.0, 0.0}

\DeclareMathAlphabet{\mathpzc}{OT1}{pzc}{m}{it}

\def\spacingset#1{\renewcommand{\baselinestretch}%
{#1}\small\normalsize} \spacingset{1}

\renewcommand{\baselinestretch}{1.66}

\title{Multi-environment Invariance Learning with Missing Data}
\author{Yiran Jia, Jelena Bradic}

\date{\today}

\begin{document}
\maketitle
\begin{abstract}
	Learning models that can handle distribution shifts is a key challenge in domain generalization. Invariance learning, an approach that focuses on identifying features invariant across environments, improves model generalization by capturing stable relationships, which may represent causal effects when the data distribution is encoded within a structural equation model (SEM) and satisfies modularity conditions. This has led to a growing body of work that builds on invariance learning, leveraging the inherent heterogeneity across environments to develop methods that provide causal explanations while enhancing robust prediction. However, in many practical scenarios, obtaining complete outcome data from each environment is challenging due to the high cost or complexity of data collection. This limitation in available data hinders the development of models that fully leverage environmental heterogeneity, making it crucial to address missing outcomes to improve both causal insights and robust prediction. In this work, we derive an estimator from the invariance objective under missing outcomes. We establish non-asymptotic guarantees on variable selection property and $\ell_2$ error convergence rates, which are influenced by the proportion of missing data and the quality of imputation models across environments. We evaluate the performance of the new estimator through extensive simulations and demonstrate its application using the UCI Bike Sharing dataset to predict the count of bike rentals. The results show that despite relying on a biased imputation model, the estimator is efficient and achieves lower prediction error, provided the bias is within a reasonable range.
\end{abstract}


\section{Introduction}\label{section:introduction}

The challenge of out-of-distribution (OOD) generalization—where models need to perform well on unseen data that comes from a different distribution than the training data—has been the focus of extensive research. Two primary approaches have emerged to tackle this issue: distributionally robust optimization (DRO) and invariance learning \citep{liu2021heterogeneous}. Stemming from robust optimization, DRO addresses the OOD problem by minimizing the worst-case error over a range of possible distributions, typically constrained by $f$-divergence or Wasserstein distance, to protect models against potential distributional shifts by focusing on the worst-case scenarios within the uncertainty set \citep{liu2021towards}. Building on causal discovery and causal inference, \cite{peters2016causal} explore invariance of causal mechanisms and propose Invariant Causal Prediction (ICP), which investigates how invariance across multiple environments can help infer causal structures. Subsequent work has relaxed some assumptions of ICP. For example, \cite{pfister2019invariant} leverage sequential data from non-stationary environments to detect causal relations in multivariate linear time series, while \cite{rothenhausler2021anchor} relax the conditions of the standard IV model, assuming a linear SCM with additive interventions.

Inspired by the idea of invariance in prediction under causal models, as exploited by ICP, \cite{arjovsky2019invariant} recognize the potential of applying this concept to improve generalization in machine learning models and shift the focus toward learning invariant data representation across environments without requiring explicit causal modeling. Following the IRM framework to handle spurious correlations, subsequent works have introduced variations and refined regularization techniques to further enhance generalization  \citep{chang2020invariant, ahuja2020invariant, krueger2021out, creager2021environment, mahajan2021domain, yin2024optimization}. Among these, \cite{fan2024environment} take a statistical approach, proposing an environment-invariant linear least squares (EILLS) objective function to eliminate linear spurious variables and select a set of important variables, including the true causal ones, thereby achieving robust prediction in unseen environments. Their method provides a non-asymptotic analysis under flexible SEM with weaker identification conditions, making it more sample-efficient, particularly in settings with fewer environments.

While previous works in invariance learning frameworks that leverage the heterogeneity across multiple environments have advanced our understanding of how to tackle distribution shifts for robust prediction and, under certain conditions, help identify sets that contain causal variables, these methods typically assume fully available outcome data across all environments. However, in many real-world applications, obtaining complete outcomes is challenging due to cost constraints or incomplete data collection. This scarcity of labeled outcome data mirrors the challenge faced in semi-supervised learning (SSL), where unlabeled data is plentiful, but acquiring labels is costly and time-consuming. While many algorithms have been developed for semi-supervised learning across various applications \citep{ouali2020overview, van2020survey, zhu2022introduction, yang2022survey, mey2022improved, duarte2023review}, there is an emerging focus on methods that emphasize estimation and statistical inference \citep{gronsbell2018semi, chakrabortty2018efficient, zhang2019semi, azriel2022semi, zhang2022high, angelopoulos2023ppi++, song2024general}.

For example, \cite{zhang2019semi} develop a general semi-supervised inference framework focused on the estimation of the population mean, with applications such as treatment effect estimation.  \cite{zhang2022high} introduce a double robust estimator for the outcome's mean, achieving root-n inference in high-dimensional settings.  \cite{chakrabortty2018efficient} propose a two-step efficient estimator for parameters in a working linear regression model, incorporating semi-nonparametric imputation of the outcome for the unlabeled data.  \cite{azriel2022semi} introduce a different method to reduce the variance and prediction error of best linear predictor by extracting additional efficiency from the least squares estimator, without requiring an imputation model. \cite{angelopoulos2023ppi++} propose PPI++, a methodology for estimating and inferring parameters defined as the minimizer of a convex loss when only a small amount of labeled data is available. They assume access to a pretrained machine learning model for imputation and use both gold-standard labeled observations and imputed ones to construct a rectified loss function. This loss function, adjusted with a debiasing term to account for imputation, remains unbiased for the true objective. Consequently, the minimizer of the rectified loss, referred to as the prediction-powered point estimate, is a consistent and asymptotically normal estimator of the true parameter of interest. The PPI++ framework proposed by \cite{angelopoulos2023ppi++} aligns with key themes from influential works in semiparametric inference, causality, and missing data, such as those by \citet{bickel1993efficient, robins1994estimation, robins1995semiparametric, tsiatis2006semiparametric, chernozhukov2018double}. These works, along with \cite{angelopoulos2023ppi++} share the spirit of doubly robust methods, where combining two models—typically one for the outcome and another for the treatment or missingness mechanism—ensures reliable inference even if one model is misspecified. Moreover, when both models are correctly specified, these methods achieve optimal efficiency.

The works in the semi-supervised learning literature shed light on how we can effectively address the issue of incomplete outcome data by appropriately utilizing all available information from each environment, adjusting for any biases introduced through imputation. This helps us pursue our ultimate goal of identifying a set of variables that exhibit invariant relationships across environments and constructing a robust predictor that maintains accuracy even in unseen environments, ultimately moving toward bridging the gap between missing outcomes and robust out-of-distribution generalization.

In this work, we propose an adaptation of the invariance learning framework that accommodates missing outcome data while ensuring consistent estimation. We leverage machine learning models to impute the missing labels and incorporate both labeled and imputed data into the objective. Crucially, instead of treating imputed values as gold-standard observations, we construct a debiased objective function that accounts for any bias introduced in the imputation process.

We first assume a given deterministic imputation model for each environment, such as a pre-trained machine learning model, which serves as a reasonable approximation for the missing labels in that environment but may not perfectly align with the true regression function. In this case, the imputation bias—defined as the mean difference between the true labels and their imputed values—is generally non-zero. Building on insights from \cite{fan2024environment}, who conducted non-asymptotic statistical analysis of their EILLS objective assuming fully observed outcomes, we extend these techniques to accommodate missing outcome data. Through non-asymptotic statistical analysis, we explore how the interaction between the ratio of missing data and the imputation bias shapes the convergence behavior of our estimator. We also identify key conditions under which the imputation bias can be mitigated, ensuring reliable variable selection and consistent estimation. 

While our work introduces an extension to invariance learning that addresses missing outcomes through debiased objectives and imputation with non-asymptotic guarantees, other efforts have tackled related challenges involving robust prediction and missing data. For instance, building on prior work that assumes fully observed spurious variables \citep{makar2022causally, veitch2021counterfactual, puli2021predictive}, \cite{goldstein2022learning} focus on learning invariant representations in settings where some spurious variables are only partially observed, rather than addressing missing outcome data. They propose a doubly robust kernel-based Maximum Mean Discrepancy (MMD) estimator, which they incorporate into the regularization term of the main objective to ensure consistency under partial observation of spurious variables. In contrast, our approach is situated within a broader framework of invariance risk minimization, which leverages the heterogeneity across multiple environments rather than requiring the explicit identification of spurious variables. Instead, we focus on addressing the challenge of missing outcomes in each environment, providing a solution with theoretical guarantees for reliable estimation of parameters associated with invariant variables and robust prediction in unseen environments.

Another line of research that may seem closely related is semi-supervised domain adaptation, with works like \cite{li2021learning} addressing the challenge of training models under distributional shifts between a source and target domain. These approaches rely on labeled data from the source domain and a small amount of labeled data from the target domain to improve performance on that specific target distribution. In contrast, we assume that data within each environment follows the same underlying distribution and use semi-supervised learning techniques to address the challenge of missing outcomes in each environment. Our primary objective is to identify invariant variables across multiple environments, enabling robust prediction in any unseen environment, without requiring a predefined target domain.

The remainder of this paper is structured as follows. In Section \ref{section:background}, we formally introduce the problem setup, beginning with the notation, data-generating process, the concept of invariance, and the missingness mechanism across environments. Section \ref{section:preliminary} reviews the objective from \cite{fan2024environment}, which relies on fully labeled observations. In Section \ref{section:methodology}, we build from the estimation of coefficients in a working linear model under missingness, gradually extending to the estimation of invariant variables and introducing our main objective. Section \ref{section:theory} provides theoretical guarantees along with their potential implications. In Section \ref{section:simulation}, we demonstrate the performance of our method using both simulated and semi-synthetic data. Finally, in Section \ref{section:conclusion}, we conclude the paper by discussing promising directions for future research.

\section{Background}\label{section:background}

\subsection{Notation}\label{subsection:notation}

We introduce the notation used throughout the paper. Let $\mathbb{R}^+$ and $\mathbb{N}^+$ denote the sets of positive real numbers and positive integers, respectively. We denote $[p] = \{1, \dots, p\}$ as the index set of $p$ variables. Let $S$ be an arbitrary subset of $[p]$, and let $|S|$ denote its cardinality. We define $a \vee b = \max \{a, b\}$ and $a \wedge b = \min \{a, b\}$. The notation $a(n) = O(b(n))$ indicates that there exists a universal constant $C > 0$ such that $a(n) \leq C \cdot b(n)$ for all $n \in \mathbb{N}^{+}$. Similarly, $a(n) = \Omega(b(n))$ means there exists a constant $c > 0$ such that $a(n) \geq c \cdot b(n)$ for all $n \in \mathbb{N}^{+}$. We write $a(n) \asymp b(n)$ if both $a(n) = O(b(n))$ and $a(n) = \Omega(b(n))$ hold. The notation $a(n) \ll b(n)$, $b(n) \gg a(n)$, or $a(n) = o(b(n))$ signifies that $\limsup _{n \to \infty} (a(n) / b(n)) = 0$.

Vectors are represented by bold lowercase letters ($\boldsymbol{x} = (x_1, \dots, x_p)^\top$), with $\|\boldsymbol{x}\|_q$ indicating the $\ell_q$ norm and $\operatorname{supp}(\boldsymbol{x})=\{j \in[p]: x_j \neq 0\}$ denoting its support set. For any subset $S = \{j_1, \dots, j_{|S|}\} \subseteq [p]$ with indices ordered as $j_1 < j_2 < \dots < j_{|S|}$, we define $\boldsymbol{x}_S = (x_{j_1}, \dots, x_{j_{|S|}})^\top$ as the restriction of $\boldsymbol{x}$ to the coordinates in $S$. We use $\boldsymbol{A} = [A_{i, j}]_{i\in[n], j\in[m]}$ to denote a matrix, with $ \boldsymbol{A}_{S,T}=[A_{i, j}]_{i\in S, j\in T} $ as the submatrix indexed by $ S, T $. We abbreviate $ \boldsymbol{A}_{S,S}$ as $ \boldsymbol{A}_{S}$. The spectral norm is given by $ \|\boldsymbol{A}\|_2 = \max _{\boldsymbol{v} \in \mathbb{R}^m, \|\boldsymbol{v}\|_2 = 1} \|\boldsymbol{A} \boldsymbol{v}\|_2 $.

For each $e\in\mathcal{E}$, we have $n^{(e)}$ number of gold-standard observations, $m^{(e)}$ number of missing-label observations, and $N^{(e)}$ total number of observations. Suppose $\{(\boldsymbol{x}_i^{(e)}, y^{(e)}_i)\}_{i=1}^{N^{(e)}}\subseteq \mathcal{R}^p \times \mathcal{R}$ are drawn i.i.d. from some distribution $\mu^{(e)} \in \mathcal{U}_{\boldsymbol{\beta}^*, \sigma^2}$. Let $f$ be any measurable function, we denote $\mathbb{E}[f(\boldsymbol{x}^{(e)}, y^{(e)})]=\int f(\boldsymbol{x}, y) \mu^{(e)}(d \boldsymbol{x}, d y)$. Further more, we let $\widehat{\mathbb{E}}_{N^{(e)}}[f(\boldsymbol{x}^{(e)}, y^{(e)})]=\frac{1}{N^{(e)}} \sum_{i=1}^{N^{(e)}} f(\boldsymbol{x}_i^{(e)}, y_i^{(e)})$, $\widehat{\mathbb{E}}_{n^{(e)}}[f(\boldsymbol{x}^{(e)}, y^{(e)})]=\frac{1}{n^{(e)}} \sum_{i=1}^{n^{(e)}} f(\boldsymbol{x}_i^{(e)}, y_i^{(e)})$, and $\widehat{\mathbb{E}}_{m^{(e)}}[f(\boldsymbol{x}^{(e)}, y^{(e)})]=\frac{1}{m^{(e)}} \sum_{i=1}^{m^{(e)}} f(\widetilde{\boldsymbol{x}}_i^{(e)}, \widetilde{y}_i^{(e)})$. When referring to quantities that include both labeled and unlabeled data, we simplify the notation by omitting the tilde. Let $\boldsymbol{\Sigma}^{(e)} = \mathbb{E}[\boldsymbol{x}^{(e)} \boldsymbol{x}^{(e)\top}]$ be the population covariance matrix for environment $e$. The pooled covariance matrix is given by $\overline{\boldsymbol{\Sigma}} = \sum_{e \in \mathcal{E}} \omega^{(e)} \boldsymbol{\Sigma}^{(e)}$. 

We denote by $\widehat{h}^{(e)}: \mathbb{R}^{p} \to \mathbb{R}$ a general prediction rule for environment $e$. The imputation error is defined as $z^{(e)} = \widehat{h}^{(e)}(\boldsymbol{x}^{(e)}) - y^{(e)}$, with its expectation $\eta^{(e)} = \mathbb{E}[z^{(e)}]$, referred to as the imputation bias. The maximum imputation bias across all environments in $\mathcal{E}$ is given by $\eta_{\max} = \max_{e\in\mathcal{E}} \eta^{(e)}$. The missingness ratio in environment $e$ is denoted by $\tau^{(e)} \in [0,1)$, with its empirical estimate given by $\widehat{\tau}^{(e)} = {m^{(e)}}/{N^{(e)}}$. The maximum empirical missingness ratio across all environments is given by $\widehat{\tau}_{\max} = \max_{e\in \mathcal{E}} \widehat{\tau}^{(e)}$.

Next we introduce two general formulas that encapsulates the different sample size notations.  
\begin{equation}\label{eq:general_samplesize}
	\mathcal{N}(k,g,h,f) = \min_{e\in\mathcal{E}} \frac{k(N^{(e)}, n^{(e)}, m^{(e)})}{g(\omega^{(e)}) h(\widehat{\tau}^{(e)}) f(|\eta^{(e)}|)} \quad \text{and}\quad \mathcal{G}(k,g,h,f) = \left(\sum_{e\in\mathcal{E}}  \frac{g(\omega^{(e)})h(\widehat{\tau}^{(e)})f(|\eta^{(e)}|)}{k(N^{(e)}, n^{(e)}, m^{(e)})}    \right)^{-1}.
\end{equation}
These formulations allow us to express various sample sizes compactly by choosing appropriate functions for $k(x,y,z)$, $g(x)$, $h(y)$, and $f(z)$. We next define key sample sizes in the main text. Additional notations used in proofs are provided in the Supplement Section \ref{section:more_notation}, Table~\ref{tab:sample_size_min}-\ref{tab:sample_size_sum_more}.
\begin{itemize}
	\item Basic Sample Sizes: $N_{\min}, n_{\min}, m_{\min}$ correspond to $\mathcal{N}(k,g,h,f)$ with $g(x) = h(y) = f(z) = 1$, and $k(x,y,z)$ set to $x, y, z$, respectively.
	\item Weighted Sample Sizes: $N_{*}, n_{*}, m_{*}$ extend this by incorporating weights, setting $g(x) = x$ while keeping $h(y) = f(z) = 1$.
	\item Missingness-Adjusted Sample Sizes: $n_*^{\widehat\tau}$ and $m_*^{\widehat\tau}$ further account for missingness, using $h(y) = y$. Similarly, $n^{\widehat\tau^2}_*$ and $m^{\widehat\tau^2}_*$ use $h(y) = y^2$ to reflect squared missingness term.
	\item Bias-Adjusted Sample Sizes: $ n_*^{\widehat\tau,|\eta|}$ and $m_*^{\widehat\tau,|\eta|}$ incorporate both missingness and imputation bias by setting $h(y) = y$ and $f(z) = z$. Similarly, $n_*^{\widehat\tau^2,|\eta|^2}$ and $m_*^{\widehat\tau^2,|\eta|^2}$ account for squared terms in both missingness and bias with $h(y) = y^2$ and $f(z) = z^2$.
	\item Aggregated Sample Sizes: $\bar{N}$, $\bar{n}$, $\bar{m}$ are defined using $\mathcal{G}(k,g, h,f)$ with $g(x) = x$, $h(y) = 1$, $f(z) = 1$, and $k(x,y,z)$ as $x$, $y$, $z$, respectively.
	\item Mixed-Aggregated Sample Sizes: $\overline{\sqrt{nN}}^{\widehat\tau}$ and $\overline{\sqrt{mN}}^{\widehat\tau}$ are defined as $\mathcal{G}(k,g, h,f)$ with $g(x) = x$, $h(y) = y$, $f(z) = 1$, and $k(x,y,z)$ being $\sqrt{xy}$, $\sqrt{xz}$, respectively.
\end{itemize}

\subsection{Data Generating Process}\label{subsection:DGP}

\noindent \textbf{Invariante Linear Structure}

We consider a set of distributions $\mathcal{U}_{\boldsymbol{\beta}^*, \sigma^2}$ such that each distribution $\mu \in \mathcal{U}_{\boldsymbol{\beta}^*, \sigma^2}$ is distinct, but the conditional expectation of the response variable $y$ given the true set of important covariates $\boldsymbol{x}_{S^*}$ follows the same linear form across all distributions in $\mathcal{U}_{\boldsymbol{\beta}^*, \sigma^2}$. Specifically, we assume that observations are independently and identically distributed (i.i.d.) from some $\mu \in \mathcal{U}_{\boldsymbol{\beta}^*, \sigma^2}$, which is characterized by:
\begin{equation}\label{eq:distribution}
	\mathcal{U}_{\boldsymbol{\beta}^*, \sigma^2}=\left\{\begin{array}{c}
		\mu: \mathbb{E}_\mu\left[y \mid \boldsymbol{x}_{S^*}\right]=\left(\boldsymbol{\beta}_{S^*}^*\right)^{\top} \boldsymbol{x}_{S^*}, \operatorname{Var}_\mu\left[y \mid \boldsymbol{x}_{S^*}\right] \leq \sigma^2, \mu \text {-a.s. } \boldsymbol{x}, \\
		\forall j \in[p], \mathbb{E}_\mu\left[x_j^2\right] \leq \sigma^2
	\end{array}\right\} .
\end{equation}

In another words, we assume that the following linear structure describes the relationship between the covariates $\boldsymbol{x}^{(e)} \in \mathbb{R}^p$ and the response variable $y^{(e)} \in \mathbb{R}$ across a set of environments, denoted by $\mathcal{E}$.
\begin{equation}\label{eq:linear_structure}
	y^{(e)}  = (\boldsymbol{\beta}^*_{S^*})^{\top} \boldsymbol{x}_{S^*}^{(e)} + \varepsilon^{(e)}, \quad \forall e \in \mathcal{E}, \quad \text{with} \quad \mathbb{E}[\varepsilon^{(e)} \mid \boldsymbol{x}_{S^*}^{(e)}] = 0,
\end{equation}
where the support of the covariates $S^* = \{j:\beta_j^*\ne 0\}$ and the corresponding coefficients $\boldsymbol{\beta}^*_{S^*}$ are the same across $\mathcal{E}$, indicating a stable relationships between these features and the response variable throughout the different environments.

This invariant linear structure allows us to leverage the heterogeneity across multiple environments to identify both the stable covariates support set $S^*$ and its associated coefficients $\boldsymbol{\beta}^*$. These estimates can then be used to construct a robust predictor that generalizes effectively to unseen environments whose distributions belong to the set $\mathcal{U}_{\boldsymbol{\beta}^*, \sigma^2}$.

\noindent \textbf{Structure Equation Model}

The above data-generating process can also be framed within the context of a structural equation model (SEM). For each environment $e \in \mathcal{E}$, let $\boldsymbol{k}^{(e)}=\left(\boldsymbol{x}^{(e)}, y^{(e)}\right) \in \mathbb{R}^{p+1}$ denote the collection of covariates and the outcome variable. In the SEM framework, each variable $k_j^{(e)}$ is influenced by its parent nodes, denoted as $\mathrm{pa}(j)$, which represent the variables that directly affect $k_j^{(e)}$ in the underlying causal graph. Given this setup, the data for each environment follow the assignments:
\begin{align}\label{eq:sem}
	x_j^{(e)} & := k_j^{(e)} \leftarrow f_j^{(e)}\left(\boldsymbol{k}_{\mathrm{pa}(j)}^{(e)}, u_j^{(e)}\right), \quad\quad j=1, \ldots, p \nonumber           \\
	y^{(e)}   & := k_{p+1}^{(e)} \leftarrow\left(\boldsymbol{\beta}^*\right)_{\mathrm{pa}(p+1)}^{\top} \boldsymbol{x}_{\mathrm{pa}(p+1)}^{(e)}+u_{p+1}^{(e)},
\end{align}
where $\boldsymbol{u}^{(e)} = \left(u_1^{(e)}, \ldots, u_{p+1}^{(e)}\right)$ consists of independent, zero-mean random variables that are independent across environments. While the covariates $x_j^{(e)}$ may follow nonlinear relationships through arbitrary functions $f_j^{(e)}(\cdot)$, we assume that the outcome $y^{(e)}$ follows a linear structure involving its direct causes $\boldsymbol{x}_{\mathrm{pa}(p+1)}^{(e)}$.

In this SEM formulation, the coefficients $\left(\boldsymbol{\beta}^*\right)_{\mathrm{pa}(p+1)}$ represent the causal effects, corresponding directly to the parameter vector $\boldsymbol{\beta}^*_{S^*}$ introduced in the linear data-generating process. Similarly, the set of direct causes $\mathrm{pa}(p+1)$ aligns with the set $S^*$ from the earlier model, both of which remain invariant across environments.

\noindent \textbf{True Regression Model}

Besides using the linear model (\ref{eq:linear_structure}) and the SEM (\ref{eq:sem}) to describe the data-generating process for $y^{(e)}$ across environments, we can also express the outcome within each specific environment $e\in\mathcal{E}$ using the environment-dependent true regression function:
\begin{equation}\label{eq:true_regression}
	h^{(e)}(\boldsymbol{x}^{(e)}) \coloneqq \mathbb{E}[y^{(e)} \mid \boldsymbol{x}^{(e)}].
\end{equation}
This function represents the conditional expectation of the outcome given the covariates, and it is not restricted to a linear form. For each environment, the outcome can be written as:
\begin{equation}\label{eq:true_regression_model}
	y^{(e)} = h^{(e)}(\boldsymbol{x}^{(e)}) + \xi^{(e)} \quad\text{with}\quad \mathbb{E}[\xi^{(e)} \mid \boldsymbol{x}^{(e)}] = 0,
\end{equation}
where $\xi^{(e)}$ a zero-mean noise term independent of the covariates $\boldsymbol{x}^{(e)}$. Unlike the previous invariant structure described by equations (\ref{eq:linear_structure}) and (\ref{eq:sem}), which holds across all environments with the same important covariates $S^*$ and stable parameters $\boldsymbol{\beta}^*$, the true regression function $h^{(e)}(\cdot)$ can differ from one environment to another, capturing environment-specific relationships between the covariates and the outcome.

We want to emphasize the distinction between the invariant models, represented by equations (\ref{eq:linear_structure}) and (\ref{eq:sem}), and the true regression model described in equation (\ref{eq:true_regression_model}). The former two assume that there is an invariant structure across all environments, satisfying the exogeneity condition globally. In contrast, the true regression function $h^{(e)}(\boldsymbol{x}^{(e)})$ in equation (\ref{eq:true_regression_model}) ensures the exogeneity condition only for a specific environment $e\in\mathcal{E}$. When focusing on prediction within a single environment, it is often preferable to estimate $h^{(e)}(\boldsymbol{x}^{(e)})\in \mathcal{L}_2(\mathbb{P}_{\boldsymbol{x}^{(e)}})$ rather than $(\boldsymbol{\beta}_{S^*}^*)^{\top}\boldsymbol{x}^{(e)}_{S^*}\in\mathcal{L}_2(\mathbb{P}_{\boldsymbol{x}_{S^*}^{(e)}})$. This preference arises because the true regression function $h^{(e)}(\boldsymbol{x}^{(e)})$ minimizes the mean squared prediction error:
\begin{align}\label{eq:mse}
	\mathbb{E}[\{y^{(e)} - g(\boldsymbol{x}^{(e)})\}^2],
\end{align}
over all square-integrable functions $g\in\mathcal{L}_2(\mathbb{P}_{\boldsymbol{x}^{(e)}})$. In other words, $h^{(e)}(\boldsymbol{x}^{(e)})$ offers the most accurate prediction for that environment $e$ by capturing all relevant relationships, including any spurious correlations, present in the data of that specific environment.

\noindent \textbf{Best Linear Predictor}

Continuing with the focus on a single environment $e\in\mathcal{E}$, it is useful to note that when restricting the model to the linear function class, the best linear predictor for that environment is given by:
\begin{equation}\label{eq:best_linear_predictor}
	\boldsymbol{\beta}^{(e)}= \arg\min_{\boldsymbol{\beta} \in \mathbb{R}^p} \mathbb{E}[\{y^{(e)} - \boldsymbol{\beta}^{\top}\boldsymbol{x}^{(e)}\}^2].
\end{equation}
This predictor yields the coefficients that minimize the mean squared error for linear approximations of the relationship between $\boldsymbol{x}^{(e)}$ and $y^{(e)}$. In cases where the true regression $h^{(e)}(\boldsymbol{x}^{(e)})$ is nonlinear and thus the linear model is misspecified, the data in that specific environment $e$ can be expressed as:
\begin{equation}\label{eq:best_linear_predictor_model}
	y^{(e)} = (\boldsymbol{\beta}^{(e)})^{\top}\boldsymbol{x}^{(e)}+ \zeta^{(e)}.
\end{equation}
Here, $\zeta^{(e)}$ captures the part of $y^{(e)}$ that is not explained by the linear predictor, and the exogeneity condition no longer holds, i.e. $\mathbb{E}[\zeta^{(e)} \mid \boldsymbol{x}^{(e)}] \ne 0$. However, a weaker condition still applies:
\begin{equation}\label{eq:weaker_error_condition}
	\mathbb{E}[\zeta^{(e)} \boldsymbol{x}^{(e)}] = 0.
\end{equation}
This weaker condition ensures that, on average, the residual $\zeta^{(e)}$ is uncorrelated with the covariates $\boldsymbol{x}^{(e)}$, making the linear predictor $(\boldsymbol{\beta}^{(e)})^{\top}\boldsymbol{x}^{(e)}$ the best approximation of the relationship between $\boldsymbol{x}^{(e)}$ and $y^{(e)}$ within the linear function space.

\noindent \textbf{The Connection}

So far, we have introduced four different but interconnected ways to describe the data-generating process across multiple environments. The first two—equation (\ref{eq:linear_structure}) and (\ref{eq:sem})—focus on capturing invariant relationships between the outcome and a subset of covariates that remain consistent across all environments. In contrast, the latter two—equations (\ref{eq:true_regression_model}) and (\ref{eq:best_linear_predictor_model})—describe environment-specific relationships that hold only within individual environments. Each framework provides unique insights that contribute in complementary ways to the methods presented in this paper.

The invariant linear model (\ref{eq:linear_structure}) offers a foundation for identifying stable covariates and their associated coefficients across environments. Alongside the best linear predictor (\ref{eq:best_linear_predictor}) in each specific environment, which serves as a practical bridge between environment-specific data and the invariant structure, we can leverage the heterogeneity across multiple environments to uncover stable patterns. These stable relationships, in turn, facilitate the development of a robust predictor that generalizes effectively to unseen environments. Under the modularity \cite{scholkopf2012causal} assumption, the SEM framework (\ref{eq:sem}) provides a complementary causal interpretation by revealing how these stable covariates correspond to the direct causes of the outcome, thereby aligning with the invariant linear model in its interpretation. 

The true regression function $h^{(e)}\left(\boldsymbol{x}^{(e)}\right)$ in equation (\ref{eq:true_regression_model}) captures the environment-specific relationships between covariates and the outcome. While such models offer the most accurate prediction within individual environments, relying solely on them for prediction can be problematic when applied to unseen environments. As discussed in the literature on robust prediction in Section \ref{section:introduction}, models that leverage all correlations—including spurious ones—tend to perform poorly when the distribution of the test data differs from that of the training environments, especially when shifts arise due to spurious correlation and anti-causal variables. However, as we will illustrate later, when dealing with missing outcome observations within each environment, models that approximate the true environment-specific regression functions are valuable. These models provide critical information that can be utilized to develop methods for robust prediction across multiple environments, ensuring reliable performance despite the presence of missing data.

\subsection{Missing Outcomes}\label{subsection:missing_outcomes}

We consider a semi-supervised setting, where for each environment, some of the data are labeled while the rest remain unlabeled. This setup can be viewed as a special case of missing data, specifically where the outcomes are missing completely at random (MCAR). That is, the missingness of outcomes is unrelated to the observed or unobserved data. Although the broader missing data literature often addresses the more general missing at random (MAR) setting \citep{scharfstein1999adjusting}, MAR relies on assumptions that are difficult to verify or fulfill in practice \citep{robins2000sensitivity}, especially in situations with a large amount of missing data \citep{gui2022causal}. Therefore, in this work, we focus on the MCAR assumption as it is more reasonable for the semi-supervised learning setting. Our aim is to develop a robust theoretical framework under MCAR, and we leave extensions to more complex missing data mechanisms, like MAR, for future work.

Specifically, for each environment $e \in \mathcal{E}$, we observe a sample of $n^{(e)}$ gold-standard observations:
$$(\boldsymbol{x}_1^{(e)}, y_1^{(e)}), \ldots, (\boldsymbol{x}_{n^{(e)}}^{(e)}, y_{n^{(e)}}^{(e)}),$$
which are drawn i.i.d according to the joint distribution $(\boldsymbol{x}^{(e)}, y^{(e)}) \sim \mu^{(e)}\in\mathcal{U}_{\boldsymbol{\beta}^*, \sigma^2}$. In addition to these complete observations, we have a sample of size $m^{(e)}$ where only the covariates $\boldsymbol{x}^{(e)}$ are observed, and the corresponding outcomes are missing. These are expressed as:
$$(\tilde{\boldsymbol{x}}_{n^{(e)}+1}^{(e)}, \tilde{y}_{n^{(e)}+1}^{(e)}), \ldots, (\tilde{\boldsymbol{x}}_{n^{(e)}+m^{(e)}}^{(e)}, \tilde{y}_{n^{(e)}+m^{(e)}}^{(e)}),$$
where the outcome variables $\tilde{y}_{n^{(e)} + 1}^{(e)}, \ldots, \tilde{y}_{n^{(e)}+m^{(e)}}^{(e)}$ are not observed. The partial observations follow the same distribution $\mu^{(e)}$ as the gold-standard data, with the key difference being the absence of observed outcomes. When referring to quantities that include both labeled and unlabeled data, we simplify the notation by omitting the tilde. The total sample size for each environment $e$ is given by $N^{(e)} = n^{(e)} + m^{(e)}$. Moreover, we define the empirical ratio of missingness in environment $e$ as
\begin{equation}\label{eq:missing_ratio}
	\widehat{\tau}^{(e)} = \frac{m^{(e)}}{N^{(e)}}.
\end{equation}
As the sample size increases, we assume that the empirical ratio of missingness converges to a constant $\tau^{(e)}$, representing the underlying proportion of missing outcomes in each environment. This can be expressed as:
\begin{equation}\label{eq:convergence_missing_ratio}
	\tau^{(e)} = \lim_{m^{(e)}, N^{(e)} \to \infty} \frac{m^{(e)}}{N^{(e)}} \in [0, 1).
\end{equation}

Our aim is to study methods that incorporate the additional unlabeled data to improve the inference of the true important set $S^*$ and the corresponding coefficient $\boldsymbol{\beta}^*$. The key lies in the relationship between the marginal distribution of the covariates $p(\boldsymbol{x}^{(e)})$ and the conditional distribution $p(y^{(e)} \mid$ $\boldsymbol{x}^{(e)})$. In semi-supervised learning, if $p(\boldsymbol{x}^{(e)})$ holds relevant information about $p(y^{(e)} \mid$ $\boldsymbol{x}^{(e)})$, then the unlabeled data can help us better estimate the marginal distribution of the covariates, which in turn sharpens our understanding of the joint distribution and supports better predictions \citep{learning2006semi}. This principle, as discussed in \cite{zhu2005semi}, highlights that the success of leveraging unlabeled data hinges on how well $p(\boldsymbol{x}^{(e)})$ captures the structure of $p(y^{(e)} \mid$ $\boldsymbol{x}^{(e)})$. If there's no such connection, the added data offer no improvement to prediction accuracy.

In the context of linear model, if the conditional expectation $\mathbb{E}[y^{(e)} \mid \boldsymbol{x}^{(e)}]$ is truly linear, then $\boldsymbol{x}^{(e)}$'s distribution does not add additional information when estimating the best linear predictor. In this case, the marginal distribution of $\boldsymbol{x}^{(e)}$ is considered ancillary and does not affect the inference for $\boldsymbol{\beta}^{(e)}$. But when $\mathbb{E}[y^{(e)} \mid \boldsymbol{x}^{(e)}]$ includes unmodeled non-linearities, leveraging unlabeled data becomes beneficial, allowing one to estimate $\boldsymbol{\beta}^{(e)}$ more efficiently \citep{buja2019models, azriel2022semi,chakrabortty2018efficient}. 


Assuming a linear form for $\mathbb{E}[y^{(e)} \mid \boldsymbol{x}^{(e)}]$ is unrealistic in our problem setting. Given the flexibility allowed for nonlinear functions $f_j^{(e)}(\cdot)$ for all $j=1,...,p$ within the SEM framework (\ref{eq:sem}) of each environment, the true regression function $h^{(e)}(\boldsymbol{x}^{(e)})$ likely captures complex, environment-specific nonlinear interactions. Thus, while the invariant linear structure ensures stability across environments, conditional expectations within individual environments may deviate from linearity.

As we will explain in Section \ref{section:methodology}, our approach is based on an objective function that combines a pooled least squares term across multiple environments with penalty terms designed to eliminate spurious variables. One can intuitively think of this as using the best linear predictor in each environment as a bridge between the environment-specific data and the invariant structure. The key opportunity to leverage the unlabeled data arises from the fact that we are applying a linear working model within each environment, while the true regression function $h^{(e)}\left(\boldsymbol{x}^{(e)}\right)$ in the same environment is likely nonlinear. This discrepancy between the linear working model and the underlying nonlinear structure allows us to exploit the marginal distribution of the covariates to improve the inference for $\boldsymbol{\beta}^{(e)}$. Consequently, we can more effectively detect spurious correlations and ultimately identify the stable variables and their corresponding coefficients across environments. 

\subsection{Imputation}\label{subsection:imputation}

Given the semi-supervised nature of our problem, where some environments may contain unlabeled data, imputing the missing outcomes becomes an essential step for harnessing the information from the covariate-only observations. We denote a general imputation model by $\widehat h^{(e)}: \mathbb{R}^{p}\to\mathbb{R}$, where the superscript $(e)$ indicates that it is a prediction rule specific to the environment $e$. Recall that the $m^{(e)}$ observations $(\tilde{\boldsymbol{x}}_{n^{(e)}+1}^{(e)}, \tilde{y}_{n^{(e)}+1}^{(e)}), \ldots, (\tilde{\boldsymbol{x}}_{n^{(e)}+m^{(e)}}^{(e)}, \tilde{y}_{n^{(e)}+m^{(e)}}^{(e)})$ are independent and identically distributed samples from the joint distribution $(\boldsymbol{x}^{(e)}, y^{(e)}) \sim \mu^{(e)}$, with the outcomes $\tilde{y}_{n^{(e)} + 1}^{(e)}, \ldots, \tilde{y}_{n^{(e)}+m^{(e)}}^{(e)}$ being unobserved. In practice, an imputation model is applied to the observed covariates $\tilde{\boldsymbol{x}}_{n^{(e)}+1}^{(e)}, \ldots, \tilde{\boldsymbol{x}}_{n^{(e)}+m^{(e)}}^{(e)}$ to produce imputed values for the missing outcomes:
$$
	\widehat h^{(e)} (\tilde{\boldsymbol{x}}_{n^{(e)}+1}^{(e)}), \ldots, \widehat h^{(e)}(\tilde{\boldsymbol{x}}_{n^{(e)}+m^{(e)}}^{(e)}).
$$
Since the true outcomes are not observed, the imputation errors corresponding to each observation, given by
$$
	\widehat h^{(e)} (\tilde{\boldsymbol{x}}_{n^{(e)}+1}^{(e)}) - \tilde{y}_{n^{(e)} + 1}^{(e)}, \ldots, \widehat h^{(e)}(\tilde{\boldsymbol{x}}_{n^{(e)}+m^{(e)}}^{(e)})-\tilde{y}_{n^{(e)}+m^{(e)}}^{(e)}
$$
remain unknown. Let's denote the underlying imputation error as
\begin{equation}\label{eq:imputation_error}
	z^{(e)} = \widehat h^{(e)} (\boldsymbol{x}^{(e)}) - y^{(e)}.
\end{equation}
The imputation bias, denoted by $\eta^{(e)}$, is defined as the expected imputation error:
\begin{equation}\label{eq:imputation_bias}
	\eta^{(e)}=\mathbb{E}[z^{(e)}].
\end{equation}
When $\eta^{(e)}=0$, we say that the imputation of the outcome in environment $e$ is unbiased. In the ideal scenario where the imputation model $\widehat{h}^{(e)}(\cdot)$ perfectly matches the true regression function $h^{(e)}(\cdot)$, the imputation bias is zero:
\begin{align*}
	\eta^{(e)} & = \mathbb{E}\left[z^{(e)}\right]                                                               \\
	           & = \mathbb{E}[ h^{(e)} (\boldsymbol{x}^{(e)}) - y^{(e)}]                                        \\
	           & = \mathbb{E}[\mathbb{E}[  h^{(e)} (\boldsymbol{x}^{(e)}) - y^{(e)}\mid \boldsymbol{x}^{(e)}] ] \\
	           & = \mathbb{E}[ h^{(e)} (\boldsymbol{x}^{(e)}) - \mathbb{E}[y^{(e)}\mid \boldsymbol{x}^{(e)}] ]  \\
	           & = 0,
\end{align*}
where the last line holds by the definition of $h^{(e)}(\boldsymbol{x}^{(e)})$. In other words, the imputed outcomes are unbiased.

However, in practice, achieving a perfect imputation model is rarely feasible. A common approach is to estimate $h^{(e)}(\cdot)$ using a machine learning algorithm, typically trained nonparametrically on a subset of gold-standard data. While these models are designed to be consistent or asymptotically unbiased, their performance in finite samples is inevitably affected by bias due to the limited amount of training data. This bias, arises from the discrepancy between the estimated imputation model and the true regression function, can propagate through the analysis and impact the inference of the stable variables and their coefficients across multiple environments. Alternatively, a pre-trained machine learning model can be used for imputation. Such models, typically trained on large datasets with considerable computational resources, may provide reasonable approximations.  However, the discrepancy between the pre-trained model and the true regression function can still introduce bias, impacting final inference results.

Rather than basing our analysis on the characteristics of a specific type of imputation model, we consider $\widehat{h}^{(e)}(\cdot)$ as an arbitrary non-random function. This approach enables us to examine how the imputation bias $\eta^{(e)}$ in a specific environment $e$ influences the entire inference process across multiple environments. Our goal is to perform a non-asymptotic analysis that clarify conditions under which the bias introduced by the imputation model $\widehat{h}^{(e)}(\cdot)$ does not hinder the selection of stable variables or compromise the convergence of the corresponding coefficient estimators.

\section{Preliminary - A Review on Environment Invariant Linary Least Squares}\label{section:preliminary}

In recent years, invariance learning has emerged as a promising approach for improving generalization in unseen environments. Within this framework, some approaches focus on leveraging the heterogeneity across multiple environments to identify stable relationships. The Environment Invariant Linear Least Squares (EILLS) approach, proposed by \cite{fan2024environment}, is one such example. It takes a statistical perspective on invariance learning and develops a non-asymptotic analysis, through which theoretical guarantees for finite sample data can be established.

Given that missing outcomes are common in practice, practitioners often select an imputation model without fully accounting for the potential bias. This motivates the need to develop methods that appropriately address the bias and understand its impact on inference in finite samples. To address this, we leverage the non-asymptotic analysis techniques developed by \cite{fan2024environment}. Below, we provide a brief overview of their approach.

Starting from the observation that for any $e\in\mathcal{E}$, $\mathbb{E}\left[y^{(e)} \mid \boldsymbol{x}_{S^*}^{(e)}\right]=\boldsymbol{\beta}_{S^*}^{\top} \boldsymbol{x}_{S^*}^{(e)}$, or conditional expectation invariant (CE-invariant) across environments, implies $\nabla_j \mathrm{R}^{(e)}(\boldsymbol{\beta}^*)=0$ for all $j\in S^*$, where $\mathrm{R}^{(e)}(\boldsymbol{\beta})$ is the population $L_2$ risk in environment $e$ and is defined as
\begin{equation}\label{eq:population_L2_risk}
	\mathrm{R}^{(e)}(\boldsymbol{\beta})=\mathbb{E}\left[\left|y^{(e)}-\boldsymbol{\beta}^{\top} \boldsymbol{x}^{(e)}\right|^2\right],
\end{equation}
\cite{fan2024environment} construct the following penalty function $\mathrm{J}(\boldsymbol{\beta} ; \boldsymbol{\omega})$ to enforce zero gradient in the direction of non-zero coordinates:
\begin{equation}\label{eq:J_pop}
	\mathrm{J}(\boldsymbol{\beta} ; \boldsymbol{\omega})=\sum_{j=1}^p\mathds{1}\left\{\beta_j \neq 0\right\} \sum_{e \in \mathcal{E}} \frac{\omega^{(e)}}{4}\left|\nabla_j \mathrm{R}^{(e)}(\boldsymbol{\beta})\right|^2,
\end{equation}
which has global minimum equals to zero because $\mathrm{J}(\boldsymbol{\beta}^* ; \boldsymbol{\omega})=0$. The authors recognize that the solutions to $\mathrm{J}(\boldsymbol{\beta} ; \boldsymbol{\omega})=0$ may not be unique, as there could exist another $\widetilde{\boldsymbol{\beta}}$ with support $\widetilde{S}$ for which $\mathrm{J}(\widetilde{\boldsymbol{\beta}} ; \boldsymbol{\omega})=0$ also holds. Although achieving the global minimum does not guarantee that the selected variables are CE-invariant across environments, the paper points out that it does imply a weaker form of invariance, known as linear least squares invariance (LLS-invariance):
\begin{equation}\label{eq:LLS_invariant}
	\forall e \in \mathcal{E}, \quad \widetilde{\boldsymbol{\beta}} \in \underset{\operatorname{supp}(\boldsymbol{\beta}) \subseteq \widetilde{S}}{\operatorname{argmin}} \mathbb{E}\left[\left|y^{(e)}-\boldsymbol{\beta}^{\top} \boldsymbol{x}^{(e)}\right|^2\right].
\end{equation}

Building on the idea that the penalty (\ref{eq:J_pop}) can screen out all linear spurious variables, \cite{fan2024environment} continue their approach by incorporating linear information to ensure LLS-invariance in the selected variables. Specifically, they propose combining the penalty function $\mathrm{J}(\boldsymbol{\beta} ; \boldsymbol{\omega})$, which helps eliminate linear spurious variables, with a pooled population $L_2$ loss function $\mathrm{R}(\boldsymbol{\beta} ; \boldsymbol{\omega})$ defined as follows:
\begin{equation}\label{eq:pool_loss_pop}
	\mathrm{R}(\boldsymbol{\beta} ; \boldsymbol{\omega})=\sum_{e \in \mathcal{E}} \omega^{(e)} \mathbb{E}\left[\left|y^{(e)}-\boldsymbol{\beta}^{\top} \boldsymbol{x}^{(e)}\right|^2\right].
\end{equation}
The population objective can be expressed as:
\begin{equation}\label{eq:fan_pop}
	\mathrm{Q}(\boldsymbol{\beta} ; \gamma, \boldsymbol{\omega})=\mathrm{R}(\boldsymbol{\beta} ; \boldsymbol{\omega})+\gamma \mathrm{J}(\boldsymbol{\beta} ; \boldsymbol{\omega}),
\end{equation}
where $\gamma\in\mathbb{R}^+$ is a regularization parameter.

The empirical counterpart of the objective function can be constructed using the $n^{(e)}$ fully observed data points available in each environment, as follows:
\begin{equation}\label{eq:fan_emp}
	\widehat{\mathrm{Q}}(\boldsymbol{\beta} ; \gamma, \boldsymbol{\omega})=\widehat{\mathrm{R}}(\boldsymbol{\beta} ; \boldsymbol{\omega})+\gamma \widehat{\mathrm{J}}(\boldsymbol{\beta} ; \boldsymbol{\omega}),
\end{equation}
where the empirical $L_2$ loss $\widehat{\mathrm{R}}(\boldsymbol{\beta} ; \boldsymbol{\omega})$ and the empirical penalty $\widehat{\mathrm{J}}(\boldsymbol{\beta} ; \boldsymbol{\omega})$ are given by:
\begin{equation}\label{eq:pool_loss_emp}
	\widehat{\mathrm{R}}(\boldsymbol{\beta} ; \boldsymbol{\omega})=\sum_{e \in \mathcal{E}} \omega^{(e)} \widehat{\mathbb{E}}_{n^{(e)}}\left[\left|y^{(e)}-\boldsymbol{\beta}^{\top} \boldsymbol{x}^{(e)}\right|^2\right]
\end{equation}
and
\begin{equation}\label{eq:J_emp}
	\widehat{\mathrm{J}}(\boldsymbol{\beta} ; \boldsymbol{\omega})= \sum_{j=1}^p\mathds{1}\left\{\beta_j \neq 0\right\} \sum_{e \in \mathcal{E}} \frac{\omega^{(e)}}{4}\left|\nabla_j \widehat{\mathrm{R}}^{(e)}(\boldsymbol{\beta})\right|^2 = \sum_{j=1}^p \mathds{1}\left\{\beta_j \neq 0\right\} \sum_{e \in \mathcal{E}} \omega^{(e)}\left|\widehat{\mathbb{E}}_{n^{(e)}}\left[x_j^{(e)}\left(y^{(e)}-\boldsymbol{\beta}^{\top} \boldsymbol{x}^{(e)}\right)\right]\right|^2,
\end{equation}
where $\widehat{\mathrm{R}}^{(e)}(\boldsymbol{\beta}) = \widehat{\mathbb{E}}_{n^{(e)}}[| y^{(e)} - \boldsymbol{\beta}^{\top} \boldsymbol{x}^{(e)}|^2]$ is the empirical $L_2$ risk in environment $e$.

It is important to note that the empirical approach described above relies exclusively on the fully observed data, $n^{(e)}$, from each environment, without utilizing any information from the observations with missing outcomes. While this strategy simplifies the estimation process, it may overlook valuable information that could potentially improve inference if incorporated appropriately. In the following section, we introduce a methodology that leverages the information from partially observed data in a way that preserves, and even enhances, the accuracy of the final results without compromising the integrity of the analysis.

\section{Methodology}\label{section:methodology}

To leverage the information from partially observed data, we begin by addressing the bias introduced during imputation. When estimating quantities of interest, such as the mean outcome or regression coefficients, the imputed values can introduce systematic errors due to inaccuracies in the imputation model. These errors can propagate through the analysis, leading to biased estimators. Therefore, a key step in our methodology is to correct for the imputation bias.

\subsection{Imputation Bias Corrected Pooled Empirical $L_2$ Loss $\widehat{\mathrm{R}}_{\mathrm{Adj}}(\boldsymbol{\beta} ; \boldsymbol{\omega})$}\label{subsection:imputation_bias_corrected_pool}

To begin addressing the imputation bias in the estimation of the pooled empirical $L_2$ loss function (\ref{eq:pool_loss_emp}), we draw inspiration from existing approaches centered around augmented inverse probability weighting (AIPW) formulations \citep{tsiatis2006semiparametric, chakrabortty2018efficient, zhang2022high, angelopoulos2023prediction}.

Let's start from the linear regression $(\boldsymbol{\beta}^{(e)})^{\top} \boldsymbol{x}^{(e)}$ in environment $e\in\mathcal{E}$, where $\boldsymbol{\beta}^{(e)}$ is the minimizer of the population $L_2$ risk
\begin{equation}\label{eq:ls_objective}
	\mathrm{R}^{(e)}(\boldsymbol{\beta}) = \mathbb{E}\left[\ell_{\boldsymbol{\beta}}(\boldsymbol{x}^{(e)}, y^{(e)}) \right] \quad \text{with}\quad \ell_{\boldsymbol{\beta}}(\boldsymbol{x}^{(e)}, y^{(e)}) = \left\{y^{(e)} - (\boldsymbol{\beta})^{\top} \boldsymbol{x}^{(e)} \right\}^2.
\end{equation}
When there is a large number of unlabeled observations and an imputation function $\widehat{h}^{(e)}(\cdot)$ is available, a simple approach to estimate $\boldsymbol{\beta}^{(e)}$, while incorporating all observations, is to minimize the following empirical objective:
\begin{equation}\label{eq:ls_objective_biased}
	\frac{1}{n^{(e)}} \sum_{i=1}^{n^{(e)}} \ell_{\boldsymbol{\beta}}(\boldsymbol{x}_i^{(e)}, y_i^{(e)})  + \frac{1}{m^{(e)}} \sum_{i=n^{(e)}+1}^{n^{(e)}+m^{(e)}} \ell_{\boldsymbol{\beta}}(\tilde{\boldsymbol{x}}_i^{(e)}, \widehat{h}^{(e)}(\tilde{\boldsymbol{x}}_i^{(e)})).
\end{equation}
Note that this empirical objective is biased for the true objective in (\ref{eq:ls_objective}). Specifically, the bias can be expressed as
\begin{equation}\label{eq:bias_objective_amount}
	\mathbb{E}\left[\ell_{\boldsymbol{\beta}}(\boldsymbol{x}^{(e)}, \widehat{h}^{(e)}(\boldsymbol{x}^{(e)})) \right].
\end{equation}
A natural next step is to perform a bias correction by subtracting an empirical estimate of the bias (\ref{eq:bias_objective_amount}) from the original objective (\ref{eq:ls_objective_biased}), resulting in the following updated and unbiased objective:
\begin{equation}\label{eq:ls_objective_correct_bias}
	\frac{1}{n^{(e)}} \sum_{i=1}^{n^{(e)}} \ell_{\boldsymbol{\beta}}(\boldsymbol{x}_i^{(e)}, y_i^{(e)})  + \frac{1}{m^{(e)}} \sum_{i=n^{(e)}+1}^{n^{(e)}+m^{(e)}} \ell_{\boldsymbol{\beta}}(\tilde{\boldsymbol{x}}_i^{(e)}, \widehat{h}^{(e)}(\tilde{\boldsymbol{x}}_i^{(e)})) -  \frac{1}{n^{(e)}} \sum_{i=1}^{n^{(e)}} \ell_{\boldsymbol{\beta}}(\boldsymbol{x}_i^{(e)}, \widehat{h}^{(e)}(\boldsymbol{x}_i^{(e)})).
\end{equation}
To take a step further, we can incorporate additional observations while still preserving the unbiasedness:
\begin{equation}\label{eq:ls_objective_correct_bias_efficient}
	\frac{1}{n^{(e)}} \sum_{i=1}^{n^{(e)}} \ell_{\boldsymbol{\beta}}(\boldsymbol{x}_i^{(e)}, y_i^{(e)})  + \frac{1}{N^{(e)}} \sum_{i=1}^{N^{(e)}} \ell_{\boldsymbol{\beta}}(\boldsymbol{x}_i^{(e)}, \widehat{h}^{(e)}(\boldsymbol{x}_i^{(e)})) -  \frac{1}{n^{(e)}} \sum_{i=1}^{n^{(e)}} \ell_{\boldsymbol{\beta}}(\boldsymbol{x}_i^{(e)}, \widehat{h}^{(e)}(\boldsymbol{x}_i^{(e)})).
\end{equation}
In the objective above, we simplify the notation by omitting the tilde when referring to quantities that include both labeled and unlabeled data.

The empirical objective in (\ref{eq:ls_objective_correct_bias_efficient}) can be directly connected to the AIPW score. To see this, first define $d^{(e)}$ as a binary random variable with support $\{0,1\}$, where $d^{(e)} = 1$ indicates that the label is observed, and $d^{(e)} = 0$ means it is missing. Under the setting of MCAR, $d^{(e)}$ is independent of $(\boldsymbol{x}^{(e)}, y^{(e)})$, and $\mathbb{E}[d^{(e)}] = 1-\tau^{(e)}$. Rearranging the terms in (\ref{eq:ls_objective_correct_bias_efficient}), we obtain:
\begin{equation}\label{eq:ls_objective_correct_bias_efficient_rearrange}
	\frac{1}{N^{(e)}} \sum_{i=1}^{N^{(e)}}\left[\frac{d_i^{(e)} \ell_{\boldsymbol{\beta}}\left(\boldsymbol{x}^{(e)}_i, y^{(e)}_i\right)-d_i^{(e)} \ell_{\boldsymbol{\beta}}\left(\boldsymbol{x}^{(e)}_i, \widehat{h}^{(e)}_i(\boldsymbol{x}^{(e)}_i)\right)}{1-\widehat{\tau}}+\ell_{\boldsymbol{\beta}}\left(\boldsymbol{x}^{(e)}_i, \widehat{h}^{(e)}_i(\boldsymbol{x}^{(e)}_i)\right)\right]
\end{equation}
The minimizer of this rearranged objective satisfies the following equation
\begin{equation}\label{eq:AIPW_score}
	\frac{1}{N^{(e)}} \sum_{i=1}^{N^{(e)}}\left[\frac{d_i^{(e)} \boldsymbol{x}_i^{(e)}\{y_i^{(e)} - \widehat{h}^{(e)}_i(\boldsymbol{x}^{(e)}_i)\}}{1-\widehat{\tau}}+\boldsymbol{x}_i^{(e)}\{ \widehat{h}^{(e)}_i(\boldsymbol{x}^{(e)}_i) - (\boldsymbol{\beta}^{(e)})^{\top} \boldsymbol{x}_i^{(e)}\}\right]=0,
\end{equation}
which is the AIPW estimating equation.

Since we are working under the MCAR assumption, $1-\widehat{\tau}^{(e)}$ is an unbiased estimator of the probability of being observed. As a result, the estimator for $\boldsymbol{\beta}^{(e)}$ remains consistent even if $\widehat{h}^{(e)}(\cdot)$ is a given non-random imputation function, regardless of its quality. While the empirical objective in (\ref{eq:ls_objective_correct_bias_efficient}) directly relates to the AIPW score, our focus is not on obtaining asymptotic results from the AIPW approach. Instead, we leverage the structure of this debiased empirical objective as a basis for developing non-asymptotic analysis tailored to finite sample settings in the context of missing data.

With the bias correction approach established for a single environment, we can now extend this approach to incorporate multiple environments. The imputation bias-corrected pooled empirical $L_2$ loss, denoted by $\widehat{\mathrm{R}}_{\mathrm {Adj }}(\boldsymbol{\beta} ; \boldsymbol{\omega})$, is formulated as follows:
\begin{equation}\label{eq:bia_corr_pool_L2_loss}
	\widehat{\mathrm{R}}_{\mathrm{Adj}}(\boldsymbol{\beta} ; \boldsymbol{\omega})=\sum_{e \in \mathcal{E}} \frac{\omega^{(e)}}{N^{(e)}} \sum_{i=1}^{N^{(e)}}\left\{\widehat{h}^{(e)}(\boldsymbol{x}_i^{(e)})-\boldsymbol{\beta}^{\top} \boldsymbol{x}_i^{(e)}\right\}^2+\sum_{e \in \mathcal{E}} \frac{\omega^{(e)}}{n^{(e)}} \sum_{i=1}^{n^{(e)}}\left(\left\{y_i^{(e)}-\boldsymbol{\beta}^{\top} \boldsymbol{x}_i^{(e)}\right\}^2-\left\{\widehat{h}^{(e)}(\boldsymbol{x}_i^{(e)})-\boldsymbol{\beta}^{\top} \boldsymbol{x}_i^{(e)}\right\}^2\right).
\end{equation}
It is easy to verify that the above empirical objective is unbiased for the pooled population $L_2$ loss function $\mathrm{R}(\boldsymbol{\beta} ; \boldsymbol{\omega})$ in (\ref{eq:pool_loss_pop}). In addition to correcting the pooled empirical $L_2$ loss for imputation bias, it is crucial to address the bias present in the regularization term (\ref{eq:J_emp}) to ensure that the entire objective function (\ref{eq:fan_emp}) remains unbiased.

\subsection{Imputation Bias Corrected Empirical Penalty $\widehat{\mathrm{J}}_{\mathrm{Adj}}(\boldsymbol{\beta} ; \boldsymbol{\omega})$}\label{subsection:imputation_bias_corrected_J}

The original penalty function, $\widehat{\mathrm{J}}(\boldsymbol{\beta} ; \boldsymbol{\omega})$, introduced by \cite{fan2024environment}, is based on the insight that the gradient of the $L_2$ risk $\mathrm{R}^{(e)}$ over all environments $\mathcal{E}$ vanishes in the directions corresponding to the coordinates associated with the LLS-invariant variables. This penalty discourages linear spurious correlations. As shown in the second equality of equation (\ref{eq:J_emp}), it not only represents the $\ell_2$ norm of the gradient in the directions of non-zero coordinates but can also be interpreted as a measure of the correlation between the features and the fitted residuals across environments. A variable is considered linearly spurious if it correlates with the model residuals in at least one environment. The penalty function specifically targets and penalizes any nonzero coefficient associated with a feature that correlates with the fitted residuals in any of the environments, thereby reinforcing the selection of variables that maintain stable relationships across all environments.

When data contains missing labels and imputation is applied, a straightforward approach is to use the following penalty function:\begin{equation}\label{eq:naive_penalty}
	\sum_{j=1}^p \mathds{1}\left\{\beta_j \neq 0\right\} \sum_{e \in \mathcal{E}} \omega^{(e)} \left\{\left|\widehat{\mathbb{E}}_{n^{(e)}}\left[x_j^{(e)}\left(y^{(e)}-\boldsymbol{\beta}^{\top} \boldsymbol{x}^{(e)}\right)\right] + \widehat{\mathbb{E}}_{m^{(e)}}\left[x_j^{(e)}\left(\widehat{h}^{(e)}(\boldsymbol{x}^{(e)})-\boldsymbol{\beta}^{\top} \boldsymbol{x}^{(e)}\right)\right] \right|^2\right\}.
\end{equation}
However, simply substituting the missing outcomes with predictions from the imputation model can lead to an inaccurate measurement of the correlation between features and the fitted values, distorting original linear spuriousness. To address this issue, we adjust the penalty function as follows:
\begin{align}
	\widehat{\mathrm{J}}_{\mathrm{Adj}}(\boldsymbol{\beta} ; \boldsymbol{\omega}) & = \sum_{j=1}^p \mathds{1}\left\{\beta_j \neq 0\right\} \sum_{e \in \mathcal{E}} \omega^{(e)} \left\{\left|\widehat{\mathbb{E}}_{n^{(e)}}\left[x_j^{(e)}\left(y^{(e)}-\boldsymbol{\beta}^{\top} \boldsymbol{x}^{(e)}\right)\right] \right.\right. \notag \\
	                                                                              &\left. \quad\quad\quad\quad\quad\quad\quad\quad\quad\quad\quad +\, \widehat{\mathbb{E}}_{m^{(e)}}\left[x_j^{(e)}\left(\widehat{h}^{(e)}(\boldsymbol{x}^{(e)})-\boldsymbol{\beta}^{\top} \boldsymbol{x}^{(e)}\right)\right]\right. \notag               \\
	                                                                              & \left.\quad\quad\quad\quad\quad\quad\quad\quad\quad\quad\quad\left. -\, \widehat{\mathbb{E}}_{n^{(e)}}\left[x_j^{(e)}\left(\widehat{h}^{(e)}(\boldsymbol{x}^{(e)})-\boldsymbol{\beta}^{\top} \boldsymbol{x}^{(e)}\right)\right] \right|^2\right\},
	\label{eq:penalty_adjust}
\end{align}
which can be simplified and extended by incorporating more observations, resulting in the final form of the adjusted penalty function:
\begin{equation}\label{eq:penalty_adjust_more}
	\widehat{\mathrm{J}}_{\mathrm{Adj}}(\boldsymbol{\beta} ; \boldsymbol{\omega}) = \sum_{j=1}^p \mathds{1}\left\{\beta_j \neq 0\right\} \sum_{e \in \mathcal{E}} \omega^{(e)}\left\{\left| \widehat{\mathbb{E}}_{N^{(e)}}\left[x_j^{(e)}\left(\widehat{h}^{(e)}(\boldsymbol{x}^{(e)})-\boldsymbol{\beta}^{\top} \boldsymbol{x}^{(e)}\right)\right]  + \widehat{\mathbb{E}}_{n^{(e)}}\left[x_j^{(e)}\left(y^{(e)}-\widehat{h}^{(e)}(\boldsymbol{x}^{(e)})\right)\right] \right|^2\right\}.
\end{equation}
In particular, one can view the term $\widehat{\mathbb{E}}_{n^{(e)}}[x_j^{(e)}(y^{(e)}-\widehat{h}^{(e)}(\boldsymbol{x}^{(e)}))]$ as the correction for the distortion caused by directly using imputed values, ensuring that the adjusted penalty better reflects the original correlations present in the data.

We note that the adjusted penalty function in (\ref{eq:penalty_adjust_more}) also shares a connection with the AIPW score in (\ref{eq:AIPW_score}). Specifically, the AIPW score can be viewed as the imputation-bias-corrected gradient of the empirical $L_2$ risk, $\widehat{\mathrm{R}}^{(e)}(\boldsymbol{\beta})$, for environment $e$. The solution of (\ref{eq:AIPW_score}), denoted by $\widehat{\boldsymbol{\beta}}^{(e)}_{\mathrm{AIPW}}$, is a consistent estimator of $\boldsymbol{\beta}^{(e)}$ and satisfies the condition that the imputation-bias-corrected gradient is zero in all directions for this specific environment $e$. When there is a distribution shift from environment $e$ to $e'$, it is unlikely that $\boldsymbol{\beta}^{(e)}$ will also minimize $\mathrm{R}^{(e')}(\boldsymbol{\beta})$, the population $L_2$ risk in environment $e'$. Therefore, it is unreasonable to expect that the estimator $\widehat{\boldsymbol{\beta}}^{(e)}_{\mathrm{AIPW}}$ would satisfy the AIPW estimating equations for environment $e'$. Instead, a more reasonable idea is to identify a subset of predictors such that there exists an estimator supported on this subset for which the AIPW estimating equations hold, meaning that the imputation-bias-corrected gradient is zero along the corresponding coordinates in both environments. While finding the minimizer $\widehat{\boldsymbol{\beta}}^{(e)}_{\mathrm{AIPW}}$ by solving a single equation for environment $e$ is straightforward, it is less obvious how to determine a subset of variables that allows for a constrained minimizer satisfying two, or even more AIPW estimating equations across environments. On closer examination, the steps taken to derive the imputation bias-corrected penalty function (\ref{eq:penalty_adjust_more}) reveal an approach to address this question.

\subsection{Imputation Bias Corrected Empirical Objective $\widehat{\mathrm{Q}}_{\mathrm{Adj}}(\boldsymbol{\beta} ; \boldsymbol{\omega})$}\label{subsection:imputation_bias_corrected_Q}

To construct the full imputation bias-corrected empirical objective, we combine the previously derived imputation bias-corrected pooled $L_2$ loss, $\widehat{\mathrm{R}}_{\mathrm{Adj}}(\boldsymbol{\beta} ; \boldsymbol{\omega})$, from equation (\ref{eq:bia_corr_pool_L2_loss}) with the imputation bias-corrected penalty term, $\widehat{\mathrm{J}}_{\mathrm{Adj}}(\boldsymbol{\beta} ; \boldsymbol{\omega})$, given in equation (\ref{eq:penalty_adjust_more}). Together, with a tuning parameter $\gamma$, these components form an adjusted objective that accounts for biases in both the loss function and the regularization term due to inaccurate imputation model. The adjusted empirical objective can be written as:
\begin{equation}\label{eq:objective_adjust_whole}
	\widehat{\mathrm{Q}}_{\mathrm{Adj}}(\boldsymbol{\beta} ; \gamma, \boldsymbol{\omega}) = \widehat{\mathrm{R}}_{\mathrm{Adj}}(\boldsymbol{\beta} ; \boldsymbol{\omega}) + \gamma \widehat{\mathrm{J}}_{\mathrm{Adj}}(\boldsymbol{\beta} ; \boldsymbol{\omega}).
\end{equation}
We denote the minimizer of the objective $\widehat{\mathrm{Q}}_{\mathrm{Adj}}(\boldsymbol{\beta} ; \boldsymbol{\omega})$ as $\widehat{\boldsymbol{\beta}}_{\mathrm{Adj}}$, and call it Imputation-Adjusted Environment Invariant (IAEI) estimator.

In the process of constructing the objective $\widehat{\mathrm{Q}}_{\mathrm{Adj}}(\boldsymbol{\beta} ; \boldsymbol{\omega})$, we effectively account for the biases introduced by imputation in both the pooled least squares loss and the regularization term. As we will demonstrate in the next section, the IAEI estimator $\widehat{\boldsymbol{\beta}}_{\mathrm{Adj}}$ has favorable statistical properties, including variable selection consistency and non-asymptotic control over the $\ell_2$ estimation error with respect to the true parameter $\boldsymbol{\beta}^*$. Specifically, variable selection consistency ensures that the selected set of variables encompasses all LLS-invariant variables while excluding those that exhibit spurious linear relationships. Additionally, the non-asymptotic guarantees provide a bounded measure of accuracy in finite samples, indicating that the estimation error remains controlled under certain conditions.

\section{Theory}\label{section:theory}

Before presenting the main theoretical results, we first introduce several conditions that form the basis of our analysis, thereby allowing us to rigorously evaluate the statistical properties of the proposed methodology. Given that our approach leverages the non-asymptotic analysis techniques from \cite{fan2024environment}, the conditions we impose bear similarities to those in the referenced work. However, our theoretical focus diverges to address the challenges posed by imputation and bias correction.

\begin{cond}\label{cond:independent}
	For every environment $e\in\mathcal{E}$, the data $$(\boldsymbol{x}_1^{(e)}, y_1^{(e)}), \ldots,(\boldsymbol{x}_{n^{(e)}}^{(e)}, y_{n^{(e)}}^{(e)}), (\tilde{\boldsymbol{x}}_{n^{(e)}+1}^{(e)}, \tilde{y}_{n^{(e)}+1}^{(e)}), \ldots,(\tilde{\boldsymbol{x}}_{n^{(e)}+ m^{(e)}}^{(e)}, \tilde{y}_{n^{(e)}+m^{(e)}}^{(e)})$$ are are independently and identically distributed (i.i.d.) samples from the distribution $(\boldsymbol{x}^{(e)}, y^{(e)}) \sim \mu^{(e)}$, where $\mu^{(e)}\in \mathcal{U}_{\beta^*, \sigma^2}$ for some variance $\sigma^2$. Furthermore, the data from different environments are assumed to be independent.
\end{cond}

\begin{cond}\label{cond:pd_covariance_matrix}
	There exists some universal constants $\kappa_L \in(0,1]$ and $\kappa_U \in[1, \infty)$ such that
	\begin{equation}\label{eq:pd_parameters}
		\forall e \in \mathcal{E}, \quad \kappa_L \boldsymbol{I}_p \preceq \boldsymbol{\Sigma}^{(e)} \preceq \kappa_U \boldsymbol{I}_p .
	\end{equation}
\end{cond}

\begin{cond}\label{cond:subg_x}
	There exists some universal constant $\sigma_x \in[1, \infty)$ such that
	\begin{equation}
		\forall e \in \mathcal{E}, \boldsymbol{v} \in \mathbb{R}^p, \quad \mathbb{E}\left[\exp \left\{\boldsymbol{v}^{\top} \overline{\boldsymbol{\Sigma}}^{-1 / 2} \boldsymbol{x}^{(e)}\right\}\right] \leq \exp \left(\frac{\sigma_x^2}{2} \cdot\|\boldsymbol{v}\|_2^2\right).
	\end{equation}
\end{cond}

\begin{cond}\label{cond:subg_e}
	There exists some universal constant $\sigma_{\varepsilon} \in \mathbb{R}^{+}$ such that,
	\begin{equation}
		\forall e \in \mathcal{E}, \lambda \in \mathbb{R}, \quad \mathbb{E}\left[e^{\lambda \varepsilon^{(e)}}\right] \leq e^{\frac{1}{2} \lambda^2 \sigma_{\varepsilon}^2}
	\end{equation}
\end{cond}

\begin{cond}\label{cond:subg_z}
	There exists some universal constant $\sigma_{z} \in \mathbb{R}^{+}$ such that,
	\begin{equation}
		\forall e \in \mathcal{E}, \lambda \in \mathbb{R}, \quad \mathbb{E}\left[e^{\lambda (z^{(e)} - \eta^{(e)})}\right] \leq e^{\frac{1}{2} \lambda^2 \sigma_{z}^2} .
	\end{equation}
\end{cond}

Condition \ref{cond:independent} aligns with the data-generating processes (\ref{eq:linear_structure}) or (\ref{eq:sem}) described in Section \ref{subsection:DGP}, which also implies that $\mathbb{E}[\varepsilon^{(e)}]=0$. We work based on sub-Gaussian Conditions \ref{cond:pd_covariance_matrix}-\ref{cond:subg_z}, which are standard in non-asymptotic analysis, to allow our results to apply to a broad class of random variables. We focus on the scenario where the covariates are centered, i.e. $\mathbb{E}[\boldsymbol{x}^{(e)}]=\boldsymbol{0}$, as it is straightforward to extend the results to non-centered covariates. As discussed in Section \ref{subsection:imputation}, we treat the imputation function $\widehat{h}^{(e)}(\cdot)$ as a fixed, non-random function. In general, the imputation bias, or the expected imputation error $\eta^{(e)}$, is not zero. To account for this, we center the random imputation error $z^{(e)}$ around its mean $\eta^{(e)}$ in Condition \ref{cond:subg_z}.

Following the approach in \cite{fan2024environment}, we continue to use the concept of pooled linear spurious variables and provide a formal definition here for completeness:
\begin{definition}\label{definition:pooled_spurious_variables}
	We let $G_{\boldsymbol{\omega}}$ be the index set of all pooled linear spurious variables in environments $\mathcal{E}$ concerning weight $\omega^{(e)}$ for each environment $e\in\mathcal{E}$, that is, $G_{\boldsymbol{\omega}}=\{j \in[p]$ : $\left.\sum_{e \in \mathcal{E}}\omega^{(e)} \mathbb{E}[x_j^{(e)} \varepsilon^{(e)}] \neq 0\right\}$. We say a variable $x_j$ is a pooled linear spurious variable if $j \in G_{\boldsymbol{\omega}}$.
\end{definition}
In general, the spurious effects across different environments do not cancel out. Thus, intuitively, a variable $x_j$ belongs to $G_{\boldsymbol{\omega}}$ if there exist at least one environment $e\in\mathcal{E}$ such that $\mathbb{E}[x_j^{(e)} \varepsilon^{(e)}]\ne 0$.

Next, we introduce some frequently used notations along with their interpretations. These notations and their derivations originally appeared in \cite{fan2024environment}, and we restate them here with brief explanations for clarity. In the following discussion, all quantities will be depend on a given subset $S\subseteq [p]$ with $S\supseteq S^*$.

We start by considering the expression
\begin{equation}\label{eq:bs}
	\mathrm{b}_S=\left\|\sum_{e \in \mathcal{E}} \omega^{(e)} \mathbb{E}\left[\varepsilon^{(e)} x_S^{(e)}\right]\right\|_2^2.
\end{equation}
We then define the pooled population $L_2$ loss function, restricted to the set $S$, as
\begin{equation}\label{eq:pool_loss_pop_constrained_S}
	\mathrm{R}_S(\boldsymbol{\beta} ; \boldsymbol{\omega})=\sum_{e \in \mathcal{E}} \omega^{(e)} \mathbb{E}\left[\left|y^{(e)}-\boldsymbol{\beta}^{\top} \boldsymbol{x}_S^{(e)}\right|^2\right].
\end{equation}
Since this objevtive is quadratic, it has a unique minimzier, denoted as $\bar{\boldsymbol{\beta}}_S^\mathrm{R}$. The minimizer $\bar{\boldsymbol{\beta}}_S^\mathrm{R}$ can be viewed as the target of the pooled empirical least squares estimator, constructed using all labeled data while being constrained to the set $S$. One closed-form expression for this minimizer, derived as follows, is
\begin{equation}\label{eq:minimizer_pooled_L2}
	\bar{\boldsymbol{\beta}}_S^\mathrm{R} = \boldsymbol{\beta}^* + \bar{\Sigma}^{-1} \sum_{e\in\mathcal{E}} \mathbb{E}\left[\varepsilon^{(e)} \boldsymbol{x}_S^{(e)}\right].
\end{equation}
This derivation also yields a formula for the $\ell_2$ distance between $\bar{\boldsymbol{\beta}}_S^\mathrm{R}$ and $\boldsymbol{\beta}^*$, which can be interpreted as a measure of the deviation of the population-level constrained pooled linear regression parameter from the true invariant parameter:
\begin{equation}\label{eq:bar_beta_minus_beta_true_norm}
	\|\bar{\boldsymbol{\beta}}_S^\mathrm{R}-\boldsymbol{\beta}^* \|_2 = \left\|\bar{\Sigma}^{-1} \sum_{e\in\mathcal{E}} \mathbb{E}\left[\varepsilon^{(e)} \boldsymbol{x}_S^{(e)}\right]\right\|_2.
\end{equation}
From equation (\ref{eq:bar_beta_minus_beta_true_norm}) and under Condition \ref{cond:pd_covariance_matrix}, we can see that $b_S$ is proportional to this $\ell_2$ distance, or the magnitude of the deviation, i.e. $b_S \asymp\|\bar{\boldsymbol{\beta}}_S^{\mathrm{R}}-\boldsymbol{\beta}^*\|_2^2$.

Next, we consider the quantity
\begin{equation}\label{eq:ds}
	\overline{\mathrm{d}}_S=\sum_{e \in \mathcal{E}} \omega^{(e)}\left\|\boldsymbol{\beta}^{(e, S)}-\overline{\boldsymbol{\beta}}^{(S)}\right\|_2^2\quad \text{with} \quad \overline{\boldsymbol{\beta}}^{(S)}= \sum_{e' \in \mathcal{E}} \omega^{(e')} \boldsymbol{\beta}^{\left(e', S\right)}.
\end{equation}
The term $\boldsymbol{\beta}^{\left(e', S\right)}$ is defined as the optimal linear predictor at population-level in environment $e$,
constrained to the set $S$, given by
\begin{equation}\label{eq:best_linear_predictor_S}
	\boldsymbol{\beta}^{(e, S)}=\underset{\boldsymbol{\beta} \in \mathbb{R}^p, \operatorname{supp}(\boldsymbol{\beta}) \subseteq S}{\operatorname{argmin}} \mathrm{R}^{(e)}(\boldsymbol{\beta}),
\end{equation}
with $\mathrm{R}^{(e)}(\boldsymbol{\beta})$ defined earilier in equation (\ref{eq:population_L2_risk}). Furthermore, note that
\begin{align}
	\overline{\mathrm{d}}_S & = \sum_{e \in \mathcal{E}} \omega^{(e)}\left\|\boldsymbol{\beta}^{(e, S)}-\overline{\boldsymbol{\beta}}^{(S)}\right\|_2^2                                                                                                                       \\
	                        & = \sum_{e \in \mathcal{E}} \omega^{(e)}\left\|\boldsymbol{\beta}^{(e, S)}- \boldsymbol{\beta}^* + \boldsymbol{\beta}^*-\sum_{e' \in \mathcal{E}} \omega^{(e')} \boldsymbol{\beta}^{\left(e', S\right)}\right\|_2^2                              \\
	                        & = \sum_{e \in \mathcal{E}} \omega^{(e)}\left\|\{\boldsymbol{\beta}^{(e, S)}- \boldsymbol{\beta}^*\} - \sum_{e' \in \mathcal{E}} \omega^{(e')}\{ \boldsymbol{\beta}^{\left(e', S\right)}-\boldsymbol{\beta}^*\}\right\|_2^2 \label{eq:ds_decom}.
\end{align}
While the term $\bar{\boldsymbol{\beta}}_S^\mathrm{R}-\boldsymbol{\beta}^*$ reflects the deviation of the pooled linear regression parameter, here $\boldsymbol{\beta}^{(e, S)}- \boldsymbol{\beta}^*$ captures how the environment-specific linear regression parameter differs from the truth. Accordingly, the quantity in (\ref{eq:ds_decom}) indicates the variation of the divergences among different environments in $\mathcal{E}$.

With some intuition established for the quantities $\mathrm{b}_S$ and $\overline{\mathrm{d}}_S$, we now proceed to explore the properties of the population objective in equation (\ref{eq:fan_pop}). \cite{fan2024environment} demonstrate that when some pooled linear spurious variables exist, this objective is strongly convex around the true parameter $\boldsymbol{\beta}^*$, provided that certain identification conditions are satisfied and the regularization parameter $\gamma$ is sufficiently large. The required magnitude of $\gamma$ depends on the ratio of $\mathrm{b}_S$ to $\overline{\mathrm{d}}_S$. For completeness, we restate the identification condition and the strong convexity result near $\boldsymbol{\beta}^*$ here.

\begin{cond}\label{cond:identification}
	For any $S \subseteq[p]$ satisfying $S \cap G_{\boldsymbol{\omega}} \neq \emptyset$, there exists some $e, e^{\prime} \in \mathcal{E}$ such that $\boldsymbol{\beta}^{(e, S)} \neq \boldsymbol{\beta}^{\left(e^{\prime}, S\right)}$.
\end{cond}

\begin{theorem}\label{theorem:strong_convex_pop_eills}
	Assume Conditions \ref{cond:independent}-\ref{cond:pd_covariance_matrix} and \ref{cond:identification} hold. Then $\boldsymbol{\beta}^*$ is the unique minimizer of $\mathrm{Q}(\boldsymbol{\beta} ; \gamma, \boldsymbol{\omega})$ for large enough $\gamma$ : for any $\epsilon \in(0,1)$ and any $\gamma \geq \epsilon^{-1} \gamma^*$ with
	\begin{equation}\label{eq:gamma_star}
		\gamma^*=\left(\kappa_L\right)^{-3} \sup _{S: S \cap G_\omega \neq \emptyset}\left(\mathrm{b}_S / \overline{\mathrm{d}}_S\right),
	\end{equation}
	we have
	\begin{equation}
		\mathrm{Q}(\boldsymbol{\beta} ; \gamma, \boldsymbol{\omega})-\mathrm{Q}\left(\boldsymbol{\beta}^* ; \gamma, \boldsymbol{\omega}\right) \geq(1-\epsilon)\left\|\bar{\boldsymbol{\Sigma}}^{1 / 2}\left(\boldsymbol{\beta}-\boldsymbol{\beta}^*\right)\right\|_2^2+\kappa_L^2\left(\gamma-\epsilon^{-1} \gamma^*\right) \overline{\mathrm{d}}_{\mathrm{supp}(\boldsymbol{\beta})} .
	\end{equation}
\end{theorem}

Having reviewed the theoretical result on the population obejctive (\ref{eq:fan_pop}), we now turn our attention to deriving the key properties of the empirical estimator $\widehat{\boldsymbol{\beta}}_{\mathrm{Adj}}$, which provides statistical guarantees in the presence of missing labeled data. Before establishing the variable selection property and $\ell_2$ error bound of $\widehat{\boldsymbol{\beta}}_{\mathrm{Adj}}$, we first turn to an intermediate step: examining two critical quantities involving both the population and the imputed empirical objectives. These quantities will play a pivotal role in establishing non-asymptotic results, particularly regarding the convergence rate of $\widehat{\boldsymbol{\beta}}_{\mathrm{Adj}}$. By first deriving high-probability upper bounds on these expressions, we gain insights into the influence of the missing mechanism. 

\subsection{High Probability Upper Bounds for Some Key Quantities}\label{subsection:hpub_TR_TJ}

To begin, let $\mathcal{D}$ denote an operator that takes two functions $f\in\mathcal{F}$ and $g\in\mathcal{G}$, where $\mathcal{F}$ and $\mathcal{G}$ are spaces of functions defined on the same domain, and outputs their pointwise difference: $\mathcal{D}(f,g)(\cdot) = f(\cdot) - g(\cdot)$. In our anlaysis, for any arbitary $\boldsymbol{\beta}\ne \boldsymbol{\beta}^*$, we define the following two quantities:
\begin{equation}\label{eq:D_R_RAdj}
	\mathcal{D}_{\mathrm{R},\widehat{\mathrm{R}}_{\mathrm{Adj}}}(\boldsymbol{\beta}) \coloneqq  \mathcal{D}(\mathrm{R}, \widehat{\mathrm{R}}_{\mathrm{Adj}})(\boldsymbol{\beta}) - \mathcal{D}(\mathrm{R}, \widehat{\mathrm{R}}_{\mathrm{Adj}})(\boldsymbol{\beta}^*)
\end{equation}
and
\begin{equation}\label{eq:D_J_JAdj}
	\mathcal{D}_{\mathrm{J},\widehat{\mathrm{J}}_{\mathrm{Adj}}}(\boldsymbol{\beta}) \coloneqq   \mathcal{D}(\mathrm{J}, \widehat{\mathrm{J}}_{\mathrm{Adj}})(\boldsymbol{\beta}) - \mathcal{D}(\mathrm{J}, \widehat{\mathrm{J}}_{\mathrm{Adj}})(\boldsymbol{\beta}^*).
\end{equation}

Intuitively, because our target parameter is $\boldsymbol{\beta}^*$, we would hope that the discrepancy between the pooled population and empirical $L_2$ risks at $\boldsymbol{\beta}$ is not substantially greater than the discrepancies at $\boldsymbol{\beta}^*$. In other words, we expect that $\mathcal{D}_{\mathrm{R},\widehat{\mathrm{R}}_{\mathrm{Adj}}}(\boldsymbol{\beta})$ to be upper bounded. Similarly, we want the discrepancy between the population and empirical spuriousness measure for any set $S$ of variables to be not substantially greater than the discrepancy for the true set $S^*$ of variables. In other words, we expect that $\mathcal{D}_{\mathrm{J},\widehat{\mathrm{J}}_{\mathrm{Adj}}}(\boldsymbol{\beta})$ to be upper bounded.

One might initially think that, to ensure accuracy, we should try to make the discrepancy between the pooled population and empirical $L_2$ risks at $\boldsymbol{\beta}$ be close to that at $\boldsymbol{\beta}^*$; specifically, by showing that $|\mathcal{D}_{\mathrm{R},\widehat{\mathrm{R}}_{\mathrm{Adj}}}(\boldsymbol{\beta})|$ is bounded with high probability. Similarly, it might seem necessary to control the discrepancy between the pooled population and empirical spuriousness measure at $\boldsymbol{\beta}$ be close to that at $\boldsymbol{\beta}^*$ by ensuring that $|\mathcal{D}_{\mathrm{J},\widehat{\mathrm{J}}_{\mathrm{Adj}}}(\boldsymbol{\beta})|$ is bounded. While a two-sided bound for $\mathcal{D}_{\mathrm{R},\widehat{\mathrm{R}}_{\mathrm{Adj}}}(\boldsymbol{\beta})$ is indeed possible, deriving one for $\mathcal{D}_{\mathrm{J},\widehat{\mathrm{J}}_{\mathrm{Adj}}}(\boldsymbol{\beta})$ is more challenging. As we will show in Section \ref{subsection:nonasy_l2_error}, however, a two-sided bound for $\mathcal{D}_{\mathrm{J},\widehat{\mathrm{J}}_{\mathrm{Adj}}}(\boldsymbol{\beta})$ is not necessary. 

The study of high-probability upper bounds for $\mathcal{D}_{\mathrm{R},\widehat{\mathrm{R}}_{\mathrm{Adj}}}(\boldsymbol{\beta})$ and $\mathcal{D}_{\mathrm{J},\widehat{\mathrm{J}}_{\mathrm{Adj}}}(\boldsymbol{\beta})$ is fundamental for establishing variable selection property, which is descriped in Section \ref{subsection:nonasy_vsc}. These quantities directly measure the discrepancies between the empirical and population risk differences at $\boldsymbol{\beta}$ and $\boldsymbol{\beta}^*$. By controlling these discrepancies with high probability, the selection process inherently avoids choosing $\boldsymbol{\beta}$ that deviate substantially from $\boldsymbol{\beta}^*$. Consequently, the final estimator identifies the correct support, including the true important variables while excluding linearly spurious ones, thereby ensuring the variable selection property. Furthermore, in Section \ref{subsection:nonasy_l2_error}, we will show that the $\ell_2$ error bound for $\widehat{\boldsymbol{\beta}}_{\mathrm{Adj}}$ is controlled by the sum of $\mathcal{D}_{\mathrm{R},\widehat{\mathrm{R}}_{\mathrm{Adj}}}(\widehat{\boldsymbol{\beta}}_{\mathrm{Adj}})$ and $\mathcal{D}_{\mathrm{J},\widehat{\mathrm{J}}_{\mathrm{Adj}}}(\widehat{\boldsymbol{\beta}}_{\mathrm{Adj}})$. Consequently, obtaining explicit high-probability upper bounds on $\mathcal{D}_{\mathrm{R},\widehat{\mathrm{R}}_{\mathrm{Adj}}}(\boldsymbol{\beta})$ and $\mathcal{D}_{\mathrm{J},\widehat{\mathrm{J}}_{\mathrm{Adj}}}(\boldsymbol{\beta})$ for any arbitrary $\boldsymbol{\beta}$ enables us to ultimately establish an $\ell_2$ error bound for $\widehat{\boldsymbol{\beta}}_{\mathrm{Adj}}$. 

These two goals motivates our current effort to derive the bounds of $\mathcal{D}_{\mathrm{R},\widehat{\mathrm{R}}_{\mathrm{Adj}}}(\boldsymbol{\beta})$ and $\mathcal{D}_{\mathrm{J},\widehat{\mathrm{J}}_{\mathrm{Adj}}}(\boldsymbol{\beta})$ in terms of the total number of observations, the number of labeled observations, the number of missing-label observations, imputation bias, and the sub-Gaussian parameter of imputation error, providing insight into how these factors interplay to influence the variable selection consistency and the convergence of our estimator $\widehat{\boldsymbol{\beta}}_{\mathrm{Adj}}$. In the following, we first list two general results in regards to the upper bounds of these two quantities. 

\begin{lemma}[Two-side Bound for $\mathcal{D}_{\mathrm{R},\widehat{\mathrm{R}}_{\mathrm{Adj}}}(\boldsymbol{\beta})$]\label{lemma:two_sided_bound_R}
	Suppose Conditions \ref{cond:independent}-\ref{cond:subg_z} hold. Define the event
	\begin{align}
		\mathcal{A}_{1,t} & = \left\{  \forall \boldsymbol{\beta}\in\mathbb{R}^p,  \left|   \mathcal{D}(\mathrm{R}, \widehat{\mathrm{R}}_{\mathrm{Adj}})(\boldsymbol{\beta}) - \mathcal{D}(\mathrm{R}, \widehat{\mathrm{R}}_{\mathrm{Adj}})(\boldsymbol{\beta}^*)  \right| \right.\notag                                                                   \\
		                  & \quad  \quad  \left. \le  c_1 \left(\kappa_U\sigma_x^2\|\boldsymbol{\beta} - \boldsymbol{\beta}^*\|_2^2\,\sqrt{\frac{t+p}{N_*}}  +\kappa_U^{1/2}\sigma_x\sigma_{\varepsilon}\|\boldsymbol{\beta} - \boldsymbol{\beta}^*\|_2\,\sqrt{\frac{t+p}{N_*}}\right.\right.\notag                                                        \\
		                  & \quad\quad  \quad \quad\left.\left. +\kappa_U^{1/2}\sigma_x\sigma_{z}\|\boldsymbol{\beta} - \boldsymbol{\beta}^*\|_2\,\sqrt{\frac{t+p}{m_*^{\widehat{\tau}}}}  +
		\kappa_U^{1/2}\sigma_x\|\boldsymbol{\beta} - \boldsymbol{\beta}^*\|_2\,\sqrt{\frac{t+p}{m_*^{\widehat{\tau},|\eta|}}}\right. \right.\notag                                                                                                                                                                                                         \\
		                  & \quad\quad  \quad \quad \left.\left.+ \kappa_U^{1/2}\sigma_x\sigma_{z}\|\boldsymbol{\beta} - \boldsymbol{\beta}^*\|_2\,\sqrt{\frac{t+p}{n_*^{\widehat{\tau}}}}  + \kappa_U^{1/2}\sigma_x\|\boldsymbol{\beta} - \boldsymbol{\beta}^*\|_2\,\sqrt{\frac{t+p}{n_*^{\widehat{\tau},|\eta|}}}\right)\right\}.\label{eq:inlemma_a1_2}
	\end{align}
	with some universal constants $c_1$. We have $\mathbb{P}(\mathcal{A}_{1, t}) \geq 1-5 e^{-t}$, for any $t \geq 0$.
\end{lemma}

\begin{lemma}[One-side Bound for $\mathrm{J}$]\label{lemma:one_sided_bound_J} 
	Suppose Conditions \ref{cond:independent}-\ref{cond:subg_z} hold. Then we have $\mathbb{P}[\mathcal{A}_{2,t}] \ge 1-32e^{-t}$, for any $t \in\left(0, n_{\text {min}}-\log (4|\mathcal{E}|)-p\right]$, where the event $\mathcal{A}_{2,t}$ is defined in (\ref{eq:J_general_bound}) in appendix. 
\end{lemma}

In the following analysis, we derive high-probability upper bounds for $\mathcal{D}_{\mathrm{R},\widehat{\mathrm{R}}_{\mathrm{Adj}}}(\boldsymbol{\beta})$ and $\mathcal{D}_{\mathrm{J},\widehat{\mathrm{J}}_{\mathrm{Adj}}}(\boldsymbol{\beta})$ across two scenarios. The first examines high missingness combined with a reliable imputation model, where we demonstrate that the derived bounds are tighter than those in \cite{fan2024environment} under specific conditions. The second explores high missingness with an imprecise imputation model, showing that under appropriate assumptions, the bounds still converge, ensuring the stability of our results in less ideal settings 

\subsubsection{Bounds Under High Missingness and Reliable Imputation}\label{subsubsection:bounds_high_miss_realiable_imp}

For completeness, we analyze the bound in (\ref{eq:inlemma_a1_2}) across four specific cases. Among these, Case 4 is particularly relevant to the setting of high missingness and reliable imputation. 

\begin{corollary}\label{corollary:R_four_cases}
We analysis the bound in (\ref{eq:inlemma_a1_2}) under four specific cases. 

Case 1: $\widehat{\tau}_{\max}\le 1/2$ and $\eta_{\mathrm{max}} \ge 1$. In this scenario, we have $\mathbb{P}(\mathcal{A}_{1,t,Case1})\geq 1-5 e^{-t},\forall t\geq 0$, where the event $\mathcal{A}_{1,t,Case1}$ is defined as follows:
	\begin{align}
		\mathcal{A}_{1,t, Case1} & = \left\{  \forall \boldsymbol{\beta}\in\mathbb{R}^p, \left|    \mathcal{D}(\mathrm{R}, \widehat{\mathrm{R}}_{\mathrm{Adj}})(\boldsymbol{\beta}) - \mathcal{D}(\mathrm{R}, \widehat{\mathrm{R}}_{\mathrm{Adj}})(\boldsymbol{\beta}^*)  \right|\right.\notag \\
		                         & \quad\left. \le  c_1\left(\kappa_U\sigma_x^2\|\boldsymbol{\beta} - \boldsymbol{\beta}^*\|_2^2\,\sqrt{\frac{t+p}{N_*}} +
		\kappa_U^{1/2}\sigma_x\sigma_{z}\sigma_{\varepsilon}\|\boldsymbol{\beta} - \boldsymbol{\beta}^*\|_2\,\sqrt{\frac{t+p}{m_*^{\widehat{\tau},|\eta|}}}\right)\right\},\label{eq:inlemma_a1_4}
	\end{align}

	Case 2: $\widehat{\tau}_{\max}\le 1/2$ and $\eta_{\mathrm{max}} < 1$. In this scenario, we have $\mathbb{P}(\mathcal{A}_{1,t,Case2})\geq 1-5 e^{-t},\forall t\geq 0$, where the event $\mathcal{A}_{1,t,Case2}$ is defined as follows:
	\begin{align}
		\mathcal{A}_{1,t, Case2} & = \left\{  \forall \boldsymbol{\beta}\in\mathbb{R}^p, \left|    \mathcal{D}(\mathrm{R}, \widehat{\mathrm{R}}_{\mathrm{Adj}})(\boldsymbol{\beta}) - \mathcal{D}(\mathrm{R}, \widehat{\mathrm{R}}_{\mathrm{Adj}})(\boldsymbol{\beta}^*)   \right|\right.\notag                              \\
		                         & \quad \left.\le  c_1\left( \kappa_U\sigma_x^2\|\boldsymbol{\beta} - \boldsymbol{\beta}^*\|_2^2\,\sqrt{\frac{t+p}{N_*}}+\kappa_U^{1/2}\sigma_x\sigma_{z}\sigma_{\varepsilon}\|\boldsymbol{\beta} - \boldsymbol{\beta}^*\|_2\,\sqrt{\frac{t+p}{N_*}}\right)\right\},\label{eq:inlemma_a1_5}
	\end{align}

	Case 3: $\widehat{\tau}_{\min}> 1/2$ and $ \eta_{\mathrm{max}} \ge 1 $. In this scenario, we have $\mathbb{P}(\mathcal{A}_{1,t,Case3})\geq 1-5 e^{-t}, \, \forall t\geq 0$, where the event $\mathcal{A}_{1,t,Case3}$ is defined as follows:
	\begin{align}
		\mathcal{A}_{1,t, Case3} & = \left\{  \forall \boldsymbol{\beta}\in\mathbb{R}^p, \left|    \mathcal{D}(\mathrm{R}, \widehat{\mathrm{R}}_{\mathrm{Adj}})(\boldsymbol{\beta}) - \mathcal{D}(\mathrm{R}, \widehat{\mathrm{R}}_{\mathrm{Adj}})(\boldsymbol{\beta}^*)   \right|\right.\notag \\
		                         & \quad \left.\le c_1 \left(\kappa_U\sigma_x^2\|\boldsymbol{\beta} - \boldsymbol{\beta}^*\|_2^2\,\sqrt{\frac{t+p}{N_*}} +
		\kappa_U^{1/2}\sigma_x\sigma_{z}\sigma_{\varepsilon}\|\boldsymbol{\beta} - \boldsymbol{\beta}^*\|_2\,\sqrt{\frac{t+p}{n_*^{\widehat{\tau},|\eta|}}}\right)\right\},\label{eq:inlemma_a1_7}
	\end{align}

	Case 4: $\widehat{\tau}_{\min}> 1/2$ and $ \eta_{\mathrm{max}} < 1 $. In this scenario, we have $\mathbb{P}(\mathcal{A}_{1,t,Case4})\geq 1-5 e^{-t}, \, \forall t\geq 0$, where the event $\mathcal{A}_{1,t,Case4}$ is defined as follows:
	\begin{align}
		\mathcal{A}_{1,t, Case4} & = \left\{  \forall \boldsymbol{\beta}\in\mathbb{R}^p,  \left|   \mathcal{D}(\mathrm{R}, \widehat{\mathrm{R}}_{\mathrm{Adj}})(\boldsymbol{\beta}) - \mathcal{D}(\mathrm{R}, \widehat{\mathrm{R}}_{\mathrm{Adj}})(\boldsymbol{\beta}^*)   \right| \right.\notag                                              \\
		                         & \quad \left.\le   c_1\left(\kappa_U\sigma_x^2\|\boldsymbol{\beta} - \boldsymbol{\beta}^*\|_2^2\,\sqrt{\frac{t+p}{N_*}}+\kappa_U^{1/2}\sigma_x\sigma_{z}\sigma_{\varepsilon}\|\boldsymbol{\beta} - \boldsymbol{\beta}^*\|_2\,\sqrt{\frac{t+p}{n_*^{\widehat{\tau}}}}\right)\right\}.\label{eq:inlemma_a1_8}
	\end{align}
\end{corollary}

Together with Lemma \ref{lemma:two_sided_bound_R}, it establishes a high-probability two-sided bound for $\mathcal{D}_{\mathrm{R},\widehat{\mathrm{R}}_{\mathrm{Adj}}}(\boldsymbol{\beta})$ that holds for any $\boldsymbol{\beta}$ across four scenarios that vary based on the ratio of observed to missing label observations and the level of imputation bias. A comparable bound appears in \cite{fan2024environment}, though it is based on a complete analysis:
\begin{align}
	\left|    \mathcal{D}(\mathrm{R}, \widehat{\mathrm{R}}_{\mathrm{Adj}})(\boldsymbol{\beta}) - \mathcal{D}(\mathrm{R}, \widehat{\mathrm{R}}_{\mathrm{Adj}})(\boldsymbol{\beta}^*)   \right| \le  c_1\left( \kappa_U\sigma_x^2\|\boldsymbol{\beta} - \boldsymbol{\beta}^*\|_2^2\,\sqrt{\frac{t+p}{n_*}}+\kappa_U^{1/2}\sigma_x\sigma_{\varepsilon}\|\boldsymbol{\beta} - \boldsymbol{\beta}^*\|_2\,\sqrt{\frac{t+p}{n_*}}\right).\label{eq:fan_inlemma_a1}
\end{align}
Next, we compare our bound in each scenario to the result in \cite{fan2024environment}, demonstrating that our imputation-adjusted approach consistently yields a tighter bound under certain conditions which is reasonable to achive in practice. This comparison highlights the improved accuracy of our method in estimating parameters in the presence of missing labels, as supported by simulations.

By the definition of $N_*$ and $n_*$, it is clear that the first term in (\ref{eq:inlemma_a1_4})-(\ref{eq:inlemma_a1_8}) are all smaller than the first term in (\ref{eq:fan_inlemma_a1}). Thus, we focus our comparison on the second term. In Case 1, when $1< |\eta^{(e)}| < N^{(e)}/ \{n^{(e)} \sigma_z^2\}$, we have $\sigma_{z}^2 / m_*^{\widehat{\tau},|\eta|} < 1/ n_*$, so the second term in (\ref{eq:inlemma_a1_4}) is smaller than the second term in (\ref{eq:fan_inlemma_a1}). In Case 2, when $1 < N^{(e)} / \{n^{(e)}\sigma_{z}^2 \}$, we find $\sigma_{z}^2 / N_* < 1/ n_*$, so the second term in (\ref{eq:inlemma_a1_5}) is smaller than the second term in (\ref{eq:fan_inlemma_a1}). We then consider Case 3. When $1< |\eta^{(e)}| <  N^{(e)}/\{m^{(e)}\sigma_{z}^2\}$, we have $\sigma_{z}^2/ n_*^{\widehat{\tau},|\eta|} < 1/ n_*$, making the second term in (\ref{eq:inlemma_a1_7}) smaller than the second term in (\ref{eq:fan_inlemma_a1}). Lastly, in Case 4, when $ 1< N^{(e)}/\{m^{(e)}\sigma_z^2\}$, we find $\sigma_{z}^2 / n_*^{\widehat{\tau}} < 1/ n_*$, implying the second term in (\ref{eq:inlemma_a1_8}) is smaller than that in (\ref{eq:fan_inlemma_a1}). Taken together, a sufficient condition to guarantee that the bounds in all four cases are tighter than the bound in (\ref{eq:fan_inlemma_a1}) is
\begin{equation}\label{eq:suffi_cond1}
	\forall e\in\mathcal{E},  \sigma_{z} < \sqrt{\frac{N^{(e)}}{ n^{(e)}}}\wedge  \sqrt{\frac{N^{(e)}}{ m^{(e)} }}.
\end{equation}
The parameter $\sigma_{z}^2$ represents the variation of the imputation error. When the imputation model is stable, condition (\ref{eq:suffi_cond1}) is more like to hold. This result demonstrates how our approach leverages missing label observations and the imputation model to achieve a tighter bound for finite sample sizes.

So far, we focused on $\mathcal{D}_{\mathrm{R},\widehat{\mathrm{R}}_{\mathrm{Adj}}}(\boldsymbol{\beta})$, providing a two-sided bound for each of the four scenarios and identifying a sufficient condition for consistent improvement over the bound in (\ref{eq:fan_inlemma_a1}). Next, we turn our attention to $\mathcal{D}_{\mathrm{J},\widehat{\mathrm{J}}_{\mathrm{Adj}}}(\boldsymbol{\beta})$. Unlike $\mathcal{D}_{\mathrm{R},\widehat{\mathrm{R}}_{\mathrm{Adj}}}(\boldsymbol{\beta})$, where the bounds could be readily separated across scenarios, the complexity of the upper bound for $\mathcal{D}_{\mathrm{J},\widehat{\mathrm{J}}_{\mathrm{Adj}}}(\boldsymbol{\beta})$, as shown in the inequality (\ref{eq:J_general_bound}) in supplementary materials, makes it impractical to separate and simplify results across multiple scenarios. Therefore, we directly focus on deriving an explicit upper bound in the scenario characterized by high missingness and an accurate imputation model.  Following this context, we then present a sufficient condition that ensures our upper bound on $\mathcal{D}_{\mathrm{J},\widehat{\mathrm{J}}_{\mathrm{Adj}}}(\boldsymbol{\beta})$ remains tighter than the counterpart in Lemma C.6 of \cite{fan2024environment}.

\begin{corollary}[One-side Bound for $\mathcal{D}_{\mathrm{J},\widehat{\mathrm{J}}_{\mathrm{Adj}}}(\boldsymbol{\beta})$]\label{corollary:oneside_boundJ_high_missing_good_imp}
	Suppose Conditions \ref{cond:independent}-\ref{cond:subg_z} hold, $\widehat{\tau}_{\min} > 0.618$, and $\eta_{\max} < 1$. Define the event  
	\begin{align}
		\mathcal{A}_{2,t} &= \left[  \forall \boldsymbol{\beta}\in\mathbb{R}^p, \frac{1}{c_1}\left\{\mathcal{D}(\mathrm{J}, \widehat{\mathrm{J}}_{\mathrm{Adj}})(\boldsymbol{\beta}) - \mathcal{D}(\mathrm{J}, \widehat{\mathrm{J}}_{\mathrm{Adj}})(\boldsymbol{\beta}^*)\right\}\right.\notag\\
		&\quad\quad\quad\quad\quad\quad \left.  \le \kappa_U^2 \sigma_x^2\left\|\boldsymbol{\beta}-\boldsymbol{\beta}^*\right\|_2^2\sqrt{\frac{t+p}{N_*}} \right.\notag\\
		&\quad\quad\quad\quad\quad\quad\left.   +   \kappa_U^{3 / 2}  \sigma_x^2 \sigma_{\varepsilon}\sigma_z \left\|\boldsymbol{\beta}-\boldsymbol{\beta}^*\right\|_2\sqrt{\frac{t+p}{n_*^{\widehat{\tau}}}}\right.\notag\\
		&\quad\quad\quad\quad\quad\quad \left.+ \kappa_U^{1 / 2} \sigma_x \sigma_{\varepsilon} \sigma_z \sqrt{\frac{t+p}{n^{\widehat{\tau}^2}_*}} \times \sqrt{\sum_{e \in \mathcal{E}} \omega^{(e)}\left\|\mathbb{E}\left[x_S^{(e)} \varepsilon^{(e)}\right]\right\|_2^2}\right.\notag  \\ 
		&\quad\quad\quad\quad\quad\quad \left. + \kappa_U \sigma_x\sigma^2_{\varepsilon}\sigma_z\frac{t+p}{n^{\widehat{\tau}}_*} \right.\notag  \\ 
		&\quad\quad\quad\quad\quad\quad \left.+\kappa_U \sigma_x^2 \sigma^2_{\varepsilon} \sigma_z^2\frac{\log \left(4|\mathcal{E}|\left|S^*\right|\right)+t}{\bar{n}}\left|S^* \backslash S\right| \right.\notag\\
		&\quad\quad\quad\quad\quad\quad \left. + \kappa_U^{3 / 2} \sigma_x^3\sigma_{\varepsilon}\sigma_z\frac{p+\log(2|\mathcal{E}|)+t}{\overline{\sqrt{nN}}^{\widehat{\tau}}}  \|\boldsymbol{\beta}-\boldsymbol{\beta}^*\|_2  \right]
		\label{eq:inlemma_boundJ_step1}
	\end{align}
	for some universal constants $c_1$. We have $\mathbb{P}[\mathcal{A}_{2,t}] \ge 1-32e^{-t}$, for any $t \in\left(0, n_{\text {min}}-\log (4|\mathcal{E}|)-p\right]$.
\end{corollary}

In Corollary \ref{corollary:oneside_boundJ_high_missing_good_imp}, we focus on the scenario characterized by high missingness and an accurate imputation model in each environment, i.e. $\widehat{\tau}_{\min} > 0.618$, and $\eta_{\max} < 1$. This scenario is closely related to Case 4 in Lemma \ref{lemma:two_sided_bound_R}, where $\widehat{\tau}_{\min}> 0.5$ and $\eta_{\mathrm{max}} < 1$. The upper bound presented in (\ref{eq:inlemma_boundJ_step1}) corresponds to a slightly more challenging scenario, assuming that in each $e\in\mathcal{E}$, more than $61.8\%$ of the observations have missing labels.

It is natural to ask what the bound would look like in the intermediate range, where $0.5 < m^{(e)}/N^{(e)} < 0.618, \forall e\in \mathcal{E}$. In this case, the sample size term $n^{\widehat{\tau}^2}_*$ exceeds $N_*$, and based on the general result in (\ref{eq:J_general_bound}) provided in the appendix, the third term in (\ref{eq:inlemma_boundJ_step1}) becomes $\kappa_U^{1 / 2} \sigma_x \sigma_{\varepsilon} \sigma_z \sqrt{{(t+p)}/{N_*}} \times \sqrt{\sum_{e \in \mathcal{E}} \omega^{(e)}\|\mathbb{E}[x_S^{(e)} \varepsilon^{(e)}]\|_2^2}$. Note that $N_*$ accounts for all observations, both those with and without labels, making it the optimal sample size quantity. As a result, the third term of the upper bound, proportional to $1/\sqrt{N_*}$, achieves its tightest form in cases where the missing ratio in each environment lies between $0.5$ and $0.618$. The scenarios with greater missingness, i.e. $\widehat{\tau}_{\min} > 0.618$ present a more interesting and critical case to study. In this setting, $n^{\widehat{\tau}^2}_*$ is smaller than $N_*$, and thus the corresponding upper bound is more impacted by the presence of large missingness. Moreover, any conclusions drawn for this more challenging scenario naturally apply to less severe cases, such as those where the missing ratio exceeds $50\%$ but remains below $61.8\%$. For these reasons, we focus our analysis on the more challenging case where $\widehat{\tau}_{\min} > 0.618$, as it not only addresses the impact of large missingness on the upper bound but also ensures that the findings extend to less severe scenarios. 

Next, we compare the derived upper bound for $\mathcal{D}_{\mathrm{J},\widehat{\mathrm{J}}_{\mathrm{Adj}}}(\boldsymbol{\beta})$ with the corresponding result in \cite{fan2024environment}, which only uses label-observed data, demonstrating how our imputation-adjusted approach yields tighter bounds under specific conditions in terms of the variability of the imputation model. For completeness, we restate the bound from \cite{fan2024environment}, which is given by:
\begin{align}
	\frac{1}{c_1}\left\{\mathcal{D}(\mathrm{J}, \widehat{\mathrm{J}}_{\mathrm{Adj}})(\boldsymbol{\beta}) - \mathcal{D}(\mathrm{J}, \widehat{\mathrm{J}}_{\mathrm{Adj}})(\boldsymbol{\beta}^*)\right\} &\leq  \kappa_U^2 \sigma_x^2\left\|\boldsymbol{\beta}-\boldsymbol{\beta}^*\right\|_2^2\sqrt{\frac{t+p}{n_*}}\notag \\
	&\quad+ \kappa_U^{3 / 2} \sigma_x^2 \sigma_{\varepsilon}\left\|\boldsymbol{\beta}-\boldsymbol{\beta}^*\right\|_2 \sqrt{\frac{t+p}{n_*}}\notag\\
	&\quad+ \kappa_U^{1 / 2} \sigma_x \sigma_{\varepsilon} \sqrt{\frac{t+p}{n_*}} \times \sqrt{\sum_{e \in \mathcal{E}} \omega^{(e)}\left\|\mathbb{E}\left[\boldsymbol{x}_{S}^{(e)}{ \varepsilon}^{(e)}\right]\right\|_2^2} \notag\\
	&\quad+\kappa_U \sigma_x \sigma_{\varepsilon}^2 \frac{t+p}{n_*}\notag\\
	&\quad+  \kappa_U \sigma_x^2 \sigma_{\varepsilon}^2 \frac{\log \left(4 |S^*||\mathcal{E}|\right)+t}{\bar{n}} \left|S^* \backslash S\right|\notag\\
	&\quad +  \kappa_U^{3 / 2} \sigma_x^3 \sigma_{\varepsilon} \frac{p+\log(2|\mathcal{E}|)+t}{\bar{n}} \left\|\boldsymbol{\beta}-\boldsymbol{\beta}^*\right\|_2 \label{eq:fan_inlemma_a2}
\end{align}

The first term in the upper bound (\ref{eq:inlemma_boundJ_step1}) is clearly smaller than the corresponding term in the upper bound (\ref{eq:fan_inlemma_a2}). To compare the second terms in those two upper bounds, observe that when $ 1< N^{(e)}/\{m^{(e)}\sigma_z^2\}$, it follows that $\sigma_{z}^2 / n_*^{\widehat{\tau}} < 1/ n_*$, which ensures that the second term in (\ref{eq:inlemma_a1_8}) is smaller than the corresponding term in (\ref{eq:fan_inlemma_a1}). For the third term, if $1 < N^{(e)}/\{ m^{(e)}  \sigma_z\}$, then $\sigma_{z}^2 / n_*^{\widehat{\tau}^2} < 1/ n_*$, implies the one in (\ref{eq:inlemma_boundJ_step1}) is smaller. Next, when $1 < N^{(e)}/ \{m^{(e)} \sigma_z\}$, we have $\sigma_{z} / n_*^{\widehat{\tau}} < 1/ n_*$, which leads to the forth term in (\ref{eq:inlemma_boundJ_step1}) is smaller than the one in (\ref{eq:fan_inlemma_a2}). For the fifth term, when $\sigma_{z}^2 <1$, we have the fifth term in (\ref{eq:inlemma_boundJ_step1}) is smaller than the corresponding term in (\ref{eq:fan_inlemma_a2}). Lastly, when $1 < \sqrt{N^{(e)}/n^{(e)}} (N^{(e)}/m^{(e)}) /\sigma_z$, we have $\sigma_{z}^2 / \overline{\sqrt{nN}}^{\widehat{\tau}} < 1/ \bar{n}$, so the last term in (\ref{eq:inlemma_boundJ_step1}) is smaller than the corresponding term in (\ref{eq:fan_inlemma_a2}). Taken together, a sufficient condition ensuring that the bound in (\ref{eq:inlemma_boundJ_step1}) is tighter than the one in (\ref{eq:fan_inlemma_a1}) is
\begin{equation}\label{eq:suffi_cond2}
	\sigma_{z} < 1,
\end{equation}
which also implies that (\ref{eq:suffi_cond1}) is satisfied. 

In summary, we analyzed the scenario of high missingness, where the proportion of missing label observations exceeds $0.618$. Under the conditions that the sub-Gaussian parameter of the imputation error satisfies $\sigma_z < 1$ and imputation bias satisfies $\eta_{\max} < 1$, our imputation-adjusted bound consistently outperforms the bound from the complete analysis in finite sample settings.

\subsubsection{Bounds Under High Missingness and Imprecise Imputation}\label{subsubsection:bounds_imprecise_imp}

Having established tighter bounds under high missingness with accurate imputation, a natural question arises: how do the bounds behave when the imputation is not highly accurate but still offers meaningful prediction for the label? In this section, we explore this scenario and show that, under reasonable assumptions, our approach continues to exhibit convergence properties despite some degree of imputation error.

Note that the scenario of high missingness with imprecise imputation aligns with the conditions outlined in Case 3 in Corollary \ref{corollary:R_four_cases}.  Specifically, when $\widehat{\tau}_{\min}> 1/2$ and $ \eta_{\mathrm{max}} \ge 1$, for any $ \boldsymbol{\beta}\in\mathbb{R}^p$, we have 
\begin{align}
	\left|    \mathcal{D}(\mathrm{R}, \widehat{\mathrm{R}}_{\mathrm{Adj}})(\boldsymbol{\beta}) - \mathcal{D}(\mathrm{R}, \widehat{\mathrm{R}}_{\mathrm{Adj}})(\boldsymbol{\beta}^*)   \right| \le c_1 \left(\kappa_U\sigma_x^2\|\boldsymbol{\beta} - \boldsymbol{\beta}^*\|_2^2\,\sqrt{\frac{t+p}{N_*}} + \kappa_U^{1/2}\sigma_x\sigma_{z}\sigma_{\varepsilon}\|\boldsymbol{\beta} - \boldsymbol{\beta}^*\|_2\,\sqrt{\frac{t+p}{n_*^{\widehat{\tau},|\eta|}}}\right)\label{eq:inlemma_boundR_case3},
\end{align}
with high probability. 

Next, we establish the upper bound for $\mathcal{D}_{\mathrm{J},\widehat{\mathrm{J}}_{\mathrm{Adj}}}(\boldsymbol{\beta})$ in this scenario as follows. 

\begin{corollary}\label{corollary:oneside_boundJ_high_missing_imprecise_imp}
	Suppose Conditions \ref{cond:independent}-\ref{cond:subg_z} hold, $\widehat{\tau}_{\min} > 0.618$, and $\eta_{\max} \ge 1$. Define the event  
	\begin{align}
		\mathcal{A}_{2,t} &= \left[  \forall \boldsymbol{\beta}\in\mathbb{R}^p, \frac{1}{c_1}\left\{\mathcal{D}(\mathrm{J}, \widehat{\mathrm{J}}_{\mathrm{Adj}})(\boldsymbol{\beta}) - \mathcal{D}(\mathrm{J}, \widehat{\mathrm{J}}_{\mathrm{Adj}})(\boldsymbol{\beta}^*)\right\}\right.\notag\\
		&\quad\quad\quad\quad\quad\quad \left.  \le \kappa_U^2 \sigma_x^2\left\|\boldsymbol{\beta}-\boldsymbol{\beta}^*\right\|_2^2\sqrt{\frac{t+p}{N_*}} \right.\notag\\
		&\quad\quad\quad\quad\quad\quad\left.   +   \kappa_U^{3 / 2}  \sigma_x^2 \sigma_{\varepsilon}\sigma_z \left\|\boldsymbol{\beta}-\boldsymbol{\beta}^*\right\|_2\sqrt{\frac{t+p}{n_*^{\widehat{\tau}, |\eta|}}}\right.\notag\\
		&\quad\quad\quad\quad\quad\quad \left.+ \kappa_U^{1 / 2} \sigma_x \sigma_{\varepsilon} \sigma_z \sqrt{\frac{t+p}{n_*^{\widehat{\tau}^2, |\eta|^2}}} \times \sqrt{\sum_{e \in \mathcal{E}} \omega^{(e)}\left\|\mathbb{E}\left[x_S^{(e)} \varepsilon^{(e)}\right]\right\|_2^2}\right.\notag  \\ 
		&\quad\quad\quad\quad\quad\quad \left. + \kappa_U \sigma_x\sigma^2_{\varepsilon}\sigma_z\frac{t+p}{n_*^{\widehat{\tau},|\eta|}} \right.\notag  \\ 
		&\quad\quad\quad\quad\quad\quad \left.+\kappa_U \sigma_x^2 \sigma^2_{\varepsilon} \sigma_z^2\frac{\log \left(4|\mathcal{E}|\left|S^*\right|\right)+t}{\bar{n}^{|\eta|^2}}\left|S^* \backslash S\right| \right.\notag\\
		&\quad\quad\quad\quad\quad\quad \left. + \kappa_U^{3 / 2} \sigma_x^3\sigma_{\varepsilon}\sigma_z\frac{p+\log(2|\mathcal{E}|)+t}{\overline{\sqrt{nN}}^{\widehat{\tau},|\eta|}}  \|\boldsymbol{\beta}-\boldsymbol{\beta}^*\|_2  \right]
		\label{eq:inlemma_boundJ_step2}
	\end{align}
	for some universal constants $c_1$. We have $\mathbb{P}[\mathcal{A}_{2,t}] \ge 1-32e^{-t}$, for any $t \in\left(0, n_{\text {min}}-\log (4|\mathcal{E}|)-p\right]$.
\end{corollary}
This result highlights how the imputation bias $|\eta^{(e)}|$, being greater than $1$, influences the upper bound and ultimately affects the later results on the variable selection property and the rate of convergence, as shown later in Sections \ref{subsection:nonasy_vsc} and \ref{subsection:nonasy_l2_error}. More specifically, under the scenario of large missingness and an imputation model yielding a bias that exceeds $1$, ensuring a smaller upper bound requires controlling the magnitude of the imputation bias to prevent it from becoming excessively large. Note that the above upper bound is controlled by $\bar{n}^{|\eta|^2}$. To ensure the convergence, it is sufficient for the imputation bias to satisfy the condition
\begin{equation}\label{eq:suf_condition_control_upperbound}
	|\eta^{(e)}| \ll \sqrt{n^{(e)}}, \forall e\in\mathcal{E}.
\end{equation}

When the imputation is moderate, this suggests that increasing $n^{(e)}$, the number of labeled observations, can improve the result. In applications where the amount of labeled data can be adjusted—often through human annotation—an estimate of the magnitude of the imputation bias $|\eta^{(e)}|$ in each environment can help practitioners decide whether it is necessary to allocate additional resources for labeling. If the imputation bias suggests that the current labeled data is insufficient, it may be worth investing in more labeled observations to achieve better results.

To conclude, we analyzed the scenario of large missingness with an imprecise imputation model and showed that controlling the magnitude of the imputation bias to be smaller than the square root of $n^{(e)}$ in each environment $e$ is crucial to ensure the convergence of the upper bound. These results highlight the importance of balancing the number of labeled observations and the quality of imputation to mitigate the impact of large missingness on the performance of the estimator, ensuring that the proposed approach continues to deliver reliable results even in the presence of imprecise imputation.

\subsection{Non-asymptotic Variable Selection Property}\label{subsection:nonasy_vsc}

In this section, we establish the non-asymptotic variable selection properties of the Imputation-Adjusted Environment Invariant (IAEI) estimator $\widehat{\boldsymbol{\beta}}_{\mathrm{Adj}}$. Our goal is to show that even in the presence of missing labeled data, the estimator can keep all the imporant variables while excluding all the pooled linear spurious ones. The analysis provides finite-sample guarantees, detailing how the estimator's variable selection performance depends on factors such as the proportion of missing data and the regularization parameter $\gamma$. This result offers insight into the conditions under which the IAEI estimator performs a good variable selection despite the challenges posed by incomplete data.

\begin{theorem}[Non-asymptotic Variable Selection Property]\label{theorem:nonasy_vsc}
	Define
	$$
		\mathbf{s}_{+}=\min _{j \in S^*}\left|\beta_j^*\right|^2 \quad \text { and } \quad \mathbf{s}_{-}=\min _{S \subseteq[p], S \cap G_\omega \neq \emptyset} \overline{\mathrm{d}}_S
	$$
	Suppose Conditions \ref{cond:independent}-\ref{cond:identification}	hold and $\gamma \geq 3 \gamma^* \vee 1$. Let $\widehat{\tau}_{\min} > 0.618$, $\eta_{\max} \ge 1$, and $n_*^{\widehat{\tau}, |\eta|} \geq n_*^{\widehat{\tau}^2, |\eta|^2}$. Then there exists some universal constants $c_1,c_2$ that depends on $(\kappa_U, \sigma_x, \sigma_{\varepsilon}, \sigma_z)$ such that for any $t>0$, if $n_{\min}\ge (p+\log (4|\mathcal{E}|)+t),\bar{n} \ge c_1\eta^2_{\max}  (\gamma/\kappa_L) (p+\log (4|\mathcal{E}|)+t)\{\mathbf{s}_{+}^{-0.5}+\mathbf{s}_{+}^{-1}+(\gamma \kappa_L \mathbf{s}_{-})^{-0.5}\}, n_*^{\widehat{\tau}^2} \ge c_2 \eta^2_{\max} (\gamma/\kappa_L)^2 (p+t) \{\mathbf{s}_{+}^{-1}+ (\gamma \kappa_L \mathbf{s}_{-})^{-1}+1\}$, then the estimator $\widehat{\boldsymbol{\beta}}_{\mathrm{Adj}}$ minimizing (\ref{eq:objective_adjust_whole}) satisfies
	\begin{equation}
		\mathbb{P}\left[S^* \subseteq \operatorname{supp}(\widehat{\boldsymbol{\beta}}_{\mathrm{Adj}}) \subseteq\left(G_{\boldsymbol{\omega}}\right)^c\right] \geq 1-37 e^{-t}.
		\label{eq:low_dim_vsc_property}
	\end{equation}
	In the case where $\widehat{\tau}_{\min} > 0.618$ and $\eta_{\max} <  1$, there exists some universal constants $c_1,c_2$ that depends on $(\kappa_U, \sigma_x, \sigma_{\varepsilon}, \sigma_z)$ such that for any $t>0$, if $n_{\min}\ge (p+\log (4|\mathcal{E}|)+t),\bar{n} \ge c_1  (\gamma/\kappa_L) (p+\log (4|\mathcal{E}|)+t)\{\mathbf{s}_{+}^{-0.5}+\mathbf{s}_{+}^{-1}+(\gamma \kappa_L \mathbf{s}_{-})^{-0.5}\}, n_*^{\widehat{\tau}} \ge c_2  (\gamma/\kappa_L)^2 (p+t) \{\mathbf{s}_{+}^{-1}+ (\gamma \kappa_L \mathbf{s}_{-})^{-1}+1\}$, then estimator $\widehat{\boldsymbol{\beta}}_{\mathrm{Adj}}$ satisfies (\ref{eq:low_dim_vsc_property}).
\end{theorem}

\newpage

\subsection{Non-asymptotic $\ell_2$ Error Bound}\label{subsection:nonasy_l2_error}

\begin{theorem}[Non-asymptotic $\ell_2$ Error Bound]\label{theorem:nonasy_l2_error}
	Assume Conditions \ref{cond:independent}-\ref{cond:identification}	hold are satisfied and $\gamma \geq 3 \gamma^* \vee 1$. Let $\widehat{\tau}_{\min} > 0.618$, $\eta_{\max} \ge 1$, and $n_*^{\widehat{\tau}, |\eta|} \geq n_*^{\widehat{\tau}^2, |\eta|^2}$. There exists some universal constants $c_1-c_3$ that depends on $(\kappa_U, \sigma_x)$ such that for any $t>0$, if $n_{\min } \geq (p+\log (4|\mathcal{E}|)+t), N_* \geq c_1(\gamma/\kappa_L)^2 (p+t)$, then $\widehat{\boldsymbol{\beta}}_{\mathrm{Adj}}$ minimizing objective $\widehat{\mathrm{Q}}_{\mathrm{Adj}}(\boldsymbol{\beta} ; \gamma, \boldsymbol{\omega})$ in (\ref{eq:objective_adjust_whole}) satisfies 
\begin{equation}\label{eq:large_eta_non-asymptotic_l2_error_bound_rate_1}
	\frac{\|\widehat{\boldsymbol{\beta}}_{\mathrm{Adj}}-\boldsymbol{\beta}^*\|_2}{\sigma_{\varepsilon}^2\sigma_z^2(\gamma/\kappa_L)}\le c_2\left(\sqrt{\frac{(p+t)\eta_{\max}^2}{n_*^{\hat\tau^2}}}+\frac{(p+\log(2|\mathcal{E}|)+t)\eta_{\max}^2}{\bar{n}}\right) + c_3 \frac{\sqrt{|S^*|}(\log(4|\mathcal{E}||S^*|)+t)\eta_{\max}^2}{(\min _{j \in S^*}\left|\beta_j^*\right|)\bar{n}}
\end{equation}
	with probability at least $1-37e^{-t}$. Moreover, when the additional conditions in Theorem \ref{theorem:nonasy_vsc} hold, then 
\begin{equation}\label{eq:large_eta_non-asymptotic_l2_error_bound_rate_2}
	\frac{\|\widehat{\boldsymbol{\beta}}_{\mathrm{Adj}}-\boldsymbol{\beta}^*\|_2}{\sigma_{\varepsilon}\sigma_z(\gamma/\kappa_L)}\le c_2\left(\sqrt{\frac{(p_0+t)\eta_{\max}^2}{n_*^{\hat\tau^2}}}+\frac{(p_0+\log(2|\mathcal{E}|)+t)\eta_{\max}^2}{\bar{n}}\right) \quad \text{ with }\quad p_0=\left|\left(G_{\boldsymbol{\omega}}\right)^c\right|
\end{equation}
	occurs with probability at least $1-74e^{-t}$. In the case where $\widehat{\tau}_{\min} > 0.618$ and $\eta_{\max} < 1$, we have $\widehat{\boldsymbol{\beta}}_{\mathrm{Adj}}$ satisfies 
\begin{equation}\label{eq:small_eta_non-asymptotic_l2_error_bound_rate_1}
		\frac{\|\widehat{\boldsymbol{\beta}}_{\mathrm{Adj}}-\boldsymbol{\beta}^*\|_2}{\sigma_{\varepsilon}^2\sigma_z^2(\gamma/\kappa_L)}\le c_2\left(\sqrt{\frac{p+t}{n_*^{\hat\tau}}}+\frac{p+\log(2|\mathcal{E}|)+t}{\bar{n}}\right) + c_3 \frac{\sqrt{|S^*|}(\log(4|\mathcal{E}||S^*|)+t)}{(\min _{j \in S^*}\left|\beta_j^*\right|)\bar{n}}.
\end{equation}
Furthermore, when the additional conditions in Theorem \ref{theorem:nonasy_vsc} hold, then 
\begin{equation}\label{eq:small_eta_non-asymptotic_l2_error_bound_rate_2}
		\frac{\|\widehat{\boldsymbol{\beta}}_{\mathrm{Adj}}-\boldsymbol{\beta}^*\|_2}{\sigma_{\varepsilon}\sigma_z(\gamma/\kappa_L)}\le c_2\left(\sqrt{\frac{p_0+t}{n_*^{\hat\tau}}}+\frac{p_0+\log(2|\mathcal{E}|)+t}{\bar{n}}\right).
\end{equation}
\end{theorem}

When $\eta_{\max} \ge 1$, we require $\eta_{\max }=o(\sqrt{\bar{n} \wedge n_*^{\widehat{\tau}^2}})$ to ensure the convergence of $\widehat{\boldsymbol{\beta}}_{\mathrm{Adj}}$. This condition is particularly relevant when the imputation model is estimated rather than treated as a fixed deterministic model, as it restricts the imputation bias from growing faster than the square root of $\bar{n} \wedge n_*^{\widehat{\tau}^2}$.  If the imputation model is fixed and deterministic, $\eta_{\max }$ remains constant and does not scale with the sample size, making this requirement automatically satisfied.

When $\eta_{\max} < 1$, recall that $n_*^{\hat{\tau}} =\min _{e \in \mathcal{E}} ({{n^{(e)}}/{m^{(e)}}} )({N^{(e)}}/{\omega^{(e)}})$. If $ {{n^{(e)}}/{m^{(e)}}}$ is small, then $n_*^{\hat{\tau}} $ is small, leading to a larger upper bound. When handling missing data, using only gold-standard observations is equivalent to replacing $ {{n^{(e)}}/{m^{(e)}}}$ with ${{n^{(e)}}/{N^{(e)}}}$  in the expression $n_*^{\hat{\tau}}$. Since ${{n^{(e)}}/{m^{(e)}}}$ is always lower bounded by ${{n^{(e)}}/{N^{(e)}}}$, incorporating unlabeled data improves the tightness of the upper bound.

We compare $\eta_{\max}$ with $1$ for simplicity in theoretical analysis. Alternatively, we could compare it with a constant $c$ that depends on $\kappa_L, \kappa_U, \sigma_x, \sigma_{\varepsilon}, \sigma_z, \gamma, \mathbf{s}_{+}, \mathbf{s}_{-}$. However, since these are population parameters that are not observed in practice, such an analysis provides little practical benefit. To keep the discussion clear, we distinguish between $\eta_{\max}\ge1$ and $\eta_{\max}<1$.

Overall, the key implication of this result is twofold: when the imputation model is estimated, hence random, the imputation bias $\eta_{\max}$ must not grow faster than $\sqrt{\bar{n} \wedge n_*^{\widehat{\tau}^2}}$; when the imputation model is deterministic, $\eta_{\max}$ should remain bounded by a constant independent of the sample size, making it a reliable choice that improves performance over the naive estimator, which relies solely on gold-standard data. 


\section{Simulation Study}\label{section:simulation}

In this section, we evaluate the empirical performance of the proposed IAEI estimator under scenarios involving missing outcomes across different environments. Specifically, we aim to examine the robustness of the estimator to spurious relationships and its ability to recover stable causal effects in increasingly complex data-generating mechanisms. To this end, we compare the proposed estimator with alternative methods under four structural equation models (SEMs) designed to mimic varying levels of nonlinearity and spurious relationships.

We begin with a baseline SEM, referred to as Model $0$, as described in \cite{fan2024environment}. In this setting, the spurious relationships are nearly linear, and the conditional expectation of the outcome $y$ given all covariates to be well-approximated by a linear model in both environments. To better reflect practical scenarios involving more complex relationships, we progressively introduce additional nonlinearities. While incorporating linear featurs is sufficient to capture the spurious variables in those settings, we later also present results obtained by modifying the objective to include a nonlinear component in the regularizer, which enhances the model's ability to capture more complex spurious relationships. 

\begin{itemize}
	\item Model $1$: Builds on Model $0$ by introducing nonlinear terms for covariates $x_7$ and $x_8$, increasing the complexity of spurious relationships. 
	\item Model $2$: Extends Model $1$ by incorporating nonlinear term $(x_{12}^{(2)})^2-1 + \sin(x_{12}^{(2)})$ into $y^{(2)}$ for environment $2$, making the conditional expectation of $y$ given all covariates less likely to follow a linear relationship.
	\item Model $3$: Further builds on Model $1$ by increasing the complexity of spurious relationships, making it more difficult to identify the invariant structure using linear features alone.
\end{itemize}

After generating data under these SEMs, we simulate scenarios with randomly missing outcomes within each environment. The missing ratios are set to 30\%, 50\%, and 70\%, representing practical constraints in data availability. To handle the missing outcomes, we explore three distinct approaches to impute using different models—Linear Regression, Random Forest, and XGBoost.

We further investigate imputation strategies with varying levels of alignment to the true environment distributions, categorized as precise imputation, bias imputation, and hbias imputation. In precise imputation, separate models are trained on data specific to each environment, ensuring strong alignment with their unique distributions. Bias imputation involves training a single model on pooled data from both environments, followed by introducing small shifts in the covariates, resulting in moderate bias. The hbias imputation model extends this approach by introducing slightly larger shifts and increased variance in the covariates, leading to a marginally higher level of bias compared to the bias imputation model.

 We will show that the proposed IAEI estimator distinguishes itself by leveraging all available observations and addressing potential imputation bias to achieve unbiased estimation. The evaluation includes the IAEI estimator alongside its oracle counterpart IAEI-oracle, which assumes access to the true labels, providing a benchmark for its efficiency compared to the ideal scenario of complete data availability. Note that this oracle estimator is equivalent to the EILLS estimator constructed under the assumption that all data points have observed labels.

To demonstrate the advantages of the IAEI estimator specifically in the context of missing data, which is the central focus of this paper, we compare its performance to three straightforward yet naive adaptations of the EILLS framework. These methods, while intuitive, either fail to fully utilize the available information or introduce bias in the resulting estimates:
\begin{itemize}
	\item EILLS-observe: EILLS with observed data only. This method excludes observations without labels, using only the subset of data with observed outcomes.
	\item EILLS-impute: This method replaces all labels, including observed ones, with imputed values before fitting the model, relying entirely on imputation for the labels. 
	\item EILLS-mixed: This method combines observed and imputed labels, using observed labels where available and imputed values for the rest.
\end{itemize}

All the previously mentioned estimators rely on a penalty function that captures linear relationships, specifically leveraging the correlation between the error and $x_j$. However, as the complexity and nonlinearity increase in Models $1-3$, it becomes beneficial to enhance the penalty function by incorporating a nonlinear component, $x_j^2$, alongside the linear terms, as suggested in \cite{fan2024environment}. To ensure a comprehensive evaluation, we compute each estimator using both the basic penalty and its enhanced counterpart, explicitly labeling methods that use the enhanced penalty with the symbol $\dagger$.

The enhanced penalty function extends the standard approach to better account for the nonlinearities in the data-generating mechanisms. For the EILLS-observe, EILLS-impute, and EILLS-mixed estimators, the empirical enhanced penalty is defined as:
\begin{equation}\label{eq:enhanced_penalty_fan}
	\mathrm{J}^{\dagger}(\boldsymbol{\beta} ; \boldsymbol{\omega})= \sum_{j=1}^p \mathds{1}\left\{\beta_j \neq 0\right\} \sum_{e \in \mathcal{E}} \omega^{(e)}\left\{\left|\widehat{\mathbb{E}}_{n^{(e)}}\left[x_j^{(e)}(y^{(e)}-\boldsymbol{\beta}^{\top} \boldsymbol{x}^{(e)})\right]\right|^2 + \left|\widehat{\mathbb{E}}_{n^{(e)}}\left[(x_j^{(e)})^2(y^{(e)}-\boldsymbol{\beta}^{\top} \boldsymbol{x}^{(e)})\right]\right|^2\right\},
\end{equation}
For the IAEI estimator, which leverages both observed and imputed data, the empirical enhanced penalty function is adjusted to account for the imputation process. The adjusted empirical enhanced penalty is given by:
\begin{align}
	\mathrm{J}^{\dagger}_{\mathrm{Adj}}(\boldsymbol{\beta} ; \boldsymbol{\omega}) &= \sum_{j=1}^p \mathds{1}\left\{\beta_j \neq 0\right\} \sum_{e \in \mathcal{E}} \omega^{(e)} \left\{ \left| \widehat{\mathbb{E}}_{N^{(e)}}\left[x_j^{(e)}\left(\widehat{h}^{(e)}(\boldsymbol{x}^{(e)})-\boldsymbol{\beta}^{\top} \boldsymbol{x}^{(e)}\right)\right]  + \widehat{\mathbb{E}}_{n^{(e)}}\left[x_j^{(e)}\left(y^{(e)}-\widehat{h}^{(e)}(\boldsymbol{x}^{(e)})\right)\right] \right|^2\right.\notag \\
	&\left.\quad \quad\quad\quad\quad\quad\quad\quad \quad\quad\quad\quad + \left| \widehat{\mathbb{E}}_{N^{(e)}}\left[(x_j^{(e)})^2\left(\widehat{h}^{(e)}(\boldsymbol{x}^{(e)})-\boldsymbol{\beta}^{\top} \boldsymbol{x}^{(e)}\right)\right]  + \widehat{\mathbb{E}}_{n^{(e)}}\left[(x_j^{(e)})^2\left(y^{(e)}-\widehat{h}^{(e)}(\boldsymbol{x}^{(e)})\right)\right] \right|^2\right\} \label{eq:enhanced_penalty_ieai}
\end{align}
These enhanced penalties ensure that both the standard and imputation-adjusted methods are evaluated comprehensively in the simulation study, particularly under the increasing complexity of the data-generating mechanisms in Models $1-3$. Consequently, we compare ten different estimators: IAEI, IAEI-oracle, EILLS-observe, EILLS-impute, EILLS-mix, IAEI$^{\dagger}$, IAEI-oracle$^{\dagger}$, EILLS-observe$^{\dagger}$, EILLS-impute$^{\dagger}$, and EILLS-mix$^{\dagger}$. 

\subsection{Data Generating Process}\label{simulation:DGP}

To provide a precise and reproducible framework for the simulation study, we now formally define the mathematical structure of the data-generating processes (DGPs) underlying each of the SEMs described above. These expressions capture the relationships between covariates, outcomes, and spurious variables, illustrating how increasing complexity and nonlinearity are introduced across models. In the following SEMs, let $u_1^{(e)}, \ldots, u_{13}^{(e)} \sim \mathcal{N}\left(\mathbf{0}, I_{13 \times 13}\right)$ for all the $e=1, 2$. 
\subsubsection*{Model 0}
\begin{multicols}{2}
	\normalsize
	\noindent
	\begin{align*}
		&\text{Environment } e=1: \\
		x_1^{(1)} &\leftarrow u_1^{(1)} \\
		x_4^{(1)} &\leftarrow u_4^{(1)}\\
		x_2^{(1)} & \leftarrow \sin (x_4^{(1)})+u_2^{(1)} \\
		x_3^{(1)} & \leftarrow \cos (x_4^{(1)})+u_3^{(1)} \\
		x_5^{(1)} & \leftarrow \sin (x_3^{(1)}+u_5^{(1)}) \\
		x_{10}^{(1)} & \leftarrow 2.5 x_1^{(1)}+1.5 x_2^{(1)}+u_{10}^{(1)} \\
		y^{(1)} & \leftarrow 3 x_1^{(1)}+2 x_2^{(1)}-0.5 x_3^{(1)}+u_{13}^{(1)} \\
		x_6^{(1)} & \leftarrow 0.8 y^{(1)} u_6^{(1)} \\
	\end{align*}
	
	\columnbreak
	
	\noindent
	\begin{align*}
		&\text{Environment } e=2: \\
		x_1^{(2)} &\leftarrow u_1^{(2)} \\
		x_4^{(2)} &\leftarrow(u_4^{(2)})^2-1\\
		x_2^{(2)} & \leftarrow \sin (x_4^{(2)})+u_2^{(2)} \\
		x_3^{(2)} & \leftarrow \cos (x_4^{(2)})+u_3^{(2)} \\
		x_5^{(2)} & \leftarrow \sin (x_3^{(2)}+u_5^{(2)}) \\
		x_{10}^{(2)} & \leftarrow 2.5 x_1^{(2)}+1.5 x_2^{(2)}+u_{10}^{(2)} \\
		y^{(2)} & \leftarrow 3 x_1^{(2)}+2 x_2^{(2)}-0.5 x_3^{(2)}+u_{13}^{(2)} \\
		x_6^{(2)} & \leftarrow 0.8 y^{(2)} u_6^{(2)} \\
	\end{align*}
\end{multicols}
\begin{multicols}{2}
	\normalsize
	\noindent
	\begin{align*}
		&\text{Environment } e=1: \\
		x_7^{(1)} & \leftarrow 0.5 x_3^{(1)}+y^{(1)}+u_7^{(1)} \\
		x_8^{(1)} & \leftarrow 0.5 x_7^{(1)}-y^{(1)}+x_{10}^{(1)}+u_8^{(1)} \\
		x_9^{(1)} & \leftarrow \tanh (x_7^{(1)})+0.1 \cos (x_8^{(1)})+u_9^{(1)} \\
		x_{11}^{(1)} & \leftarrow 0.4(x_7^{(1)}+x_8^{(1)}) * u_{11}^{(1)} \\
		x_{12}^{(1)} & \leftarrow u_{12}^{(1)}
	\end{align*}
	
	\columnbreak
	
	\noindent
	\begin{align*}
		&\text{Environment } e=2: \\
		x_7^{(2)} &\leftarrow 4 x_3^{(2)}+\tanh (y^{(2)})+u_7^{(2)} \\
		x_8^{(2)} & \leftarrow 0.5 x_7^{(2)}-y^{(2)}+x_{10}^{(2)}+u_8^{(2)} \\
		x_9^{(2)} & \leftarrow \tanh (x_7^{(2)})+0.1 \cos (x_8^{(2)})+u_9^{(2)} \\
		x_{11}^{(2)} & \leftarrow 0.4(x_7^{(2)}+x_8^{(2)}) * u_{11}^{(2)} \\
		x_{12}^{(2)} & \leftarrow u_{12}^{(2)}
	\end{align*}
\end{multicols}

\subsubsection*{Model 1}

\begin{multicols}{2}
	\normalsize
	\noindent
	\begin{align*}
		&\text{Environment } e=1: \\
		x_1^{(1)} &\leftarrow u_1^{(1)} \\
		x_4^{(1)} &\leftarrow u_4^{(1)}\\
		x_2^{(1)} & \leftarrow \sin (x_4^{(1)})+u_2^{(1)} \\
		x_3^{(1)} & \leftarrow \cos (x_4^{(1)})+u_3^{(1)} \\
		x_5^{(1)} & \leftarrow \sin (x_3^{(1)}+u_5^{(1)}) \\
		x_{10}^{(1)} & \leftarrow 2.5 x_1^{(1)}+1.5 x_2^{(1)}+u_{10}^{(1)} \\
		y^{(1)} & \leftarrow 3 x_1^{(1)}+2 x_2^{(1)}-0.5 x_3^{(1)}+u_{13}^{(1)} \\
		x_6^{(1)} & \leftarrow 0.8 y^{(1)} u_6^{(1)} \\
		x_7^{(1)} & \leftarrow 0.5\sin((x_3^{(1)})^2) + 8 (y^{(1)})^3 + u_7^{(1)}\\
		x_8^{(1)} & \leftarrow \log(| x_7^{(1)} y^{(1)} + 1|)+ 5\sin(y^{(1)}) + u_8^{(1)}\\
		x_9^{(1)} & \leftarrow \tanh (x_7^{(1)})+0.1 \cos (x_8^{(1)})+u_9^{(1)} \\
		x_{11}^{(1)} & \leftarrow 0.4(x_7^{(1)}+x_8^{(1)}) * u_{11}^{(1)} \\
		x_{12}^{(1)} & \leftarrow u_{12}^{(1)}
	\end{align*}
	
	\columnbreak
	
	\noindent
	\begin{align*}
		&\text{Environment } e=2: \\
		x_1^{(2)} &\leftarrow u_1^{(2)} \\
		x_4^{(2)} &\leftarrow(u_4^{(2)})^2-1\\
		x_2^{(2)} & \leftarrow \sin (x_4^{(2)})+u_2^{(2)} \\
		x_3^{(2)} & \leftarrow \cos (x_4^{(2)})+u_3^{(2)} \\
		x_5^{(2)} & \leftarrow \sin (x_3^{(2)}+u_5^{(2)}) \\
		x_{10}^{(2)} & \leftarrow 2.5 x_1^{(2)}+1.5 x_2^{(2)}+u_{10}^{(2)} \\
		y^{(2)} & \leftarrow 3 x_1^{(2)}+2 x_2^{(2)}-0.5 x_3^{(2)}+u_{13}^{(2)} \\
		x_6^{(2)} & \leftarrow 0.8 y^{(2)} u_6^{(2)} \\
		x_7^{(2)} &\leftarrow \tanh(x_3^{(2)}) + 4(|y^{(2)}|)^{1/2}+u_7^{(2)} \\
		x_8^{(2)} & \leftarrow 0.5 (x_7^{(2)})^2 + (y^{(2)})^3 + \cos(y^{(2)})+u_8^{(2)} \\
		x_9^{(2)} & \leftarrow \tanh (x_7^{(2)})+0.1 \cos (x_8^{(2)})+u_9^{(2)} \\
		x_{11}^{(2)} & \leftarrow 0.4(x_7^{(2)}+x_8^{(2)}) * u_{11}^{(2)} \\
		x_{12}^{(2)} & \leftarrow u_{12}^{(2)}
	\end{align*}
\end{multicols}

\newpage

\subsubsection*{Model 2}

\begin{multicols}{2}
	\normalsize
	\noindent
	\begin{align*}
		&\text{Environment } e=1: \\
		x_1^{(1)} &\leftarrow u_1^{(1)} \\
		x_4^{(1)} &\leftarrow u_4^{(1)}\\
		x_2^{(1)} & \leftarrow \sin (x_4^{(1)})+u_2^{(1)} \\
		x_3^{(1)} & \leftarrow \cos (x_4^{(1)})+u_3^{(1)} \\
		x_5^{(1)} & \leftarrow \sin (x_3^{(1)}+u_5^{(1)}) \\
		x_{10}^{(1)} & \leftarrow 2.5 x_1^{(1)}+1.5 x_2^{(1)}+u_{10}^{(1)} \\
		y^{(1)} & \leftarrow 3 x_1^{(1)}+2 x_2^{(1)}-0.5 x_3^{(1)}+u_{13}^{(1)} \\
		x_6^{(1)} & \leftarrow 0.8 y^{(1)} u_6^{(1)} \\
		x_7^{(1)} & \leftarrow 0.5\sin((x_3^{(1)})^2) + 8 (y^{(1)})^3 + u_7^{(1)}\\
		x_8^{(1)} & \leftarrow \log(| x_7^{(1)} y^{(1)} + 1|)+ 5\sin(y^{(1)}) + u_8^{(1)}\\
		x_9^{(1)} & \leftarrow \tanh (x_7^{(1)})+0.1 \cos (x_8^{(1)})+u_9^{(1)} \\
		x_{11}^{(1)} & \leftarrow 0.4(x_7^{(1)}+x_8^{(1)}) * u_{11}^{(1)} \\
		x_{12}^{(1)} & \leftarrow u_{12}^{(1)}
	\end{align*}
	
	\columnbreak
	
	\noindent
	\begin{align*}
		&\text{Environment } e=2: \\
		x_1^{(2)} &\leftarrow u_1^{(2)} \\
		x_4^{(2)} &\leftarrow(u_4^{(2)})^2-1\\
		x_2^{(2)} & \leftarrow \sin (x_4^{(2)})+u_2^{(2)} \\
		x_3^{(2)} & \leftarrow \cos (x_4^{(2)})+u_3^{(2)} \\
		x_5^{(2)} & \leftarrow \sin (x_3^{(2)}+u_5^{(2)}) \\
		x_{10}^{(2)} & \leftarrow 2.5 x_1^{(2)}+1.5 x_2^{(2)}+u_{10}^{(2)} \\
		y^{(2)} & \leftarrow 3 x_1^{(2)}+2 x_2^{(2)}-0.5 x_3^{(2)}+ \sin(x_{12}^{(2)}) + \{(x_{12}^{(2)})^2 -1\} + u_{13}^{(2)} \\
		x_6^{(2)} & \leftarrow 0.8 y^{(2)} u_6^{(2)} \\
		x_7^{(2)} &\leftarrow \tanh(x_3^{(2)}) + 4(|y^{(2)}|)^{1/2}+u_7^{(2)} \\
		x_8^{(2)} & \leftarrow 0.5 (x_7^{(2)})^2 + (y^{(2)})^3 + \cos(y^{(2)})+u_8^{(2)} \\
		x_9^{(2)} & \leftarrow \tanh (x_7^{(2)})+0.1 \cos (x_8^{(2)})+u_9^{(2)} \\
		x_{11}^{(2)} & \leftarrow 0.4(x_7^{(2)}+x_8^{(2)}) * u_{11}^{(2)} \\
		x_{12}^{(2)} & \leftarrow u_{12}^{(2)}
	\end{align*}
\end{multicols}

\subsubsection*{Model 3}

\begin{multicols}{2}
	\normalsize
	\noindent
	\begin{align*}
		&\text{Environment } e=1: \\
		x_1^{(1)} &\leftarrow u_1^{(1)} \\
		x_4^{(1)} &\leftarrow u_4^{(1)}\\
		x_2^{(1)} & \leftarrow \sin (x_4^{(1)})+u_2^{(1)} \\
		x_3^{(1)} & \leftarrow \cos (x_4^{(1)})+u_3^{(1)} \\
		x_5^{(1)} & \leftarrow \sin (x_3^{(1)}+u_5^{(1)}) \\
		x_{10}^{(1)} & \leftarrow 2.5 x_1^{(1)}+1.5 x_2^{(1)}+u_{10}^{(1)} \\
		y^{(1)} & \leftarrow 3 x_1^{(1)}+2 x_2^{(1)}-0.5 x_3^{(1)}+u_{13}^{(1)} \\
		x_6^{(1)} & \leftarrow 0.8 y^{(1)} u_6^{(1)} \\
		x_7^{(1)} & \leftarrow 0.5\sin((x_3^{(1)})^2) + 8 (y^{(1)})^3 + u_7^{(1)}\\
		x_8^{(1)} & \leftarrow \log(| x_7^{(1)} y^{(1)} + 1|)+ 5\sin(y^{(1)}) + u_8^{(1)}\\
		x_9^{(1)} & \leftarrow \tanh (x_7^{(1)})+0.1 \cos (x_8^{(1)})+u_9^{(1)} \\
		x_{11}^{(1)} & \leftarrow 0.4(x_7^{(1)}+x_8^{(1)}) * u_{11}^{(1)} \\
		x_{12}^{(1)} & \leftarrow u_{12}^{(1)}
	\end{align*}
	
	\columnbreak
	
	\noindent
	\begin{align*}
		&\text{Environment } e=2: \\
		x_1^{(2)} &\leftarrow u_1^{(2)} \\
		x_4^{(2)} &\leftarrow(u_4^{(2)})^2-1\\
		x_2^{(2)} & \leftarrow \sin (x_4^{(2)})+u_2^{(2)} \\
		x_3^{(2)} & \leftarrow \cos (x_4^{(2)})+u_3^{(2)} \\
		x_5^{(2)} & \leftarrow \sin (x_3^{(2)}+u_5^{(2)}) \\
		x_{10}^{(2)} & \leftarrow 2.5 x_1^{(2)}+1.5 x_2^{(2)}+u_{10}^{(2)} \\
		y^{(2)} & \leftarrow3 x_1^{(2)}+2 x_2^{(2)}-0.5 x_3^{(2)}+ 0.5x_{12}^{(2)} + u_{13}^{(2)} + v_{13}^{(2)}\\
		x_6^{(2)} & \leftarrow 0.8 y^{(2)} u_6^{(2)} \\
		x_7^{(2)} &\leftarrow \tanh(x_3^{(2)}) + 4(|y^{(2)}|)^{1/2}+u_7^{(2)} \\
		x_8^{(2)} & \leftarrow 0.5 (x_7^{(2)})^2 + (y^{(2)})^3 + \cos(y^{(2)})+u_8^{(2)} \\
		x_9^{(2)} & \leftarrow \tanh (x_7^{(2)})+0.1 \cos (x_8^{(2)})+u_9^{(2)} \\
		x_{11}^{(2)} & \leftarrow 0.4(x_7^{(2)}+x_8^{(2)}) * u_{11}^{(2)} \\
		x_{12}^{(2)} & \leftarrow u_{12}^{(2)},
	\end{align*}
\end{multicols}
where $v_{13}^{(2)} = -0.5(x_{12}^{(2)})^3 + v_{14}^{(2)},\text{ and }  v_{14}^{(2)}\sim  \mathcal{N}(0,1)$.

\subsection{Performance on Variable Selection}\label{simulation:variable_selection}

In this section, we evaluate the ability of the proposed IAEI estimator and other naive estimators to correctly identify important set of variables across the SEMs defined in Section \ref{simulation:DGP}. Accurate variable selection is critical for understanding the causal structure of the data and for ensuring that downstream predictions and inferences are unbiased and interpretable. 

To quantify variable selection performance, we use the false discovery rate (FDR), which quantifies the proportion of irrelevant variables incorrectly identified as relevant. We computed over $500$ iterations, and the reported results represent the average FDR across these repetitions. 

In general, methods with the enhanced penalty consistently outperform those using the original penalty. Under precise imputation, EILLS-impute$^{\dagger}$ and EILLS-mix$^{\dagger}$ achieve performance close to the oracle and outperform IAEI$^{\dagger}$ in Model 0. However, in Models 1 through 3, IAEI$^{\dagger}$ with nonlinear imputation methods like XGBoost or RandomForest closes the gap with EILLS-impute$^{\dagger}$ and EILLS-mix$^{\dagger}$, demonstrating competitive performance. In contrast, IAEI$^{\dagger}$ with linear imputation struggles to perform well, even with precise imputation. Under biased imputation, IAEI, EILLS-observe, and IAEI-oracle converge to zero, while EILLS-impute and EILLS-mix diverge regardless of whether the original or enhanced penalty is applied. Notably, IAEI$^{\dagger}$ demonstrates the best performance across all Models 0-3 and imputation methods, showcasing its robustness and effectiveness.

Here, we present results for both Model 0 and Model 3. Model 0 represents a simple setting with predominantly linear relationships, where IAEI$^{\dagger}$ performs less effectively compared to EILLS-impute$^{\dagger}$ and EILLS-mix$^{\dagger}$ under precise imputation. In contrast, Model 3 introduces significant nonlinear dependencies and spurious relationships, highlighting the advantages of using nonlinear imputation methods in more complex settings. Detailed results for Models 1 and 2, which exhibit intermediate complexity, are provided in the supplementary material.

\begin{figure}[H]
    \centering
    \begin{subfigure}[t]{0.32\textwidth} 
        \centering
        \includegraphics[width=\textwidth]{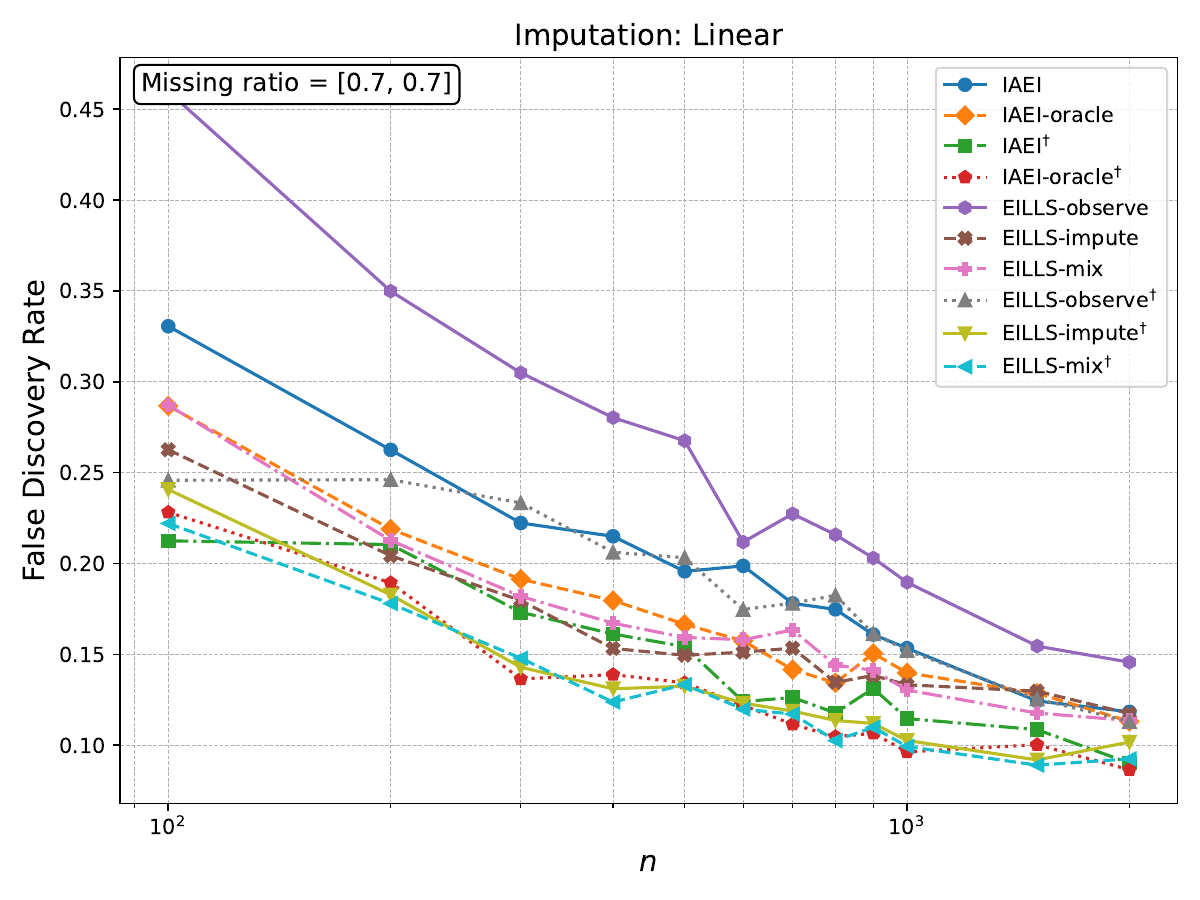} 
        \caption{FDR results under Model 0 with precie Linear imputation.}
        \label{fig:./figures/fig3b_and_fig3b1_and_fig3a1/precise_imputation/model0/fig3a1_precise_model0_model0_0.7_linear_all_fdr.pdf}
    \end{subfigure}
    \hfill 
    \begin{subfigure}[t]{0.32\textwidth} 
        \centering
        \includegraphics[width=\textwidth]{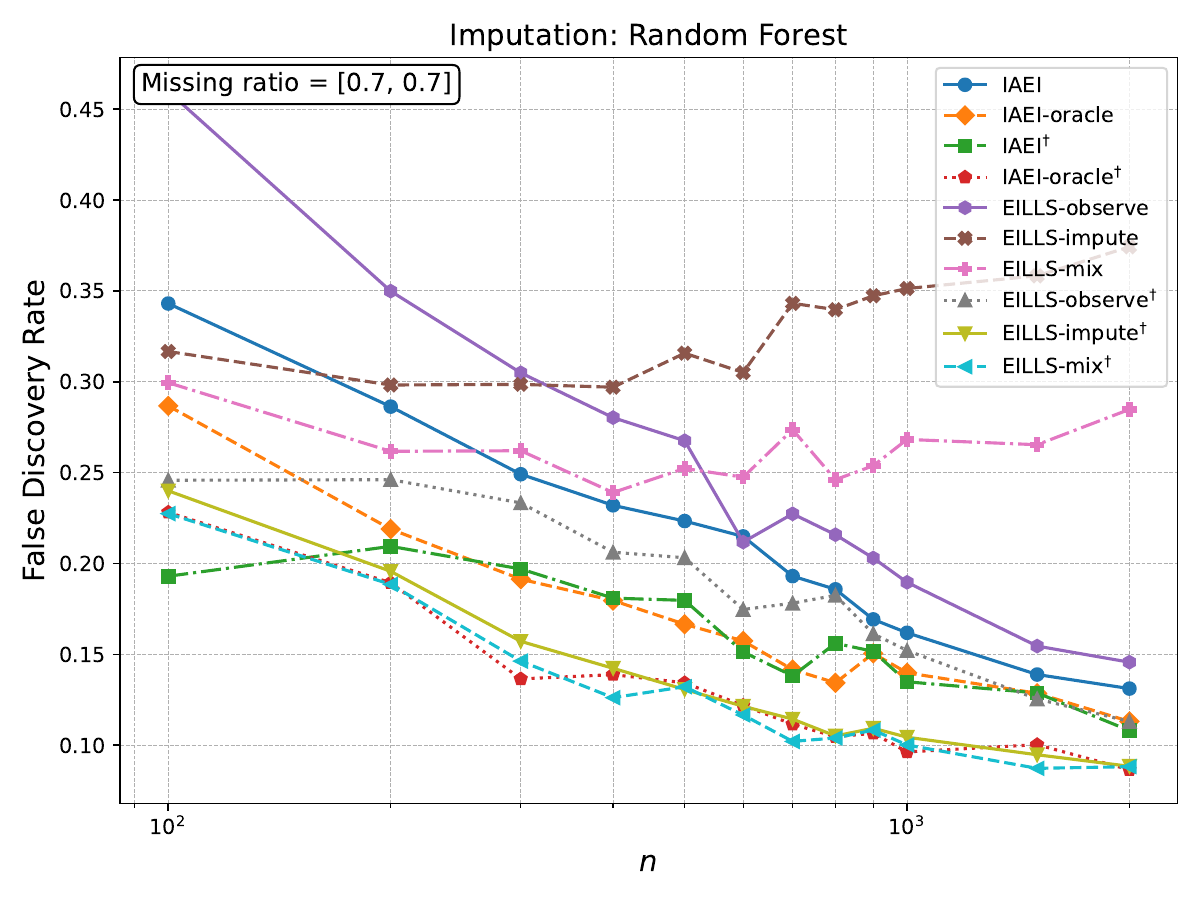} 
        \caption{FDR results under Model 0 with precise RandomForest imputation.}
        \label{fig:./figures/fig3b_and_fig3b1_and_fig3a1/precise_imputation/model0/fig3a1_precise_model0_model0_0.7_rf_all_fdr.pdf}
    \end{subfigure}
	\hfill 
    \begin{subfigure}[t]{0.32\textwidth} 
        \centering
        \includegraphics[width=\textwidth]{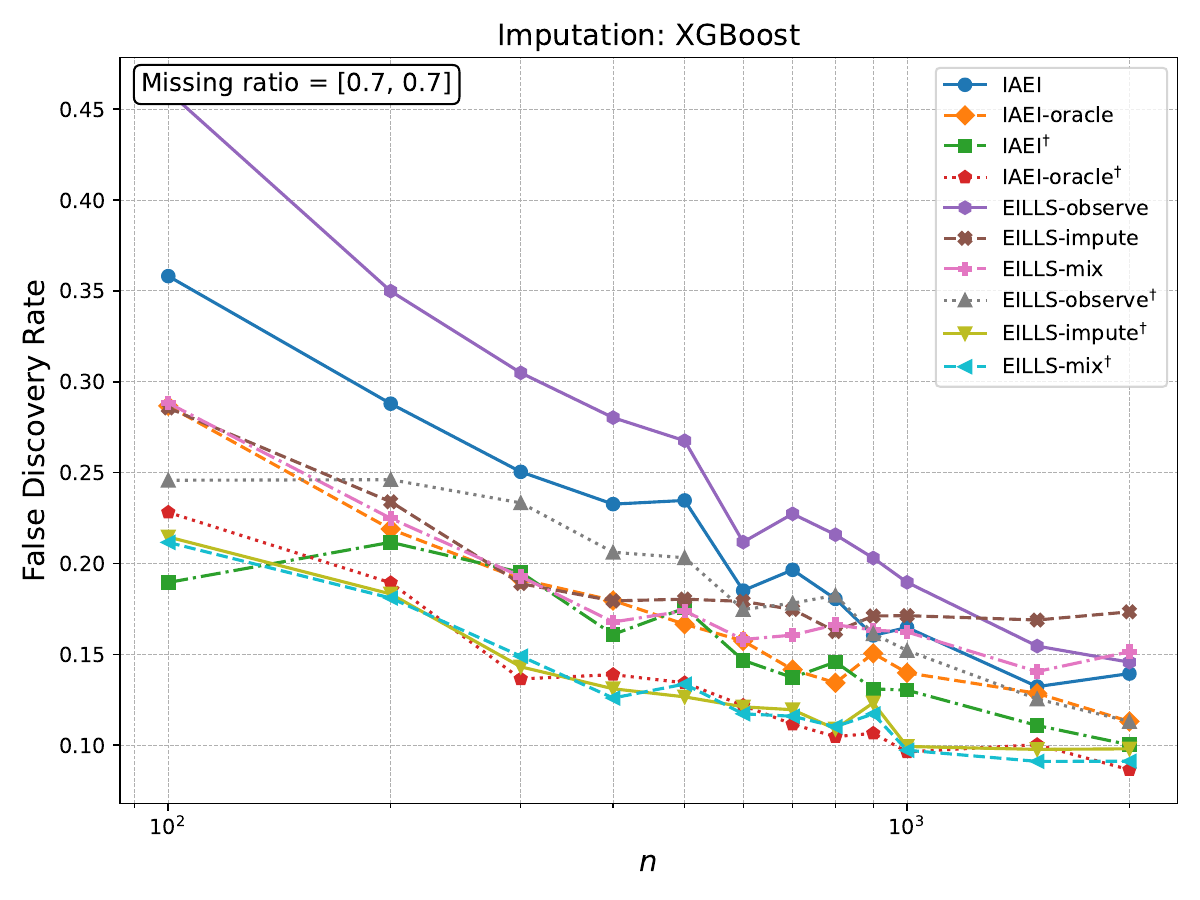} 
        \caption{FDR results under Model 0 with precise XGBoost imputation.}
        \label{fig:./figures/fig3b_and_fig3b1_and_fig3a1/precise_imputation/model0/fig3a1_precise_model0_model0_0.7_xgboost_all_fdr.pdf}
    \end{subfigure}
    \caption{This demonstrates that methods with the enhanced penalty achieve relatively lower FDR. Under precise imputation, EILLS-impute$^{\dagger}$ and EILLS-mix$^{\dagger}$ perform close to the enhanced oracle estimator. In contrast, IAEI$^{\dagger}$ consistently exhibits a noticeable gap from the oracle, regardless of the imputation method employed.}
    \label{fig:./figures/fig3b_and_fig3b1_and_fig3a1/precise_imputation/model0/fig3a1_precise_model0_model0_0.7_linear_all_fdr.pdf./figures/fig3b_and_fig3b1_and_fig3a1/precise_imputation/model0/fig3a1_precise_model0_model0_0.7_rf_all_fdr.pdffig:./figures/fig3b_and_fig3b1_and_fig3a1/precise_imputation/model0/fig3a1_precise_model0_model0_0.7_xgboost_all_fdr.pdf}
\end{figure}

\begin{figure}[H]
    \centering
    \begin{subfigure}[t]{0.32\textwidth} 
        \centering
        \includegraphics[width=\textwidth]{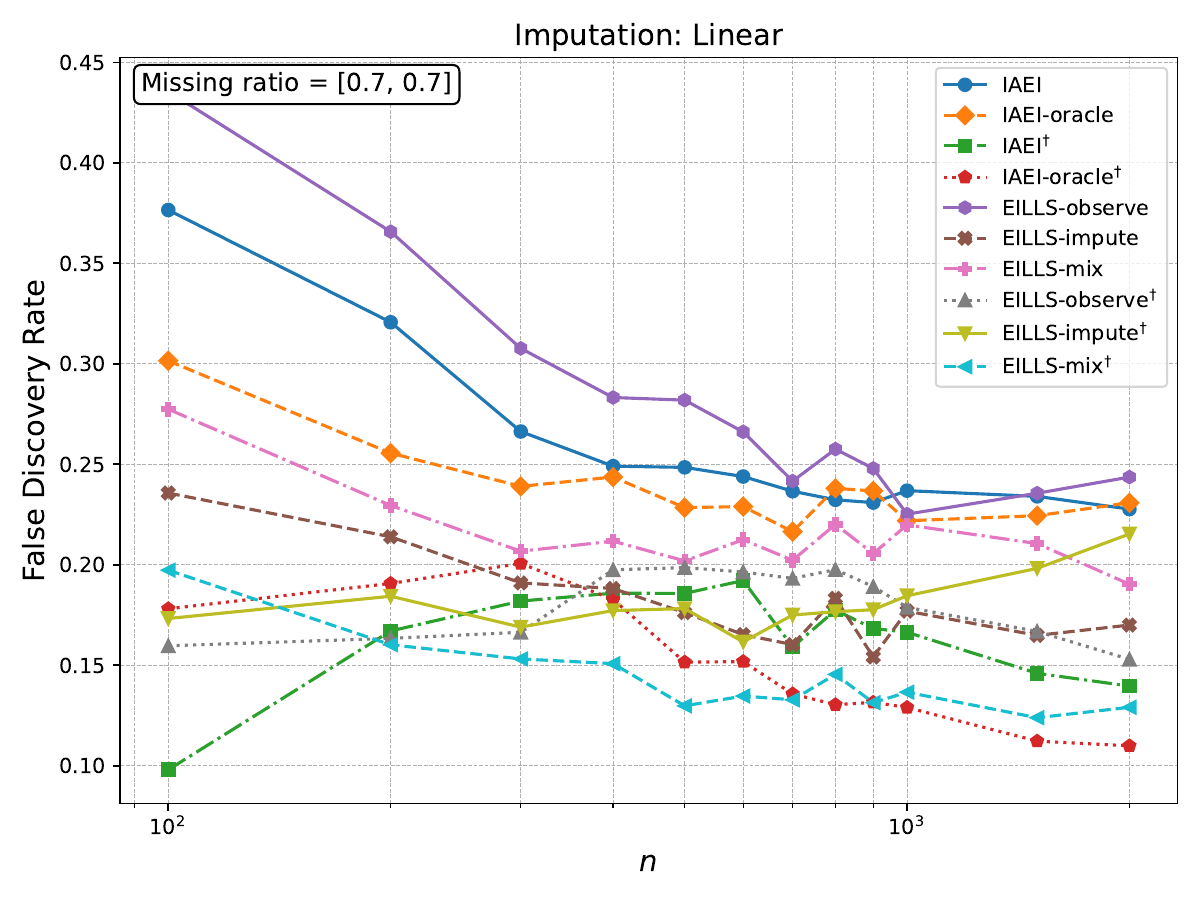} 
        \caption{FDR results under Model 3 with precise Linear imputation.}
        \label{fig:./figures/fig3b_and_fig3b1_and_fig3a1/precise_imputation/model3/fig3a1_precise_model3_model0_0.7_linear_all_fdr.pdf}
    \end{subfigure}
    \hfill 
    \begin{subfigure}[t]{0.32\textwidth} 
        \centering
        \includegraphics[width=\textwidth]{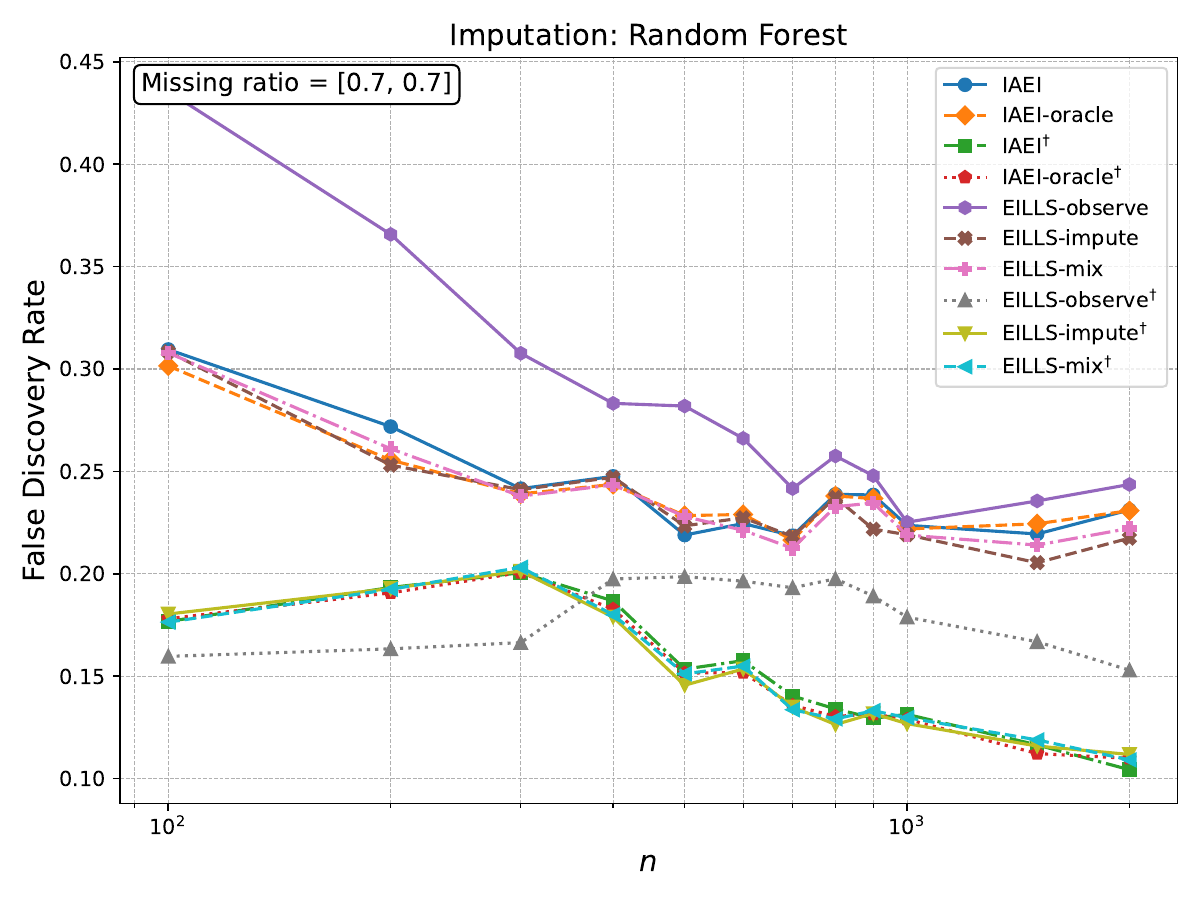} 
        \caption{FDR results under Model 3 with precise RandomForest imputation.}
        \label{fig:./figures/fig3b_and_fig3b1_and_fig3a1/precise_imputation/model3/fig3a1_precise_model3_model0_0.7_rf_all_fdr.pdf}
    \end{subfigure}
	\hfill 
    \begin{subfigure}[t]{0.32\textwidth} 
        \centering
        \includegraphics[width=\textwidth]{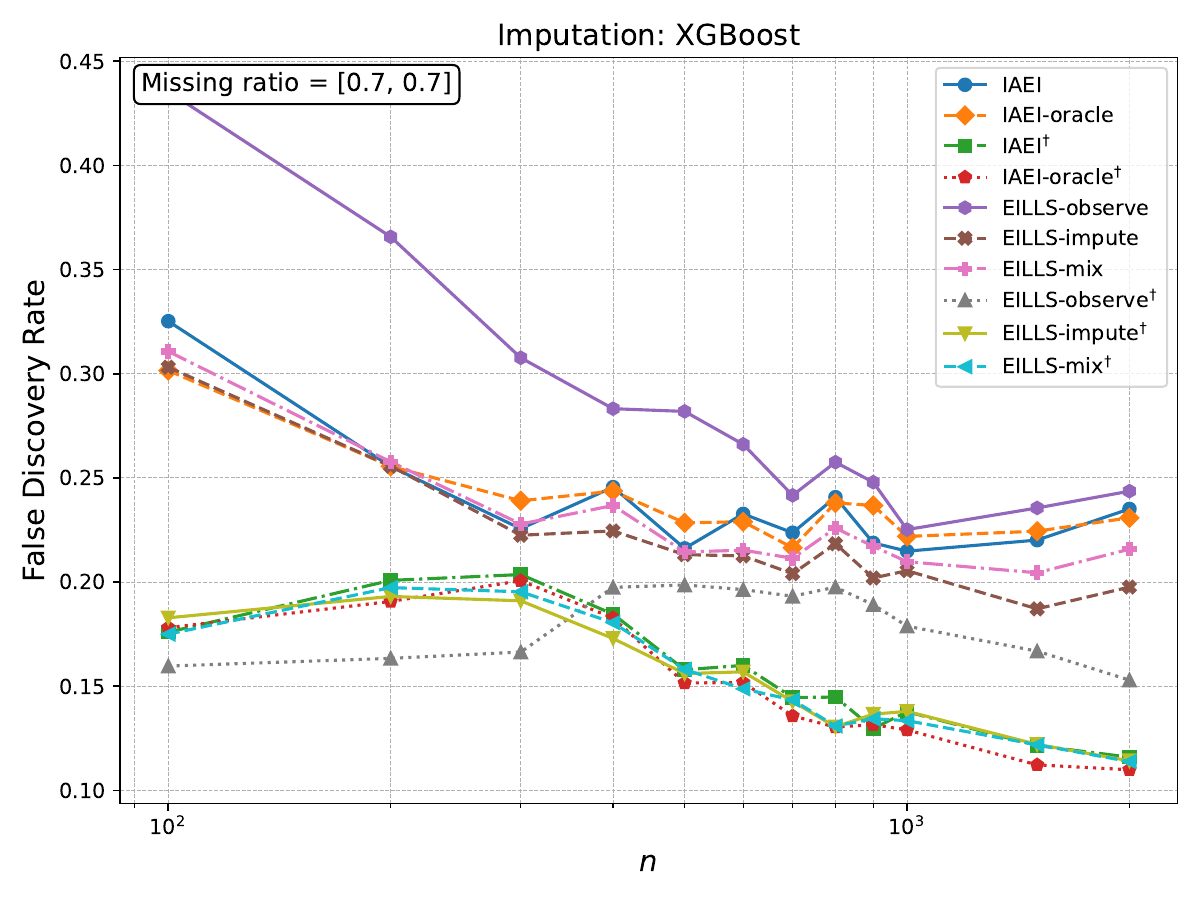} 
        \caption{FDR results under Model 3 with precise XGBoost imputation.}
        \label{fig:./figures/fig3b_and_fig3b1_and_fig3a1/precise_imputation/model3/fig3a1_precise_model3_model0_0.7_xgboost_all_fdr.pdf}
    \end{subfigure}
    \caption{In Model 3, a complex scenario with nonlinear spurious relationships, the advantage of the enhanced penalty becomes more evident. While the results with linear imputation are less clear, the RandomForest and XGBoost imputations highlights the effectiveness of the enhanced penalty in reducing FDR. Notably, EILLS-impute$^{\dagger}$, EILLS-mix$^{\dagger}$, and IAEI$^{\dagger}$ demonstrate performance close to the oracle, underscoring the advantages of combining nonlinear imputation methods with the enhanced penalty.}
    \label{fig:./figures/fig3b_and_fig3b1_and_fig3a1/precise_imputation/model3/fig3a1_precise_model3_model0_0.7_linear_all_fdr.pdf./figures/fig3b_and_fig3b1_and_fig3a1/precise_imputation/model3/fig3a1_precise_model3_model0_0.7_rf_all_fdr.pdf./figures/fig3b_and_fig3b1_and_fig3a1/precise_imputation/model3/fig3a1_precise_model3_model0_0.7_xgboost_all_fdr.pdf}
\end{figure}

\begin{figure}[H]
    \centering
    \begin{subfigure}[t]{0.32\textwidth} 
        \centering
        \includegraphics[width=\textwidth]{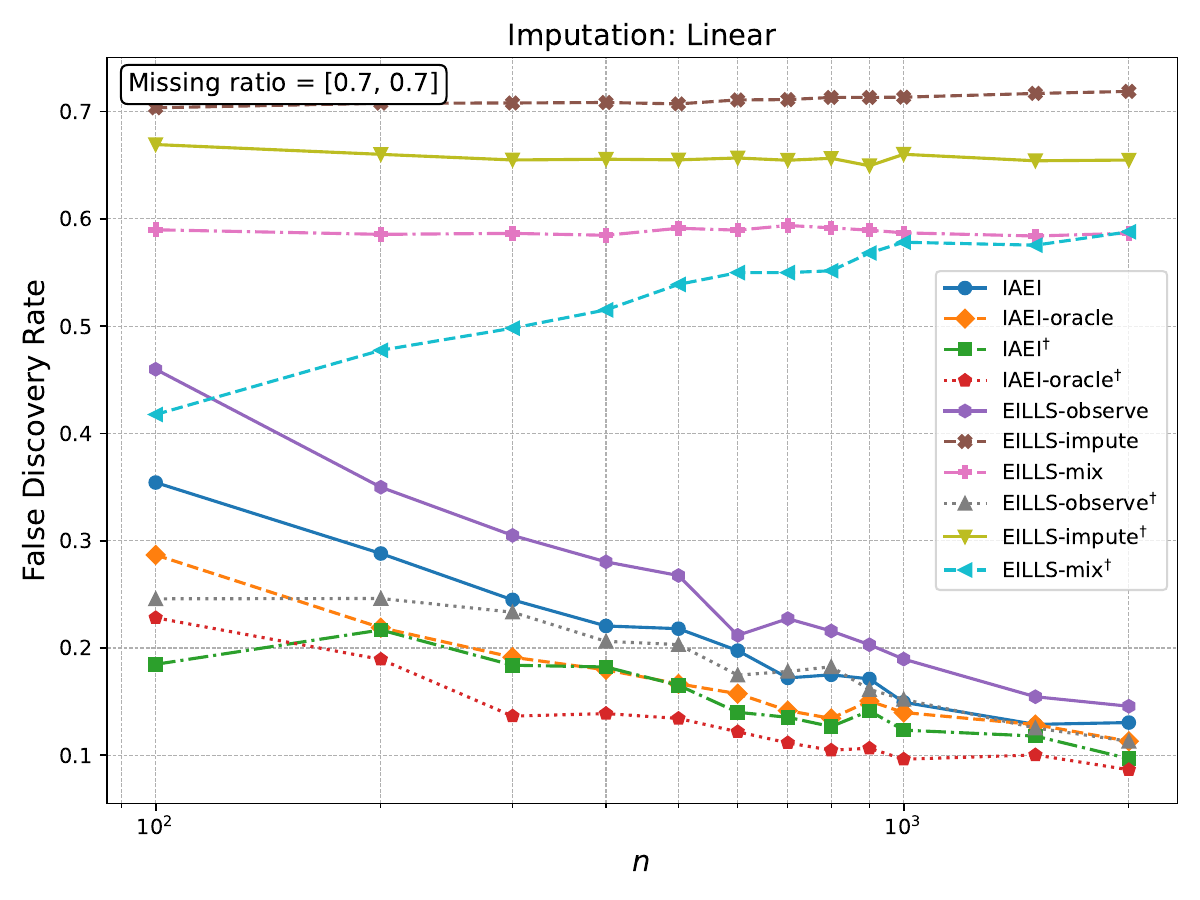} 
        \caption{FDR results under Model 0 with bias Linear imputation.}
        \label{fig:./figures/fig3b_and_fig3b1_and_fig3a1/bias_imputation/model0/fig3a1_bias_model0_model0_0.7_linear_all_fdr.pdf}
    \end{subfigure}
    \hfill 
    \begin{subfigure}[t]{0.32\textwidth} 
        \centering
        \includegraphics[width=\textwidth]{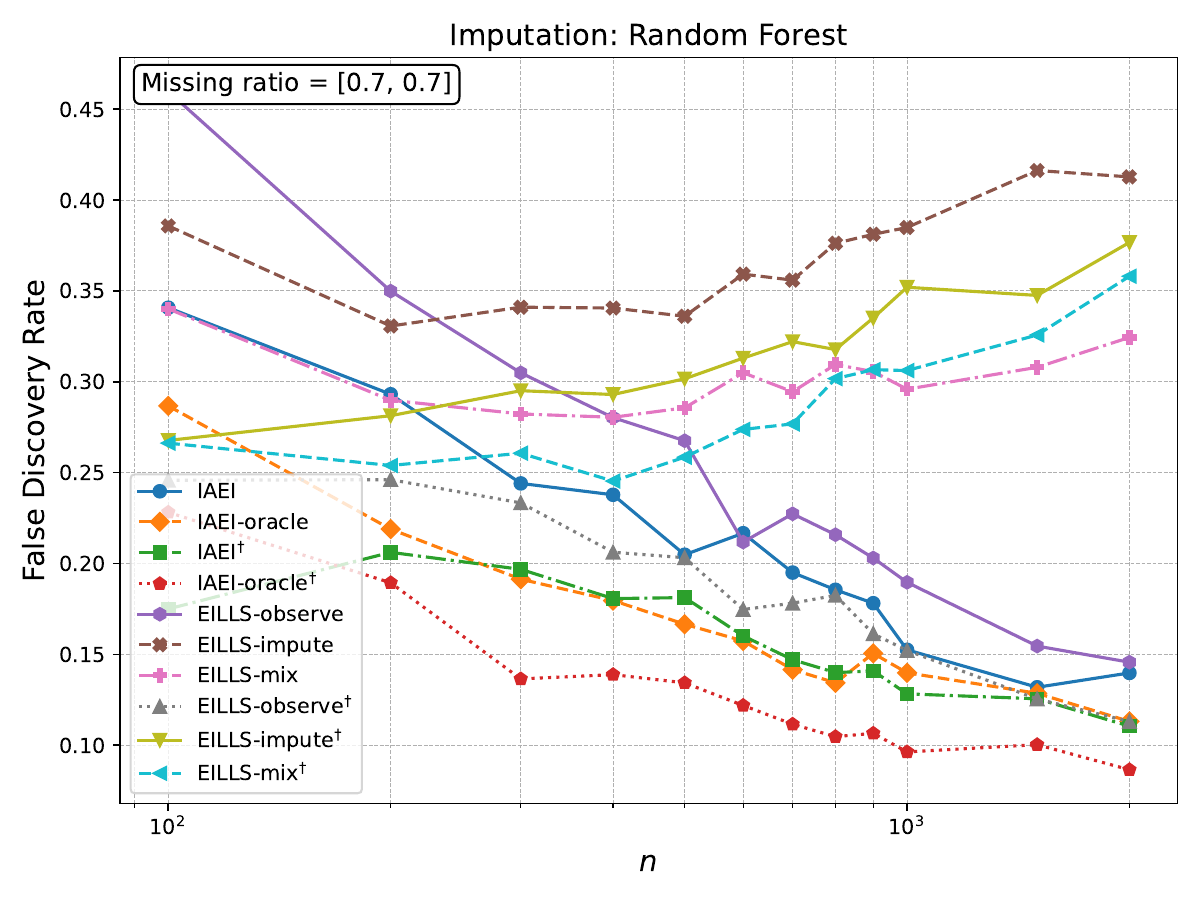} 
        \caption{FDR results under Model 0 with bias RandomForest imputation.}
        \label{fig:./figures/fig3b_and_fig3b1_and_fig3a1/bias_imputation/model0/fig3a1_bias_model0_model0_0.7_rf_all_fdr.pdf}
    \end{subfigure}
	\hfill 
    \begin{subfigure}[t]{0.32\textwidth} 
        \centering
        \includegraphics[width=\textwidth]{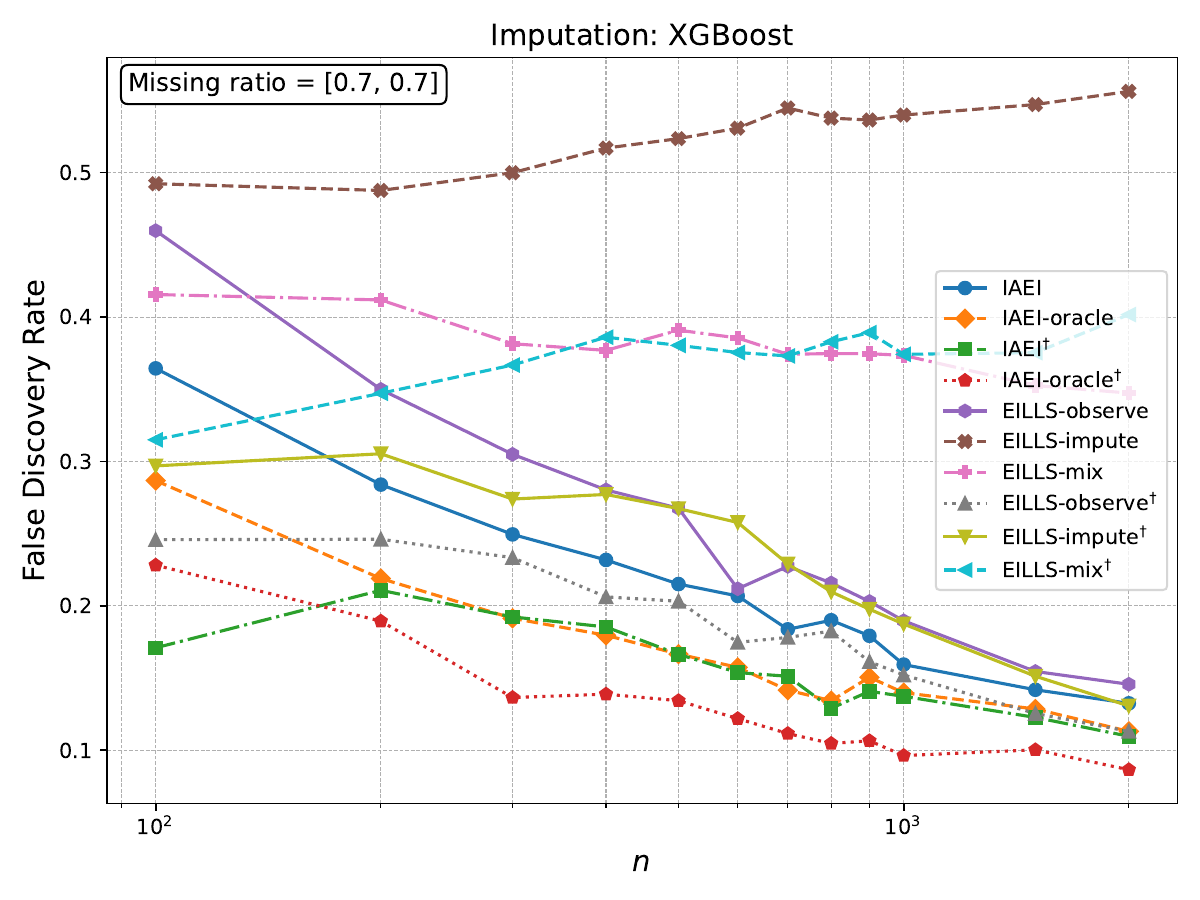} 
        \caption{FDR results under Model 0 with bias XGBoost imputation.}
        \label{fig:./figures/fig3b_and_fig3b1_and_fig3a1/bias_imputation/model0/fig3a1_bias_model0_model0_0.7_xgboost_all_fdr.pdf}
    \end{subfigure}
    \caption{With biased imputation, the naive methods fail to correctly identify the relevant set of variables, leading to high FDR. We observe IAEI$^{\dagger}$ still shows a noticeable gap from the oracle, regardless of the imputation method used.}
    \label{fig:./figures/fig3b_and_fig3b1_and_fig3a1/hbias_imputation/model0/fig3a1_hbias_model0_model0_0.7_linear_all_fdr.pdf./figures/fig3b_and_fig3b1_and_fig3a1/hbias_imputation/model0/fig3a1_hbias_model0_model0_0.7_rf_all_fdr.pdffig:./figures/fig3b_and_fig3b1_and_fig3a1/hbias_imputation/model0/fig3a1_hbias_model0_model0_0.7_xgboost_all_fdr.pdf}
\end{figure}

\begin{figure}[H]
    \centering
    \begin{subfigure}[t]{0.32\textwidth} 
        \centering
        \includegraphics[width=\textwidth]{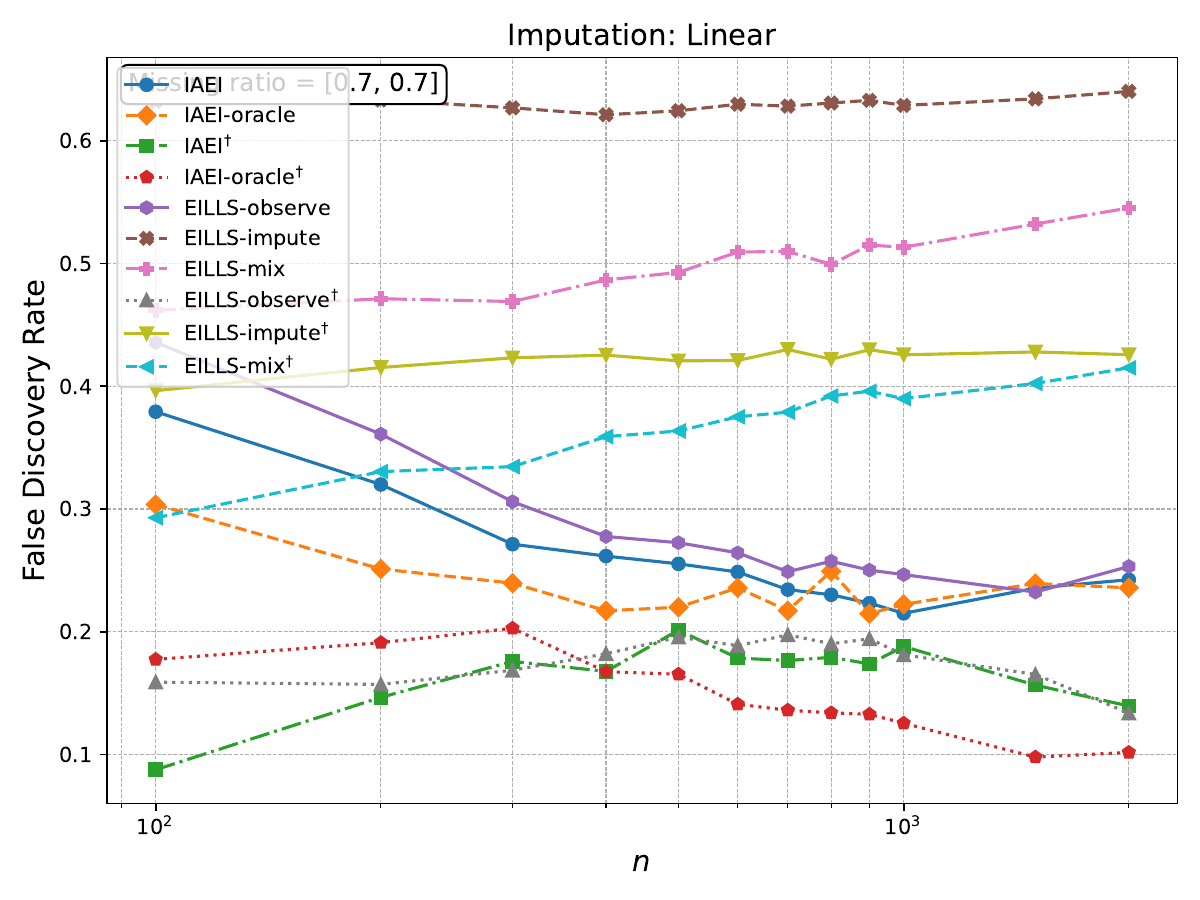} 
        \caption{FDR results under Model 3 with bias Linear imputation.}
        \label{fig:./figures/fig3b_and_fig3b1_and_fig3a1/bias_imputation/model3/fig3a1_bias_model3_model0_0.7_linear_all_fdr.pdf}
    \end{subfigure}
    \hfill 
    \begin{subfigure}[t]{0.32\textwidth} 
        \centering
        \includegraphics[width=\textwidth]{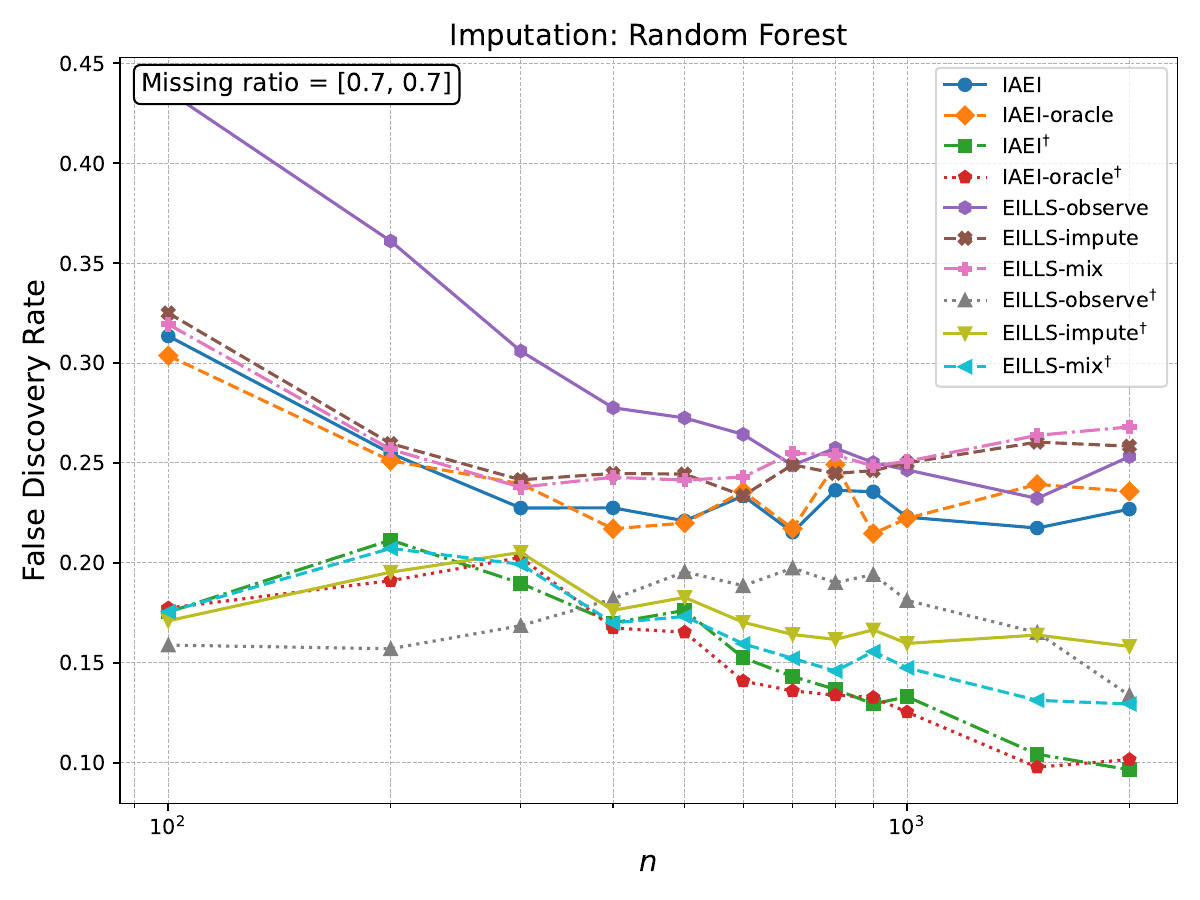} 
        \caption{FDR results under Model 3 with hbias RandomForest imputation.}
        \label{fig:./figures/fig3b_and_fig3b1_and_fig3a1/bias_imputation/model3/fig3a1_bias_model3_model0_0.7_rf_all_fdr.pdf}
    \end{subfigure}
	\hfill 
    \begin{subfigure}[t]{0.32\textwidth} 
        \centering
        \includegraphics[width=\textwidth]{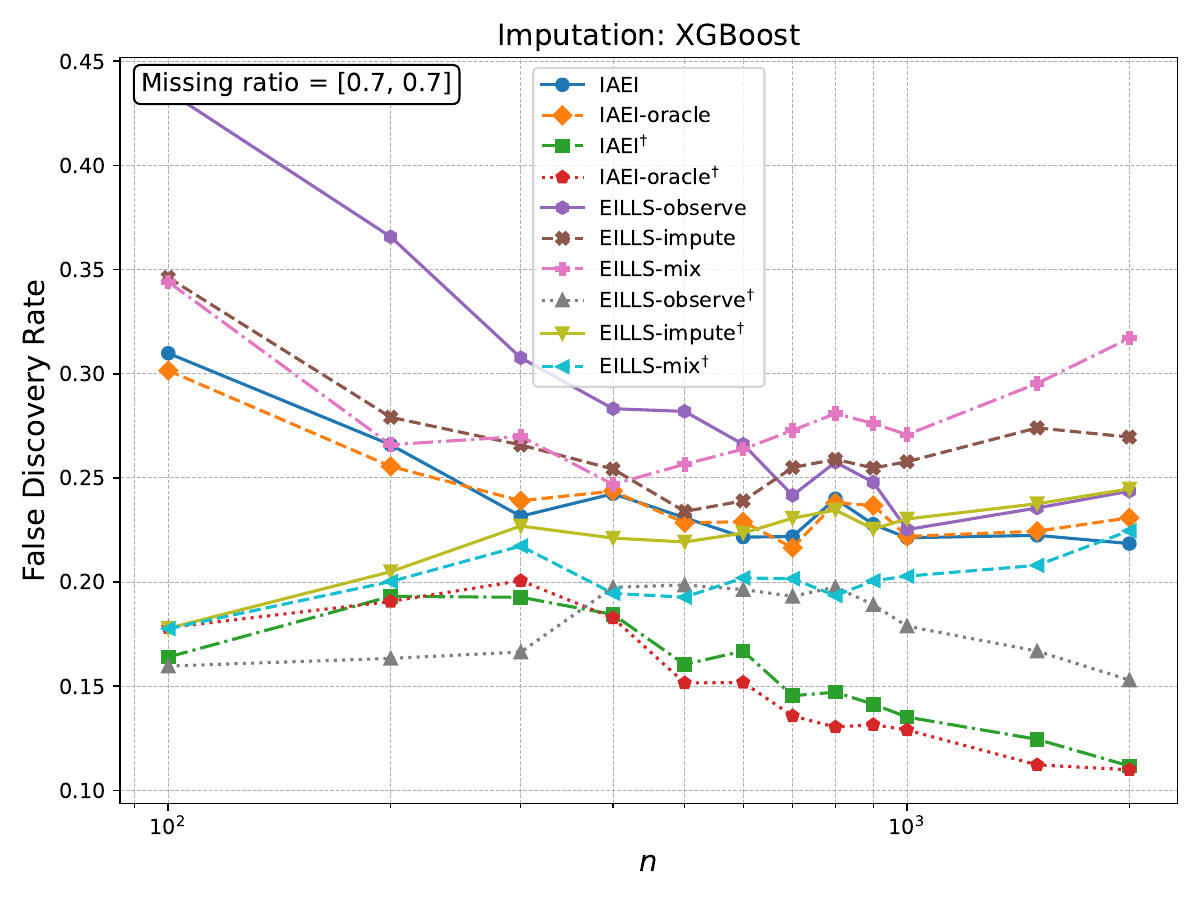} 
        \caption{FDR results under Model 3 with bias XGBoost imputation.}
        \label{fig:./figures/fig3b_and_fig3b1_and_fig3a1/bias_imputation/model3/fig3a1_bias_model3_model0_0.7_xgboost_all_fdr.pdf}
    \end{subfigure}
    \caption{Under Model 3, the advantage of employing IAEI$^{\dagger}$ with nonlinear imputation methods becomes evident, as it achieves an FDR comparable to the oracle. In contrast, the naive methods, EILLS-impute$^{\dagger}$ and EILLS-mix$^{\dagger}$, fail to select the correct variables due to biased labels, while EILLS-observe struggles by not fully utilizing the available information.}
    \label{fig:./figures/fig3b_and_fig3b1_and_fig3a1/bias_imputation/model3/fig3a1_bias_model3_model0_0.7_linear_all_fdr.pdf./figures/fig3b_and_fig3b1_and_fig3a1/bias_imputation/model3/fig3a1_bias_model3_model0_0.7_rf_all_fdr.pdf./figures/fig3b_and_fig3b1_and_fig3a1/bias_imputation/model3/fig3a1_bias_model3_model0_0.7_xgboost_all_fdr.pdf}
\end{figure}

The results have consistently demonstrated that methods with the enhanced penalty outperform those using the original penalty in terms of average FDR over 500 iterations. To offer a complementary perspective, we present a single simulation result under Model 3 with precise imputation, highlighting how the enhanced penalty influences variable selection performance. Results for other models are provided in the supplementary material.

\begin{figure}[H]
    \centering
    \begin{subfigure}[t]{0.45\textwidth} 
        \centering
        \includegraphics[width=\textwidth]{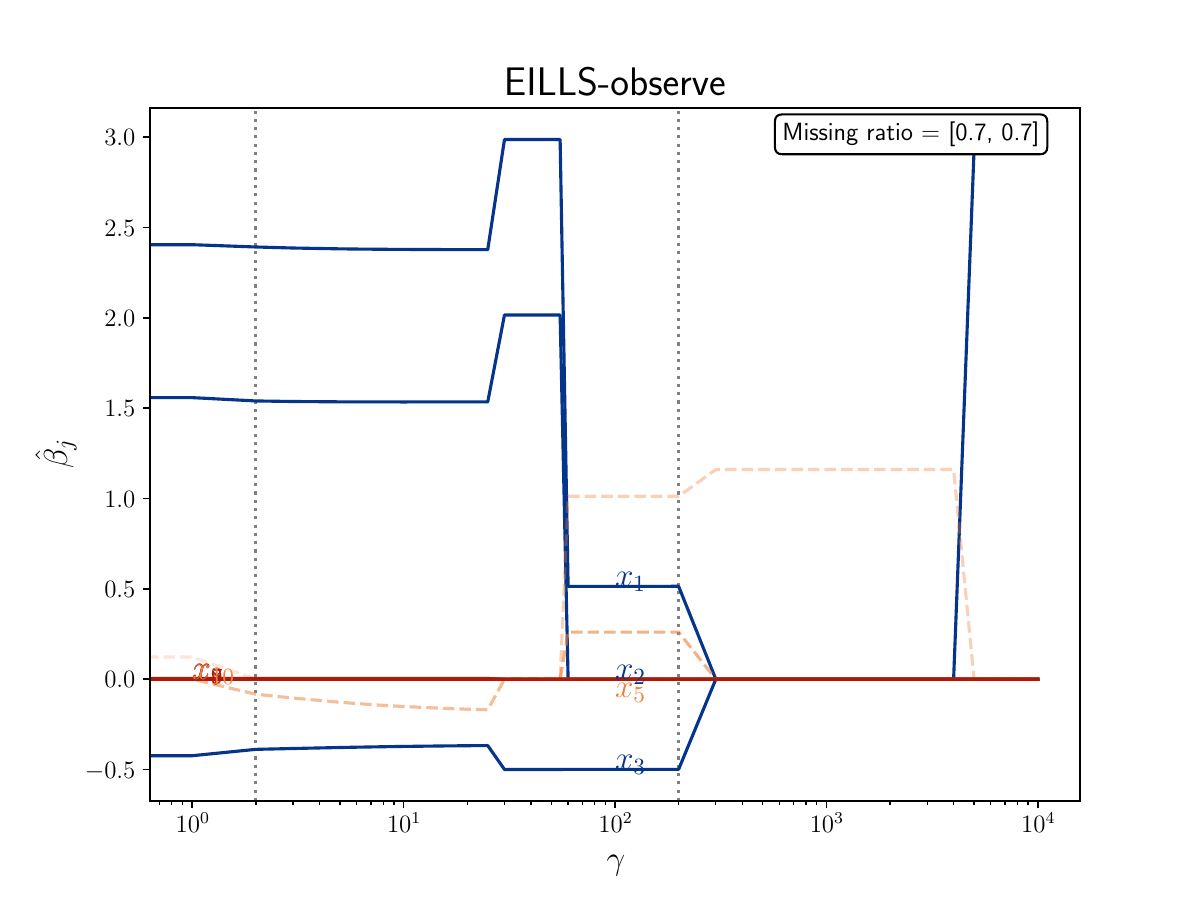} 
        \caption{EILLS-observe.}
        \label{fig:./figures/fig3a/precise_imputation/model3/fig3a_xgboost_eills_observe.pdf}
    \end{subfigure}
    \hfill
    \begin{subfigure}[t]{0.45\textwidth} 
        \centering
        \includegraphics[width=\textwidth]{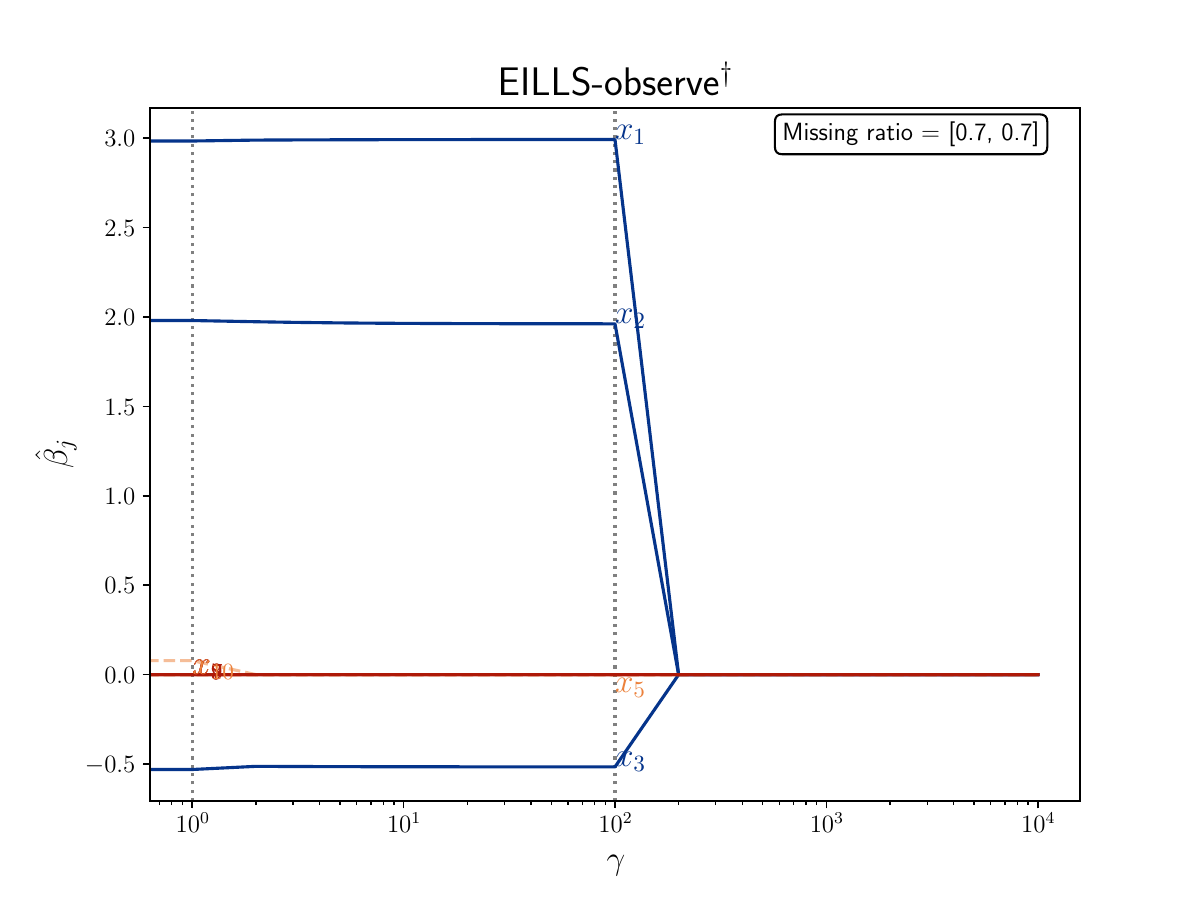} 
        \caption{EILLS-observe$^{\dagger}$.}
        \label{fig:./figures/fig3a/precise_imputation/model3/fig3a_xgboost_eills_observe_ce.pdf}
    \end{subfigure}

	\caption{Variable selection performance of EILLS-observe compared to EILLS-observe$^{\dagger}$ under Model 3. The left plot illustrates the performance using the original penalty, while the right plot highlights the improved stability and accuracy achieved with the enhanced penalty.}
    
    \label{fig:variable_selection_one_simulation1}
\end{figure}

\begin{figure}[H]
    \centering
    
    \begin{subfigure}[t]{0.45\textwidth} 
        \centering
        \includegraphics[width=\textwidth]{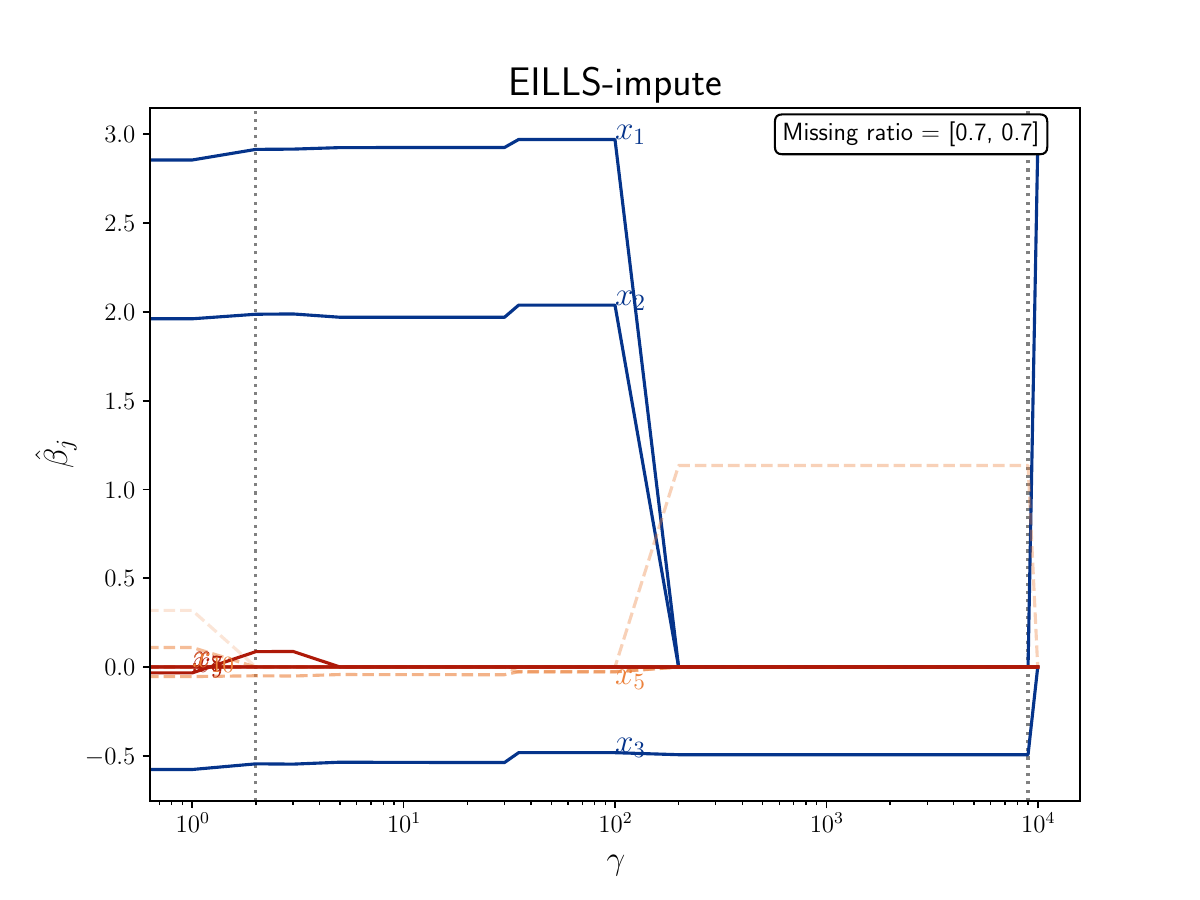} 
        \caption{EILLS-impute.}
        \label{fig:./figures/fig3a/precise_imputation/model3/fig3a_xgboost_eills_impute.pdf}
    \end{subfigure}
    \hfill
    \begin{subfigure}[t]{0.45\textwidth} 
        \centering
        \includegraphics[width=\textwidth]{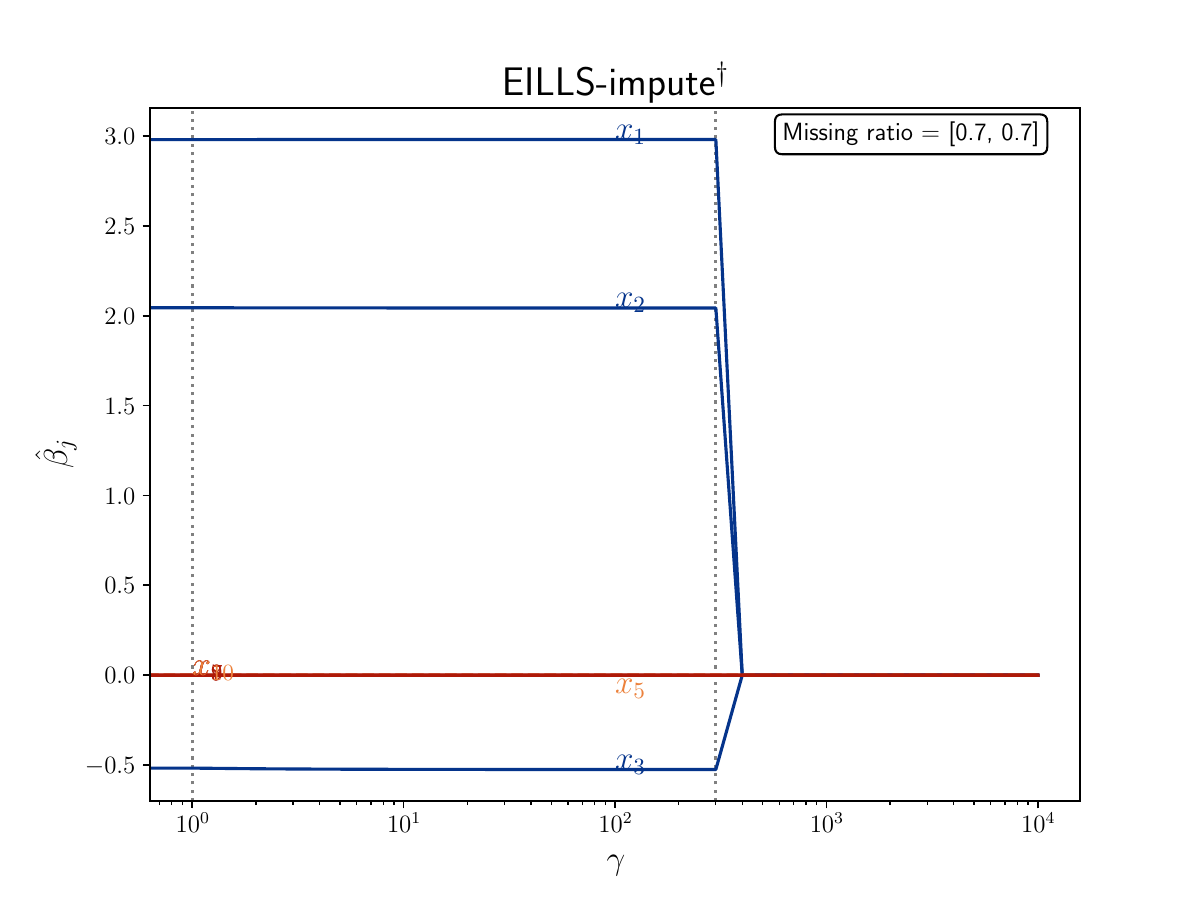} 
        \caption{EILLS-impute$^{\dagger}$.}
        \label{fig:./figures/fig3a/precise_imputation/model3/fig3a_xgboost_eills_impute_ce.pdf}
    \end{subfigure}

	\caption{Variable selection performance of EILLS-impute compared to EILLS-impute$^{\dagger}$ under Model 3 with precise imputation. The suboptimal performance in the left plot is not attributable to biased imputation, as this analysis focuses on precise imputation. The left plot illustrates the performance with the original penalty, while the right plot demonstrates the enhanced stability and accuracy achieved with the improved penalty.}
    
    \label{fig:variable_selection_one_simulation2}
\end{figure}

\begin{figure}[H]
    \centering

    \begin{subfigure}[t]{0.45\textwidth} 
        \centering
        \includegraphics[width=\textwidth]{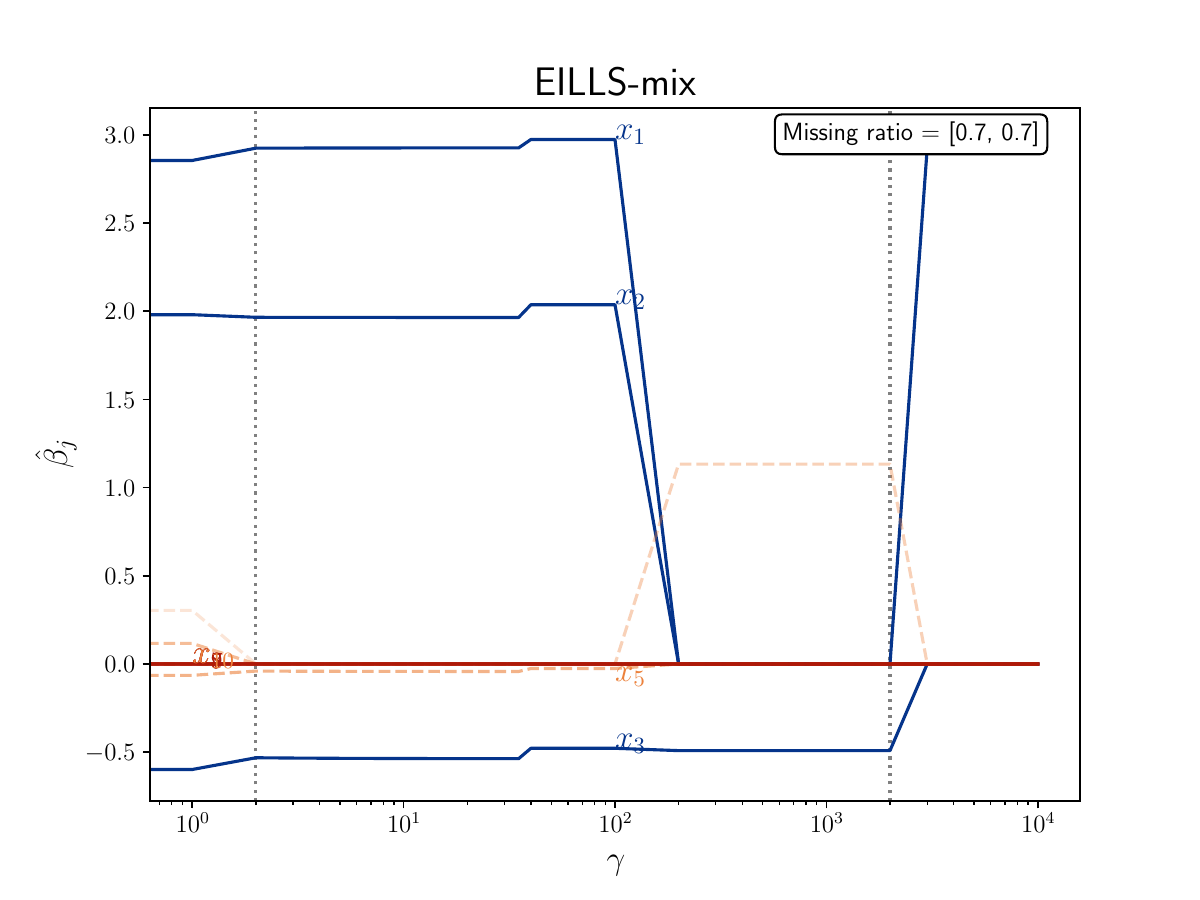} 
        \caption{EILLS-mix.}
        \label{fig:./figures/fig3a/precise_imputation/model3/fig3a_xgboost_eills_mix.pdf}
    \end{subfigure}
    \hfill
    \begin{subfigure}[t]{0.45\textwidth} 
        \centering
        \includegraphics[width=\textwidth]{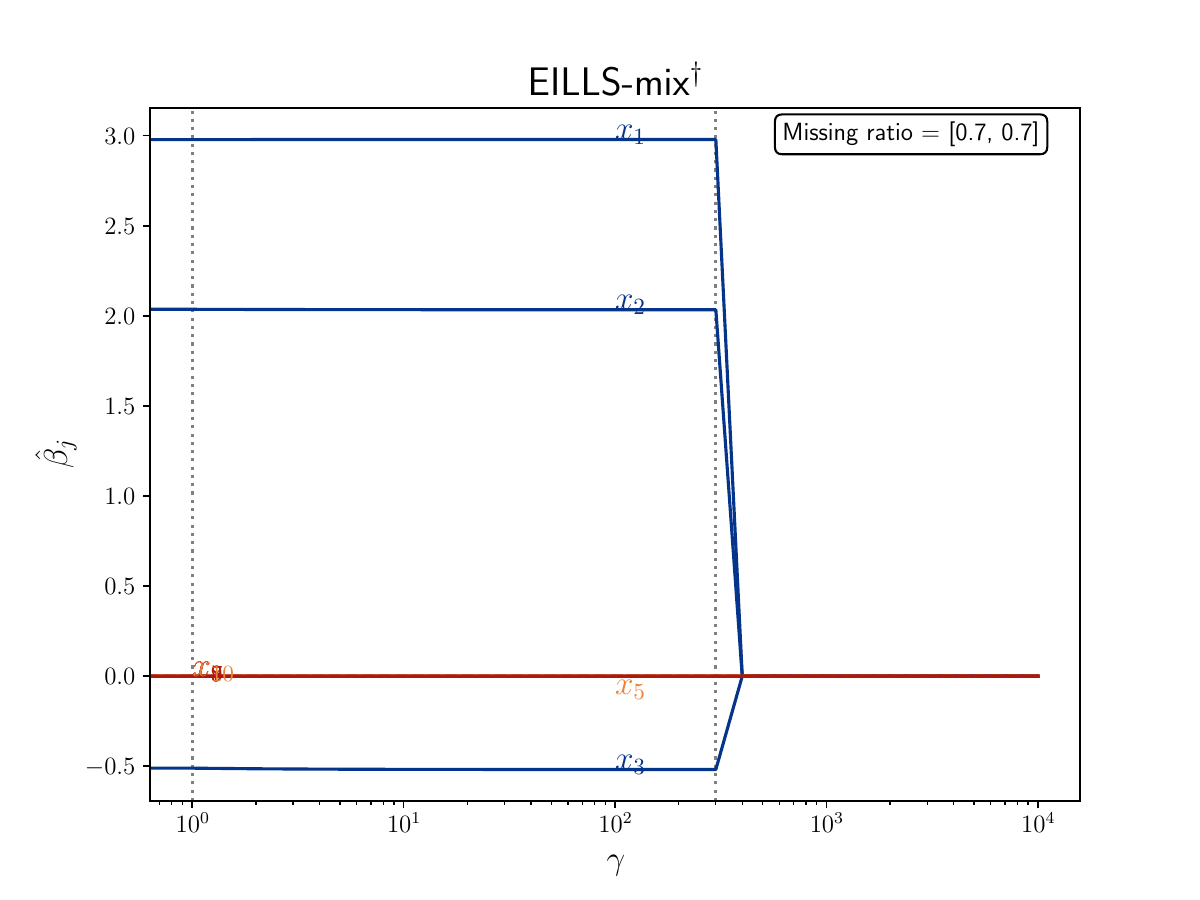} 
        \caption{EILLS-mix$^{\dagger}$.}
        \label{fig:./figures/fig3a/precise_imputation/model3/fig3a_xgboost_eills_mix_ce.pdf.}
    \end{subfigure}

	\caption{Variable selection performance of EILLS-mix vs. EILLS-mix$^{\dagger}$ under Model 3 with precise imputation, where the right plot demonstrates the enhanced penalty’s improved stability and accuracy.}
    
    \label{fig:variable_selection_one_simulation3}
\end{figure}

\begin{figure}[H]
    \centering

    \begin{subfigure}[t]{0.45\textwidth} 
        \centering
        \includegraphics[width=\textwidth]{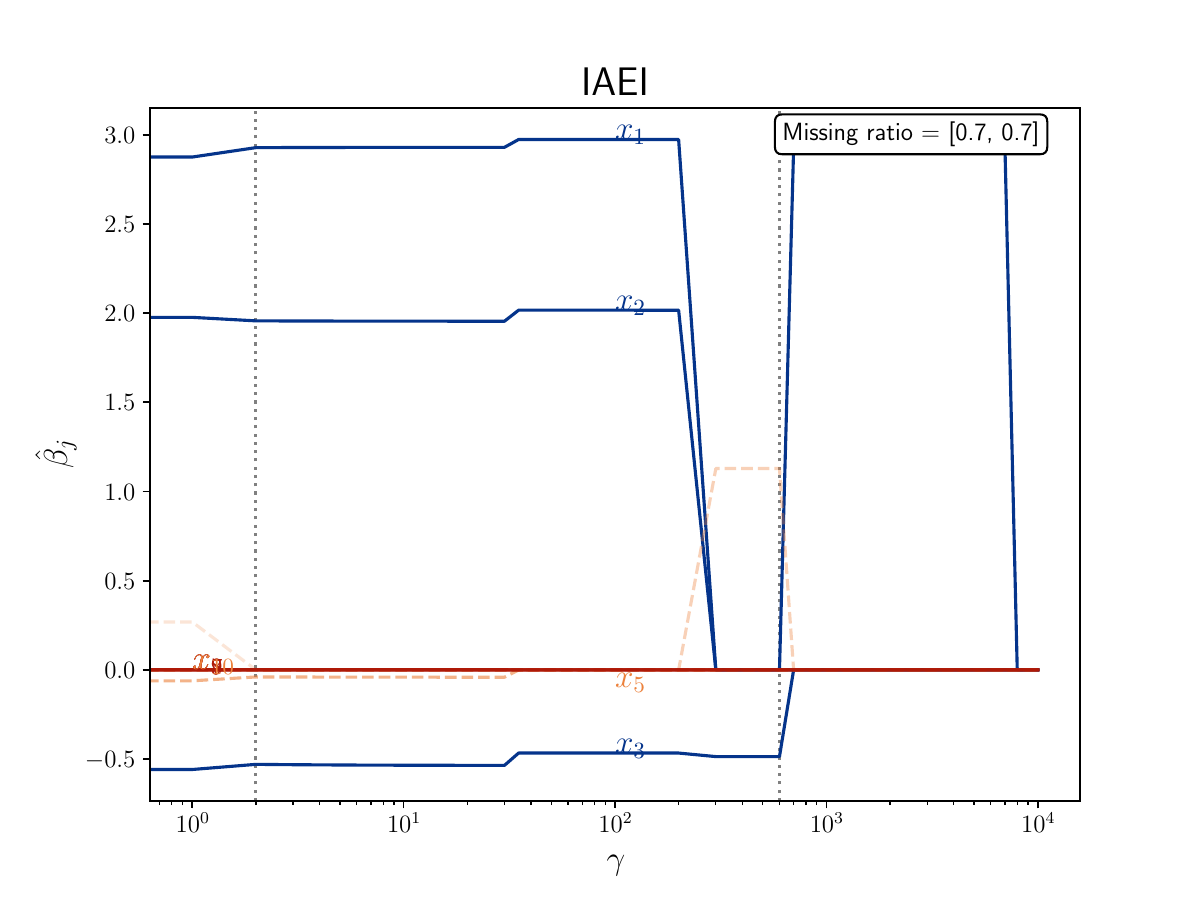} 
        \caption{IAEI.}
        \label{fig:./figures/fig3a/precise_imputation/model3/fig3a_xgboost_iaei.pdf}
    \end{subfigure}
    \hfill
    \begin{subfigure}[t]{0.45\textwidth} 
        \centering
        \includegraphics[width=\textwidth]{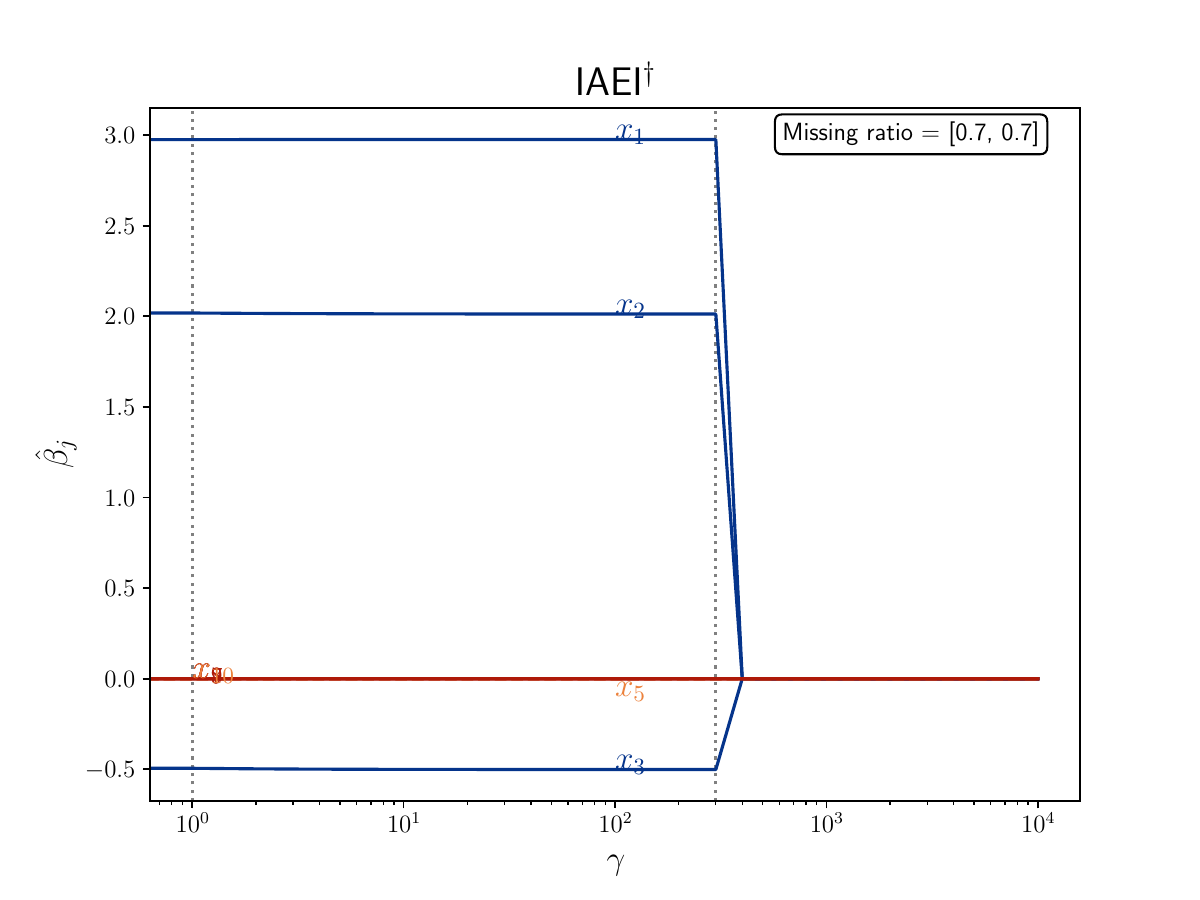} 
        \caption{IAEI$^{\dagger}$.}
        \label{fig:./figures/fig3a/precise_imputation/model3/fig3a_xgboost_iaei_ce.pdf}
    \end{subfigure}
    
    \caption{Variable selection performance of IAEI vs IAEI$^{\dagger}$ under Model 3. While the left plot shows the result with original penalty, the right plot demonstrates the improved stable performance using the enhanced penalty.}
    \label{fig:variable_selection_one_simulation4}
\end{figure}

\subsection{Performance on $\ell_2$ Error Convergence}\label{simulation:l2_error_convergence}

In this section, we evaluate the empirical Mean Squared Error (MSE) of the proposed IAEI estimator and its counterparts. Specifically, the empirical MSE is computed as the $\ell_2$ norm of the different between the estimator and the true coefficient $\boldsymbol{\beta}^*$, averaged over 500 replications for each simulation scenario. This approach provides a robust measure of the estimators’ accuracy and convergence behavior under varying complexities of data-generating mechanisms and imputation strategies.

Unlike the findings in Section \ref{simulation:variable_selection}, methods with the enhanced penalty exhibit greater variability. 
Overall, methods with the original penalty demonstrate lower empirical MSE, making them more reliable in this context. Therefore, our focus is on evaluating the performance of methods using the original penalty. 
\begin{figure}[H]
    \centering
    \begin{subfigure}[t]{0.32\textwidth} 
        \centering
        \includegraphics[width=\textwidth]{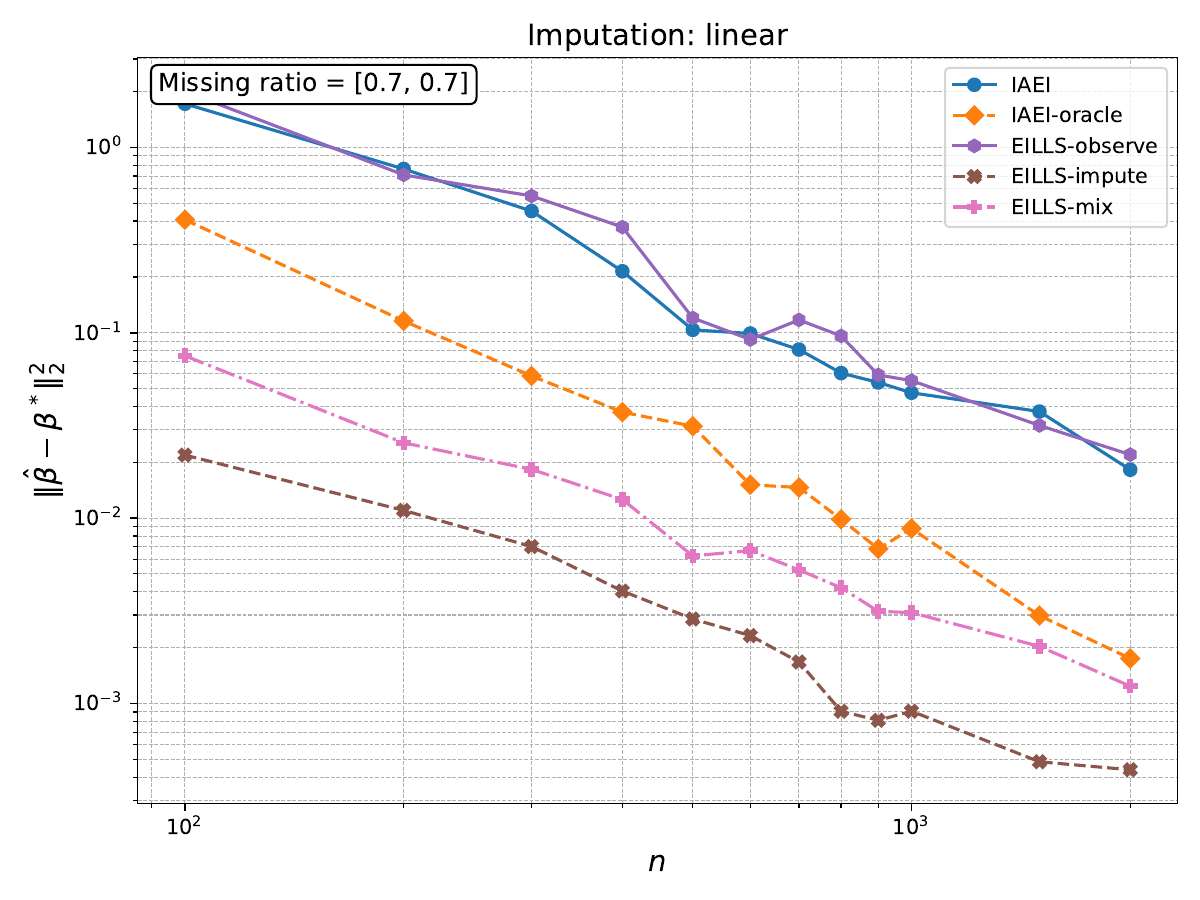}
        \caption{Compare methods with Linear imputation.}
        \label{fig:./figures/fig3b_and_fig3b1_and_fig3a1/precise_imputation/model1/fig3b_precise_model1_model0_0.7_linear_simple.pdf}
    \end{subfigure}
    \hfill 
    \begin{subfigure}[t]{0.32\textwidth} 
        \centering
        \includegraphics[width=\textwidth]{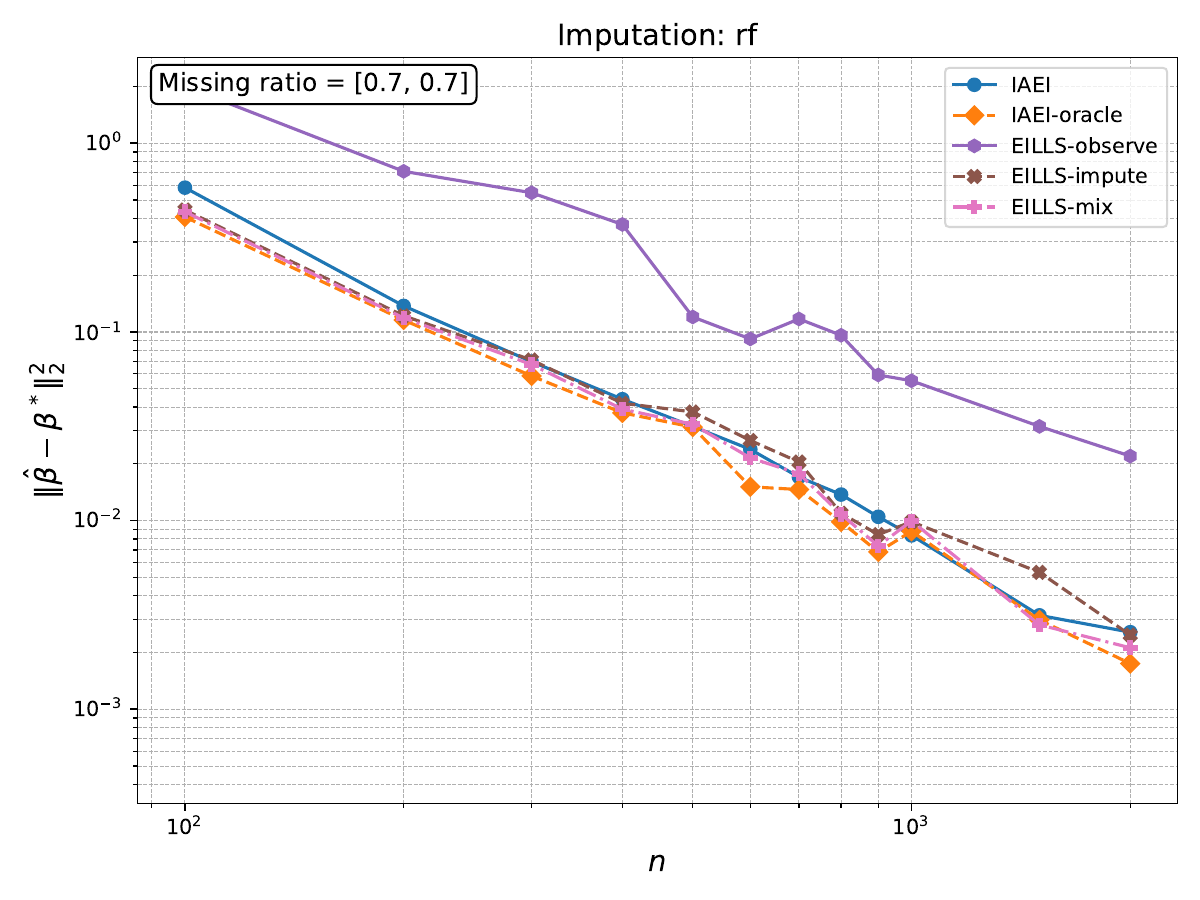}
        \caption{Compare methods with RandomForest imputation.}
        \label{fig:./figures/fig3b_and_fig3b1_and_fig3a1/precise_imputation/model1/fig3b_precise_model1_model0_0.7_rf_simple.pdf}
    \end{subfigure}
    \hfill 
    \begin{subfigure}[t]{0.32\textwidth} 
        \centering
        \includegraphics[width=\textwidth]{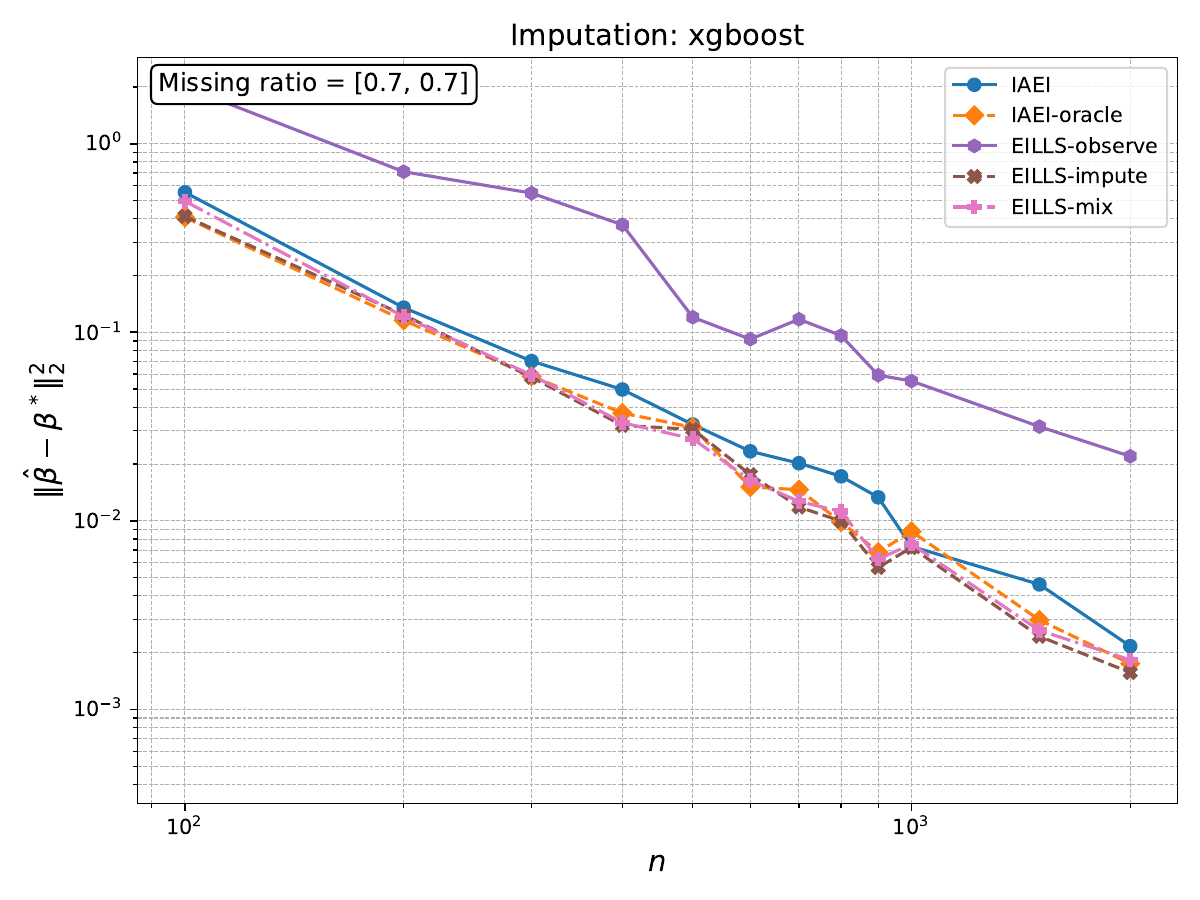}
        \caption{Compare methods with XGBoost imputation.}
        \label{fig:./figures/fig3b_and_fig3b1_and_fig3a1/precise_imputation/model1/fig3b_precise_model1_model0_0.7_xgboost_simple.pdf}
    \end{subfigure}

    \caption{Compare methods with Linear, RandomForest, and XGBoost imputation under data generating process Model 1.}
    \label{fig:./figures/fig3b_and_fig3b1_and_fig3a1/precise_imputation/model1/fig3b_precise_model1_model0_0.7_linear_simple.pdf./figures/fig3b_and_fig3b1_and_fig3a1/precise_imputation/model1/fig3b_precise_model1_model0_0.7_rf_simple.pdf./figures/fig3b_and_fig3b1_and_fig3a1/precise_imputation/model1/fig3b_precise_model1_model0_0.7_xgboost_simple.pdf}
\end{figure}

When precise linear regression models are available for each environment, EILLS-impute, which replaces all labels with imputed ones, achieves the best performance. This is because, first, the error of the best linear predictor is uncorrelated with linear functions of predictors, avoiding spurious biases in the penalty component of the objective. Meanwhile, unlike nonlinear imputation methods, such as RandomForest or XGBoost, which estimate the true regression $E[y^{(e)}\mid \boldsymbol{x}^{(e)}]$ and capture additional information about $y$ beyond the linear structure, linear imputation provides exactly the necessary and sufficient information for identifying the best invariant linear predictor, together with the unbiased penalty.  In other words, while $E[y^{(e)}\mid \boldsymbol{x}^{(e)}]$ encompasses richer information, this added complexity introduces unnecessary variability when the goal is restricted to identifying linear relationships. By replacing $y$ with the best linear predictor, EILLS-impute eliminates this extraneous variability, ensuring that the estimation process is both efficient and tightly aligned with the inherent linear structure of the problem. From the simulation, we also observe that a smaller missing ratio further amplifies the performance gap between EILLS-impute with precise linear imputation and methods that only fill missing labels.

\begin{figure}[H]
    \centering
    \begin{subfigure}[t]{0.45\textwidth}
        \centering
        \includegraphics[width=\textwidth]{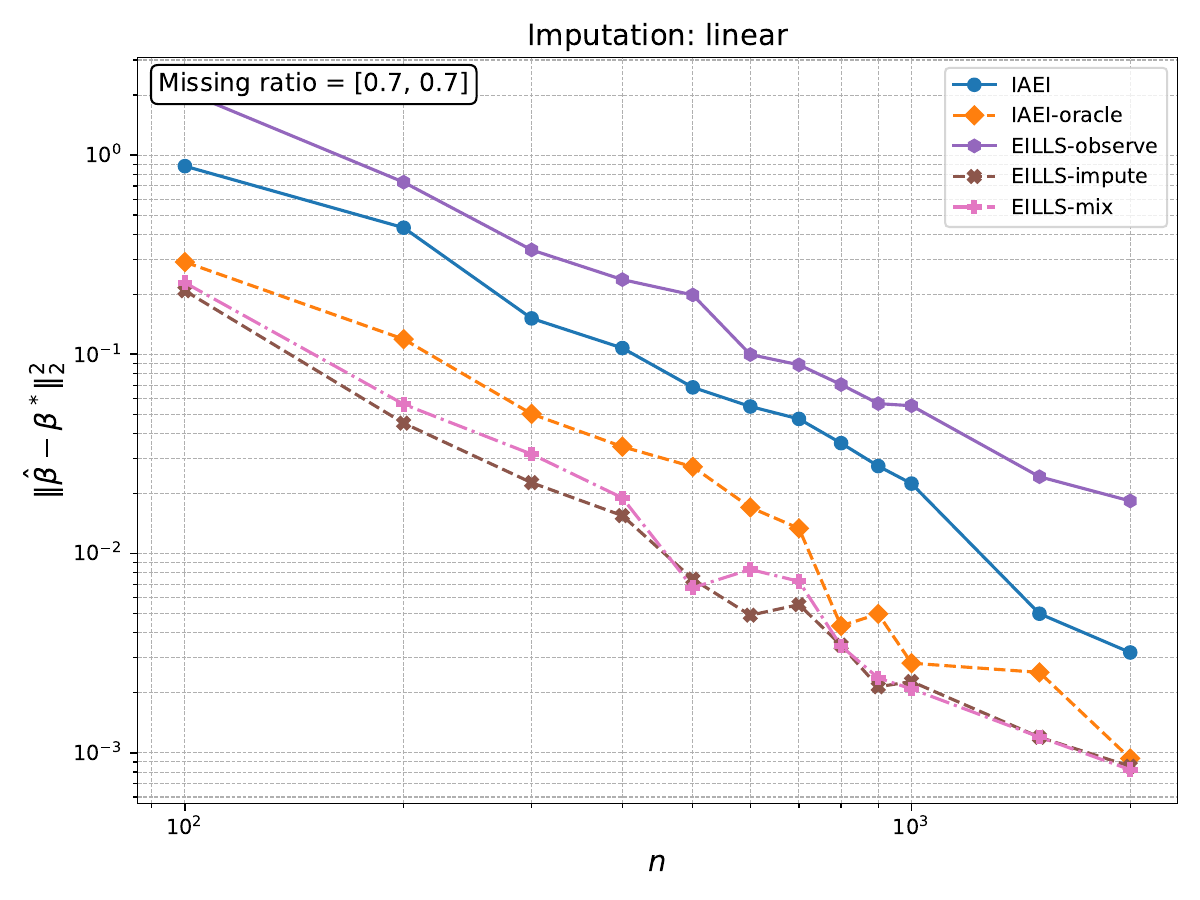}
        \caption{Methods under Model 0 with missing ratio $0.7$.}
        \label{fig:./figures/fig3b_and_fig3b1_and_fig3a1/precise_imputation/model0/fig3b_precise_model0_model0_0.7_linear_simple.pdf}
    \end{subfigure}
    \hfill
    \begin{subfigure}[t]{0.45\textwidth}
        \centering
        \includegraphics[width=\textwidth]{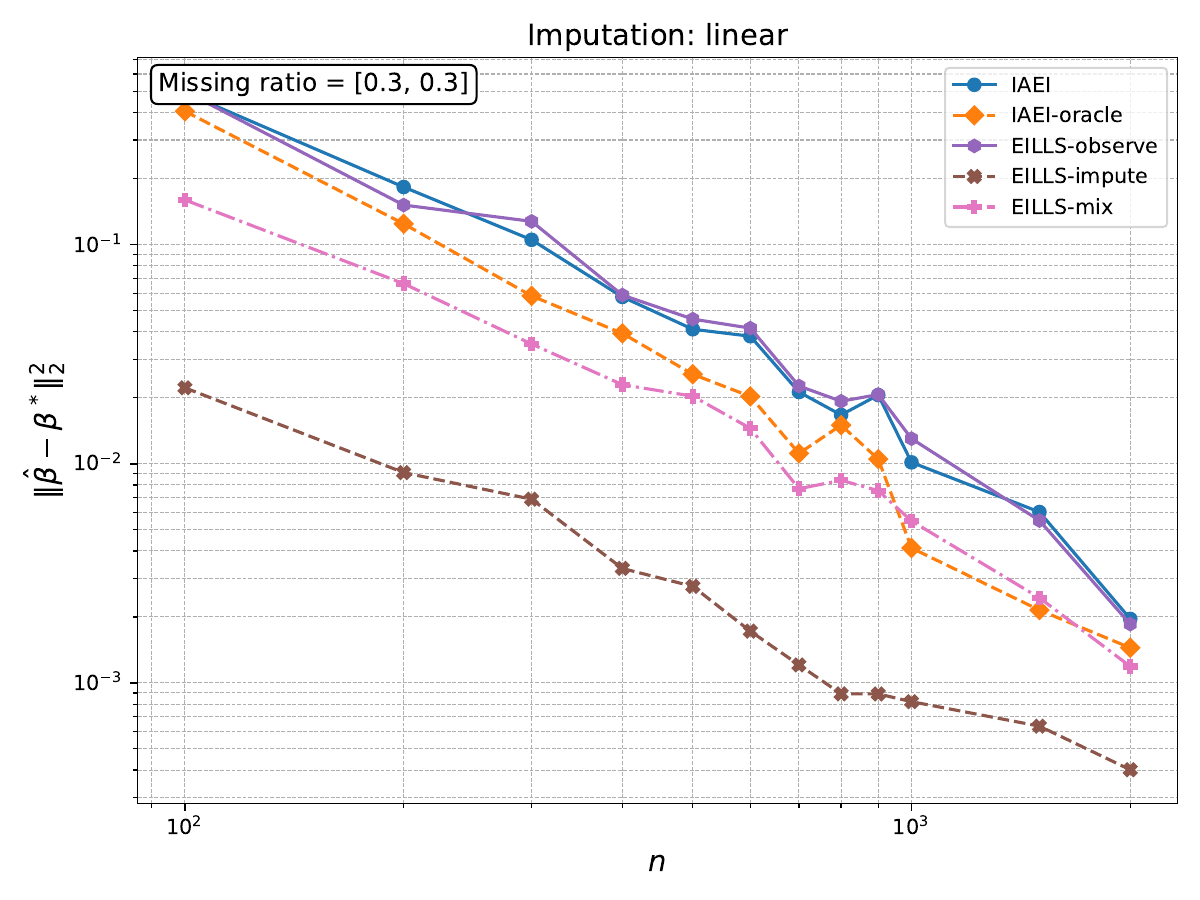}
        \caption{Methods under Model 0 with missing ratio $0.3$.}
        \label{fig:./figures/fig3b_and_fig3b1_and_fig3a1/precise_imputation/model1/fig3b_precise_model1_model0_0.3_linear_simple.pdf}
    \end{subfigure}
	\caption{When the missing ratio decreases from $0.7$ to $0.3$, more labels are replaced with the best linear estimates under the EILLS-impute method, resulting in lower empirical MSE.}
\end{figure}

In the following plots, we can see that IAEI with linear imputation performs only marginally better than EILLS-observe across all settings, with the performance gap being more noticeable in simpler scenarios like Model 0. However, as the complexity of the settings increases with the introduction of nonlinearity in Models 1–3, the advantage of IAEI with linear imputation diminishes, making it nearly comparable to EILLS-observe.

\begin{figure}[H]
    \centering
    \begin{subfigure}[t]{0.45\textwidth}
        \centering
        \includegraphics[width=\textwidth]{./figures/fig3b_and_fig3b1_and_fig3a1/precise_imputation/model0/fig3b_precise_model0_model0_0.7_linear_simple.pdf}
        \caption{Compare methods in Model 0.}
        \label{fig:./figures/fig3b_and_fig3b1_and_fig3a1/precise_imputation/model0/fig3b_precise_model0_model0_0.7_linear_simple.pdf}
    \end{subfigure}
    \hfill
    \begin{subfigure}[t]{0.45\textwidth}
        \centering
        \includegraphics[width=\textwidth]{./figures/fig3b_and_fig3b1_and_fig3a1/precise_imputation/model1/fig3b_precise_model1_model0_0.7_linear_simple.pdf}
        \caption{Compare methods in Model 1.}
        \label{fig:./figures/fig3b_and_fig3b1_and_fig3a1/precise_imputation/model1/fig3b_precise_model1_model0_0.7_linear_simple.pdf}
    \end{subfigure}

    \vspace{1em} 

    \begin{subfigure}[t]{0.45\textwidth}
        \centering
        \includegraphics[width=\textwidth]{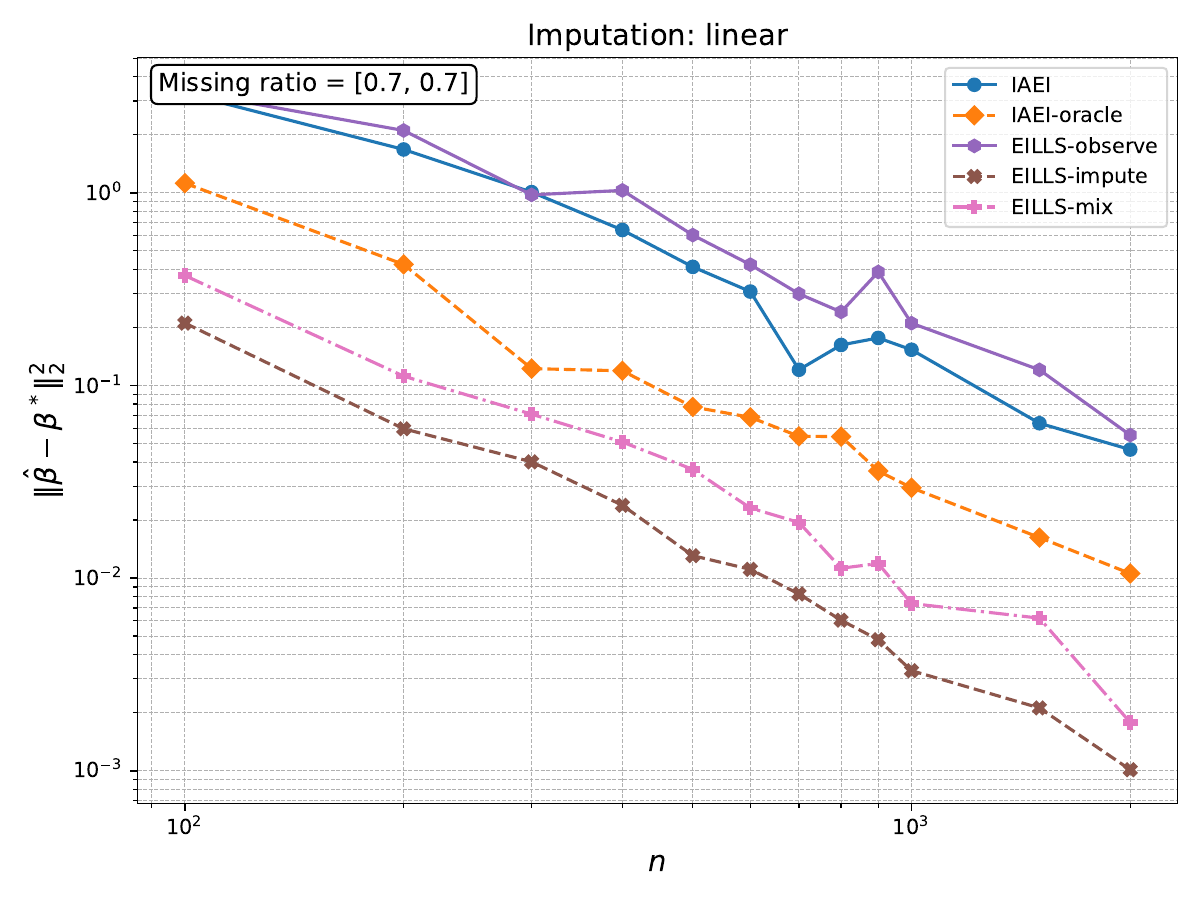}
        \caption{Compare methods in Model 2.}
        \label{fig:./figures/fig3b_and_fig3b1_and_fig3a1/precise_imputation/model2/fig3b_precise_model2_model0_0.7_linear_simple.pdf}
    \end{subfigure}
    \hfill
    \begin{subfigure}[t]{0.45\textwidth}
        \centering
        \includegraphics[width=\textwidth]{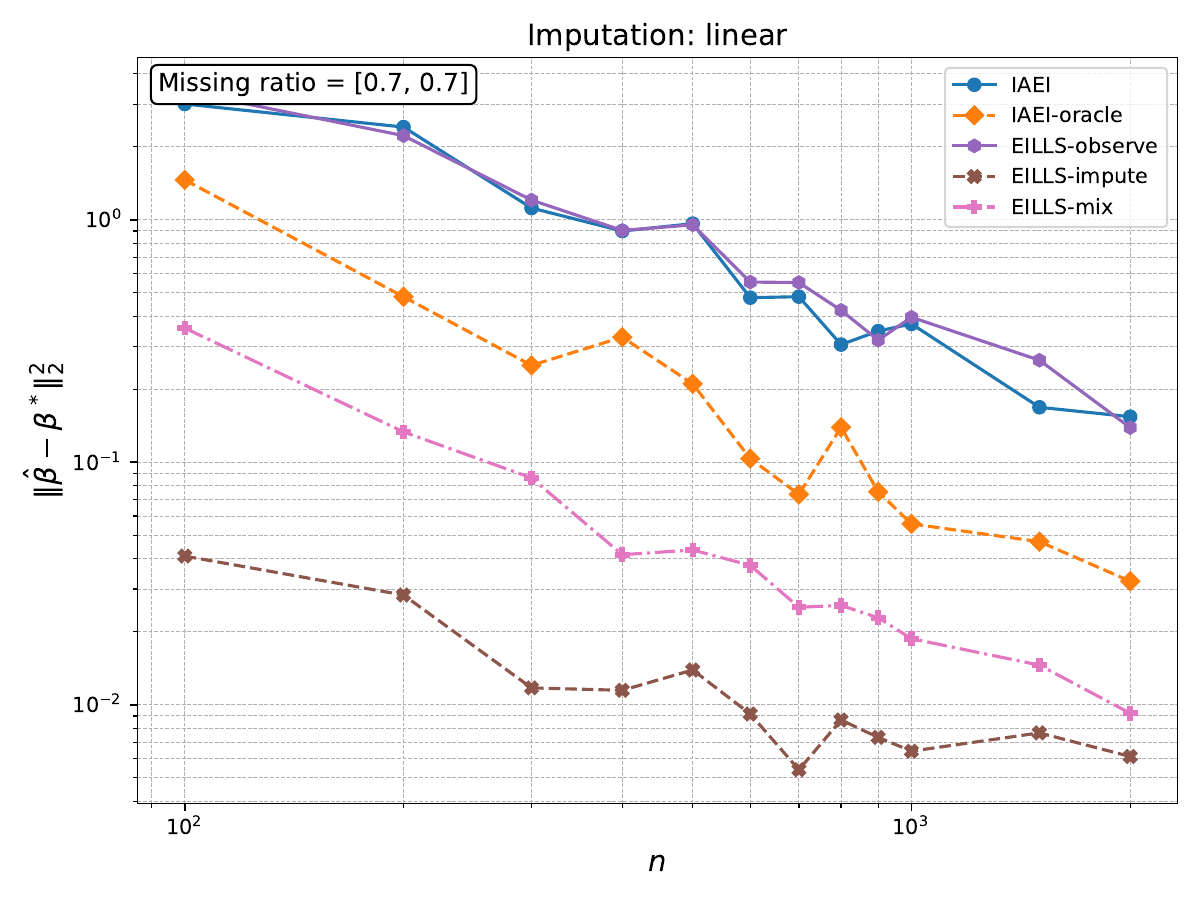}
        \caption{Compare methods in Model 3.}
        \label{fig:./figures/fig3b_and_fig3b1_and_fig3a1/precise_imputation/model3/fig3b_precise_model3_model0_0.7_linear_simple.pdf}
    \end{subfigure}

    \caption{Comapre methods across Models 0-3 with precise linear imputation.}
    \label{fig:ff}
\end{figure}

In contrast, as we will see below, IAEI with nonlinear imputation methods such as XGBoost or RandomForest shows a significant performance boost over EILLS-observe in Models 1–3, effectively leveraging the added nonlinearity to handle the complexity of these scenarios. This highlights the effectiveness of nonlinear imputation in addressing the limitations of EILLS-observe in more complex settings. Notably, with nonlinear imputation, EILLS-impute and EILLS-mix no longer exceed the oracle’s performance but instead align closely with it. This suggests that in nonlinear settings Models 1–3, IAEI, EILLS-impute, and EILLS-mix with nonlinear imputation achieve performance comparable to the oracle, while EILLS-observe remains the only method that consistently underperforms.

\begin{figure}[H]
    \centering
    \begin{subfigure}[t]{0.45\textwidth}
        \centering
        \includegraphics[width=\textwidth]{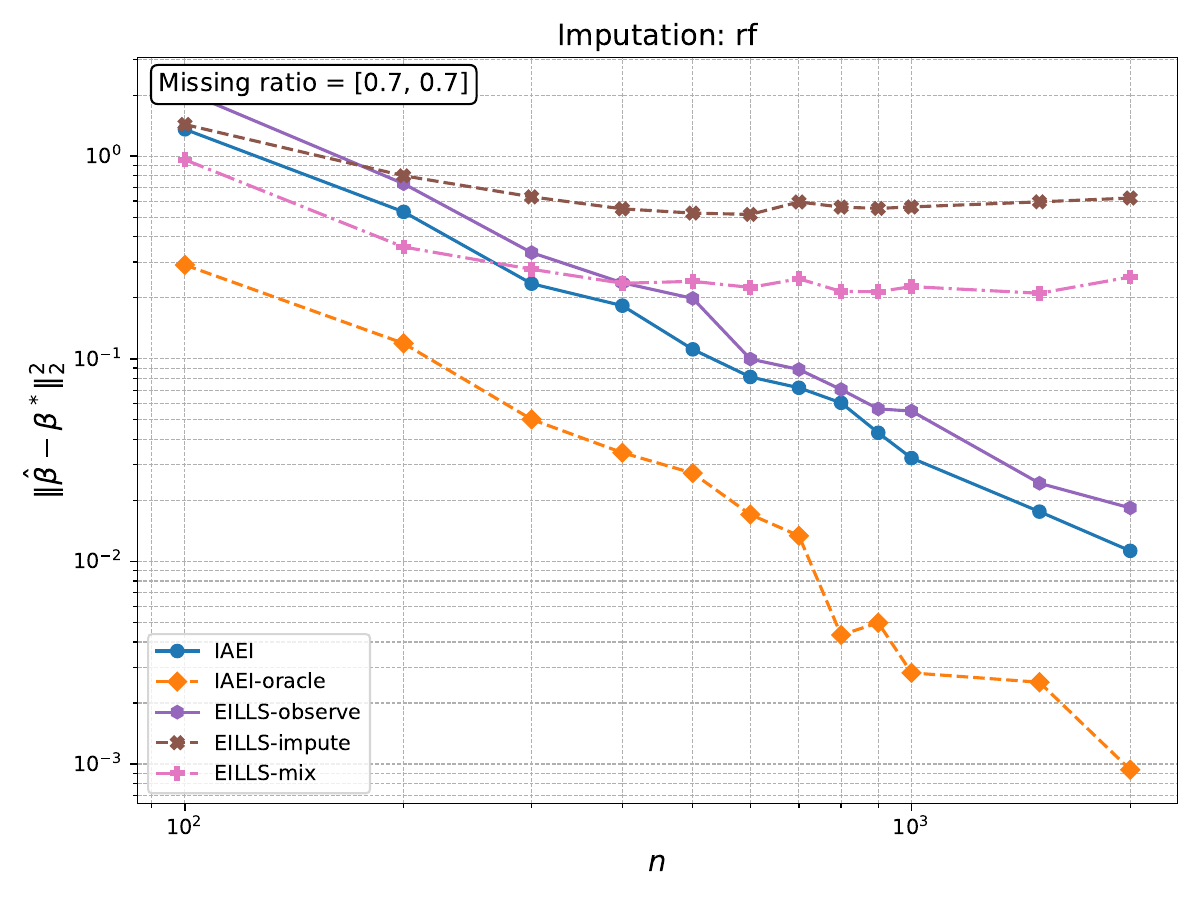}
        \caption{Compare methods in Model 0.}
        \label{fig:./figures/fig3b_and_fig3b1_and_fig3a1/precise_imputation/model0/fig3b_precise_model0_model0_0.7_rf_simple.pdf}
    \end{subfigure}
    \hfill
    \begin{subfigure}[t]{0.45\textwidth}
        \centering
        \includegraphics[width=\textwidth]{./figures/fig3b_and_fig3b1_and_fig3a1/precise_imputation/model1/fig3b_precise_model1_model0_0.7_rf_simple.pdf}
        \caption{Compare methods in Model 1.}
        \label{fig:./figures/fig3b_and_fig3b1_and_fig3a1/precise_imputation/model1/fig3b_precise_model1_model0_0.7_rf_simple.pdf}
    \end{subfigure}

    \vspace{1em} 

    \begin{subfigure}[t]{0.45\textwidth}
        \centering
        \includegraphics[width=\textwidth]{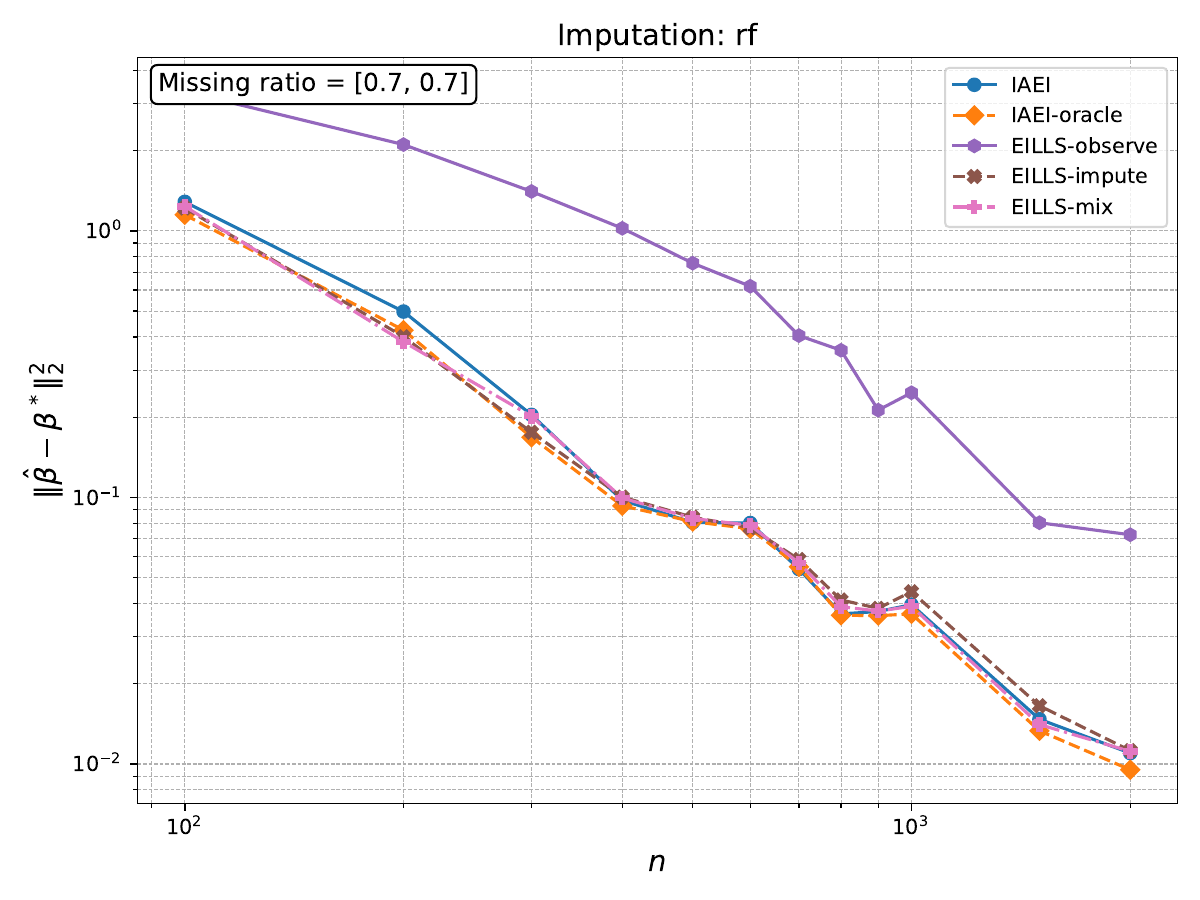}
        \caption{Compare methods in Model 2.}
        \label{fig:./figures/fig3b_and_fig3b1_and_fig3a1/precise_imputation/model2/fig3b_precise_model2_model0_0.7_rf_simple.pdf}
    \end{subfigure}
    \hfill
    \begin{subfigure}[t]{0.45\textwidth}
        \centering
        \includegraphics[width=\textwidth]{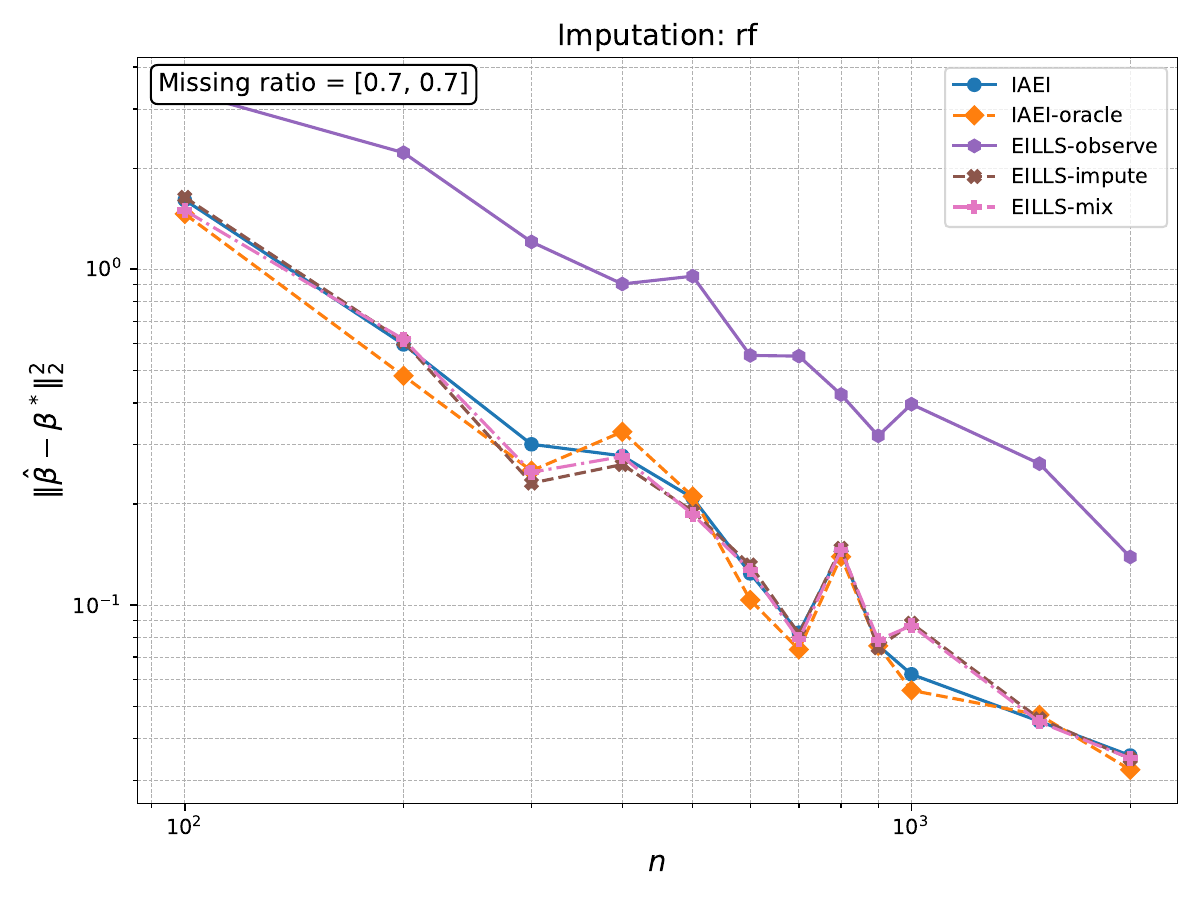}
        \caption{Compare methods in Model 3.}
        \label{fig:./figures/fig3b_and_fig3b1_and_fig3a1/precise_imputation/model3/fig3b_precise_model3_model0_0.7_rf_simple.pdf}
    \end{subfigure}

    \caption{Comapre methods across Models 0-3 with precise RandomForest imputation.}
    \label{fig:dd}
\end{figure}

While the results under precise imputation demonstrate the advantages of methods like EILLS-impute with linear imputation, the assumption of perfect precision and alignment with true environment distributions is often unrealistic in practice. In the following, we explore scenarios involving biased imputation and observe how the performance of EILLS-impute can deviate largely from the truth. To that end, we highlight the robustness of the proposed IAEI estimator under biased imputation strategies and analyze how variations in imputation accuracy affect its performance.

\begin{figure}[H]
    \centering
    \begin{subfigure}[t]{0.45\textwidth}
        \centering
        \includegraphics[width=\textwidth]{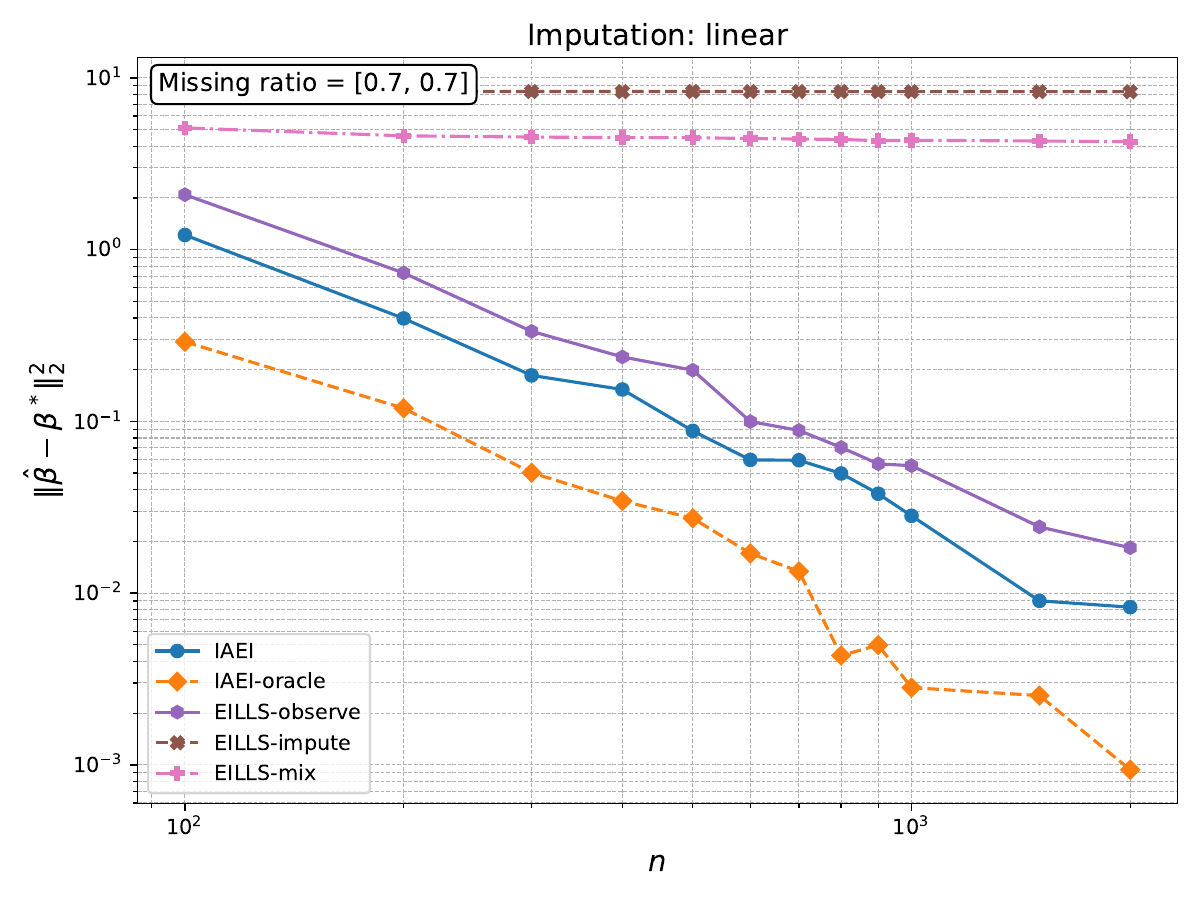}
        \caption{Compare methods in Model 0.}
        \label{fig:./figures/fig3b_and_fig3b1_and_fig3a1/bias_imputation/model0/fig3b_bias_model0_model0_0.7_linear_simple.pdf}
    \end{subfigure}
    \hfill
    \begin{subfigure}[t]{0.45\textwidth}
        \centering
        \includegraphics[width=\textwidth]{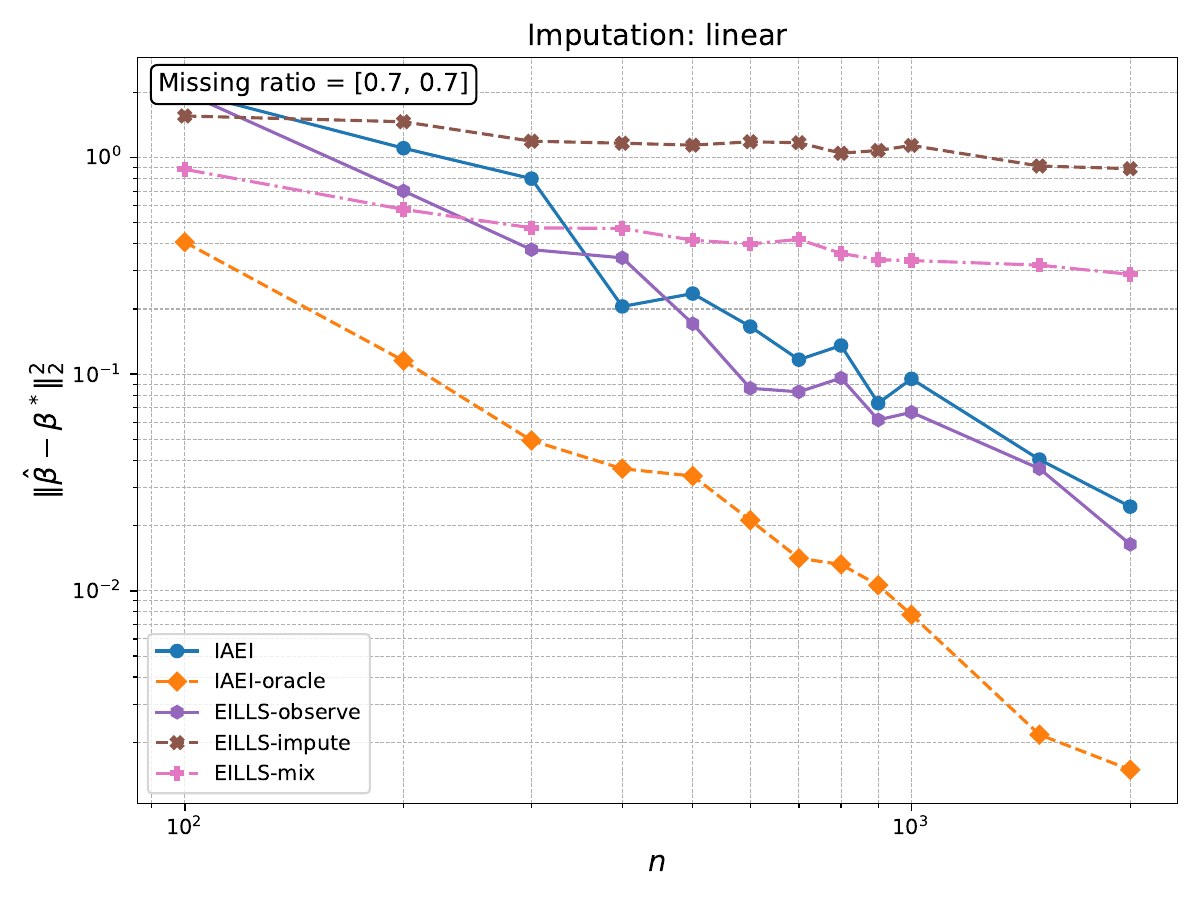}
        \caption{Compare methods in Model 1.}
        \label{fig:./figures/fig3b_and_fig3b1_and_fig3a1/bias_imputation/model1/fig3b_bias_model1_model0_0.7_linear_simple.pdf}
    \end{subfigure}

    \vspace{1em} 

    \begin{subfigure}[t]{0.45\textwidth}
        \centering
        \includegraphics[width=\textwidth]{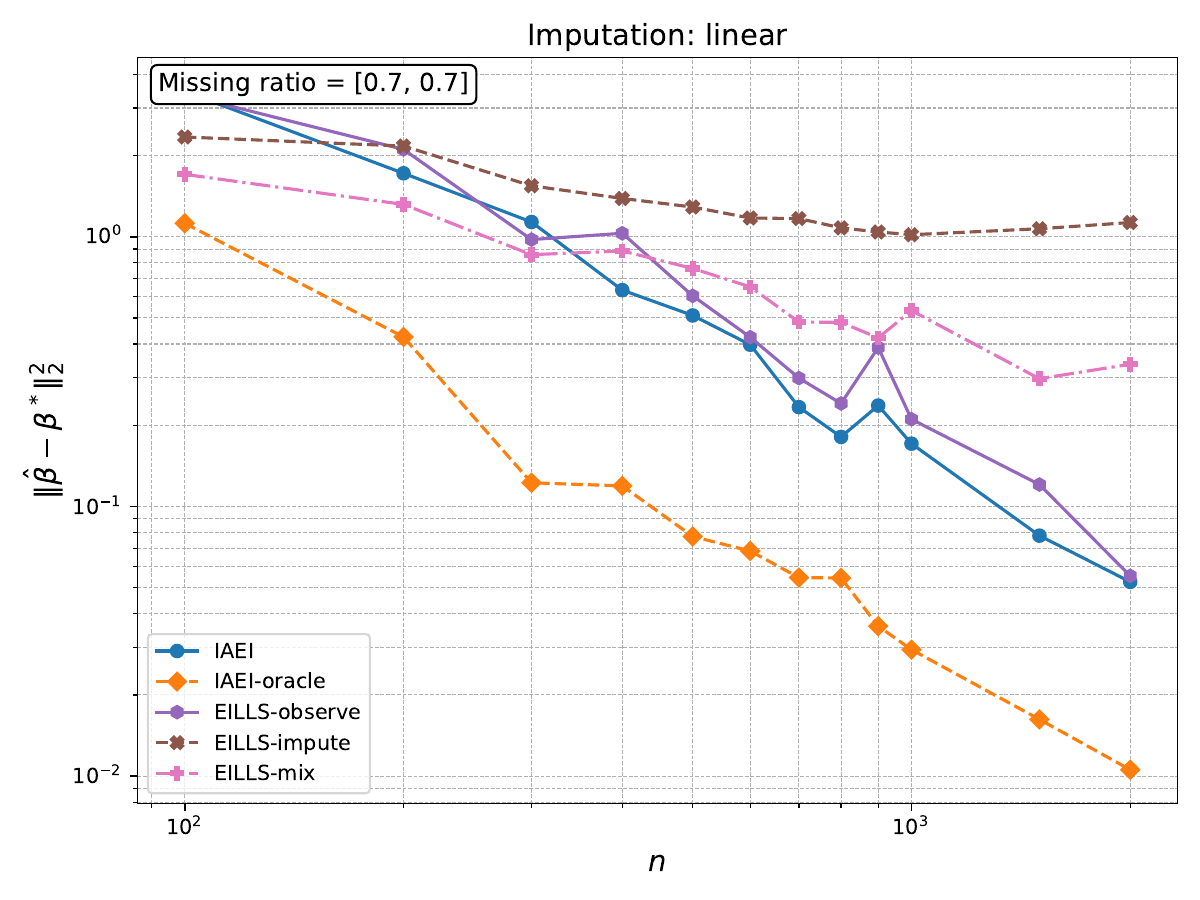}
        \caption{Compare methods in Model 2.}
        \label{fig:./figures/fig3b_and_fig3b1_and_fig3a1/bias_imputation/model2/fig3b_bias_model2_model0_0.7_linear_simple.pdf}
    \end{subfigure}
    \hfill
    \begin{subfigure}[t]{0.45\textwidth}
        \centering
        \includegraphics[width=\textwidth]{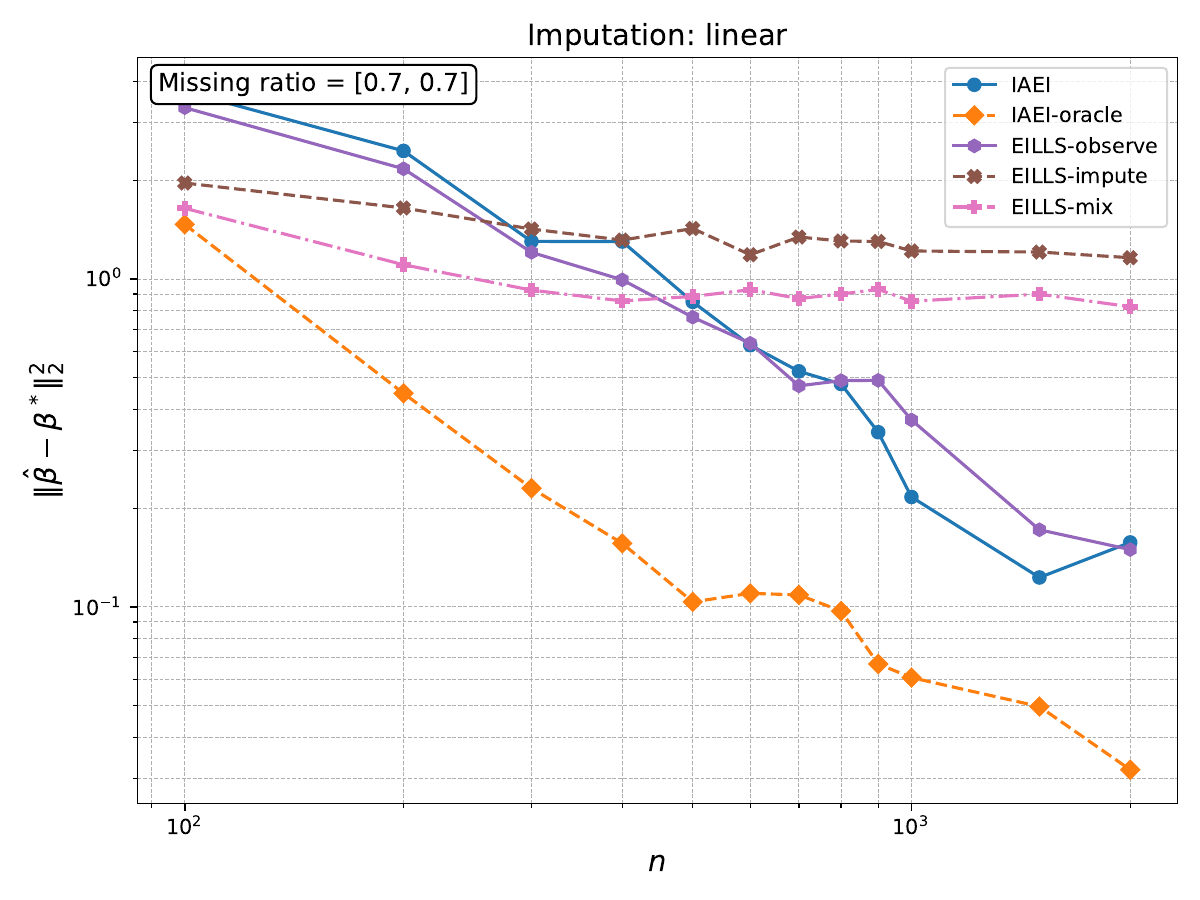}
        \caption{Compare methods in Model 3.}
        \label{fig:./figures/fig3b_and_fig3b1_and_fig3a1/bias_imputation/model3/fig3b_bias_model3_model0_0.7_linear_simple.pdf}
    \end{subfigure}

    \caption{While previously we observe that EILLS-impute and EILLS-mix with precise linear imputation outperform other methods, here we see that a biased imputation leads to a significant deterioration in their convergence to the truth $\boldsymbol{\beta}^*$.}
    \label{fig:fig:./figures/fig3b_and_fig3b1_and_fig3a1/bias_imputation/model2/fig3b_bias_model2_model0_0.7_linear_simple.pdffig:./figures/fig3b_and_fig3b1_and_fig3a1/bias_imputation/model3/fig3b_bias_model3_model0_0.7_linear_simple.pdf}
\end{figure}

Given the strong performance of EILLS-impute with accurate linear imputation, one might wonder if correcting the bias in imputation could improve its performance under biased settings. Figure \ref{fig:fig:./figures/fig3b_and_fig3b1_and_fig3a1/bias_imputation/model2/fig3b_bias_model2_model0_0.7_linear_simple.pdffig:./figures/fig3b_and_fig3b1_and_fig3a1/bias_imputation/model3/fig3b_bias_model3_model0_0.7_linear_simple.pdf} validate this intuition, demonstrating that IAEI with linear imputation outperforms EILLS-observe and the biased estimators EILLS-impute and EILLS-mix across Models 0–3. Building on this, we now focus on the second observation: incorporating nonlinear imputation significantly enhances the performance of IAEI, especially in more complex settings. The following plots under XGBoost imputation illustrate this improvement, while the results with RandomForest imputation, which show a similar pattern, are provided in the supplementary material. 

\begin{figure}[H]
    \centering
    \begin{subfigure}[t]{0.45\textwidth}
        \centering
        \includegraphics[width=\textwidth]{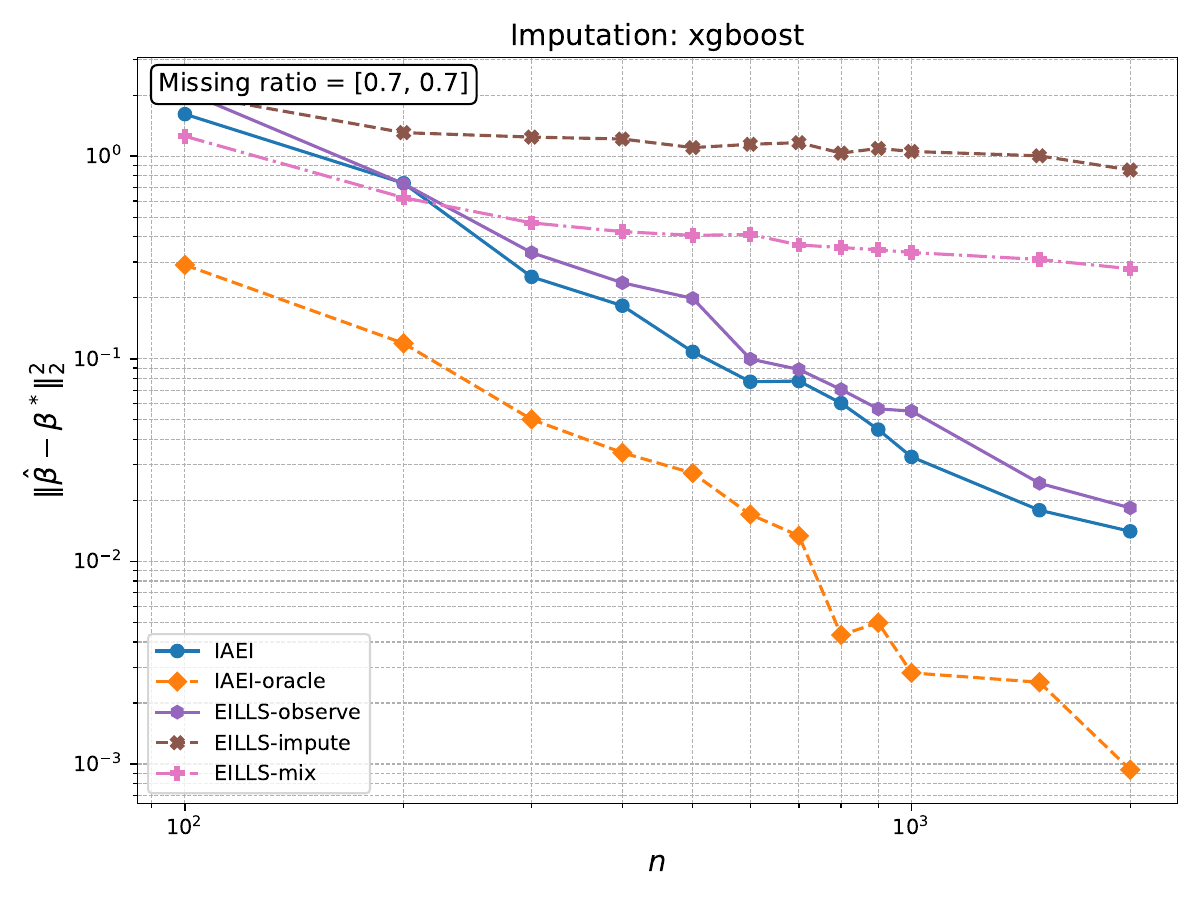}
        \caption{Compare methods in Model 0.}
        \label{fig:./figures/fig3b_and_fig3b1_and_fig3a1/bias_imputation/model0/fig3b_bias_model0_model0_0.7_xgboost_simple.pdf}
    \end{subfigure}
    \hfill
    \begin{subfigure}[t]{0.45\textwidth}
        \centering
        \includegraphics[width=\textwidth]{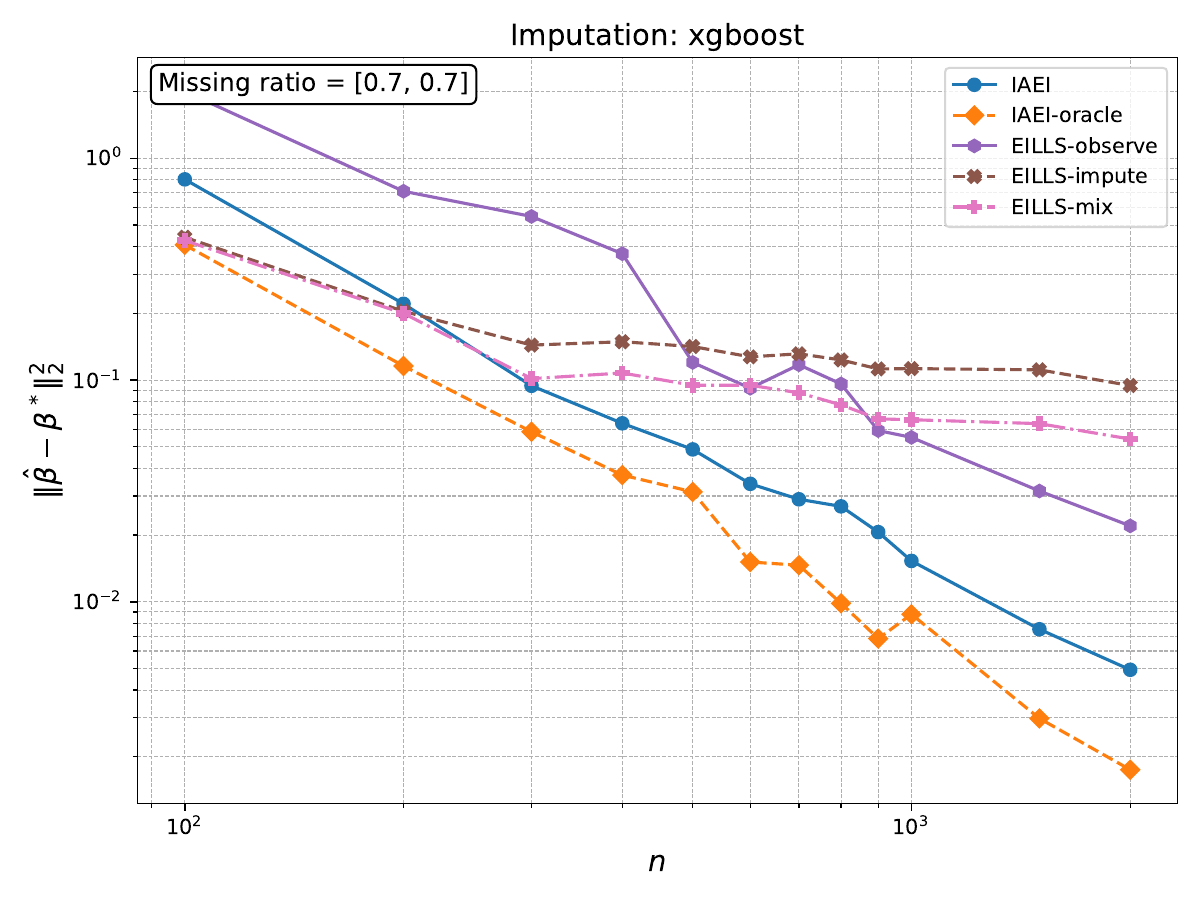}
        \caption{Compare methods in Model 1.}
        \label{fig:./figures/fig3b_and_fig3b1_and_fig3a1/bias_imputation/model1/fig3b_bias_model1_model0_0.7_xgboost_simple.pdf}
    \end{subfigure}

    \vspace{1em} 

    \begin{subfigure}[t]{0.45\textwidth}
        \centering
        \includegraphics[width=\textwidth]{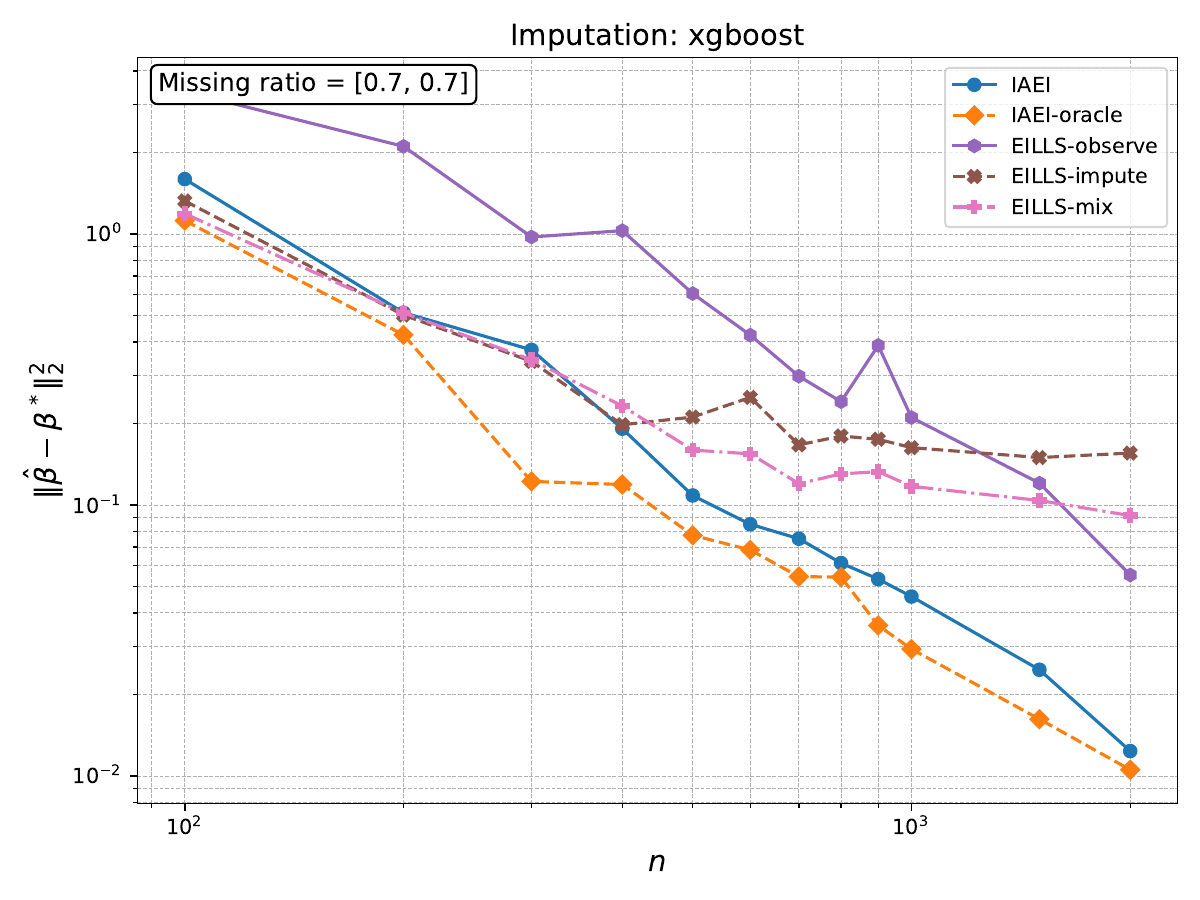}
        \caption{Compare methods in Model 2.}
        \label{fig:./figures/fig3b_and_fig3b1_and_fig3a1/bias_imputation/model2/fig3b_bias_model2_model0_0.7_xgboost_simple.pdf}
    \end{subfigure}
    \hfill
    \begin{subfigure}[t]{0.45\textwidth}
        \centering
        \includegraphics[width=\textwidth]{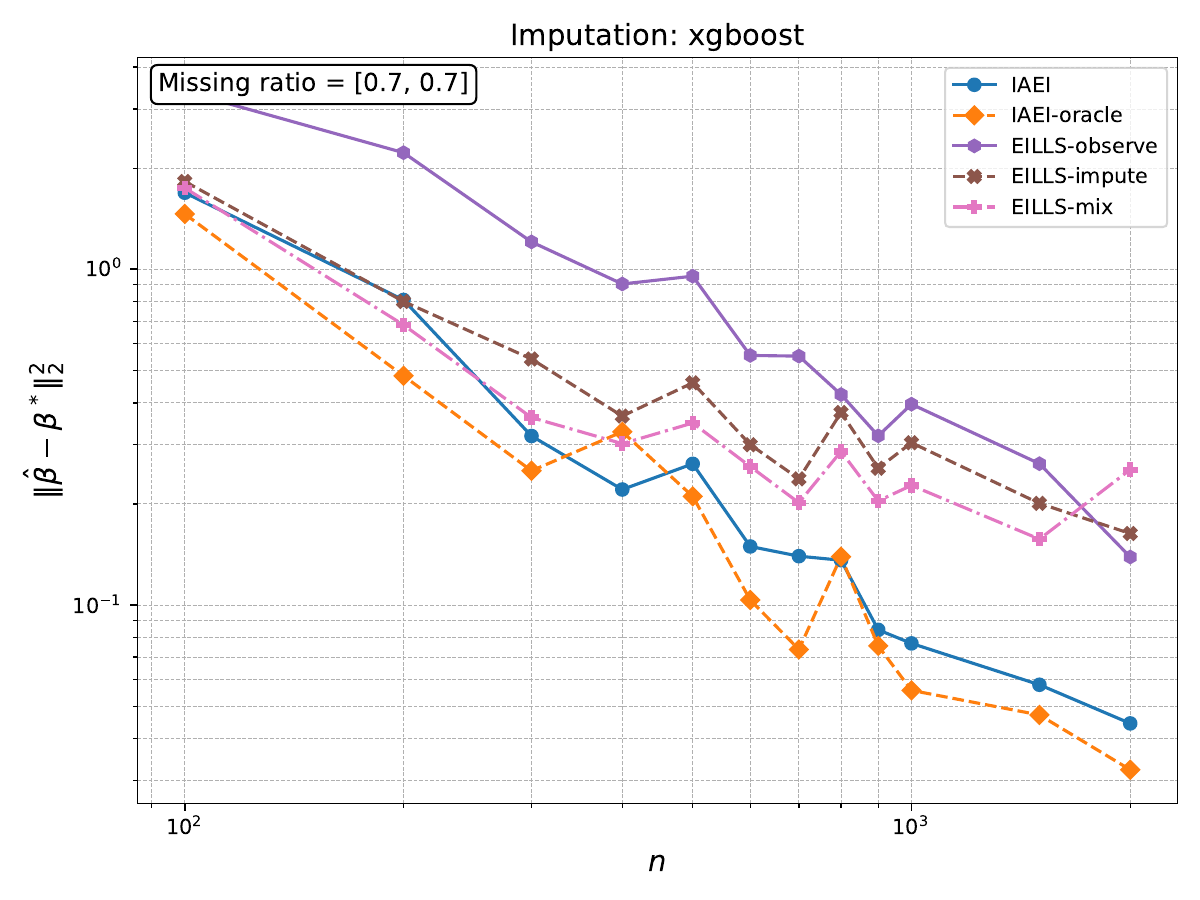}
        \caption{Compare methods in Model 3.}
        \label{fig:./figures/fig3b_and_fig3b1_and_fig3a1/bias_imputation/model3/fig3b_bias_model3_model0_0.7_xgboost_simple.pdf}
    \end{subfigure}
    \caption{In all Models 0–3, IAEI with XGBoost imputation consistently outperforms the other methods. Notably, in Models 1–3, its performance is significantly superior.}
    \label{fig:fig:./figures/fig3b_and_fig3b1_and_fig3a1/bias_imputation/model2/fig3b_bias_model2_model0_0.7_xgboost_simple.pdffig:./figures/fig3b_and_fig3b1_and_fig3a1/bias_imputation/model3/fig3b_bias_model3_model0_0.7_xgboost_simple.pdf}
\end{figure}

As expected, higher imputation bias results in a larger performance gap between the IAEI estimator and the oracle. Below, we compare the performance of methods under XGBoost bias and hbias imputation in Model 1.

\begin{figure}[H]
    \centering
    \begin{subfigure}[t]{0.45\textwidth}
        \centering
        \includegraphics[width=\textwidth]{./figures/fig3b_and_fig3b1_and_fig3a1/bias_imputation/model1/fig3b_bias_model1_model0_0.7_xgboost_simple.pdf}
        \caption{Methods with XGBoost bias imputation.}
        \label{fig:./figures/fig3b_and_fig3b1_and_fig3a1/bias_imputation/model1/fig3b_bias_model1_model0_0.7_xgboost_simple.pdf}
    \end{subfigure}
    \hfill
    \begin{subfigure}[t]{0.45\textwidth}
        \centering
        \includegraphics[width=\textwidth]{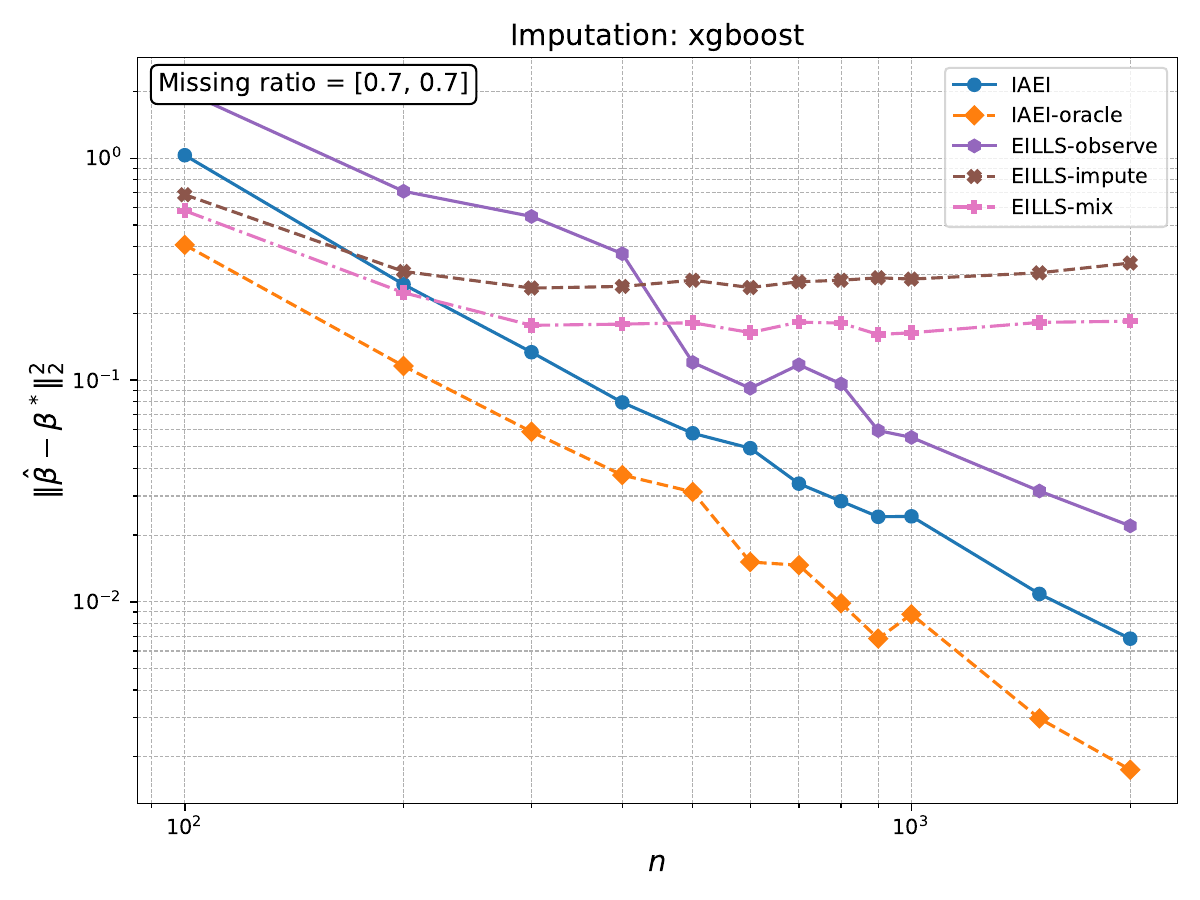}
        \caption{Methods with XGBoost hbias imputation.}
        \label{fig:./figures/fig3b_and_fig3b1_and_fig3a1/hbias_imputation/model1/fig3b_hbias_model1_model0_0.7_xgboost_simple.pdf}
    \end{subfigure}
    \caption{Methods dependent on imputation, such as EILLS-impute, EILLS-mix, and IAEI, show decreased performance as imputation bias increases.}
    \label{fig:./figures/fig3b_and_fig3b1_and_fig3a1/bias_imputation/model1/fig3b_bias_model1_model0_0.7_xgboost_simple.pdf./figures/fig3b_and_fig3b1_and_fig3a1/hbias_imputation/model1/fig3b_hbias_model1_model0_0.7_xgboost_simple.pdf}
\end{figure}

Next, we explicitly demonstrate the benefits of employing a nonlinear imputation model in complex scenarios and its impact on the performance of the IAEI estimator.

\begin{figure}[H]
    \centering
    \begin{subfigure}[t]{0.45\textwidth}
        \centering
        \includegraphics[width=\textwidth]{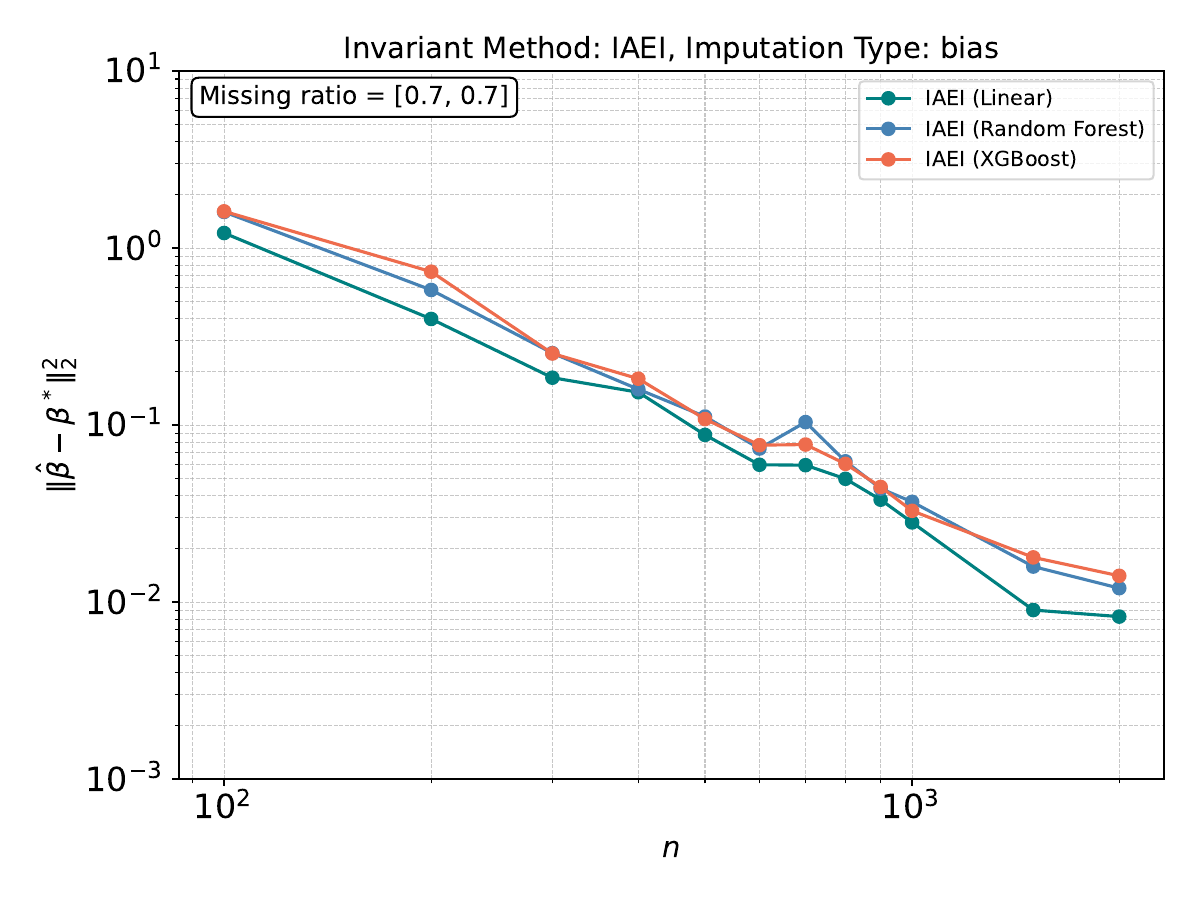}
        \caption{IAEI with different imputation methods in Model 0.}
        \label{fig:./figures/fig3b_and_fig3b1_and_fig3a1/bias_imputation/model0/fig3b1_bias_model0_model0_0.7_IAEI.pdf}
    \end{subfigure}
    \hfill
    \begin{subfigure}[t]{0.45\textwidth}
        \centering
        \includegraphics[width=\textwidth]{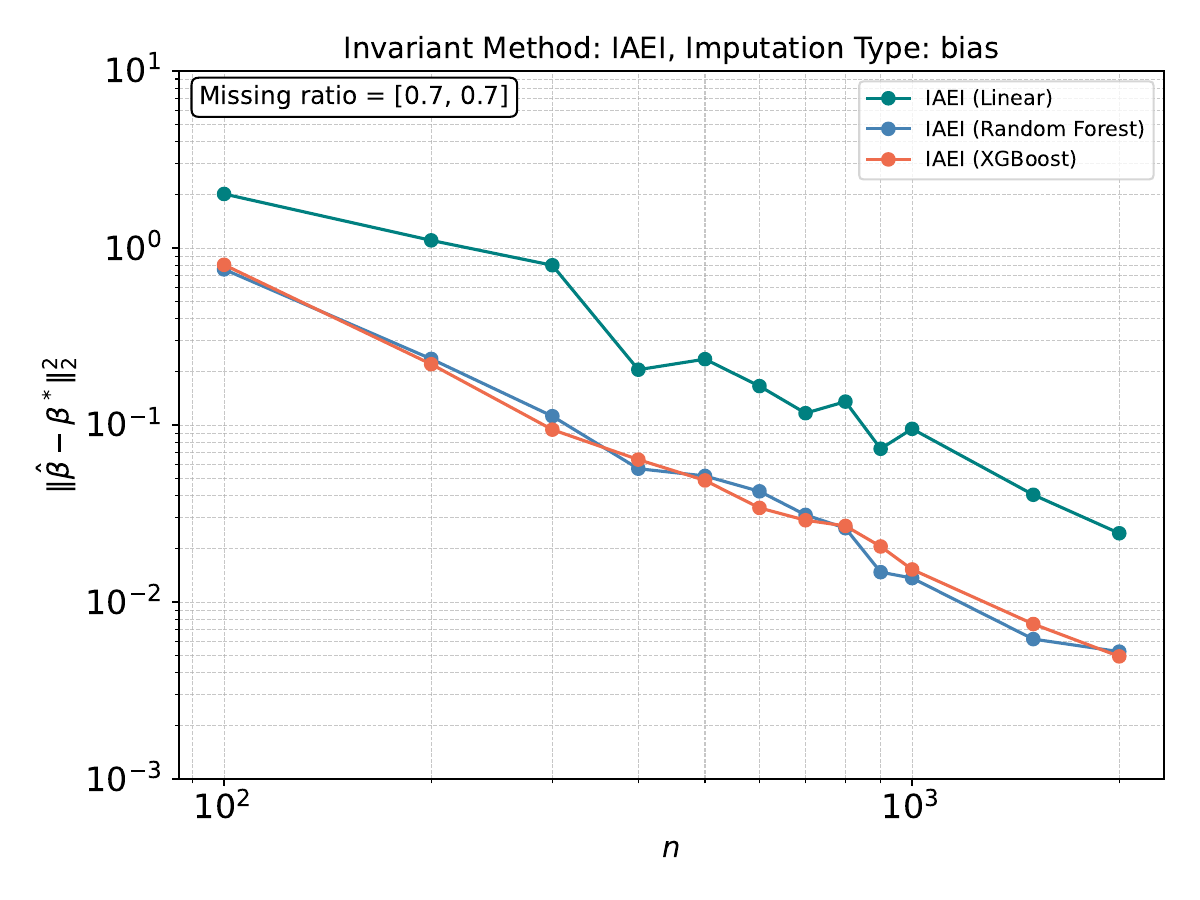}
        \caption{IAEI with different imputation methods in Model 1.}
        \label{fig:./figures/fig3b_and_fig3b1_and_fig3a1/bias_imputation/model1/fig3b1_bias_model1_model0_0.7_IAEI.pdf}
    \end{subfigure}

    \vspace{1em} 

    \begin{subfigure}[t]{0.45\textwidth}
        \centering
        \includegraphics[width=\textwidth]{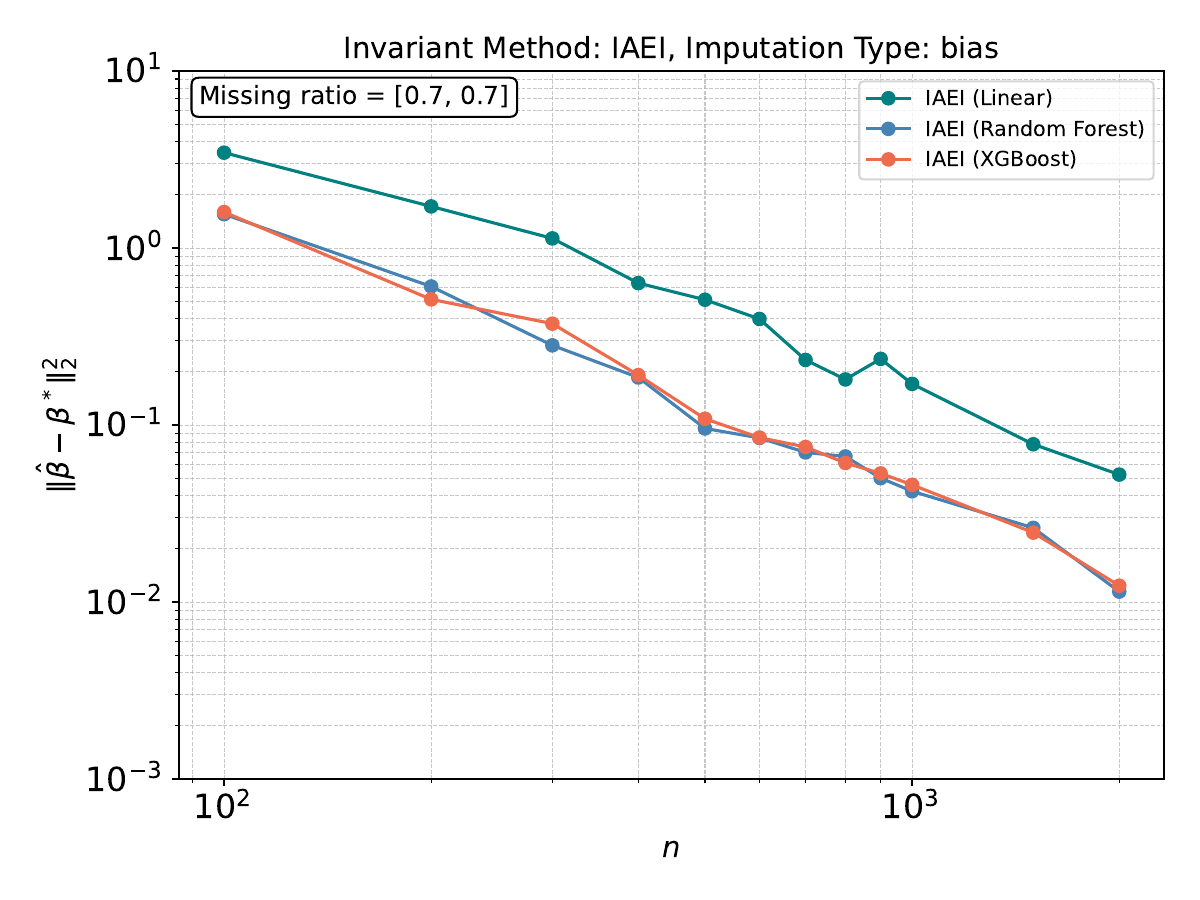}
        \caption{IAEI with different imputation methods in Model 2.}
        \label{fig:./figures/fig3b_and_fig3b1_and_fig3a1/bias_imputation/model2/fig3b1_bias_model2_model0_0.7_IAEI.pdf}
    \end{subfigure}
    \hfill
    \begin{subfigure}[t]{0.45\textwidth}
        \centering
        \includegraphics[width=\textwidth]{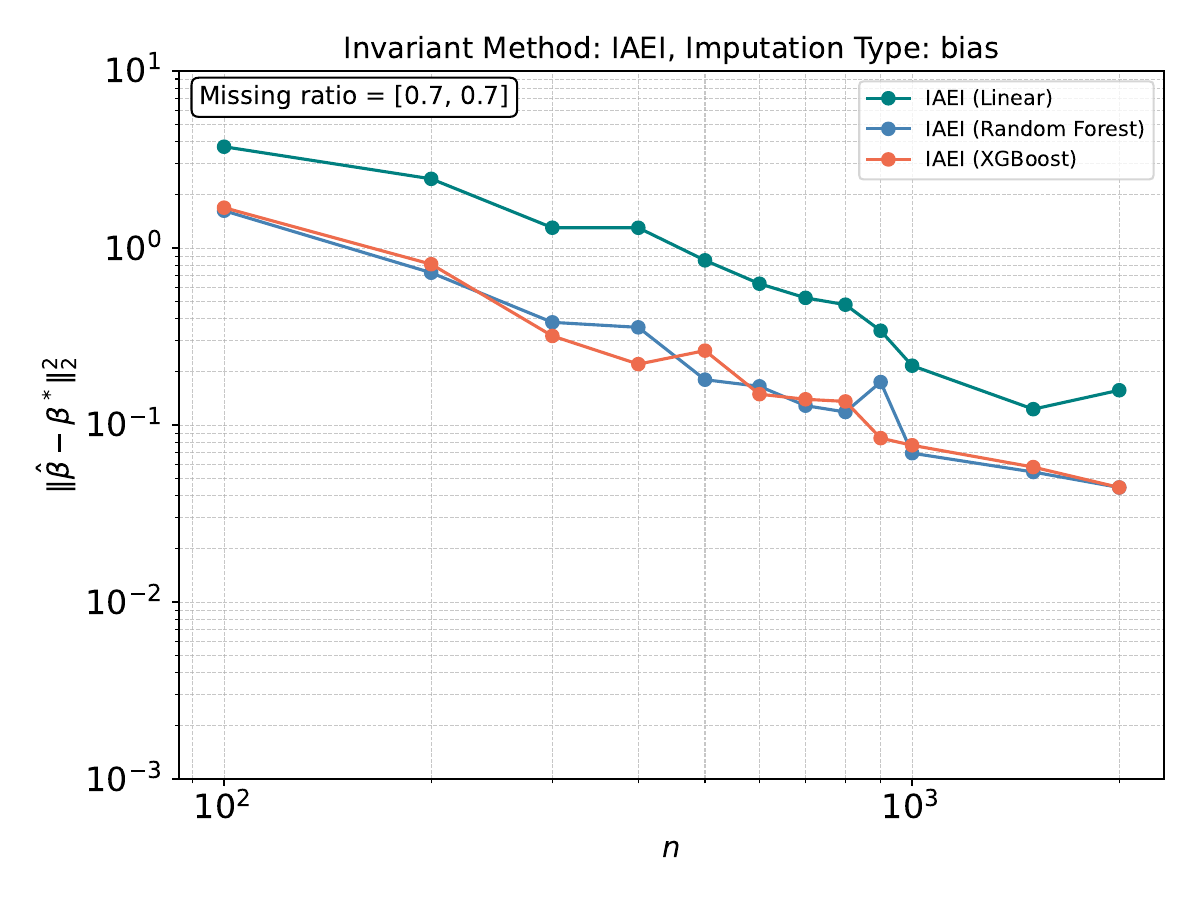}
        \caption{IAEI with different imputation methods in Model 3.}
        \label{fig:./figures/fig3b_and_fig3b1_and_fig3a1/bias_imputation/model3/fig3b1_bias_model3_model0_0.7_IAEI.pdf}
    \end{subfigure}
    \caption{Plots shows IAEI with nonlinear imputation consistently outperforms the linear methods espeically in complex scenarios Model 1-3.}
    \label{fig:./figures/fig3b_and_fig3b1_and_fig3a1/bias_imputation/model3/fig3b1_bias_model3_model0_0.7_IAEI.pdf./figures/fig3b_and_fig3b1_and_fig3a1/bias_imputation/model2/fig3b1_bias_model2_model0_0.7_IAEI.pdf}
\end{figure}

\section{Real Data Analysis}\label{section:real_data_analysis}

\subsection{Bike Sharing Dataset}

We evaluate the predictive performance of the IAEI estimator compared to naive estimators using the Bike Sharing Dataset, available from the UCI Machine Learning Repository \citep{bike_sharing_275}. This dataset contains 17,389 hourly records of bike rentals from the Capital Bikeshare system in Washington, D.C., spanning the years 2011 and 2012. The objective is to predict daily bike rental counts based on weather data. To introduce missing data, we randomly omit $85\%$ of the outcomes from the year 2012 records. An imputation model is trained using data from year 2011. 

For the year 2012 data, we divide it into 12 subsets, each corresponding to a specific month from January to December. In each fold, the estimator is trained on data from 11 months and evaluated on the remaining month. Predictions are made for each hourly bike count, and the mean squared error (MSE) is computed at the daily level by averaging the errors across all hours within a day. The MSE for each day is then averaged across all 12 folds, yielding 31 final daily MSE values. This process is repeated for a range of $\gamma$ hyperparameter values. This process is repeated across a range of $\gamma$ values. For each day, we select the $\gamma$ that yields the lowest MSE in predicting bike rental counts. 

\begin{figure}[H]
    \centering
    \includegraphics[width=0.8\textwidth]{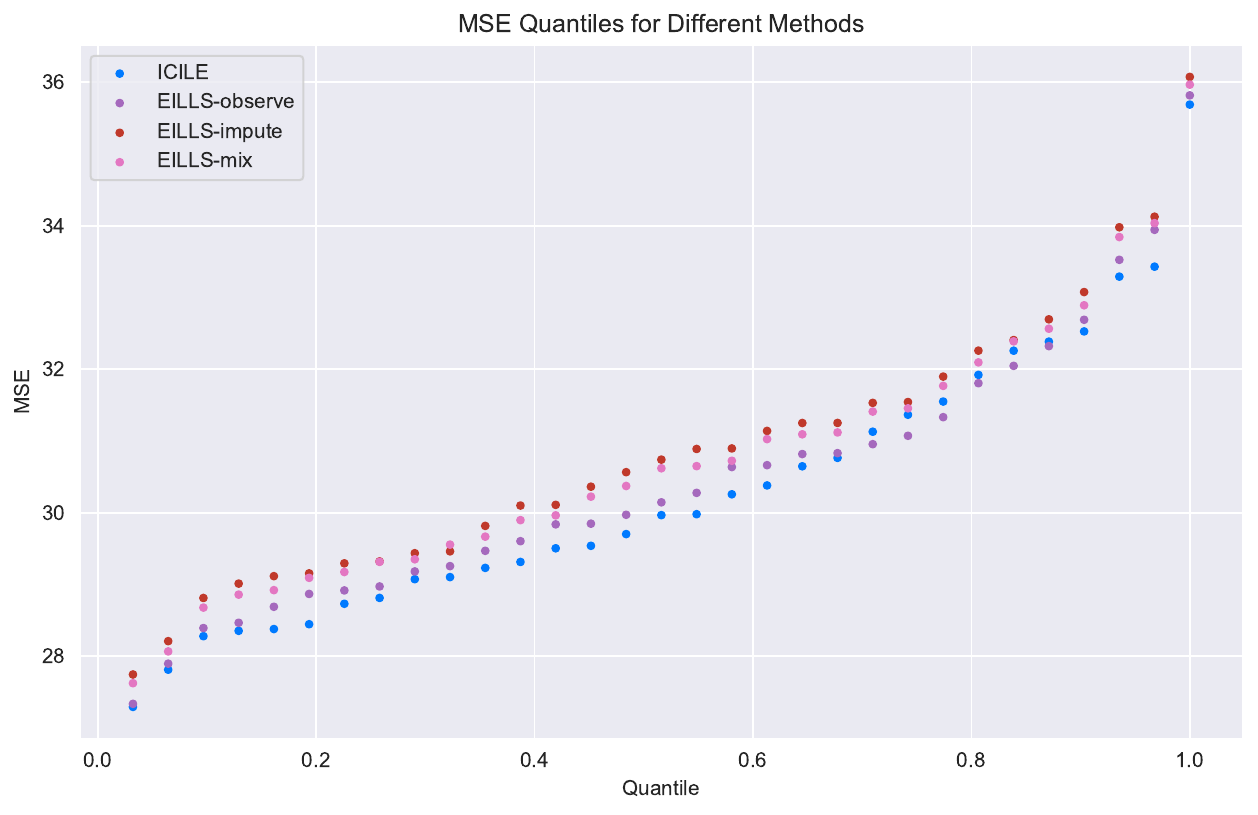}
    \caption{Plot the quantiles of the conditional mean squared error of the predicted bike rental counts for each day with the optimal gamma.}
    \label{fig:bike_mse}
\end{figure}

This procedure is applied to all methods—IAEI, EILLS-observe, EILLS-impute, and EILLS-mix—resulting in 31 MSE values per method, each computed as the average across 12 folds using the optimal $\gamma$. We rank the MSE values for each method from lowest to highest and compute their empirical quantiles. In Figure \ref{fig:bike_mse}, the y-axis represents these ordered MSE values, while the x-axis denotes their normalized rank, corresponding to their quantile position among the 31 days.

Figure \ref{fig:bike_mse} shows that the IAEI estimator achieves lower MSE than EILLS-observe, EILLS-impute, and EILLS-mix, especially at lower quantiles, demonstrating greater stability and accuracy on well-predicted days. As the quantile increases, IAEI maintains an advantage over EILLS-impute and EILLS-mix, but its gap with EILLS-observe narrows, indicating that prediction becomes more challenging, reducing performance differences across methods. 

\begin{figure}[H]
    \centering
    \includegraphics[width=0.8\textwidth]{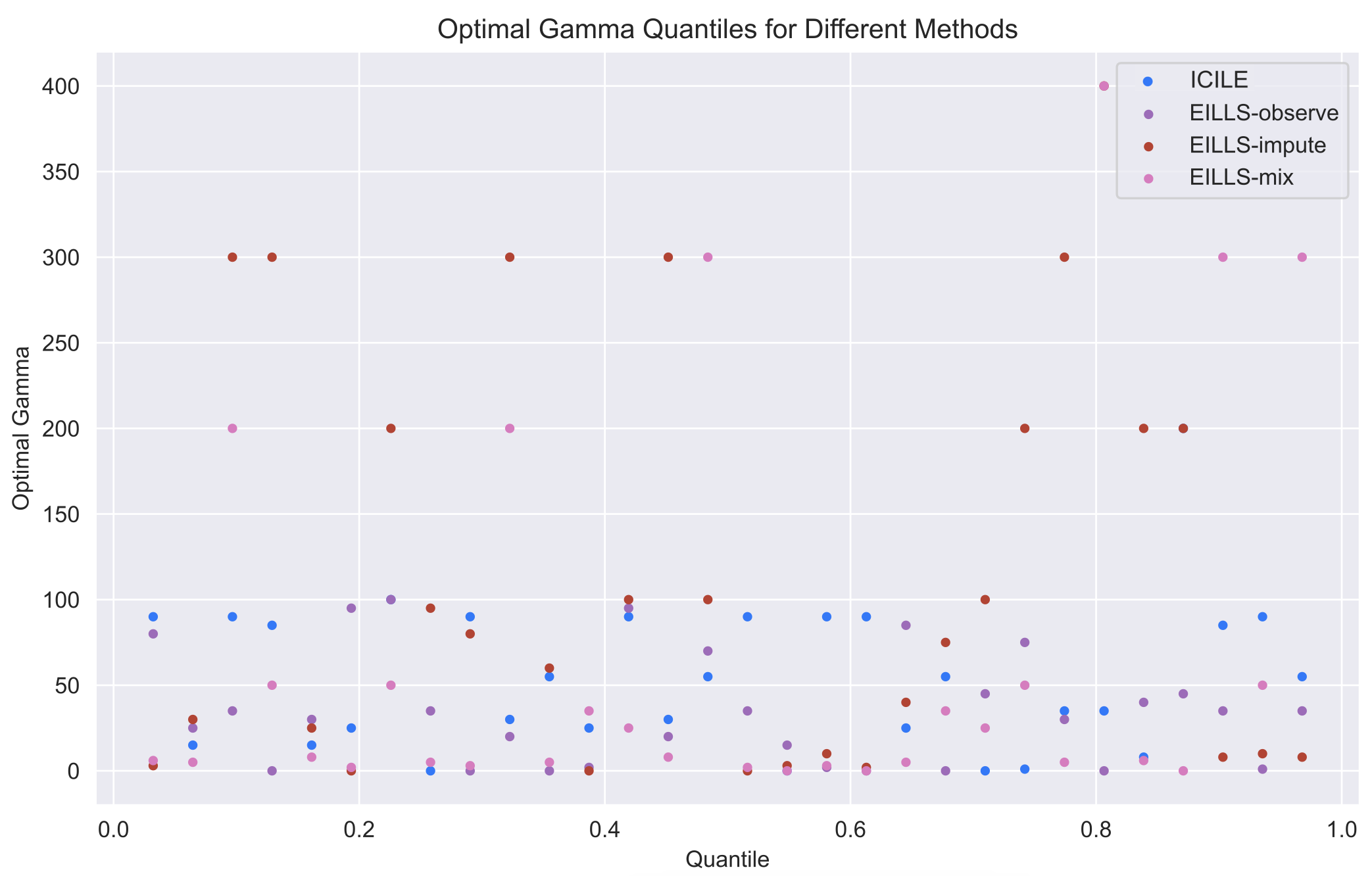}
    \caption{Plot of the optimal gamma.}
    \label{fig:bike_gamma}
\end{figure}

Additionally, we plot the optimal gamma values chosen for each day. As shown in Figure \ref{fig:bike_gamma}, IAEI selects gamma values within a similar range as EILLS-observe, typically between $0$ and $100$. In contrast, EILLS-impute and EILLS-mix, which operate with biased imputed outcomes, occasionally rely on higher gamma values, leading to overall instability in gamma selection.

\section{Conclusion}\label{section:conclusion}

In this paper, we tackle the challenge of learning invariant predictors in the presence of missing labels by introducing an estimator that effectively integrates both observed and imputed outcomes. We incorporate imputation into the invariance learning framework, allowing for a potentially biased imputation model, and establish sufficient conditions for achieving the variable selection property while controlling the $\ell_2$ error bound in the estimation of $\boldsymbol{\beta}^*$.

We consider the invariance learning setup, where researchers aim for both reliable prediction and robust attribution. The goal is not only to ensure predictive accuracy on an unseen environment but also to identify the key variables that drive invariant relationships with the outcome. This motivates the assumption that the true invariant relationship between the outcome of interest and a key set of variables is linear. At the same time, it is beneficial to allow for flexible or complex SEMs, where covariates may be influenced by others in a nonlinear way. The more complex the SEM, the more likely it is that the true regression function within each environment exhibits nonlinearity.  In such cases, when working with a linear model, leveraging the marginal distribution of covariates can improve the inference of the best linear predictor. Recognizing this opportunity, we address the challenge of large amounts of missing outcome data in practice. Rather than discarding imperfect data, we allow the use of a potentially biased imputation model and develop a method that carefully integrates imputation into the invariance learning framework, leveraging both observed and imputed outcomes to improve variable selection and estimation.

In our simulation study, our method demonstrates strong performance across various data settings. When the imtputation bias is within a reasonable range, our estimator consistently outperforms other methods across all scenarios. In complex settings with high nonlinearity, using a nonlinear imputation model greatly improves performance. For variable selection in such cases, adding enhanced penalty terms is beneficial as it better captures nonlinear spurious relationships. However, for estimation, the original penalty terms are preferable, as the enhanced version increases variability, leading to higher MSE.

We also apply our method to a real-world bike-sharing dataset, using 2011 data to train the imputation model. Our analysis reveals that simply filling in missing labels with imputed values, derived from a model trained on 2011 data, introduces bias and results in higher MSE across all 31 days. Additionally, our estimator achieves the lowest MSE at lower quantiles, indicating superior performance on well-predicted days, while at higher quantiles, its performance aligns with that of the method using only gold-standard data.

While the current work focuses on leveraging missing outcome data under the assumption of missing completely at random (MCAR), in many practical settings, missingness may depend on observed covariates or even unmeasured variables. A natural extension would be to develop more robust approaches that account for missing at random (MAR) and missing not at random (MNAR) mechanisms. Additionally, our method assumes a linear invariant relationship with a continuous outcome. Future work could extend this framework to generalized linear models, enabling its application to binary or categorical outcomes.

\section*{Acknowledgements}
The author would like to thank Prof.~Bradic for early discussions and initial guidance on this project. The author also thanks Yihong Gu for helpful discussions related to invariance learning.

\newpage
\bibliographystyle{apalike}
\bibliography{reference}

\newpage
\textbf{\Large Supplementary Materials}

\appendix

\begin{center} {\large Summary} \end{center}
~~~~~~~~The Supplementary Material contains the following:
\begin{enumerate}
	\item Additional Notation Details
	\item Preliminaries
	\item Technical Lemmas
	\item Proofs for Non-asymmptotic Results
	      \begin{enumerate}
		      \item Proof of Lemma \ref{lemma:two_sided_bound_R}
		      \item Proof of Lemma \ref{lemma:one_sided_bound_J}
		      \item Proof of Corollary \ref{corollary:oneside_boundJ_high_missing_good_imp}
		      \item Proof of Corollary \ref{corollary:oneside_boundJ_high_missing_imprecise_imp}
		      \item Proof of Theorem \ref{theorem:nonasy_vsc}
		      \item Proof of Theorem \ref{theorem:nonasy_l2_error}
	      \end{enumerate}
	\item Additional Simulation Results
		  \begin{enumerate}
			\item Performance on Variable Selection
			\item Performance on $\ell_2$ Error Convergence
		  \end{enumerate}
\end{enumerate}

\newpage


\section{Additional Notation Details}\label{section:more_notation}

In this section, we provide detailed expressions for the sample size notations introduced in the main text, along with those used in the proof. Recall that we previously defined two general formulas, 
\begin{equation}\label{eq:general_samplesize_supplementary}
	\mathcal{N}(k,g,h,f) = \min_{e\in\mathcal{E}} \frac{k(N^{(e)}, n^{(e)}, m^{(e)})}{g(\omega^{(e)}) h(\widehat{\tau}^{(e)}) f(|\eta^{(e)}|)} \quad \text{and}\quad \mathcal{G}(k,g,h,f) = \left(\sum_{e\in\mathcal{E}}  \frac{g(\omega^{(e)})h(\widehat{\tau}^{(e)})f(|\eta^{(e)}|)}{k(N^{(e)}, n^{(e)}, m^{(e)})}    \right)^{-1},
\end{equation}
which encapsulate various sample size notations. The specific choices of functions in $\mathcal{N}(k,g,h,f)$ determine the different sample size definitions. Tables~\ref{tab:sample_size_min}, \ref{tab:sample_size_sum}, and \ref{tab:sample_size_sum_more} together provide a structured summary of these expressions.

\begin{table}[H]
    \centering
    \renewcommand{\arraystretch}{0.5}
	\captionsetup{position=bottom}
    \begin{tabular}{lccccccc} 
        \toprule
         Sample size &  Full Expression& Function & $k(x,y,z)$ &  $g(x)$ & $h(y)$ &  $f(z)$ \\ 
        \midrule
        $N_{\min}$  &  $\displaystyle\min _{e \in \mathcal{E}} N^{(e)}$  &$\mathcal{N}$ & $x$  &  $1$&  $1$& $1$ \\  
        $n_{\min}$  & $\displaystyle\min_{e \in \mathcal{E}} n^{(e)}$   &$ \mathcal{N}$ & $y$ & $1$ &  $1$& $1$ \\  
        $m_{\min}$  & $\displaystyle\min_{e \in \mathcal{E}} m^{(e)}$   &$\mathcal{N}$ &  $z$ & $1$ & $1$ &$1$  \\  
        $N_{*}$ & $\displaystyle\min _{e \in \mathcal{E}} \frac{N^{(e)}}{\omega^{(e)}}$    &$\mathcal{N}$ & $x$  &  $x$ &$1$  &$1$  \\  
        $n_{*}$  &  $\displaystyle\min _{e \in \mathcal{E}} \frac{n^{(e)}}{\omega^{(e)}}$   &$ \mathcal{N}$ &  $y$ &  $x$ &  $1$&$1$  \\  
        $m_{*}$ &  $\displaystyle\min _{e \in \mathcal{E}} \frac{m^{(e)}}{\omega^{(e)}}$   &$ \mathcal{N}$ & $z$ & $x$  & $1$ &$1 $ \\  
        $n_*^{\widehat\tau}$  &  $\displaystyle \min _{e \in \mathcal{E}} \frac{n^{(e)}}{\omega^{(e)}\widehat\tau^{(e)}}$  & $\mathcal{N}$  & $y$  &  $x$&  $y$ & $1$ \\  
        $m_*^{\widehat\tau}$  & $\displaystyle \min _{e \in \mathcal{E}} \frac{m^{(e)}}{\omega^{(e)}\widehat\tau^{(e)}}$   & $\mathcal{N}$  & $ z$ &  $x$&  $y$ &  $1$\\  
        $n^{\widehat\tau^2}_*$  & $\displaystyle \min _{e \in \mathcal{E}} \frac{n^{(e)}}{\omega^{(e)}(\widehat\tau^{(e)})^2}$   & $\mathcal{N}$  &  $y$&  $x$&  $y^2$& $1$ \\  
        $m^{\widehat\tau^2}_*$ & $\displaystyle \min _{e \in \mathcal{E}} \frac{m^{(e)}}{\omega^{(e)}(\widehat\tau^{(e)})^2}$   & $\mathcal{N}$  & $z$ & $ x$ &  $y^2$& $1$ \\  
        $ n_*^{\widehat\tau,|\eta|}$ &$\displaystyle \min _{e \in \mathcal{E}} \frac{n^{(e)}}{\omega^{(e)}\widehat\tau^{(e)}|\eta^{(e)}|}$ &  $\mathcal{N}$& $y$ &$x$  &$y$&$z$ \\ 
         $m_*^{\widehat\tau,|\eta|}$& $\displaystyle \min _{e \in \mathcal{E}} \frac{m^{(e)}}{\omega^{(e)}\widehat{\tau}^{(e)}|\eta^{(e)}|} $& $\mathcal{N}$ &$z$  &$x$ &$y$&$z$ \\ 
         $n_*^{\widehat\tau^2,|\eta|^2}$ & $\displaystyle \min _{e \in \mathcal{E}} \frac{n^{(e)}}{\omega^{(e)}(\widehat\tau^{(e)})^2|\eta^{(e)}|^2}$ & $\mathcal{N}$ &$y$ & $x$ & $y^2$&$z^2$ \\
         $m_*^{\widehat\tau^2,|\eta|^2}$ &$\displaystyle \min _{e \in \mathcal{E}} \frac{m^{(e)}}{\omega^{(e)}(\widehat\tau^{(e)})^2|\eta^{(e)}|^2}$  & $\mathcal{N}$ &$z$ & $x$ &  $y^2$& $z^2$\\
        \bottomrule
    \end{tabular}
    \caption{Summary of sample size notations based on minimization across environments. Each notation follows the general function $\mathcal{N}(k,g,h,f)$, where $k,g,h,f$ are functions that determine how sample sizes are scaled and adjusted based on environment weights $\omega^{(e)}$, missing ratio $\widehat{\tau}^{(e)}$, and absolute imputation bias $|\eta^{(e)}|$}
    \label{tab:sample_size_min}
\end{table}

\begin{table}[H]
    \centering
    \renewcommand{\arraystretch}{0.5}
    \begin{tabular}{lccccccc} 
        \toprule
         Sample size &  Full Expression& Function & $k(x,y,z)$ &  $g(x)$ & $h(y)$ &  $f(z)$ \\ 
        \midrule
         $\bar{N}$ & $\displaystyle\left(\sum_{e \in \mathcal{E}} \frac{\omega^{(e)}}{N^{(e)}}\right)^{-1} $ &$\mathcal{G}(k,g,h,f)$  & $x$ & $x$ & 1& 1\\
         $\bar{n}$ & $\displaystyle\left(\sum_{e \in \mathcal{E}} \frac{\omega^{(e)}}{n^{(e)}}\right)^{-1} $  &  $\mathcal{G}(k,g,h,f)$& $y$  &$x$ &1&1 \\ 
          $\bar{m}$&  $\displaystyle\left(\sum_{e \in \mathcal{E}} \frac{\omega^{(e)}}{m^{(e)}}\right)^{-1} $ &  $\mathcal{G}(k,g,h,f)$& $z$ & $x$ &1&1 \\ 
         $ \bar{N}^{|\eta|^2}$ & $\displaystyle \left( \sum_{e \in \mathcal{E}} \frac{\omega^{(e)}|\eta^{(e)}|^2}{N^{(e)}}\right)^{-1}$ & $\mathcal{G}(k,g,h,f)$   & $x$& $x$ & $1$ & $z^2$\\
         $\bar{n}^{|\eta|^2}$ &  $\displaystyle \left(\sum_{e \in \mathcal{E}} \frac{\omega^{(e)}|\eta^{(e)}|^2}{n^{(e)}}\right)^{-1}$ & $\mathcal{G}(k,g,h,f)$    & $y$ &  $x$ &$1$  &$z^2$ \\
         $\bar{n}^{\widehat\tau}$ & $\displaystyle \left(\sum_{e \in \mathcal{E}} \frac{\omega^{(e)}\widehat\tau^{(e)}}{n^{(e)}}\right)^{-1} $ & $\mathcal{G}(k,g,h,f)$  &  $y$ & $x$ & $y$ &$1$ \\
         $\bar{m}^{\widehat\tau}$ & $\displaystyle \left(\sum_{e \in \mathcal{E}} \frac{\omega^{(e)}\widehat\tau^{(e)}}{m^{(e)}}\right)^{-1}$ & $\mathcal{G}(k,g,h,f)$  & $z$ & $x$ & $y$  &$1$  \\ 
         $\bar{n}^{\widehat\tau,|\eta|}$  & $\displaystyle \left(\sum_{e \in \mathcal{E}} \frac{\omega^{(e)}\widehat\tau^{(e)}|\eta^{(e)}|}{n^{(e)}}\right)^{-1}$ & $\mathcal{G}(k,g,h,f)$ & $y$&$x$  & $y$&$z$ \\
        $\bar{m}^{\widehat\tau,|\eta|}$  & $\displaystyle\left(\sum_{e \in \mathcal{E}} \frac{\omega^{(e)}\widehat\tau^{(e)}|\eta^{(e)}|}{m^{(e)}}\right)^{-1}$ & $\mathcal{G}(k,g,h,f)$ &$z$& $x$ &$y$ &$z$ \\ 
         $\bar{ n}^{\widehat\tau^2, |\eta|}$ & $\displaystyle \left(\sum_{e \in \mathcal{E}}\frac{\omega^{(e)}( \widehat\tau^{(e)})^2|\eta^{(e)}| }{n^{(e)}}\right)^{-1}$ &$\mathcal{G}(k,g,h,f)$  &  $y$ & $x$&$y^2$ & $z$ \\
        $\bar{m}^{\widehat\tau^2,|\eta|}$  & $\displaystyle \left(\sum_{e \in \mathcal{E}}\frac{\omega^{(e)}( \widehat\tau^{(e)})^2|\eta^{(e)}| }{m^{(e)}}\right)^{-1}$ &$\mathcal{G}(k,g,h,f)$  &$z$  &$x$  & $y^2$& $z$\\
        $\overline{\sqrt{nN}}$ & $\displaystyle\left(\sum_{e \in \mathcal{E}} \frac{\omega^{(e)} }{\sqrt{n^{(e)} N^{(e)}}}\right)^{-1}$ & $\mathcal{G}(k,g,h,f)$  & $\sqrt{xy}$& $x$ &$1$ &$1$ \\ 
         $\overline{\sqrt{mN}}$ & $\displaystyle \left(\sum_{e \in \mathcal{E}} \frac{\omega^{(e)} }{\sqrt{m^{(e)} N^{(e)}}}\right)^{-1}$ & $\mathcal{G}(k,g,h,f)$  &$\sqrt{xz}$ &  $x$&$1$ &$1$ \\ 
         $\overline{\sqrt{nN}}^{\widehat\tau}$ &  $\displaystyle \left(\sum_{e \in \mathcal{E}} \frac{\omega^{(e)} \widehat{\tau}^{(e)}}{\sqrt{n^{(e)} N^{(e)}}}\right)^{-1}$ & $\mathcal{G}(k,g,h,f)$  & $\sqrt{xy}$& $x$& $y$ & $1$\\
         $\overline{\sqrt{mN}}^{\widehat\tau}$ &  $\displaystyle \left(\sum_{e \in \mathcal{E}} \frac{\omega^{(e)} \widehat{\tau}^{(e)}}{\sqrt{m^{(e)} N^{(e)}}}\right)^{-1}$&  $\mathcal{G}(k,g,h,f)$ & $\sqrt{xz}$ & $x$ & $y$ &$1$ \\
        \bottomrule
    \end{tabular}
    \caption{Summary of sample size notations based on averaging across environments. These notations are formulated using $\mathcal{G}(k,g,h,f)$, where $k,g,h,f$ define how the sample sizes are aggregated and weighted.}
    \label{tab:sample_size_sum}
\end{table}

\begin{table}[H]
    \centering
    \renewcommand{\arraystretch}{0.5}
    \begin{tabular}{lccccccc} 
        \toprule
         Sample size &  Full Expression& Function & $k(x,y,z)$ &  $g(x)$ & $h(y)$ &  $f(z)$ \\ 
        \midrule
        $\overline{\sqrt{nN}}^{\widehat\tau,|\eta|}$ &  $\displaystyle\left(\sum_{e \in \mathcal{E}} \frac{\omega^{(e)} \widehat{\tau}^{(e)}|\eta^{(e)}|}{\sqrt{n^{(e)} N^{(e)}}}\right)^{-1}  $ & $\mathcal{G}(k,g,h,f)$ &  $\sqrt{xy}$ & $x$ & $y$&$z$ \\ 
       $\overline{\sqrt{mN}}^{\widehat\tau,|\eta|}$ &  $\displaystyle\left(\sum_{e \in \mathcal{E}} \frac{\omega^{(e)} \widehat{\tau}^{(e)}|\eta^{(e)}|}{\sqrt{m^{(e)} N^{(e)}}}\right)^{-1}$ &$\mathcal{G}(k,g,h,f)$   &   $\sqrt{xz}$ &$x$  & $y$  & $z$\\ 
        $\overline{\sqrt{n} N}^{\widehat\tau}$ &  $\displaystyle \left(\sum_{e \in \mathcal{E}} \frac{ \omega^{(e)}\widehat\tau^{(e)}}{\sqrt{n^{(e)}}N^{(e)}}\right)^{-1}$ & $\mathcal{G}(k,g,h,f)$ &$x\sqrt{y}$  & $x$ &$y$&1 \\ 
        $\overline{\sqrt{m} N}^{\widehat\tau}$ & $\displaystyle \left(\sum_{e \in \mathcal{E}} \frac{ \omega^{(e)}\widehat\tau^{(e)}}{\sqrt{m^{(e)}}N^{(e)}}\right)^{-1}$ & $\mathcal{G}(k,g,h,f)$ & $x\sqrt{z}$ & $x$ & $y$&1\\ 
       $\overline{\sqrt{n} N}^{\widehat\tau, |\eta|}$  &  $\displaystyle\left(\sum_{e \in \mathcal{E}} \frac{ \omega^{(e)}\widehat\tau^{(e)}|\eta^{(e)}|}{\sqrt{n^{(e)}}N^{(e)}}\right)^{-1}$ & $\mathcal{G}(k,g,h,f)$ & $x\sqrt{y}$  &$x$ & $y$&$z$\\ 
        $\overline{\sqrt{m} N}^{\widehat\tau, |\eta|}$ & $\displaystyle\left( \sum_{e \in \mathcal{E}} \frac{ \omega^{(e)}\widehat\tau^{(e)} |\eta^{(e)}|}{\sqrt{m^{(e)}}N^{(e)}}\right)^{-1}$  & $\mathcal{G}(k,g,h,f)$ &  $x\sqrt{z}$ &$x$  &$y$& $z$\\ 
       $\overline{\sqrt{Nn }}^{|\eta|^2}$  & $\displaystyle \left(\sum_{e \in \mathcal{E}} \frac{\omega^{(e)}|\eta^{(e)}|^2}{\sqrt{N^{(e)}n^{(e)}}}\right)^{-1} $ & $\mathcal{G}(k,g,h,f)$ &  $\sqrt{xy}$& $x $ &$1$ & $z^2$ \\ 
        $\overline{\sqrt{nm}}^{\widehat\tau^2, |\eta|} $ & $\displaystyle\left(\sum_{e \in \mathcal{E}}\frac{\omega^{(e)}( \widehat\tau^{(e)})^2|\eta^{(e)} |}{\sqrt{n^{(e)}m^{(e)}}}\right)^{-1}$ & $\mathcal{G}(k,g,h,f)$ &$\sqrt{yz}$  &  $x $ & $y^2$ & $z$\\ 
       $N_{\dagger}$  &  $\displaystyle \left(\sum_{e \in \mathcal{E}} \frac{\omega^{(e)}}{(N^{(e)})^{2/3}}\right)^{-1}$ & $\mathcal{G}(k,g,h,f)$ & $x^{2/3}$ &$x$ &$1$ & $1$ \\  
        $N_{\omega}$ & $\displaystyle\left(\sum_{e \in \mathcal{E}} \frac{\left(\omega^{(e)}\right)^2}{N^{(e)}}\right)^{-1}$  & $\mathcal{G}(k,g,h,f)$ & $x$ & $x^2$ & $1$ & $1$\\
        $n_\omega^{\widehat\tau^2}$ & $\displaystyle \left(\sum_{e \in \mathcal{E}} \frac{\left(\omega^{(e)}\widehat\tau^{(e)}\right)^2}{n^{(e)}}\right)^{-1}$ & $\mathcal{G}(k,g,h,f)$ &$y$   &$x^2$ &$y^2$& $1$ \\ 
        $ m_{\omega}^{\widehat\tau^2}$ & $\displaystyle \left(\sum_{e \in \mathcal{E}} \frac{\left(\omega^{(e)}\widehat\tau^{(e)}\right)^2}{m^{(e)}}\right)^{-1}$ & $\mathcal{G}(k,g,h,f)$ & $z$  &$x^2$  &$y^2$& $1$   \\ 
        $n_\omega^{\widehat\tau^2,|\eta|^2}$ & $\displaystyle \left(\sum_{e \in \mathcal{E}} \frac{\left(\omega^{(e)}\widehat\tau^{(e)}|\eta^{(e)}|\right)^2}{n^{(e)}}\right)^{-1}$ & $\mathcal{G}(k,g,h,f)$ &$y$&  $x^2$  &$y^2$ & $z^2$ \\ 
       $m_\omega^{\widehat\tau^2,|\eta|^2}$ & $\displaystyle \left(\sum_{e \in \mathcal{E}} \frac{\left(\omega^{(e)}\widehat\tau^{(e)}|\eta^{(e)}|\right)^2}{m^{(e)}}\right)^{-1}$ & $\mathcal{G}(k,g,h,f)$  &  $z$& $x^2$  &$y^2$ &$z^2$ \\      
       \bottomrule
    \end{tabular}
    \caption{Extended summary of sample size notations based on averaging across environments, incorporating additional adjustments for missingness, weighting, and absolute imputation bias.}
    \label{tab:sample_size_sum_more}
\end{table}

\section{Preliminaries}\label{section:preliminaries}

We introduce some basic definitions and lemmas that will be useful for the proofs.

\begin{definition}\label{definition:sub-gaussian_random_variable}
	\textnormal{(Sub-Gaussian Random Variable)} A random variable $X$ is a sub-Gaussian random variable with parameter $\sigma \in \mathbb{R}^{+}$if
	$$
		\forall \lambda \in \mathbb{R}, \quad \mathbb{E}[\exp (\lambda X)] \leq \exp \left(\frac{\lambda^2}{2} \sigma^2\right).
	$$
\end{definition}

\begin{definition}\label{definition:sub-exp_random_variable}
	\textnormal{(Sub-exponential Random Variable)}A random variable $X$ is a sub-Exponential random variable with parameter $(\nu, \alpha) \in \mathbb{R}^{+} \times \mathbb{R}^{+}$if
	$$
		\forall|\lambda|<1 / \alpha, \quad \mathbb{E}[\exp (\lambda X)] \leq \exp \left(\frac{\lambda^2}{2} \nu^2\right).
	$$
\end{definition}

\begin{lemma}\label{lemma:product_two_sub-gaussian_random_variables}
	\textnormal{(Product of Two Sub-Gaussian Random Variables)} Suppose $X_1$ and $X_2$ are two zero-mean sub-Gaussian random variables with parameters $\sigma_1$ and $\sigma_2$, respectively. Then $X_1 X_2$ is a sub-exponential random variable with parameter $\left(c_1 \sigma_1 \sigma_2, c_2 \sigma_1 \sigma_2\right)$, where $c_1, c_2>0$ are some universal constants.
\end{lemma}

\begin{lemma}\label{lemma:sum_independent_sub-exponential_random_variables}
	\textnormal{(Sum of Independent Sub-exponential Random Variables)} Suppose $X_1, \ldots, X_N$ are independent sub-exponential random variables with parameters $\left\{\left(\nu_i, \alpha_i\right)\right\}_{i=1}^N$, respectively. There exists some universal constant $c_1$ such that the following holds,
	$$
		\mathbb{P}\left[\left|\sum_{i=1}^N\left(X_i-\mathbb{E}\left[X_i\right]\right)\right| \geq c_1\left\{\sqrt{t \times \sum_{i=1}^N \nu_i^2}+t \times \max _{i \in[N]} \alpha_i\right\}\right] \leq 2 e^{-t}
	$$
\end{lemma}

\begin{definition}\label{definition:power_mean}
	\textnormal{(Power mean)} Let $p\in\mathcal{R}$ be nonzero, $k_1,...,k_M$ are positive real numbers. The generalized mean with exponent $p$ of values $k_1,...,k_M$ is
	$$
		M_p(k_1,...,k_M) = \left(\frac{1}{M} \sum_{e=1}^{M} k_e^p\right)^{\frac{1}{p}}
	$$
\end{definition}

\begin{definition}\label{definition:weighted_power_mean}
	\textnormal{(Weighted power mean)} Consider a sequence of positive weights $\nu_1,...,\nu_M$, the weighted power mean with exponent $p$ of positive real numbers $k_1,...,k_M$ is defined as
	$$
		WM_p(k_1,...,k_M) = \left(\frac{ \sum_{e=1}^{M}\nu_e \,k_e^p}{\sum_{e=1}^M\nu_e}\right)^{\frac{1}{p}}
	$$
\end{definition}

\section{Technical Lemmas}\label{section:technical_lemma}

\begin{prop}\label{prop:pop_eills_continue}
	wait.
\end{prop}

\begin{lemma}\label{lemma:weighted_power_mean_inequality}
	\label{lemma:weighted_power_mean_inequality} Suppose $p<q$, then we have $WM_p(k_1,...,k_M) \le WM_q(k_1,...,k_M) $. In particular, when $p=-\infty$ and. $q=-1$, we have
	$$
		\frac{\sum_{e=1}^M\nu_e}{\sum_{e=1}^M\nu_ek_e^{-1}} = WM_{-1}(k_1,...,k_M) \ge  WM_{-\infty}(k_1,...,k_M) = \min\{k_1,...,k_M\}
	$$
	Notice this inequality holds without requiring the sum of positive weights to be $1$.
\end{lemma}

{\it Proof.}
Since $k_1,...,k_M$ are positive real numbers and $\nu_1,...,\nu_M$ are positive weights,
$$
	\frac{\partial}{\partial p} WM_p(k_1,...,k_M) \ge 0.
$$
\qed

\begin{remark}\label{remark:specific_weighted_power_mean_inequality}
	Recall that $n^{(e)}$ represents the number of labeled observations in each environment, and $\omega^{(e)}$ denotes the weights assigned to the environments, with $\sum_{e\in\mathcal{E}} \omega^{(e)} = 1$. Let us define $k_e = \frac{n^{(e)}}{\omega^{(e)}}$ and set $\nu_e = \omega^{(e)}$. Then, by applying Lemma \ref{lemma:weighted_power_mean_inequality}, we obtain the inequality
	\begin{equation}\label{eq:specific_weighted_power_mean_inequality}
		n_{\omega}=\frac{1}{\sum_{e\in\mathcal{E}}  \frac{ (\omega^{(e)})^2}{n^{(e)}} }\ge \underset{e\in\mathcal{E}}{\operatorname{min}} \frac{n^{(e)}}{(\omega^{(e)})} = n_*.
	\end{equation}
\end{remark}

\section{Proofs for Non-asymmptotic Results}\label{section:proofs_nonasymptotic}

\subsection{Proof of Lemma \ref{lemma:two_sided_bound_R}}

It follows from the definitions of the pooled population $L_2$ risk $\mathrm{R}$ in (\ref{eq:pool_loss_pop}), the pooled empirical $L_2$ risk $\widehat{\mathrm{R}}_{\mathrm{Adj}}(\boldsymbol{\beta} ; \boldsymbol{\omega})$ in (\ref{eq:bia_corr_pool_L2_loss}) with imputation bias adjusted, and the imputation model $\widehat{h}^{(e)}$ that
\begin{align*}
	\big| \mathrm{R}(\boldsymbol{\beta}) & -\mathrm{R}(\boldsymbol{\beta}^*)   - \widehat{\mathrm{R}}_{\mathrm{Adj}}(\boldsymbol{\beta})+\widehat{\mathrm{R}}_{\mathrm{Adj}}(\boldsymbol{\beta}^*)\big|                                                                                                                                                                                                                                                                      \\
	                                     & = \Big| (\boldsymbol{\beta}-\boldsymbol{\beta}^*)^{\top}(\boldsymbol{\Sigma}-\widehat{\boldsymbol{\Sigma}}_{N})(\boldsymbol{\beta}-\boldsymbol{\beta}^*) -  2(\boldsymbol{\beta}-\boldsymbol{\beta}^*)^{\top} \sum_{e\in\mathcal{E}}\omega^{(e)} \left\{ \mathbb{E}[\boldsymbol{x}^{(e)} \varepsilon^{(e)}] - \widehat{\mathbb{E}}_{N^{(e)}}[{\boldsymbol{x}}^{(e)}\varepsilon^{(e)}]  \right\}                                   \\
	                                     & \quad - 2(\boldsymbol{\beta}-\boldsymbol{\beta}^*)^{\top} \sum_{e\in\mathcal{E}}\omega^{(e)}\left\{\mathbb{E}[\boldsymbol{x}^{(e)} \{\widehat{h}^{(e)}(\boldsymbol{x}^{(e)}) - y^{(e)}\}] -\widehat{\mathbb{E}}_{N^{(e)}}[{\boldsymbol{x}}^{(e)} \{\widehat{h}^{(e)}({\boldsymbol{x}}^{(e)} )- {y}^{(e)}\}] \right\}                                                                                                              \\
	                                     & \quad  -2(\boldsymbol{\beta}-\boldsymbol{\beta}^*)^{\top} \sum_{e\in\mathcal{E}}\omega^{(e)}\left\{ \mathbb{E}[\boldsymbol{x}^{(e)} \{y^{(e)} - \widehat{h}^{(e)}(\boldsymbol{x}^{(e)} )\}] -\widehat{\mathbb{E}}_{n^{(e)}}[\boldsymbol{x}^{(e)} \{y^{(e)} - \widehat{h}^{(e)}(\boldsymbol{x}^{(e)})\}] \right\} \Big |                                                                                                           \\
	                                     & \stackrel{(a)}{=}\Big| (\boldsymbol{\beta}-\boldsymbol{\beta}^*)^{\top}(\boldsymbol{\Sigma}-\widehat{\boldsymbol{\Sigma}}_{N})(\boldsymbol{\beta}-\boldsymbol{\beta}^*) -  2(\boldsymbol{\beta}-\boldsymbol{\beta}^*)^{\top} \sum_{e\in\mathcal{E}}\omega^{(e)} \left\{ \mathbb{E}[\boldsymbol{x}^{(e)} \varepsilon^{(e)}] - \widehat{\mathbb{E}}_{N^{(e)}}[{\boldsymbol{x}}^{(e)}\varepsilon^{(e)}]  \right\}                    \\
	                                     & \quad - 2(\boldsymbol{\beta}-\boldsymbol{\beta}^*)^{\top} \sum_{e\in\mathcal{E}}\omega^{(e)}\left\{\mathbb{E}[\boldsymbol{x}^{(e)} (z^{(e)}-\eta^{(e)})] +\mathbb{E}[\boldsymbol{x}^{(e)} \eta^{(e)}]-\widehat{\mathbb{E}}_{N^{(e)}}[{\boldsymbol{x}}^{(e)} (z^{(e)}-\eta^{(e)})] -\widehat{\mathbb{E}}_{N^{(e)}}[{\boldsymbol{x}}^{(e)} (\eta^{(e)})] \right\}                                                                   \\
	                                     & \quad  -2(\boldsymbol{\beta}-\boldsymbol{\beta}^*)^{\top} \sum_{e\in\mathcal{E}}\omega^{(e)}\left\{ \mathbb{E}[\boldsymbol{x}^{(e)} (-z^{(e)}+\eta^{(e)})]+ \mathbb{E}[\boldsymbol{x}^{(e)} (-\eta^{(e)} )] -\widehat{\mathbb{E}}_{n^{(e)}}[\boldsymbol{x}^{(e)} (-z^{(e)}+\eta^{(e)})] -\widehat{\mathbb{E}}_{n^{(e)}}[\boldsymbol{x}^{(e)} (-\eta^{(e)})] \right\} \Big |                                                       \\
	                                     & \stackrel{(b)}{=}\Big| (\boldsymbol{\beta}-\boldsymbol{\beta}^*)^{\top}(\boldsymbol{\Sigma}-\widehat{\boldsymbol{\Sigma}}_{N})(\boldsymbol{\beta}-\boldsymbol{\beta}^*) -  2(\boldsymbol{\beta}-\boldsymbol{\beta}^*)^{\top} \sum_{e\in\mathcal{E}}\omega^{(e)} \left\{ \mathbb{E}[\boldsymbol{x}^{(e)} \varepsilon^{(e)}] - \widehat{\mathbb{E}}_{N^{(e)}}[{\boldsymbol{x}}^{(e)}\varepsilon^{(e)}]  \right\}                    \\
	                                     & \quad - 2(\boldsymbol{\beta}-\boldsymbol{\beta}^*)^{\top} \sum_{e\in\mathcal{E}}\omega^{(e)}\left\{\widehat{\tau}^{(e)}\left(\mathbb{E}[\boldsymbol{x}^{(e)} (z^{(e)}-\eta^{(e)})] -\widehat{\mathbb{E}}_{m^{(e)}}[{\boldsymbol{x}}^{(e)} (z^{(e)}-\eta^{(e)})]\right) \right\}                                                                                                                                                   \\
	                                     & \quad - 2(\boldsymbol{\beta}-\boldsymbol{\beta}^*)^{\top} \sum_{e\in\mathcal{E}}\omega^{(e)}\left\{\widehat{\tau}^{(e)}\left(\mathbb{E}[\boldsymbol{x}^{(e)} ] -\widehat{\mathbb{E}}_{m^{(e)}}[{\boldsymbol{x}}^{(e)}  ]\right)\eta^{(e)}\right\}                                                                                                                                                                                 \\
	                                     & \quad  -2(\boldsymbol{\beta}-\boldsymbol{\beta}^*)^{\top} \sum_{e\in\mathcal{E}}\omega^{(e)}\left\{\widehat{\tau}^{(e)}\left(- \mathbb{E}[\boldsymbol{x}^{(e)} (z^{(e)}-\eta^{(e)})] +\widehat{\mathbb{E}}_{n^{(e)}}[\boldsymbol{x}^{(e)} (z^{(e)}-\eta^{(e)})] \right)\right\}                                                                                                                                                   \\
	                                     & \quad  -2(\boldsymbol{\beta}-\boldsymbol{\beta}^*)^{\top} \sum_{e\in\mathcal{E}}\omega^{(e)}\left\{\widehat{\tau}^{(e)}\left(- \mathbb{E}[\boldsymbol{x}^{(e)} ] +\widehat{\mathbb{E}}_{n^{(e)}}[\boldsymbol{x}^{(e)}]\right) \eta^{(e)} \right\} \Big |                                                                                                                                                                          \\
	                                     & \stackrel{(c)}{\leq} \Big| (\boldsymbol{\beta}-\boldsymbol{\beta}^*)^{\top}(\boldsymbol{\Sigma}-\widehat{\boldsymbol{\Sigma}}_{N})(\boldsymbol{\beta}-\boldsymbol{\beta}^*) \Big|+2\Big|  (\boldsymbol{\beta}-\boldsymbol{\beta}^*)^{\top} \sum_{e\in\mathcal{E}}\omega^{(e)} \left\{ \mathbb{E}[\boldsymbol{x}^{(e)} \varepsilon^{(e)}] - \widehat{\mathbb{E}}_{N^{(e)}}[{\boldsymbol{x}}^{(e)}\varepsilon^{(e)}]  \right\}\Big| \\
	                                     & \quad +2\Big| (\boldsymbol{\beta}-\boldsymbol{\beta}^*)^{\top} \sum_{e\in\mathcal{E}}\omega^{(e)}\widehat{\tau}^{(e)}\left\{\mathbb{E}[\boldsymbol{x}^{(e)} (z^{(e)}-\eta^{(e)})] -\widehat{\mathbb{E}}_{m^{(e)}}[{\boldsymbol{x}}^{(e)} (z^{(e)}-\eta^{(e)})] \right\}\Big|                                                                                                                                                      \\
	                                     & \quad +2\Big| (\boldsymbol{\beta}-\boldsymbol{\beta}^*)^{\top} \sum_{e\in\mathcal{E}}\omega^{(e)}\widehat{\tau}^{(e)}\eta^{(e)}\left\{\mathbb{E}[\boldsymbol{x}^{(e)} ] -\widehat{\mathbb{E}}_{m^{(e)}}[{\boldsymbol{x}}^{(e)}  ]\right\}\Big|                                                                                                                                                                                    \\
	                                     & \quad  +2\Big|(\boldsymbol{\beta}-\boldsymbol{\beta}^*)^{\top} \sum_{e\in\mathcal{E}}\omega^{(e)}\widehat{\tau}^{(e)}\left\{- \mathbb{E}[\boldsymbol{x}^{(e)} (z^{(e)}-\eta^{(e)})] +\widehat{\mathbb{E}}_{n^{(e)}}[\boldsymbol{x}^{(e)} (z^{(e)}-\eta^{(e)})] \right\}\Big|                                                                                                                                                      \\
	                                     & \quad  +2\Big|(\boldsymbol{\beta}-\boldsymbol{\beta}^*)^{\top} \sum_{e\in\mathcal{E}}\omega^{(e)}\widehat{\tau}^{(e)}\eta^{(e)} \left\{- \mathbb{E}[\boldsymbol{x}^{(e)} ] +\widehat{\mathbb{E}}_{n^{(e)}}[\boldsymbol{x}^{(e)}]  \right\}\Big |                                                                                                                                                                                  \\
	                                     & = H_1^{(e)}(\boldsymbol{\beta}) + H_2^{(e)}(\boldsymbol{\beta}) +H_3^{(e)}(\boldsymbol{\beta})+H_4^{(e)}(\boldsymbol{\beta})+H_5^{(e)}(\boldsymbol{\beta})+H_6^{(e)}(\boldsymbol{\beta}),
\end{align*}
where at the step $(a)$ we replace imputation $\widehat{h}^{(e)}({\boldsymbol{x}}^{(e)} )- {y}^{(e)}$ with its short notation $z^{(e)}$. To prepare for the next step, we also center the random variable $z^{(e)}$ or $-z^{(e)}$ by subtracting or plus its mean $\eta^{(e)}$, respectively. In step $(b)$, we use two observations. First, we have
\begin{align*}
	 & \left\{\left(\mathbb{E}[\boldsymbol{x}^{(e)} (z^{(e)}-\eta^{(e)})] -\widehat{\mathbb{E}}_{N^{(e)}}[{\boldsymbol{x}}^{(e)} (z^{(e)}-\eta^{(e)})]\right) \right\} +\left\{\left(\mathbb{E}[\boldsymbol{x}^{(e)} (-z^{(e)}+\eta^{(e)})] -\widehat{\mathbb{E}}_{n^{(e)}}[{\boldsymbol{x}}^{(e)} (-z^{(e)}+\eta^{(e)})]\right) \right\}                                                   \\
	 & \quad = \left\{\left(\mathbb{E}[\boldsymbol{x}^{(e)} (z^{(e)}-\eta^{(e)})] -\widehat{\mathbb{E}}_{N^{(e)}}[{\boldsymbol{x}}^{(e)} (z^{(e)}-\eta^{(e)})]\right) \right\} +\left\{\left(-\mathbb{E}[\boldsymbol{x}^{(e)} (z^{(e)}-\eta^{(e)})] +\widehat{\mathbb{E}}_{n^{(e)}}[{\boldsymbol{x}}^{(e)} (z^{(e)}-\eta^{(e)})]\right) \right\}                                            \\
	 & \quad = \left\{\left(\mathbb{E}[\boldsymbol{x}^{(e)} (z^{(e)}-\eta^{(e)})]  - \frac{1}{N^{(e)}} \sum_{\ell=1}^{N^{(e)}}\boldsymbol{x}_{\ell}^{(e)} (z^{(e)}_{\ell}-\eta^{(e)}) \right) \right\} +  \left\{\left( -\mathbb{E}[\boldsymbol{x}^{(e)} (z^{(e)}-\eta^{(e)})]  + \frac{1}{n^{(e)}}\sum_{i=1}^{n^{(e)}}\boldsymbol{x}_{i}^{(e)} (z^{(e)}_{i}-\eta^{(e)}) \right) \right\}   \\
	 & \quad =\mathbb{E}[\boldsymbol{x}^{(e)} (z^{(e)}-\eta^{(e)})]  - \frac{1}{N^{(e)}} \sum_{\ell=1}^{m^{(e)}}\boldsymbol{x}_{\ell}^{(e)} (z^{(e)}_{\ell}-\eta^{(e)}) - \frac{1}{N^{(e)}}\sum_{i=1}^{n^{(e)}}\boldsymbol{x}_{i}^{(e)} (z^{(e)}_{i}-\eta^{(e)})                                                                                                                            \\
	 & \quad \quad  - \mathbb{E}[\boldsymbol{x}^{(e)} (z^{(e)}-\eta^{(e)})]  + \frac{1}{n^{(e)}}\sum_{i=1}^{n^{(e)}}\boldsymbol{x}_{i}^{(e)} (z^{(e)}_{i}-\eta^{(e)})                                                                                                                                                                                                                       \\
	 & \quad =\mathbb{E}[\boldsymbol{x}^{(e)} (z^{(e)}-\eta^{(e)})]  - \frac{m^{(e)}}{N^{(e)}} \frac{1}{m^{(e)}}\sum_{\ell=1}^{m^{(e)}}\boldsymbol{x}_{\ell}^{(e)} (z^{(e)}_{\ell}-\eta^{(e)}) - \mathbb{E}[\boldsymbol{x}^{(e)} (z^{(e)}-\eta^{(e)})] +\left( - \frac{1}{N^{(e)}}+ \frac{1}{n^{(e)}}\right)\sum_{i=1}^{n^{(e)}}\boldsymbol{x}_{i}^{(e)} (z^{(e)}_{i}-\eta^{(e)})           \\
	 & \quad = \mathbb{E}[\boldsymbol{x}^{(e)} (z^{(e)}-\eta^{(e)})]  - \frac{m^{(e)}}{N^{(e)}} \frac{1}{m^{(e)}}\sum_{\ell=1}^{m^{(e)}}\boldsymbol{x}_{\ell}^{(e)} (z^{(e)}_{\ell}-\eta^{(e)}) - \mathbb{E}[\boldsymbol{x}^{(e)} (z^{(e)}-\eta^{(e)})] + \frac{m^{(e)}}{N^{(e)}}\frac{1}{n^{(e)}}\sum_{i=1}^{n^{(e)}}\boldsymbol{x}_{i}^{(e)} (z^{(e)}_{i}-\eta^{(e)})                     \\
	 & \quad = \mathbb{E}[\boldsymbol{x}^{(e)} (z^{(e)}-\eta^{(e)})]  - \widehat{\tau}^{(e)} \frac{1}{m^{(e)}}\sum_{\ell=1}^{m^{(e)}}\boldsymbol{x}_{\ell}^{(e)} (z^{(e)}_{\ell}-\eta^{(e)}) - \mathbb{E}[\boldsymbol{x}^{(e)} (z^{(e)}-\eta^{(e)})] + \widehat{\tau}^{(e)} \frac{1}{n^{(e)}}\sum_{i=1}^{n^{(e)}}\boldsymbol{x}_{i}^{(e)} (z^{(e)}_{i}-\eta^{(e)})                          \\
	 & \quad = \left\{\widehat{\tau}^{(e)} \left(\mathbb{E}[\boldsymbol{x}^{(e)} (z^{(e)}-\eta^{(e)})] -\widehat{\mathbb{E}}_{m^{(e)}}[{\boldsymbol{x}}^{(e)} (z^{(e)}-\eta^{(e)})]\right) \right\} +\left\{\widehat{\tau}^{(e)} \left(-\mathbb{E}[\boldsymbol{x}^{(e)} (z^{(e)}-\eta^{(e)})] +\widehat{\mathbb{E}}_{n^{(e)}}[{\boldsymbol{x}}^{(e)} (z^{(e)}-\eta^{(e)})]\right) \right\}.
\end{align*}
Moreover, we have
\begin{align*}
	 & \left\{\left(\mathbb{E}[\boldsymbol{x}^{(e)} ] -\widehat{\mathbb{E}}_{N^{(e)}}[{\boldsymbol{x}}^{(e)}  ]\right)\eta^{(e)}\right\} + \left\{\left(\mathbb{E}[\boldsymbol{x}^{(e)} ] -\widehat{\mathbb{E}}_{n^{(e)}}[\boldsymbol{x}^{(e)}]\right) (-\eta^{(e)}) \right\}                                              \\
	 & \quad = \left\{\left(\mathbb{E}[\boldsymbol{x}^{(e)} ] -\widehat{\mathbb{E}}_{N^{(e)}}[{\boldsymbol{x}}^{(e)}  ]\right)\eta^{(e)}\right\} + \left\{\left(-\mathbb{E}[\boldsymbol{x}^{(e)} ] +\widehat{\mathbb{E}}_{n^{(e)}}[\boldsymbol{x}^{(e)}]\right) \eta^{(e)} \right\}                                        \\
	 & \quad= \left\{\mathbb{E}[\boldsymbol{x}^{(e)} ] -\widehat{\mathbb{E}}_{N^{(e)}}[{\boldsymbol{x}}^{(e)}  ]-\mathbb{E}[\boldsymbol{x}^{(e)} ] +\widehat{\mathbb{E}}_{n^{(e)}}[\boldsymbol{x}^{(e)}]\right\} \eta^{(e)}                                                                                                \\
	 & \quad= \left\{\mathbb{E}[\boldsymbol{x}^{(e)} ] -\frac{1}{N^{(e)}}\sum_{\ell=1}^{N^{(e)}}{\boldsymbol{x}}_{\ell}^{(e)}  -\mathbb{E}[\boldsymbol{x}^{(e)} ] +\frac{1}{n^{(e)}}\sum_{i=1}^{n^{(e)}} \boldsymbol{x}_i^{(e)}\right\} \eta^{(e)}                                                                         \\
	 & \quad = \left\{\mathbb{E}[\boldsymbol{x}^{(e)} ] -\frac{1}{N^{(e)}}\sum_{\ell=1}^{m^{(e)}}{\boldsymbol{x}}_{\ell}^{(e)}-\frac{1}{N^{(e)}}\sum_{i=1}^{n^{(e)}}{\boldsymbol{x}}_{i}^{(e)}  -\mathbb{E}[\boldsymbol{x}^{(e)} ] +\frac{1}{n^{(e)}}\sum_{i=1}^{n^{(e)}} \boldsymbol{x}_i^{(e)}\right\} \eta^{(e)}        \\
	 & \quad = \left\{\mathbb{E}[\boldsymbol{x}^{(e)} ] -\frac{1}{N^{(e)}}\sum_{\ell=1}^{m^{(e)}}{\boldsymbol{x}}_{\ell}^{(e)} -\mathbb{E}[\boldsymbol{x}^{(e)} ]+\left(-\frac{1}{N^{(e)}}+\frac{1}{n^{(e)}}\right)\sum_{i=1}^{n^{(e)}}{\boldsymbol{x}}_{i}^{(e)}  \right\} \eta^{(e)}                                     \\
	 & \quad = \left\{\mathbb{E}[\boldsymbol{x}^{(e)} ] -\frac{m^{(e)}}{N^{(e)}}\frac{1}{m^{(e)}}\sum_{\ell=1}^{m^{(e)}}{\boldsymbol{x}}_{\ell}^{(e)} -\mathbb{E}[\boldsymbol{x}^{(e)} ]+\frac{-n^{(e)}+N^{(e)}}{n^{(e)}N^{(e)}}\sum_{i=1}^{n^{(e)}}{\boldsymbol{x}}_{i}^{(e)}  \right\} \eta^{(e)}                        \\
	 & \quad = \left\{\mathbb{E}[\boldsymbol{x}^{(e)} ] -\frac{m^{(e)}}{N^{(e)}}\frac{1}{m^{(e)}}\sum_{\ell=1}^{m^{(e)}}{\boldsymbol{x}}_{\ell}^{(e)} -\mathbb{E}[\boldsymbol{x}^{(e)} ]+\frac{m^{(e)}}{N^{(e)}}\frac{1}{n^{(e)}}\sum_{i=1}^{n^{(e)}}{\boldsymbol{x}}_{i}^{(e)}  \right\} \eta^{(e)}                       \\
	 & \quad= \left\{\widehat{\tau}^{(e)}\left(\mathbb{E}[\boldsymbol{x}^{(e)} ] -\widehat{\mathbb{E}}_{m^{(e)}}[{\boldsymbol{x}}^{(e)}  ]\right)\eta^{(e)}\right\} + \left\{\widehat{\tau}^{(e)}\left(-\mathbb{E}[\boldsymbol{x}^{(e)} ] +\widehat{\mathbb{E}}_{n^{(e)}}[\boldsymbol{x}^{(e)}]\right)\eta^{(e)} \right\}.
\end{align*}
This is achievable because the objective $\widehat{\mathrm{Q}}_{\mathrm{Adj}}$ in (\ref{eq:objective_adjust_whole}) utilizes all available observations and properly accounts for the imputation bias with labeled data. In the last step (c), we simply apply triangle inequality. Now we are ready to derive high-probability upper bounds on $H_k^{(e)}(\boldsymbol{\beta})$ for each $k\in[6]$. Note that the first two terms are similar as the expression in Lemma C.7 of \cite{fan2024environment}, but we include them here for completeness.

\noindent{\it Step 1.} UPPER BOUND ON $H_1^{(e)}(\boldsymbol{\beta})$. Define the event
$$
	\mathcal{C}_{1,t} = \left\{\forall \boldsymbol{\beta}\in\mathbb{R}^p, \quad H_1^{(e)}(\boldsymbol{\beta}) \le C_1\kappa_U\sigma_x^2\|\boldsymbol{\beta} - \boldsymbol{\beta}^*\|_2^2\left\{\sqrt{\frac{t+p}{N_{\omega}}} + \frac{t+p}{N_*}\right\}\right\}.
$$
We claim that this event occurs with probability at least $1-e^{-t}$ for any $t>0$.

With out loss of generality, let $\boldsymbol{\beta}\ne \boldsymbol{\beta}^*$, observe
\begin{align}
	\sup _{\boldsymbol{\beta}\in\mathbb{R}^p, \boldsymbol{\beta} \neq \boldsymbol{\beta}^*} \frac{\left(\boldsymbol{\beta}-\boldsymbol{\beta}^*\right)^{\top}}{\left\|\boldsymbol{\beta}-\boldsymbol{\beta}^*\right\|_2} \left(\boldsymbol\Sigma - \widehat{\boldsymbol\Sigma}_N\right)\frac{\left(\boldsymbol{\beta}-\boldsymbol{\beta}^*\right)^{\top}}{\left\|\boldsymbol{\beta}-\boldsymbol{\beta}^*\right\|_2} & = \sup _{\boldsymbol{\beta}\in\mathbb{R}^p, \boldsymbol{\beta} \neq \boldsymbol{\beta}^*} \frac{\left(\boldsymbol{\beta}-\boldsymbol{\beta}^*\right)^{\top}}{\left\|\boldsymbol{\beta}-\boldsymbol{\beta}^*\right\|_2} \sum_{e\in\mathcal{E}}\omega^{(e)}\left(\boldsymbol\Sigma^{(e)}_{S\cup S*} - \widehat{\boldsymbol\Sigma}^{(e)}_{N^{(e)}, S\cup S*}\right)\frac{\left(\boldsymbol{\beta}-\boldsymbol{\beta}^*\right)^{\top}}{\left\|\boldsymbol{\beta}-\boldsymbol{\beta}^*\right\|_2}\notag \\
	                                                                                                                                                                                                                                                                                                                                                                                                                & \le \sup _{\boldsymbol{\beta}\in\mathbb{R}^p, \boldsymbol{\beta} \neq \boldsymbol{\beta}^*}  \left\|\sum_{e\in\mathcal{E}}\omega^{(e)}\left(\boldsymbol\Sigma^{(e)}_{S\cup S*} - \widehat{\boldsymbol\Sigma}^{(e)}_{N^{(e)}, S\cup S*}\right)\right\|_2\notag                                                                                                                                                                                                                                      \\
	                                                                                                                                                                                                                                                                                                                                                                                                                & \equiv \sup _{|S| \leq p}\left\|\boldsymbol{A}_S\right\|_2
	\label{eq:pf_lemma_a1_1}
\end{align}

For any $S\subset [p]$, let $\left\{\left(\boldsymbol{v}_k^{(e)}, \boldsymbol{u}_k^{(e)}\right)\right\}_{k=1}^{N_S} \in \mathcal{B}\left(S \cup S^*\right) \times \mathcal{B}\left(S \cup S^*\right)\coloneq \mathcal{B}^2\left(S \cup S^*\right)
$ be a $1/4$-covering of $\mathcal{B}^2\left(S \cup S^*\right)$ in a sense that for any $(\boldsymbol{u}, \boldsymbol{v}) \in \mathcal{B}^2\left(S \cup S^*\right)$, there exists some $\pi(\boldsymbol{u}, \boldsymbol{v}) \in\left[N_S\right]$ such that
$$
	\left\|\boldsymbol{u}-\boldsymbol{u}_{\pi(\boldsymbol{u}, \boldsymbol{v})}^{(S)}\right\|_2+\left\|\boldsymbol{v}-\boldsymbol{v}_{\pi(\boldsymbol{u}, \boldsymbol{v})}^{(S)}\right\|_2 \leq \frac{1}{4}
$$
It follows from standard empirical process theory that $N_S \leq 9^{2\left|S \cup S^*\right|}$, then
$$
	N=\sum_{S \subset[p]} N_S \leq \sum_{S \subset[p]}  9^{2\left|S \cup S^*\right|} \leq \sum_{i=0}^p 81^{i+s^*}\left(\begin{array}{c}
			p \\
			i
		\end{array}\right) \leq 81^{s^*}\left(81 e\right)^p.
$$
At the same time, denote $\boldsymbol{u}^{\dagger}=\boldsymbol{u}_{\pi(\boldsymbol{u}, \boldsymbol{v})}^{(S)}$ and $\boldsymbol{v}^{\dagger}=\boldsymbol{v}_{\pi(\boldsymbol{u}, \boldsymbol{v})}^{(S)}$. It follows from the variational representation of the matrix $\ell_2$ norm that
$$
	\begin{aligned}
		\left\|\boldsymbol{A}_S\right\|_2= & \sup _{(\boldsymbol{u}, \boldsymbol{v}) \in \mathcal{B}^2\left(S \cup S^*\right)} \boldsymbol{u}^{\top} \boldsymbol{A}_S \boldsymbol{v}                                                                                                                                                                                                                                                                                    \\
		\leq                               & \sup _{(\boldsymbol{u}, \boldsymbol{v}) \in \mathcal{B}^2\left(S \cup S^*\right)}\left(\boldsymbol{u}^{\dagger}\right)^{\top} \boldsymbol{A}_S \boldsymbol{v}^{\dagger}+\sup _{(\boldsymbol{u}, \boldsymbol{v}) \in \mathcal{B}^2\left(S \cup S^*\right)}\left(\boldsymbol{u}-\boldsymbol{u}^{\dagger}\right)^{\top} \boldsymbol{A}_S \boldsymbol{v}^{\dagger}                                                             \\
		                                   & \quad+\sup _{(\boldsymbol{u}, \boldsymbol{v}) \in \mathcal{B}^2\left(S \cup S^*\right)}\left(\boldsymbol{u}-\boldsymbol{u}^{\dagger}\right)^{\top} \boldsymbol{A}_S\left(\boldsymbol{v}-\boldsymbol{v}^{\dagger}\right)+\sup _{(\boldsymbol{u}, \boldsymbol{v}) \in \mathcal{B}^2\left(S \cup S^*\right)}\left(\boldsymbol{u}^{\dagger}\right)^{\top} \boldsymbol{A}_S\left(\boldsymbol{v}-\boldsymbol{v}^{\dagger}\right) \\
		                                   & \leq \sup _{k \in\left[N_S\right]}\left(\boldsymbol{u}_k^{(S)}\right)^{\top} \boldsymbol{A}_S \boldsymbol{v}_k^{(S)}+\left(\frac{1}{4}+\frac{1}{4}+\frac{1}{16}\right)\left\|\boldsymbol{A}_S\right\|_2,
	\end{aligned}
$$
which implies $\left\|\boldsymbol{A}_S\right\|_2 \leq  \sup _{k \in\left[N_S\right]}\left(\boldsymbol{u}_k^{(S)}\right)^{\top} \boldsymbol{A}_S \boldsymbol{v}_k^{(S)}$, thus
\begin{equation}
	\sup _{|S| \leq p}\left\|\boldsymbol{A}_S\right\|_2 \leq  \sup _{|S| \leq p, k \in\left[N_S\right]}4\left(\boldsymbol{u}_k^{(S)}\right)^{\top} \boldsymbol{A}_S\left(\boldsymbol{v}_k^{(S)}\right)\equiv\sup _{|S| \leq p, k \in\left[N_S\right]} 4Z(S, k).
	\label{eq:pf_lemma_a1_2}
\end{equation}
For fixed $k$ and $S$, $Z(S, k)$ can be written as the sum of independent zero-mean random variables as
\begin{align*}
	Z(S, k) & = \left(\boldsymbol{u}_k^{(S)}\right)^{\top}\left\{ \sum_{e\in\mathcal{E}}\omega^{(e)}\left(\boldsymbol\Sigma^{(e)}_{S\cup S*} - \widehat{\boldsymbol\Sigma}^{(e)}_{N^{(e)}, S\cup S*}\right)\right\}\left(\boldsymbol{v}_k^{(S)}\right)                                                                                                                                                                                                                                       \\
	        & = \left(\boldsymbol{u}_k^{(S)}\right)^{\top}\left\{ \sum_{e\in\mathcal{E}}\omega^{(e)}\left(\mathbb{E}[\boldsymbol{x}^{(e)}_{S\cup S^*} (\boldsymbol{x}^{(e)}_{S\cup S^*})^{\top} ] - \frac{1}{N^{(e)}} \sum_{\ell=1}^{N^{(e)}}[\boldsymbol{x}^{(e)}_{\ell}]_{S\cup S^*}[\boldsymbol{x}^{(e)}_{\ell}]_{S\cup S^*}^{\top}\right)\right\}\left(\boldsymbol{v}_k^{(S)}\right)                                                                                                     \\
	        & = \sum_{e\in\mathcal{E}}\sum_{\ell=1}^{N^{(e)}} \frac{\omega^{(e)}}{N^{(e)}}\left\{\mathbb{E}\left[\left(\boldsymbol{u}_k^{(S)}\right)^{\top}  [\boldsymbol{x}^{(e)}_{\ell}]_{S\cup S^*}[\boldsymbol{x}^{(e)}_{\ell}]_{S\cup S^*}^{\top}   \left(\boldsymbol{v}_k^{(S)}\right)\right] -  \left(\boldsymbol{u}_k^{(S)}\right)^{\top}  [\boldsymbol{x}^{(e)}_{\ell}]_{S\cup S^*}[\boldsymbol{x}^{(e)}_{\ell}]_{S\cup S^*}^{\top}   \left(\boldsymbol{v}_k^{(S)}\right) \right\}.
\end{align*}
Notice $\left\{\left(\boldsymbol{u}_k^{(S)}\right)^{\top}  [\boldsymbol{x}^{(e)}_{\ell}]_{S\cup S^*}\right\}\left\{[\boldsymbol{x}^{(e)}_{\ell}]_{S\cup S^*}^{\top}  \left(\boldsymbol{v}_k^{(S)}\right)\right\} $ is the product of two zero mean sub-Gaussian random variables with parameter $\kappa_U^{1/2}\sigma_x$, therefore $Z(S, k)$ is sum of sub-Exponential random variable with parameter $(c_1\frac{\omega^{(e)}}{N^{(e)}} \kappa_U\sigma_x^2, c_2 \frac{\omega^{(e)}}{N^{(e)}}\kappa_U\sigma_x^2)$. Therefore, we have
$$
	\mathbb{P}\left[|Z(S, k)| \geq c_3 \kappa_U \sigma_x^2\left(\sqrt{\frac{u}{N_\omega}}+\frac{u}{N_*}\right)\right] \leq 2 e^{-u}.
$$
Apply union bound over all $k\in[N_S]$ and $S\subset [p]$, we have
\begin{align*}
	 & \mathbb{P}\left[ \sup _{|S| \leq p, k \in\left[N_S\right]}4Z(S,k)\ge 4 c_3 \kappa_U \sigma_x^2\left(\sqrt{\frac{u}{N_\omega}}+\frac{u}{N_*}\right) \right]                         \\
	 & \quad \le \sum_{|S| \leq p, k\in[N_S]} \mathbb{P}\left[ \left\{Z(S,k)\ge c_3 \kappa_U \sigma_x^2\left(\sqrt{\frac{u}{N_\omega}}+\frac{u}{N_*}\right)\right\} \right]\le 2N e^{-u}.
\end{align*}
Let $u = t + \log (2N)\le C'(t+p)$, for some universal constant $C'$. Then we get
$$
	\mathbb{P}\left[ \sup _{|S| \leq p, k \in\left[N_S\right]}4Z(S,k)\le 4 C'c_3 \kappa_U \sigma_x^2\left(\sqrt{\frac{t + p}{N_\omega}}+\frac{t+p}{N_*}\right) \right] \ge 1-e^{-t}.
$$
Together with (\ref{eq:pf_lemma_a1_1}) and (\ref{eq:pf_lemma_a1_2}), we are done.

\noindent{\it Step 2.} UPPER BOUND ON $H_2^{(e)}(\boldsymbol{\beta})$. We argue in this step that the event
$$
	\mathcal{C}_{2,t} =\left\{\forall \boldsymbol{\beta}\in\mathbb{R}^p, H_2^{(e)}(\boldsymbol{\beta}) \le C_2\kappa_U^{1/2}\sigma_x\sigma_{\varepsilon}\|\boldsymbol{\beta} - \boldsymbol{\beta}^*\|_2\left\{\sqrt{\frac{t+p}{N_{\omega}}} + \frac{t+p}{N_*}\right\}\right\}.
$$
occurs with probability at least $1-e^{-t}$ for any $t>0$.

Observe
\begin{align}
	 & \sup _{\boldsymbol{\beta} \in \mathbb{R}^p, \boldsymbol{\beta} \neq \boldsymbol{\beta}^*} \frac{\left(\boldsymbol{\beta}-\boldsymbol{\beta}^*\right)^{\top}}{\left\|\boldsymbol{\beta}-\boldsymbol{\beta}^*\right\|_2} \sum_{e \in \mathcal{E}} \omega^{(e)} \left\{\mathbb{E}\left[\boldsymbol{x}^{(e)} \varepsilon^{(e)}\right]-\widehat{\mathbb{E}}_{N^{(e)}}\left[\boldsymbol{x}^{(e)} \varepsilon^{(e)}\right]\right\}\notag                                 \\
	 & \quad = \sup _{\boldsymbol{\beta} \in \mathbb{R}^p, \boldsymbol{\beta} \neq \boldsymbol{\beta}^*} \frac{\left(\boldsymbol{\beta}-\boldsymbol{\beta}^*\right)^{\top}}{\left\|\boldsymbol{\beta}-\boldsymbol{\beta}^*\right\|_2} \sum_{e \in \mathcal{E}} \omega^{(e)} \left\{\mathbb{E}\left[\boldsymbol{x}_{S\cup S^*}^{(e)} \varepsilon^{(e)}\right]-\widehat{\mathbb{E}}_{N^{(e)}}\left[\boldsymbol{x}_{S\cup S^*}^{(e)} \varepsilon^{(e)}\right]\right\}\notag \\
	 & \quad \le  \sup _{\boldsymbol{\beta} \in \mathbb{R}^p, \boldsymbol{\beta} \neq \boldsymbol{\beta}^*}  \left\|   \sum_{e \in \mathcal{E}} \omega^{(e)}  \left\{\mathbb{E}\left[\boldsymbol{x}_{S\cup S^*}^{(e)} \varepsilon^{(e)}\right]-\widehat{\mathbb{E}}_{N^{(e)}}\left[\boldsymbol{x}_{S\cup S^*}^{(e)} \varepsilon^{(e)}\right]\right\}   \right\|_2\notag                                                                                                  \\
	 & \quad \equiv \sup _{|S| \leq p}\left\|\boldsymbol{\xi}_S\right\|_2
	\label{eq:step2_observe}
\end{align}
For any $S \subset[p]$ with $|S| \leq s$, let $\boldsymbol{v}_1^{(S)}, \ldots, \boldsymbol{v}_{N_S}^{(S)}$ be an 1/4-covering of $\mathcal{B}\left(S \cup S^*\right)$, that is, for any $\boldsymbol{v} \in \mathcal{B}\left(S \cup S^*\right)$, there exists some $\pi(\boldsymbol{v}) \in\left[N_S\right]$ such that
$$
	\left\|\boldsymbol{v}-\boldsymbol{v}_{\pi(v)}^{(S)}\right\|_2 \leq 1 / 4.
$$
It follows from standard empirical process result that $N_S \leq 9^{\left|S \cup S^*\right|}$, then $N = \sum_{|S| \leq p} N_S \le 9^{s^*}(9e)^p$.

For any $S\in[p]$ with $|S| \leq s$, it follows from the variational representation of the $\ell_2$ norm that
$$
	\|\boldsymbol{\xi}_S\|_2=\sup _{\boldsymbol{v} \in \mathcal{B}\left(S \cup S^*\right)} \boldsymbol{v}^{\top} \boldsymbol{\xi}_S=\sup _{k \in\left[N_S\right]}\left(\boldsymbol{v}_k^{(S)}\right)^{\top} \boldsymbol{\xi}_S+\sup _{\boldsymbol{v} \in \mathcal{B}\left(S \cup S^*\right)}\left(\boldsymbol{v}-\boldsymbol{v}_{\pi(v)}^{(S)}\right)^{\top} \boldsymbol{\xi}_S \leq \sup _{k \in\left[N_S\right]}\left(\boldsymbol{v}_k^{(S)}\right)^{\top} \boldsymbol{\xi}_S+\frac{1}{4}\|\xi_S\|_2,
$$
which implies $\|\boldsymbol{\xi}\|_2 \leq 2 \sup _{k \in\left[N_S\right]}\left(\boldsymbol{v}_k^{(S)}\right)^{\top} \boldsymbol{\xi}_S$, thus
\begin{equation}
	\sup _{|S| \leq p}\left\|\boldsymbol{\xi}_S\right\|_2\leq  \sup _{|S| \leq p, k \in\left[N_S\right]} 2\left(\boldsymbol{v}_k^{(S)}\right)^{\top}\boldsymbol{\xi}_S \equiv  \sup _{|S| \leq p, k \in\left[N_S\right]} 2Z(S,k).
	\label{eq:step2_norm_trick}
\end{equation}
For given fixed $\boldsymbol{v}_k^{(S)} \in \mathcal{B}\left(S \cup S^*\right), Z(S, k)$ can be written as the sum of independent zero-mean random variables as
\begin{align*}
	Z(S, k) & = \left(\boldsymbol{v}_k^{(S)}\right)^{\top}\boldsymbol{\xi}_S                                                                                                                                                                                                                                                                                       \\
	        & = \left(\boldsymbol{v}_k^{(S)}\right)^{\top} \sum_{e \in \mathcal{E}} \omega^{(e)}  \left\{\mathbb{E}\left[\boldsymbol{x}_{S\cup S^*}^{(e)} \varepsilon^{(e)}\right]-\widehat{\mathbb{E}}_{N^{(e)}}\left[\boldsymbol{x}_{S\cup S^*}^{(e)} \varepsilon^{(e)}\right]\right\}                                                                           \\
	        & = \sum_{e\in\mathcal{E}}\sum_{\ell=1}^{N^{(e)}} \frac{\omega^{(e)}}{N^{(e)}}\left\{\mathbb{E}\left[\left(\boldsymbol{v}_k^{(S)}\right)^{\top}  [\boldsymbol{x}^{(e)}_{\ell}]_{S\cup S^*}\varepsilon_{\ell}^{(e)}  \right] -  \left(\boldsymbol{v}_k^{(S)}\right)^{\top}  [\boldsymbol{x}^{(e)}_{\ell}]_{S\cup S^*}\varepsilon_{\ell}^{(e)} \right\}.
\end{align*}
Observe that $\varepsilon_{\ell}^{(e)}$ is a zero-mean sub-Gaussian random variable with parameter $\sigma_{\varepsilon}$ by Condition 3.3.4, and $\left(\boldsymbol{v}_k^{(S)}\right)^{\top}  [\boldsymbol{x}^{(e)}_{\ell}]_{S\cup S^*}$ is a zero-mean sub-Gaussian random variable with parameter $\kappa_U^{1/2}\sigma_x$. Hence, $Z(S, k)$ is a sum of sub-Exponential random variables with parameter $(c_1\frac{\omega^{(e)}}{N^{(e)}} \kappa_U\sigma_x\sigma_{\varepsilon}, c_2 \frac{\omega^{(e)}}{N^{(e)}}\kappa_U\sigma_x\sigma_{\varepsilon})$. Thus we have
$$
	\mathbb{P}\left[|Z(S, k)| \geq c_3 \kappa_U \sigma_x\sigma_{\varepsilon} \left(\sqrt{\frac{u}{N_\omega}}+\frac{u}{N_*}\right)\right] \leq 2 e^{-u}.
$$
Apply union bound over all $k \in\left[N_S\right]$ and $S\subset [p]$, we have
\begin{align*}
	 & \mathbb{P}\left[\sup _{|S| \leq p, k \in\left[N_S\right]} 2 Z(S, k) \geq 2 c_3 \kappa_U \sigma_x\sigma_{\varepsilon}\left(\sqrt{\frac{u}{N_\omega}}+\frac{u}{N_*}\right)\right]                                       \\
	 & \quad \leq \sum_{|S| \leq p, k \in\left[N_S\right]} \mathbb{P}\left[\left\{Z(S, k) \geq c_3 \kappa_U \sigma_x\sigma_{\varepsilon}\left(\sqrt{\frac{u}{N_\omega}}+\frac{u}{N_*}\right)\right\}\right] \leq 2 N e^{-u}.
\end{align*}
Let $u=t+\log (2 N) < C'(t+p)$, we get
$$
	\mathbb{P}\left[\sup _{|S| \leq p, k \in\left[N_S\right]} 2 Z(S, k) \leq 2 c_3 \kappa_U \sigma_x\sigma_{\varepsilon}\left(\sqrt{\frac{t+p}{N_\omega}}+\frac{t+p}{N_*}\right)\right] \geq 1-e^{-t}.
$$
Together with (\ref{eq:step2_observe}) and (\ref{eq:step2_norm_trick}), we are done.

\noindent{\it Step 3.} UPPER BOUND ON $H_3^{(e)}(\boldsymbol{\beta})$.  Define the event
$$
	\mathcal{C}_{3,t} =\left\{\forall \boldsymbol{\beta}\in\mathbb{R}^p, H_3^{(e)}(\boldsymbol{\beta}) \le C_3\kappa_U^{1/2}\sigma_x\sigma_{z}\|\boldsymbol{\beta} - \boldsymbol{\beta}^*\|_2\left\{\sqrt{\frac{t+p}{m_\omega^{\widehat{\tau}^2}}} + \frac{t+p}{m_*^{\widehat{\tau}}}\right\} \right\}.
$$
We will claim that $\mathbb{P}(\mathcal{C}_{3,t})\ge1-e^{-t}$ for any $t>0$. Observe
\begin{align}
	 & \sup _{\boldsymbol{\beta} \in \mathbb{R}^p, \boldsymbol{\beta} \neq \boldsymbol{\beta}^*} \frac{\left(\boldsymbol{\beta}-\boldsymbol{\beta}^*\right)^{\top}}{\left\|\boldsymbol{\beta}-\boldsymbol{\beta}^*\right\|_2} \sum_{e \in \mathcal{E}} \omega^{(e)}\widehat{\tau}^{(e)}\left\{\mathbb{E}\left[\boldsymbol{x}^{(e)} (z^{(e)}-\eta^{(e)})\right]-\widehat{\mathbb{E}}_{m^{(e)}}\left[\boldsymbol{x}^{(e)} (z^{(e)}-\eta^{(e)})\right]\right\}\notag                                \\
	 & \quad =\sup _{\boldsymbol{\beta} \in \mathbb{R}^p, \boldsymbol{\beta} \neq \boldsymbol{\beta}^*} \frac{\left(\boldsymbol{\beta}-\boldsymbol{\beta}^*\right)^{\top}}{\left\|\boldsymbol{\beta}-\boldsymbol{\beta}^*\right\|_2} \sum_{e \in \mathcal{E}} \omega^{(e)}\widehat{\tau}^{(e)}\left\{\mathbb{E}\left[\boldsymbol{x}_{S\cup S^*}^{(e)} (z^{(e)}-\eta^{(e)})\right]-\widehat{\mathbb{E}}_{m^{(e)}}\left[\boldsymbol{x}_{S\cup S^*}^{(e)} (z^{(e)}-\eta^{(e)})\right]\right\}\notag \\
	 & \quad \le \sup _{\boldsymbol{\beta} \in \mathbb{R}^p, \boldsymbol{\beta} \neq \boldsymbol{\beta}^*} \left\|\sum_{e \in \mathcal{E}} \omega^{(e)}\widehat{\tau}^{(e)}\left\{\mathbb{E}\left[\boldsymbol{x}_{S\cup S^*}^{(e)} (z^{(e)}-\eta^{(e)})\right]-\widehat{\mathbb{E}}_{m^{(e)}}\left[\boldsymbol{x}_{S\cup S^*}^{(e)} (z^{(e)}-\eta^{(e)})\right]\right\}\right\|_2\notag                                                                                                          \\
	 & \quad \equiv \sup _{|S| \leq p}\left\|\boldsymbol{\xi}_S\right\|_2.
	\label{eq:step3_observe}
\end{align}
For any $S \subset[p]$ with $|S| \leq s$, let $\boldsymbol{v}_1^{(S)}, \ldots, \boldsymbol{v}_{N_S}^{(S)}$ be an 1/4-covering of $\mathcal{B}\left(S \cup S^*\right)$. Follows from the variational representation of the $\ell_2$ norm that
\begin{equation}
	\sup _{|S| \leq p}\left\|\boldsymbol{\xi}_S\right\|_2\leq  \sup _{|S| \leq p, k \in\left[N_S\right]} 2\left(\boldsymbol{v}_k^{(S)}\right)^{\top}\boldsymbol{\xi}_S \equiv  \sup _{|S| \leq p, k \in\left[N_S\right]} 2Z(S,k).
	\label{eq:step3_norm_trick}
\end{equation}
For given fixed $\boldsymbol{v}_k^{(S)} \in \mathcal{B}\left(S \cup S^*\right), Z(S, k)$ can be written as the sum of independent zero-mean random variables as
\begin{align*}
	Z(S, k) & = \left(\boldsymbol{v}_k^{(S)}\right)^{\top}\boldsymbol{\xi}_S                                                                                                                                                                                                                                                                                                                     \\
	        & = \left(\boldsymbol{v}_k^{(S)}\right)^{\top} \sum_{e \in \mathcal{E}} \omega^{(e)} \widehat{\tau}^{(e)} \left\{\mathbb{E}\left[\boldsymbol{x}_{S\cup S^*}^{(e)}(z^{(e)} - \eta^{(e)})\right]-\widehat{\mathbb{E}}_{m^{(e)}}\left[\boldsymbol{x}_{S\cup S^*}^{(e)} (z^{(e)} - \eta^{(e)})\right]\right\}                                                                            \\
	        & = \sum_{e\in\mathcal{E}}\sum_{\ell=1}^{m^{(e)}} \frac{\omega^{(e)}\widehat{\tau}^{(e)}}{m^{(e)}}\left\{\mathbb{E}\left[\left(\boldsymbol{v}_k^{(S)}\right)^{\top}  [\boldsymbol{x}^{(e)}_{\ell}]_{S\cup S^*}(z_{\ell}^{(e)} - \eta^{(e)})  \right] -  \left(\boldsymbol{v}_k^{(S)}\right)^{\top}  [\boldsymbol{x}^{(e)}_{\ell}]_{S\cup S^*}(z_{\ell}^{(e)} - \eta^{(e)}) \right\}.
\end{align*}
Observe that $(z_{\ell}^{(e)} - \eta^{(e)})$ is a zero-mean sub-Gaussian random variable with parameter $\sigma_{z}$ by Condition 3.3.5, and $\left(\boldsymbol{v}_k^{(S)}\right)^{\top}  [\boldsymbol{x}^{(e)}_{\ell}]_{S\cup S^*}$ is a zero-mean sub-Gaussian random variable with parameter $\kappa_U^{1/2}\sigma_x$. Hence, $Z(S, k)$ is a sum of sub-Exponential random variables with parameter $(c_1\frac{\omega^{(e)}\widehat{\tau}^{(e)}}{m^{(e)}} \kappa_U\sigma_x\sigma_{z}, c_2 \frac{\omega^{(e)}\widehat{\tau}^{(e)}}{m^{(e)}}\kappa_U\sigma_x\sigma_{z})$. Thus we have
$$
	\mathbb{P}\left[|Z(S, k)| \geq c_3 \kappa_U \sigma_x\sigma_{z} \left(\sqrt{\frac{u}{m_\omega^{\widehat{\tau}^2}}}+\frac{u}{m_*^{\widehat{\tau}}}\right)\right] \leq 2 e^{-u}.
$$
Apply union bound over all $k \in\left[N_S\right]$ and $S\subset [p]$, we have
\begin{align*}
	 & \mathbb{P}\left[\sup _{|S| \leq p, k \in\left[N_S\right]} 2 Z(S, k) \geq 2 c_3 \kappa_U \sigma_x\sigma_{z}\left(\sqrt{\frac{u}{m_\omega^{\widehat{\tau}^2}}}+\frac{u}{m_*^{\widehat{\tau}}}\right)\right]                                       \\
	 & \quad \leq \sum_{|S| \leq p, k \in\left[N_S\right]} \mathbb{P}\left[\left\{Z(S, k) \geq c_3 \kappa_U \sigma_x\sigma_{z}\left(\sqrt{\frac{u}{m_\omega^{\widehat{\tau}^2}}}+\frac{u}{m_*^{\widehat{\tau}}}\right)\right\}\right] \leq 2 N e^{-u}.
\end{align*}
Let $u=t+\log (2 N)\le C'(t+p)$, we get
$$
	\mathbb{P}\left[\sup _{|S| \leq p, k \in\left[N_S\right]} 2 Z(S, k) \leq 2 C'c_3 \kappa_U \sigma_x\sigma_z\left(\sqrt{\frac{t+p}{m_\omega^{\widehat{\tau}^2}}}+\frac{t+p}{m_*^{\widehat{\tau}}}\right)\right] \geq 1-e^{-t}.
$$
Together with (\ref{eq:step3_observe}) and (\ref{eq:step3_norm_trick}), we are done.

\noindent{\it Step 4.} UPPER BOUND ON $H_4^{(e)}(\boldsymbol{\beta})$. The goal of this step is to derive a high-probability bound for the event
$$
	\mathcal{C}_{4,t} =\left\{\forall \boldsymbol{\beta}\in\mathbb{R}^p, H_4^{(e)}(\boldsymbol{\beta}) \le C_4\kappa_U^{1/2}\sigma_x\|\boldsymbol{\beta} - \boldsymbol{\beta}^*\|_2\left\{\sqrt{\frac{t+p}{m_\omega^{\widehat{\tau}^2,|\eta|^2}}} + \frac{t+p}{m_*^{\widehat{\tau},|\eta|}}\right\} \right\}.
$$
Observe
\begin{align*}
	 & \sup _{\boldsymbol{\beta} \in \mathbb{R}^p, \boldsymbol{\beta} \neq \boldsymbol{\beta}^*} \frac{\left(\boldsymbol{\beta}-\boldsymbol{\beta}^*\right)^{\top}}{\left\|\boldsymbol{\beta}-\boldsymbol{\beta}^*\right\|_2} \sum_{e \in \mathcal{E}} \omega^{(e)}\widehat{\tau}^{(e)}\eta^{(e)}\left\{\mathbb{E}[\boldsymbol{x}^{(e)}]-\widehat{\mathbb{E}}_{m^{(e)}}[\tilde{\boldsymbol{x}}^{(e)}]\right\}                    \\
	 & =\sup _{\boldsymbol{\beta} \in \mathbb{R}^p, \boldsymbol{\beta} \neq \boldsymbol{\beta}^*} \frac{\left(\boldsymbol{\beta}-\boldsymbol{\beta}^*\right)^{\top}}{\left\|\boldsymbol{\beta}-\boldsymbol{\beta}^*\right\|_2} \sum_{e \in \mathcal{E}} \omega^{(e)}\widehat{\tau}^{(e)}\eta^{(e)}\left\{\mathbb{E}[\boldsymbol{x}_{S\cup S^*}^{(e)}]-\widehat{\mathbb{E}}_{m^{(e)}}[{\boldsymbol{x}}_{S\cup S^*}^{(e)}]\right\} \\
	 & \leq \sup _{\boldsymbol{\beta} \in \mathbb{R}^p, \boldsymbol{\beta} \neq \boldsymbol{\beta}^*}\left\|\sum_{e \in \mathcal{E}} \omega^{(e)}\widehat{\tau}^{(e)}\eta^{(e)}\left\{\mathbb{E}[\boldsymbol{x}_{S\cup S^*}^{(e)}]-\widehat{\mathbb{E}}_{m^{(e)}}[{\boldsymbol{x}}_{S\cup S^*}^{(e)}]\right\}\right\|_2                                                                                                          \\
	 & \equiv \sup _{|S| \leq p}\left\|\boldsymbol{\xi}_S\right\|_2 .
\end{align*}
The claim can be shown by using Condition \ref{cond:subg_x} and following similar procedure as previous steps.

\noindent{\it Step 5.} UPPER BOUND ON $H_5^{(e)}(\boldsymbol{\beta})$. We will claim that $\mathbb{P}(\mathcal{C}_{5,t})\ge1-e^{-t}$ for any $t>0$, where
$$
	\mathcal{C}_{5,t} =\left\{\forall \boldsymbol{\beta}\in\mathbb{R}^p, H_5^{(e)}(\boldsymbol{\beta}) \le C_5\kappa_U^{1/2}\sigma_x\sigma_{z}\|\boldsymbol{\beta} - \boldsymbol{\beta}^*\|_2\left\{\sqrt{\frac{t+p}{ n_\omega^{\widehat{\tau}^2}}} + \frac{t+p}{n_*^{\widehat{\tau}}}\right\} \right\}.
$$
We omit the proof since it is very similar as the procedure in step 3.

\noindent{\it Step 6.} UPPER BOUND ON $H_6^{(e)}(\boldsymbol{\beta})$. We will claim that $\mathbb{P}(\mathcal{C}_{6,t})\ge1-e^{-t}$ for any $t>0$, where
$$
	\mathcal{C}_{6,t} =\left\{\forall \boldsymbol{\beta}\in\mathbb{R}^p, H_6^{(e)}(\boldsymbol{\beta}) \le C_6\kappa_U^{1/2}\sigma_x\|\boldsymbol{\beta} - \boldsymbol{\beta}^*\|_2\left\{\sqrt{\frac{t+p}{n_\omega^{\widehat{\tau}^2,|\eta|^2}}} + \frac{t+p}{n_*^{\widehat{\tau},|\eta|}}\right\} \right\}.
$$
The proof procedure in this step is exactly same as the one in step 4 except the sample size now is depends on $n^{(e)}, \forall e\in\mathcal{E}$.

\noindent{\it Step 7.}  CONCLUSION. Under the event $\bigcap_{k=1}^5 C_{k, t}$, which occurs with probability $1-5e^{-t}, \forall t>0$, the following inequality holds.

\begin{align}
	 & \frac{1}{c_1}\left|   \mathrm{R}(\boldsymbol{\beta})-\mathrm{R}(\boldsymbol{\beta}^*)   - \widehat{\mathrm{R}}_{\mathrm{Adj}}(\boldsymbol{\beta})+\widehat{\mathrm{R}}_{\mathrm{Adj}}(\boldsymbol{\beta}^*)    \right|\le  \notag                                                                                                                                                                                      \\
	 & \quad  \quad  \kappa_U\sigma_x^2\|\boldsymbol{\beta} - \boldsymbol{\beta}^*\|_2^2\left\{\sqrt{\frac{t+p}{N_{\omega}}} + \frac{t+p}{N_*}\right\} +\kappa_U^{1/2}\sigma_x\sigma_{\varepsilon}\|\boldsymbol{\beta} - \boldsymbol{\beta}^*\|_2\left\{\sqrt{\frac{t+p}{N_{\omega}}} + \frac{t+p}{N_*}\right\}\notag                                                                                                         \\
	 & \quad\quad  \quad \quad +\kappa_U^{1/2}\sigma_x\sigma_{z}\|\boldsymbol{\beta} - \boldsymbol{\beta}^*\|_2\left\{\sqrt{\frac{t+p}{m_\omega^{\widehat{\tau}^2}}} + \frac{t+p}{m_*^{\widehat{\tau}}}\right\} +
	\kappa_U^{1/2}\sigma_x\|\boldsymbol{\beta} - \boldsymbol{\beta}^*\|_2\left\{\sqrt{\frac{t+p}{m_\omega^{\widehat{\tau}^2,|\eta|^2}}} + \frac{t+p}{m_*^{\widehat{\tau},|\eta|}}\right\}\notag                                                                                                                                                                                                                               \\
	 & \quad\quad  \quad \quad + \kappa_U^{1/2}\sigma_x\sigma_{z}\|\boldsymbol{\beta} - \boldsymbol{\beta}^*\|_2\left\{\sqrt{\frac{t+p}{ n_\omega^{\widehat{\tau}^2}}} + \frac{t+p}{n_*^{\widehat{\tau}}}\right\}  + \kappa_U^{1/2}\sigma_x\|\boldsymbol{\beta} - \boldsymbol{\beta}^*\|_2\left\{\sqrt{\frac{t+p}{n_\omega^{\widehat{\tau}^2,|\eta|^2}}} + \frac{t+p}{n_*^{\widehat{\tau},|\eta|}}\right\}\label{eq:lemma_a1}
\end{align}

\noindent{\it Step 8.} FURTHER SIMPLICATION. First note that by Lemma \ref{lemma:weighted_power_mean_inequality} and Remark \ref{remark:specific_weighted_power_mean_inequality}, we have the following inequalities:
\begin{equation}
	N_{\omega} > N_*\quad \text{and}\quad n_\omega^{\widehat{\tau}^2,|\eta|^2} > n_*^{\widehat{\tau},|\eta|}\quad \text{and}\quad m_\omega^{\widehat{\tau}^2,|\eta|^2} > m_*^{\widehat{\tau},|\eta|}
\end{equation}
Thus we have
\begin{align}
	 & \frac{1}{c_1}\left|   \mathrm{R}(\boldsymbol{\beta})-\mathrm{R}(\boldsymbol{\beta}^*)   - \widehat{\mathrm{R}}_{\mathrm{Adj}}(\boldsymbol{\beta})+\widehat{\mathrm{R}}_{\mathrm{Adj}}(\boldsymbol{\beta}^*)    \right|\le  \notag                                                                 \\
	 & \quad  \quad  \kappa_U\sigma_x^2\|\boldsymbol{\beta} - \boldsymbol{\beta}^*\|_2^2\,\sqrt{\frac{t+p}{N_*}}  +\kappa_U^{1/2}\sigma_x\sigma_{\varepsilon}\|\boldsymbol{\beta} - \boldsymbol{\beta}^*\|_2\,\sqrt{\frac{t+p}{N_*}}\notag                                                               \\
	 & \quad\quad  \quad \quad +\kappa_U^{1/2}\sigma_x\sigma_{z}\|\boldsymbol{\beta} - \boldsymbol{\beta}^*\|_2\,\sqrt{\frac{t+p}{m_*^{\widehat{\tau}}}}  +
	\kappa_U^{1/2}\sigma_x\|\boldsymbol{\beta} - \boldsymbol{\beta}^*\|_2\,\sqrt{\frac{t+p}{m_*^{\widehat{\tau},|\eta|}}}\notag                                                                                                                                                                          \\
	 & \quad\quad  \quad \quad + \kappa_U^{1/2}\sigma_x\sigma_{z}\|\boldsymbol{\beta} - \boldsymbol{\beta}^*\|_2\,\sqrt{\frac{t+p}{n_*^{\widehat{\tau}}}}  + \kappa_U^{1/2}\sigma_x\|\boldsymbol{\beta} - \boldsymbol{\beta}^*\|_2\,\sqrt{\frac{t+p}{n_*^{\widehat{\tau},|\eta|}}}.\label{eq:lemma_a1_2}
\end{align}

\begin{itemize}
	\item \text{Case 1:} $\forall e\in\mathcal{E}, {n^{(e)}}/{m^{(e)}} > 1 $, i.e. $\widehat{\tau}_{\max} \le 1/2$, and $ \eta_{\mathrm{max}} \ge 1 $,
	\item \text{Case 2:} $\forall e\in\mathcal{E}, {n^{(e)}}/{m^{(e)}} > 1 $, i.e. $\widehat{\tau}_{\max} \le 1/2$, and $ \eta_{\mathrm{max}} \ge 1 $,
	\item \text{Case 3:} $\forall e\in\mathcal{E}, {n^{(e)}}/{m^{(e)}} < 1 $, i.e. $\widehat{\tau}_{\min} > 1/2$, and $ \eta_{\mathrm{max}} < 1 $,
	\item \text{Case 4:} $\forall e\in\mathcal{E}, {n^{(e)}}/{m^{(e)}} < 1 $, i.e. $\widehat{\tau}_{\min} > 1/2$, and $ \eta_{\mathrm{max}} < 1 $.
\end{itemize}

For each case, we present the implications for the bound accordingly. First let's plug $\widehat{\tau}^{(e)} = m^{(e)}/ N^{(e)}$ in the quantities $m_*^{\widehat{\tau}}, m_*^{\widehat{\tau},|\eta|}, n_*^{\widehat{\tau}}$, and $n_*^{\widehat{\tau},|\eta|}$, we get
\begin{equation}\label{eq:replug_hat_tau_samples}
	m_*^{\widehat{\tau}} = \min_{e\in\mathcal{E}} \frac{N^{(e)}}{\omega^{(e)}} = N_*, \quad  m_*^{\widehat{\tau},|\eta|} = \min_{e\in\mathcal{E}} \frac{1}{|\eta^{(e)}|}\frac{N^{(e)}}{\omega^{(e)}},\quad n_*^{\widehat{\tau}} =\min_{e\in\mathcal{E}} \frac{n^{(e)}}{m^{(e)}}\frac{N^{(e)}}{\omega^{(e)}},\quad  n_*^{\widehat{\tau},|\eta|} = \min_{e\in\mathcal{E}} \frac{n^{(e)}}{m^{(e)}} \frac{1}{|\eta^{(e)}|}\frac{N^{(e)}}{\omega^{(e)}}.
\end{equation}

Recall that $n^{(e)}$ is the number of observations with label observed in environment $e$, and $m^{(e)}$ is the number of missing-label observations. Thus the ratio $n^{(e)}/m^{(e)}$ represents the missing mechanism, or the balance within environment $e$. When there are more observations with label recorded than the ones without, $\widehat{\tau}_{\max} < 1/2$, so $ n_*^{\widehat{\tau}} > m_*^{\widehat{\tau}}$ and $n_*^{\widehat{\tau},|\eta|}> m_*^{\widehat{\tau},|\eta|}$. Together with the fact that $m_*^{\widehat{\tau}}=N_*$, we can continue rewriting the bound (\ref{eq:lemma_a1_2}) to be
\begin{align}
	 & \frac{1}{c_1}\left|   \mathrm{R}(\boldsymbol{\beta})-\mathrm{R}(\boldsymbol{\beta}^*)   - \widehat{\mathrm{R}}_{\mathrm{Adj}}(\boldsymbol{\beta})+\widehat{\mathrm{R}}_{\mathrm{Adj}}(\boldsymbol{\beta}^*)    \right|\le  \notag                         \\
	 & \quad  \quad  \kappa_U\sigma_x^2\|\boldsymbol{\beta} - \boldsymbol{\beta}^*\|_2^2\,\sqrt{\frac{t+p}{N_*}}+\kappa_U^{1/2}\sigma_x\sigma_{z}\sigma_{\varepsilon}\|\boldsymbol{\beta} - \boldsymbol{\beta}^*\|_2\,\sqrt{\frac{t+p}{m_*^{\widehat{\tau}}}}  +
	\kappa_U^{1/2}\sigma_x\|\boldsymbol{\beta} - \boldsymbol{\beta}^*\|_2\,\sqrt{\frac{t+p}{m_*^{\widehat{\tau},|\eta|}}}\label{eq:lemma_a1_3}
\end{align}
Furthermore, if we assume $\eta_{\mathrm{max}} > 1$, from equalities (\ref{eq:replug_hat_tau_samples}), we get
\begin{align}
	 & \frac{1}{c_1}\left|   \mathrm{R}(\boldsymbol{\beta})-\mathrm{R}(\boldsymbol{\beta}^*)   - \widehat{\mathrm{R}}_{\mathrm{Adj}}(\boldsymbol{\beta})+\widehat{\mathrm{R}}_{\mathrm{Adj}}(\boldsymbol{\beta}^*)    \right|\le  \kappa_U\sigma_x^2\|\boldsymbol{\beta} - \boldsymbol{\beta}^*\|_2^2\,\sqrt{\frac{t+p}{N_*}} +
	\kappa_U^{1/2}\sigma_x\sigma_{z}\sigma_{\varepsilon}\|\boldsymbol{\beta} - \boldsymbol{\beta}^*\|_2\,\sqrt{\frac{t+p}{m_*^{\widehat{\tau},|\eta|}}},\label{eq:lemma_a1_4}
\end{align}
which is the upper bound in Case 1.

On the other hand, if we assume $\eta_{\mathrm{max}} \le 1$, then we have
\begin{align}
	\frac{1}{c_1}\left|   \mathrm{R}(\boldsymbol{\beta})-\mathrm{R}(\boldsymbol{\beta}^*)   - \widehat{\mathrm{R}}_{\mathrm{Adj}}(\boldsymbol{\beta})+\widehat{\mathrm{R}}_{\mathrm{Adj}}(\boldsymbol{\beta}^*)    \right| & \le   \kappa_U\sigma_x^2\|\boldsymbol{\beta} - \boldsymbol{\beta}^*\|_2^2\,\sqrt{\frac{t+p}{N_*}}+\kappa_U^{1/2}\sigma_x\sigma_{z}\sigma_{\varepsilon}\|\boldsymbol{\beta} - \boldsymbol{\beta}^*\|_2\,\sqrt{\frac{t+p}{m_*^{\widehat{\tau}}}}\notag     \\
	                                                                                                                                                                                                                       & \quad\le   \kappa_U\sigma_x^2\|\boldsymbol{\beta} - \boldsymbol{\beta}^*\|_2^2\,\sqrt{\frac{t+p}{N_*}}+\kappa_U^{1/2}\sigma_x\sigma_{z}\sigma_{\varepsilon}\|\boldsymbol{\beta} - \boldsymbol{\beta}^*\|_2\,\sqrt{\frac{t+p}{N_*}}.\label{eq:lemma_a1_5}
\end{align}
where the last line follows from the fact that $m_*^{\widehat{\tau}} = N_*$ in (\ref{eq:replug_hat_tau_samples}). Hence we obtain the upper bound in Case 2.

Now, suppose we are in the regime that there are less labeled observations than the covariates-only observations in every environment, then $\widehat{\tau}_{\max} \ge 1/2$, so $ n_*^{\widehat{\tau}} < m_*^{\widehat{\tau}}$ and $n_*^{\widehat{\tau},|\eta|}< m_*^{\widehat{\tau},|\eta|}$.
Together with the fact that $m_*^{\widehat{\tau}}=N_*$, we can continue rewriting the bound (\ref{eq:lemma_a1_2}) to be
\begin{align}
	 & \frac{1}{c_1}\left|   \mathrm{R}(\boldsymbol{\beta})-\mathrm{R}(\boldsymbol{\beta}^*)   - \widehat{\mathrm{R}}_{\mathrm{Adj}}(\boldsymbol{\beta})+\widehat{\mathrm{R}}_{\mathrm{Adj}}(\boldsymbol{\beta}^*)    \right|\le  \notag                         \\
	 & \quad  \quad  \kappa_U\sigma_x^2\|\boldsymbol{\beta} - \boldsymbol{\beta}^*\|_2^2\,\sqrt{\frac{t+p}{N_*}}+\kappa_U^{1/2}\sigma_x\sigma_{z}\sigma_{\varepsilon}\|\boldsymbol{\beta} - \boldsymbol{\beta}^*\|_2\,\sqrt{\frac{t+p}{n_*^{\widehat{\tau}}}}  +
	\kappa_U^{1/2}\sigma_x\|\boldsymbol{\beta} - \boldsymbol{\beta}^*\|_2\,\sqrt{\frac{t+p}{n_*^{\widehat{\tau},|\eta|}}}\label{eq:lemma_a1_6}
\end{align}

Furthermore, if we assume $\eta_{\mathrm{max}} > 1$, from equalities (\ref{eq:replug_hat_tau_samples}), we get
\begin{align}
	 & \frac{1}{c_1}\left|   \mathrm{R}(\boldsymbol{\beta})-\mathrm{R}(\boldsymbol{\beta}^*)   - \widehat{\mathrm{R}}_{\mathrm{Adj}}(\boldsymbol{\beta})+\widehat{\mathrm{R}}_{\mathrm{Adj}}(\boldsymbol{\beta}^*)    \right|\le  \kappa_U\sigma_x^2\|\boldsymbol{\beta} - \boldsymbol{\beta}^*\|_2^2\,\sqrt{\frac{t+p}{N_*}} +
	\kappa_U^{1/2}\sigma_x\sigma_{z}\sigma_{\varepsilon}\|\boldsymbol{\beta} - \boldsymbol{\beta}^*\|_2\,\sqrt{\frac{t+p}{n_*^{\widehat{\tau},|\eta|}}},\label{eq:lemma_a1_7}
\end{align}
which is the upper bound in Case 3.

On the other hand, if we assume $\eta_{\mathrm{max}} \le 1$, then we have
\begin{align}
	 & \frac{1}{c_1}\left|   \mathrm{R}(\boldsymbol{\beta})-\mathrm{R}(\boldsymbol{\beta}^*)   - \widehat{\mathrm{R}}_{\mathrm{Adj}}(\boldsymbol{\beta})+\widehat{\mathrm{R}}_{\mathrm{Adj}}(\boldsymbol{\beta}^*)    \right|\le   \kappa_U\sigma_x^2\|\boldsymbol{\beta} - \boldsymbol{\beta}^*\|_2^2\,\sqrt{\frac{t+p}{N_*}}+\kappa_U^{1/2}\sigma_x\sigma_{z}\sigma_{\varepsilon}\|\boldsymbol{\beta} - \boldsymbol{\beta}^*\|_2\,\sqrt{\frac{t+p}{n_*^{\widehat{\tau}}}}\label{eq:lemma_a1_8}
\end{align}
which is the upper bound in Case 4.

This completes the proof.

\qed

\subsection{Proof of Lemma \ref{lemma:one_sided_bound_J}}

Let $S = \text{supp}(\boldsymbol{\beta}), I = S\cap S^*$. It follows from the fact that $J(\boldsymbol{\beta}^*)=0$, the definition of $\mathrm{J}$ and $ \widehat{\mathrm{J}}_{\mathrm{Adj}}$, we have
$$
	\mathrm{J}(\boldsymbol{\beta}) - \mathrm{J}(\boldsymbol{\beta}^*) - \widehat{\mathrm{J}}_{\mathrm{Adj}}(\boldsymbol{\beta}) +  \widehat{\mathrm{J}}_{\mathrm{Adj}}(\boldsymbol{\beta}^*)  = \sum_{e\in\mathcal{E}} \frac{\omega^{(e)}}{4}\{ \|\nabla_S \mathrm{R}^{(e)}(\boldsymbol{\beta}) \|_2^2 - \|\nabla_S  \widehat{\mathrm{R}}^{\mathrm{PP},(e)}(\boldsymbol{\beta}) \|_2^2 + \|\nabla_{S^*} \widehat{\mathrm{R}}^{\mathrm{PP},(e)}(\boldsymbol{\beta}^*)\|_2^2  \}.
$$
Let's focus on an arbitrary $e\in\mathcal{E}$,

\begin{align}
	 & \frac{1}{8}\left\{     \|\nabla_S \mathrm{R}^{(e)}(\boldsymbol{\beta}) \|_2^2 - \|\nabla_S  \widehat{\mathrm{R}}^{\mathrm{PP},(e)}(\boldsymbol{\beta}) \|_2^2 + \|\nabla_{S^*} \widehat{\mathrm{R}}^{\mathrm{PP},(e)}(\boldsymbol{\beta}^*)\|_2^2     \right\}\notag                                                              \\
	 & = -\frac{1}{4} \{\nabla_S \mathrm{R}^{(e)}(\boldsymbol{\beta})\}^{\top} \left\{  \nabla_S  \widehat{\mathrm{R}}^{\mathrm{PP},(e)}(\boldsymbol{\beta}) - \nabla_S\mathrm{R}^{(e)}(\boldsymbol{\beta}) \right\} + \frac{1}{8}\|  \nabla_{S^*\setminus S} \widehat{\mathrm{R}}^{\mathrm{PP},(e)}(\boldsymbol{\beta}^*)  \|_2^2\notag \\
	 & \quad -\frac{1}{8}\|   \nabla_{S\setminus S^*} \widehat{\mathrm{R}}^{\mathrm{PP},(e)}(\boldsymbol{\beta}) -   \nabla_{S\setminus S^*} \mathrm{R}^{(e)}(\boldsymbol{\beta})      \|_2^2\notag                                                                                                                                      \\
	 & \quad + \frac{1}{8}\left\{\| \nabla_{I} \widehat{\mathrm{R}}^{\mathrm{PP},(e)}(\boldsymbol{\beta}^*) \|_2^2  - \|\nabla_{I} \widehat{\mathrm{R}}^{\mathrm{PP},(e)}(\boldsymbol{\beta}) - \nabla_{I} \mathrm{R}^{(e)}(\boldsymbol{\beta})\|_2^2 \right\}\notag                                                                     \\
\end{align}
\begin{align}
	 & \le (\boldsymbol{\beta} - \boldsymbol{\beta}^*)^{\top}\boldsymbol{\Sigma}^{(e)}_{:,S}  \left\{   \widehat{\mathbb{E}}_{N^{(e)}} [{\boldsymbol{x}}^{(e)}_S {\varepsilon}^{(e)} ] - \mathbb{E}[\boldsymbol{x}^{(e)}_S \varepsilon^{(e)} ]   \right\} - \{\mathbb{E}[\boldsymbol{x}^{(e)}_S\varepsilon^{(e)}]\}^{\top}\left\{   \widehat{\mathbb{E}}_{N^{(e)}} [{\boldsymbol{x}}^{(e)}_S {\varepsilon}^{(e)} ] - \mathbb{E}[\boldsymbol{x}^{(e)}_S \varepsilon^{(e)} ]   \right\}\notag \\
	 & \quad - (\boldsymbol{\beta} - \boldsymbol{\beta}^*)^{\top} \boldsymbol{\Sigma}^{(e)}_{:,S}  \left\{    \widehat{\mathbb{E}}_{N^{(e)}}[{\boldsymbol{x}}^{(e)}_S ({\boldsymbol{x}}^{(e)})^{\top}  (\boldsymbol{\beta} - \boldsymbol{\beta}^*)]- \boldsymbol{\Sigma}^{(e)}_{S,:}  (\boldsymbol{\beta} - \boldsymbol{\beta}^*)   \right\}\notag                                                                                                                                          \\
	 & \quad + \{\mathbb{E}[\boldsymbol{x}^{(e)}_S\varepsilon^{(e)}]\}^{\top} \left\{    \widehat{\mathbb{E}}_{N^{(e)}}[{\boldsymbol{x}}^{(e)}_S ({\boldsymbol{x}}^{(e)})^{\top}  (\boldsymbol{\beta} - \boldsymbol{\beta}^*)]- \boldsymbol{\Sigma}^{(e)}_{S,:} (\boldsymbol{\beta} - \boldsymbol{\beta}^*)    \right\}\notag                                                                                                                                                               \\
	 & \quad +  (\boldsymbol{\beta} - \boldsymbol{\beta}^*)^{\top} \boldsymbol{\Sigma}^{(e)}_{:,S}  \left\{ \widehat{\tau}^{(e)}  \left(\widehat{\mathbb{E}}_{m^{(e)}}[{\boldsymbol{x}}^{(e)}_S(z^{(e)} - \eta^{(e)})]  -   \mathbb{E}[\boldsymbol{x}^{(e)}_S(z^{(e)} - \eta^{(e)})]\right) \right\}\notag                                                                                                                                                                                  \\
	 & \quad + (\boldsymbol{\beta} - \boldsymbol{\beta}^*)^{\top}  \boldsymbol{\Sigma}^{(e)}_{:,S}  \left\{ \widehat{\tau}^{(e)}\left( -  \widehat{\mathbb{E}}_{n^{(e)}}[\boldsymbol{x}^{(e)}_S(z^{(e)}-\eta^{(e)} )]  +\mathbb{E}[\boldsymbol{x}^{(e)}_S(z^{(e)}-\eta^{(e)} )]  \right)  \right\}\notag                                                                                                                                                                                    \\
	 & \quad  -\{ \mathbb{E}[\boldsymbol{x}^{(e)}_S\varepsilon^{(e)}]\}^{\top} \left\{ \widehat{\tau}^{(e)} \left( \widehat{\mathbb{E}}_{m^{(e)}}[{\boldsymbol{x}}^{(e)}_S(z^{(e)} - \eta^{(e)})]  -   \mathbb{E}[\boldsymbol{x}^{(e)}_S(z^{(e)} - \eta^{(e)})] \right)\right\}\notag                                                                                                                                                                                                       \\
	 & \quad  - \{\mathbb{E}[\boldsymbol{x}^{(e)}_S\varepsilon^{(e)}]\}^{\top}  \left\{ \widehat{\tau}^{(e)}\left(   - \widehat{\mathbb{E}}_{n^{(e)}}[\boldsymbol{x}^{(e)}_S(z^{(e)}-\eta^{(e)} )]  +\mathbb{E}[\boldsymbol{x}^{(e)}_S(z^{(e)}-\eta^{(e)} )]    \right)\right\}\notag                                                                                                                                                                                                       \\
	 & \quad + (\boldsymbol{\beta} - \boldsymbol{\beta}^*)^{\top}  \boldsymbol{\Sigma}^{(e)}_{:,S}  \left\{  \widehat{\tau}^{(e)}\left(  \widehat{\mathbb{E}}_{m^{(e)}}[\boldsymbol{x}^{(e)}_S]  -\mathbb{E}[\boldsymbol{x}^{(e)}_S]\right)   \eta^{(e)}   \right\}\notag                                                                                                                                                                                                                   \\
	 & \quad + (\boldsymbol{\beta} - \boldsymbol{\beta}^*)^{\top}  \boldsymbol{\Sigma}^{(e)}_{:,S}  \left\{  \widehat{\tau}^{(e)} \left(  -\widehat{\mathbb{E}}_{n^{(e)}}[\boldsymbol{x}^{(e)}_S]  +\mathbb{E}[\boldsymbol{x}^{(e)}_S]\right)   \eta^{(e)}   \right\}\notag                                                                                                                                                                                                                 \\
	 & \quad  - \{\mathbb{E}[\boldsymbol{x}^{(e)}_S\varepsilon^{(e)}]\}^{\top}  \left\{ \widehat{\tau}^{(e)}  \left( \widehat{\mathbb{E}}_{m^{(e)}}[\boldsymbol{x}^{(e)}_S]  -\mathbb{E}[\boldsymbol{x}^{(e)}_S]  \right)  \eta^{(e)} \right\}\notag                                                                                                                                                                                                                                        \\
	 & \quad  - \{\mathbb{E}[\boldsymbol{x}^{(e)}_S\varepsilon^{(e)}]\}^{\top}  \left\{ \widehat{\tau}^{(e)}  \left(- \widehat{\mathbb{E}}_{n^{(e)}}[\boldsymbol{x}^{(e)}_S]  +\mathbb{E}[\boldsymbol{x}^{(e)}_S]  \right)  \eta^{(e)} \right\}\notag                                                                                                                                                                                                                                       \\
	 & \quad  + \frac{1}{2} \| \widehat{\mathbb{E}}_{N^{(e)}}[{\boldsymbol{x}}^{(e)}_{S^*\setminus S} {\varepsilon}^{(e)} ] \|_2^2\notag                                                                                                                                                                                                                                                                                                                                                    \\
	 & \quad + \frac{1}{2} \|     \widehat{\mathbb{E}}_{N^{(e)}}[{\boldsymbol{x}}^{(e)}_{S^*\setminus S}(z^{(e)} - \eta^{(e)})] -   \mathbb{E}[\boldsymbol{x}^{(e)}_{S^*\setminus S}(z^{(e)} - \eta^{(e)}  )]  \|_2^2\notag                                                                                                                                                                                                                                                                 \\
	 & \quad +\frac{1}{2} \|  -  \widehat{\mathbb{E}}_{n^{(e)}}[\boldsymbol{x}^{(e)}_{S^*\setminus S}(z^{(e)} - \eta^{(e)} )]  +\mathbb{E}[\boldsymbol{x}^{(e)}_{S^*\setminus S}(z^{(e)} - \eta^{(e)}  )] \|_2^2\notag                                                                                                                                                                                                                                                                      \\
	 & \quad + \frac{1}{2} \| \left\{\widehat{\mathbb{E}}_{N^{(e)}}[{\boldsymbol{x}}^{(e)}_{S^*\setminus S}]   -\mathbb{E}[\boldsymbol{x}^{(e)}_{S^*\setminus S}]\right\}\eta^{(e)}   \|_2^2\notag                                                                                                                                                                                                                                                                                          \\
	 & \quad +\frac{1}{2} \|  \left\{-  \widehat{\mathbb{E}}_{n^{(e)}}[\boldsymbol{x}^{(e)}_{S^*\setminus S}]  + \mathbb{E}[\boldsymbol{x}^{(e)}_{S^*\setminus S}]   \right\}\eta^{(e)}  \|_2^2\notag                                                                                                                                                                                                                                                                                       \\
	 & \quad  + \{\widehat{\mathbb{E}}_{N^{(e)}}[{\boldsymbol{x}}^{(e)}_{S^*\setminus S} {\varepsilon}^{(e)} ]\}^{\top} \left\{ \widehat{\tau}^{(e)}\left( \widehat{\mathbb{E}}_{m^{(e)}}[{\boldsymbol{x}}^{(e)}_{S^*\setminus S}(z^{(e)} - \eta^{(e)}  )]-   \mathbb{E}[\boldsymbol{x}^{(e)}_{S^*\setminus S}(z^{(e)} - \eta^{(e)}  )] \right) \right\}\notag                                                                                                                              \\
	 & \quad + \{\widehat{\mathbb{E}}_{N^{(e)}}[{\boldsymbol{x}}^{(e)}_{S^*\setminus S} {\varepsilon}^{(e)} ]\}^{\top} \left\{ \widehat{\tau}^{(e)}\left(-\widehat{\mathbb{E}}_{n^{(e)}}[\boldsymbol{x}^{(e)}_{S^*\setminus S}(z^{(e)} - \eta^{(e)} )] +\mathbb{E}[\boldsymbol{x}^{(e)}_{S^*\setminus S}(z^{(e)} - \eta^{(e)} ) )] \right) \right\}\notag                                                                                                                                   \\
	 & \quad + \{\widehat{\mathbb{E}}_{N^{(e)}}[{\boldsymbol{x}}^{(e)}_{S^*\setminus S} {\varepsilon}^{(e)} ]\}^{\top} \left\{ \widehat{\tau}^{(e)}\left(\widehat{\mathbb{E}}_{m^{(e)}}[\boldsymbol{x}^{(e)}_{S^*\setminus S}] -\mathbb{E}[\boldsymbol{x}^{(e)}_{S^*\setminus S}]\right) \eta^{(e)}  \right\}\notag                                                                                                                                                                         \\
	 & \quad + \{\widehat{\mathbb{E}}_{N^{(e)}}[{\boldsymbol{x}}^{(e)}_{S^*\setminus S} {\varepsilon}^{(e)} ]\}^{\top} \left\{ \widehat{\tau}^{(e)}\left(-\widehat{\mathbb{E}}_{n^{(e)}}[\boldsymbol{x}^{(e)}_{S^*\setminus S}] +\mathbb{E}[\boldsymbol{x}^{(e)}_{S^*\setminus S}]\right) \eta^{(e)}  \right\}\notag                                                                                                                                                                        \\
	 & \quad+  \left\{ \widehat{\mathbb{E}}_{N^{(e)}}[{\boldsymbol{x}}^{(e)}_{S^*\setminus S}(z^{(e)} - \eta^{(e)}) ] -   \mathbb{E}[\boldsymbol{x}^{(e)}_{S^*\setminus S}(z^{(e)} - \eta^{(e)}) ] \right\}^{\top} \left\{ -\widehat{\mathbb{E}}_{n^{(e)}}[\boldsymbol{x}^{(e)}_{S^*\setminus S}(z^{(e)} - \eta^{(e)} )] +\mathbb{E}[\boldsymbol{x}^{(e)}_{S^*\setminus S}(z^{(e)} - \eta^{(e)} )]\right\}\notag                                                                            \\
	 & \quad + \left\{\left(\widehat{\mathbb{E}}_{N^{(e)}}[{\boldsymbol{x}}^{(e)}_{S^*\setminus S}]   -\mathbb{E}[\boldsymbol{x}^{(e)}_{S^*\setminus S}]\right)\eta^{(e)}  \right\}^{\top}\left\{\left(- \widehat{\mathbb{E}}_{n^{(e)}}[\boldsymbol{x}^{(e)}_{S^*\setminus S}]   +\mathbb{E}[\boldsymbol{x}^{(e)}_{S^*\setminus S}]\right)\eta^{(e)}\right\}\notag                                                                                                                          \\
	 & \quad + \left\{ \widehat{\tau}^{(e)}\left(\widehat{\mathbb{E}}_{m^{(e)}}[{\boldsymbol{x}}^{(e)}_{S^*\setminus S}(z^{(e)} - \eta^{(e)} )] -   \mathbb{E}[\boldsymbol{x}^{(e)}_{S^*\setminus S}(z^{(e)} - \eta^{(e)}) ] \right)\right\}^{\top} \left\{ \widehat{\tau}^{(e)}\left( \widehat{\mathbb{E}}_{m^{(e)}}[\boldsymbol{x}^{(e)}_{S^*\setminus S}]   -\mathbb{E}[\boldsymbol{x}^{(e)}_{S^*\setminus S}]\right) \eta^{(e)} \right\}\notag                                          \\
	 & \quad + \left\{\widehat{\tau}^{(e)}\left(\widehat{\mathbb{E}}_{m^{(e)}}[{\boldsymbol{x}}^{(e)}_{S^*\setminus S}(z^{(e)} - \eta^{(e)} )] -   \mathbb{E}[\boldsymbol{x}^{(e)}_{S^*\setminus S}(z^{(e)} - \eta^{(e)}) ] \right)\right\}^{\top} \left\{\widehat{\tau}^{(e)} \left(- \widehat{\mathbb{E}}_{n^{(e)}}[\boldsymbol{x}^{(e)}_{S^*\setminus S}]   +\mathbb{E}[\boldsymbol{x}^{(e)}_{S^*\setminus S}]\right) \eta^{(e)} \right\}\notag                                          \\
	 & \quad + \left\{\widehat{\tau}^{(e)} \left(\widehat{\mathbb{E}}_{m^{(e)}}[{\boldsymbol{x}}^{(e)}_{S^*\setminus S}]   -\mathbb{E}[\boldsymbol{x}^{(e)}_{S^*\setminus S}]\right)\eta^{(e)}  \right\}^{\top} \left\{\widehat{\tau}^{(e)} \left(-\widehat{\mathbb{E}}_{n^{(e)}}[\boldsymbol{x}^{(e)}_{S^*\setminus S}(z^{(e)} - \eta^{(e)} )] +\mathbb{E}[\boldsymbol{x}^{(e)}_{S^*\setminus S}(z^{(e)} - \eta^{(e)} )] \right)\right\}\notag                                             \\
	 & \quad + \left\{\widehat{\tau}^{(e)} \left(-\widehat{\mathbb{E}}_{n^{(e)}}[{\boldsymbol{x}}^{(e)}_{S^*\setminus S}]   +\mathbb{E}[\boldsymbol{x}^{(e)}_{S^*\setminus S}]\right)\eta^{(e)}  \right\}^{\top} \left\{\widehat{\tau}^{(e)}\left(- \widehat{\mathbb{E}}_{n^{(e)}}[\boldsymbol{x}^{(e)}_{S^*\setminus S}(z^{(e)} - \eta^{(e)} )] +\mathbb{E}[\boldsymbol{x}^{(e)}_{S^*\setminus S}(z^{(e)} - \eta^{(e)} )] \right)\right\}\notag                                            \\
\end{align}
\begin{align}
	 & \quad  + \left\{\widehat{\mathbb{E}}_{N^{(e)}}[{\boldsymbol{x}}^{(e)}_I \varepsilon^{(e)}]\right\}^{\top}\left\{    \widehat{\mathbb{E}}_{N^{(e)}}[{\boldsymbol{x}}^{(e)}_I ({\boldsymbol{x}}^{(e)})^{\top} (\boldsymbol{\beta} - \boldsymbol{\beta}^*) ]- \boldsymbol{\Sigma}^{(e)}_{I,:}  (\boldsymbol{\beta} - \boldsymbol{\beta}^*) \right\} \notag                                                                                                            \\
	 & \quad +  \left\{ \widehat{\tau}^{(e)}\left( \widehat{\mathbb{E}}_{m^{(e)}}[{\boldsymbol{x}}^{(e)}_I(z^{(e)} - \eta^{(e)}  )] -   \mathbb{E}[\boldsymbol{x}^{(e)}_I(z^{(e)} - \eta^{(e)}  )] \right)\right\}^{\top} \left\{    \widehat{\mathbb{E}}_{N^{(e)}}[{\boldsymbol{x}}^{(e)}_I ({\boldsymbol{x}}^{(e)})^{\top}  (\boldsymbol{\beta} - \boldsymbol{\beta}^*)]- \boldsymbol{\Sigma}^{(e)}_{I,:} (\boldsymbol{\beta} - \boldsymbol{\beta}^*)    \right\}\notag \\
	 & \quad + \left\{\widehat{\tau}^{(e)}\left(- \widehat{\mathbb{E}}_{n^{(e)}}[\boldsymbol{x}^{(e)}_I(z^{(e)} - \eta^{(e)} )] +\mathbb{E}[\boldsymbol{x}^{(e)}_I(z^{(e)} - \eta^{(e)} )] \right) \right\}^{\top}\left\{    \widehat{\mathbb{E}}_{N^{(e)}}[{\boldsymbol{x}}^{(e)}_I ({\boldsymbol{x}}^{(e)})^{\top}  (\boldsymbol{\beta} - \boldsymbol{\beta}^*)]- \boldsymbol{\Sigma}^{(e)}_{I,:}(\boldsymbol{\beta} - \boldsymbol{\beta}^*)    \right\}\notag          \\
	 & \quad +\left\{\widehat{\tau}^{(e)}\left(\widehat{\mathbb{E}}_{m^{(e)}}[\boldsymbol{x}^{(e)}_I] -\mathbb{E}[\boldsymbol{x}^{(e)}_I] \right) \eta^{(e)}  \right\}^{\top} \left\{    \widehat{\mathbb{E}}_{N^{(e)}}[{\boldsymbol{x}}^{(e)}_I ({\boldsymbol{x}}^{(e)})^{\top}  (\boldsymbol{\beta} - \boldsymbol{\beta}^*)]- \boldsymbol{\Sigma}^{(e)}_{I,:}(\boldsymbol{\beta} - \boldsymbol{\beta}^*)    \right\}\notag                                              \\
	 & \quad +\left\{\widehat{\tau}^{(e)}\left(-\widehat{\mathbb{E}}_{n^{(e)}}[\boldsymbol{x}^{(e)}_I] +\mathbb{E}[\boldsymbol{x}^{(e)}_I] \right) \eta^{(e)}  \right\}^{\top} \left\{    \widehat{\mathbb{E}}_{N^{(e)}}[{\boldsymbol{x}}^{(e)}_I ({\boldsymbol{x}}^{(e)})^{\top}  (\boldsymbol{\beta} - \boldsymbol{\beta}^*)]- \boldsymbol{\Sigma}^{(e)}_{I,:}(\boldsymbol{\beta} - \boldsymbol{\beta}^*)    \right\}\notag                                             \\
	 & \le \sum_{j=1}^{12}\mathrm{T}_{1,j}^{(e)}(\boldsymbol{\beta}) + \sum_{j=1}^{15}\mathrm{T}_{2,j}^{(e)}(\boldsymbol{\beta}) +\sum_{j=1}^{5}\mathrm{T}_{3,j}^{(e)}(\boldsymbol{\beta}),
	\label{eq:Jdecomp}
\end{align}
where each summand is the absolute value of the corresponding term in the previous equality.

Next, we will derive high-probability upper bounds on $\sum_{e\in\mathcal{E}} \omega^{(e)} \mathrm{T}_{k,\ell}^{(e)}(\boldsymbol{\beta})$. Notice that the terms $\mathrm{T}_{1,1}^{(e)}(\boldsymbol{\beta}), \mathrm{T}_{1,2}^{(e)}(\boldsymbol{\beta}), \mathrm{T}_{1,3}^{(e)}(\boldsymbol{\beta})$, $\mathrm{T}_{1,4}^{(e)}(\boldsymbol{\beta}),  \mathrm{T}_{2,1}^{(e)}(\boldsymbol{\beta})$ and $\mathrm{T}_{3,1}^{(e)}(\boldsymbol{\beta})$ are similar as the terms $\mathrm{T}_k^{(e)}(\boldsymbol{\beta}),k\in[6]$ in Lemma C.6 \cite{fan2024environment}, but we will include them for completeness.

\noindent{\it Step 1.1} Upper bound on $T_{1,1}^{(e)}(\boldsymbol{\beta})$. Define the event
\begin{equation}
	C_{1, t}=\left\{\forall \boldsymbol{\beta} \in \mathbb{R}^p, \quad \sum_{e \in \mathcal{E}} \omega^{(e)} \mathrm{T}_{1,1}^{(e)}(\boldsymbol{\beta}) \leq D_{1,1} \kappa_U^{3 / 2} \sigma_x \sigma_{\varepsilon}\left\|\boldsymbol{\beta}-\boldsymbol{\beta}^*\right\|_2\left(\sqrt{\frac{t+p}{N_\omega}}+\frac{t+p}{N_*}\right)\right\}\label{eq:event_c1t}
\end{equation}
for some constant $D_{1,1}\ge 0$. We claim that $\mathbb{P}(C_{1, t}) \le 1-e^{-t}$. Without loss of generality, we assume $\boldsymbol{\beta}\ne \boldsymbol{\beta}^*$ because the inequality holds trivially when they are equal. Next observe
\begin{align}
	\sup _{\boldsymbol{\beta} \in \mathbb{R}^p, \boldsymbol{\beta} \neq \boldsymbol{\beta}^*} \frac{\sum_{e \in \mathcal{E}} \omega^{(e)} \mathrm{T}_{1,1}^{(e)}}{\left\|\boldsymbol{\beta}-\boldsymbol{\beta}^*\right\|_2} & =\sup _{\boldsymbol{\beta} \in \mathbb{R}^p, \boldsymbol{\beta} \neq \boldsymbol{\beta}^*} \frac{\left(\boldsymbol{\beta}-\boldsymbol{\beta}^*\right)^{\top}}{\left\|\boldsymbol{\beta}-\boldsymbol{\beta}^*\right\|_2} \sum_{e \in \mathcal{E}} \omega^{(e)} \boldsymbol{\Sigma}_{S \cup S^*, S}^{(e)}\left\{\widehat{\mathbb{E}}_{N^{(e)}}\left[\boldsymbol{x}_S^{(e)} \varepsilon^{(e)}\right]-\mathbb{E}\left[\boldsymbol{x}^{(e)} \varepsilon^{(e)}\right]\right\} \notag \\
	                                                                                                                                                                                                                        & \leq \sup _{|S| \leq p}\left\|\sum_{e \in \mathcal{E}} \omega^{(e)} \boldsymbol{\Sigma}_{S \cup S^*, S}^{(e)}\left\{\widehat{\mathbb{E}}_{N^{(e)}}\left[\boldsymbol{x}_S^{(e)} {\varepsilon}^{(e)}\right]-\mathbb{E}\left[\boldsymbol{x}^{(e)} {\varepsilon}^{(e)}\right]\right\}\right\|_2\notag                                                                                                                                                                              \\
	                                                                                                                                                                                                                        & \equiv \sup _{|S| \leq p}\left\|\boldsymbol{\xi}_S\right\|_2
	\label{eq:decom_for_t11}
\end{align}

For any $S \subset[p]$, let $\boldsymbol{v}_1^{(S)}, \ldots, \boldsymbol{v}_{N_S}^{(S)}$ be an 1/4-covering of $\mathcal{B}\left(S \cup S^*\right)$, that is, for any $\boldsymbol{v} \in \mathcal{B}\left(S \cup S^*\right)$, there exists some $\pi(\boldsymbol{v}) \in\left[N_S\right]$ such that $\left\|\boldsymbol{v}-\boldsymbol{v}_{\pi(v)}^{(S)}\right\|_2 \leq 1 / 4$. It follows from standard empirical process result that $N_S \leq 9^{\left|S \cup S^*\right|}$, then $N=\sum_{|S| \leq p} N_S\le 9^{s^*}(9 e)^p$.

At the same time, denote $\boldsymbol{\xi}=\sum_{e \in \mathcal{\mathcal { E }}} \omega^{(e)} \boldsymbol{\Sigma}_{S \cup S^*, S}^{(e)}\left\{\widehat{\mathbb{E}}_{N^{(e)}}\left[\boldsymbol{x}_S^{(e)} {\varepsilon}^{(e)}\right]-\mathbb{E}\left[\boldsymbol{x}^{(e)} {\varepsilon}^{(e)}\right]\right\}$. For any $|S| \in[p]$, it follows from the variational representation of the $\ell_2$ norm that
$$
	\|\boldsymbol{\xi}\|_2=\sup _{\boldsymbol{v} \in \mathcal{B}\left(S \cup S^*\right)} \boldsymbol{v}^{\top} \boldsymbol{\xi}=\sup _{k \in\left[N_S\right]}\left(\boldsymbol{v}_k^{(S)}\right)^{\top} \boldsymbol{\xi}+\sup _{\boldsymbol{v} \in \mathcal{B}\left(S \cup S^*\right)}\left(\boldsymbol{v}-\boldsymbol{v}_{\pi(v)}^{(S)}\right)^{\top} \boldsymbol{\xi} \leq \sup _{k \in\left[N_S\right]}\left(\boldsymbol{v}_k^{(S)}\right)^{\top} \boldsymbol{\xi}+\frac{1}{4}\|\boldsymbol{\xi}\|_2,
$$
where the last inequality follows from the Cauchy-Schwarz inequality and our construction of covering above. This implies $\|\boldsymbol{\xi}\|_2 \leq 2 \sup _{k \in\left[N_S\right]}\left(\boldsymbol{v}_k^{(S)}\right)^{\top} \boldsymbol{\xi}$, thus
\begin{align}
	\sup _{|S| \leq s} & \left\|\sum_{e \in \mathcal{E}} \omega^{(e)} \boldsymbol{\Sigma}_{S \cup S^*, S}^{(e)}\left\{\widehat{\mathbb{E}}_{N^{(e)}}\left[\boldsymbol{x}_S^{(e)} \varepsilon^{(e)}\right]-\mathbb{E}\left[\boldsymbol{x}^{(e)} \varepsilon^{(e)}\right]\right\}\right\|_2 \notag                                                                             \\
	                   & \leq 2 \sup _{|S| \leq s, k \in\left[N_S\right]}\left(\boldsymbol{v}_k^{(S)}\right)^{\top} \sum_{e \in \mathcal{E}} \omega^{(e)} \boldsymbol{\Sigma}_{S \cup S^*, S}^{(e)}\left\{\widehat{\mathbb{E}}_{N^{(e)}}\left[\boldsymbol{x}_S^{(e)} \varepsilon^{(e)}\right]-\mathbb{E}\left[\boldsymbol{x}^{(e)} {\varepsilon}^{(e)}\right]\right\} \notag \\
	                   & \equiv \sup _{|S| \leq p, k \in\left[N_S\right]} 2 Z(S, k)
	\label{eq:norm_t11}
\end{align}
For given fixed $\boldsymbol{v}_k^{(S)} \in \mathcal{B}\left(S \cup S^*\right), Z(S, k)$ can be written as the sum of independent zero-mean random variables as
\begin{align*}
	Z(S, k) & = \left(\boldsymbol{v}_k^{(S)}\right)^{\top} \sum_{e \in \mathcal{E}} \omega^{(e)} \boldsymbol{\Sigma}_{S \cup S^*, S}^{(e)}\left\{\widehat{\mathbb{E}}_{N^{(e)}}\left[\boldsymbol{x}_S^{(e)} \varepsilon^{(e)}\right]-\mathbb{E}\left[\boldsymbol{x}^{(e)} {\varepsilon}^{(e)}\right]\right\}                                                                                                                                                   \\\
	        & = \sum_{e \in \mathcal{E}} \sum_{\ell=1}^{N^{(e)}} \frac{\omega^{(e)} }{N^{(e)}}\left\{\mathbb{E}\left[\left(\boldsymbol{v}_k^{(S)}\right)^{\top}\boldsymbol{\Sigma}_{S \cup S^*, S}^{(e)}\left[\boldsymbol{x}_{\ell}^{(e)}\right]_{S}\varepsilon^{(e)}_{\ell}\right]-\left(\boldsymbol{v}_k^{(S)}\right)^{\top}\boldsymbol{\Sigma}_{S \cup S^*, S}^{(e)}\left[\boldsymbol{x}_{\ell}^{(e)}\right]_{S \cup S^*}\varepsilon_{\ell}^{(e)}\right\} .
\end{align*}
Observe that $\varepsilon^{(e)}$ is a zero-mean sub-Gaussian random variable with parameter $\sigma_{\varepsilon}$ by Condition \ref{cond:subg_e}, and $\left(\boldsymbol{v}_k^{(S)}\right)^{\top} \boldsymbol{\Sigma}_{S \cup S^*, S}^{(e)} \boldsymbol{x}_S$ is a zero-mean sub-Gaussian random variable with parameter $\kappa_U^{3 / 2} \sigma_x$ by Condition \ref{cond:subg_x}. Then we have
$$
	\mathbb{P}\left[|Z(S, k)| \geq d_{1,1} \kappa_U^{3 / 2} \sigma_x \sigma_{\varepsilon}\left(\sqrt{\frac{u}{N_\omega}}+\frac{u}{N_*}\right)\right] \leq 2 e^{-u}, \quad \forall u>0.
$$
Apply union bound over all $k\in [N_S]$ and $S\subseteq [p]$, we have
\begin{align*}
	 & \mathbb{P}\left[\sup _{|S| \leq p, k \in\left[N_S\right]} 2 Z(S, k) \geq 2d_{1,1} \kappa_U^{3 / 2} \sigma_x \sigma_{\varepsilon}\left(\sqrt{\frac{u}{N_\omega}}+\frac{u}{N_*}\right)\right]                          \\
	 & \quad \leq \sum_{|S| \leq p, k \in\left[N_S\right]} \mathbb{P}\left[|Z(S, k) |\geq d_{1,1} \kappa_U^{3 / 2} \sigma_x \sigma_{\varepsilon}\left(\sqrt{\frac{u}{N_\omega}}+\frac{u}{N_*}\right)\right] \leq 2 N e^{-u}
\end{align*}
Letting $u=t+\log (10 N)<C'(t+p)$, and together with (\ref{eq:decom_for_t11}) and (\ref{eq:norm_t11}), we are done.

\noindent{\it Step 1.2}. UPPER BOUND ON $T_{1,2}^{(e)}(\boldsymbol{\beta})$.  We claim that $\mathbb{P}\left(C_{2, t}\right) \geq 1-e^{-t}$ for any $t>0$, where
\begin{align}
	C_{2, t}=\left\{\forall \boldsymbol{\beta}\in \mathbb{R}^p, \quad \sum_{e \in \mathcal{E}} \omega^{(e)} \mathrm{T}_{1,2}^{(e)}(\boldsymbol{\beta}) \leq D_{1,2} \kappa_U^{1 / 2} \sigma_x \sigma_{\varepsilon} \sqrt{\frac{t+p}{N_*}} \times \sqrt{\sum_{e \in \mathcal{E}} \omega^{(e)}\left\|\mathbb{E}\left[x_S^{(e)} \varepsilon^{(e)}\right]\right\|_2^2}\right.\notag \\
	\left.+D_{1,2}  \kappa_U \sigma_x \sigma_{\varepsilon}^2 \frac{t+p}{N_*}\right\}.
	\label{eq:event_c2t}
\end{align}
for some universal constant $D_{1,2} $ to be determined. For a fixed $S$, we can write down $\sum_{e \in \mathcal{E}} \omega^{(e)} \mathrm{T}_{1,2}^{(e)}(\boldsymbol{\beta})$ as sum of independent random variables as
\begin{align*}
	Z(S) \equiv \sum_{e \in \mathcal{E}} \omega^{(e)} \mathrm{T}_{1,2}^{(e)}(\boldsymbol{\beta}) & = \sum_{e \in \mathcal{E}} \sum_{\ell=1}^{N^{(e)}} \frac{\omega^{(e)}}{N^{(e)}}\left(\mathbb{E}[\boldsymbol{x}_S^{(e)} \varepsilon^{(e)}]^{\top}[\boldsymbol{x}_{\ell}^{(e)}]_S\varepsilon_{\ell}^{(e)}-\mathbb{E}[\mathbb{E}[\boldsymbol{x}_S^{(e)} \varepsilon^{(e)}]^{\top}[\boldsymbol{x}_{\ell}^{(e)}]_S\varepsilon_{\ell}^{(e)}]\right).
\end{align*}
By Condition \ref{cond:subg_x} and \ref{cond:subg_e}, it is a sum of sub-Exponential random variables with parameters
$$\left(\frac{\omega^{(e)}}{N^{(e)}} \kappa_U^{1/2}\sigma_x\sigma_{\varepsilon}\left\|\mathbb{E}[\boldsymbol{x}_S^{(e)}\varepsilon^{(e)}]\right\|_2, \frac{\omega^{(e)}}{N^{(e)}} \kappa_U^{1/2}\sigma_x\sigma_{\varepsilon}\left\|\mathbb{E}[\boldsymbol{x}_S^{(e)}\varepsilon^{(e)}]\right\|_2\right).$$
Then it follows that
\begin{align*}
	\mathbb{P}\left[      |Z(S)|   \ge  d_{1,2}  k_U^{1/2} \sigma_x\sigma_{\varepsilon}  \left\{\sqrt{\sum_{e \in \mathcal{E}}\left(\omega^{(e)}\right)^2 \frac{1}{N^{(e)}}\left\|\mathbb{E}\left[x_S^{(e)} \varepsilon^{(e)}\right]\right\|_2^2} \times \sqrt{x}+\max _{e \in \mathcal{E}} \frac{\omega^{(e)}}{N^{(e)}}\left\|\mathbb{E}\left[x_S^{(e)} \varepsilon^{(e)}\right]\right\|_2 \times x\right\} \right] \le 2e^{-x},
\end{align*}
for any $x>0$. Next observe:
\begin{align*}
	 & d_{1,2}  k_U^{1/2} \sigma_x\sigma_{\varepsilon}  \left\{\sqrt{\sum_{e \in \mathcal{E}}\left(\omega^{(e)}\right)^2 \frac{1}{N^{(e)}}\left\|\mathbb{E}\left[x_S^{(e)} \varepsilon^{(e)}\right]\right\|_2^2} \times \sqrt{x}+\max _{e \in \mathcal{E}} \frac{\omega^{(e)}}{N^{(e)}}\left\|\mathbb{E}\left[x_S^{(e)} \varepsilon^{(e)}\right]\right\|_2 \times x\right\} \\
	 & \le d_{1,2}  k_U^{1/2} \sigma_x\sigma_{\varepsilon} \left\{\sqrt{\frac{x}{N_*}} \sqrt{\sum_{e \in \mathcal{E}} \omega^{(e)}\left\|\mathbb{E}\left[\boldsymbol{x}_S^{(e)} \varepsilon^{(e)}\right]\right\|_2^2}+\frac{x}{N_*} \max _{e \in \mathcal{E}}\left\|\mathbb{E}\left[\boldsymbol{x}^{(e)} \varepsilon^{(e)}\right]\right\|_2\right\}                         \\
	 & \le   d_{1,2} \kappa_U^{1 / 2} \sigma_x \sigma_{\varepsilon}\left\{\sqrt{\frac{x}{N_*}} \sqrt{\overline{\mathrm{b}}_S}+\kappa_U^{1 / 2} \sigma_{\varepsilon} \frac{x}{N_*}\right\},
\end{align*}
where the last line is by the notation $\overline{\mathrm{b}}_S$ and the fact that $\left\|\mathbb{E}\left[\varepsilon^{(e)} \boldsymbol{x}^{(e)}\right]\right\|_2 \leq \sigma_{\varepsilon} \kappa_U^{1 / 2}$, which can be found in Lemma C.11. of \cite{fan2024environment}. Therefore, we have
$$
	\mathbb{P}\left[      |Z(S)|   \ge  d_{1,2} \kappa_U^{1 / 2} \sigma_x \sigma_{\varepsilon}\left\{\sqrt{\frac{x}{N_*}} \sqrt{\overline{\mathrm{b}}_S}+\kappa_U^{1 / 2} \sigma_{\varepsilon} \frac{x}{N_*}\right\} \right] \le 2e^{-x},\quad\forall x>0.
$$

Notice the above high probability bound is true for $\boldsymbol{\beta}\in\mathbb{R}^p$ with fixed support set $S$. The total number of $S$ satisfying $\left|S \backslash S^*\right|=r$ can be upper bounded by
$$
	N_r=2^{s^*} \times\left(\begin{array}{c}
			p-s^* \\
			r
		\end{array}\right) \leq 2^{s^*}(e p / r)^r.
$$

Let $x = u + \log(2N_r)$, we have
\begin{equation}
	\mathbb{P}\left[      |Z(S)|   \ge  d_{1,2} \kappa_U^{1 / 2} \sigma_x \sigma_{\varepsilon}\left\{\sqrt{\frac{u + \log(2N_r)}{N_*}} \sqrt{\overline{\mathrm{b}}_S}+\kappa_U^{1 / 2} \sigma_{\varepsilon} \frac{u + \log(2N_r)}{N_*}\right\} \right] \le \frac{e^{-u}}{N_r},\quad\forall x>0.
	\label{eq:ulog2N_r}
\end{equation}

Now we are ready to define a new event that holds true simultaneously for a group of $S$ that satisfies $\left|S \backslash S^*\right|=r\in\{1,...,p\}$:
$$
	\mathcal{K}_u(r) = \left\{\forall S,\left|S \backslash S^*\right|=r, \quad|Z(S)| \leq  d_{1,2} \kappa_U^{1 / 2} \sigma_x \sigma_{\varepsilon}\left\{\sqrt{\frac{u + \log(2N_r)}{N_*}} \sqrt{\overline{\mathrm{b}}_S}+\kappa_U^{1 / 2} \sigma_{\varepsilon} \frac{u + \log(2N_r)}{N_*}\right\}\right\},
$$
where $u>0$. In order to study the probability of this event, it is more intuitive to work with its complement:
$$
	\mathcal{K}^c_u(r) = \left\{\exists S,\left|S \backslash S^*\right|=r\quad s.t.\quad|Z(S)| \geq  d_{1,2} \kappa_U^{1 / 2} \sigma_x \sigma_{\varepsilon}\left\{\sqrt{\frac{u + \log(2N_r)}{N_*}} \sqrt{\overline{\mathrm{b}}_S}+\kappa_U^{1 / 2} \sigma_{\varepsilon} \frac{u + \log(2N_r)}{N_*}\right\}\right\}.
$$

Using equation \ref{eq:ulog2N_r} and applying a union bound over $S$ that satisfies $\left|S \backslash S^*\right|=r$, we have
\begin{align*}
	\mathbb{P}[\mathcal{K}^c_u(r) ] & = \sum_{\left|S \backslash S^*\right|=r} \mathbb{P}\left[      |Z(S)|   \ge  d_{1,2} \kappa_U^{1 / 2} \sigma_x \sigma_{\varepsilon}\left\{\sqrt{\frac{u + \log(2N_r)}{N_*}} \sqrt{\overline{\mathrm{b}}_S}+\kappa_U^{1 / 2} \sigma_{\varepsilon} \frac{u + \log(2N_r)}{N_*}\right\} \right] \\
	                                & \le \sum_{\left|S \backslash S^*\right|=r} \frac{e^{-u}}{N_r}                                                                                                                                                                                                                               \\
	                                & = e^{-u},
\end{align*}
and hence $\mathbb{P}[\mathcal{K}_u(r) ]\ge 1-e^{-u}$. However, so far the high probability bound only concerns a specific $r$ rather than all $r\in\{1,...,p\}$. Now consider the event $\bigcap_{r=1}^p \mathcal{K}_{t+\log p}(r)$, and notice
\begin{align*}
	\mathbb{P}\left[\bigcap_{r=1}^p \mathcal{K}_{t+\log p}(r)\right] & = 1- \mathbb{P}\left[\bigcup_{r=1}^p \mathcal{K}^c_{t+\log p}(r)\right] \\
	                                                                 & \ge 1-\sum_{r=1}^p\mathbb{P}\left[ \mathcal{K}^c_{t+\log p}(r)\right]   \\
	                                                                 & \ge 1-\sum_{r=1}^p e^{-t-\log p}                                        \\
	                                                                 & = 1-e^{-t}.
\end{align*}
Along with the following fact that $\forall r\ge 1$,
\begin{align*}
	(t + \log p)+ \log(2N_r) & \le 2(t +\log p + r\log(ep/r)+s^*) \\
	                         & \le 4r\log(ep/r)+2s^*+2t           \\
	                         & \le 4(r\log(ep/r)+s^*+t)           \\
	                         & \le 4(p+s^*+t)                     \\
	                         & \le 8(p+t),
\end{align*}
we have the event
\begin{align*}
	C_{2, t}=\left\{\forall \boldsymbol{\beta}\in \mathbb{R}^p, \quad \sum_{e \in \mathcal{E}} \omega^{(e)} \mathrm{T}_{1,2}^{(e)}(\boldsymbol{\beta}) \leq D_{1,2} \kappa_U^{1 / 2} \sigma_x \sigma_{\varepsilon} \sqrt{\frac{p+t}{N_*}} \times \sqrt{\sum_{e \in \mathcal{E}} \omega^{(e)}\left\|\mathbb{E}\left[x_S^{(e)} \varepsilon^{(e)}\right]\right\|_2^2}\right. \\
	\left.+D_{1,2}  \kappa_U \sigma_x \sigma_{\varepsilon}^2 \frac{p+t}{N_*}\right\}
\end{align*}
takes place with probability at least $1-e^{-t}$ for arbitrary $t>0$.

\noindent{\it Step 1.3} UPPER BOUND ON $T_{1,3}^{(e)}(\boldsymbol{\beta})$. We argue in this step that the event
\begin{equation}
	C_{3, t}=\left\{\forall \boldsymbol{\beta} \in \mathbb{R}^p, \quad \sum_{e \in \mathcal{E}} \omega^{(e)} \mathrm{T}_3^{(e)}(\boldsymbol{\beta}) \leq D_{1,3} \kappa_U^2 \sigma_x^2\left\|\boldsymbol{\beta}-\boldsymbol{\beta}^*\right\|_2^2\left(\sqrt{\frac{t+p}{N_\omega}}+\frac{t+p}{N_*}\right)\right\}
	\label{eq:event_c3t}
\end{equation}
occurs with probability at least $1-e^{-t}$ for any $t>0$, where $ D_{1,3}$ is some universal constatn to be determined. Without loss of generality, let $\boldsymbol{\beta} \neq \boldsymbol{\beta}^*$, then it suffices to establish an upper bound for
\begin{align*}
	 & \sup _{\boldsymbol{\beta} \neq \boldsymbol{\beta}^*} \frac{\left(\boldsymbol{\beta}-\boldsymbol{\beta}^*\right)^{\top}}{\left\|\boldsymbol{\beta}-\boldsymbol{\beta}^*\right\|_2} \sum_{e \in \mathcal{E}} \omega^{(e)} \boldsymbol{\Sigma}_{:, S}^{(e)}\left(\widehat{\mathbb{E}}_{N^{(e)}}\left[\boldsymbol{x}_S^{(e)}\left(\boldsymbol{x}^{(e)}\right)^{\top}\right]-\mathbb{E}\left[\boldsymbol{x}_S^{(e)}\left(\boldsymbol{x}^{(e)}\right)^{\top}\right]\right) \frac{\left(\boldsymbol{\beta}-\boldsymbol{\beta}^*\right)}{\left\|\boldsymbol{\beta}-\boldsymbol{\beta}^*\right\|_2^2} \\
	 & \leq \sup _{|S| \leq p}\left\|\sum_{e \in \mathcal{E}} \omega^{(e)} \boldsymbol{\Sigma}_{S \cup S^*, S}^{(e)}\left(\widehat{\mathbb{E}}_{N^{(e)}}\left[\boldsymbol{x}_S^{(e)}\left(\boldsymbol{x}_{S \cup S^*}^{(e)}\right)^{\top}\right]-\mathbb{E}\left[\boldsymbol{x}_S^{(e)}\left(\boldsymbol{x}_{S \cup S^*}^{(e)}\right)^{\top}\right]\right)\right\|_2                                                                                                                                                                                                                                \\
	 & \equiv \sup _{|S| \leq p}\left\|\boldsymbol{A}_S\right\|_2 .
\end{align*}
Following a similar strategy as in Step 1 of the proof of Lemma A.1, for any $S\subset [p]$, let $\left\{\left(\boldsymbol{v}_k^{(e)}, \boldsymbol{u}_k^{(e)}\right)\right\}_{k=1}^{N_S} \in \mathcal{B}^2\left(S \cup S^*\right)$ be a 1/4-covering of $\mathcal{B}^2\left(S \cup S^*\right)$. It follows from standard empirical process theory that $N_S \leq 9^{2\left|S \cup S^*\right|}$, then $N=\sum_{S \subset[p]} N_S \leq 81^{s^*}\left(\frac{81 e p}{s}\right)^s$. Moreover, by variational representation of the matrix $\ell_2$ norm, $\left\|\boldsymbol{A}_S\right\|_2 \leq 4 \sup _{k \in\left[N_S\right]}\left(\boldsymbol{u}_k^{(S)}\right)^{\top} \boldsymbol{A}_S \boldsymbol{v}_k^{(S)}$, thus
\begin{equation}
	\sup _{|S| \leq p}\left\|\boldsymbol{A}_S\right\|_2 \leq \sup _{|S| \leq p, k \in\left[N_S\right]}4\left(\boldsymbol{u}_k^{(e)}\right)^{\top} \boldsymbol{A}_S\left(\boldsymbol{v}_k^{(e)}\right)\equiv\sup _{|S| \leq p, k \in\left[N_S\right]} 4 Z(S, k).
	\label{eq:step3T}
\end{equation}
For fixed $k$ and $S, Z(S, k)$ can be written as the sum of independent zero-mean random variables as
\begin{align*}
	Z(S, k) & =\left(\boldsymbol{u}_k^{(S)}\right)^{\top}\left\{\sum_{e \in \mathcal{E}} \omega^{(e)} \boldsymbol{\Sigma}_{S \cup S^*, S}^{(e)}\left(\widehat{\mathbb{E}}_{N^{(e)}}\left[\boldsymbol{x}_S^{(e)}\left(\boldsymbol{x}_{S \cup S^*}^{(e)}\right)^{\top}\right]-\mathbb{E}\left[\boldsymbol{x}_S^{(e)}\left(\boldsymbol{x}_{S \cup S^*}^{(e)}\right)^{\top}\right]\right)\right\}\left(\boldsymbol{v}_k^{(S)}\right)                             \\
	        & =\left(\boldsymbol{u}_k^{(S)}\right)^{\top}\left\{\sum_{e \in \mathcal{E}} \omega^{(e)}\boldsymbol{\Sigma}_{S \cup S^*, S}^{(e)}\left(\frac{1}{N^{(e)}} \sum_{\ell=1}^{N^{(e)}}\left[\boldsymbol{x}_{\ell}^{(e)}\right]_{S}\left[\boldsymbol{x}_{\ell}^{(e)}\right]_{S \cup S^*}^{\top}-\mathbb{E}\left[\boldsymbol{x}_{S}^{(e)}\left(\boldsymbol{x}_{S \cup S^*}^{(e)}\right)^{\top}\right]\right)\right\}\left(\boldsymbol{v}_k^{(S)}\right) \\
	        & =\sum_{e \in \mathcal{E}} \sum_{\ell=1}^{N^{(e)}} \frac{\omega^{(e)}}{N^{(e)}}\left\{\left(\boldsymbol{u}_k^{(S)}\right)^{\top}\boldsymbol{\Sigma}_{S \cup S^*, S}^{(e)}\left[\boldsymbol{x}_{\ell}^{(e)}\right]_{S }\left[\boldsymbol{x}_{\ell}^{(e)}\right]_{S \cup S^*}^{\top}\left(\boldsymbol{v}_k^{(S)}\right)\right\}                                                                                                                   \\
	        & \quad \quad\quad\quad\quad\quad\quad\quad-\mathbb{E}\left[\left(\boldsymbol{u}_k^{(S)}\right)^{\top}\boldsymbol{\Sigma}_{S \cup S^*, S}^{(e)}\left[\boldsymbol{x}_{\ell}^{(e)}\right]_{S }\left[\boldsymbol{x}_{\ell}^{(e)}\right]_{S \cup S^*}^{\top}\left(\boldsymbol{v}_k^{(S)}\right)\right] .
\end{align*}
Observe that $\left(\boldsymbol{u}_k^{(e)}\right)^{\top} \Sigma_{S \cup S^*, S}^{(e)} \boldsymbol{x}_S^{(e)}$ and $\left(\boldsymbol{x}_{S \cup S^*}^{(e)}\right)^{\top} \boldsymbol{v}_k^{(e)}$ are sub-Gaussian random variables with parameter $\kappa_U^{3 / 2} \sigma_x$ and $\kappa_U^{1 / 2} \sigma_x$, respectively. Therefore, for any $u>0$,
$$
	\mathbb{P}\left[|Z(S, k)| \leq d_{1,3} \kappa_U^2 \sigma_x^2\left(\sqrt{\frac{u}{N_\omega}}+\frac{u}{N_*}\right)\right] \geq 1-2 e^{-u}.
$$
Next apply union bound over $k\in[N_S], S\subset [p]$, we have
\begin{align*}
	\mathbb{P} & {\left[\sup _{|S| \leq p, k \in\left[N_S\right]} 4 Z(S, k) \geq 4 d_{1,3} \kappa_U^{2} \sigma_x^2 \left(\sqrt{\frac{u}{N_\omega}}+\frac{u}{N_*}\right)\right] }                        \\
	           & \leq \sum_{|S| \leq p, k \in\left[N_S\right]} \mathbb{P}\left[|Z(S, k)| \geq d_{1,1} \kappa_U^2 \sigma_x^2 \left(\sqrt{\frac{u}{N_\omega}}+\frac{u}{N_*}\right)\right] \leq 2 N e^{-u}
\end{align*}
Now let $u=t+\log(2N)\le C^{\prime} (t+p)$, we have
$$
	\mathbb{P}  {\left[\sup _{|S| \leq p, k \in\left[N_S\right]} 4 Z(S, k) \leq 4 C^{\prime}d_{1,3} \kappa_U^{2} \sigma_x^2 \left(\sqrt{\frac{t+p}{N_\omega}}+\frac{t+p}{N_*}\right)\right] } \geq 1-e^{-t}
$$
Together with (\ref{eq:step3T}), we have shown the claim.

\bigskip
In some of the following Steps 1.4-1.12, we follow similar procedures as previous steps, and therefore those proofs are omitted.

\noindent{\it Step 1.4} UPPER BOUND ON $T_{1,4}^{(e)}(\boldsymbol{\beta})$. Follow similar procedure as step 1.1, for any $t>0$,
\begin{equation}
	\mathbb{P}\left(C_{4, t}\right)=\mathbb{P}\left[\forall \boldsymbol{\beta} \in \mathbb{R}^p, \quad \sum_{e \in \mathcal{E}} \omega^{(e)} \mathrm{T}_4^{(e)}(\boldsymbol{\beta}) \leq D_{1,4} \kappa_U^{3 / 2} \sigma_x^2 \sigma_{\varepsilon}\left\|\boldsymbol{\beta}-\boldsymbol{\beta}^*\right\|_2\left(\sqrt{\frac{t+p}{N_\omega}}+\frac{t+p}{N_*}\right)\right] \geq 1-e^{-t}.
	\label{eq:event_c4t}
\end{equation}

\noindent{\it Step 1.5} UPPER BOUND ON $T_{1,5}^{(e)}(\boldsymbol{\beta})$. In this step, the event
\begin{equation}
	C_{5, t}=\left\{\forall \boldsymbol{\beta} \in \mathbb{R}^p, \quad \sum_{e \in \mathcal{E}} \omega^{(e)} \mathrm{T}_{1,5}^{(e)}(\boldsymbol{\beta}) \leq D_{1,5} \kappa_U^{3 / 2} \sigma_x \sigma_z\left\|\boldsymbol{\beta}-\boldsymbol{\beta}^*\right\|_2\left(\sqrt{\frac{t+p}{m_\omega^{\widehat{\tau}^2}}}+\frac{t+p}{m_*^{\widehat{\tau}}}\right)\right\}
	\label{eq:event_c5t}
\end{equation}
occurs with probability at least $1-e^{-t},\forall t>0$.

\noindent{\it Step 1.6}. UPPER BOUND ON $T_{1,6}^{(e)}(\boldsymbol{\beta})$. In this step, the event
\begin{equation}
	C_{6, t}=\left\{\forall \boldsymbol{\beta} \in \mathbb{R}^p, \quad \sum_{e \in \mathcal{E}} \omega^{(e)} \mathrm{T}_{1,6}^{(e)}(\boldsymbol{\beta}) \leq D_{1,6} \kappa_U^{3 / 2} \sigma_x \sigma_z\left\|\boldsymbol{\beta}-\boldsymbol{\beta}^*\right\|_2\left(\sqrt{\frac{t+p}{n_\omega^{\widehat{\tau}^2}}}+\frac{t+p}{n_*^{\widehat{\tau}}}\right)\right\}
	\label{eq:event_c6t}
\end{equation}
occurs with probability at least $1-e^{-t}, \forall t>0$.

\noindent{\it Step 1.7}. UPPER BOUND ON $T_{1,7}^{(e)}(\boldsymbol{\beta})$. In this step we claim that $\mathbb{P}\left(C_{7, t}\right) \geq 1-e^{-t}$ for any $t>0$, where

\begin{align}
	C_{7, t}=\left\{\forall \boldsymbol{\beta} \in \mathbb{R}^p, \quad \sum_{e \in \mathcal{E}} \omega^{(e)} \mathrm{T}_{1,7}^{(e)}(\boldsymbol{\beta}) \leq D_{1,7} \kappa_U^{1 / 2} \sigma_x \sigma_{z} \sqrt{\frac{t+p}{m^{\widehat{\tau}^2}_*}} \right.\notag \\
	\left. \times \sqrt{\sum_{e \in \mathcal{E}} \omega^{(e)}\left\|\mathbb{E}\left[x_S^{(e)} \varepsilon^{(e)}\right]\right\|_2^2}+D_{1,7} \kappa_U \sigma_x \sigma_{z}\sigma_{\varepsilon}\frac{t+p}{m^{\widehat{\tau}}_*}\right\}
	\label{eq:event_c7t}
\end{align}
for some universal constant $D_{1,7} $ to be determined. For a fixed $S$, we can write down $\sum_{e \in \mathcal{E}} \omega^{(e)} \mathrm{T}_{1,7}^{(e)}(\boldsymbol{\beta})$ as sum of independent random variables as
\begin{align*}
	Z(S) \equiv \sum_{e \in \mathcal{E}} \omega^{(e)} \mathrm{T}_{1,7}^{(e)}(\boldsymbol{\beta}) & = \sum_{e \in \mathcal{E}} \sum_{\ell=1}^{m^{(e)}} \frac{\omega^{(e)}\widehat{\tau}^{(e)}}{m^{(e)}}\left(\mathbb{E}[\boldsymbol{x}_S^{(e)}\varepsilon^{(e)}]^{\top}[\boldsymbol{x}_{\ell}^{(e)}]_S (z_{\ell}^{(e)}-\eta^{(e)})-\mathbb{E}\left[\mathbb{E}[\boldsymbol{x}_S^{(e)} \varepsilon^{(e)}]^{\top}[\boldsymbol{x}_{\ell}^{(e)}](z_{\ell}^{(e)}-\eta^{(e)})\right]\right).
\end{align*}
By Conditions \ref{cond:subg_x} and \ref{cond:subg_e}, it is a sum of sub-Exponential random variables with parameters
$$\left(\frac{\omega^{(e)}\widehat{\tau}^{(e)}}{m^{(e)}} \kappa_U^{1/2}\sigma_x\sigma_{z}\left\|\mathbb{E}[\boldsymbol{x}_S^{(e)}\varepsilon^{(e)}]\right\|_2, \frac{\omega^{(e)}\widehat{\tau}^{(e)}}{m^{(e)}} \kappa_U^{1/2}\sigma_x\sigma_{z}\left\|\mathbb{E}[\boldsymbol{x}_S^{(e)}\varepsilon^{(e)}]\right\|_2\right).$$
Then it follows that
\begin{align*}
	\mathbb{P}\left[      |Z(S)|   \ge  d_{1,7}  k_U^{1/2} \sigma_x\sigma_{z}  \left\{\sqrt{\sum_{e \in \mathcal{E}}\left(\omega^{(e)}\widehat{\tau}^{(e)}\right)^2 \frac{1}{m^{(e)}}\left\|\mathbb{E}\left[x_S^{(e)} \varepsilon^{(e)}\right]\right\|_2^2} \times \sqrt{x}+\max _{e \in \mathcal{E}} \frac{\omega^{(e)}\widehat{\tau}^{(e)}}{m^{(e)}}\left\|\mathbb{E}\left[x_S^{(e)} \varepsilon^{(e)}\right]\right\|_2 \times x\right\} \right] \le 2e^{-x},
\end{align*}
for any $x>0$. Next observe:
\begin{align*}
	 & d_{1,7}  k_U^{1/2} \sigma_x\sigma_{z}  \left\{\sqrt{\sum_{e \in \mathcal{E}}\left(\omega^{(e)}\widehat{\tau}^{(e)}\right)^2 \frac{1}{m^{(e)}}\left\|\mathbb{E}\left[x_S^{(e)} \varepsilon^{(e)}\right]\right\|_2^2} \times \sqrt{x}+\max _{e \in \mathcal{E}} \frac{\omega^{(e)}\widehat{\tau}^{(e)}}{m^{(e)}}\left\|\mathbb{E}\left[x_S^{(e)} \varepsilon^{(e)}\right]\right\|_2 \times x\right\} \\
	 & \le d_{1,7}  k_U^{1/2} \sigma_x\sigma_{z} \left\{\sqrt{\frac{x}{m_*^{\widehat{\tau}^2}}} \sqrt{\sum_{e \in \mathcal{E}} \omega^{(e)}\left\|\mathbb{E}\left[\boldsymbol{x}_S^{(e)} \varepsilon^{(e)}\right]\right\|_2^2}+\frac{x}{m_*^{\widehat{\tau}}} \max _{e \in \mathcal{E}}\left\|\mathbb{E}\left[\boldsymbol{x}^{(e)} \varepsilon^{(e)}\right]\right\|_2\right\}                             \\
	 & \le   d_{1,7} \kappa_U^{1 / 2} \sigma_x \sigma_{z}\left\{\sqrt{\frac{x}{m_*^{\widehat{\tau}^2}}} \sqrt{\overline{\mathrm{b}}_S}+\kappa_U^{1 / 2} \sigma_{\varepsilon} \frac{x}{m_*^{\widehat{\tau}}}\right\},
\end{align*}
where the last line is by the notation $\overline{\mathrm{b}}_S$ and the fact that $\left\|\mathbb{E}\left[\varepsilon^{(e)} \boldsymbol{x}^{(e)}\right]\right\|_2 \leq \sigma_{\varepsilon} \kappa_U^{1 / 2}$, which is proved in Lemma C.11 in \cite{fan2024environment}. Therefore, we have
$$
	\mathbb{P}\left[      |Z(S)|   \ge  d_{1,7} \kappa_U^{1 / 2} \sigma_x \sigma_{z}\left\{\sqrt{\frac{x}{m_*^{\widehat{\tau}^2}}} \sqrt{\overline{\mathrm{b}}_S}+\kappa_U^{1 / 2} \sigma_{\varepsilon} \frac{x}{m_*^{\widehat{\tau}}}\right\} \right] \le 2e^{-x},\quad\forall x>0.
$$

Notice the above high probability bound is true for $\boldsymbol{\beta}\in\mathbb{R}^p$ with fixed support set $S$. The total number of $S$ satisfying $\left|S \backslash S^*\right|=r$ can be upper bounded by
$$
	N_r=2^{s^*} \times\left(\begin{array}{c}
			p-s^* \\
			r
		\end{array}\right) \leq 2^{s^*}(e p / r)^r.
$$

Let $x = u + \log(2N_r)$, we have
\begin{equation}
	\mathbb{P}\left[      |Z(S)|   \ge  d_{1,7} \kappa_U^{1 / 2} \sigma_x \sigma_{z}\left\{\sqrt{\frac{u + \log(2N_r)}{m^{\widehat{\tau}^2}_*}} \sqrt{\overline{\mathrm{b}}_S}+\kappa_U^{1 / 2} \sigma_{\varepsilon} \frac{u + \log(2N_r)}{m^{\widehat{\tau}}_*}\right\} \right] \le \frac{e^{-u}}{N_r},\quad\forall u>0.
	\label{eq:ulog2N_r_stept16}
\end{equation}

Now we are ready to define a new event that holds true simultaneously for a group of $S$ that satisfies $\left|S \backslash S^*\right|=r\in\{1,...,p\}$:
$$
	\mathcal{K}_u(r) = \left\{\forall S,\left|S \backslash S^*\right|=r, \quad|Z(S)| \leq  d_{1,7} \kappa_U^{1 / 2} \sigma_x \sigma_{z}\left\{\sqrt{\frac{u + \log(2N_r)}{m^{\widehat{\tau}^2}_*}} \sqrt{\overline{\mathrm{b}}_S}+\kappa_U^{1 / 2} \sigma_{\varepsilon} \frac{u + \log(2N_r)}{m^{\widehat{\tau}}_*}\right\}\right\},
$$
where $u>0$. In order to study the probability of this event, it is more intuitive to work with its complement:
$$
	\mathcal{K}^c_u(r) = \left\{\exists S,\left|S \backslash S^*\right|=r, \quad|Z(S)| \geq  d_{1,7} \kappa_U^{1 / 2} \sigma_x \sigma_{z}\left\{\sqrt{\frac{u + \log(2N_r)}{m^{\widehat{\tau}^2}_*}} \sqrt{\overline{\mathrm{b}}_S}+\kappa_U^{1 / 2} \sigma_{\varepsilon} \frac{u + \log(2N_r)}{m^{\widehat{\tau}}_*}\right\}\right\}
$$

Using equation \ref{eq:ulog2N_r_stept16} and applying a union bound over $S$ that satisfies $\left|S \backslash S^*\right|=r$, we have
\begin{align*}
	\mathbb{P}[\mathcal{K}^c_u(r) ] & = \sum_{\left|S \backslash S^*\right|=r} \mathbb{P}\left[      |Z(S)|   \ge  d_{1,7} \kappa_U^{1 / 2} \sigma_x \sigma_{z}\left\{\sqrt{\frac{u + \log(2N_r)}{m^{\widehat{\tau}^2}_*}} \sqrt{\overline{\mathrm{b}}_S}+\kappa_U^{1 / 2} \sigma_{\varepsilon} \frac{u + \log(2N_r)}{m^{\widehat{\tau}}_*}\right\} \right] \\
	                                & \le \sum_{\left|S \backslash S^*\right|=r} \frac{e^{-u}}{N_r}                                                                                                                                                                                                                                                         \\
	                                & = e^{-u},
\end{align*}
and hence $\mathbb{P}[\mathcal{K}_u(r) ]\ge 1-e^{-u}$. However, so far the high probability bound only concerns a specific $r$ rather than all $r\in\{1,...,p\}$. Now consider the event $\bigcap_{r=1}^p \mathcal{K}_{t+\log p}(r)$, and notice
\begin{align*}
	\mathbb{P}\left[\bigcap_{r=1}^p \mathcal{K}_{t+\log p}(r)\right] & = 1- \mathbb{P}\left[\bigcup_{r=1}^p \mathcal{K}^c_{t+\log p}(r)\right] \\
	                                                                 & \ge 1-\sum_{r=1}^p\mathbb{P}\left[ \mathcal{K}^c_{t+\log p}(r)\right]   \\
	                                                                 & \ge 1-\sum_{r=1}^p e^{-t-\log p}                                        \\
	                                                                 & = 1-e^{-t}.
\end{align*}
Along with the following fact that $\forall r\ge 1$,
\begin{align*}
	(t + \log p)+ \log(2N_r) & \le 2(t +\log p + r\log(ep/r)+s^*) \\
	                         & \le 4r\log(ep/r)+2s^*+2t           \\
	                         & \le 4(r\log(ep/r)+s^*+t)           \\
	                         & \le 4(p+s^*+t)                     \\
	                         & \le 8(p+t),
\end{align*}
with $D_{1,7} = 8d_{1,7}$, we have the event
\begin{align*}
	C_{7, t}=\left\{\forall \boldsymbol{\beta}\in \mathbb{R}^p, \quad \sum_{e \in \mathcal{E}} \omega^{(e)} \mathrm{T}_{1,7}^{(e)}(\boldsymbol{\beta}) \leq D_{1,7} \kappa_U^{1 / 2} \sigma_x \sigma_{z} \sqrt{\frac{p+t}{m^{\widehat{\tau}^2}_*}}\right. \\
	\left.\times \sqrt{\sum_{e \in \mathcal{E}} \omega^{(e)}\left\|\mathbb{E}\left[x_S^{(e)} \varepsilon^{(e)}\right]\right\|_2^2}+D_{1,7}  \kappa_U \sigma_x \sigma_z\sigma_{\varepsilon} \frac{p+t}{m^{\widehat{\tau}}_*}\right\}
\end{align*}
takes place with probability at least $1-e^{-t}$ for arbitrary $t>0$.

\noindent{\it Step 1.8} UPPER BOUND ON $T_{1,8}^{(e)}(\boldsymbol{\beta})$. Here we have that $\mathbb{P}\left(C_{8, t}\right) \geq 1-e^{-t}$ for any $t>0$, where
\begin{align}
	C_{8, t}=\left\{\forall \boldsymbol{\beta} \in \mathbb{R}^p, \quad \sum_{e \in \mathcal{E}} \omega^{(e)} \mathrm{T}_{1,8}^{(e)}(\boldsymbol{\beta}) \leq D_{1,8} \kappa_U^{1 / 2} \sigma_x \sigma_{z} \sqrt{\frac{t+p}{n^{\widehat{\tau}^2}_*}} \right.\notag \\
	\left. \times \sqrt{\sum_{e \in \mathcal{E}} \omega^{(e)}\left\|\mathbb{E}\left[x_S^{(e)} \varepsilon^{(e)}\right]\right\|_2^2}+D_{1,8} \kappa_U \sigma_x \sigma_{z}\sigma_{\varepsilon}\frac{t+p}{n^{\widehat{\tau}}_*}\right\}
	\label{eq:event_c8t}
\end{align}
where $D_{1,8}$ is some universal constant.

\noindent{\it Step 1.9} UPPER BOUND ON $T_{1,9}^{(e)}(\boldsymbol{\beta})$. In this step, the event
\begin{equation}
	C_{9, t}=\left\{\forall \boldsymbol{\beta} \in \mathbb{R}^p, \quad \sum_{e \in \mathcal{E}} \omega^{(e)} \mathrm{T}_{1,9}^{(e)}(\boldsymbol{\beta}) \leq D_{1,9} \kappa_U^{3 / 2} \sigma_x \left\|\boldsymbol{\beta}-\boldsymbol{\beta}^*\right\|_2\left(\sqrt{\frac{t+p}{m_\omega^{\widehat{\tau}^2,|\eta|^2}}}+\frac{t+p}{m_*^{\widehat{\tau},|\eta|}}\right)\right\}
	\label{eq:event_c9t}
\end{equation}
occurs with probability at least $1-e^{-t},\forall t>0$.

\noindent{\it Step 1.10} UPPER BOUND ON $T_{1,10}^{(e)}(\boldsymbol{\beta})$. In this step, the event
\begin{equation}
	C_{10, t}=\left\{\forall \boldsymbol{\beta} \in \mathbb{R}^p, \quad \sum_{e \in \mathcal{E}} \omega^{(e)} \mathrm{T}_{1,10}^{(e)}(\boldsymbol{\beta}) \leq D_{1,10} \kappa_U^{3 / 2} \sigma_x \left\|\boldsymbol{\beta}-\boldsymbol{\beta}^*\right\|_2\left(\sqrt{\frac{t+p}{n_\omega^{\widehat{\tau}^2,|\eta|^2}}}+\frac{t+p}{n_*^{\widehat{\tau},|\eta|}}\right)\right\},
	\label{eq:event_c10t}
\end{equation}
which occurs with probability at least $1-e^{-t},\forall t>0$.

\noindent{\it Step 1.11} UPPER BOUND ON $T_{1,11}^{(e)}(\boldsymbol{\beta})$. In this step we claim that $\mathbb{P}\left(C_{11, t}\right) \geq 1-e^{-t}$ for any $t>0$, where
\begin{align}
	C_{11, t}=\left\{\forall \boldsymbol{\beta} \in \mathbb{R}^p, \quad \sum_{e \in \mathcal{E}} \omega^{(e)} \mathrm{T}_{1,11}^{(e)}(\boldsymbol{\beta}) \leq D_{1,11} \kappa_U^{1 / 2} \sigma_x  \sqrt{\frac{t+p}{m^{\widehat{\tau}^2,|\eta|^2}_*}} \right.\notag \\
	\left. \times \sqrt{\sum_{e \in \mathcal{E}} \omega^{(e)}\left\|\mathbb{E}\left[x_S^{(e)} \varepsilon^{(e)}\right]\right\|_2^2}+D_{1,11} \kappa_U \sigma_x\sigma_{\varepsilon}\frac{t+p}{m^{\widehat{\tau},|\eta|}_*}\right\}
	\label{eq:event_c11t}
\end{align}
for some universal constant $D_{1,11} $. For a fixed $S$, we can write down $\sum_{e \in \mathcal{E}} \omega^{(e)} \mathrm{T}_{1,11}^{(e)}(\boldsymbol{\beta})$ as sum of independent random variables as
\begin{align*}
	Z(S) \equiv \sum_{e \in \mathcal{E}} \omega^{(e)} \mathrm{T}_{1,11}^{(e)}(\boldsymbol{\beta}) & = \sum_{e \in \mathcal{E}} \sum_{\ell=1}^{m^{(e)}} \frac{\omega^{(e)}\widehat{\tau}^{(e)}\eta^{(e)}}{m^{(e)}}\left(\mathbb{E}[\boldsymbol{x}_S^{(e)}\varepsilon^{(e)}]^{\top}[\boldsymbol{x}_{\ell}^{(e)}]_S -\mathbb{E}\left[\mathbb{E}[\boldsymbol{x}_S^{(e)} \varepsilon^{(e)}]^{\top}[\boldsymbol{x}_{\ell}^{(e)}]\right]\right).
\end{align*}
By Condition \ref{cond:subg_x}, it is a sum of sub-Exponential random variables with parameters
$$\left(\frac{\omega^{(e)}\widehat{\tau}^{(e)}|\eta^{(e)}|}{m^{(e)}} \kappa_U^{1/2}\sigma_x\left\|\mathbb{E}[\boldsymbol{x}_S^{(e)}\varepsilon^{(e)}]\right\|_2, \frac{\omega^{(e)}\widehat{\tau}^{(e)}|\eta^{(e)}|}{m^{(e)}} \kappa_U^{1/2}\sigma_x\left\|\mathbb{E}[\boldsymbol{x}_S^{(e)}\varepsilon^{(e)}]\right\|_2\right).$$
Then it follows that
\begin{align*}
	 & \mathbb{P}\left[      |Z(S)|   \ge  d_{1,11}  k_U^{1/2} \sigma_x \left\{\sqrt{\sum_{e \in \mathcal{E}}\left(\omega^{(e)}\widehat{\tau}^{(e)}|\eta^{(e)}|\right)^2 \frac{1}{m^{(e)}}\left\|\mathbb{E}\left[x_S^{(e)} \varepsilon^{(e)}\right]\right\|_2^2} \times \sqrt{x} \right.\right. \\
	 & \quad \quad\quad\quad\quad\quad\quad\quad\quad\quad\quad\quad\quad\left.\left.+\max _{e \in \mathcal{E}} \frac{\omega^{(e)}\widehat{\tau}^{(e)}|\eta^{(e)}|}{m^{(e)}}\left\|\mathbb{E}\left[x_S^{(e)} \varepsilon^{(e)}\right]\right\|_2 \times x\right\} \right] \le 2e^{-x},
\end{align*}
for any $x>0$. Next observe:
\begin{align*}
	 & d_{1,11}  k_U^{1/2} \sigma_x  \left\{\sqrt{\sum_{e \in \mathcal{E}}\left(\omega^{(e)}\widehat{\tau}^{(e)}|\eta^{(e)}|\right)^2 \frac{1}{m^{(e)}}\left\|\mathbb{E}\left[x_S^{(e)} \varepsilon^{(e)}\right]\right\|_2^2} \times \sqrt{x}+\max _{e \in \mathcal{E}} \frac{\omega^{(e)}\widehat{\tau}^{(e)}|\eta^{(e)}|}{m^{(e)}}\left\|\mathbb{E}\left[x_S^{(e)} \varepsilon^{(e)}\right]\right\|_2 \times x\right\} \\
	 & \le d_{1,11}  k_U^{1/2} \sigma_x \left\{\sqrt{\frac{x}{m_*^{\widehat{\tau}^2,|\eta|^2}}} \sqrt{\sum_{e \in \mathcal{E}} \omega^{(e)}\left\|\mathbb{E}\left[\boldsymbol{x}_S^{(e)} \varepsilon^{(e)}\right]\right\|_2^2}+\frac{x}{m_*^{\widehat{\tau},|\eta|}} \max _{e \in \mathcal{E}}\left\|\mathbb{E}\left[\boldsymbol{x}^{(e)} \varepsilon^{(e)}\right]\right\|_2\right\}                                     \\
	 & \le   d_{1,11} \kappa_U^{1 / 2} \sigma_x \left\{\sqrt{\frac{x}{m_*^{\widehat{\tau}^2,|\eta|^2}}} \sqrt{\overline{\mathrm{b}}_S}+\kappa_U^{1 / 2} \sigma_{\varepsilon} \frac{x}{m_*^{\widehat{\tau},|\eta|}}\right\},
\end{align*}
where the last line is by the notation $\overline{\mathrm{b}}_S$ and the fact that $\left\|\mathbb{E}\left[\varepsilon^{(e)} \boldsymbol{x}^{(e)}\right]\right\|_2 \leq \sigma_{\varepsilon} \kappa_U^{1 / 2}$, which is proved as lemma 8 in \cite{fan2024environment}. Therefore, we have
$$
	\mathbb{P}\left[      |Z(S)|   \ge  d_{1,11} \kappa_U^{1 / 2} \sigma_x\left\{\sqrt{\frac{x}{m_*^{\widehat{\tau}^2,|\eta|^2}}} \sqrt{\overline{\mathrm{b}}_S}+\kappa_U^{1 / 2} \sigma_{\varepsilon} \frac{x}{m_*^{\widehat{\tau},|\eta|}}\right\} \right] \le 2e^{-x},\quad\forall x>0.
$$

Notice the above high probability bound is true for $\boldsymbol{\beta}\in\mathbb{R}^p$ with fixed support set $S$. The total number of $S$ satisfying $\left|S \backslash S^*\right|=r$ can be upper bounded by
$$
	N_r=2^{s^*} \times\left(\begin{array}{c}
			p-s^* \\
			r
		\end{array}\right) \leq 2^{s^*}(e p / r)^r.
$$

Let $x = u + \log(2N_r)$, we have
\begin{equation}
	\mathbb{P}\left[      |Z(S)|   \ge  d_{1,11} \kappa_U^{1 / 2} \sigma_x \left\{\sqrt{\frac{u + \log(2N_r)}{m^{\widehat{\tau}^2,|\eta|^2}_*}} \sqrt{\overline{\mathrm{b}}_S}+\kappa_U^{1 / 2} \sigma_{\varepsilon} \frac{u + \log(2N_r)}{m^{\widehat{\tau},|\eta|}_*}\right\} \right] \le \frac{e^{-u}}{N_r},\quad\forall u>0.
	\label{eq:ulog2N_r_stept16}
\end{equation}

Now we are ready to define a new event that holds true simultaneously for a group of $S$ that satisfies $\left|S \backslash S^*\right|=r\in\{1,...,p\}$:
$$
	\mathcal{K}_u(r) = \left\{\forall S,\left|S \backslash S^*\right|=r, \quad|Z(S)| \leq  d_{1,11} \kappa_U^{1 / 2} \sigma_x \left\{\sqrt{\frac{u + \log(2N_r)}{m^{\widehat{\tau}^2,|\eta|^2}_*}} \sqrt{\overline{\mathrm{b}}_S}+\kappa_U^{1 / 2} \sigma_{\varepsilon} \frac{u + \log(2N_r)}{m^{\widehat{\tau},|\eta|}_*}\right\}\right\},
$$
where $u>0$. In order to study the probability of this event, it is more intuitive to work with its complement:
$$
	\mathcal{K}^c_u(r) = \left\{\exists S,\left|S \backslash S^*\right|=r, \quad|Z(S)| \geq  d_{1,11} \kappa_U^{1 / 2} \sigma_x \left\{\sqrt{\frac{u + \log(2N_r)}{m^{\widehat{\tau}^2,|\eta|^2}_*}} \sqrt{\overline{\mathrm{b}}_S}+\kappa_U^{1 / 2} \sigma_{\varepsilon} \frac{u + \log(2N_r)}{m^{\widehat{\tau},|\eta|}_*}\right\}\right\}
$$

Using equation \ref{eq:ulog2N_r_stept16} and applying a union bound over $S$ that satisfies $\left|S \backslash S^*\right|=r$, we have
\begin{align*}
	\mathbb{P}[\mathcal{K}^c_u(r) ] & = \sum_{\left|S \backslash S^*\right|=r} \mathbb{P}\left[      |Z(S)|   \ge  d_{1,11} \kappa_U^{1 / 2} \sigma_x \left\{\sqrt{\frac{u + \log(2N_r)}{m^{\widehat{\tau}^2,|\eta|^2}_*}} \sqrt{\overline{\mathrm{b}}_S}+\kappa_U^{1 / 2} \sigma_{\varepsilon} \frac{u + \log(2N_r)}{m^{\widehat{\tau},|\eta|}_*}\right\} \right] \\
	                                & \le \sum_{\left|S \backslash S^*\right|=r} \frac{e^{-u}}{N_r}                                                                                                                                                                                                                                                                \\
	                                & = e^{-u},
\end{align*}
and hence $\mathbb{P}[\mathcal{K}_u(r) ]\ge 1-e^{-u}$. However, so far the high probability bound only concerns a specific $r$ rather than all $r\in\{1,...,p\}$. Now consider the event $\bigcap_{r=1}^p \mathcal{K}_{t+\log p}(r)$, and notice
\begin{align*}
	\mathbb{P}\left[\bigcap_{r=1}^p \mathcal{K}_{t+\log p}(r)\right] & = 1- \mathbb{P}\left[\bigcup_{r=1}^p \mathcal{K}^c_{t+\log p}(r)\right] \\
	                                                                 & \ge 1-\sum_{r=1}^p\mathbb{P}\left[ \mathcal{K}^c_{t+\log p}(r)\right]   \\
	                                                                 & \ge 1-\sum_{r=1}^p e^{-t-\log p}                                        \\
	                                                                 & = 1-e^{-t}.
\end{align*}
Along with the following fact that $\forall r\ge 1$,
\begin{align*}
	(t + \log p)+ \log(2N_r) & \le 2(t +\log p + r\log(ep/r)+s^*) \\
	                         & \le 4r\log(ep/r)+2s^*+2t           \\
	                         & \le 4(r\log(ep/r)+s^*+t)           \\
	                         & \le 4(p+s^*+t)                     \\
	                         & \le 8(p+t),
\end{align*}
with $D_{1,7} = 8d_{1,7}$, we have the event
\begin{align*}
	C_{11, t}=\left\{\forall \boldsymbol{\beta}\in \mathbb{R}^p, \quad \sum_{e \in \mathcal{E}} \omega^{(e)} \mathrm{T}_{1,11}^{(e)}(\boldsymbol{\beta}) \leq D_{1,11} \kappa_U^{1 / 2} \sigma_x  \sqrt{\frac{p+t}{m^{\widehat{\tau}^2,|\eta|^2}_*}}\right. \\
	\left.\times \sqrt{\sum_{e \in \mathcal{E}} \omega^{(e)}\left\|\mathbb{E}\left[x_S^{(e)} \varepsilon^{(e)}\right]\right\|_2^2}+D_{1,11}  \kappa_U \sigma_x \sigma_{\varepsilon} \frac{p+t}{m^{\widehat{\tau},|\eta|}_*}\right\}
\end{align*}
takes place with probability at least $1-e^{-t}$ for arbitrary $t>0$.

\noindent{\it Step 1.12} UPPER BOUND ON $T_{1,12}^{(e)}(\boldsymbol{\beta})$. In this step we claim that $\mathbb{P}\left(C_{12, t}\right) \geq 1-e^{-t}$ for any $t>0$, where
\begin{align}
	C_{12, t}=\left\{\forall \boldsymbol{\beta} \in \mathbb{R}^p, \quad \sum_{e \in \mathcal{E}} \omega^{(e)} \mathrm{T}_{1,12}^{(e)}(\boldsymbol{\beta}) \leq D_{1,12} \kappa_U^{1 / 2} \sigma_x  \sqrt{\frac{t+p}{n^{\widehat{\tau}^2,|\eta|^2}_*}} \right.\notag \\
	\left. \times \sqrt{\sum_{e \in \mathcal{E}} \omega^{(e)}\left\|\mathbb{E}\left[x_S^{(e)} \varepsilon^{(e)}\right]\right\|_2^2}+D_{1,12} \kappa_U \sigma_x\sigma_{\varepsilon}\frac{t+p}{n^{\widehat{\tau},|\eta|}_*}\right\}
	\label{eq:event_c12t}
\end{align}
for some universal constant $D_{1,12} $.

\noindent{\it Step 2.1}. UPPER BOUND ON $T_{2,1}^{(e)}(\boldsymbol{\beta})$. The goal of this step is to derive a high-probability bound for the event
\begin{equation}
	\mathcal{U}_{1, t}=\left\{\forall \boldsymbol{\beta} \in \mathbb{R}^p, \quad \sum_{e \in \mathcal{E}} \omega^{(e)} \mathrm{T}_{2,1}^{(e)}(\boldsymbol{\beta}) \leq E_{2,1} \kappa_U \sigma_x^2 \sigma_{\varepsilon}^2 \frac{t+\log \left(2|\mathcal{E}|\left|S^*\right|\right)}{\bar{N}}\left|S^* \backslash S\right|\right\}
	\label{eq:event_u1t}
\end{equation}
for any $t \in\left(0, N_{\min }-\log \left(2\left|\mathcal{E} \| S^*\right|\right)\right]$. Note that both the L.H.S. and R.H.S. of the above inequality depends on $\boldsymbol{\beta}$, or more precisely, $S=\operatorname{supp}(\boldsymbol{\beta})$. Denoting $\delta=E_{2,1} \kappa_U \sigma_x^2 \sigma_{\varepsilon}^2\left\{t+\log \left(2|\mathcal{E}|\left|S^*\right|\right)\right\} / \bar{N}$, observe the following decomposition
\begin{align*}
	\mathcal{U}_{1, t} & =\bigcup_{T \subset S^*}\left\{\forall \boldsymbol{\beta} \in \mathbb{R}^p, S^* \backslash \operatorname{supp}(\boldsymbol{\beta})=T, \quad \sum_{e \in \mathcal{E}} \omega^{(e)} T_{2,1}^{(e)}(\boldsymbol{\beta}) \leq \delta|T|\right\}                                                                              \\
	                   & =\bigcup_{T \subset S^*}\left\{\forall \boldsymbol{\beta} \in \mathbb{R}^p, S^* \backslash \operatorname{supp}(\boldsymbol{\beta})=T, \quad \sum_{j \in T} \sum_{e \in \mathcal{E}} \omega^{(e)}\frac{1}{2}\left|\widehat{\mathbb{E}}_{N^{(e)}}\left[x_j^{(e)} \varepsilon^{(e)}\right]\right|^2 \leq \delta|T|\right\} \\
	                   & =\bigcup_{T \subset S^*} \mathcal{K}(T) .
\end{align*}
At the same time, given fixed $j \in S^*$ and $e \in \mathcal{E}$, it follows from Condition \ref{cond:subg_x} and Condition \ref{cond:subg_e} that,
$$
	\mathbb{P}\left[\mathcal{K}_{1, x}(e, j)\right]=\mathbb{P}\left[\left|\widehat{\mathbb{E}}_{N^{(e)}}\left[x_j^{(e)} \varepsilon^{(e)}\right]\right| \leq C^{\prime} \kappa_U^{1 / 2} \sigma_x \sigma_\epsilon\left(\sqrt{\frac{x}{N^{(e)}}}+\frac{x}{N^{(e)}}\right)\right] \geq 1-2 e^{-x}
$$
for some universal constant $C^{\prime}$. Meanwhile, we can write
$$
	\mathbb{P}\left[\mathcal{K}^c_{1, x}(e, j)\right]=\mathbb{P}\left[\left|\widehat{\mathbb{E}}_{N^{(e)}}\left[x_j^{(e)} \varepsilon^{(e)}\right]\right| \geq C^{\prime} \kappa_U^{1 / 2} \sigma_x \sigma_\epsilon\left(\sqrt{\frac{x}{N^{(e)}}}+\frac{x}{N^{(e)}}\right)\right] \leq 2 e^{-x}.
$$
Under the event $\bigcap_{j\in S^*, e\in\mathcal{E}}\mathcal{K}_{1, t+\log(2s^*|\mathcal{E}|)}(e, j)$, we have
\begin{align*}
	\frac{1}{2}\sum_{j \in T} \sum_{e \in \mathcal{E}} \omega^{(e)}\left|\widehat{\mathbb{E}}_{N^{(e)}}\left[x_j^{(e)} \varepsilon^{(e)}\right]\right|^2 & \le \sum_{j \in T} \sum_{e \in \mathcal{E}} \omega^{(e)} \frac{1}{2}\left\{C^{\prime} \kappa_U^{1 / 2} \sigma_x \sigma_\epsilon\left(\sqrt{\frac{t+\log(2s^*|\mathcal{E}|)}{N^{(e)}}}+\frac{t+\log(2s^*|\mathcal{E}|)}{N^{(e)}}\right)\right\}^2 \\
	                                                                                                                                                     & \le \sum_{j \in T} \sum_{e \in \mathcal{E}}\left(\frac{1}{2}\right)2(C^{\prime})^2  \kappa_U \sigma_x^2 \sigma_{\varepsilon}^2\left(\frac{t+\log(2s^*|\mathcal{E}|)}{N^{(e)}} \omega^{(e)}\right)                                                \\
	                                                                                                                                                     & =  \sum_{j \in T}(C^{\prime})^2  \kappa_U \sigma_x^2 \sigma_{\varepsilon}^2(t+\log(2s^*|\mathcal{E}|)\left( \sum_{e \in \mathcal{E}} \frac{\omega^{(e)}}{N^{(e)}}\right)                                                                         \\
	                                                                                                                                                     & \stackrel{(a)}{=}  \left\{(C^{\prime})^2  \kappa_U \sigma_x^2 \sigma_{\varepsilon}^2\left(\frac{t+\log(2s^*|\mathcal{E}|}{\bar{N}}\right)\right\}|T|                                                                                             \\
	                                                                                                                                                     & \stackrel{(b)}{=}  \delta |T|,
\end{align*}
where $(a)$ holds provided $t+\log \left(2 s^*|\mathcal{E}|\right) \leq N_{\min }$, and $(b)$ holds if we choose $E_{2,1} = (C^{\prime})^2 $. Therefore, we have
$$
	\bigcap_{j\in S^*, e\in\mathcal{E}}\mathcal{K}_{1, t+\log(2s^*|\mathcal{E}|)}(e, j)\subset \mathcal{K}(T)\subset \bigcup_{T \subset S^*} \mathcal{K}(T).
$$
By this fact, we can continue deriving the probability of event $\mathcal{U}_{1, t}$ as following:
\begin{align*}
	\mathbb{P}\left(\mathcal{U}_{1, t}\right)=\mathbb{P}\left[\bigcup_{T \subset S^*} \mathcal{K}(T)\right] & \geq \mathbb{P}\left[\bigcap_{j \in S^*, e \in \mathcal{E}} \mathcal{K}_{1, t+\log \left(2 s^* | \mathcal{E}|\right)}(e, j)\right]  \\
	                                                                                                        & =1- \mathbb{P}\left[\bigcup_{j \in S^*, e \in \mathcal{E}} \mathcal{K}^c_{1, t+\log \left(2 s^* | \mathcal{E}|\right)}(e, j)\right] \\
	                                                                                                        & \ge 1-\sum_{j \in S^*, e \in \mathcal{E}} \mathbb{P}\left[\mathcal{K}^c_{1, t+\log \left(2 s^* | \mathcal{E}|\right)}(e, j)\right]  \\
	                                                                                                        & \ge 1- \sum_{j \in S^*, e \in \mathcal{E}} 2e^{-t}e^{-\log \left(2 s^* | \mathcal{E}|\right)}                                       \\
	                                                                                                        & = 1- e^{-t}, \quad \forall t \in\left(0, N_{\min }-\log \left(2\left|\mathcal{E} \| S^*\right|\right)\right]
\end{align*}
as desired.

\noindent{\it Step 2.2} UPPER BOUND ON $T_{2,2}^{(e)}(\boldsymbol{\beta})$. The goal of this step is to derive a high-probability bound for the event
\begin{equation}
	\mathcal{U}_{2, t}=\left\{\forall \boldsymbol{\beta} \in \mathbb{R}^p, \quad \sum_{e \in \mathcal{E}} \omega^{(e)} \mathrm{T}_{2,2}^{(e)}(\boldsymbol{\beta}) \leq E_{2,2} \kappa_U \sigma_x^2 \sigma_{z}^2 \frac{t+\log \left(2|\mathcal{E}|\left|S^*\right|\right)}{\bar{N}}\left|S^* \backslash S\right|\right\}
	\label{eq:eq:event_u2t}
\end{equation}
for any $t \in\left(0, N_{\min }-\log \left(2\left|\mathcal{E} \| S^*\right|\right)\right]$. Note that both the L.H.S. and R.H.S. of the above inequality depends on $\boldsymbol{\beta}$, or more precisely, $S=\operatorname{supp}(\boldsymbol{\beta})$. Denoting $\delta=E_{2,2} \kappa_U \sigma_x^2 \sigma_{z}^2\left\{t+\log \left(2|\mathcal{E}|\left|S^*\right|\right)\right\} / \bar{N}$, observe the following decomposition
\begin{align*}
	\mathcal{U}_{2, t} & =\bigcup_{T \subset S^*}\left\{\forall \boldsymbol{\beta} \in \mathbb{R}^p, S^* \backslash \operatorname{supp}(\boldsymbol{\beta})=T, \quad \sum_{e \in \mathcal{E}} \omega^{(e)} T_{2,2}^{(e)}(\boldsymbol{\beta}) \leq \delta|T|\right\}                                                                                                                                                                              \\
	                   & =\bigcup_{T \subset S^*}\left\{\forall \boldsymbol{\beta} \in \mathbb{R}^p, S^* \backslash \operatorname{supp}(\boldsymbol{\beta})=T, \quad \sum_{j \in T} \sum_{e \in \mathcal{E}} \omega^{(e)}\left|\widehat{\mathbb{E}}_{N^{(e)}}\left[\boldsymbol{x}_{j}^{(e)}\left(z^{(e)}-\eta^{(e)}\right)\right]-\mathbb{E}\left[\boldsymbol{x}_{j}^{(e)}\left(z^{(e)}-\eta^{(e)}\right)\right]\right|^2 \leq \delta|T|\right\} \\
	                   & =\bigcup_{T \subset S^*} \mathcal{K}(T) .
\end{align*}
At the same time, given fixed $j \in S^*$ and $e \in \mathcal{E}$, it follows from Conditions Condition \ref{cond:subg_x} and Condition \ref{cond:subg_z} that,
$$
	\mathbb{P}\left[\mathcal{K}_{2, x}(e, j)\right]=\mathbb{P}\left[\left|\widehat{\mathbb{E}}_{N^{(e)}}\left[\boldsymbol{x}_{j}^{(e)}\left(z^{(e)}-\eta^{(e)}\right)\right]-\mathbb{E}\left[\boldsymbol{x}_{j}^{(e)}\left(z^{(e)}-\eta^{(e)}\right)\right]\right| \leq C^{\prime} \kappa_U^{1 / 2} \sigma_x \sigma_z\left(\sqrt{\frac{x}{N^{(e)}}}+\frac{x}{N^{(e)}}\right)\right] \geq 1-2 e^{-x}
$$
for some universal constant $C^{\prime}$. Meanwhile, we can write
$$
	\mathbb{P}\left[\mathcal{K}^c_{2, x}(e, j)\right]=\mathbb{P}\left[\left|\widehat{\mathbb{E}}_{N^{(e)}}\left[\boldsymbol{x}_{j}^{(e)}\left(z^{(e)}-\eta^{(e)}\right)\right]-\mathbb{E}\left[\boldsymbol{x}_{j}^{(e)}\left(z^{(e)}-\eta^{(e)}\right)\right]\right| \geq C^{\prime} \kappa_U^{1 / 2} \sigma_x \sigma_z\left(\sqrt{\frac{x}{N^{(e)}}}+\frac{x}{N^{(e)}}\right)\right] \leq 2 e^{-x}.
$$
Under the event $\bigcap_{j\in S^*, e\in\mathcal{E}}\mathcal{K}_{2, t+\log(2s^*|\mathcal{E}|)}(e, j)$, we have
\begin{align*}
	 & \sum_{j \in T} \sum_{e \in \mathcal{E}} \omega^{(e)}\left(\frac{1}{2}\right)\left|\widehat{\mathbb{E}}_{N^{(e)}}\left[\boldsymbol{x}_{j}^{(e)}\left(z^{(e)}-\eta^{(e)}\right)\right]-\mathbb{E}\left[\boldsymbol{x}_{j}^{(e)}\left(z^{(e)}-\eta^{(e)}\right)\right]\right|^2 \\
	 & \quad \le \sum_{j \in T} \sum_{e \in \mathcal{E}} \omega^{(e)} \left(\frac{1}{2}\right)\left\{C^{\prime} \kappa_U^{1 / 2} \sigma_x \sigma_z\left(\sqrt{\frac{t+\log(2s^*|\mathcal{E}|)}{N^{(e)}}}+\frac{t+\log(2s^*|\mathcal{E}|)}{N^{(e)}}\right)\right\}^2                 \\
	 & \quad\le \sum_{j \in T} \sum_{e \in \mathcal{E}}\left(\frac{1}{2}\right)2(C^{\prime})^2  \kappa_U \sigma_x^2 \sigma_{z}^2\left(\frac{t+\log(2s^*|\mathcal{E}|)}{N^{(e)}} \omega^{(e)}\right)                                                                                 \\
	 & \quad=  \sum_{j \in T}(C^{\prime})^2  \kappa_U \sigma_x^2 \sigma_{z}^2(t+\log(2s^*|\mathcal{E}|)\left( \sum_{e \in \mathcal{E}} \frac{\omega^{(e)}}{N^{(e)}}\right)                                                                                                          \\
	 & \quad\stackrel{(a)}{=}  \left\{(C^{\prime})^2  \kappa_U \sigma_x^2 \sigma_{z}^2\left(\frac{t+\log(2s^*|\mathcal{E}|}{\bar{N}}\right)\right\}|T|                                                                                                                              \\
	 & \quad\stackrel{(b)}{=}  \delta |T|,
\end{align*}
where $(a)$ holds provided $t+\log \left(2 s^*|\mathcal{E}|\right) \leq N_{\min }$, and $(b)$ holds if we choose $E_{2,2} = (C^{\prime})^2 $. Therefore, we have
$$
	\bigcap_{j\in S^*, e\in\mathcal{E}}\mathcal{K}_{2, t+\log(2s^*|\mathcal{E}|)}(e, j)\subset \mathcal{K}(T)\subset \bigcup_{T \subset S^*} \mathcal{K}(T).
$$
By this fact, we can continue deriving the probability of event $\mathcal{U}_{1, t}$ as following:
\begin{align*}
	\mathbb{P}\left(\mathcal{U}_{2, t}\right)=\mathbb{P}\left[\bigcup_{T \subset S^*} \mathcal{K}(T)\right] & \geq \mathbb{P}\left[\bigcap_{j \in S^*, e \in \mathcal{E}} \mathcal{K}_{2, t+\log \left(2 s^* | \mathcal{E}|\right)}(e, j)\right]  \\
	                                                                                                        & =1- \mathbb{P}\left[\bigcup_{j \in S^*, e \in \mathcal{E}} \mathcal{K}^c_{2, t+\log \left(2 s^* | \mathcal{E}|\right)}(e, j)\right] \\
	                                                                                                        & \ge 1-\sum_{j \in S^*, e \in \mathcal{E}} \mathbb{P}\left[\mathcal{K}^c_{2, t+\log \left(2 s^* | \mathcal{E}|\right)}(e, j)\right]  \\
	                                                                                                        & \ge 1- \sum_{j \in S^*, e \in \mathcal{E}} 2e^{-t}e^{-\log \left(2 s^* | \mathcal{E}|\right)}                                       \\
	                                                                                                        & = 1- e^{-t}, \quad \forall t \in\left(0, N_{\min }-\log \left(2\left|\mathcal{E} \| S^*\right|\right)\right]
\end{align*}
as desired.

\noindent{\it Step 2.3} UPPER BOUND ON $T_{2,3}^{(e)}(\boldsymbol{\beta})$. The goal of this step is to derive a high-probability bound for the event
\begin{equation}
	\mathcal{U}_{3, t}=\left\{\forall \boldsymbol{\beta} \in \mathbb{R}^p, \quad \sum_{e \in \mathcal{E}} \omega^{(e)} \mathrm{T}_{2,3}^{(e)}(\boldsymbol{\beta}) \leq E_{2,3} \kappa_U \sigma_x^2 \sigma_{z}^2 \frac{t+\log \left(2|\mathcal{E}|\left|S^*\right|\right)}{\bar{n}}\left|S^* \backslash S\right|\right\}
	\label{eq:event_u3t}
\end{equation}
for any $t \in\left(0, n_{\min }-\log \left(2\left|\mathcal{E} \| S^*\right|\right)\right]$.

\noindent{\it Step 2.4}. UPPER BOUND ON $T_{2,4}^{(e)}(\boldsymbol{\beta})$. The goal of this step is to derive a high-probability bound for the event
\begin{equation}
	\mathcal{U}_{4, t}=\left\{\forall \boldsymbol{\beta} \in \mathbb{R}^p, \quad \sum_{e \in \mathcal{E}} \omega^{(e)} \mathrm{T}_{2,4}^{(e)}(\boldsymbol{\beta}) \leq E_{2,4} \kappa_U \sigma_x^2  \frac{t+\log \left(2|\mathcal{E}|\left|S^*\right|\right)}{\bar{N}^{|\eta|^2}}\left|S^* \backslash S\right|\right\}
	\label{eq:event_u4t}
\end{equation}
for any $t \in\left(0, N_{\min }-\log \left(2\left|\mathcal{E} \| S^*\right|\right)\right]$. Note that both the L.H.S. and R.H.S. of the above inequality depends on $\boldsymbol{\beta}$, or more precisely, $S=\operatorname{supp}(\boldsymbol{\beta})$. Denoting $\delta=E_{2,4}  \kappa_U \sigma_x^2\{t+\log(2s^*|\mathcal{E}|\}/\bar{N}^{|\eta|^2}$, observe the following decomposition
\begin{align*}
	\mathcal{U}_{4, t} & =\bigcup_{T \subset S^*}\left\{\forall \boldsymbol{\beta} \in \mathbb{R}^p, S^* \backslash \operatorname{supp}(\boldsymbol{\beta})=T, \quad \sum_{e \in \mathcal{E}} \omega^{(e)} T_{2,4}^{(e)}(\boldsymbol{\beta}) \leq \delta|T|\right\}                                                                                                         \\
	                   & =\bigcup_{T \subset S^*}\left\{\forall \boldsymbol{\beta} \in \mathbb{R}^p, S^* \backslash \operatorname{supp}(\boldsymbol{\beta})=T, \quad \sum_{j \in T} \sum_{e \in \mathcal{E}} \omega^{(e)} |\eta^{(e)}|^2\left|\widehat{\mathbb{E}}_{N^{(e)}}[\boldsymbol{x}_{j}^{(e)}]-\mathbb{E}[\boldsymbol{x}_{j}^{(e)}]\right|^2 \leq \delta|T|\right\} \\
	                   & =\bigcup_{T \subset S^*} \mathcal{K}(T) .
\end{align*}
At the same time, given fixed $j \in S^*$ and $e \in \mathcal{E}$, it follows from Condition \ref{cond:subg_x} that,
$$
	\mathbb{P}\left[\mathcal{K}_{4, x}(e, j)\right]=\mathbb{P}\left[\left|\widehat{\mathbb{E}}_{N^{(e)}}[\boldsymbol{x}_{j}^{(e)}]-\mathbb{E}[\boldsymbol{x}_{j}^{(e)}]\right| \leq C^{\prime} \kappa_U^{1 / 2} \sigma_x \left(\sqrt{ \frac{x}{N^{(e)}}}+ \frac{x}{N^{(e)}}\right)\right] \geq 1-2 e^{-x}
$$
for some universal constant $C^{\prime}$. Meanwhile, we can write
$$
	\mathbb{P}\left[\mathcal{K}^c_{4, x}(e, j)\right]=\mathbb{P}\left[\left|\widehat{\mathbb{E}}_{N^{(e)}}[\boldsymbol{x}_{j}^{(e)}]-\mathbb{E}[\boldsymbol{x}_{j}^{(e)}]\right| \geq C^{\prime} \kappa_U^{1 / 2} \sigma_x \left(\sqrt{ \frac{x}{N^{(e)}}}+ \frac{x}{N^{(e)}}\right)\right] \leq 2 e^{-x}.
$$
Under the event $\bigcap_{j\in S^*, e\in\mathcal{E}}\mathcal{K}_{4, t+\log(2s^*|\mathcal{E}|)}(e, j)$, we have
\begin{align*}
	 & \sum_{j \in T} \sum_{e \in \mathcal{E}} \omega^{(e)}|\eta^{(e)}|^2\left(\frac{1}{2}\right)\left|\widehat{\mathbb{E}}_{N^{(e)}}[\boldsymbol{x}_{j}^{(e)}]-\mathbb{E}[\boldsymbol{x}_{j}^{(e)}]\right|^2                                                               \\
	 & \quad \le \sum_{j \in T} \sum_{e \in \mathcal{E}} \omega^{(e)} |\eta^{(e)}|^2\left(\frac{1}{2}\right)\left\{C^{\prime} \kappa_U^{1 / 2} \sigma_x \left(\sqrt{ \frac{t+\log(2s^*|\mathcal{E}|)}{N^{(e)}}}+ \frac{t+\log(2s^*|\mathcal{E}|)}{N^{(e)}}\right)\right\}^2 \\
	 & \quad\le \sum_{j \in T} \sum_{e \in \mathcal{E}}\left(\frac{1}{2}\right)2(C^{\prime})^2  \kappa_U \sigma_x^2 \left(\frac{t+\log(2s^*|\mathcal{E}|)}{N^{(e)}} \omega^{(e)}|\eta^{(e)}|^2\right)                                                                       \\
	 & \quad=  \sum_{j \in T}(C^{\prime})^2  \kappa_U \sigma_x^2 (t+\log(2s^*|\mathcal{E}|)\left( \sum_{e \in \mathcal{E}} \frac{\omega^{(e)}|\eta^{(e)}|^2}{N^{(e)}}\right)                                                                                                \\
	 & \quad\stackrel{(a)}{=}  \left\{(C^{\prime})^2  \kappa_U \sigma_x^2\left(\frac{t+\log(2s^*|\mathcal{E}|}{\bar{N}^{|\eta|^2}}\right)\right\}|T|                                                                                                                        \\
	 & \quad\stackrel{(b)}{=}  \delta |T|,
\end{align*}
where $(a)$ holds provided $t+\log \left(2 s^*|\mathcal{E}|\right) \leq N_{\min }$, and $(b)$ holds if we choose $E_{2,2} = (C^{\prime})^2 $. Therefore, we have
$$
	\bigcap_{j\in S^*, e\in\mathcal{E}}\mathcal{K}_{4, t+\log(2s^*|\mathcal{E}|)}(e, j)\subset \mathcal{K}(T)\subset \bigcup_{T \subset S^*} \mathcal{K}(T).
$$
By this fact, we can continue deriving the probability of event $\mathcal{U}_{1, t}$ as following:
\begin{align*}
	\mathbb{P}\left(\mathcal{U}_{4, t}\right)=\mathbb{P}\left[\bigcup_{T \subset S^*} \mathcal{K}(T)\right] & \geq \mathbb{P}\left[\bigcap_{j \in S^*, e \in \mathcal{E}} \mathcal{K}_{4, t+\log \left(2 s^* | \mathcal{E}|\right)}(e, j)\right]  \\
	                                                                                                        & =1- \mathbb{P}\left[\bigcup_{j \in S^*, e \in \mathcal{E}} \mathcal{K}^c_{4, t+\log \left(2 s^* | \mathcal{E}|\right)}(e, j)\right] \\
	                                                                                                        & \ge 1-\sum_{j \in S^*, e \in \mathcal{E}} \mathbb{P}\left[\mathcal{K}^c_{4, t+\log \left(2 s^* | \mathcal{E}|\right)}(e, j)\right]  \\
	                                                                                                        & \ge 1- \sum_{j \in S^*, e \in \mathcal{E}} 2e^{-t}e^{-\log \left(2 s^* | \mathcal{E}|\right)}                                       \\
	                                                                                                        & = 1- e^{-t}, \quad \forall t \in\left(0, N_{\min }-\log \left(2\left|\mathcal{E} \| S^*\right|\right)\right]
\end{align*}

\noindent{\it Step 2.6} UPPER BOUND ON $T_{2,6}^{(e)}(\boldsymbol{\beta})$. In this step, we claim that the following event

\begin{equation}
	\mathcal{U}_{6,t} = \left\{ \forall \boldsymbol{\beta}\in\mathbb{R}^p,\quad  \sum_{e \in \mathcal{E}} \omega^{(e)} \mathrm{T}_{2,6}^{(e)}(\boldsymbol{\beta}) \leq E_{2,6} \kappa_U \sigma_x^2 \sigma_{\varepsilon}\sigma_{z} \frac{t+ \log(4|\mathcal{E}||S^*|)}{\overline{\sqrt{mN}}^{\widehat{\tau}}}|S^*\backslash S|\right\}
	\label{eq:event_u6t}
\end{equation}
occurs with probability at least $1-e^{-t}$ for any $t \in (0,m_{\min }-\log \left(4|\mathcal{E}| |S^*|\right)]$. Observe the following, $\forall \boldsymbol{\beta} \in \mathbb{R}^p$

\begin{align*}
	\sum_{e \in \mathcal{E}} \omega^{(e)} T_{2,6}^{(e)}(\boldsymbol{\beta}) & =\sum_{e \in \mathcal{E}} \omega^{(e)}\left\{\widehat{\mathbb{E}}_{N^{(e)}}\left[\boldsymbol{x}_{S^* \setminus S}^{(e)} \varepsilon^{(e)}\right]\right\}^{\top}\left\{\widehat{\tau}^{(e)}\left(\widehat{\mathbb{E}}_{m^{(e)}}\left[\boldsymbol{x}_{S^* \setminus S}^{(e)}\left(z^{(e)}-\eta^{(e)}\right)\right]-\mathbb{E}\left[\boldsymbol{x}_{S^* \setminus S}^{(e)}\left(z^{(e)}-\eta^{(e)}\right)\right]\right)\right\} \\
	                                                                        & =\sum_{e \in \mathcal{E}} \omega^{(e)}\widehat{\tau}^{(e)}\left\{\widehat{\mathbb{E}}_{N^{(e)}}\left[\boldsymbol{x}_{S^* \setminus S}^{(e)} \varepsilon^{(e)}\right]\right\}^{\top}\left\{\widehat{\mathbb{E}}_{m^{(e)}}\left[\boldsymbol{x}_{S^* \setminus S}^{(e)}\left(z^{(e)}-\eta^{(e)}\right)\right]-\mathbb{E}\left[\boldsymbol{x}_{S^* \setminus S}^{(e)}\left(z^{(e)}-\eta^{(e)}\right]\right)\right\}              \\
	                                                                        & \leq \sum_{e \in \mathcal{E}} \omega^{(e)}\widehat{\tau}^{(e)}  \left\|\widehat{\mathbb{E}}_{N^{(e)}}\left[\boldsymbol{x}_{S^* \setminus S}^{(e)} \varepsilon^{(e)}\right]\right\|_2\left\|\widehat{\mathbb{E}}_{m^{(e)}}\left[\boldsymbol{x}_{S^* \setminus S}^{(e)}\left(z^{(e)}-\eta^{(e)}\right)\right]-\mathbb{E}\left[\boldsymbol{x}_{S^* \setminus S}^{(e)}\left(z^{(e)}-\eta^{(e)}\right]\right)\right\|_2
\end{align*}

Next, we are going to study the following two events, denote $\delta_1=2\left(C^{\prime}\right)^2 \kappa_U \sigma_x^2 \sigma_{\varepsilon}^2\left(\frac{u+\log(2s^*)}{N^{(e)}}\right)$ and $\delta_2 = 2\left(C^{\prime}\right)^2 \kappa_U \sigma_x^2 \sigma_{z}^2\left(\frac{u+\log(2s^*)}{m^{(e)}}\right)$, we have
\begin{align*}
	\mathcal{K}_{1, u}(e) & = \left\{\forall \boldsymbol{\beta}\in\mathbb{R}^p, \quad \left\|\widehat{\mathbb{E}}_{N^{(e)}}\left[\boldsymbol{x}_{S^* \setminus S}^{(e)} \varepsilon^{(e)}\right]\right\|_2 \leq \sqrt{ \delta_1|T|}\right\}                                                                                                          \\
	\mathcal{K}_{2, u}(e) & =\left\{\forall \boldsymbol{\beta}\in\mathbb{R}^p, \quad \left\|\widehat{\mathbb{E}}_{m^{(e)}}\left[\boldsymbol{x}_{S^* \setminus S}^{(e)}\left(z^{(e)}-\eta^{(e)}\right)\right]-\mathbb{E}\left[\boldsymbol{x}_{S^* \setminus S}^{(e)}\left(z^{(e)}-\eta^{(e)}\right]\right)\right\|_2 \leq \sqrt{\delta_2|T|}\right\},
\end{align*}
and we claim that each of these two events occurs with probability at least $1-e^{-u}$, where $u>0$ satisfies $u+\log \left(2 s^*\right) \leq N_{\min }$ and $u+\log \left(2 s^*\right) \leq m_{\min }$, respectively.

Given the above claims are true, under the event $\mathcal{K}_u=\bigcap_{e \in \mathcal{E}}\left\{\mathcal{K}_{1, u}(e) \cap \mathcal{K}_{2, u}(e)\right\}$, with probability
\begin{align*}
	\mathbb{P}\left[\bigcap_{e \in \mathcal{E}}\left\{\mathcal{K}_{1, u}(e) \cap \mathcal{K}_{2, u}(e)\right\}\right] & =  1- \mathbb{P}\left[\bigcup_{e \in \mathcal{E}}\left\{\mathcal{K}^c_{1, u}(e) \cup \mathcal{K}^c_{2, u}(e)\right\}\right]       \\
	                                                                                                                  & \ge 1- \sum_{e\in\mathcal{E}} \mathbb{P}[ \mathcal{K}^c_{1, u}(e)] - \sum_{e\in\mathcal{E}}  \mathbb{P}[ \mathcal{K}^c_{2, u}(e)] \\
	                                                                                                                  & \ge 1-2|\mathcal{E}| e^{-u},
\end{align*}
we obtain
\begin{align*}
	\forall \boldsymbol{\beta} \in \mathbb{R}^p, \quad \sum_{e \in \mathcal{E}} \omega^{(e)} T_{2,6}^{(e)}(\boldsymbol{\beta}) & \le  \sum_{e \in \mathcal{E}} \omega^{(e)} \widehat{\tau}^{(e)} \sqrt{\delta_1|T|} \sqrt{\delta_2|T|}                                                                                                                                                                                                                       \\
	                                                                                                                           & =\sum_{e \in \mathcal{E}} \omega^{(e)} \widehat{\tau}^{(e)} \left\{2\left(C^{\prime}\right)^2 \kappa_U \sigma_x^2 \sigma_{\varepsilon}^2\left(\frac{u+\log(2s^*)}{N^{(e)}}\right)\right\}^{1/2}\left\{2\left(C^{\prime}\right)^2 \kappa_U \sigma_x^2 \sigma_{z}^2\left(\frac{u+\log(2s^*)}{m^{(e)}}\right)\right\}^{1/2}|T| \\
	                                                                                                                           & = \sum_{e \in \mathcal{E}} \omega^{(e)} \widehat{\tau}^{(e)} \left\{\sqrt{2}C^{\prime} \kappa_U^{1/2} \sigma_x \sigma_{\varepsilon}\sqrt{\frac{u+\log(2s^*)}{N^{(e)}}}\right\}\left\{\sqrt{2}C^{\prime} \kappa_U^{1/2} \sigma_x \sigma_{z}\sqrt{\frac{u+\log(2s^*)}{m^{(e)}}}\right\}|T|                                    \\
	                                                                                                                           & = \sum_{e \in \mathcal{E}} \omega^{(e)} \widehat{\tau}^{(e)} 2(C^{\prime})^2 \kappa_U \sigma_x^2 \sigma_{\varepsilon}\sigma_{z} \frac{u+\log(2s^*)}{\sqrt{N^{(e)}m^{(e)}}}|T|                                                                                                                                               \\
	                                                                                                                           & = 2(C^{\prime})^2 \kappa_U \sigma_x^2 \sigma_{\varepsilon}\sigma_{z}  \{u+\log(2s^*)\}\sum_{e \in \mathcal{E}}\frac{\omega^{(e)} \widehat{\tau}^{(e)} }{\sqrt{m^{(e)}N^{(e)}}}|T|                                                                                                                                           \\
	                                                                                                                           & = E_{2,6} \kappa_U \sigma_x^2 \sigma_{\varepsilon}\sigma_{z} \frac{u+\log(2s^*)}{\overline{\sqrt{mN}}^{\widehat{\tau}}}|T|
\end{align*}
provided that $u+\log(2s^*) \le m_{\min}$. Finally, set $u = \log(2|\mathcal{E}|)+t$, we get
$$
	\mathbb{P}\left[ \forall \boldsymbol{\beta} \in \mathbb{R}^p,\quad   \sum_{e \in \mathcal{E}} \omega^{(e)} T_{2,6}^{(e)}(\boldsymbol{\beta}) \le E_{2,6} \kappa_U \sigma_x^2 \sigma_{\varepsilon}\sigma_{z} \frac{t+ \log(4|\mathcal{E}||S^*|)}{\overline{\sqrt{mN}}^{\widehat{\tau}}}|T| \right] \ge 1-2|\mathcal{E}| e^{-t- \log(2|\mathcal{E}|)} = 1-e^{-t}.
$$
for any $t$ satisfies $ \{\log(2|\mathcal{E}|)+t\}+\log(2s^*) \le m_{\min}$. This completes the proof. It remains to prove the two events $\mathcal{K}_{1, u}(e)$ and $\mathcal{K}_{2, u}(e)$ occur with high probability.

Now fix $u>0, e\in \mathcal{E}$, let's prove $ \mathbb{P}[\mathcal{K}_{1, u}(e)]\ge 1-e^{-u}$. Note that both the L.H.S. and R.H.S. of the above inequality depends on $\boldsymbol{\beta}$, or more precisely, $S = \text{supp}(\boldsymbol{\beta})$. Observe the following decomposition
$$
	\begin{aligned}
		\mathcal{K}_{1, u}(e) & =\bigcup_{T \subset S^*}\left\{\forall \boldsymbol{\beta} \in \mathbb{R}^p, S^* \backslash \operatorname{supp}(\boldsymbol{\beta})=T, \quad \left\|\widehat{\mathbb{E}}_{N^{(e)}}\left[\boldsymbol{x}_{S^* \setminus S}^{(e)} \varepsilon^{(e)}\right]\right\|_2\leq \sqrt{\delta_1|T|} \right\} \\
		                      & =\bigcup_{T \subset S^*}\left\{\forall \boldsymbol{\beta} \in \mathbb{R}^p, S^* \backslash \operatorname{supp}(\boldsymbol{\beta})=T, \quad \sum_{j \in T} \left|\widehat{\mathbb{E}}_{N^{(e)}}\left[x_j^{(e)} \varepsilon^{(e)}\right]\right|^2 \leq \delta_1|T|\right\}                        \\
		                      & =\bigcup_{T \subset S^*} \mathcal{K}(T,e) .
	\end{aligned}
$$

At the same time, given fixed $j \in S^*$ and $e \in \mathcal{E}$, it follows from Condition \ref{cond:subg_x} and Condition \ref{cond:subg_e} that,
$$
	\mathbb{P}\left[\mathcal{K}_{1, x}(e, j)\right]=\mathbb{P}\left[\left|\widehat{\mathbb{E}}_{N^{(e)}}\left[\boldsymbol{x}_j^{(e)} \varepsilon^{(e)}\right]\right| \leq C^{\prime} \kappa_U^{1 / 2} \sigma_x \sigma_\varepsilon\left(\sqrt{\frac{x}{N^{(e)}}}+\frac{x}{N^{(e)}}\right)\right] \geq 1-2 e^{-x},
$$
for some universal constant $C^{\prime}$. Meanwhile, we can write
$$
	\mathbb{P}\left[\mathcal{K}_{1, x}^c(e, j)\right]=\mathbb{P}\left[\left|\widehat{\mathbb{E}}_{N^{(e)}}\left[\boldsymbol{x}_j^{(e)} \varepsilon^{(e)}\right]\right| \geq C^{\prime} \kappa_U^{1 / 2} \sigma_x \sigma_\varepsilon\left(\sqrt{\frac{x}{N^{(e)}}}+\frac{x}{N^{(e)}}\right)\right] \leq 2 e^{-x}.
$$

Under the event $\bigcap_{j \in S^*} \mathcal{K}_{1, u+\log(2s^*)}(e, j)$, we have
$$
	\begin{aligned}
		\sum_{j \in T} \left|\widehat{\mathbb{E}}_{N^{(e)}}\left[\boldsymbol{x}_j^{(e)} \varepsilon^{(e)}\right]\right|^2 & \leq \sum_{j \in T} \left\{C^{\prime} \kappa_U^{1 / 2} \sigma_x \sigma_\varepsilon\left(\sqrt{\frac{u+\log(2s^*)}{N^{(e)}}}+\frac{u+\log(2s^*)}{N^{(e)}}\right)\right\}^2 \\
		                                                                                                                  & \leq 2\left(C^{\prime}\right)^2 \kappa_U \sigma_x^2 \sigma_{\varepsilon}^2\left(\frac{u+\log(2s^*)}{N^{(e)}}\right) |T|                                                   \\
		                                                                                                                  & =\delta_1|T|,
	\end{aligned}
$$
provided that $u+\log(2s^*) \le N_{\min}$. Therefore, we have
$$
	\bigcap_{j \in S^*} \mathcal{K}_{1, u+\log(2s^*)}(e, j) \subset \mathcal{K}(T,e) \subset \bigcup_{T \subset S^*} \mathcal{K}(T,e)
$$
By this fact, we can continue deriving the probability of event $\mathcal{K}_{1, u}(e)$ as following:
$$
	\begin{aligned}
		\mathbb{P}\left(\mathcal{K}_{1, u}(e)\right)=\mathbb{P}\left[\bigcup_{T \subset S^*} \mathcal{K}(T,e)\right] & \geq \mathbb{P}\left[\bigcap_{j \in S^*} \mathcal{K}_{1, u+\log(2s^*)}(e, j)\right]             \\
		                                                                                                             & =1-\mathbb{P}\left[\bigcup_{j \in S^*} \mathcal{K}_{1, u+\log(2s^*)}^c(e, j)\right]             \\
		                                                                                                             & \geq 1-\sum_{j \in S^*} \mathbb{P}\left[\mathcal{K}_{1, u+\log(2s^*)}^c(e, j)\right]            \\
		                                                                                                             & \geq 1-e^{-u},\quad \forall u \in\left(0, N_{\min }-\log \left(2\left|S^*\right|\right)\right],
	\end{aligned}
$$
as desired. Next, let's prove $\mathbb{P}[\mathcal{K}_{2, u}(e)]\ge 1-e^{-u}$. Fix $u>0, e\in \mathcal{E}$, notice
\begin{align*}
	\mathcal{K}_{2, u}(e) & =\bigcup_{T \subset S^*}\left\{\forall \boldsymbol{\beta} \in \mathbb{R}^p, S^* \backslash \operatorname{supp}(\boldsymbol{\beta})=T, \quad \left\|\widehat{\mathbb{E}}_{m^{(e)}}\left[\boldsymbol{x}_{S^* \setminus S}^{(e)}(z^{(e)}-\eta^{(e)})\right]-\mathbb{E}\left[\boldsymbol{x}_{S^* \setminus S}^{(e)}(z^{(e)}-\eta^{(e)})\right]\right\|_2\leq \sqrt{\delta_2|T|} \right\} \\
	                      & =\bigcup_{T \subset S^*}\left\{\forall \boldsymbol{\beta} \in \mathbb{R}^p, S^* \backslash \operatorname{supp}(\boldsymbol{\beta})=T, \quad \sum_{j \in T} \left|\widehat{\mathbb{E}}_{m^{(e)}}\left[\boldsymbol{x}_j^{(e)} (z^{(e)}-\eta^{(e)})\right]- \mathbb{E}\left[\boldsymbol{x}_{j}^{(e)}(z^{(e)}-\eta^{(e)})\right]\right|^2 \leq \delta_2|T|\right\}                       \\
	                      & =\bigcup_{T \subset S^*} \mathcal{K}(T,e) .
\end{align*}
At the same time, given fixed $j \in S^*$ and $e \in \mathcal{E}$, since
$$
	\widehat{\mathbb{E}}_{m^{(e)}}\left[\boldsymbol{x}_j^{(e)} (z^{(e)}-\eta^{(e)})\right]-\mathbb{E}\left[\boldsymbol{x}_{j}^{(e)}(z^{(e)}-\eta^{(e)})\right] =
	\sum_{\ell=1}^{m^{(e)}}  \frac{1}{m^{(e)}} [\boldsymbol{x}_{\ell}^{(e)}]_j (z_{\ell}^{(e)}-\eta^{(e)})-\mathbb{E}\left[ \frac{1}{m^{(e)}} [\boldsymbol{x}_{\ell}^{(e)}]_j (z_{\ell}^{(e)}-\eta^{(e)})\right],
$$
where $\frac{1}{m^{(e)}} [\boldsymbol{x}_{\ell}^{(e)}]_j (z_{\ell}^{(e)}-\eta^{(e)})$ is a sub-Exponential random variable with parameter $(1/m^{(e)})\kappa_U^{1/2}\sigma_x\sigma_z$ by Conditions \ref{cond:subg_x} and \ref{cond:subg_z}, we have
\begin{align*}
	\mathbb{P}\left[\mathcal{K}_{2, x}(e, j)\right] & =\mathbb{P}\left[\left|\widehat{\mathbb{E}}_{m^{(e)}}\left[\boldsymbol{x}_j^{(e)} (z^{(e)}-\eta^{(e)})\right]-\mathbb{E}\left[\boldsymbol{x}_{j}^{(e)}(z^{(e)}-\eta^{(e)})\right]\right| \leq C^{\prime} \kappa_U^{1 / 2} \sigma_x \sigma_z\left(\sqrt{\frac{x}{m^{(e)}}}+\frac{x}{m^{(e)}}\right)\right]\geq 1-2 e^{-x}
\end{align*}
for some universal constant $C^{\prime}$. Meanwhile, we can write
\begin{align*}
	\mathbb{P}\left[\mathcal{K}^c_{2, x}(e, j)\right] & =\mathbb{P}\left[\left|\widehat{\mathbb{E}}_{m^{(e)}}\left[\boldsymbol{x}_j^{(e)} (z^{(e)}-\eta^{(e)})\right]-\mathbb{E}\left[\boldsymbol{x}_{j}^{(e)}(z^{(e)}-\eta^{(e)})\right]\right| \geq C^{\prime} \kappa_U^{1 / 2} \sigma_x \sigma_z\left(\sqrt{\frac{x}{m^{(e)}}}+\frac{x}{m^{(e)}}\right)\right]\leq 2 e^{-x}.
\end{align*}

Under the event $\bigcap_{j \in S^*} \mathcal{K}_{2, u+\log(2s^*)}(e, j)$, we have
$$
	\begin{aligned}
		\sum_{j \in T} \left|\widehat{\mathbb{E}}_{m^{(e)}}\left[\boldsymbol{x}_j^{(e)} (z^{(e)}-\eta^{(e)})\right]-\mathbb{E}\left[\boldsymbol{x}_{j}^{(e)}(z^{(e)}-\eta^{(e)})\right]\right|^2 & \leq \sum_{j \in T} \left\{C^{\prime} \kappa_U^{1 / 2} \sigma_x \sigma_z\left(\sqrt{\frac{u+\log(2s^*)}{m^{(e)}}}+\frac{u+\log(2s^*)}{m^{(e)}}\right)\right\}^2 \\
		                                                                                                                                                                                         & \leq 2\left(C^{\prime}\right)^2 \kappa_U \sigma_x^2 \sigma_{z}^2\left(\frac{u+\log(2s^*)}{m^{(e)}}\right) |T|                                                   \\
		                                                                                                                                                                                         & =\delta_2|T|,
	\end{aligned}
$$
provided that $u+\log(2s^*) \le m_{\min}$. Therefore, we have
$$
	\bigcap_{j \in S^*} \mathcal{K}_{2, u+\log(2s^*)}(e, j) \subset \mathcal{K}(T,e) \subset \bigcup_{T \subset S^*} \mathcal{K}(T,e)
$$
By this fact, we can continue deriving the probability of event $\mathcal{K}_{2, u}(e)$ as following:
$$
	\begin{aligned}
		\mathbb{P}\left(\mathcal{K}_{2, u}(e)\right)=\mathbb{P}\left[\bigcup_{T \subset S^*} \mathcal{K}(T,e)\right] & \geq \mathbb{P}\left[\bigcap_{j \in S^*} \mathcal{K}_{1, u+\log(2s^*)}(e, j)\right]             \\
		                                                                                                             & =1-\mathbb{P}\left[\bigcup_{j \in S^*} \mathcal{K}_{2, u+\log(2s^*)}^c(e, j)\right]             \\
		                                                                                                             & \geq 1-\sum_{j \in S^*} \mathbb{P}\left[\mathcal{K}_{2, u+\log(2s^*)}^c(e, j)\right]            \\
		                                                                                                             & \geq 1-e^{-u},\quad \forall u \in\left(0, m_{\min }-\log \left(2\left|S^*\right|\right)\right],
	\end{aligned}
$$
as desired.

\noindent{\it Step 2.7}. UPPER BOUND ON $T_{2,7}^{(e)}(\boldsymbol{\beta})$. In this step, we claim that the following event
\begin{equation}
	\mathcal{U}_{7,t} = \left\{ \forall \boldsymbol{\beta}\in\mathbb{R}^p,\quad  \sum_{e \in \mathcal{E}} \omega^{(e)} \mathrm{T}_{2,7}^{(e)}(\boldsymbol{\beta}) \leq E_{2,7} \kappa_U \sigma_x^2 \sigma_{\varepsilon}\sigma_{z} \frac{t+ \log(4|\mathcal{E}||S^*|)}{\overline{\sqrt{nN}}^{\widehat{\tau}}}|S^*\backslash S|\right\}
	\label{eq:event_u7t}
\end{equation}
occurs with probability at least $1-e^{-t}$ for any $t \in (0,n_{\min }-\log \left(4|\mathcal{E}| |S^*|\right)]$.

\noindent{\it Step 2.8} UPPER BOUND ON $T_{2,8}^{(e)}(\boldsymbol{\beta})$. In this step, we claim that the following event
\begin{equation}
	\mathcal{U}_{8,t} = \left\{ \forall \boldsymbol{\beta}\in\mathbb{R}^p,\quad  \sum_{e \in \mathcal{E}} \omega^{(e)} \mathrm{T}_{2,8}^{(e)}(\boldsymbol{\beta}) \leq E_{2,8} \kappa_U \sigma_x^2 \sigma_{\varepsilon} \frac{t+ \log(4|\mathcal{E}||S^*|)}{\overline{\sqrt{mN}}^{\widehat{\tau},|\eta|}}|S^*\backslash S|\right\}
	\label{eq:event_u8t}
\end{equation}
occurs with probability at least $1-e^{-t}$ for any $t \in (0,m_{\min }-\log \left(4|\mathcal{E}| |S^*|\right)]$. Observe the following, $\forall \boldsymbol{\beta} \in \mathbb{R}^p$

\begin{align*}
	\sum_{e \in \mathcal{E}} \omega^{(e)} T_{2,8}^{(e)}(\boldsymbol{\beta}) & =\sum_{e \in \mathcal{E}} \omega^{(e)}\left\{\widehat{\mathbb{E}}_{N^{(e)}}\left[\boldsymbol{x}_{S^* \setminus S}^{(e)} \varepsilon^{(e)}\right]\right\}^{\top}\left\{\widehat{\tau}^{(e)}\left(\widehat{\mathbb{E}}_{m^{(e)}}\left[\boldsymbol{x}_{S^* \backslash S}^{(e)}\right]-\mathbb{E}\left[\boldsymbol{x}_{S^* \backslash S}^{(e)}\right]\right) \eta^{(e)}\right\} \\
	                                                                        & =\sum_{e \in \mathcal{E}} \omega^{(e)}\widehat{\tau}^{(e)}\eta^{(e)}\left\{\widehat{\mathbb{E}}_{N^{(e)}}\left[\boldsymbol{x}_{S^* \setminus S}^{(e)} \varepsilon^{(e)}\right]\right\}^{\top}\left\{\widehat{\mathbb{E}}_{m^{(e)}}\left[\boldsymbol{x}_{S^* \backslash S}^{(e)}\right]-\mathbb{E}\left[\boldsymbol{x}_{S^* \backslash S}^{(e)}\right]\right\}               \\
	                                                                        & \leq \sum_{e \in \mathcal{E}} \omega^{(e)}\widehat{\tau}^{(e)} |\eta^{(e)}| \left\|\widehat{\mathbb{E}}_{N^{(e)}}\left[\boldsymbol{x}_{S^* \setminus S}^{(e)} \varepsilon^{(e)}\right]\right\|_2\left\|\widehat{\mathbb{E}}_{m^{(e)}}\left[\boldsymbol{x}_{S^* \backslash S}^{(e)}\right]-\mathbb{E}\left[\boldsymbol{x}_{S^* \backslash S}^{(e)}\right]\right\|_2
\end{align*}

Next, we are going to study the following two events, denote $\delta_1=2\left(C^{\prime}\right)^2 \kappa_U \sigma_x^2 \sigma_{\varepsilon}^2\left(\frac{u+\log(2s^*)}{N^{(e)}}\right)$ and $\delta_2 = 2\left(C^{\prime}\right)^2 \kappa_U \sigma_x^2 \left(\frac{u+\log(2s^*)}{m^{(e)}}\right)$, we have
\begin{align*}
	\mathcal{K}_{1, u}(e) & = \left\{\forall \boldsymbol{\beta}\in\mathbb{R}^p, \quad \left\|\widehat{\mathbb{E}}_{N^{(e)}}\left[\boldsymbol{x}_{S^* \setminus S}^{(e)} \varepsilon^{(e)}\right]\right\|_2 \leq \sqrt{ \delta_1|T|}\right\}                                             \\
	\mathcal{K}_{2, u}(e) & =\left\{\forall \boldsymbol{\beta}\in\mathbb{R}^p, \quad\left\|\widehat{\mathbb{E}}_{m^{(e)}}\left[\boldsymbol{x}_{S^* \backslash S}^{(e)}\right]-\mathbb{E}\left[\boldsymbol{x}_{S^* \backslash S}^{(e)}\right]\right\|_2 \leq \sqrt{\delta_2|T|}\right\},
\end{align*}
and we claim that each of these two events occurs with probability at least $1-e^{-u}$, we claim that each of these two events occurs with probability at least $1-e^{-u}$, where $u>0$ satisfies $u+\log \left(2 s^*\right) \leq N_{\min }$ and $u+\log \left(2 s^*\right) \leq m_{\min }$, respectively.

Given the above claims are true, under the event $\mathcal{K}_u=\bigcap_{e \in \mathcal{E}}\left\{\mathcal{K}_{1, u}(e) \cap \mathcal{K}_{2, u}(e)\right\}$, with probability
\begin{align*}
	\mathbb{P}\left[\bigcap_{e \in \mathcal{E}}\left\{\mathcal{K}_{1, u}(e) \cap \mathcal{K}_{2, u}(e)\right\}\right] & =  1- \mathbb{P}\left[\bigcup_{e \in \mathcal{E}}\left\{\mathcal{K}^c_{1, u}(e) \cup \mathcal{K}^c_{2, u}(e)\right\}\right]       \\
	                                                                                                                  & \ge 1- \sum_{e\in\mathcal{E}} \mathbb{P}[ \mathcal{K}^c_{1, u}(e)] - \sum_{e\in\mathcal{E}}  \mathbb{P}[ \mathcal{K}^c_{2, u}(e)] \\
	                                                                                                                  & \ge 1-2|\mathcal{E}| e^{-u},
\end{align*}
we obtain
\begin{align*}
	\forall \boldsymbol{\beta} \in \mathbb{R}^p, \quad \sum_{e \in \mathcal{E}} \omega^{(e)} T_{2,6}^{(e)}(\boldsymbol{\beta}) & \le  \sum_{e \in \mathcal{E}} \omega^{(e)} \widehat{\tau}^{(e)}|\eta^{(e)}| \sqrt{\delta_1|T|} \sqrt{\delta_2|T|}                                                                                                                                                                                                           \\
	                                                                                                                           & =\sum_{e \in \mathcal{E}} \omega^{(e)} \widehat{\tau}^{(e)} |\eta^{(e)}|\left\{2\left(C^{\prime}\right)^2 \kappa_U \sigma_x^2 \sigma_{\varepsilon}^2\left(\frac{u+\log(2s^*)}{N^{(e)}}\right)\right\}^{1/2}\left\{2\left(C^{\prime}\right)^2 \kappa_U \sigma_x^2 \left(\frac{u+\log(2s^*)}{m^{(e)}}\right)\right\}^{1/2}|T| \\
	                                                                                                                           & = \sum_{e \in \mathcal{E}} \omega^{(e)} \widehat{\tau}^{(e)}| \eta^{(e)} |\left\{\sqrt{2}C^{\prime} \kappa_U^{1/2} \sigma_x \sigma_{\varepsilon}\sqrt{\frac{u+\log(2s^*)}{N^{(e)}}}\right\}\left\{\sqrt{2}C^{\prime} \kappa_U^{1/2} \sigma_x \sqrt{\frac{u+\log(2s^*)}{m^{(e)}}}\right\}|T|                                 \\
	                                                                                                                           & = \sum_{e \in \mathcal{E}} \omega^{(e)} \widehat{\tau}^{(e)} | \eta^{(e)}|2(C^{\prime})^2 \kappa_U \sigma_x^2 \sigma_{\varepsilon} \frac{u+\log(2s^*)}{\sqrt{N^{(e)}m^{(e)}}}|T|                                                                                                                                            \\
	                                                                                                                           & = 2(C^{\prime})^2 \kappa_U \sigma_x^2 \sigma_{\varepsilon}  \{u+\log(2s^*)\}\sum_{e \in \mathcal{E}}\frac{\omega^{(e)} \widehat{\tau}^{(e)}| \eta^{(e)}|}{\sqrt{m^{(e)}N^{(e)}}}|T|                                                                                                                                         \\
	                                                                                                                           & = E_{2,8} \kappa_U \sigma_x^2 \sigma_{\varepsilon} \frac{u+\log(2s^*)}{\overline{\sqrt{mN}}^{\widehat{\tau},|\eta|}}|T|
\end{align*}
provided that $u+\log(2s^*) \le m_{\min}$. Finally, set $u = \log(2|\mathcal{E}|)+t$, we get
$$
	\mathbb{P}\left[ \forall \boldsymbol{\beta} \in \mathbb{R}^p,\quad   \sum_{e \in \mathcal{E}} \omega^{(e)} T_{2,8}^{(e)}(\boldsymbol{\beta}) \le E_{2,8} \kappa_U \sigma_x^2 \sigma_{\varepsilon}\frac{t+ \log(4|\mathcal{E}||S^*|)}{\overline{\sqrt{mN}}^{\widehat{\tau},|\eta|}}|T| \right] \ge 1-2|\mathcal{E}| e^{-t- \log(2|\mathcal{E}|)} = 1-e^{-t}.
$$
for any $t$ satisfies $ \{\log(2|\mathcal{E}|)+t\}+\log(2s^*) \le m_{\min}$. This completes the proof. It remains to prove the two events $\mathcal{K}_{1, u}(e)$ and $\mathcal{K}_{2, u}(e)$ occur with high probability.

Now fix $u>0, e\in \mathcal{E}$, let's prove $ \mathbb{P}[\mathcal{K}_{1, u}(e)]\ge 1-e^{-u}$. Note that both the L.H.S. and R.H.S. of the above inequality depends on $\boldsymbol{\beta}$, or more precisely, $S = \text{supp}(\boldsymbol{\beta})$. Observe the following decomposition
$$
	\begin{aligned}
		\mathcal{K}_{1, u}(e) & =\bigcup_{T \subset S^*}\left\{\forall \boldsymbol{\beta} \in \mathbb{R}^p, S^* \backslash \operatorname{supp}(\boldsymbol{\beta})=T, \quad \left\|\widehat{\mathbb{E}}_{N^{(e)}}\left[\boldsymbol{x}_{S^* \setminus S}^{(e)} \varepsilon^{(e)}\right]\right\|_2\leq \sqrt{\delta_1|T|} \right\} \\
		                      & =\bigcup_{T \subset S^*}\left\{\forall \boldsymbol{\beta} \in \mathbb{R}^p, S^* \backslash \operatorname{supp}(\boldsymbol{\beta})=T, \quad \sum_{j \in T} \left|\widehat{\mathbb{E}}_{N^{(e)}}\left[\boldsymbol{x}_j^{(e)} \varepsilon^{(e)}\right]\right|^2 \leq \delta_1|T|\right\}           \\
		                      & =\bigcup_{T \subset S^*} \mathcal{K}(T,e) .
	\end{aligned}
$$

At the same time, given fixed $j \in S^*$ and $e \in \mathcal{E}$, it follows from Condition \ref{cond:subg_x} and Condition \ref{cond:subg_e} that,
$$
	\mathbb{P}\left[\mathcal{K}_{1, x}(e, j)\right]=\mathbb{P}\left[\left|\widehat{\mathbb{E}}_{N^{(e)}}\left[\boldsymbol{x}_j^{(e)} \varepsilon^{(e)}\right]\right| \leq C^{\prime} \kappa_U^{1 / 2} \sigma_x \sigma_\varepsilon\left(\sqrt{\frac{x}{N^{(e)}}}+\frac{x}{N^{(e)}}\right)\right] \geq 1-2 e^{-x},
$$
for some universal constant $C^{\prime}$. Meanwhile, we can write
$$
	\mathbb{P}\left[\mathcal{K}_{1, x}^c(e, j)\right]=\mathbb{P}\left[\left|\widehat{\mathbb{E}}_{N^{(e)}}\left[\boldsymbol{x}_j^{(e)} \varepsilon^{(e)}\right]\right| \geq C^{\prime} \kappa_U^{1 / 2} \sigma_x \sigma_\varepsilon\left(\sqrt{\frac{x}{N^{(e)}}}+\frac{x}{N^{(e)}}\right)\right] \leq 2 e^{-x}.
$$

Under the event $\bigcap_{j \in S^*} \mathcal{K}_{1, u+\log(2s^*)}(e, j)$, we have
$$
	\begin{aligned}
		\sum_{j \in T} \left|\widehat{\mathbb{E}}_{N^{(e)}}\left[\boldsymbol{x}_j^{(e)} \varepsilon^{(e)}\right]\right|^2 & \leq \sum_{j \in T} \left\{C^{\prime} \kappa_U^{1 / 2} \sigma_x \sigma_\varepsilon\left(\sqrt{\frac{u+\log(2s^*)}{N^{(e)}}}+\frac{u+\log(2s^*)}{N^{(e)}}\right)\right\}^2 \\
		                                                                                                                  & \leq 2\left(C^{\prime}\right)^2 \kappa_U \sigma_x^2 \sigma_{\varepsilon}^2\left(\frac{u+\log(2s^*)}{N^{(e)}}\right) |T|                                                   \\
		                                                                                                                  & =\delta_1|T|,
	\end{aligned}
$$
provided that $u+\log(2s^*) \le N_{\min}$. Therefore, we have
$$
	\bigcap_{j \in S^*} \mathcal{K}_{1, u+\log(2s^*)}(e, j) \subset \mathcal{K}(T,e) \subset \bigcup_{T \subset S^*} \mathcal{K}(T,e)
$$
By this fact, we can continue deriving the probability of event $\mathcal{K}_{1, u}(e)$ as following:
$$
	\begin{aligned}
		\mathbb{P}\left(\mathcal{K}_{1, u}(e)\right)=\mathbb{P}\left[\bigcup_{T \subset S^*} \mathcal{K}(T,e)\right] & \geq \mathbb{P}\left[\bigcap_{j \in S^*} \mathcal{K}_{1, u+\log(2s^*)}(e, j)\right]             \\
		                                                                                                             & =1-\mathbb{P}\left[\bigcup_{j \in S^*} \mathcal{K}_{1, u+\log(2s^*)}^c(e, j)\right]             \\
		                                                                                                             & \geq 1-\sum_{j \in S^*} \mathbb{P}\left[\mathcal{K}_{1, u+\log(2s^*)}^c(e, j)\right]            \\
		                                                                                                             & \geq 1-e^{-u},\quad \forall u \in\left(0, N_{\min }-\log \left(2\left|S^*\right|\right)\right],
	\end{aligned}
$$
as desired. Next, let's prove $\mathbb{P}[\mathcal{K}_{2, u}(e)]\ge 1-e^{-u}$. Fix $u>0, e\in \mathcal{E}$, notice
\begin{align*}
	\mathcal{K}_{2, u}(e) & =\bigcup_{T \subset S^*}\left\{\forall \boldsymbol{\beta} \in \mathbb{R}^p, S^* \backslash \operatorname{supp}(\boldsymbol{\beta})=T, \quad \left\|\widehat{\mathbb{E}}_{m^{(e)}}\left[\boldsymbol{x}_{S^* \backslash S}^{(e)}\right]-\mathbb{E}\left[\boldsymbol{x}_{S^* \backslash S}^{(e)}\right]\right\|_2\leq \sqrt{\delta_2|T|} \right\} \\
	                      & =\bigcup_{T \subset S^*}\left\{\forall \boldsymbol{\beta} \in \mathbb{R}^p, S^* \backslash \operatorname{supp}(\boldsymbol{\beta})=T, \quad \sum_{j \in T} \left|\widehat{\mathbb{E}}_{m^{(e)}}\left[\boldsymbol{x}_j^{(e)}\right]- \mathbb{E}\left[\boldsymbol{x}_{j}^{(e)}\right]\right|^2 \leq \delta_2|T|\right\}                          \\
	                      & =\bigcup_{T \subset S^*} \mathcal{K}(T,e) .
\end{align*}
At the same time, given fixed $j \in S^*$ and $e \in \mathcal{E}$, since
$$
	\widehat{\mathbb{E}}_{m^{(e)}}\left[\boldsymbol{x}_j^{(e)} \right]-\mathbb{E}\left[\boldsymbol{x}_{j}^{(e)}\right] =
	\sum_{\ell=1}^{m^{(e)}}  \frac{1}{m^{(e)}} [\boldsymbol{x}_{\ell}^{(e)}]_j -\mathbb{E}\left[ \frac{1}{m^{(e)}} [x_{\ell}^{(e)}]_j\right],
$$
where $\frac{1}{m^{(e)}} [\boldsymbol{x}_{\ell}^{(e)}]_j $ is a sub-Exponential random variable with parameter $(1/m^{(e)})\kappa_U^{1/2}\sigma_x$ by Condition \ref{cond:subg_x}, we have
\begin{align*}
	\mathbb{P}\left[\mathcal{K}_{2, x}(e, j)\right] & =\mathbb{P}\left[\left|\widehat{\mathbb{E}}_{m^{(e)}}\left[\boldsymbol{x}_j^{(e)}\right]-\mathbb{E}\left[\boldsymbol{x}_{j}^{(e)}\right]\right| \leq C^{\prime} \kappa_U^{1 / 2} \sigma_x \left(\sqrt{\frac{x}{m^{(e)}}}+\frac{x}{m^{(e)}}\right)\right]\geq 1-2 e^{-x}
\end{align*}
for some universal constant $C^{\prime}$. Meanwhile, we can write
\begin{align*}
	\mathbb{P}\left[\mathcal{K}^c_{2, x}(e, j)\right] & =\mathbb{P}\left[\left|\widehat{\mathbb{E}}_{m^{(e)}}\left[\boldsymbol{x}_j^{(e)} \right]-\mathbb{E}\left[\boldsymbol{x}_{j}^{(e)}\right]\right| \geq C^{\prime} \kappa_U^{1 / 2} \sigma_x \left(\sqrt{\frac{x}{m^{(e)}}}+\frac{x}{m^{(e)}}\right)\right]\leq 2 e^{-x}.
\end{align*}

Under the event $\bigcap_{j \in S^*} \mathcal{K}_{2, u+\log(2s^*)}(e, j)$, we have
$$
	\begin{aligned}
		\sum_{j \in T} \left|\widehat{\mathbb{E}}_{m^{(e)}}\left[\boldsymbol{x}_j^{(e)} \right]-\mathbb{E}\left[\boldsymbol{x}_{j}^{(e)}\right]\right|^2 & \leq \sum_{j \in T} \left\{C^{\prime} \kappa_U^{1 / 2} \sigma_x \left(\sqrt{\frac{u+\log(2s^*)}{m^{(e)}}}+\frac{u+\log(2s^*)}{m^{(e)}}\right)\right\}^2 \\
		                                                                                                                                                 & \leq 2\left(C^{\prime}\right)^2 \kappa_U \sigma_x^2\left(\frac{u+\log(2s^*)}{m^{(e)}}\right) |T|                                                        \\
		                                                                                                                                                 & =\delta_2|T|,
	\end{aligned}
$$
provided that $u+\log(2s^*) \le m_{\min}$. Therefore, we have
$$
	\bigcap_{j \in S^*} \mathcal{K}_{2, u+\log(2s^*)}(e, j) \subset \mathcal{K}(T,e) \subset \bigcup_{T \subset S^*} \mathcal{K}(T,e)
$$
By this fact, we can continue deriving the probability of event $\mathcal{K}_{2, u}(e)$ as following:
$$
	\begin{aligned}
		\mathbb{P}\left(\mathcal{K}_{2, u}(e)\right)=\mathbb{P}\left[\bigcup_{T \subset S^*} \mathcal{K}(T,e)\right] & \geq \mathbb{P}\left[\bigcap_{j \in S^*} \mathcal{K}_{1, u+\log(2s^*)}(e, j)\right]             \\
		                                                                                                             & =1-\mathbb{P}\left[\bigcup_{j \in S^*} \mathcal{K}_{2, u+\log(2s^*)}^c(e, j)\right]             \\
		                                                                                                             & \geq 1-\sum_{j \in S^*} \mathbb{P}\left[\mathcal{K}_{2, u+\log(2s^*)}^c(e, j)\right]            \\
		                                                                                                             & \geq 1-e^{-u},\quad \forall u \in\left(0, m_{\min }-\log \left(2\left|S^*\right|\right)\right],
	\end{aligned}
$$
as desired.

\noindent{\it Step 2.9} UPPER BOUND ON $T_{2,9}^{(e)}(\boldsymbol{\beta})$. In this step, we claim that the following event

\begin{equation}
	\mathcal{U}_{9,t} = \left\{ \forall \boldsymbol{\beta}\in\mathbb{R}^p,\quad  \sum_{e \in \mathcal{E}} \omega^{(e)} \mathrm{T}_{2,9}^{(e)}(\boldsymbol{\beta}) \leq E_{2,9} \kappa_U \sigma_x^2 \sigma_{\varepsilon} \frac{t+ \log(4|\mathcal{E}||S^*|)}{\overline{\sqrt{nN}}^{\widehat{\tau},|\eta|}}|S^*\backslash S|\right\}
	\label{eq:event_u9t}
\end{equation}
occurs with probability at least $1-e^{-t}$ for any $t \in (0,n_{\min }-\log \left(4|\mathcal{E}| |S^*|\right)]$.

\noindent{\it Step 2.10} UPPER BOUND ON $T_{2,10}^{(e)}(\boldsymbol{\beta})$. In this step, we claim that the following event

\begin{equation}
	\mathcal{U}_{10,t} = \left\{ \forall \boldsymbol{\beta}\in\mathbb{R}^p,\quad  \sum_{e \in \mathcal{E}} \omega^{(e)} \mathrm{T}_{2,10}^{(e)}(\boldsymbol{\beta}) \leq E_{2,10} \kappa_U \sigma_x^2 \sigma_{z}^2 \frac{t+ \log(4|\mathcal{E}||S^*|)}{\overline{\sqrt{nN}}}|S^*\backslash S|\right\}
	\label{eq:event_u10t}
\end{equation}
occurs with probability at least $1-e^{-t}$ for any $t \in (0,m_{\min }-\log \left(4|\mathcal{E}| |S^*|\right)]$. Observe the following, $\forall \boldsymbol{\beta} \in \mathbb{R}^p$

\begin{align*}
	 & \sum_{e \in \mathcal{E}} \omega^{(e)} T_{2,10}^{(e)}(\boldsymbol{\beta})                                                                                                                                                                                                                                              \\
	 & \quad =\sum_{e \in \mathcal{E}} \omega^{(e)}\left\{\widehat{\mathbb{E}}_{N^{(e)}}\left[\boldsymbol{x}_{S^* \backslash S}^{(e)}\left(z^{(e)}-\eta^{(e)}\right)\right]-\mathbb{E}\left[\boldsymbol{x}_{S^* \backslash S}^{(e)}\left(z^{(e)}-\eta^{(e)}\right)\right]\right\}^{\top}                                     \\
	 & \quad \quad \quad \quad \quad \quad \quad \quad\quad \quad \quad \quad \quad \left\{-\widehat{\mathbb{E}}_{n^{(e)}}\left[\boldsymbol{x}_{S^* \backslash S}^{(e)}\left(z^{(e)}-\eta^{(e)}\right)\right]+\mathbb{E}\left[\boldsymbol{x}_{S^* \backslash S}^{(e)}\left(z^{(e)}-\eta^{(e)}\right)\right]\right\}          \\
	 & \quad \leq \sum_{e \in \mathcal{E}} \omega^{(e)} \left\|\widehat{\mathbb{E}}_{N^{(e)}}\left[\boldsymbol{x}_{S^* \backslash S}^{(e)}\left(z^{(e)}-\eta^{(e)}\right)\right]-\mathbb{E}\left[\boldsymbol{x}_{S^* \backslash S}^{(e)}\left(z^{(e)}-\eta^{(e)}\right)\right]\right\|_2                                     \\
	 & \quad \quad \quad \quad \quad \quad \quad \quad\quad \quad \quad \quad \quad \times \left\|-\widehat{\mathbb{E}}_{n^{(e)}}\left[\boldsymbol{x}_{S^* \backslash S}^{(e)}\left(z^{(e)}-\eta^{(e)}\right)\right]+\mathbb{E}\left[\boldsymbol{x}_{S^* \backslash S}^{(e)}\left(z^{(e)}-\eta^{(e)}\right)\right]\right\|_2
\end{align*}

Next, we are going to study the following two events, denote $\delta_1=2\left(C^{\prime}\right)^2 \kappa_U \sigma_x^2 \sigma_{z}^2\left(\frac{u+\log(2s^*)}{N^{(e)}}\right)$ and $\delta_2 = 2\left(C^{\prime}\right)^2 \kappa_U \sigma_x^2 \sigma_{z}^2\left(\frac{u+\log(2s^*)}{n^{(e)}}\right)$, we have
\begin{align*}
	\mathcal{K}_{1, u}(e) & = \left\{\forall \boldsymbol{\beta}\in\mathbb{R}^p, \quad \left\|\widehat{\mathbb{E}}_{N^{(e)}}\left[\boldsymbol{x}_{S^* \backslash S}^{(e)}\left(z^{(e)}-\eta^{(e)}\right)\right]-\mathbb{E}\left[\boldsymbol{x}_{S^* \backslash S}^{(e)}\left(z^{(e)}-\eta^{(e)}\right)\right]\right\|_2\leq \sqrt{ \delta_1|T|}\right\} \\
	\mathcal{K}_{2, u}(e) & =\left\{\forall \boldsymbol{\beta}\in\mathbb{R}^p, \quad \left\|-\widehat{\mathbb{E}}_{n^{(e)}}\left[\boldsymbol{x}_{S^* \backslash S}^{(e)}\left(z^{(e)}-\eta^{(e)}\right)\right]+\mathbb{E}\left[\boldsymbol{x}_{S^* \backslash S}^{(e)}\left(z^{(e)}-\eta^{(e)}\right)\right]\right\| \leq \sqrt{\delta_2|T|}\right\},
\end{align*}
and we claim that each of these two events occurs with probability at least $1-e^{-u}$, where $u>0$ satisfies $u+\log \left(2 s^*\right) \leq N_{\min }$ and $u+\log \left(2 s^*\right) \leq n_{\min }$, respectively.

Given the above claims are true, under the event $\mathcal{K}_u=\bigcap_{e \in \mathcal{E}}\left\{\mathcal{K}_{1, u}(e) \cap \mathcal{K}_{2, u}(e)\right\}$, with probability
\begin{align*}
	\mathbb{P}\left[\bigcap_{e \in \mathcal{E}}\left\{\mathcal{K}_{1, u}(e) \cap \mathcal{K}_{2, u}(e)\right\}\right] & =  1- \mathbb{P}\left[\bigcup_{e \in \mathcal{E}}\left\{\mathcal{K}^c_{1, u}(e) \cup \mathcal{K}^c_{2, u}(e)\right\}\right]       \\
	                                                                                                                  & \ge 1- \sum_{e\in\mathcal{E}} \mathbb{P}[ \mathcal{K}^c_{1, u}(e)] - \sum_{e\in\mathcal{E}}  \mathbb{P}[ \mathcal{K}^c_{2, u}(e)] \\
	                                                                                                                  & \ge 1-2|\mathcal{E}| e^{-u},
\end{align*}
we obtain
\begin{align*}
	\forall \boldsymbol{\beta} \in \mathbb{R}^p, \quad \sum_{e \in \mathcal{E}} \omega^{(e)} T_{2,10}^{(e)}(\boldsymbol{\beta}) & \le  \sum_{e \in \mathcal{E}} \omega^{(e)}\sqrt{\delta_1|T|} \sqrt{\delta_2|T|}                                                                                                                                                                                                              \\
	                                                                                                                            & =\sum_{e \in \mathcal{E}} \omega^{(e)} \left\{2\left(C^{\prime}\right)^2 \kappa_U \sigma_x^2 \sigma_{z}^2\left(\frac{u+\log(2s^*)}{N^{(e)}}\right)\right\}^{1/2}\left\{2\left(C^{\prime}\right)^2 \kappa_U \sigma_x^2 \sigma_{z}^2\left(\frac{u+\log(2s^*)}{n^{(e)}}\right)\right\}^{1/2}|T| \\
	                                                                                                                            & = \sum_{e \in \mathcal{E}} \omega^{(e)}  \left\{\sqrt{2}C^{\prime} \kappa_U^{1/2} \sigma_x \sigma_{z}\sqrt{\frac{u+\log(2s^*)}{N^{(e)}}}\right\}\left\{\sqrt{2}C^{\prime} \kappa_U^{1/2} \sigma_x \sigma_{z}\sqrt{\frac{u+\log(2s^*)}{n^{(e)}}}\right\}|T|                                   \\
	                                                                                                                            & = \sum_{e \in \mathcal{E}} \omega^{(e)}  2(C^{\prime})^2 \kappa_U \sigma_x^2 \sigma_{z}^2 \frac{u+\log(2s^*)}{\sqrt{N^{(e)}n^{(e)}}}|T|                                                                                                                                                      \\
	                                                                                                                            & = 2(C^{\prime})^2 \kappa_U \sigma_x^2 \sigma_{z}^2  \{u+\log(2s^*)\}\sum_{e \in \mathcal{E}}\frac{\omega^{(e)}}{\sqrt{n^{(e)}N^{(e)}}}|T|                                                                                                                                                    \\
	                                                                                                                            & = E_{2,10} \kappa_U \sigma_x^2 \sigma_{z}^2 \frac{u+\log(2s^*)}{\overline{\sqrt{nN}}}|T|
\end{align*}
provided that $u+\log(2s^*) \le n_{\min}$. Finally, set $u = \log(2|\mathcal{E}|)+t$, we get
$$
	\mathbb{P}\left[ \forall \boldsymbol{\beta} \in \mathbb{R}^p,\quad   \sum_{e \in \mathcal{E}} \omega^{(e)} T_{2,10}^{(e)}(\boldsymbol{\beta}) \le E_{2,10} \kappa_U \sigma_x^2 \sigma_{z}^2 \frac{t+ \log(4|\mathcal{E}||S^*|)}{\overline{\sqrt{nN}}}|T| \right] \ge 1-2|\mathcal{E}| e^{-t- \log(2|\mathcal{E}|)} = 1-e^{-t}.
$$
for any $t$ satisfies $ \{\log(2|\mathcal{E}|)+t\}+\log(2s^*) \le n_{\min}$. This completes the proof. It remains to prove the two events $\mathcal{K}_{1, u}(e)$ and $\mathcal{K}_{2, u}(e)$ occur with high probability.

Now fix $u>0, e\in \mathcal{E}$, let's prove $ \mathbb{P}[\mathcal{K}_{1, u}(e)]\ge 1-e^{-u}$. Note that both the L.H.S. and R.H.S. of the above inequality depends on $\boldsymbol{\beta}$, or more precisely, $S = \text{supp}(\boldsymbol{\beta})$. Observe the following decomposition
$$
	\begin{aligned}
		\mathcal{K}_{1, u}(e) & =\bigcup_{T \subset S^*}\left\{\forall \boldsymbol{\beta} \in \mathbb{R}^p, S^* \backslash \operatorname{supp}(\boldsymbol{\beta})=T, \quad \left\|\widehat{\mathbb{E}}_{N^{(e)}}\left[\boldsymbol{x}_{S^* \backslash S}^{(e)}\left(z^{(e)}-\eta^{(e)}\right)\right]-\mathbb{E}\left[\boldsymbol{x}_{S^* \backslash S}^{(e)}\left(z^{(e)}-\eta^{(e)}\right)\right]\right\|_2\leq \sqrt{\delta_1|T|} \right\}       \\
		                      & =\bigcup_{T \subset S^*}\left\{\forall \boldsymbol{\beta} \in \mathbb{R}^p, S^* \backslash \operatorname{supp}(\boldsymbol{\beta})=T, \quad \sum_{j \in T} \left|\widehat{\mathbb{E}}_{N^{(e)}}\left[\boldsymbol{x}_{S^* \backslash S}^{(e)}\left(z^{(e)}-\eta^{(e)}\right)\right]-\mathbb{E}\left[\boldsymbol{x}_{S^* \backslash S}^{(e)}\left(z^{(e)}-\eta^{(e)}\right)\right]\right|^2 \leq \delta_1|T|\right\} \\
		                      & =\bigcup_{T \subset S^*} \mathcal{K}(T,e) .
	\end{aligned}
$$

At the same time, given fixed $j \in S^*$ and $e \in \mathcal{E}$, it follows from Conditions \ref{cond:subg_x} and \ref{cond:subg_z} that,
$$
	\mathbb{P}\left[\mathcal{K}_{1, x}(e, j)\right]=\mathbb{P}\left[\left|\widehat{\mathbb{E}}_{N^{(e)}}\left[\boldsymbol{x}_{S^* \backslash S}^{(e)}\left(z^{(e)}-\eta^{(e)}\right)\right]-\mathbb{E}\left[\boldsymbol{x}_{S^* \backslash S}^{(e)}\left(z^{(e)}-\eta^{(e)}\right)\right]\right| \leq C^{\prime} \kappa_U^{1 / 2} \sigma_x \sigma_z\left(\sqrt{\frac{x}{N^{(e)}}}+\frac{x}{N^{(e)}}\right)\right] \geq 1-2 e^{-x},
$$
for some universal constant $C^{\prime}$. Meanwhile, we can write
$$
	\mathbb{P}\left[\mathcal{K}_{1, x}^c(e, j)\right]=\mathbb{P}\left[\left|\widehat{\mathbb{E}}_{N^{(e)}}\left[\boldsymbol{x}_j^{(e)} \varepsilon^{(e)}\right]\right| \geq C^{\prime} \kappa_U^{1 / 2} \sigma_x \sigma_z\left(\sqrt{\frac{x}{N^{(e)}}}+\frac{x}{N^{(e)}}\right)\right] \leq 2 e^{-x}.
$$

Under the event $\bigcap_{j \in S^*} \mathcal{K}_{1, u+\log(2s^*)}(e, j)$, we have
$$
	\begin{aligned}
		\sum_{j \in T} \left|\widehat{\mathbb{E}}_{N^{(e)}}\left[\boldsymbol{x}_{S^* \backslash S}^{(e)}\left(z^{(e)}-\eta^{(e)}\right)\right]-\mathbb{E}\left[\boldsymbol{x}_{S^* \backslash S}^{(e)}\left(z^{(e)}-\eta^{(e)}\right)\right]\right|^2 & \leq \sum_{j \in T} \left\{C^{\prime} \kappa_U^{1 / 2} \sigma_x \sigma_z\left(\sqrt{\frac{u+\log(2s^*)}{N^{(e)}}}+\frac{u+\log(2s^*)}{N^{(e)}}\right)\right\}^2 \\
		                                                                                                                                                                                                                                              & \leq 2\left(C^{\prime}\right)^2 \kappa_U \sigma_x^2 \sigma_{z}^2\left(\frac{u+\log(2s^*)}{N^{(e)}}\right) |T|                                                   \\
		                                                                                                                                                                                                                                              & =\delta_1|T|,
	\end{aligned}
$$
provided that $u+\log(2s^*) \le N_{\min}$. Therefore, we have
$$
	\bigcap_{j \in S^*} \mathcal{K}_{1, u+\log(2s^*)}(e, j) \subset \mathcal{K}(T,e) \subset \bigcup_{T \subset S^*} \mathcal{K}(T,e)
$$
By this fact, we can continue deriving the probability of event $\mathcal{K}_{1, u}(e)$ as following:
$$
	\begin{aligned}
		\mathbb{P}\left(\mathcal{K}_{1, u}(e)\right)=\mathbb{P}\left[\bigcup_{T \subset S^*} \mathcal{K}(T,e)\right] & \geq \mathbb{P}\left[\bigcap_{j \in S^*} \mathcal{K}_{1, u+\log(2s^*)}(e, j)\right]             \\
		                                                                                                             & =1-\mathbb{P}\left[\bigcup_{j \in S^*} \mathcal{K}_{1, u+\log(2s^*)}^c(e, j)\right]             \\
		                                                                                                             & \geq 1-\sum_{j \in S^*} \mathbb{P}\left[\mathcal{K}_{1, u+\log(2s^*)}^c(e, j)\right]            \\
		                                                                                                             & \geq 1-e^{-u},\quad \forall u \in\left(0, N_{\min }-\log \left(2\left|S^*\right|\right)\right],
	\end{aligned}
$$
as desired. Next, let's prove $\mathbb{P}[\mathcal{K}_{2, u}(e)]\ge 1-e^{-u}$. Fix $u>0, e\in \mathcal{E}$, notice
\begin{align*}
	\mathcal{K}_{2, u}(e) & =\bigcup_{T \subset S^*}\left\{\forall \boldsymbol{\beta} \in \mathbb{R}^p, S^* \backslash \operatorname{supp}(\boldsymbol{\beta})=T, \quad \left\|-\widehat{\mathbb{E}}_{n^{(e)}}\left[\boldsymbol{x}_{S^* \backslash S}^{(e)}\left(z^{(e)}-\eta^{(e)}\right)\right]+\mathbb{E}\left[\boldsymbol{x}_{S^* \backslash S}^{(e)}\left(z^{(e)}-\eta^{(e)}\right)\right]\right\|_2\leq \sqrt{\delta_2|T|} \right\}       \\
	                      & =\bigcup_{T \subset S^*}\left\{\forall \boldsymbol{\beta} \in \mathbb{R}^p, S^* \backslash \operatorname{supp}(\boldsymbol{\beta})=T, \quad \sum_{j \in T} \left|-\widehat{\mathbb{E}}_{n^{(e)}}\left[\boldsymbol{x}_{S^* \backslash S}^{(e)}\left(z^{(e)}-\eta^{(e)}\right)\right]+\mathbb{E}\left[\boldsymbol{x}_{S^* \backslash S}^{(e)}\left(z^{(e)}-\eta^{(e)}\right)\right]\right|^2 \leq \delta_2|T|\right\} \\
	                      & =\bigcup_{T \subset S^*} \mathcal{K}(T,e) .
\end{align*}
At the same time, given fixed $j \in S^*$ and $e \in \mathcal{E}$, since
$$
	\widehat{\mathbb{E}}_{n^{(e)}}\left[\boldsymbol{x}_j^{(e)} (z^{(e)}-\eta^{(e)})\right]-\mathbb{E}\left[\boldsymbol{x}_{j}^{(e)}(z^{(e)}-\eta^{(e)})\right] =
	\sum_{\ell=1}^{n^{(e)}}  \frac{1}{n^{(e)}} [\boldsymbol{x}_{\ell}^{(e)}]_j (z_{\ell}^{(e)}-\eta^{(e)})-\mathbb{E}\left[ \frac{1}{n^{(e)}} [\boldsymbol{x}_{\ell}^{(e)}]_j (z_{\ell}^{(e)}-\eta^{(e)})\right],
$$
where $\frac{1}{n^{(e)}} [\boldsymbol{x}_{\ell}^{(e)}]_j (z_{\ell}^{(e)}-\eta^{(e)})$ is a sub-Exponential random variable with parameter $(1/n^{(e)})\kappa_U^{1/2}\sigma_x\sigma_z$ by Conditions \ref{cond:subg_x} and \ref{cond:subg_z}, we have
\begin{align*}
	\mathbb{P}\left[\mathcal{K}_{2, x}(e, j)\right] & =\mathbb{P}\left[\left|-\widehat{\mathbb{E}}_{n^{(e)}}\left[\boldsymbol{x}_j^{(e)} (z^{(e)}-\eta^{(e)})\right]+\mathbb{E}\left[\boldsymbol{x}_{j}^{(e)}(z^{(e)}-\eta^{(e)})\right]\right| \leq C^{\prime} \kappa_U^{1 / 2} \sigma_x \sigma_z\left(\sqrt{\frac{x}{n^{(e)}}}+\frac{x}{n^{(e)}}\right)\right]\geq 1-2 e^{-x}
\end{align*}
for some universal constant $C^{\prime}$. Meanwhile, we can write
\begin{align*}
	\mathbb{P}\left[\mathcal{K}^c_{2, x}(e, j)\right] & =\mathbb{P}\left[\left|-\widehat{\mathbb{E}}_{n^{(e)}}\left[\boldsymbol{x}_j^{(e)} (z^{(e)}-\eta^{(e)})\right]+\mathbb{E}\left[\boldsymbol{x}_{j}^{(e)}(z^{(e)}-\eta^{(e)})\right]\right| \geq C^{\prime} \kappa_U^{1 / 2} \sigma_x \sigma_z\left(\sqrt{\frac{x}{n^{(e)}}}+\frac{x}{n^{(e)}}\right)\right]\leq 2 e^{-x}.
\end{align*}

Under the event $\bigcap_{j \in S^*} \mathcal{K}_{2, u+\log(2s^*)}(e, j)$, we have
$$
	\begin{aligned}
		\sum_{j \in T} \left|-\widehat{\mathbb{E}}_{n^{(e)}}\left[\boldsymbol{x}_j^{(e)} (z^{(e)}-\eta^{(e)})\right]+\mathbb{E}\left[\boldsymbol{x}_{j}^{(e)}(z^{(e)}-\eta^{(e)})\right]\right|^2 & \leq \sum_{j \in T} \left\{C^{\prime} \kappa_U^{1 / 2} \sigma_x \sigma_z\left(\sqrt{\frac{u+\log(2s^*)}{n^{(e)}}}+\frac{u+\log(2s^*)}{n^{(e)}}\right)\right\}^2 \\
		                                                                                                                                                                                          & \leq 2\left(C^{\prime}\right)^2 \kappa_U \sigma_x^2 \sigma_{z}^2\left(\frac{u+\log(2s^*)}{n^{(e)}}\right) |T|                                                   \\
		                                                                                                                                                                                          & =\delta_2|T|,
	\end{aligned}
$$
provided that $u+\log(2s^*) \le n_{\min}$. Therefore, we have
$$
	\bigcap_{j \in S^*} \mathcal{K}_{2, u+\log(2s^*)}(e, j) \subset \mathcal{K}(T,e) \subset \bigcup_{T \subset S^*} \mathcal{K}(T,e)
$$
By this fact, we can continue deriving the probability of event $\mathcal{K}_{2, u}(e)$ as following:
$$
	\begin{aligned}
		\mathbb{P}\left(\mathcal{K}_{2, u}(e)\right)=\mathbb{P}\left[\bigcup_{T \subset S^*} \mathcal{K}(T,e)\right] & \geq \mathbb{P}\left[\bigcap_{j \in S^*} \mathcal{K}_{1, u+\log(2s^*)}(e, j)\right]             \\
		                                                                                                             & =1-\mathbb{P}\left[\bigcup_{j \in S^*} \mathcal{K}_{2, u+\log(2s^*)}^c(e, j)\right]             \\
		                                                                                                             & \geq 1-\sum_{j \in S^*} \mathbb{P}\left[\mathcal{K}_{2, u+\log(2s^*)}^c(e, j)\right]            \\
		                                                                                                             & \geq 1-e^{-u},\quad \forall u \in\left(0, n_{\min }-\log \left(2\left|S^*\right|\right)\right],
	\end{aligned}
$$
as desired.

\noindent{\it Step 2.11} UPPER BOUND ON $T_{2,11}^{(e)}(\boldsymbol{\beta})$. In this step, we claim that the following event

\begin{equation}
	\mathcal{U}_{11,t} = \left\{ \forall \boldsymbol{\beta}\in\mathbb{R}^p,\quad  \sum_{e \in \mathcal{E}} \omega^{(e)} \mathrm{T}_{2,11}^{(e)}(\boldsymbol{\beta}) \leq E_{2,11} \kappa_U \sigma_x^2 \frac{t+ \log(4|\mathcal{E}||S^*|)}{\overline{\sqrt{Nn}}^{|\eta|^2}}|S^*\backslash S|\right\}
	\label{eq:event_u11t}
\end{equation}
occurs with probability at least $1-e^{-t}$ for any $t \in (0,n_{\min }-\log \left(4|\mathcal{E}| |S^*|\right)]$. Observe the following, $\forall \boldsymbol{\beta} \in \mathbb{R}^p$

\begin{align*}
	\sum_{e \in \mathcal{E}} \omega^{(e)} T_{2,11}^{(e)}(\boldsymbol{\beta}) & =\sum_{e \in \mathcal{E}} \omega^{(e)}\left\{\left(\widehat{\mathbb{E}}_{N^{(e)}}\left[\boldsymbol{x}_{S^* \backslash S}^{(e)}\right]-\mathbb{E}\left[\boldsymbol{x}_{S^* \backslash S}^{(e)}\right]\right) \eta^{(e)}\right\}^{\top}\left\{\left(-\widehat{\mathbb{E}}_{n^{(e)}}\left[\boldsymbol{x}_{S^* \backslash S}^{(e)}\right]+\mathbb{E}\left[\boldsymbol{x}_{S^* \backslash S}^{(e)}\right]\right) \eta^{(e)}\right\} \\
	                                                                         & =\sum_{e \in \mathcal{E}} \omega^{(e)}|\eta^{(e)}|^2\left\{\widehat{\mathbb{E}}_{N^{(e)}}\left[\boldsymbol{x}_{S^* \backslash S}^{(e)}\right]-\mathbb{E}\left[\boldsymbol{x}_{S^* \backslash S}^{(e)}\right]\right\}^{\top}\left\{-\widehat{\mathbb{E}}_{n^{(e)}}\left[\boldsymbol{x}_{S^* \backslash S}^{(e)}\right]+\mathbb{E}\left[\boldsymbol{x}_{S^* \backslash S}^{(e)}\right]\right\}                                   \\
	                                                                         & \leq \sum_{e \in \mathcal{E}} \omega^{(e)} |\eta^{(e)}|^2 \left\|\widehat{\mathbb{E}}_{N^{(e)}}\left[\boldsymbol{x}_{S^* \backslash S}^{(e)}\right]-\mathbb{E}\left[\boldsymbol{x}_{S^* \backslash S}^{(e)}\right]\right\|_2\left\|-\widehat{\mathbb{E}}_{n^{(e)}}\left[\boldsymbol{x}_{S^* \backslash S}^{(e)}\right]+\mathbb{E}\left[\boldsymbol{x}_{S^* \backslash S}^{(e)}\right]\right\|_2
\end{align*}

Next, we are going to study the following two events, denote $\delta_1=2\left(C^{\prime}\right)^2 \kappa_U \sigma_x^2 \left(\frac{u+\log(2s^*)}{N^{(e)}}\right)$ and $\delta_2 = 2\left(C^{\prime}\right)^2 \kappa_U \sigma_x^2 \left(\frac{u+\log(2s^*)}{n^{(e)}}\right)$, we have
\begin{align*}
	\mathcal{K}_{1, u}(e) & = \left\{\forall \boldsymbol{\beta}\in\mathbb{R}^p, \quad  \left\|\widehat{\mathbb{E}}_{N^{(e)}}\left[\boldsymbol{x}_{S^* \backslash S}^{(e)}\right]-\mathbb{E}\left[\boldsymbol{x}_{S^* \backslash S}^{(e)}\right]\right\|_2 \leq \sqrt{ \delta_1|T|}\right\} \\
	\mathcal{K}_{2, u}(e) & =\left\{\forall \boldsymbol{\beta}\in\mathbb{R}^p, \quad\left\|-\widehat{\mathbb{E}}_{n^{(e)}}\left[\boldsymbol{x}_{S^* \backslash S}^{(e)}\right]+\mathbb{E}\left[\boldsymbol{x}_{S^* \backslash S}^{(e)}\right]\right\|_2 \leq \sqrt{\delta_2|T|}\right\},
\end{align*}
and we claim that each of these two events occurs with probability at least $1-e^{-u}$, we claim that each of these two events occurs with probability at least $1-e^{-u}$, where $u>0$ satisfies $u+\log \left(2 s^*\right) \leq N_{\min }$ and $u+\log \left(2 s^*\right) \leq n_{\min }$, respectively.

Given the above claims are true, under the event $\mathcal{K}_u=\bigcap_{e \in \mathcal{E}}\left\{\mathcal{K}_{1, u}(e) \cap \mathcal{K}_{2, u}(e)\right\}$, with probability
\begin{align*}
	\mathbb{P}\left[\bigcap_{e \in \mathcal{E}}\left\{\mathcal{K}_{1, u}(e) \cap \mathcal{K}_{2, u}(e)\right\}\right] & =  1- \mathbb{P}\left[\bigcup_{e \in \mathcal{E}}\left\{\mathcal{K}^c_{1, u}(e) \cup \mathcal{K}^c_{2, u}(e)\right\}\right]       \\
	                                                                                                                  & \ge 1- \sum_{e\in\mathcal{E}} \mathbb{P}[ \mathcal{K}^c_{1, u}(e)] - \sum_{e\in\mathcal{E}}  \mathbb{P}[ \mathcal{K}^c_{2, u}(e)] \\
	                                                                                                                  & \ge 1-2|\mathcal{E}| e^{-u},
\end{align*}
we obtain
\begin{align*}
	\forall \boldsymbol{\beta} \in \mathbb{R}^p, \quad \sum_{e \in \mathcal{E}} \omega^{(e)} T_{2,11}^{(e)}(\boldsymbol{\beta}) & \le  \sum_{e \in \mathcal{E}} \omega^{(e)}|\eta^{(e)}|^2  \sqrt{\delta_1|T|} \sqrt{\delta_2|T|}                                                                                                                                                                                   \\
	                                                                                                                            & =\sum_{e \in \mathcal{E}} \omega^{(e)} |\eta^{(e)}|^2\left\{2\left(C^{\prime}\right)^2 \kappa_U \sigma_x^2\left(\frac{u+\log(2s^*)}{N^{(e)}}\right)\right\}^{1/2}\left\{2\left(C^{\prime}\right)^2 \kappa_U \sigma_x^2 \left(\frac{u+\log(2s^*)}{n^{(e)}}\right)\right\}^{1/2}|T| \\
	                                                                                                                            & = \sum_{e \in \mathcal{E}} \omega^{(e)} |\eta^{(e)}|^2\left\{\sqrt{2}C^{\prime} \kappa_U^{1/2} \sigma_x\sqrt{\frac{u+\log(2s^*)}{N^{(e)}}}\right\}\left\{\sqrt{2}C^{\prime} \kappa_U^{1/2} \sigma_x \sqrt{\frac{u+\log(2s^*)}{n^{(e)}}}\right\}|T|                                \\
	                                                                                                                            & = \sum_{e \in \mathcal{E}} \omega^{(e)} |\eta^{(e)}|^2 2(C^{\prime})^2 \kappa_U \sigma_x^2 \frac{u+\log(2s^*)}{\sqrt{N^{(e)}n^{(e)}}}|T|                                                                                                                                          \\
	                                                                                                                            & = 2(C^{\prime})^2 \kappa_U \sigma_x^2   \{u+\log(2s^*)\}\sum_{e \in \mathcal{E}}\frac{\omega^{(e)} |\eta^{(e)}|^2}{\sqrt{n^{(e)}N^{(e)}}}|T|                                                                                                                                      \\
	                                                                                                                            & = E_{2,11} \kappa_U \sigma_x^2  \frac{u+\log(2s^*)}{\overline{\sqrt{nN}}^{|\eta|^2}}|T|
\end{align*}
provided that $u+\log(2s^*) \le n_{\min}$. Finally, set $u = \log(2|\mathcal{E}|)+t$, we get
$$
	\mathbb{P}\left[ \forall \boldsymbol{\beta} \in \mathbb{R}^p,\quad   \sum_{e \in \mathcal{E}} \omega^{(e)} T_{2,11}^{(e)}(\boldsymbol{\beta}) \le E_{2,11} \kappa_U \sigma_x^2\frac{t+ \log(4|\mathcal{E}||S^*|)}{\overline{\sqrt{nN}}^{|\eta|^2}}|T| \right] \ge 1-2|\mathcal{E}| e^{-t- \log(2|\mathcal{E}|)} = 1-e^{-t}.
$$
for any $t$ satisfies $ \{\log(2|\mathcal{E}|)+t\}+\log(2s^*) \le n_{\min}$. This completes the proof. It remains to prove the two events $\mathcal{K}_{1, u}(e)$ and $\mathcal{K}_{2, u}(e)$ occur with high probability.

Now fix $u>0, e\in \mathcal{E}$, let's prove $ \mathbb{P}[\mathcal{K}_{1, u}(e)]\ge 1-e^{-u}$. Note that both the L.H.S. and R.H.S. of the above inequality depends on $\boldsymbol{\beta}$, or more precisely, $S = \text{supp}(\boldsymbol{\beta})$. Observe the following decomposition
$$
	\begin{aligned}
		\mathcal{K}_{1, u}(e) & =\bigcup_{T \subset S^*}\left\{\forall \boldsymbol{\beta} \in \mathbb{R}^p, S^* \backslash \operatorname{supp}(\boldsymbol{\beta})=T, \quad \left\|\widehat{\mathbb{E}}_{N^{(e)}}\left[\boldsymbol{x}_{S^* \backslash S}^{(e)}\right]-\mathbb{E}\left[\boldsymbol{x}_{S^* \backslash S}^{(e)}\right]\right\|_2\leq \sqrt{\delta_1|T|} \right\}       \\
		                      & =\bigcup_{T \subset S^*}\left\{\forall \boldsymbol{\beta} \in \mathbb{R}^p, S^* \backslash \operatorname{supp}(\boldsymbol{\beta})=T, \quad \sum_{j \in T} \left|\widehat{\mathbb{E}}_{N^{(e)}}\left[\boldsymbol{x}_{S^* \backslash S}^{(e)}\right]-\mathbb{E}\left[\boldsymbol{x}_{S^* \backslash S}^{(e)}\right]\right|^2 \leq \delta_1|T|\right\} \\
		                      & =\bigcup_{T \subset S^*} \mathcal{K}(T,e) .
	\end{aligned}
$$

At the same time, given fixed $j \in S^*$ and $e \in \mathcal{E}$, it follows from Condition \ref{cond:subg_x} $$
	\mathbb{P}\left[\mathcal{K}_{1, x}(e, j)\right]=\mathbb{P}\left[\left|\widehat{\mathbb{E}}_{N^{(e)}}\left[\boldsymbol{x}_{S^* \backslash S}^{(e)}\right]-\mathbb{E}\left[\boldsymbol{x}_{S^* \backslash S}^{(e)}\right]\right| \leq C^{\prime} \kappa_U^{1 / 2} \sigma_x \left(\sqrt{\frac{x}{N^{(e)}}}+\frac{x}{N^{(e)}}\right)\right] \geq 1-2 e^{-x},
$$
for some universal constant $C^{\prime}$. Meanwhile, we can write
$$
	\mathbb{P}\left[\mathcal{K}_{1, x}^c(e, j)\right]=\mathbb{P}\left[\left|\widehat{\mathbb{E}}_{N^{(e)}}\left[\boldsymbol{x}_{S^* \backslash S}^{(e)}\right]-\mathbb{E}\left[\boldsymbol{x}_{S^* \backslash S}^{(e)}\right]\right| \geq C^{\prime} \kappa_U^{1 / 2} \sigma_x\left(\sqrt{\frac{x}{N^{(e)}}}+\frac{x}{N^{(e)}}\right)\right] \leq 2 e^{-x}.
$$

Under the event $\bigcap_{j \in S^*} \mathcal{K}_{1, u+\log(2s^*)}(e, j)$, we have
$$
	\begin{aligned}
		\sum_{j \in T}\left|\widehat{\mathbb{E}}_{N^{(e)}}\left[\boldsymbol{x}_{S^* \backslash S}^{(e)}\right]-\mathbb{E}\left[\boldsymbol{x}_{S^* \backslash S}^{(e)}\right]\right|^2 & \leq \sum_{j \in T} \left\{C^{\prime} \kappa_U^{1 / 2} \sigma_x\left(\sqrt{\frac{u+\log(2s^*)}{N^{(e)}}}+\frac{u+\log(2s^*)}{N^{(e)}}\right)\right\}^2 \\
		                                                                                                                                                                               & \leq 2\left(C^{\prime}\right)^2 \kappa_U \sigma_x^2 \left(\frac{u+\log(2s^*)}{N^{(e)}}\right) |T|                                                      \\
		                                                                                                                                                                               & =\delta_1|T|,
	\end{aligned}
$$
provided that $u+\log(2s^*) \le N_{\min}$. Therefore, we have
$$
	\bigcap_{j \in S^*} \mathcal{K}_{1, u+\log(2s^*)}(e, j) \subset \mathcal{K}(T,e) \subset \bigcup_{T \subset S^*} \mathcal{K}(T,e)
$$
By this fact, we can continue deriving the probability of event $\mathcal{K}_{1, u}(e)$ as following:
$$
	\begin{aligned}
		\mathbb{P}\left(\mathcal{K}_{1, u}(e)\right)=\mathbb{P}\left[\bigcup_{T \subset S^*} \mathcal{K}(T,e)\right] & \geq \mathbb{P}\left[\bigcap_{j \in S^*} \mathcal{K}_{1, u+\log(2s^*)}(e, j)\right]             \\
		                                                                                                             & =1-\mathbb{P}\left[\bigcup_{j \in S^*} \mathcal{K}_{1, u+\log(2s^*)}^c(e, j)\right]             \\
		                                                                                                             & \geq 1-\sum_{j \in S^*} \mathbb{P}\left[\mathcal{K}_{1, u+\log(2s^*)}^c(e, j)\right]            \\
		                                                                                                             & \geq 1-e^{-u},\quad \forall u \in\left(0, N_{\min }-\log \left(2\left|S^*\right|\right)\right],
	\end{aligned}
$$
as desired. Next, let's prove $\mathbb{P}[\mathcal{K}_{2, u}(e)]\ge 1-e^{-u}$. Fix $u>0, e\in \mathcal{E}$, notice
\begin{align*}
	\mathcal{K}_{2, u}(e) & =\bigcup_{T \subset S^*}\left\{\forall \boldsymbol{\beta} \in \mathbb{R}^p, S^* \backslash \operatorname{supp}(\boldsymbol{\beta})=T, \quad \left\|-\widehat{\mathbb{E}}_{n^{(e)}}\left[\boldsymbol{x}_{S^* \backslash S}^{(e)}\right]+\mathbb{E}\left[\boldsymbol{x}_{S^* \backslash S}^{(e)}\right] \right\|_2\leq \sqrt{\delta_2|T|} \right\}      \\
	                      & =\bigcup_{T \subset S^*}\left\{\forall \boldsymbol{\beta} \in \mathbb{R}^p, S^* \backslash \operatorname{supp}(\boldsymbol{\beta})=T, \quad \sum_{j \in T} \left|-\widehat{\mathbb{E}}_{n^{(e)}}\left[\boldsymbol{x}_{S^* \backslash S}^{(e)}\right]+\mathbb{E}\left[\boldsymbol{x}_{S^* \backslash S}^{(e)}\right]\right|^2 \leq \delta_2|T|\right\} \\
	                      & =\bigcup_{T \subset S^*} \mathcal{K}(T,e) .
\end{align*}
At the same time, given fixed $j \in S^*$ and $e \in \mathcal{E}$, since
$$
	\widehat{\mathbb{E}}_{n^{(e)}}\left[\boldsymbol{x}_j^{(e)} \right]-\mathbb{E}\left[\boldsymbol{x}_{j}^{(e)}\right] =
	\sum_{\ell=1}^{n^{(e)}}  \frac{1}{n^{(e)}} [\boldsymbol{x}_{\ell}^{(e)}]_j -\mathbb{E}\left[ \frac{1}{n^{(e)}} [x_{\ell}^{(e)}]_j\right],
$$
where $\frac{1}{n^{(e)}} [\boldsymbol{x}_{\ell}^{(e)}]_j $ is a sub-Exponential random variable with parameter $(1/n^{(e)})\kappa_U^{1/2}\sigma_x$ by Condition \ref{cond:subg_x}, we have
\begin{align*}
	\mathbb{P}\left[\mathcal{K}_{2, x}(e, j)\right] & =\mathbb{P}\left[\left|-\widehat{\mathbb{E}}_{n^{(e)}}\left[\boldsymbol{x}_j^{(e)}\right]+\mathbb{E}\left[\boldsymbol{x}_{j}^{(e)}\right]\right| \leq C^{\prime} \kappa_U^{1 / 2} \sigma_x \left(\sqrt{\frac{x}{n^{(e)}}}+\frac{x}{n^{(e)}}\right)\right]\geq 1-2 e^{-x}
\end{align*}
for some universal constant $C^{\prime}$. Meanwhile, we can write
\begin{align*}
	\mathbb{P}\left[\mathcal{K}^c_{2, x}(e, j)\right] & =\mathbb{P}\left[\left|\widehat{\mathbb{E}}_{n^{(e)}}\left[\boldsymbol{x}_j^{(e)} \right]-\mathbb{E}\left[\boldsymbol{x}_{j}^{(e)}\right]\right| \geq C^{\prime} \kappa_U^{1 / 2} \sigma_x \left(\sqrt{\frac{x}{n^{(e)}}}+\frac{x}{n^{(e)}}\right)\right]\leq 2 e^{-x}.
\end{align*}

Under the event $\bigcap_{j \in S^*} \mathcal{K}_{2, u+\log(2s^*)}(e, j)$, we have
$$
	\begin{aligned}
		\sum_{j \in T} \left|-\widehat{\mathbb{E}}_{n^{(e)}}\left[\boldsymbol{x}_j^{(e)} \right]+\mathbb{E}\left[\boldsymbol{x}_{j}^{(e)}\right]\right|^2 & \leq \sum_{j \in T} \left\{C^{\prime} \kappa_U^{1 / 2} \sigma_x \left(\sqrt{\frac{u+\log(2s^*)}{n^{(e)}}}+\frac{u+\log(2s^*)}{n^{(e)}}\right)\right\}^2 \\
		                                                                                                                                                  & \leq 2\left(C^{\prime}\right)^2 \kappa_U \sigma_x^2\left(\frac{u+\log(2s^*)}{n^{(e)}}\right) |T|                                                        \\
		                                                                                                                                                  & =\delta_2|T|,
	\end{aligned}
$$
provided that $u+\log(2s^*) \le n_{\min}$. Therefore, we have
$$
	\bigcap_{j \in S^*} \mathcal{K}_{2, u+\log(2s^*)}(e, j) \subset \mathcal{K}(T,e) \subset \bigcup_{T \subset S^*} \mathcal{K}(T,e)
$$
By this fact, we can continue deriving the probability of event $\mathcal{K}_{2, u}(e)$ as following:
$$
	\begin{aligned}
		\mathbb{P}\left(\mathcal{K}_{2, u}(e)\right)=\mathbb{P}\left[\bigcup_{T \subset S^*} \mathcal{K}(T,e)\right] & \geq \mathbb{P}\left[\bigcap_{j \in S^*} \mathcal{K}_{1, u+\log(2s^*)}(e, j)\right]             \\
		                                                                                                             & =1-\mathbb{P}\left[\bigcup_{j \in S^*} \mathcal{K}_{2, u+\log(2s^*)}^c(e, j)\right]             \\
		                                                                                                             & \geq 1-\sum_{j \in S^*} \mathbb{P}\left[\mathcal{K}_{2, u+\log(2s^*)}^c(e, j)\right]            \\
		                                                                                                             & \geq 1-e^{-u},\quad \forall u \in\left(0, n_{\min }-\log \left(2\left|S^*\right|\right)\right],
	\end{aligned}
$$
as desired.

\noindent{\it Step 2.12} UPPER BOUND ON $T_{2,12}^{(e)}(\boldsymbol{\beta})$. In this step, we claim that the following event
\begin{equation}
	\mathcal{U}_{12,t} = \left\{ \forall \boldsymbol{\beta}\in\mathbb{R}^p,\quad  \sum_{e \in \mathcal{E}} \omega^{(e)} \mathrm{T}_{2,12}^{(e)}(\boldsymbol{\beta}) \leq E_{2,12} \kappa_U \sigma_x^2 \sigma_{z} \frac{t+ \log(4|\mathcal{E}||S^*|)}{\bar{m}^{\widehat{\tau}^2, |\eta|}}|S^*\backslash S|\right\}
	\label{eq:event_u12t}
\end{equation}
occurs with probability at least $1-e^{-t}$ for any $t \in (0,m_{\min }-\log \left(4|\mathcal{E}| |S^*|\right)]$. Observe the following, $\forall \boldsymbol{\beta} \in \mathbb{R}^p$

\begin{align*}
	 & \sum_{e \in \mathcal{E}} \omega^{(e)} T_{2,12}^{(e)}(\boldsymbol{\beta})                                                                                                                                                                                                                                                                                                                                                                                                                \\
	 & \quad =\sum_{e \in \mathcal{E}} \omega^{(e)}(\widehat{\tau}^{(e)})^2\eta^{(e)} \left\{\widehat{\mathbb{E}}_{m^{(e)}}\left[\boldsymbol{x}_{S^* \backslash S}^{(e)}\left(z^{(e)}-\eta^{(e)}\right)\right]-\mathbb{E}\left[\boldsymbol{x}_{S^* \backslash S}^{(e)}\left(z^{(e)}-\eta^{(e)}\right)\right]\right\}^{\top}\left\{\widehat{\mathbb{E}}_{m^{(e)}}\left[\boldsymbol{x}_{S^* \backslash S}^{(e)}\right]-\mathbb{E}\left[\boldsymbol{x}_{S^* \backslash S}^{(e)}\right] \right\}   \\
	 & \quad \leq \sum_{e \in \mathcal{E}}\omega^{(e)}(\widehat{\tau}^{(e)})^2 |\eta^{(e)}| \left\|\widehat{\mathbb{E}}_{m^{(e)}}\left[\boldsymbol{x}_{S^* \backslash S}^{(e)}\left(z^{(e)}-\eta^{(e)}\right)\right]-\mathbb{E}\left[\boldsymbol{x}_{S^* \backslash S}^{(e)}\left(z^{(e)}-\eta^{(e)}\right)\right]\right\|_2\left\|\widehat{\mathbb{E}}_{m^{(e)}}\left[\boldsymbol{x}_{S^* \backslash S}^{(e)}\right]-\mathbb{E}\left[\boldsymbol{x}_{S^* \backslash S}^{(e)}\right]\right\|_2
\end{align*}

Next, we are going to study the following two events, denote $\delta_1=2\left(C^{\prime}\right)^2 \kappa_U \sigma_x^2 \sigma_{z}^2\left(\frac{u+\log(2s^*)}{m^{(e)}}\right)$ and $\delta_2 = 2\left(C^{\prime}\right)^2 \kappa_U \sigma_x^2 \left(\frac{u+\log(2s^*)}{m^{(e)}}\right)$, we have
\begin{align*}
	\mathcal{K}_{1, u}(e) & = \left\{\forall \boldsymbol{\beta}\in\mathbb{R}^p, \quad\left\|\widehat{\mathbb{E}}_{m^{(e)}}\left[\boldsymbol{x}_{S^* \backslash S}^{(e)}\left(z^{(e)}-\eta^{(e)}\right)\right]-\mathbb{E}\left[\boldsymbol{x}_{S^* \backslash S}^{(e)}\left(z^{(e)}-\eta^{(e)}\right)\right]\right\|_2 \leq \sqrt{ \delta_1|T|}\right\} \\
	\mathcal{K}_{2, u}(e) & =\left\{\forall \boldsymbol{\beta}\in\mathbb{R}^p, \quad \left\|\widehat{\mathbb{E}}_{m^{(e)}}\left[\boldsymbol{x}_{S^* \backslash S}^{(e)}\right]-\mathbb{E}\left[\boldsymbol{x}_{S^* \backslash S}^{(e)}\right]\right\|_2  \leq \sqrt{\delta_2|T|}\right\},
\end{align*}
and we claim that each of these two events occurs with probability at least $1-e^{-u}$, where $u>0$ satisfies $u+\log \left(2 s^*\right) \leq m_{\min }$.

Given the above claims are true, under the event $\mathcal{K}_u=\bigcap_{e \in \mathcal{E}}\left\{\mathcal{K}_{1, u}(e) \cap \mathcal{K}_{2, u}(e)\right\}$, with probability
\begin{align*}
	\mathbb{P}\left[\bigcap_{e \in \mathcal{E}}\left\{\mathcal{K}_{1, u}(e) \cap \mathcal{K}_{2, u}(e)\right\}\right] & =  1- \mathbb{P}\left[\bigcup_{e \in \mathcal{E}}\left\{\mathcal{K}^c_{1, u}(e) \cup \mathcal{K}^c_{2, u}(e)\right\}\right]       \\
	                                                                                                                  & \ge 1- \sum_{e\in\mathcal{E}} \mathbb{P}[ \mathcal{K}^c_{1, u}(e)] - \sum_{e\in\mathcal{E}}  \mathbb{P}[ \mathcal{K}^c_{2, u}(e)] \\
	                                                                                                                  & \ge 1-2|\mathcal{E}| e^{-u},
\end{align*}
we obtain
\begin{align*}
	\forall \boldsymbol{\beta} \in \mathbb{R}^p, & \quad \sum_{e \in \mathcal{E}} \omega^{(e)} T_{2,12}^{(e)}(\boldsymbol{\beta})                                                                                                                                                                                                                                             \\
	                                             & \quad \le  \sum_{e \in \mathcal{E}} \omega^{(e)} (\widehat{\tau}^{(e)})^2|\eta^{(e)}| \sqrt{\delta_1|T|} \sqrt{\delta_2|T|}                                                                                                                                                                                                \\
	                                             & \quad=\sum_{e \in \mathcal{E}} \omega^{(e)} (\widehat{\tau}^{(e)})^2|\eta^{(e)}| \left\{2\left(C^{\prime}\right)^2 \kappa_U \sigma_x^2 \sigma_{z}^2\left(\frac{u+\log(2s^*)}{m^{(e)}}\right)\right\}^{1/2}\left\{2\left(C^{\prime}\right)^2 \kappa_U \sigma_x^2 \left(\frac{u+\log(2s^*)}{m^{(e)}}\right)\right\}^{1/2}|T| \\
	                                             & \quad= \sum_{e \in \mathcal{E}} \omega^{(e)} (\widehat{\tau}^{(e)})^2|\eta^{(e)}|\left\{\sqrt{2}C^{\prime} \kappa_U^{1/2} \sigma_x\sigma_z\sqrt{\frac{u+\log(2s^*)}{m^{(e)}}}\right\}\left\{\sqrt{2}C^{\prime} \kappa_U^{1/2} \sigma_x \sqrt{\frac{u+\log(2s^*)}{m^{(e)}}}\right\}|T|                                      \\
	                                             & \quad= \sum_{e \in \mathcal{E}} \omega^{(e)} (\widehat{\tau}^{(e)})^2|\eta^{(e)}| 2(C^{\prime})^2 \kappa_U \sigma_x^2\sigma_{z} \frac{u+\log(2s^*)}{m^{(e)}}|T|                                                                                                                                                            \\
	                                             & \quad= 2(C^{\prime})^2 \kappa_U \sigma_x^2\sigma_{z}  \{u+\log(2s^*)\}\sum_{e \in \mathcal{E}}\frac{\omega^{(e)}( \widehat{\tau}^{(e)})^2|\eta^{(e)}| }{m^{(e)}}|T|                                                                                                                                                        \\
	                                             & \quad= E_{2,12} \kappa_U \sigma_x^2 \sigma_{z} \frac{u+\log(2s^*)}{\bar{m}^{\widehat{\tau}^2,|\eta|}}|T|
\end{align*}
provided that $u+\log(2s^*) \le m_{\min}$. Finally, set $u = \log(2|\mathcal{E}|)+t$, we get
$$
	\mathbb{P}\left[ \forall \boldsymbol{\beta} \in \mathbb{R}^p,\quad   \sum_{e \in \mathcal{E}} \omega^{(e)} T_{2,12}^{(e)}(\boldsymbol{\beta}) \le E_{2,12} \kappa_U \sigma_x^2 \sigma_{z} \frac{t+ \log(4|\mathcal{E}||S^*|)}{\bar{m}^{\widehat{\tau}^2,|\eta|}}|T| \right] \ge 1-2|\mathcal{E}| e^{-t- \log(2|\mathcal{E}|)} = 1-e^{-t}.
$$
for any $t$ satisfies $ \{\log(2|\mathcal{E}|)+t\}+\log(2s^*) \le m_{\min}$. This completes the proof. It remains to prove the two events $\mathcal{K}_{1, u}(e)$ and $\mathcal{K}_{2, u}(e)$ occur with high probability.

Now fix $u>0, e\in \mathcal{E}$, let's prove $ \mathbb{P}[\mathcal{K}_{1, u}(e)]\ge 1-e^{-u}$. Note that both the L.H.S. and R.H.S. of the above inequality depends on $\boldsymbol{\beta}$, or more precisely, $S = \text{supp}(\boldsymbol{\beta})$. Observe the following decomposition
$$
	\begin{aligned}
		\mathcal{K}_{1, u}(e) & =\bigcup_{T \subset S^*}\left\{\forall \boldsymbol{\beta} \in \mathbb{R}^p, S^* \backslash \operatorname{supp}(\boldsymbol{\beta})=T, \quad \left\|\left(\widehat{\mathbb{E}}_{m^{(e)}}\left[\boldsymbol{x}_{S^* \backslash S}^{(e)}\left(z^{(e)}-\eta^{(e)}\right)\right]-\mathbb{E}\left[\boldsymbol{x}_{S^* \backslash S}^{(e)}\left(z^{(e)}-\eta^{(e)}\right)\right]\right)\right\|_2\leq \sqrt{\delta_1|T|} \right\} \\
		                      & =\bigcup_{T \subset S^*}\left\{\forall \boldsymbol{\beta} \in \mathbb{R}^p, S^* \backslash \operatorname{supp}(\boldsymbol{\beta})=T, \quad \sum_{j \in T} \left|\widehat{\mathbb{E}}_{m^{(e)}}\left[\boldsymbol{x}_{S^* \backslash S}^{(e)}\left(z^{(e)}-\eta^{(e)}\right)\right]-\mathbb{E}\left[\boldsymbol{x}_{S^* \backslash S}^{(e)}\left(z^{(e)}-\eta^{(e)}\right)\right]\right|^2 \leq \delta_1|T|\right\}        \\
		                      & =\bigcup_{T \subset S^*} \mathcal{K}(T,e) .
	\end{aligned}
$$

At the same time, given fixed $j \in S^*$ and $e \in \mathcal{E}$, it follows from Conditions \ref{cond:subg_x} that,
$$
	\mathbb{P}\left[\mathcal{K}_{1, x}(e, j)\right]=\mathbb{P}\left[\left|\widehat{\mathbb{E}}_{m^{(e)}}\left[\boldsymbol{x}_{S^* \backslash S}^{(e)}\left(z^{(e)}-\eta^{(e)}\right)\right]-\mathbb{E}\left[\boldsymbol{x}_{S^* \backslash S}^{(e)}\left(z^{(e)}-\eta^{(e)}\right)\right]\right| \leq C^{\prime} \kappa_U^{1 / 2} \sigma_x \sigma_z\left(\sqrt{\frac{x}{m^{(e)}}}+\frac{x}{m^{(e)}}\right)\right] \geq 1-2 e^{-x},
$$
for some universal constant $C^{\prime}$. Meanwhile, we can write
$$
	\mathbb{P}\left[\mathcal{K}^c_{1, x}(e, j)\right]=\mathbb{P}\left[\left|\widehat{\mathbb{E}}_{m^{(e)}}\left[\boldsymbol{x}_{S^* \backslash S}^{(e)}\left(z^{(e)}-\eta^{(e)}\right)\right]-\mathbb{E}\left[\boldsymbol{x}_{S^* \backslash S}^{(e)}\left(z^{(e)}-\eta^{(e)}\right)\right]\right| \geq C^{\prime} \kappa_U^{1 / 2} \sigma_x \sigma_z\left(\sqrt{\frac{x}{m^{(e)}}}+\frac{x}{m^{(e)}}\right)\right] \leq 1-2 e^{-x},
$$

Under the event $\bigcap_{j \in S^*} \mathcal{K}_{1, u+\log(2s^*)}(e, j)$, we have
$$
	\begin{aligned}
		\sum_{j \in T} \left|\widehat{\mathbb{E}}_{m^{(e)}}\left[\boldsymbol{x}_{S^* \backslash S}^{(e)}\left(z^{(e)}-\eta^{(e)}\right)\right]-\mathbb{E}\left[\boldsymbol{x}_{S^* \backslash S}^{(e)}\left(z^{(e)}-\eta^{(e)}\right)\right]\right|^2 & \leq \sum_{j \in T} \left\{C^{\prime} \kappa_U^{1 / 2} \sigma_x \sigma_z\left(\sqrt{\frac{u+\log(2s^*)}{m^{(e)}}}+\frac{u+\log(2s^*)}{m^{(e)}}\right)\right\}^2 \\
		                                                                                                                                                                                                                                              & \leq 2\left(C^{\prime}\right)^2 \kappa_U \sigma_x^2 \sigma_{z}^2\left(\frac{u+\log(2s^*)}{m^{(e)}}\right) |T|                                                   \\
		                                                                                                                                                                                                                                              & =\delta_1|T|,
	\end{aligned}
$$
provided that $u+\log(2s^*) \le m_{\min}$. Therefore, we have
$$
	\bigcap_{j \in S^*} \mathcal{K}_{1, u+\log(2s^*)}(e, j) \subset \mathcal{K}(T,e) \subset \bigcup_{T \subset S^*} \mathcal{K}(T,e)
$$
By this fact, we can continue deriving the probability of event $\mathcal{K}_{1, u}(e)$ as following:
$$
	\begin{aligned}
		\mathbb{P}\left(\mathcal{K}_{1, u}(e)\right)=\mathbb{P}\left[\bigcup_{T \subset S^*} \mathcal{K}(T,e)\right] & \geq \mathbb{P}\left[\bigcap_{j \in S^*} \mathcal{K}_{1, u+\log(2s^*)}(e, j)\right]             \\
		                                                                                                             & =1-\mathbb{P}\left[\bigcup_{j \in S^*} \mathcal{K}_{1, u+\log(2s^*)}^c(e, j)\right]             \\
		                                                                                                             & \geq 1-\sum_{j \in S^*} \mathbb{P}\left[\mathcal{K}_{1, u+\log(2s^*)}^c(e, j)\right]            \\
		                                                                                                             & \geq 1-e^{-u},\quad \forall u \in\left(0, m_{\min }-\log \left(2\left|S^*\right|\right)\right],
	\end{aligned}
$$
as desired. Next, let's prove $\mathbb{P}[\mathcal{K}_{2, u}(e)]\ge 1-e^{-u}$. Fix $u>0, e\in \mathcal{E}$, notice
\begin{align*}
	\mathcal{K}_{2, u}(e) & =\bigcup_{T \subset S^*}\left\{\forall \boldsymbol{\beta} \in \mathbb{R}^p, S^* \backslash \operatorname{supp}(\boldsymbol{\beta})=T, \quad \left\|\widehat{\mathbb{E}}_{m^{(e)}}\left[\boldsymbol{x}_{S^* \backslash S}^{(e)}\right]-\mathbb{E}\left[\boldsymbol{x}_{S^* \backslash S}^{(e)}\right] \right\|_2\leq \sqrt{\delta_2|T|} \right\}       \\
	                      & =\bigcup_{T \subset S^*}\left\{\forall \boldsymbol{\beta} \in \mathbb{R}^p, S^* \backslash \operatorname{supp}(\boldsymbol{\beta})=T, \quad \sum_{j \in T} \left|\widehat{\mathbb{E}}_{m^{(e)}}\left[\boldsymbol{x}_{S^* \backslash S}^{(e)}\right]-\mathbb{E}\left[\boldsymbol{x}_{S^* \backslash S}^{(e)}\right] \right|^2 \leq \delta_2|T|\right\} \\
	                      & =\bigcup_{T \subset S^*} \mathcal{K}(T,e) .
\end{align*}
At the same time, given fixed $j \in S^*$ and $e \in \mathcal{E}$, since
$$
	\widehat{\mathbb{E}}_{m^{(e)}}\left[\boldsymbol{x}_{j}^{(e)}\right]-\mathbb{E}\left[\boldsymbol{x}_{j}^{(e)}\right]  =
	\sum_{\ell=1}^{m^{(e)}}  \frac{1}{m^{(e)}} [\boldsymbol{x}_{\ell}^{(e)}]_j -\mathbb{E}\left[ \frac{1}{m^{(e)}} [\boldsymbol{x}_{\ell}^{(e)}]_j \right],
$$
where $\frac{1}{m^{(e)}} [\boldsymbol{x}_{\ell}^{(e)}]_j $ is a sub-Exponential random variable with parameter $(1/m^{(e)})\kappa_U^{1/2}\sigma_x$ by Condition \ref{cond:subg_x}, we have
\begin{align*}
	\mathbb{P}\left[\mathcal{K}_{2, x}(e, j)\right] & =\mathbb{P}\left[\left|\widehat{\mathbb{E}}_{m^{(e)}}\left[\boldsymbol{x}_j^{(e)} \right]-\mathbb{E}\left[\boldsymbol{x}_{j}^{(e)}\right]\right| \leq C^{\prime} \kappa_U^{1 / 2} \sigma_x\left(\sqrt{\frac{x}{m^{(e)}}}+\frac{x}{m^{(e)}}\right)\right]\geq 1-2 e^{-x}
\end{align*}
for some universal constant $C^{\prime}$. Meanwhile, we can write
\begin{align*}
	\mathbb{P}\left[\mathcal{K}^c_{2, x}(e, j)\right] & =\mathbb{P}\left[\left|\widehat{\mathbb{E}}_{m^{(e)}}\left[\boldsymbol{x}_j^{(e)} \right]-\mathbb{E}\left[\boldsymbol{x}_{j}^{(e)}\right]\right| \geq C^{\prime} \kappa_U^{1 / 2} \sigma_x \left(\sqrt{\frac{x}{m^{(e)}}}+\frac{x}{m^{(e)}}\right)\right]\leq 2 e^{-x}.
\end{align*}

Under the event $\bigcap_{j \in S^*} \mathcal{K}_{2, u+\log(2s^*)}(e, j)$, we have
$$
	\begin{aligned}
		\sum_{j \in T} \left|\widehat{\mathbb{E}}_{m^{(e)}}\left[\boldsymbol{x}_j^{(e)} \right]-\mathbb{E}\left[\boldsymbol{x}_{j}^{(e)}\right]\right|^2 & \leq \sum_{j \in T} \left\{C^{\prime} \kappa_U^{1 / 2} \sigma_x\left(\sqrt{\frac{u+\log(2s^*)}{m^{(e)}}}+\frac{u+\log(2s^*)}{m^{(e)}}\right)\right\}^2 \\
		                                                                                                                                                 & \leq 2\left(C^{\prime}\right)^2 \kappa_U \sigma_x^2\left(\frac{u+\log(2s^*)}{m^{(e)}}\right) |T|                                                       \\
		                                                                                                                                                 & =\delta_2|T|,
	\end{aligned}
$$
provided that $u+\log(2s^*) \le m_{\min}$. Therefore, we have
$$
	\bigcap_{j \in S^*} \mathcal{K}_{2, u+\log(2s^*)}(e, j) \subset \mathcal{K}(T,e) \subset \bigcup_{T \subset S^*} \mathcal{K}(T,e)
$$
By this fact, we can continue deriving the probability of event $\mathcal{K}_{2, u}(e)$ as following:
$$
	\begin{aligned}
		\mathbb{P}\left(\mathcal{K}_{2, u}(e)\right)=\mathbb{P}\left[\bigcup_{T \subset S^*} \mathcal{K}(T,e)\right] & \geq \mathbb{P}\left[\bigcap_{j \in S^*} \mathcal{K}_{1, u+\log(2s^*)}(e, j)\right]             \\
		                                                                                                             & =1-\mathbb{P}\left[\bigcup_{j \in S^*} \mathcal{K}_{2, u+\log(2s^*)}^c(e, j)\right]             \\
		                                                                                                             & \geq 1-\sum_{j \in S^*} \mathbb{P}\left[\mathcal{K}_{2, u+\log(2s^*)}^c(e, j)\right]            \\
		                                                                                                             & \geq 1-e^{-u},\quad \forall u \in\left(0, m_{\min }-\log \left(2\left|S^*\right|\right)\right],
	\end{aligned}
$$
as desired.

\noindent{\it Step 2.13}. UPPER BOUND ON $T_{2,13}^{(e)}(\boldsymbol{\beta})$. We have the following event
\begin{equation}
	\mathcal{U}_{13,t} = \left\{ \forall \boldsymbol{\beta}\in\mathbb{R}^p,\quad  \sum_{e \in \mathcal{E}} \omega^{(e)} \mathrm{T}_{2,13}^{(e)}(\boldsymbol{\beta}) \leq E_{2,13} \kappa_U \sigma_x^2 \sigma_{z} \frac{t+ \log(4|\mathcal{E}||S^*|)}{\overline{\sqrt{mn}}^{\widehat{\tau}^2, |\eta|}}|S^*\backslash S|\right\}
	\label{eq:event_u13t}
\end{equation}
occurs with probability at least $1-e^{-t}$ for any $t \in (0,n_{\min }-\log \left(4|\mathcal{E}| |S^*|\right)]$.

\noindent{\it Step 2.14} UPPER BOUND ON $T_{2,14}^{(e)}(\boldsymbol{\beta})$. We have the following event
\begin{equation}
	\mathcal{U}_{14,t} = \left\{ \forall \boldsymbol{\beta}\in\mathbb{R}^p,\quad  \sum_{e \in \mathcal{E}} \omega^{(e)} \mathrm{T}_{2,14}^{(e)}(\boldsymbol{\beta}) \leq E_{2,14} \kappa_U \sigma_x^2 \sigma_{z} \frac{t+ \log(4|\mathcal{E}||S^*|)}{\overline{\sqrt{mn}}^{\widehat{\tau}^2, |\eta|}}|S^*\backslash S|\right\}
	\label{eq:event_u14t}
\end{equation}
occurs with probability at least $1-e^{-t}$ for any $t \in (0,n_{\min }-\log \left(4|\mathcal{E}| |S^*|\right)]$.

\noindent{\it Step 2.15} UPPER BOUND ON $T_{2,15}^{(e)}(\boldsymbol{\beta})$. We have the following event
\begin{equation}
	\mathcal{U}_{15,t} = \left\{ \forall \boldsymbol{\beta}\in\mathbb{R}^p,\quad  \sum_{e \in \mathcal{E}} \omega^{(e)} \mathrm{T}_{2,15}^{(e)}(\boldsymbol{\beta}) \leq E_{2,15} \kappa_U \sigma_x^2 \sigma_{z} \frac{t+ \log(4|\mathcal{E}||S^*|)}{\bar{ n}^{\widehat{\tau}^2, |\eta|}}|S^*\setminus S|\right\}
	\label{eq:event_u15t}
\end{equation}
occurs with probability at least $1-e^{-t}$ for any $t \in (0,n_{\min }-\log \left(4|\mathcal{E}| |S^*|\right)]$.

\noindent{\it Step 3.1} UPPER BOUND ON $T_{3,1}^{(e)}(\boldsymbol{\beta})$. In this step, we claim that the following event
\begin{align}
	W_{1, t} & =\left\{\forall \boldsymbol{\beta} \in \mathbb{R}^p, \quad \sum_{e \in \mathcal{E}} \omega^{(e)} \mathrm{T}_{3,1}^{(e)}(\boldsymbol{\beta}) \leq F_{3,1} \kappa_U^{3 / 2} \sigma_x^3 \sigma_{\varepsilon} \frac{\sqrt{(s^*+\log(2|\mathcal{E}|)+t)(p+\log(2|\mathcal{E}|)+t)}}{\bar{N}}\|\boldsymbol{\beta}-\boldsymbol{\beta}^*\|_2 \right.\notag \\
	         & \quad \quad\quad\quad\quad\quad\quad\quad\quad\quad\quad\quad \quad\quad\quad\quad\quad\quad\quad\quad\quad\quad\left.+ F_{3,1} \kappa_U^{3 / 2} \sigma_x^3 \sigma_{\varepsilon}  \frac{\sqrt{|S^*|+\log(2|\mathcal{E}|)+t} (p+\log(2|\mathcal{E}|)+t)}{N_{\dagger}}\|\boldsymbol{\beta}-\boldsymbol{\beta}^*\|_2\right\}
	\label{eq:event_w1t}
\end{align}
occurs with probability at least $1-e^{-t}$ if $s^*+t+ \log(2|\mathcal{E}|)\le N_{\min}$. Observe the following
\begin{align*}
	\forall \boldsymbol{\beta} \in \mathbb{R}^p, \quad \sum_{e \in \mathcal{E}} \omega^{(e)} T_{3,1}^{(e)}(\boldsymbol{\beta}) & =\sum_{e \in \mathcal{E}} \omega^{(e)}\left(\widehat{\mathbb{E}}_{N^{(e)}}\left[\boldsymbol{x}_I^{(e)} \varepsilon^{(e)}\right]\right)^{\top}\left(\widehat{\mathbb{E}}_{N^{(e)}}\left[\boldsymbol{x}_I^{(e)}\left(\boldsymbol{x}^{(e)}_{S\cup S^*}\right)^{\top}\right]-\boldsymbol{\Sigma}_{I,:}^{(e)}\right)\left(\boldsymbol{\beta}-\boldsymbol{\beta}^*\right)                            \\
	                                                                                                                           & \leq \sum_{e \in \mathcal{E}} \omega^{(e)}\left\|\widehat{\mathbb{E}}_{N^{(e)}}\left[\boldsymbol{x}_{I}^{(e)} \varepsilon^{(e)}\right]\right\|_2\left\|\widehat{\mathbb{E}}_{N^{(e)}}\left[\boldsymbol{x}_{I}^{(e)}\left(\boldsymbol{x}^{(e)}_{S\cup S^*}\right)^{\top}\right]-\boldsymbol{\Sigma}_{I,S\cup S^*}^{(e)}\right\|_2\left\|\boldsymbol{\beta}-\boldsymbol{\beta}^*\right\|_2       \\
	                                                                                                                           & \leq \sum_{e \in \mathcal{E}} \omega^{(e)}\left\|\widehat{\mathbb{E}}_{N^{(e)}}\left[\boldsymbol{x}_{S^*}^{(e)} \varepsilon^{(e)}\right]\right\|_2\left\|\widehat{\mathbb{E}}_{N^{(e)}}\left[\boldsymbol{x}_{S^*}^{(e)}\left(\boldsymbol{x}^{(e)}_{S\cup S^*}\right)^{\top}\right]-\boldsymbol{\Sigma}_{S^*,S\cup S^*}^{(e)}\right\|_2\left\|\boldsymbol{\beta}-\boldsymbol{\beta}^*\right\|_2
\end{align*}
Next, we are going to study the following two events, define
\begin{align*}
	\mathcal{K}_{1, u}(e) & =\left\{\left\|\widehat{\mathbb{E}}_{N^{(e)}}\left[x_{S^*}^{(e)} \varepsilon^{(e)}\right]\right\|_2 \leq C_1^{\prime} \kappa_U^{1 / 2} \sigma_x \sigma_{\varepsilon}\left(\sqrt{\frac{s^*+u}{N^{(e)}}}+\frac{s^*+u}{N^{(e)}}\right)\right\}                                                                                                                           \\
	\mathcal{K}_{2, u}(e) & =\left\{\forall \boldsymbol{\beta} \in \mathbb{R}^p, \quad\left\|\widehat{\mathbb{E}}_{N^{(e)}}\left[\boldsymbol{x}_{S^*}^{(e)}\left(\left[\boldsymbol{x}^{(e)}\right]_{S\cup S^*}\right)^{\top}\right]-\boldsymbol{\Sigma}_{S^*,S\cup S^*}^{(e)}\right\|_2 \leq C_2^{\prime} \kappa_U \sigma_x^2\left(\sqrt{\frac{u+p}{N^{(e)}}}+\frac{u+p}{N^{(e)}}\right)\right\},
\end{align*}
and we claim that each of these two events occurs with probability at least $1-e^{-u}$, for any fixed $u>0$ and $e\in\mathcal{E}$.

Given the above claims are true, under the event $\mathcal{K}_u=\bigcap_{e \in \mathcal{E}}\left\{\mathcal{K}_{1, u}(e) \cap \mathcal{K}_{2, u}(e)\right\}$, with probability
\begin{align*}
	\mathbb{P}\left[\bigcap_{e \in \mathcal{E}}\left\{\mathcal{K}_{1, u}(e) \cap \mathcal{K}_{2, u}(e)\right\}\right] & =  1- \mathbb{P}\left[\bigcup_{e \in \mathcal{E}}\left\{\mathcal{K}^c_{1, u}(e) \cup \mathcal{K}^c_{2, u}(e)\right\}\right]       \\
	                                                                                                                  & \ge 1- \sum_{e\in\mathcal{E}} \mathbb{P}[ \mathcal{K}^c_{1, u}(e)] - \sum_{e\in\mathcal{E}}  \mathbb{P}[ \mathcal{K}^c_{2, u}(e)] \\
	                                                                                                                  & \ge 1-2|\mathcal{E}| e^{-u},
\end{align*}
we obtain
\begin{align*}
	\forall \boldsymbol{\beta} \in \mathbb{R}^p, \quad \sum_{e \in \mathcal{E}} \omega^{(e)} T_{3,1}^{(e)}(\boldsymbol{\beta}) & \le  \sum_{e \in \mathcal{E}} \omega^{(e)} C_1^{\prime} \kappa_U^{1 / 2} \sigma_x \sigma_{\varepsilon}\left(\sqrt{\frac{s^*+u}{N^{(e)}}}+\frac{s^*+u}{N^{(e)}}\right)C_2^{\prime} \kappa_U \sigma_x^2\left(\sqrt{\frac{u+p}{N^{(e)}}}+\frac{u+p}{N^{(e)}}\right)\|\boldsymbol{\beta}-\boldsymbol{\beta}^*\|_2 \\
	                                                                                                                           & \le  F_{3,1} \kappa_U^{3 / 2} \sigma_x^3 \sigma_{\varepsilon}  \sum_{e \in \mathcal{E}} \omega^{(e)} \sqrt{\frac{s^*+u}{N^{(e)}}}\left(\sqrt{\frac{u+p}{N^{(e)}}}+\frac{u+p}{N^{(e)}}\right)\|\boldsymbol{\beta}-\boldsymbol{\beta}^*\|_2                                                                     \\
	                                                                                                                           & \le  F_{3,1} \kappa_U^{3 / 2} \sigma_x^3 \sigma_{\varepsilon}  \sum_{e \in \mathcal{E}} \omega^{(e)} \frac{\sqrt{(s^*+u)(u+p)}}{N^{(e)}}\|\boldsymbol{\beta}-\boldsymbol{\beta}^*\|_2                                                                                                                         \\
	                                                                                                                           & \quad+  F_{3,1} \kappa_U^{3 / 2} \sigma_x^3 \sigma_{\varepsilon}  \sum_{e \in \mathcal{E}} \omega^{(e)}\frac{\sqrt{s^*+u} (u+p)}{(N^{(e)})^{2/3}}\|\boldsymbol{\beta}-\boldsymbol{\beta}^*\|_2                                                                                                                \\
	                                                                                                                           & = F_{3,1} \kappa_U^{3 / 2} \sigma_x^3 \sigma_{\varepsilon} \frac{\sqrt{(s^*+u)(u+p)}}{\bar{N}}\|\boldsymbol{\beta}-\boldsymbol{\beta}^*\|_2 + F_{3,1} \kappa_U^{3 / 2} \sigma_x^3 \sigma_{\varepsilon}  \frac{\sqrt{|S^*|+u} (u+p)}{N_{\dagger}}\|\boldsymbol{\beta}-\boldsymbol{\beta}^*\|_2
\end{align*}
provided $s^*+u\le N_{\min}$. Finally, we finish the proof by setting $u = \log(2|\mathcal{E}|)+t$. It remains to prove the two events $\mathcal{K}_{1, u}(e)$ and $\mathcal{K}_{2, u}(e)$ occur with high probability.

Now fix $u>0, e\in \mathcal{E}$, let's prove $ \mathbb{P}[\mathcal{K}_{1, u}(e)]\ge 1-e^{-u}$. Notice that
\begin{align}
	\left\|\widehat{\mathbb{E}}_{N^{(e)}}\left[\boldsymbol{x}_{S^*}^{(e)} \varepsilon^{(e)}\right]\right\|_2 & =\sup _{\boldsymbol{v} \in \mathbb{R}^{s^*},\|\boldsymbol{v}\|_2=1} \boldsymbol{v}^{\top} \widehat{\mathbb{E}}_{N^{(e)}}\left[\boldsymbol{x}_{S^*}^{(e)} \varepsilon^{(e)}\right]\notag                                                                                                                                     \\
	                                                                                                         & = \sup _{\boldsymbol{v} \in \mathbb{R}^{s^*},\|\boldsymbol{v}\|_2=1} \boldsymbol{v}^{\top} \left\{ \frac{1}{N^{(e)}}\sum_{\ell=1}^{N^{(e)}}\left[\boldsymbol{x}_{\ell}^{(e)}\right]_{S^*}\varepsilon_{\ell}^{(e)} - \mathbb{E}\left[\left[\boldsymbol{x}_{\ell}^{(e)}\right]_{S^*} \varepsilon_{\ell}^{(e)}\right]\right\}.
	\label{eq:step31l2norm}
\end{align}

Let $\boldsymbol{v} _1^{(S^*)}, ..., \boldsymbol{v} _{N_{S^*}}^{(S^*)}$ be an $1/4$-covering of $\mathcal{B}(S^*)$, that is, for any $\boldsymbol{v}\in \mathcal{B}(S^*)$, there exists some $\pi(\boldsymbol{v})\in[N_{S^*}]$ such that $\left\|\boldsymbol{v}-\boldsymbol{v}_{\pi(v)}^{(S^*)}\right\|_2 \leq 1 / 4.$ It follows from standard empirical process result that $N_{S^*}\le 9^{s^*}$.

Denote $\boldsymbol{\xi} =  \frac{1}{N^{(e)}}\sum_{\ell=1}^{N^{(e)}}\left[\boldsymbol{x}_{\ell}^{(e)}\right]_{S^*} \left[ \varepsilon_{\ell}^{(e)}\right] - \mathbb{E}\left[\left[\boldsymbol{x}_{\ell}^{(e)}\right]_{S^*} \left[ \varepsilon_{\ell}^{(e)}\right]\right]$. By the observation
$$
	\sup _{\boldsymbol{v} \in \mathcal{B}\left(S^*\right)} \boldsymbol{v}^{\top} \boldsymbol{\xi}=\sup _{k \in\left[N_{S^*}\right]}\left(\boldsymbol{v}_k^{(S^*)}\right)^{\top} \boldsymbol{\xi}+\sup _{\boldsymbol{v} \in \mathcal{B}\left( S^*\right)}\left(\boldsymbol{v}-\boldsymbol{v}_{\pi(v)}^{(S^*)}\right)^{\top} \boldsymbol{\xi} \leq \sup _{k \in\left[N_{S^*}\right]}\left(\boldsymbol{v}_k^{(S^*)}\right)^{\top} \boldsymbol{\xi}+\frac{1}{4}\left\{\sup _{\boldsymbol{v} \in \mathcal{B}\left(S^*\right)} \boldsymbol{v}^{\top} \boldsymbol{\xi}\right\},
$$
which implies $\{\sup _{\boldsymbol{v} \in \mathcal{B}\left(S^*\right)} \boldsymbol{v}^{\top} \boldsymbol{\xi}\} \leq 2 \sup _{k \in\left[N_{S^*}\right]}\left(\boldsymbol{v}_k^{(S^*)}\right)^{\top} \boldsymbol{\xi}$, thus
\begin{align}
	 & \sup _{\boldsymbol{v} \in \mathbb{R}^{s^*},\|\boldsymbol{v}\|_2=1} \boldsymbol{v}^{\top} \left\{ \frac{1}{N^{(e)}}\sum_{\ell=1}^{N^{(e)}}\left[\boldsymbol{x}_{\ell}^{(e)}\right]_{S^*} \varepsilon_{\ell}^{(e)} - \mathbb{E}\left[\left[\boldsymbol{x}_{\ell}^{(e)}\right]_{S^*}  \varepsilon_{\ell}^{(e)}\right]\right\}\notag                                                \\
	 & \quad \le  2 \sup _{k \in\left[N_{S^*}\right]}\frac{1}{N^{(e)}}\sum_{\ell=1}^{N^{(e)}}\left\{\left(\boldsymbol{v}_k^{(S^*)}\right)^{\top} \left[\boldsymbol{x}_{\ell}^{(e)}\right]_{S^*}  \varepsilon_{\ell}^{(e)}- \mathbb{E}\left[\left(\boldsymbol{v}_k^{(S^*)}\right)^{\top} \left[\boldsymbol{x}_{\ell}^{(e)}\right]_{(S^*)} \varepsilon_{\ell}^{(e)}\right]\right\}\notag \\
	 & \quad \equiv 2 \sup _{k \in\left[N_{S^*}\right]} Z(k)\label{eq:step31CSinequality},
\end{align}
Notice that for given fixed $\boldsymbol{v}_k^{(S^*)}\in\mathcal{B}(S^*)$, $Z(k)$ can be written as the sum of independent zero-mean random variables with parameter $(1/N^{(e)})\kappa_U^{1 / 2} \sigma_x\sigma_{\varepsilon}$, because each summand is $(1/N^{(e)})$ times a centered product of two sub-Gaussian random variables $\left(\boldsymbol{v}_k^{(S^*)}\right)^{\top} \left[\boldsymbol{x}_{\ell}^{(e)}\right]_{(S^*)} $ and $\varepsilon_{\ell}^{(e)}$ with parameters $\kappa_U^{1 / 2} \sigma_x$ and $\sigma_{\varepsilon}$, respectively. Therefore, we have
$$
	\mathbb{P}\left[|Z( k)| \geq C^{\prime}\kappa_U^{1/2} \sigma_x \sigma_{\varepsilon}\left\{\sqrt{\frac{x}{N^{(e)}}}+\frac{x}{N^{(e)}} \right\}\right] \leq 2 e^{-x},\quad \forall x>0,
$$
which tells us that
\begin{align*}
	 & \mathbb{P}\left[ 2 \sup _{k \in\left[N_{S^*}\right]}Z(k)\ge 2C^{\prime} \kappa_U^{1/2}\sigma_x \sigma_{\varepsilon}\left\{\sqrt{\frac{x}{N^{(e)}}}+\frac{x}{N^{(e)}} \right\}\right]          \\
	 & \quad\le\sum_{k \in\left[N_{S^*}\right]} \mathbb{P}\left[  |Z(k)|\ge C^{\prime} \kappa_U^{1/2}\sigma_x \sigma_{\varepsilon}\left\{\sqrt{\frac{x}{N^{(e)}}}+ \frac{x}{N^{(e)}} \right\}\right] \\
	 & \quad\leq \sum_{k \in\left[N_{S^*}\right]}2 e^{-x}                                                                                                                                            \\
	 & = 2N_{S^*}e^{-x}
\end{align*}
With fixed $u>0$, let $x = u + \log(2N_{S^*})\le 3(u +s^*) $ and facts \ref{eq:step31l2norm}, \ref{eq:step31CSinequality}, we have
$$
	\mathbb{P}\left[ \left\|\widehat{\mathbb{E}}_{N^{(e)}}\left[\boldsymbol{x}_{S^*}^{(e)} \varepsilon^{(e)}\right]\right\|_2\le C_1^{\prime}\kappa_U^{1/2} \sigma_x \sigma_{\varepsilon}\left\{\sqrt{\frac{u+s^*}{N^{(e)}}}+\frac{u + s^*}{N^{(e)}} \right\}\right]\ge 1-e^{-u},
$$
where $C_1^{\prime} = 6 C^{\prime}$.

\bigskip
Next, let's prove $\mathbb{P}[\mathcal{K}_{2, u}(e)]\ge 1-e^{-u}$. Fix $u>0, e\in \mathcal{E}$, notice that
\begin{align}
	 & \left\|\widehat{\mathbb{E}}_{N^{(e)}}\left[\boldsymbol{x}_{S^*}^{(e)}\left(\boldsymbol{x}^{(e)}_{S\cup S^*}\right)^{\top}\right]-\boldsymbol{\Sigma}_{S^*,S\cup S^*}^{(e)}\right\|_2\notag                                                                                                                                                                                                                                                                           \\
	 & \quad =\sup _{\boldsymbol{u}\in \mathcal{B}(S^*), \boldsymbol{v} \in \mathcal{B}(S\cup S^*)} \boldsymbol{u}^{\top}\left\{\widehat{\mathbb{E}}_{N^{(e)}}\left[\boldsymbol{x}_{S^*}^{(e)}\left(\boldsymbol{x}^{(e)}_{S\cup S^*}\right)^{\top}\right]-\boldsymbol{\Sigma}_{S^*,S\cup S^*}^{(e)}\right\}\boldsymbol{v}\notag                                                                                                                                             \\
	 & \quad= \sup _{\boldsymbol{u}\in \mathcal{B}(S^*), \boldsymbol{v} \in \mathcal{B}(S\cup S^*)} \boldsymbol{u}^{\top} \left\{ \frac{1}{N^{(e)}}\sum_{\ell=1}^{N^{(e)}}\left[\boldsymbol{x}_{\ell}^{(e)}\right]_{S^*} \left(\left[\boldsymbol{x}_{\ell}^{(e)}\right]_{S\cup S^*}\right)^{\top}  - \mathbb{E}\left[\left[\boldsymbol{x}_{\ell}^{(e)}\right]_{S^*} \left(\left[\boldsymbol{x}_{\ell}^{(e)}\right]_{S\cup S^*}\right)^{\top}  \right]\right\}\boldsymbol{v}
	\label{eq:step31l2norm}.
\end{align}
For any $S\subset [p]$, let $\left\{\left(\boldsymbol{u}_k^{(e)}, \boldsymbol{v}_k^{(e)}\right)\right\}_{k=1}^{N_S} \in \mathcal{B}\left(S^*\right)\times\mathcal{B}\left(S \cup S^*\right)$ be a 1/4-covering of $\mathcal{B}\left(S^*\right)\times\mathcal{B}\left(S \cup S^*\right)$. It follows from standard empirical process theory that $N_S \leq 9^{2\left|S \cup S^*\right|}$, then $N=\sum_{S \subset[p]} N_S \leq 81^{s^*}\left(\frac{81 e p}{s}\right)^s$. Moreover, by variational representation of the matrix $\ell_2$ norm, we have
\begin{align}
	 & \sup_{|S|\le p}\left\|\widehat{\mathbb{E}}_{N^{(e)}}\left[\boldsymbol{x}_{S^*}^{(e)}\left(\boldsymbol{x}^{(e)}_{S\cup S^*}\right)^{\top}\right]-\boldsymbol{\Sigma}_{S^*,S\cup S^*}^{(e)}\right\|_2 \notag                                                                                                                                                                                                                                                                                    \\
	 & \quad \le  4\sup_{|S|\le p,k\in[N_S]}(\boldsymbol{u_k}^{(S)})^{\top} \left\{ \frac{1}{N^{(e)}}\sum_{\ell=1}^{N^{(e)}}\left[\boldsymbol{x}_{\ell}^{(e)}\right]_{S^*} \left(\left[\boldsymbol{x}_{\ell}^{(e)}\right]_{S\cup S^*}\right)^{\top}  - \mathbb{E}\left[\left[\boldsymbol{x}_{\ell}^{(e)}\right]_{S^*} \left(\left[\boldsymbol{x}_{\ell}^{(e)}\right]_{S\cup S^*}\right)^{\top}  \right]\right\}(\boldsymbol{v_k}^{(S)})\notag                                                        \\
	 & \quad \le  4\sup_{|S|\le p,k\in[N_S]} \frac{1}{N^{(e)}}\sum_{\ell=1}^{N^{(e)}}\left\{ (\boldsymbol{u_k}^{(S)})^{\top}\left[\boldsymbol{x}_{\ell}^{(e)}\right]_{S^*} \left(\left[\boldsymbol{x}_{\ell}^{(e)}\right]_{S\cup S^*}\right)^{\top}(\boldsymbol{v_k}^{(S)})  - \mathbb{E}\left[(\boldsymbol{u_k}^{(S)})^{\top}\left[\boldsymbol{x}_{\ell}^{(e)}\right]_{S^*} \left(\left[\boldsymbol{x}_{\ell}^{(e)}\right]_{S\cup S^*}\right)^{\top} (\boldsymbol{v_k}^{(S)}) \right]\right\}\notag \\
	 & \quad \equiv 4\sup_{|S|\le p,k\in[N_S]} Z(S,k)\label{eq:step31CSinequality}.
\end{align}
For fixed $k$ and $S$, $Z(S,k)$ can be written as the sum of independent zero-mean random variable with parameter $(1/N^{(e)})\kappa_U\sigma_x^2$, because each summand is $(1/N^{(e)})$ times a centered product of two sub-Gaussian random variables $(\boldsymbol{u}_k^{(S)})^{\top} [\boldsymbol{x}_{\ell}^{(e)}]_{S^*} $ and $([\boldsymbol{x}_{\ell}^{(e)}]_{S\cup S^*})^{\top} \boldsymbol{v}_k^{(S)}$ with same parameter $\kappa_U^{1/2}\sigma_x$.
Therefore, we have
$$
	\mathbb{P}\left[|Z( S,k)| \geq C^{''}\kappa_U \sigma_x^2 \left\{\sqrt{\frac{x}{N^{(e)}}}+\frac{x}{N^{(e)}} \right\}\right] \leq 2 e^{-x},\quad \forall x>0,
$$
which tells us that
\begin{align*}
	 & \mathbb{P}\left[ 4 \sup _{|S|\le p,k \in\left[N_{S}\right]}Z(S,k)\ge 4C^{''} \kappa_U\sigma_x^2 \left\{\sqrt{\frac{x}{N^{(e)}}}+\frac{x}{N^{(e)}} \right\}\right]           \\
	 & \quad\le\sum_{|S|\le p, k \in\left[N_{S^*}\right]} \mathbb{P}\left[  |Z(k)|\ge C^{''} \kappa_U\sigma_x^2 \left\{\sqrt{\frac{x}{N^{(e)}}}+ \frac{x}{N^{(e)}} \right\}\right] \\
	 & \quad\leq \sum_{|S|\le p,k \in\left[N_{S^*}\right]}2 e^{-x}                                                                                                                 \\
	 & = 2Ne^{-x}
\end{align*}
With fixed $u>0$, let $x = u + \log(2N)\le 6(u + p + s^*) \le 12(u + p)$ and facts \ref{eq:step31l2norm}, \ref{eq:step31CSinequality}, we have
$$
	\mathbb{P}\left[ \forall \boldsymbol{\beta}\in\mathbb{R}^p, \left\|\widehat{\mathbb{E}}_{N^{(e)}}\left[\boldsymbol{x}_{S^*}^{(e)}\left(\boldsymbol{x}_{S \cup S^*}^{(e)}\right)^{\top}\right]-\boldsymbol{\Sigma}_{S^*, S \cup S^*}^{(e)}\right\|_2\ge C_2^{'} \kappa_U\sigma_x^2 \left\{\sqrt{\frac{u + p}{N^{(e)}}}+\frac{u + p}{N^{(e)}} \right\}\right]\le e^{-u},
$$
where $C_2^{'} = 48C^{''}$. Therefore, we finished proving the claim about event $\mathcal{K}_{2, u}(e)$.

\noindent{\it Step 3.2} UPPER BOUND ON $T_{3,2}^{(e)}(\boldsymbol{\beta})$. In this step, we claim that the following event

\begin{align}
	W_{2, t} & =\left\{\forall \boldsymbol{\beta} \in \mathbb{R}^p, \quad \sum_{e \in \mathcal{E}} \omega^{(e)} \mathrm{T}_{3,2}^{(e)}(\boldsymbol{\beta}) \leq F_{3,2} \kappa_U^{3 / 2} \sigma_x^3 \sigma_{z} \frac{\sqrt{(s^* +\log(2|\mathcal{E}|)+t)(p+\log(2|\mathcal{E}|)+t)}}{\overline{\sqrt{mN}}^{\widehat{\tau}}} \|\boldsymbol{\beta}-\boldsymbol{\beta}^*\|_2\right.\notag \\
	         & \quad \quad\quad\quad\quad\quad\quad\quad\quad\quad\quad\quad \quad\quad\quad\quad\quad\quad\quad\quad\quad\quad\left.+   F_{3,2} \kappa_U^{3 / 2} \sigma_x^3 \sigma_{z} \frac{\sqrt{s^* +\log(2|\mathcal{E}|)+t}(p+\log(2|\mathcal{E}|)+t)}{\overline{\sqrt{m} N}^{\widehat{\tau}} } \|\boldsymbol{\beta}-\boldsymbol{\beta}^*\|_2\right\}
	\label{eq:event_w2t}
\end{align}
occurs with probability at least $1-e^{-t}$ if $s^*+t+\log (|2 \mathcal{E}|) \leq m_{\min }$. Observe the following, $\forall \boldsymbol{\beta} \in \mathbb{R}^p$
\begin{align*}
	 & \quad \sum_{e \in \mathcal{E}} \omega^{(e)} T_{3,2}^{(e)}(\boldsymbol{\beta})                                                                                                                                                                                                                                                                                                                                                                                                                                        \\
	 & \quad =\sum_{e \in \mathcal{E}} \omega^{(e)} \left\{ \widehat{\tau}^{(e)}\left( \widehat{\mathbb{E}}_{m^{(e)}}[{\boldsymbol{x}}^{(e)}_I(z^{(e)} - \eta^{(e)}  )] -   \mathbb{E}[\boldsymbol{x}^{(e)}_I(z^{(e)} - \eta^{(e)}  )] \right)\right\}^{\top} \left\{    \widehat{\mathbb{E}}_{N^{(e)}}[{\boldsymbol{x}}^{(e)}_I ({\boldsymbol{x}}^{(e)})^{\top}  (\boldsymbol{\beta} - \boldsymbol{\beta}^*)]- \boldsymbol{\Sigma}^{(e)}_{I,:} (\boldsymbol{\beta} - \boldsymbol{\beta}^*)    \right\}                     \\
	 & \quad =\sum_{e \in \mathcal{E}} \omega^{(e)} \left\{ \widehat{\tau}^{(e)}\left( \widehat{\mathbb{E}}_{m^{(e)}}[{\boldsymbol{x}}^{(e)}_I(z^{(e)} - \eta^{(e)}  )] -   \mathbb{E}[\boldsymbol{x}^{(e)}_I(z^{(e)} - \eta^{(e)}  )] \right)\right\}^{\top} \left\{    \widehat{\mathbb{E}}_{N^{(e)}}[{\boldsymbol{x}}^{(e)}_I ({\boldsymbol{x}}_{S\cup S^*}^{(e)})^{\top}  (\boldsymbol{\beta} - \boldsymbol{\beta}^*)]- \boldsymbol{\Sigma}^{(e)}_{I,S\cup S^*} (\boldsymbol{\beta} - \boldsymbol{\beta}^*)    \right\} \\
	 & \quad \le\sum_{e \in \mathcal{E}} \omega^{(e)} \widehat{\tau}^{(e)} \left\| \widehat{\mathbb{E}}_{m^{(e)}}[{\boldsymbol{x}}^{(e)}_I(z^{(e)} - \eta^{(e)}  )] -   \mathbb{E}[\boldsymbol{x}^{(e)}_I(z^{(e)} - \eta^{(e)}  )] \right\|_2 \left\|    \widehat{\mathbb{E}}_{N^{(e)}}[{\boldsymbol{x}}^{(e)}_I ({\boldsymbol{x}}_{S\cup S^*}^{(e)})^{\top}  (\boldsymbol{\beta} - \boldsymbol{\beta}^*)]- \boldsymbol{\Sigma}^{(e)}_{I,S\cup S^*} (\boldsymbol{\beta} - \boldsymbol{\beta}^*)    \right\|_2               \\
	 & \quad \le\sum_{e \in \mathcal{E}} \omega^{(e)} \widehat{\tau}^{(e)} \left\| \widehat{\mathbb{E}}_{m^{(e)}}[{\boldsymbol{x}}^{(e)}_{S^*}(z^{(e)} - \eta^{(e)}  )] -   \mathbb{E}[\boldsymbol{x}^{(e)}_{S^*}(z^{(e)} - \eta^{(e)}  )] \right\|_2 \left\|    \widehat{\mathbb{E}}_{N^{(e)}}[{\boldsymbol{x}}^{(e)}_{S^*} ({\boldsymbol{x}}_{S\cup S^*}^{(e)})^{\top}  (\boldsymbol{\beta} - \boldsymbol{\beta}^*)]- \boldsymbol{\Sigma}^{(e)}_{S^*,S\cup S^*} (\boldsymbol{\beta} - \boldsymbol{\beta}^*)    \right\|_2
\end{align*}
Next, we are going to study the following two events, define
\begin{align*}
	\mathcal{K}_{1, u}(e) & =\left\{\left\| \widehat{\mathbb{E}}_{m^{(e)}}[{\boldsymbol{x}}^{(e)}_{S^*}(z^{(e)} - \eta^{(e)}  )] -   \mathbb{E}[\boldsymbol{x}^{(e)}_{S^*}(z^{(e)} - \eta^{(e)}  )] \right\|_2 \leq C_1^{\prime} \kappa_U^{1 / 2} \sigma_x \sigma_{z}\left(\sqrt{\frac{s^*+u}{m^{(e)}}}+\frac{s^*+u}{m^{(e)}}\right)\right\}                                                      \\
	\mathcal{K}_{2, u}(e) & =\left\{\forall \boldsymbol{\beta} \in \mathbb{R}^p, \quad\left\|\widehat{\mathbb{E}}_{N^{(e)}}\left[\boldsymbol{x}_{S^*}^{(e)}\left(\left[\boldsymbol{x}^{(e)}\right]_{S\cup S^*}\right)^{\top}\right]-\boldsymbol{\Sigma}_{S^*,S\cup S^*}^{(e)}\right\|_2 \leq C_2^{\prime} \kappa_U \sigma_x^2\left(\sqrt{\frac{u+p}{N^{(e)}}}+\frac{u+p}{N^{(e)}}\right)\right\},
\end{align*}
We have proved $\mathbb{P}[\mathcal{K}_{2, u}(e)]\ge 1-e^{-u}$, for any fixed $u>0$ and $e\in\mathcal{E}$ in Step 3.1. We will prove at the end that the event $\mathcal{K}_{1, u}(e)$ occurs also with large probability.

Given the above claims are true, under the event $\mathcal{K}_u=\bigcap_{e \in \mathcal{E}}\left\{\mathcal{K}_{1, u}(e) \cap \mathcal{K}_{2, u}(e)\right\}$, with probability greater than $1-2|\mathcal{E}| e^{-u}$, we obtain

\begin{align*}
	\forall \boldsymbol{\beta} \in \mathbb{R}^p, \quad \sum_{e \in \mathcal{E}} \omega^{(e)} T_{3,2}^{(e)}(\boldsymbol{\beta}) & \le  \sum_{e \in \mathcal{E}} \omega^{(e)} \widehat{\tau}^{(e)}  C_1^{\prime} \kappa_U^{1 / 2} \sigma_x \sigma_{z}\left(\sqrt{\frac{s^*+u}{m^{(e)}}}+\frac{s^*+u}{m^{(e)}}\right)C_2^{\prime} \kappa_U \sigma_x^2\left(\sqrt{\frac{u+p}{N^{(e)}}}+\frac{u+p}{N^{(e)}}\right)\|\boldsymbol{\beta}-\boldsymbol{\beta}^*\|_2            \\
	                                                                                                                           & \le F_{3,2} \kappa_U^{3 / 2} \sigma_x^3 \sigma_{z}  \sum_{e \in \mathcal{E}} \omega^{(e)} \widehat{\tau}^{(e)}  \sqrt{\frac{s^*+u}{m^{(e)}}}\left(\sqrt{\frac{u+p}{N^{(e)}}}+\frac{u+p}{N^{(e)}}\right)\|\boldsymbol{\beta}-\boldsymbol{\beta}^*\|_2                                                                                 \\
	                                                                                                                           & \le  F_{3,2} \kappa_U^{3 / 2} \sigma_x^3 \sigma_{z}\sum_{e \in \mathcal{E}} \omega^{(e)}\widehat{\tau}^{(e)} \sqrt{\frac{(s^* +u)(u+p)}{m^{(e)}N^{(e)}}} \|\boldsymbol{\beta}-\boldsymbol{\beta}^*\|_2                                                                                                                               \\
	                                                                                                                           & \quad  + F_{3,2} \kappa_U^{3 / 2} \sigma_x^3 \sigma_{z}\sum_{e \in \mathcal{E}} \omega^{(e)}\widehat{\tau}^{(e)} \sqrt{\frac{(s^* +u)}{m^{(e)}}} \frac{u+p}{N^{(e)}}\|\boldsymbol{\beta}-\boldsymbol{\beta}^*\|_2                                                                                                                    \\
	                                                                                                                           & =  F_{3,2} \kappa_U^{3 / 2} \sigma_x^3 \sigma_{z} \sqrt{(s^* +u)(u+p)}\sum_{e \in \mathcal{E}} \frac{\omega^{(e)}\widehat{\tau}^{(e)}}{\sqrt{m^{(e)}N^{(e)}}}\|\boldsymbol{\beta}-\boldsymbol{\beta}^*\|_2                                                                                                                           \\
	                                                                                                                           & \quad  + F_{3,2} \kappa_U^{3 / 2} \sigma_x^3 \sigma_{z} \sqrt{s^* +u}(u+p) \sum_{e \in \mathcal{E}} \frac{ \omega^{(e)}\widehat{\tau}^{(e)}}{\sqrt{m^{(e)}}N^{(e)}}\|\boldsymbol{\beta}-\boldsymbol{\beta}^*\|_2                                                                                                                     \\
	                                                                                                                           & =  F_{3,2} \kappa_U^{3 / 2} \sigma_x^3 \sigma_{z} \frac{\sqrt{(s^* +u)(u+p)}}{\overline{\sqrt{mN}}^{\widehat{\tau}}} \|\boldsymbol{\beta}-\boldsymbol{\beta}^*\|_2 + F_{3,2} \kappa_U^{3 / 2} \sigma_x^3 \sigma_{z} \frac{\sqrt{s^* +u}(u+p)}{\overline{\sqrt{m} N}^{\widehat{\tau}} } \|\boldsymbol{\beta}-\boldsymbol{\beta}^*\|_2
\end{align*}

provided $s^*+u\le m_{\min}$. Finally, we finish the proof by setting $u = \log(2|\mathcal{E}|)+t$. It remains to prove the two events $\mathcal{K}_{1, u}(e)$ and $\mathcal{K}_{2, u}(e)$ occur with high probability.

Now fix $u>0, e\in \mathcal{E}$, let's prove $ \mathbb{P}[\mathcal{K}_{1, u}(e)]\ge 1-e^{-u}$. Notice that
\begin{align}
	 & \left\| \widehat{\mathbb{E}}_{m^{(e)}}[{\boldsymbol{x}}^{(e)}_{S^*}(z^{(e)} - \eta^{(e)}  )] -   \mathbb{E}[\boldsymbol{x}^{(e)}_{S^*}(z^{(e)} - \eta^{(e)}  )] \right\|_2\notag                                                                                                                                                                                   \\
	 & \quad =\sup _{\boldsymbol{v} \in \mathbb{R}^{s^*},\|\boldsymbol{v}\|_2=1} \boldsymbol{v}^{\top}\left( \widehat{\mathbb{E}}_{m^{(e)}}[{\boldsymbol{x}}^{(e)}_{S^*}(z^{(e)} - \eta^{(e)}  )] -   \mathbb{E}[\boldsymbol{x}^{(e)}_{S^*}(z^{(e)} - \eta^{(e)}  )] \right)\notag                                                                                        \\
	 & \quad = \sup _{\boldsymbol{v} \in \mathbb{R}^{s^*},\|\boldsymbol{v}\|_2=1} \boldsymbol{v}^{\top} \left\{ \frac{1}{m^{(e)}}\sum_{\ell=1}^{m^{(e)}}\left[\boldsymbol{x}_{\ell}^{(e)}\right]_{S^*} ( z_{\ell}^{(e)} - \eta^{(e)}) - \mathbb{E}\left[\left[\boldsymbol{x}_{\ell}^{(e)}\right]_{S^*} (z_{\ell}^{(e)}-\eta^{(e)})\right]\right\}\label{eq:step32l2norm}.
\end{align}
Let $\boldsymbol{v} _1^{(S^*)}, ..., \boldsymbol{v} _{N_{S^*}}^{(S^*)}$ be an $1/4$-covering of $\mathcal{B}(S^*)$, that is, for any $\boldsymbol{v}\in \mathcal{B}(S^*)$, there exists some $\pi(\boldsymbol{v})\in[N_{S^*}]$ such that $\left\|\boldsymbol{v}-\boldsymbol{v}_{\pi(v)}^{(S^*)}\right\|_2 \leq 1 / 4.$ It follows from standard empirical process result that $N_{S^*}\le 9^{s^*}$.

Denote $\boldsymbol{\xi} =  \frac{1}{m^{(e)}}\sum_{\ell=1}^{m^{(e)}}\left[\boldsymbol{x}_{\ell}^{(e)}\right]_{S^*} ( z_{\ell}^{(e)} - \eta^{(e)}) - \mathbb{E}\left[\left[\boldsymbol{x}_{\ell}^{(e)}\right]_{S^*} (z_{\ell}^{(e)}-\eta^{(e)})\right]$. By the observation
$$
	\sup _{\boldsymbol{v} \in \mathcal{B}\left(S^*\right)} \boldsymbol{v}^{\top} \boldsymbol{\xi}=\sup _{k \in\left[N_{S^*}\right]}\left(\boldsymbol{v}_k^{(S^*)}\right)^{\top} \boldsymbol{\xi}+\sup _{\boldsymbol{v} \in \mathcal{B}\left( S^*\right)}\left(\boldsymbol{v}-\boldsymbol{v}_{\pi(v)}^{(S^*)}\right)^{\top} \boldsymbol{\xi} \leq \sup _{k \in\left[N_{S^*}\right]}\left(\boldsymbol{v}_k^{(S^*)}\right)^{\top} \boldsymbol{\xi}+\frac{1}{4}\left\{\sup _{\boldsymbol{v} \in \mathcal{B}\left(S^*\right)} \boldsymbol{v}^{\top} \boldsymbol{\xi}\right\},
$$
which implies $\{\sup _{\boldsymbol{v} \in \mathcal{B}\left(S^*\right)} \boldsymbol{v}^{\top} \boldsymbol{\xi}\} \leq 2 \sup _{k \in\left[N_{S^*}\right]}\left(\boldsymbol{v}_k^{(S^*)}\right)^{\top} \boldsymbol{\xi}$, thus
\begin{align}
	 & \sup _{\boldsymbol{v} \in \mathbb{R}^{s^*},\|\boldsymbol{v}\|_2=1} \boldsymbol{v}^{\top} \left\{ \frac{1}{m^{(e)}}\sum_{\ell=1}^{m^{(e)}}\left[\boldsymbol{x}_{\ell}^{(e)}\right]_{S^*} ( z_{\ell}^{(e)} - \eta^{(e)}) - \mathbb{E}\left[\left[\boldsymbol{x}_{\ell}^{(e)}\right]_{S^*} (z_{\ell}^{(e)}-\eta^{(e)})\right]\right\}\notag                                              \\
	 & \quad \le  2 \sup _{k \in\left[N_{S^*}\right]}\frac{1}{m^{(e)}}\sum_{\ell=1}^{m^{(e)}}\left\{\left(\boldsymbol{v}_k^{(S^*)}\right)^{\top} \left[\boldsymbol{x}_{\ell}^{(e)}\right]_{S^*}  (z_{\ell}^{(e)}-\eta^{(e)})- \mathbb{E}\left[\left(\boldsymbol{v}_k^{(S^*)}\right)^{\top} \left[\boldsymbol{x}_{\ell}^{(e)}\right]_{(S^*)} (z_{\ell}^{(e)}-\eta^{(e)})\right]\right\}\notag \\
	 & \quad \equiv 2 \sup _{k \in\left[N_{S^*}\right]} Z(k)
	\label{eq:step32CSinequality},
\end{align}
Notice that for given fixed $\boldsymbol{v}_k^{(S^*)}\in\mathcal{B}(S^*)$, $Z(k)$ can be written as the sum of independent zero-mean random variables with parameter $(1/m^{(e)})\kappa_U^{1 / 2} \sigma_x\sigma_{z}$, because each summand is $(1/m^{(e)})$ times a centered product of two sub-Gaussian random variables $\left(\boldsymbol{v}_k^{(S^*)}\right)^{\top} \left[\boldsymbol{x}_{\ell}^{(e)}\right]_{(S^*)} $ and $z_{\ell}^{(e)}-\eta^{(e)}$ with parameters $\kappa_U^{1 / 2} \sigma_x$ and $\sigma_{z}$, respectively. Therefore, we have
$$
	\mathbb{P}\left[|Z( k)| \geq C^{\prime}\kappa_U^{1/2} \sigma_x \sigma_{z}\left\{\sqrt{\frac{x}{m^{(e)}}}+\frac{x}{m^{(e)}} \right\}\right] \leq 2 e^{-x},\quad \forall x>0,
$$
which tells us that
\begin{align*}
	 & \mathbb{P}\left[ 2 \sup _{k \in\left[N_{S^*}\right]}Z(k)\ge 2C^{\prime} \kappa_U^{1/2}\sigma_x \sigma_{z}\left\{\sqrt{\frac{x}{m^{(e)}}}+\frac{x}{m^{(e)}} \right\}\right]          \\
	 & \quad\le\sum_{k \in\left[N_{S^*}\right]} \mathbb{P}\left[  |Z(k)|\ge C^{\prime} \kappa_U^{1/2}\sigma_x \sigma_{z}\left\{\sqrt{\frac{x}{m^{(e)}}}+ \frac{x}{m^{(e)}} \right\}\right] \\
	 & \quad\leq \sum_{k \in\left[N_{S^*}\right]}2 e^{-x}                                                                                                                                  \\
	 & = 2N_{S^*}e^{-x}
\end{align*}
With fixed $u>0$, let $x = u + \log(2N_{S^*})\le 3(u +s^*) $ and facts \ref{eq:step32l2norm}, \ref{eq:step32CSinequality}, we have
$$
	\mathbb{P}\left[ \left\| \widehat{\mathbb{E}}_{m^{(e)}}[{\boldsymbol{x}}^{(e)}_{S^*}(z^{(e)} - \eta^{(e)}  )] -   \mathbb{E}[\boldsymbol{x}^{(e)}_{S^*}(z^{(e)} - \eta^{(e)}  )] \right\|_2\le C_1^{\prime}\kappa_U^{1/2} \sigma_x \sigma_{z}\left\{\sqrt{\frac{u+s^*}{m^{(e)}}}+\frac{u + s^*}{m^{(e)}} \right\}\right]\ge 1-e^{-u},
$$
where $C_1^{\prime} = 6 C^{\prime}$.

\noindent{\it Step 3.3}. UPPER BOUND ON $T_{3,3}^{(e)}(\boldsymbol{\beta})$.  We claim that the following event

\begin{align}
	W_{3, t} & =\left\{\forall \boldsymbol{\beta} \in \mathbb{R}^p, \quad \sum_{e \in \mathcal{E}} \omega^{(e)} \mathrm{T}_{3,3}^{(e)}(\boldsymbol{\beta}) \leq F_{3,3} \kappa_U^{3 / 2} \sigma_x^3 \sigma_{z} \frac{\sqrt{(s^* +\log(2|\mathcal{E}|)+t)(p+\log(2|\mathcal{E}|)+t)}}{\overline{\sqrt{nN}}^{\widehat{\tau}}} \|\boldsymbol{\beta}-\boldsymbol{\beta}^*\|_2\right.\notag \\
	         & \quad \quad\quad\quad\quad\quad\quad\quad\quad\quad\quad\quad \quad\quad\quad\quad\quad\quad\quad\quad\quad\quad\left.  + F_{3,3} \kappa_U^{3 / 2} \sigma_x^3 \sigma_{z} \frac{\sqrt{s^* +\log(2|\mathcal{E}|)+t}(p+\log(2|\mathcal{E}|)+t)}{\overline{\sqrt{n} N}^{\widehat{\tau}} } \|\boldsymbol{\beta}-\boldsymbol{\beta}^*\|_2\right\}
	\label{eq:event_w3t}
\end{align}
occurs with probability at least $1-e^{-t}$ if $s^*+t+\log (|2 \mathcal{E}|) \leq n_{\min }$.

\noindent{\it Step 3.4} UPPER BOUND ON $T_{3,4}^{(e)}(\boldsymbol{\beta})$. In this step, we claim that the following event
\begin{align}
	W_{4, t} & =\left\{\forall \boldsymbol{\beta} \in \mathbb{R}^p, \quad \sum_{e \in \mathcal{E}} \omega^{(e)} \mathrm{T}_{3,4}^{(e)}(\boldsymbol{\beta}) \leq F_{3,4} \kappa_U^{3 / 2} \sigma_x^3\frac{\sqrt{(s^*+\log(2|\mathcal{E}|)+t)(p+\log(2|\mathcal{E}|)+t)}}{\overline{\sqrt{mN}}^{\widehat{\tau},|\eta|}} \|\boldsymbol{\beta}-\boldsymbol{\beta}^*\|_2\right.\notag \\
	         & \quad \quad\quad\quad\quad\quad\quad\quad\quad\quad\quad\quad \quad\quad\quad\quad\quad\quad\quad\quad\quad\quad\left. + F_{3,4} \kappa_U^{3 / 2} \sigma_x^3 \frac{\sqrt{s^*+\log(2|\mathcal{E}|)+t}(p+\log(2|\mathcal{E}|)+t)}{\overline{\sqrt{m}N}^{\widehat{\tau},|\eta|}} \|\boldsymbol{\beta}-\boldsymbol{\beta}^*\|_2\right\}
	\label{eq:event_w4t}
\end{align}
occurs with probability at least $1-e^{-t}$ if $s^*+t+\log (|2 \mathcal{E}|) \leq m_{\min }$. Observe the following, $\forall \boldsymbol{\beta} \in \mathbb{R}^p$
\begin{align*}
	 & \quad \sum_{e \in \mathcal{E}} \omega^{(e)} T_{3,4}^{(e)}(\boldsymbol{\beta})                                                                                                                                                                                                                                                                                                                                                                                                                        \\
	 & \quad =\sum_{e \in \mathcal{E}} \omega^{(e)} \left\{\widehat{\tau}^{(e)}\left(\widehat{\mathbb{E}}_{m^{(e)}}\left[\boldsymbol{x}_I^{(e)}\right]-\mathbb{E}\left[\boldsymbol{x}_I^{(e)}\right]\right) \eta^{(e)}\right\}^{\top}\left\{\widehat{\mathbb{E}}_{N^{(e)}}\left[\boldsymbol{x}_I^{(e)}\left(\boldsymbol{x}^{(e)}\right)^{\top}\left(\boldsymbol{\beta}-\boldsymbol{\beta}^*\right)\right]-\boldsymbol{\Sigma}_{I,:}^{(e)}\left(\boldsymbol{\beta}-\boldsymbol{\beta}^*\right)\right\}       \\
	 & \quad=\sum_{e \in \mathcal{E}} \omega^{(e)} \widehat{\tau}^{(e)}\eta^{(e)} \left\{\widehat{\mathbb{E}}_{m^{(e)}}\left[\boldsymbol{x}_I^{(e)}\right]-\mathbb{E}\left[\boldsymbol{x}_I^{(e)}\right]\right\}^{\top}\left\{\widehat{\mathbb{E}}_{N^{(e)}}\left[\boldsymbol{x}_I^{(e)}\left(\boldsymbol{x}_{S\cup S^*}^{(e)}\right)^{\top}\left(\boldsymbol{\beta}-\boldsymbol{\beta}^*\right)\right]-\boldsymbol{\Sigma}_{I,S\cup S^*}^{(e)}\left(\boldsymbol{\beta}-\boldsymbol{\beta}^*\right)\right\} \\
	 & \quad \le\sum_{e \in \mathcal{E}} \omega^{(e)} \widehat{\tau}^{(e)}|\eta^{(e)}| \left\|\widehat{\mathbb{E}}_{m^{(e)}}\left[\boldsymbol{x}_I^{(e)}\right]-\mathbb{E}\left[\boldsymbol{x}_I^{(e)}\right] \right\|_2 \left\|    \widehat{\mathbb{E}}_{N^{(e)}}[{\boldsymbol{x}}^{(e)}_I ({\boldsymbol{x}}_{S\cup S^*}^{(e)})^{\top}  (\boldsymbol{\beta} - \boldsymbol{\beta}^*)]- \boldsymbol{\Sigma}^{(e)}_{I,S\cup S^*} (\boldsymbol{\beta} - \boldsymbol{\beta}^*)    \right\|_2                    \\
	 & \quad \le\sum_{e \in \mathcal{E}} \omega^{(e)} \widehat{\tau}^{(e)} |\eta^{(e)}|\left\| \widehat{\mathbb{E}}_{m^{(e)}}\left[\boldsymbol{x}_{S^*}^{(e)}\right]-\mathbb{E}\left[\boldsymbol{x}_{S^*}^{(e)}\right]\right\|_2 \left\|    \widehat{\mathbb{E}}_{N^{(e)}}[{\boldsymbol{x}}^{(e)}_{S^*} ({\boldsymbol{x}}_{S\cup S^*}^{(e)})^{\top}  (\boldsymbol{\beta} - \boldsymbol{\beta}^*)]- \boldsymbol{\Sigma}^{(e)}_{S^*,S\cup S^*} (\boldsymbol{\beta} - \boldsymbol{\beta}^*)    \right\|_2
\end{align*}

Next, we are going to study the following two events, define
\begin{align*}
	\mathcal{K}_{1, u}(e) & =\left\{\left\| \widehat{\mathbb{E}}_{m^{(e)}}\left[\boldsymbol{x}_{S^*}^{(e)}\right]-\mathbb{E}\left[\boldsymbol{x}_{S^*}^{(e)}\right]\right\|_2 \leq C_1^{\prime} \kappa_U^{1 / 2} \sigma_x \left(\sqrt{\frac{s^*+u}{m^{(e)}}}+\frac{s^*+u}{m^{(e)}}\right)\right\}                                                                                                 \\
	\mathcal{K}_{2, u}(e) & =\left\{\forall \boldsymbol{\beta} \in \mathbb{R}^p, \quad\left\|\widehat{\mathbb{E}}_{N^{(e)}}\left[\boldsymbol{x}_{S^*}^{(e)}\left(\left[\boldsymbol{x}^{(e)}\right]_{S\cup S^*}\right)^{\top}\right]-\boldsymbol{\Sigma}_{S^*,S\cup S^*}^{(e)}\right\|_2 \leq C_2^{\prime} \kappa_U \sigma_x^2\left(\sqrt{\frac{u+p}{N^{(e)}}}+\frac{u+p}{N^{(e)}}\right)\right\},
\end{align*}
We have proved $\mathbb{P}[\mathcal{K}_{2, u}(e)]\ge 1-e^{-u}$, for any fixed $u>0$ and $e\in\mathcal{E}$ in Step 3.1. We will prove at the end that the event $\mathcal{K}_{1, u}(e)$ occurs also with large probability.

Given the above claims are true, under the event $\mathcal{K}_u=\bigcap_{e \in \mathcal{E}}\left\{\mathcal{K}_{1, u}(e) \cap \mathcal{K}_{2, u}(e)\right\}$, with probability greater than $1-2|\mathcal{E}| e^{-u}$, we obtain
\begin{align*}
	\forall \boldsymbol{\beta} \in \mathbb{R}^p, \quad \sum_{e \in \mathcal{E}} \omega^{(e)} T_{3,4}^{(e)}(\boldsymbol{\beta}) & \le  \sum_{e \in \mathcal{E}} \omega^{(e)} \widehat{\tau}^{(e)}| \eta^{(e)}| C_1^{\prime} \kappa_U^{1 / 2} \sigma_x \left(\sqrt{\frac{s^*+u}{m^{(e)}}}+\frac{s^*+u}{m^{(e)}}\right)C_2^{\prime} \kappa_U \sigma_x^2\left(\sqrt{\frac{u+p}{N^{(e)}}}+\frac{u+p}{N^{(e)}}\right)\|\boldsymbol{\beta}-\boldsymbol{\beta}^*\|_2 \\
	                                                                                                                           & \le  F_{3,4} \kappa_U^{3 / 2} \sigma_x^3 \sum_{e \in \mathcal{E}} \omega^{(e)} \widehat{\tau}^{(e)} |\eta^{(e)}|\sqrt{\frac{s^*+u}{m^{(e)}}}\left(\sqrt{\frac{u+p}{N^{(e)}}}+\frac{u+p}{N^{(e)}}\right)\|\boldsymbol{\beta}-\boldsymbol{\beta}^*\|_2                                                                        \\
	                                                                                                                           & \le  F_{3,4} \kappa_U^{3 / 2} \sigma_x^3\sqrt{(s^*+u)(u+p)}  \sum_{e \in \mathcal{E}}  \frac{\omega^{(e)} \widehat{\tau}^{(e)} |\eta^{(e)}|}{\sqrt{m^{(e)}N^{(e)}}}
	\|\boldsymbol{\beta}-\boldsymbol{\beta}^*\|_2                                                                                                                                                                                                                                                                                                                                                                                                            \\
	                                                                                                                           & \quad + F_{3,4} \kappa_U^{3 / 2} \sigma_x^3 \sqrt{s^*+u}(u+p) \sum_{e \in \mathcal{E}}  \frac{\omega^{(e)} \widehat{\tau}^{(e)} |\eta^{(e)}|}{\sqrt{m^{(e)}}N^{(e)}}\|\boldsymbol{\beta}-\boldsymbol{\beta}^*\|_2                                                                                                           \\
	                                                                                                                           & =  F_{3,4} \kappa_U^{3 / 2} \sigma_x^3\frac{\sqrt{(s^*+u)(u+p)}}{\overline{\sqrt{mN}}^{\widehat{\tau},|\eta|}} \|\boldsymbol{\beta}-\boldsymbol{\beta}^*\|_2 + F_{3,4} \kappa_U^{3 / 2} \sigma_x^3 \frac{\sqrt{s^*+u}(u+p)}{\overline{\sqrt{m}N}^{\widehat{\tau},|\eta|}} \|\boldsymbol{\beta}-\boldsymbol{\beta}^*\|_2
\end{align*}
provided $p+u\le m_{\min}$. Finally, we finish the proof by setting $u = \log(2|\mathcal{E}|)+t$. It remains to prove the two events $\mathcal{K}_{1, u}(e)$ and $\mathcal{K}_{2, u}(e)$ occur with high probability.

Now fix $u>0, e\in \mathcal{E}$, let's prove $ \mathbb{P}[\mathcal{K}_{1, u}(e)]\ge 1-e^{-u}$. Notice that
\begin{align}
	\left\| \widehat{\mathbb{E}}_{m^{(e)}}\left[\boldsymbol{x}_{S^*}^{(e)}\right]-\mathbb{E}\left[\boldsymbol{x}_{S^*}^{(e)}\right]\right\|_2 & =\sup _{\boldsymbol{v} \in \mathbb{R}^{s^*},\|\boldsymbol{v}\|_2=1} \boldsymbol{v}^{\top}\left( \widehat{\mathbb{E}}_{m^{(e)}}\left[\boldsymbol{x}_{S^*}^{(e)}\right]-\mathbb{E}\left[\boldsymbol{x}_{S^*}^{(e)}\right]\right)\notag                                       \\
	                                                                                                                                          & = \sup _{\boldsymbol{v} \in \mathbb{R}^{s^*},\|\boldsymbol{v}\|_2=1} \boldsymbol{v}^{\top} \left\{ \frac{1}{m^{(e)}}\sum_{\ell=1}^{m^{(e)}}\left[\boldsymbol{x}_{\ell}^{(e)}\right]_{S^*}- \mathbb{E}\left[\left[\boldsymbol{x}_{\ell}^{(e)}\right]_{S^*} \right]\right\}.
	\label{eq:step34l2norm}
\end{align}
Let $\boldsymbol{v} _1^{(S^*)}, ..., \boldsymbol{v} _{N_{S^*}}^{(S^*)}$ be an $1/4$-covering of $\mathcal{B}(S^*)$, that is, for any $\boldsymbol{v}\in \mathcal{B}(S^*)$, there exists some $\pi(\boldsymbol{v})\in[N_{S^*}]$ such that $\left\|\boldsymbol{v}-\boldsymbol{v}_{\pi(v)}^{(S^*)}\right\|_2 \leq 1 / 4.$ It follows from standard empirical process result that $N_{S^*}\le 9^{s^*}$.

Denote $\boldsymbol{\xi} = \frac{1}{m^{(e)}}\sum_{\ell=1}^{m^{(e)}}[\boldsymbol{x}_{\ell}^{(e)}]_{S^*}- \mathbb{E}[[\boldsymbol{x}_{\ell}^{(e)}]_{S^*} ]$. By the observation
$$
	\sup _{\boldsymbol{v} \in \mathcal{B}\left(S^*\right)} \boldsymbol{v}^{\top} \boldsymbol{\xi}=\sup _{k \in\left[N_{S^*}\right]}\left(\boldsymbol{v}_k^{(S^*)}\right)^{\top} \boldsymbol{\xi}+\sup _{\boldsymbol{v} \in \mathcal{B}\left( S^*\right)}\left(\boldsymbol{v}-\boldsymbol{v}_{\pi(v)}^{(S^*)}\right)^{\top} \boldsymbol{\xi} \leq \sup _{k \in\left[N_{S^*}\right]}\left(\boldsymbol{v}_k^{(S^*)}\right)^{\top} \boldsymbol{\xi}+\frac{1}{4}\left\{\sup _{\boldsymbol{v} \in \mathcal{B}\left(S^*\right)} \boldsymbol{v}^{\top} \boldsymbol{\xi}\right\},
$$
which implies $\{\sup _{\boldsymbol{v} \in \mathcal{B}\left(S^*\right)} \boldsymbol{v}^{\top} \boldsymbol{\xi}\} \leq 2 \sup _{k \in\left[N_{S^*}\right]}\left(\boldsymbol{v}_k^{(S^*)}\right)^{\top} \boldsymbol{\xi}$, thus
\begin{align}
	 & \sup _{\boldsymbol{v} \in \mathbb{R}^{s^*},\|\boldsymbol{v}\|_2=1} \boldsymbol{v}^{\top} \left\{ \frac{1}{m^{(e)}}\sum_{\ell=1}^{m^{(e)}}\left[\boldsymbol{x}_{\ell}^{(e)}\right]_{S^*} - \mathbb{E}\left[\left[\boldsymbol{x}_{\ell}^{(e)}\right]_{S^*} \right]\right\}\notag                                                 \\
	 & \quad \le  2 \sup _{k \in\left[N_{S^*}\right]}\frac{1}{m^{(e)}}\sum_{\ell=1}^{m^{(e)}}\left\{\left(\boldsymbol{v}_k^{(S^*)}\right)^{\top} \left[\boldsymbol{x}_{\ell}^{(e)}\right]_{S^*} - \mathbb{E}\left[\left(\boldsymbol{v}_k^{(S^*)}\right)^{\top} \left[\boldsymbol{x}_{\ell}^{(e)}\right]_{(S^*)} \right]\right\}\notag \\
	 & \quad \equiv 2 \sup _{k \in\left[N_{S^*}\right]} Z(k)\label{eq:step34CSinequality},
\end{align}
Notice that for given fixed $\boldsymbol{v}_k^{(S^*)}\in\mathcal{B}(S^*)$, $Z(k)$ can be written as the sum of independent zero-mean random variables with parameter $(1/m^{(e)})\kappa_U^{1 / 2} \sigma_x$. Therefore, we have
$$
	\mathbb{P}\left[|Z( k)| \geq C^{\prime}\kappa_U^{1/2} \sigma_x \left\{\sqrt{\frac{x}{m^{(e)}}}+\frac{x}{m^{(e)}} \right\}\right] \leq 2 e^{-x},\quad \forall x>0,
$$
which tells us that
\begin{align*}
	 & \mathbb{P}\left[ 2 \sup _{k \in\left[N_{S^*}\right]}Z(k)\ge 2C^{\prime} \kappa_U^{1/2}\sigma_x \left\{\sqrt{\frac{x}{m^{(e)}}}+\frac{x}{m^{(e)}} \right\}\right]          \\
	 & \quad\le\sum_{k \in\left[N_{S^*}\right]} \mathbb{P}\left[  |Z(k)|\ge C^{\prime} \kappa_U^{1/2}\sigma_x \left\{\sqrt{\frac{x}{m^{(e)}}}+ \frac{x}{m^{(e)}} \right\}\right] \\
	 & \quad\leq \sum_{k \in\left[N_{S^*}\right]}2 e^{-x}                                                                                                                        \\
	 & = 2N_{S^*}e^{-x}
\end{align*}
With fixed $u>0$, let $x = u + \log(2N_{S^*})\le 3(u +s^*) $ and facts \ref{eq:step34l2norm}, \ref{eq:step34CSinequality}, we have
$$
	\mathbb{P}\left[   \left\| \widehat{\mathbb{E}}_{m^{(e)}}\left[\boldsymbol{x}_{S^*}^{(e)}\right]-\mathbb{E}\left[\boldsymbol{x}_{S^*}^{(e)}\right]\right\|_2\le C_1^{\prime}\kappa_U^{1/2} \sigma_x\left\{\sqrt{\frac{u+s^*}{m^{(e)}}}+\frac{u + s^*}{m^{(e)}} \right\}\right]\ge 1-e^{-u},
$$
where $C_1^{\prime} = 6 C^{\prime}$.

\noindent{\it Step 3.5}. UPPER BOUND ON $T_{3,5}^{(e)}(\boldsymbol{\beta})$. In this step, we claim that the following event
\begin{align}
	W_{5, t} & =\left\{\forall \boldsymbol{\beta} \in \mathbb{R}^p, \quad \sum_{e \in \mathcal{E}} \omega^{(e)} \mathrm{T}_{3,5}^{(e)}(\boldsymbol{\beta}) \leq F_{3,5} \kappa_U^{3 / 2} \sigma_x^3\frac{\sqrt{(s^*+\log(2|\mathcal{E}|)+t)(p+\log(2|\mathcal{E}|)+t)}}{\overline{\sqrt{nN}}^{\widehat{\tau},|\eta|}} \|\boldsymbol{\beta}-\boldsymbol{\beta}^*\|_2\right.\notag \\
	         & \quad \quad\quad\quad\quad\quad\quad\quad\quad\quad\quad\quad \quad\quad\quad\quad\quad\quad\quad\quad\quad\quad\left. + F_{3,5} \kappa_U^{3 / 2} \sigma_x^3 \frac{\sqrt{s^*+\log(2|\mathcal{E}|)+t}(p+\log(2|\mathcal{E}|)+t)}{\overline{\sqrt{n}N}^{\widehat{\tau},|\eta|}} \|\boldsymbol{\beta}-\boldsymbol{\beta}^*\|_2\right\}
	\label{eq:event_w5t}
\end{align}
occurs with probability at least $1-e^{-t}$ if $s^*+t+\log (|2 \mathcal{E}|) \leq n_{\min }$.

\noindent{\it Step 4}. CONCLUSION. We now conclude the proof by combining results obtained from Steps 1.1-1.12, Steps 2.1-2.15, and Steps 3.1-3.5. Plugging these upper bounds back into the decomposition in (\ref{eq:Jdecomp}). Under the event
$$
	\left(\bigcap_{j=1}^{12} C_{j,t}\right) \cap \left(\bigcap_{j=1}^{15} U_{j,t}\right) \cap \left(\bigcap_{j=1}^{5} W_{j,t} \right),
$$
which occurs with probability at least $1-32e^{-t}$, we have the following inequality:

\begin{align*}
	 & \frac{1}{c_1}\left(\mathrm{J}(\boldsymbol{\beta})-\mathrm{J}(\boldsymbol{\beta}^*)-\widehat{\mathrm{J}}_{\mathrm{Adj}}(\boldsymbol{\beta})+\widehat{\mathrm{J}}_{\mathrm{Adj}}\left(\boldsymbol{\beta}^*\right)\right)                                                                                          \\
	 & \le \kappa_U^{3 / 2} \sigma_x \sigma_{\varepsilon}\left\|\boldsymbol{\beta}-\boldsymbol{\beta}^*\right\|_2\left(\sqrt{\frac{t+p}{N_\omega}}+\frac{t+p}{N_*}\right)                                                                                                                                  \\
	 & \quad +  \kappa_U^{1 / 2} \sigma_x \sigma_{\varepsilon} \sqrt{\frac{t+p}{N_*}} \times \sqrt{\sum_{e \in \mathcal{E}} \omega^{(e)}\left\|\mathbb{E}\left[x_S^{(e)} \varepsilon^{(e)}\right]\right\|_2^2}+\kappa_U \sigma_x \sigma_{\varepsilon}^2 \frac{t+p}{N_*}                                    \\
	 & \quad + \kappa_U^2 \sigma_x^2\left\|\boldsymbol{\beta}-\boldsymbol{\beta}^*\right\|_2^2\left(\sqrt{\frac{t+p}{N_\omega}}+\frac{t+p}{N_*}\right)                                                                                                                                                     \\
	 & \quad +   \kappa_U^{3 / 2} \sigma_x^2 \sigma_{\varepsilon}\left\|\boldsymbol{\beta}-\boldsymbol{\beta}^*\right\|_2\left(\sqrt{\frac{t+p}{N_\omega}}+\frac{t+p}{N_*}\right)                                                                                                                          \\
	 & \quad +  \kappa_U^{3 / 2} \sigma_x \sigma_z\left\|\boldsymbol{\beta}-\boldsymbol{\beta}^*\right\|_2\left(\sqrt{\frac{t+p}{m_\omega^{\widehat{\tau}^2}}}+\frac{t+p}{m_*^{\widehat{\tau}}}\right)                                                                                                     \\
	 & \quad +  \kappa_U^{3 / 2} \sigma_x \sigma_z\left\|\boldsymbol{\beta}-\boldsymbol{\beta}^*\right\|_2\left(\sqrt{\frac{t+p}{n_\omega^{\widehat{\tau}^2}}}+\frac{t+p}{n_*^{\widehat{\tau}}}\right)                                                                                                     \\
	 & \quad +  \kappa_U^{1 / 2} \sigma_x \sigma_{z} \sqrt{\frac{t+p}{m^{\widehat{\tau}^2}_*}} \times \sqrt{\sum_{e \in \mathcal{E}} \omega^{(e)}\left\|\mathbb{E}\left[x_S^{(e)} \varepsilon^{(e)}\right]\right\|_2^2}+\kappa_U \sigma_x \sigma_{z}\sigma_{\varepsilon}\frac{t+p}{m^{\widehat{\tau}}_*}   \\
	 & \quad +  \kappa_U^{1 / 2} \sigma_x \sigma_{z} \sqrt{\frac{t+p}{n^{\widehat{\tau}^2}_*}}  \times \sqrt{\sum_{e \in \mathcal{E}} \omega^{(e)}\left\|\mathbb{E}\left[x_S^{(e)} \varepsilon^{(e)}\right]\right\|_2^2}+ \kappa_U \sigma_x \sigma_{z}\sigma_{\varepsilon}\frac{t+p}{n^{\widehat{\tau}}_*} \\
	 & \quad + \kappa_U^{3 / 2} \sigma_x \left\|\boldsymbol{\beta}-\boldsymbol{\beta}^*\right\|_2\left(\sqrt{\frac{t+p}{m_\omega^{\widehat{\tau}^2,|\eta|^2}}}+\frac{t+p}{m_*^{\widehat{\tau},|\eta|}}\right)                                                                                              \\
	 & \quad +    \kappa_U^{3 / 2} \sigma_x \left\|\boldsymbol{\beta}-\boldsymbol{\beta}^*\right\|_2\left(\sqrt{\frac{t+p}{n_\omega^{\widehat{\tau}^2,|\eta|^2}}}+\frac{t+p}{n_*^{\widehat{\tau},|\eta|}}\right)                                                                                           \\
	 & \quad +  \kappa_U^{1 / 2} \sigma_x  \sqrt{\frac{t+p}{m^{\widehat{\tau}^2,|\eta|^2}_*}} \times \sqrt{\sum_{e \in \mathcal{E}} \omega^{(e)}\left\|\mathbb{E}\left[x_S^{(e)} \varepsilon^{(e)}\right]\right\|_2^2}+ \kappa_U \sigma_x\sigma_{\varepsilon}\frac{t+p}{m^{\widehat{\tau},|\eta|}_*}       \\
	 & \quad +  \kappa_U^{1 / 2} \sigma_x  \sqrt{\frac{t+p}{n^{\widehat{\tau}^2,|\eta|^2}_*}} \times \sqrt{\sum_{e \in \mathcal{E}} \omega^{(e)}\left\|\mathbb{E}\left[x_S^{(e)} \varepsilon^{(e)}\right]\right\|_2^2}+ \kappa_U \sigma_x\sigma_{\varepsilon}\frac{t+p}{n^{\widehat{\tau},|\eta|}_*}       \\
	 & \quad +    \kappa_U \sigma_x^2 \sigma_{\varepsilon}^2 \frac{t+\log \left(2|\mathcal{E}|\left|S^*\right|\right)}{\bar{N}}\left|S^* \backslash S\right|                                                                                                                                               \\
	 & \quad +    \kappa_U \sigma_x^2 \sigma_{z}^2 \frac{t+\log \left(2|\mathcal{E}|\left|S^*\right|\right)}{\bar{N}}\left|S^* \backslash S\right|                                                                                                                                                         \\
	 & \quad +  \kappa_U \sigma_x^2 \sigma_{z}^2 \frac{t+\log \left(2|\mathcal{E}|\left|S^*\right|\right)}{\bar{n}}\left|S^* \backslash S\right|                                                                                                                                                           \\
	 & \quad +   \kappa_U \sigma_x^2  \frac{t+\log \left(2|\mathcal{E}|\left|S^*\right|\right)}{\bar{N}^{|\eta|^2}}\left|S^* \backslash S\right|                                                                                                                                                           \\
	 & \quad +   \kappa_U \sigma_x^2  \frac{t+\log \left(2|\mathcal{E}|\left|S^*\right|\right)}{\bar{n}^{|\eta|^2}}\left|S^* \backslash S\right|                                                                                                                                                           \\
	 & \quad +   \kappa_U \sigma_x^2 \sigma_{\varepsilon}\sigma_{z} \frac{t+ \log(4|\mathcal{E}||S^*|)}{\overline{\sqrt{mN}}^{\widehat{\tau}}}|S^* \backslash S|                                                                                                                                           \\
	 & \quad +  \kappa_U \sigma_x^2 \sigma_{\varepsilon}\sigma_{z} \frac{t+ \log(4|\mathcal{E}||S^*|)}{\overline{\sqrt{nN}}^{\widehat{\tau}}}|S^* \backslash S|                                                                                                                                            \\
\end{align*}
\begin{align*}
	 & \quad +  \kappa_U \sigma_x^2 \sigma_{\varepsilon} \frac{t+ \log(4|\mathcal{E}||S^*|)}{\overline{\sqrt{mN}}^{\widehat{\tau},|\eta|}}|S^* \backslash S|                                                                                                                                                                              \\
	 & \quad +   \kappa_U \sigma_x^2 \sigma_{\varepsilon} \frac{t+ \log(4|\mathcal{E}||S^*|)}{\overline{\sqrt{nN}}^{\widehat{\tau},|\eta|}}|S^* \backslash S|                                                                                                                                                                             \\
	 & \quad + \kappa_U \sigma_x^2 \sigma_{z}^2 \frac{t+ \log(4|\mathcal{E}||S^*|)}{\overline{\sqrt{nN}}}|S^* \backslash S|                                                                                                                                                                                                               \\
	 & \quad +  \kappa_U \sigma_x^2 \frac{t+ \log(4|\mathcal{E}||S^*|)}{\overline{\sqrt{Nn}}^{|\eta|^2}}|S^* \backslash S|                                                                                                                                                                                                                \\
	 & \quad +  \kappa_U \sigma_x^2 \sigma_{z} \frac{t+ \log(4|\mathcal{E}||S^*|)}{\bar{m}^{\widehat{\tau}^2, |\eta|}}|S^* \backslash S|                                                                                                                                                                                                  \\
	 & \quad +   \kappa_U \sigma_x^2 \sigma_{z} \frac{t+ \log(4|\mathcal{E}||S^*|)}{\overline{\sqrt{mn}}^{\widehat{\tau}^2, |\eta|}}|S^* \backslash S|                                                                                                                                                                                    \\
	 & \quad +   \kappa_U \sigma_x^2 \sigma_{z} \frac{t+ \log(4|\mathcal{E}||S^*|)}{\overline{\sqrt{mn}}^{\widehat{\tau}^2, |\eta|}}|S^* \backslash S|                                                                                                                                                                                    \\
	 & \quad +  \kappa_U \sigma_x^2 \sigma_{z} \frac{t+ \log(4|\mathcal{E}||S^*|)}{\bar{ n}^{\widehat{\tau}^2, |\eta|}}|S^* \backslash S|                                                                                                                                                                                                 \\
	 & \quad + \kappa_U^{3 / 2} \sigma_x^3 \sigma_{\varepsilon} \frac{\sqrt{(s^*+\log(2|\mathcal{E}|)+t)(p+\log(2|\mathcal{E}|)+t)}}{\bar{N}}\|\boldsymbol{\beta}-\boldsymbol{\beta}^*\|_2                                                                                                                                                \\
	 & \quad\quad\quad\quad\quad\quad\quad\quad\quad\quad\quad\quad\quad\quad\quad\quad\quad\quad\quad\quad\quad\quad\quad\quad\quad + \kappa_U^{3 / 2} \sigma_x^3 \sigma_{\varepsilon}  \frac{\sqrt{s^* + \log(2|\mathcal{E}|)+t}(p+\log(2|\mathcal{E}|)+t)}{N_{\dagger}}\|\boldsymbol{\beta}-\boldsymbol{\beta}^*\|_2                   \\
	 & \quad + \kappa_U^{3 / 2} \sigma_x^3 \sigma_{z} \frac{\sqrt{(s^* +\log(2|\mathcal{E}|)+t)(p+\log(2|\mathcal{E}|)+t)}}{\overline{\sqrt{mN}}^{\widehat{\tau}}} \|\boldsymbol{\beta}-\boldsymbol{\beta}^*\|_2                                                                                                                          \\
	 & \quad\quad\quad\quad\quad\quad\quad\quad\quad\quad\quad\quad\quad\quad\quad\quad\quad\quad\quad\quad\quad\quad\quad\quad\quad + \kappa_U^{3 / 2} \sigma_x^3 \sigma_{z} \frac{\sqrt{s^* +\log(2|\mathcal{E}|)+t}(p+\log(2|\mathcal{E}|)+t)}{\overline{\sqrt{m} N}^{\widehat{\tau}} } \|\boldsymbol{\beta}-\boldsymbol{\beta}^*\|_2  \\
	 & \quad +   \kappa_U^{3 / 2} \sigma_x^3 \sigma_{z} \frac{\sqrt{(s^* +\log(2|\mathcal{E}|)+t)(p+\log(2|\mathcal{E}|)+t)}}{\overline{\sqrt{nN}}^{\widehat{\tau}}} \|\boldsymbol{\beta}-\boldsymbol{\beta}^*\|_2                                                                                                                        \\
	 & \quad\quad\quad\quad\quad\quad\quad\quad\quad\quad\quad\quad\quad\quad\quad\quad\quad\quad\quad\quad\quad\quad\quad\quad\quad +  \kappa_U^{3 / 2} \sigma_x^3 \sigma_{z} \frac{\sqrt{s^* +\log(2|\mathcal{E}|)+t}(p+\log(2|\mathcal{E}|)+t)}{\overline{\sqrt{n} N}^{\widehat{\tau}} } \|\boldsymbol{\beta}-\boldsymbol{\beta}^*\|_2 \\
	 & \quad +    \kappa_U^{3 / 2} \sigma_x^3\frac{\sqrt{(s^*+\log(2|\mathcal{E}|)+t)(p+\log(2|\mathcal{E}|)+t)}}{\overline{\sqrt{mN}}^{\widehat{\tau},|\eta|}} \|\boldsymbol{\beta}-\boldsymbol{\beta}^*\|_2                                                                                                                             \\
	 & \quad\quad\quad\quad\quad\quad\quad\quad\quad\quad\quad\quad\quad\quad\quad\quad\quad\quad\quad\quad\quad\quad\quad\quad\quad+ \kappa_U^{3 / 2} \sigma_x^3 \frac{\sqrt{s^*+\log(2|\mathcal{E}|)+t}(p+\log(2|\mathcal{E}|)+t)}{\overline{\sqrt{m}N}^{\widehat{\tau},|\eta|}}\|\boldsymbol{\beta}-\boldsymbol{\beta}^*\|_2           \\
	 & \quad +  \kappa_U^{3 / 2} \sigma_x^3\frac{\sqrt{(s^*+\log(2|\mathcal{E}|)+t)(p+\log(2|\mathcal{E}|)+t)}}{\overline{\sqrt{nN}}^{\widehat{\tau},|\eta|}}  \|\boldsymbol{\beta}-\boldsymbol{\beta}^*\|_2                                                                                                                              \\
	 & \quad\quad\quad\quad\quad\quad\quad\quad\quad\quad\quad\quad\quad\quad\quad\quad\quad\quad\quad\quad\quad\quad\quad\quad\quad  + \kappa_U^{3 / 2} \sigma_x^3 \frac{\sqrt{s^*+\log(2|\mathcal{E}|)+t}(p+\log(2|\mathcal{E}|)+t)}{\overline{\sqrt{n}N}^{\widehat{\tau},|\eta|}} \|\boldsymbol{\beta}-\boldsymbol{\beta}^*\|_2
\end{align*}

Since we are in low dimension regime, we can simplify our result by making a reasonable assumption that $n_{\min }>\log (2|\mathcal{E}|)+p+t$.

\begin{align}
	 & \frac{1}{c_1}\left(\mathrm{J}(\boldsymbol{\beta})-\mathrm{J}(\boldsymbol{\beta}^*)-\widehat{\mathrm{J}}_{\mathrm{Adj}}(\boldsymbol{\beta})+\widehat{\mathrm{J}}_{\mathrm{Adj}}\left(\boldsymbol{\beta}^*\right)\right)\notag                                                                                          \\
	 & \le \kappa_U^{3 / 2} \sigma_x \sigma_{\varepsilon}\left\|\boldsymbol{\beta}-\boldsymbol{\beta}^*\right\|_2\sqrt{\frac{t+p}{N_*}}  \notag                                                                                                                                                                     \\
	 & \quad +  \kappa_U^{1 / 2} \sigma_x \sigma_{\varepsilon} \sqrt{\frac{t+p}{N_*}} \times \sqrt{\sum_{e \in \mathcal{E}} \omega^{(e)}\left\|\mathbb{E}\left[x_S^{(e)} \varepsilon^{(e)}\right]\right\|_2^2}+\kappa_U \sigma_x \sigma_{\varepsilon}^2 \frac{t+p}{N_*}    \notag                                   \\
	 & \quad + \kappa_U^2 \sigma_x^2\left\|\boldsymbol{\beta}-\boldsymbol{\beta}^*\right\|_2^2\sqrt{\frac{t+p}{N_*}}    \notag                                                                                                                                                                                      \\
	 & \quad +   \kappa_U^{3 / 2} \sigma_x^2 \sigma_{\varepsilon}\left\|\boldsymbol{\beta}-\boldsymbol{\beta}^*\right\|_2\sqrt{\frac{t+p}{N_*}}     \notag                                                                                                                                                          \\
	 & \quad +  \kappa_U^{3 / 2} \sigma_x \sigma_z\left\|\boldsymbol{\beta}-\boldsymbol{\beta}^*\right\|_2\sqrt{\frac{t+p}{m_*^{\widehat{\tau}}}}               \notag                                                                                                                                              \\
	 & \quad +  \kappa_U^{3 / 2} \sigma_x \sigma_z\left\|\boldsymbol{\beta}-\boldsymbol{\beta}^*\right\|_2\sqrt{\frac{t+p}{n_*^{\widehat{\tau}}}}            \notag                                                                                                                                                 \\
	 & \quad +  \kappa_U^{1 / 2} \sigma_x \sigma_{z} \sqrt{\frac{t+p}{m^{\widehat{\tau}^2}_*}} \times \sqrt{\sum_{e \in \mathcal{E}} \omega^{(e)}\left\|\mathbb{E}\left[x_S^{(e)} \varepsilon^{(e)}\right]\right\|_2^2}+\kappa_U \sigma_x \sigma_{z}\sigma_{\varepsilon}\frac{t+p}{m^{\widehat{\tau}}_*} \notag     \\
	 & \quad +  \kappa_U^{1 / 2} \sigma_x \sigma_{z} \sqrt{\frac{t+p}{n^{\widehat{\tau}^2}_*}}  \times \sqrt{\sum_{e \in \mathcal{E}} \omega^{(e)}\left\|\mathbb{E}\left[x_S^{(e)} \varepsilon^{(e)}\right]\right\|_2^2}+ \kappa_U \sigma_x \sigma_{z}\sigma_{\varepsilon}\frac{t+p}{n^{\widehat{\tau}}_*} \notag   \\
	 & \quad + \kappa_U^{3 / 2} \sigma_x \left\|\boldsymbol{\beta}-\boldsymbol{\beta}^*\right\|_2\sqrt{\frac{t+p}{m_*^{\widehat{\tau},|\eta|}}}          \notag                                                                                                                                                     \\
	 & \quad +    \kappa_U^{3 / 2} \sigma_x \left\|\boldsymbol{\beta}-\boldsymbol{\beta}^*\right\|_2\sqrt{\frac{t+p}{n_*^{\widehat{\tau},|\eta|}}}       \notag                                                                                                                                                     \\
	 & \quad +  \kappa_U^{1 / 2} \sigma_x  \sqrt{\frac{t+p}{m^{\widehat{\tau}^2,|\eta|^2}_*}} \times \sqrt{\sum_{e \in \mathcal{E}} \omega^{(e)}\left\|\mathbb{E}\left[x_S^{(e)} \varepsilon^{(e)}\right]\right\|_2^2}+ \kappa_U \sigma_x\sigma_{\varepsilon}\frac{t+p}{m^{\widehat{\tau},|\eta|}_*}  \notag        \\
	 & \quad +  \kappa_U^{1 / 2} \sigma_x  \sqrt{\frac{t+p}{n^{\widehat{\tau}^2,|\eta|^2}_*}} \times \sqrt{\sum_{e \in \mathcal{E}} \omega^{(e)}\left\|\mathbb{E}\left[x_S^{(e)} \varepsilon^{(e)}\right]\right\|_2^2}+ \kappa_U \sigma_x\sigma_{\varepsilon}\frac{t+p}{n^{\widehat{\tau},|\eta|}_*}   \notag       \\
	 & \quad +    \kappa_U \sigma_x^2 \sigma_{\varepsilon}^2 \frac{t+\log \left(2|\mathcal{E}|\left|S^*\right|\right)}{\bar{N}}\left|S^* \backslash S\right|     \notag                                                                                                                                             \\
	 & \quad +    \kappa_U \sigma_x^2 \sigma_{z}^2 \frac{t+\log \left(2|\mathcal{E}|\left|S^*\right|\right)}{\bar{N}}\left|S^* \backslash S\right|    \notag                                                                                                                                                        \\
	 & \quad +  \kappa_U \sigma_x^2 \sigma_{z}^2 \frac{t+\log \left(2|\mathcal{E}|\left|S^*\right|\right)}{\bar{n}}\left|S^* \backslash S\right|    \notag                                                                                                                                                          \\
	 & \quad +   \kappa_U \sigma_x^2  \frac{t+\log \left(2|\mathcal{E}|\left|S^*\right|\right)}{\bar{N}^{|\eta|^2}}\left|S^* \backslash S\right|    \notag                                                                                                                                                          \\
	 & \quad +   \kappa_U \sigma_x^2  \frac{t+\log \left(2|\mathcal{E}|\left|S^*\right|\right)}{\bar{n}^{|\eta|^2}}\left|S^* \backslash S\right|  \notag                                                                                                                                                            \\
	 & \quad +   \kappa_U \sigma_x^2 \sigma_{\varepsilon}\sigma_{z} \frac{t+ \log(4|\mathcal{E}||S^*|)}{\overline{\sqrt{mN}}^{\widehat{\tau}}}|S^* \backslash S|                    \notag                                                                                                                          \\
	 & \quad +  \kappa_U \sigma_x^2 \sigma_{\varepsilon}\sigma_{z} \frac{t+ \log(4|\mathcal{E}||S^*|)}{\overline{\sqrt{nN}}^{\widehat{\tau}}}|S^* \backslash S|      \notag                                                                                                                                         \\
	 & \quad +   \kappa_U \sigma_x^2 \sigma_{\varepsilon} \frac{t+ \log(4|\mathcal{E}||S^*|)}{\overline{\sqrt{nN}}^{\widehat{\tau},|\eta|}}|S^* \backslash S|          \notag                                                                                                                                       \\
\end{align}
\begin{align}
	 & \quad + \kappa_U \sigma_x^2 \sigma_{z}^2 \frac{t+ \log(4|\mathcal{E}||S^*|)}{\overline{\sqrt{nN}}}|S^* \backslash S|      \notag                                               \\
	 & \quad +  \kappa_U \sigma_x^2 \frac{t+ \log(4|\mathcal{E}||S^*|)}{\overline{\sqrt{Nn}}^{|\eta|^2}}|S^* \backslash S|      \notag                                                \\
	 & \quad +  \kappa_U \sigma_x^2 \sigma_{z} \frac{t+ \log(4|\mathcal{E}||S^*|)}{\bar{m}^{\widehat{\tau}^2, |\eta|}}|S^* \backslash S|      \notag                                  \\
	 & \quad +   \kappa_U \sigma_x^2 \sigma_{z} \frac{t+ \log(4|\mathcal{E}||S^*|)}{\overline{\sqrt{mn}}^{\widehat{\tau}^2, |\eta|}}|S^* \backslash S|   \notag                       \\
	 & \quad +   \kappa_U \sigma_x^2 \sigma_{z} \frac{t+ \log(4|\mathcal{E}||S^*|)}{\overline{\sqrt{mn}}^{\widehat{\tau}^2, |\eta|}}|S^* \backslash S|      \notag                    \\
	 & \quad +  \kappa_U \sigma_x^2 \sigma_{z} \frac{t+ \log(4|\mathcal{E}||S^*|)}{\bar{ n}^{\widehat{\tau}^2, |\eta|}}|S^* \backslash S|         \notag                              \\
	 & \quad + \kappa_U^{3 / 2} \sigma_x^3 \sigma_{\varepsilon} \frac{p+\log(2|\mathcal{E}|)+t}{\bar{N}}\|\boldsymbol{\beta}-\boldsymbol{\beta}^*\|_2     \notag                      \\
	 & \quad + \kappa_U^{3 / 2} \sigma_x^3 \sigma_{z} \frac{p+\log(2|\mathcal{E}|)+t}{\overline{\sqrt{mN}}^{\widehat{\tau}}} \|\boldsymbol{\beta}-\boldsymbol{\beta}^*\|_2 \notag     \\
	 & \quad +   \kappa_U^{3 / 2} \sigma_x^3 \sigma_{z} \frac{p+\log(2|\mathcal{E}|)+t}{\overline{\sqrt{nN}}^{\widehat{\tau}}} \|\boldsymbol{\beta}-\boldsymbol{\beta}^*\|_2 \notag   \\
	 & \quad +    \kappa_U^{3 / 2} \sigma_x^3\frac{p+\log(2|\mathcal{E}|)+t}{\overline{\sqrt{mN}}^{\widehat{\tau},|\eta|}} \|\boldsymbol{\beta}-\boldsymbol{\beta}^*\|_2   \notag     \\
	 & \quad +  \kappa_U^{3 / 2} \sigma_x^3\frac{p+\log(2|\mathcal{E}|)+t}{\overline{\sqrt{nN}}^{\widehat{\tau},|\eta|}}  \|\boldsymbol{\beta}-\boldsymbol{\beta}^*\|_2\label{eq:J_general_bound}
\end{align}

This complete the proof. 

\qed

\subsection{Proof of Corollary \ref{corollary:oneside_boundJ_high_missing_good_imp}}

Here, we focus on simplifying the upper bound in (\ref{eq:J_general_bound}) under the scenario of substantial missingness in each environment, with an accurate imputation model. First, let's plug $\widehat{\tau}^{(e)}=m^{(e)} / N^{(e)}$ in the quantities $m_*^{\widehat{\tau}}$, $n_*^{\widehat{\tau}}$, $m_*^{\widehat{\tau}^2}$, $n_*^{\widehat{\tau}^2}$, $m_*^{\widehat{\tau},|\eta|}$, $n_*^{\widehat{\tau},|\eta|}$, $m_*^{\widehat{\tau}^2,|\eta|^2}$, $n_*^{\widehat{\tau}^2,|\eta|^2}$, $\overline{\sqrt{m N}}^{\widehat{\tau}}$, $\overline{\sqrt{n N}}^{\widehat{\tau}}$, $\overline{\sqrt{m N}}^{\widehat{\tau},|\eta|}$, $\overline{\sqrt{n N}}^{\widehat{\tau},|\eta|}$, $\bar{m}^{\widehat{\tau}^2,|\eta|}$, $\bar{n}^{\widehat{\tau}^2,|\eta|}$, $\overline{\sqrt{m n}}^{\widehat{\tau}^2,|\eta|}$, we get

\begin{align*}
	m_*^{\widehat{\tau}}                            & = \min _{e \in \mathcal{E}} \frac{m^{(e)}}{\omega^{(e)} \widehat{\tau}^{(e)}} = \min _{e \in \mathcal{E}} \frac{m^{(e)}}{\omega^{(e)} (m^{(e)}/ N^{(e)})}=\min_{e\in\mathcal{E}} \frac{N^{(e)}}{\omega^{(e)}} = N_*
	,                                                                                                                                                                                                                                                                                                                                       \\
	n_*^{\widehat{\tau}}                            & = \min _{e \in \mathcal{E}} \frac{n^{(e)}}{\omega^{(e)} \widehat{\tau}^{(e)}}=\min _{e \in \mathcal{E}} \frac{n^{(e)}}{\omega^{(e)} (m^{(e)}/ N^{(e)})}=\min_{e\in\mathcal{E}} \frac{n^{(e)}}{m^{(e)}}\frac{N^{(e)}}{\omega^{(e)}},                                                                                                                              \\
	m_*^{\widehat{\tau}^2}                          & = \min _{e \in \mathcal{E}} \frac{m^{(e)}}{\omega^{(e)}\left(\widehat{\tau}^{(e)}\right)^2}=\min _{e \in \mathcal{E}} \frac{m^{(e)}}{\omega^{(e)}(m^{(e)}/ N^{(e)})^2} = \min_{e\in\mathcal{E}} \frac{N^{(e)}}{m^{(e)}}\frac{N^{(e)}}{\omega^{(e)}}
	,                                                                                                                                                                                                                                                                                                                                       \\
	n_*^{\widehat{\tau}^2}                          & = \min _{e \in \mathcal{E}} \frac{n^{(e)}}{\omega^{(e)}\left(\widehat{\tau}^{(e)}\right)^2}= \min _{e \in \mathcal{E}} \frac{n^{(e)}}{\omega^{(e)}(m^{(e)}/ N^{(e)})^2}= \min_{e\in\mathcal{E}} \frac{n^{(e)}}{m^{(e)}}\frac{N^{(e)}}{m^{(e)}} \frac{N^{(e)}}{\omega^{(e)}}
	,                                                                                                                                                                                                                                                                                                                                       \\
	m_*^{\widehat{\tau},|\eta|}                     & = \min _{e \in \mathcal{E}} \frac{m^{(e)}}{\omega^{(e)} \widehat{\tau}^{(e)}\left|\eta^{(e)}\right|}= \min _{e \in \mathcal{E}} \frac{m^{(e)}}{\omega^{(e)} (m^{(e)}/ N^{(e)})\left|\eta^{(e)}\right|}= \min_{e\in\mathcal{E}} \frac{1}{|\eta^{(e)}|}\frac{N^{(e)}}{\omega^{(e)}},                                                                               \\
	n_*^{\widehat{\tau},|\eta|}                     & = \min _{e \in \mathcal{E}} \frac{n^{(e)}}{\omega^{(e)} \widehat{\tau}^{(e)}\left|\eta^{(e)}\right|}= \min _{e \in \mathcal{E}} \frac{n^{(e)}}{\omega^{(e)} (m^{(e)}/ N^{(e)})\left|\eta^{(e)}\right|}= \min_{e\in\mathcal{E}} \frac{n^{(e)}}{m^{(e)}} \frac{1}{|\eta^{(e)}|}\frac{N^{(e)}}{\omega^{(e)}},                                                                               \\
	m_*^{\widehat{\tau}^2,|\eta|^2}                 & = \min _{e \in \mathcal{E}} \frac{m^{(e)}}{\omega^{(e)}\left(\widehat{\tau}^{(e)}\right)^2\left|\eta^{(e)}\right|^2}= \min _{e \in \mathcal{E}} \frac{m^{(e)}}{\omega^{(e)}(m^{(e)}/ N^{(e)})^2\left|\eta^{(e)}\right|^2} = \min_{e\in\mathcal{E}} \frac{1}{|\eta^{(e)|^2}}\frac{N^{(e)}}{m^{(e)}}\frac{N^{(e)}}{\omega^{(e)}},                                                            \\
	n_*^{\widehat{\tau}^2,|\eta|^2}                 & = \min _{e \in \mathcal{E}} \frac{n^{(e)}}{\omega^{(e)}\left(\widehat{\tau}^{(e)}\right)^2\left|\eta^{(e)}\right|^2}= \min _{e \in \mathcal{E}} \frac{n^{(e)}}{\omega^{(e)}(m^{(e)}/ N^{(e)})^2\left|\eta^{(e)}\right|^2}=\min_{e\in\mathcal{E}}  \frac{n^{(e)}}{m^{(e)}}\frac{1}{|\eta^{(e)|^2}}\frac{N^{(e)}}{m^{(e)}} \frac{N^{(e)}}{\omega^{(e)}},                                                            \\
	\overline{\sqrt{m N}}^{\widehat{\tau}}          & = \left(\sum_{e \in \mathcal{E}} \frac{\omega^{(e)} \widehat{\tau}^{(e)}}{\sqrt{m^{(e)} N^{(e)}}}\right)^{-1}= \left(\sum_{e \in \mathcal{E}} \frac{\omega^{(e)} (m^{(e)}/ N^{(e)})}{\sqrt{m^{(e)} N^{(e)}}}\right)^{-1}=\left(\sum_{e \in \mathcal{E}} \frac{\omega^{(e)}\sqrt{m^{(e)}}}{(N^{(e)})^{3/2}}\right)^{-1}
	,                                                                                                                                                                                                                                                                                                                                       \\
	\overline{\sqrt{n N}}^{\widehat{\tau}}          & = \left(\sum_{e \in \mathcal{E}} \frac{\omega^{(e)} \widehat{\tau}^{(e)}}{\sqrt{n^{(e)} N^{(e)}}}\right)^{-1}= \left(\sum_{e \in \mathcal{E}} \frac{\omega^{(e)} (m^{(e)}/ N^{(e)})}{\sqrt{n^{(e)} N^{(e)}}}\right)^{-1}= \left(\sum_{e \in \mathcal{E}} \frac{1}{\sqrt{n^{(e)}/m^{(e)}}} \frac{\omega^{(e)}\sqrt{m^{(e)}}}{(N^{(e)})^{3/2}}\right)^{-1}
	,                                                                                                                                                                                                                                                                                                                                       \\
	\overline{\sqrt{m N}}^{\widehat{\tau},|\eta|}   & = \left(\sum_{e \in \mathcal{E}} \frac{\omega^{(e)} \widehat{\tau}^{(e)}\left|\eta^{(e)}\right|}{\sqrt{m^{(e)} N^{(e)}}}\right)^{-1}= \left(\sum_{e \in \mathcal{E}} \frac{\omega^{(e)} (m^{(e)}/ N^{(e)})\left|\eta^{(e)}\right|}{\sqrt{m^{(e)} N^{(e)}}}\right)^{-1}=\left(\sum_{e \in \mathcal{E}} \frac{|\eta^{(e)}|\omega^{(e)}\sqrt{m^{(e)}}}{(N^{(e)})^{3/2}}\right)^{-1},               \\
	\overline{\sqrt{n N}}^{\widehat{\tau},|\eta|}   & = \left(\sum_{e \in \mathcal{E}} \frac{\omega^{(e)} \widehat{\tau}^{(e)}\left|\eta^{(e)}\right|}{\sqrt{n^{(e)} N^{(e)}}}\right)^{-1}= \left(\sum_{e \in \mathcal{E}} \frac{\omega^{(e)} (m^{(e)}/ N^{(e)})\left|\eta^{(e)}\right|}{\sqrt{n^{(e)} N^{(e)}}}\right)^{-1}=\left(\sum_{e \in \mathcal{E}} \frac{1}{\sqrt{n^{(e)}/m^{(e)}}} \frac{|\eta^{(e)}|\omega^{(e)}\sqrt{m^{(e)}}}{(N^{(e)})^{3/2}}\right)^{-1},               \\
	\bar{m}^{\widehat{\tau}^2,|\eta|}               & = \left(\sum_{e \in \mathcal{E}} \frac{\omega^{(e)}\left(\widehat{\tau}^{(e)}\right)^2\left|\eta^{(e)}\right|}{m^{(e)}}\right)^{-1}= \left(\sum_{e \in \mathcal{E}} \frac{\omega^{(e)}(m^{(e)}/ N^{(e)})^2\left|\eta^{(e)}\right|}{m^{(e)}}\right)^{-1}=\left(\sum_{e \in \mathcal{E}} \frac{\left|\eta^{(e)}\right|\omega^{(e)}m^{(e)}}{(N^{(e)})^2}\right)^{-1},                              \\
	\bar{n}^{\widehat{\tau}^2,|\eta|}               & = \left(\sum_{e \in \mathcal{E}} \frac{\omega^{(e)}\left(\widehat{\tau}^{(e)}\right)^2\left|\eta^{(e)}\right|}{n^{(e)}}\right)^{-1}= \left(\sum_{e \in \mathcal{E}} \frac{\omega^{(e)}(m^{(e)}/ N^{(e)})^2\left|\eta^{(e)}\right|}{n^{(e)}}\right)^{-1}=\left(\sum_{e \in \mathcal{E}} \frac{1}{n^{(e)}/m^{(e)}}\frac{\left|\eta^{(e)}\right|\omega^{(e)}m^{(e)}}{(N^{(e)})^2}\right)^{-1},                              \\
	\overline{\sqrt{m n}}^{\widehat{\tau}^2,|\eta|} & = \left(\sum_{e \in \mathcal{E}} \frac{\omega^{(e)}\left(\widehat{\tau}^{(e)}\right)^2\left|\eta^{(e)}\right|}{\sqrt{n^{(e)} m^{(e)}}}\right)^{-1}= \left(\sum_{e \in \mathcal{E}} \frac{\omega^{(e)}(m^{(e)}/ N^{(e)})^2\left|\eta^{(e)}\right|}{\sqrt{n^{(e)} m^{(e)}}}\right)^{-1}=\left(\sum_{e \in \mathcal{E}} \frac{1}{\sqrt{n^{(e)}/m^{(e)}}} \frac{|\eta^{(e)}|\omega^{(e)}m^{(e)}}{(N^{(e)})^2}\right)^{-1}.
\end{align*}

When $n^{(e)}/m^{(e)} < 0.618, \forall e\in\mathcal{E}$, which is equivalent to $\widehat{\tau}_{\min} > 0.618$, $n_*^{\widehat{\tau}}\le m_*^{\widehat{\tau}}=N_*$ and $n_*^{\widehat{\tau},|\eta|} \le m_*^{\widehat{\tau},|\eta|}$. Furthermore, suppose $\eta_{\max} < 1$, then we have $\kappa_U^{3 / 2}  \sigma_x^2 \sigma_{\varepsilon}\sigma_z\|\boldsymbol{\beta}-\boldsymbol{\beta}^*\|_2\sqrt{{(t+p)}/{n_*^{\widehat{\tau}}}}$ and $\kappa_U \sigma_x\sigma_{\varepsilon}^2\sigma_z {(t+p)}/{n^{\widehat{\tau}}_*}$ control the upper bound. 

Moreover, $n_*^{\widehat{\tau}^2} \le N_*$, which can be seen from the fact that $(n^{(e)}/m^{(e)})(N^{(e)}/m^{(e)})<1$ is equivalent to $n^{(e)}/m^{(e)} < 0.618$. Under the assumption that $\eta_{\max} < 1$, we have $ \kappa_U^{1 / 2} \sigma_x \sigma_{\varepsilon} \sigma_z \sqrt{{(t+p)}/{n^{\widehat{\tau}^2}_*}} \times \sqrt{\sum_{e \in \mathcal{E}} \omega^{(e)}\|\mathbb{E}[x_S^{(e)} \varepsilon^{(e)}]\|_2^2}$ in the final expression of the upper bound. 

Next, it is clear to see that $\bar{n} < \min\{ \bar{N}, \bar{N}^{|\eta|^2},  \bar{n}^{|\eta|^2},  \overline{\sqrt{m N}}^{\widehat{\tau}}, \overline{\sqrt{n N}}^{\widehat{\tau}}, \overline{\sqrt{n N}}, \bar{n}^{\widehat{\tau}^2,|\eta|}\}$. Note that $\eta_{\max} < 1$ is a sufficient condition for $|\eta^{(e)}| < 1/ \{\widehat{\tau}^{(e)}(1- \widehat{\tau}^{(e)})^{1/2}\}$, $|\eta^{(e)}| < 1/ [\widehat{\tau}^{(e)}\{\widehat{\tau}^{(e)}(1-\widehat{\tau}^{(e)})\}^{1/2}]$, $|\eta^{(e)}| < 1/ \{\widehat{\tau}^{(e)}(1-\widehat{\tau}^{(e)})\}$, and $|\eta^{(e)}| < 1/ (1- \widehat{\tau}^{(e)})^{1/4}$ for all $e\in \mathcal{E}$. Thus, $\bar{n} <  \min\{\overline{\sqrt{n N}}^{\widehat{\tau}, |\eta|},\overline{\sqrt{n N}}^{|\eta|^2},\bar{m}^{\widehat{\tau}^2,|\eta|},  \overline{\sqrt{m n}}^{\widehat{\tau}^2,|\eta|} \}$ as well. Therefore, we have $\kappa_U \sigma_x^2 \sigma_{\varepsilon}^2 \sigma_z^2 {\{t+\log \left(2|\mathcal{E}|\left|S^*\right|\right)\}}\left|S^* \backslash S\right|/{\bar{n}}$ control the upper bound. 

Finally, since $n^{(e)}/m^{(e)} < 0.618, \forall e\in\mathcal{E}$, we have $\overline{\sqrt{n N}}^{\widehat{\tau}}  < \overline{\sqrt{m N}}^{\widehat{\tau}}$ and $\overline{\sqrt{n N}}^{\widehat{\tau},|\eta|} < \overline{\sqrt{m N}}^{\widehat{\tau},|\eta|} $. Moreover, since $\eta_{\max} < 1$, $\overline{\sqrt{n N}}^{\widehat{\tau}} < \overline{\sqrt{n N}}^{\widehat{\tau},|\eta|}$. Therefore, it remains to compare $\overline{\sqrt{n N}}^{\widehat{\tau}}$ with $\bar{N}$. To show $\overline{\sqrt{n N}}^{\widehat{\tau}}$ is smaller than $\bar{N}$, it is equivalent to show $([1/\{(1/\widehat{\tau}{(e)})-1\}]\widehat{\tau}{(e)} )^{1/2}$ is greater than $1$, which is true since $ \widehat{\tau}^{(e)}> 0.618$ and the fact that for any $x>0.618, ([1/\{(1/x)-1\}]x )^{1/2} > 1$. Thus, $\kappa_U^{3 / 2} \sigma_x^3 \sigma_{\varepsilon} \sigma_z {(p+\log (2|\mathcal{E}|)+t)}/{\overline{\sqrt{n N}}^{\widehat{\tau}}}\left\|\boldsymbol{\beta}-\boldsymbol{\beta}^*\right\|_2$ appears in the final upper bound. 

Together, we can simplify the bound in (\ref{eq:J_general_bound}) as follows: 
\begin{align}
	\frac{1}{c_1}\left(\mathrm{J}(\boldsymbol{\beta})-\mathrm{J}(\boldsymbol{\beta}^*)-\widehat{\mathrm{J}}_{\mathrm{Adj}}(\boldsymbol{\beta})+\widehat{\mathrm{J}}_{\mathrm{Adj}}\left(\boldsymbol{\beta}^*\right)\right) &\le + \kappa_U^2 \sigma_x^2\left\|\boldsymbol{\beta}-\boldsymbol{\beta}^*\right\|_2^2\sqrt{\frac{t+p}{N_*}}   \notag\\  
	&\quad +  \kappa_U^{3 / 2}  \sigma_x^2 \sigma_{\varepsilon}\sigma_z \left\|\boldsymbol{\beta}-\boldsymbol{\beta}^*\right\|_2\sqrt{\frac{t+p}{n_*^{\widehat{\tau}}}}\notag\\                               
	& \quad +  \kappa_U^{1 / 2} \sigma_x \sigma_{\varepsilon} \sigma_z \sqrt{\frac{t+p}{n^{\widehat{\tau}^2}_*}} \times \sqrt{\sum_{e \in \mathcal{E}} \omega^{(e)}\left\|\mathbb{E}\left[x_S^{(e)} \varepsilon^{(e)}\right]\right\|_2^2}  \notag \\
	& \quad + \kappa_U \sigma_x\sigma_{\varepsilon}^2\sigma_z\frac{t+p}{n^{\widehat{\tau}}_*}\notag\\
	& \quad +\kappa_U \sigma_x^2 \sigma_{\varepsilon}^2 \sigma_z^2 \frac{t+\log \left(4|\mathcal{E}|\left|S^*\right|\right)}{\bar{n}}\left|S^* \backslash S\right|\notag\\
	&\quad +\kappa_U^{3 / 2} \sigma_x^3 \sigma_{\varepsilon} \sigma_z \frac{p+\log (2|\mathcal{E}|)+t}{\overline{\sqrt{n N}}^{\widehat{\tau}}}\left\|\boldsymbol{\beta}-\boldsymbol{\beta}^*\right\|_2\label{eq:boundJ_step1}
\end{align}

This completes the proof.
\qed

\subsection{Proof of Corollary \ref{corollary:oneside_boundJ_high_missing_imprecise_imp}}

We focus on simplifying the upper bound in (\ref{eq:J_general_bound}) under the scenario of great missingness in each environment, with moderate imputation model.

When $\widehat{\tau}_{\min} > 0.618$, $n_*^{\widehat{\tau}}\le m_*^{\widehat{\tau}}=N_*$ and $n_*^{\widehat{\tau},|\eta|} \le m_*^{\widehat{\tau},|\eta|}$. Since $\eta_{\min} > 1$, we have $\kappa_U^{3 / 2}  \sigma_x^2 \sigma_{\varepsilon}\sigma_z\|\boldsymbol{\beta}-\boldsymbol{\beta}^*\|_2\sqrt{{(t+p)}/{n_*^{\widehat{\tau},|\eta|}}}$ and $\kappa_U \sigma_x\sigma_{\varepsilon}^2\sigma_z {(t+p)}/{n^{\widehat{\tau},|\eta|}_*}$ control the upper bound.

Moreover, $n_*^{\widehat{\tau}^2} \le N_*$, which can be seen from the fact that $(n^{(e)}/m^{(e)})(N^{(e)}/m^{(e)})<1$ is equivalent to $n^{(e)}/m^{(e)} < 0.618$. Since $\eta_{\min} > 1$, we have $ \kappa_U^{1 / 2} \sigma_x \sigma_{\varepsilon} \sigma_z \sqrt{{(t+p)}/{n^{\widehat{\tau}^2,|\eta|^2}_*}} \times \sqrt{\sum_{e \in \mathcal{E}} \omega^{(e)}\|\mathbb{E}[x_S^{(e)} \varepsilon^{(e)}]\|_2^2}$ in the final expression of the upper bound. 

Next, it is clear to see that $\bar{n}^{|\eta|^2} < \min\{\bar{N}, \bar{N}^{|\eta|^2}, \bar{n}, \overline{\sqrt{n N}}^{|\eta|^2}, \bar{n}^{\widehat{\tau}^2,|\eta|}\}$,  $ \overline{\sqrt{n N}}^{\widehat{\tau}, |\eta|}> \overline{\sqrt{n N}}^{|\eta|^2}$, and $\bar{m}^{\widehat{\tau}^2,|\eta|} > \bar{n}^{\widehat{\tau}^2,|\eta|}$. Moreover, since $n^{(e)}/m^{(e)} < 0.618$, we have $m^{(e)}/n^{(e)} > 1.618 > 1$, which implies $\overline{\sqrt{m N}}^{\widehat{\tau}} > \overline{\sqrt{n N}}^{\widehat{\tau}}$ and $\overline{\sqrt{m n}}^{\widehat{\tau}^2,|\eta|} > \bar{n}^{\widehat{\tau}^2,|\eta|}$. By the assumption $\eta_{\min} > 1$, we have $\overline{\sqrt{n N}}^{\widehat{\tau}} > \overline{\sqrt{n N}}^{\widehat{\tau}, |\eta|}$ and $\overline{\sqrt{n N}} > \overline{\sqrt{n N}}^{|\eta|^2}$. Therefore, we have $\kappa_U \sigma_x^2 \sigma_{\varepsilon}^2 \sigma_z^2 {\{t+\log \left(2|\mathcal{E}|\left|S^*\right|\right)\}}\left|S^* \backslash S\right|/{\bar{n}^{|\eta|^2}}$ control the upper bound. 

Finally, since $n^{(e)}/m^{(e)} < 0.618, \forall e\in\mathcal{E}$, we have $\overline{\sqrt{n N}}^{\widehat{\tau}}  < \overline{\sqrt{m N}}^{\widehat{\tau}}$, $\overline{\sqrt{n N}}^{\widehat{\tau}} < \bar{N} $, and $\overline{\sqrt{n N}}^{\widehat{\tau},|\eta|} < \overline{\sqrt{m N}}^{\widehat{\tau},|\eta|} $. Moreover, since $\eta_{\min} > 1$, $ \overline{\sqrt{n N}}^{\widehat{\tau},|\eta|} < \overline{\sqrt{n N}}^{\widehat{\tau}}$.  Thus, $\kappa_U^{3 / 2} \sigma_x^3 \sigma_{\varepsilon} \sigma_z {(p+\log (2|\mathcal{E}|)+t)}/{\overline{\sqrt{n N}}^{\widehat{\tau}, |\eta|}}\left\|\boldsymbol{\beta}-\boldsymbol{\beta}^*\right\|_2$ appears in the final upper bound. 

Together, we can simplify the bound in (\ref{eq:J_general_bound}) as follows: 
\begin{align}
	\frac{1}{c_1}\left(\mathrm{J}(\boldsymbol{\beta})-\mathrm{J}(\boldsymbol{\beta}^*)-\widehat{\mathrm{J}}_{\mathrm{Adj}}(\boldsymbol{\beta})+\widehat{\mathrm{J}}_{\mathrm{Adj}}\left(\boldsymbol{\beta}^*\right)\right) &\le + \kappa_U^2 \sigma_x^2\left\|\boldsymbol{\beta}-\boldsymbol{\beta}^*\right\|_2^2\sqrt{\frac{t+p}{N_*}}   \notag\\  
	&\quad +  \kappa_U^{3 / 2}  \sigma_x^2 \sigma_{\varepsilon}\sigma_z \left\|\boldsymbol{\beta}-\boldsymbol{\beta}^*\right\|_2\sqrt{\frac{t+p}{n_*^{\widehat{\tau}, |\eta|}}}\notag\\                               
	& \quad +  \kappa_U^{1 / 2} \sigma_x \sigma_{\varepsilon} \sigma_z \sqrt{\frac{t+p}{n_*^{\widehat{\tau}^2, |\eta|^2}}} \times \sqrt{\sum_{e \in \mathcal{E}} \omega^{(e)}\left\|\mathbb{E}\left[x_S^{(e)} \varepsilon^{(e)}\right]\right\|_2^2}  \notag \\
	& \quad + \kappa_U \sigma_x\sigma_{\varepsilon}^2\sigma_z\frac{t+p}{n^{\widehat{\tau}, |\eta|}_*}\notag\\
	& \quad +\kappa_U \sigma_x^2 \sigma_{\varepsilon}^2 \sigma_z^2 \frac{t+\log \left(4|\mathcal{E}|\left|S^*\right|\right)}{\bar{n}^{|\eta|^2}}\left|S^* \backslash S\right|\notag\\
	&\quad +\kappa_U^{3 / 2} \sigma_x^3 \sigma_{\varepsilon} \sigma_z \frac{p+\log (2|\mathcal{E}|)+t}{\overline{\sqrt{n N}}^{\widehat{\tau}, |\eta|}}\left\|\boldsymbol{\beta}-\boldsymbol{\beta}^*\right\|_2
\end{align}

This completes the proof.
\qed

\subsection{Proof of Theorem \ref{theorem:nonasy_vsc}}

We start by utilizing the following decomposition established in \cite{fan2024environment}. 
\begin{align}
	\widehat{\mathrm{Q}}_{\mathrm{Adj}}(\boldsymbol{\beta} ; \gamma, \boldsymbol{\omega})-\widehat{\mathrm{Q}}_{\mathrm{Adj}}\left(\boldsymbol{\beta}^* ; \gamma, \boldsymbol{\omega}\right) &= \widehat{\mathrm{R}}_{\mathrm{Adj}}(\boldsymbol{\beta}) + \gamma \widehat{\mathrm{J}}_{\mathrm{Adj}}(\boldsymbol{\beta}) - \left\{\widehat{\mathrm{R}}_{\mathrm{Adj}}(\boldsymbol{\beta}^*) + \gamma \widehat{\mathrm{J}}_{\mathrm{Adj}}(\boldsymbol{\beta}^*)\right\}\notag\\
	&= \widehat{\mathrm{R}}_{\mathrm{Adj}}(\boldsymbol{\beta})-\widehat{\mathrm{R}}_{\mathrm{Adj}}(\boldsymbol{\beta}^*) + \gamma \left\{ \widehat{\mathrm{J}}_{\mathrm{Adj}}(\boldsymbol{\beta}) - \widehat{\mathrm{J}}_{\mathrm{Adj}}(\boldsymbol{\beta}^*)  \right\}\notag\\
	&= - \left[ \left\{ \mathrm{R}(\boldsymbol{\beta})-\mathrm{R}(\boldsymbol{\beta}^*)\right\}     -\left\{ \widehat{\mathrm{R}}_{\mathrm{Adj}}(\boldsymbol{\beta})-\widehat{\mathrm{R}}_{\mathrm{Adj}}(\boldsymbol{\beta}^*)\right\}   \right]\notag\\
	&\quad  - \gamma  \left[ \left\{ \mathrm{J}(\boldsymbol{\beta})-\mathrm{J}(\boldsymbol{\beta}^*)\right\}     -\left\{ \widehat{\mathrm{J}}_{\mathrm{Adj}}(\boldsymbol{\beta})-\widehat{\mathrm{J}}_{\mathrm{Adj}}(\boldsymbol{\beta}^*)\right\}   \right] \notag\\
	&\quad + \mathrm{Q}(\boldsymbol{\beta} ; \gamma, \boldsymbol{\omega}) -  \mathrm{Q}(\boldsymbol{\beta}^* ; \gamma, \boldsymbol{\omega})\notag \\
	&= - \mathcal{D}_{\mathrm{R}, \widehat{\mathrm{R}}_{\mathrm{Adj}}}(\boldsymbol{\beta}) - \gamma \mathcal{D}_{\mathrm{J}, \widehat{J}_{\mathrm{Adj}}}(\boldsymbol{\beta}) + \mathrm{Q}(\boldsymbol{\beta} ; \gamma, \boldsymbol{\omega}) -  \mathrm{Q}(\boldsymbol{\beta}^* ; \gamma, \boldsymbol{\omega}),\label{eq:decomQhatQ}
\end{align}
where $\mathcal{D}_{\mathrm{R}, \widehat{\mathrm{R}}_{\mathrm{Adj}}}(\boldsymbol{\beta})$ and $\mathcal{D}_{\mathrm{J}, \widehat{J}_{\mathrm{Adj}}}(\boldsymbol{\beta})$ are defined in (\ref{eq:D_R_RAdj}) and (\ref{eq:D_J_JAdj}), respectively.

To establish the variable selection consistency guarantee, we aim to derive a high-probability lower bound for the difference $\widehat{\mathrm{Q}}_{\mathrm{Adj}}(\boldsymbol{\beta} ; \gamma, \boldsymbol{\omega})-\widehat{\mathrm{Q}}_{\mathrm{Adj}}\left(\boldsymbol{\beta}^* ; \gamma, \boldsymbol{\omega}\right)$, given certain characteristics of $\boldsymbol{\beta}$. This is equivalent to obtaining high-probability upper bounds for $\mathcal{D}_{\mathrm{R}, \widehat{\mathrm{R}}_{\mathrm{Adj}}}(\boldsymbol{\beta})$ and $\mathcal{D}_{\mathrm{J}, \widehat{J}_{\mathrm{Adj}}}(\boldsymbol{\beta})$, and a lower bound for $\mathrm{Q}(\boldsymbol{\beta} ; \gamma, \boldsymbol{\omega}) -  \mathrm{Q}(\boldsymbol{\beta}^* ; \gamma, \boldsymbol{\omega})$. Since the quantity $\mathrm{Q}(\boldsymbol{\beta} ; \gamma, \boldsymbol{\omega}) -  \mathrm{Q}(\boldsymbol{\beta}^* ; \gamma, \boldsymbol{\omega})$ is a population quantity, we directly leverage the result in Proposition C.5 of \cite{fan2024environment}. For completeness, we state the result here. 

\begin{prop}\label{prop:lower_bound_Qbeta_Qbetastar}
Assume \ref{cond:independent}, \ref{cond:pd_covariance_matrix}, and \ref{cond:identification} holds. If $\gamma \geq 3 \gamma^*$, where $\gamma^*$ is defined in (\ref{eq:gamma_star}), then
\begin{align}
\mathrm{Q}(\boldsymbol{\beta} ; \gamma, \boldsymbol{\omega})-\mathrm{Q}\left(\boldsymbol{\beta}^* ; \gamma, \boldsymbol{\omega}\right) & \geq \frac{\kappa_L}{2}\left\|\boldsymbol{\beta}-\boldsymbol{\beta}^{\star}\right\|_2^2+\frac{\gamma}{3} \mathrm{~J}(\boldsymbol{\beta} ; \boldsymbol{\omega})\notag \\
& \geq \frac{\kappa_L}{2}\left\|\boldsymbol{\beta}-\boldsymbol{\beta}^*\right\|_2^2+\frac{\gamma}{6} \kappa_L^2 \overline{\mathrm{d}}_{\operatorname{supp}(\boldsymbol{\beta})}+\frac{\gamma}{6} \mathrm{~J}(\boldsymbol{\beta} ; \boldsymbol{\omega}).\label{eq:lower_bound_Qbeta_Qbetastar}
\end{align}
\end{prop}

In the following proof, we assume that $n_{\text {min }}>\log (4|\mathcal{E}|)+p+t$ and $\gamma>1+\kappa_L$ are satisfied. We begin by proving the case where $\widehat\tau_{\min} > 0.618$ and $\eta_{\max}\ge 1$, which corresponds to the scenario of high missingness and imprecise imputation, as discussed in Section \ref{subsubsection:bounds_imprecise_imp}. Under this setting, the event $\mathcal{A}_{1, t}$ correspond to the event $\mathcal{A}_{1, t, Case 3}$ in Corollary \ref{corollary:R_four_cases}, and the event $\mathcal{A}_{2, t}$ has the property listed in Corollary \ref{corollary:oneside_boundJ_high_missing_imprecise_imp}. 

Under the event $\mathcal{A}_{1, t} \cap \mathcal{A}_{2, t}$, which occurs with high probability, we have the following upper bound:
\begin{align}
	\mathcal{D}_{\mathrm{R}, \widehat{\mathrm{R}}_{\mathrm{Adj}}}(\boldsymbol{\beta}) + \gamma \mathcal{D}_{\mathrm{J}, \widehat{J}_{\mathrm{Adj}}}(\boldsymbol{\beta}) &\le C\gamma \left\{ \kappa_U^2 \sigma_x^2\left\|\boldsymbol{\beta}-\boldsymbol{\beta}^*\right\|_2^2\sqrt{\frac{t+p}{N_*}} \right.\notag\\
	&\quad + \left.  \kappa_U^{3 / 2}  \sigma_x^2 \sigma_{\varepsilon}\sigma_z \left\|\boldsymbol{\beta}-\boldsymbol{\beta}^*\right\|_2\left\{\sqrt{\frac{t+p}{n_*^{\widehat\tau,|\eta|}}} +\sigma_x \frac{p+\log(2|\mathcal{E}|)+t}{\overline{\sqrt{nN}}^{\widehat\tau,|\eta|}}\right\} \right.\notag\\
	&\quad+ \left. \kappa_U^{1 / 2} \sigma_x \sigma_{\varepsilon} \sigma_z \sqrt{\frac{t+p}{n^{\widehat\tau^2,|\eta|^2}_*}} \times \sqrt{\sum_{e \in \mathcal{E}} \omega^{(e)}\left\|\mathbb{E}\left[x_S^{(e)} \varepsilon^{(e)}\right]\right\|_2^2} \right.\notag\\
	&\quad+ \left. \left|S^* \backslash S\right| \kappa_U \sigma_x^2 \sigma^2_{\varepsilon} \sigma_z^2\frac{\log \left(4|\mathcal{E}|\left|S^*\right|\right)+t}{\bar{n}^{|\eta|^2}}+ \kappa_U \sigma_x\sigma^2_{\varepsilon}\sigma_z\frac{t+p}{n^{\widehat\tau,|\eta|}_*}\right\}\notag\\
	&\quad= C\gamma (I_1 + I_2 +  I_3 + I_4 + I_5)\label{eq:decomposition_C12345}
\end{align}
First, we get an upper bound for $I_2$ by using the Young's inequality
\begin{align}
I_2 &= \kappa_U^{3 / 2}  \sigma_x^2 \sigma_{\varepsilon}\sigma_z \left\|\boldsymbol{\beta}-\boldsymbol{\beta}^*\right\|_2\left\{\sqrt{\frac{t+p}{n_*^{\widehat\tau,|\eta|}}} +\sigma_x\frac{p+\log(2|\mathcal{E}|)+t}{\overline{\sqrt{nN}}^{\widehat\tau,|\eta|}}\right\}\notag\\
&= \left\|\boldsymbol{\beta}-\boldsymbol{\beta}^*\right\|_2 \sqrt{\frac{\kappa_L}{8C'\gamma} }\sqrt{\frac{8C'\gamma}{\kappa_L}}\kappa_U^{3 / 2}  \sigma_x^2 \sigma_{\varepsilon}\sigma_z \left\{\sqrt{\frac{t+p}{n_*^{\widehat\tau,|\eta|}}} +\sigma_x\frac{p+\log(2|\mathcal{E}|)+t}{\overline{\sqrt{nN}}^{\widehat\tau,|\eta|}}\right\}\notag\\
&\le \frac{\kappa_L}{16 C'\gamma} \left\|\boldsymbol{\beta}-\boldsymbol{\beta}^*\right\|_2^2 +  \frac{8C'\gamma}{\kappa_L}\kappa_U^3\sigma_x^4\sigma_{\varepsilon}^2\sigma_z^2  \left\{\frac{t+p}{n_*^{\widehat\tau,|\eta|}} + \sigma_x^2\left(\frac{p+\log(2|\mathcal{E}|)+t}{\overline{\sqrt{nN}}^{\widehat\tau,|\eta|}}\right)^2\right\}\label{eq:I2_ub}
\end{align}
Next, let's upper bound $I_3$ using lemma C.10 of \cite{fan2024environment}.
\begin{align}
	I_3  &=  \kappa_U^{1 / 2} \sigma_x \sigma_{\varepsilon} \sigma_z \sqrt{\frac{t+p}{n^{\widehat\tau^2,|\eta|^2}_*}} \times \sqrt{\sum_{e \in \mathcal{E}} \omega^{(e)}\left\|\mathbb{E}\left[x_S^{(e)} \varepsilon^{(e)}\right]\right\|_2^2}\notag \\
	&\le  \kappa_U^{1 / 2} \sigma_x \sigma_{\varepsilon} \sigma_z \sqrt{\frac{t+p}{n^{\widehat\tau^2,|\eta|^2}_*}} \times \sqrt{2 \mathrm{~J}(\boldsymbol{\beta} ; \boldsymbol{\omega})+2 \kappa_U^2\left\|\boldsymbol{\beta}-\boldsymbol{\beta}^{\star}\right\|_2^2} \notag\\
	&\le  \sqrt{2}\kappa_U^{1 / 2} \sigma_x \sigma_{\varepsilon} \sigma_z \sqrt{\frac{t+p}{n^{\widehat\tau^2,|\eta|^2}_*}} \left\{\mathrm{~J}(\boldsymbol{\beta} ; \boldsymbol{\omega})^{1/2}+ \kappa_U\left\|\boldsymbol{\beta}-\boldsymbol{\beta}^{\star}\right\|_2\right\}\notag\\
	&=  \sqrt{2}\kappa_U^{1 / 2} \sigma_x \sigma_{\varepsilon} \sigma_z \sqrt{\frac{t+p}{n^{\widehat\tau^2,|\eta|^2}_*}}\mathrm{~J}(\boldsymbol{\beta} ; \boldsymbol{\omega})^{1/2}+ \sqrt{2}\kappa_U^{3/ 2} \sigma_x \sigma_{\varepsilon} \sigma_z \sqrt{\frac{t+p}{n^{\widehat\tau^2,|\eta|^2}_*}}\left\|\boldsymbol{\beta}-\boldsymbol{\beta}^{\star}\right\|_2\notag\\
	&= \sqrt{2}\kappa_U^{1 / 2} \sigma_x \sigma_{\varepsilon} \sigma_z \sqrt{\frac{t+p}{n^{\widehat\tau^2,|\eta|^2}_*}}\sqrt{\frac{6C'}{1}} \sqrt{\frac{1}{6C'}}\mathrm{~J}(\boldsymbol{\beta} ; \boldsymbol{\omega})^{1/2} + \sqrt{2}\kappa_U^{3/ 2} \sigma_x \sigma_{\varepsilon} \sigma_z \sqrt{\frac{t+p}{n^{\widehat\tau^2,|\eta|^2}_*}}\sqrt{\frac{16C'\gamma}{\kappa_L}}\sqrt{\frac{\kappa_L}{16C'\gamma}}\left\|\boldsymbol{\beta}-\boldsymbol{\beta}^{\star}\right\|_2\notag\\
	&\le 6C'\kappa_U\sigma_x^2 \sigma_{\varepsilon}^2 \sigma_z^2 \frac{t+p}{n^{\widehat\tau^2,|\eta|^2}_*} +\frac{1}{6C'}\mathrm{~J}(\boldsymbol{\beta} ; \boldsymbol{\omega}) + \frac{16C'\gamma}{\kappa_L}\kappa_U^{3} \sigma_x^2 \sigma_{\varepsilon}^2 \sigma_z^2 \frac{t+p}{n^{\widehat\tau^2,|\eta|^2}_*}+\frac{\kappa_L}{16C'\gamma}\left\|\boldsymbol{\beta}-\boldsymbol{\beta}^{\star}\right\|_2^2.\label{eq:I3_ub}
\end{align}
In the next step, we are going to upper bound $I_4$, observe that 
\begin{align}
\left|S^* \backslash \operatorname{supp}(\boldsymbol{\beta})\right| \min _{j \in S^*}\left|\beta_j^*\right|^2 & =\sum_{j \in S^*, \beta_j=0} \min _{j \in S^*}\left|\beta_j^*\right|^2 \notag\\
& \leq \sum_{j \in S^*, \beta_j=0}\left|\beta_j-\beta_j^*\right|^2 \leq \sum_{j=1}^p\left|\beta_j-\beta_j^*\right|^2=\left\|\boldsymbol{\beta}-\boldsymbol{\beta}^*\right\|_2^2.\label{eq:I4_inequality}
\end{align} 
To that end, we have
\begin{align}
I_4 &= \left|S^* \backslash S\right| \kappa_U \sigma_x^2 \sigma^2_{\varepsilon} \sigma_z^2\frac{\log \left(4|\mathcal{E}|\left|S^*\right|\right)+t}{\bar{n}^{|\eta|^2}}\notag\\
&= \left\{\left|S^* \backslash S\right|\min _{j \in S^*}\left|\beta_j^*\right|^2\right\}\left\{\frac{\kappa_U \sigma_x^2 \sigma^2_{\varepsilon} \sigma_z^2}{\min _{j \in S^*}\left|\beta_j^*\right|^2} \frac{\log \left(4|\mathcal{E}|\left|S^*\right|\right)+t}{\bar{n}^{|\eta|^2}}\right\}\notag\\
&\le \left\|\boldsymbol{\beta}-\boldsymbol{\beta}^*\right\|_2^2\left\{\frac{\kappa_U \sigma_x^2 \sigma^2_{\varepsilon} \sigma_z^2}{\min _{j \in S^*}\left|\beta_j^*\right|^2} \frac{\log \left(4|\mathcal{E}|\left|S^*\right|\right)+t}{\bar{n}^{|\eta|^2}}\right\}.\label{eq:I4_ub1}
\end{align}
Consider the following assumption in regards to the sample size
\begin{equation}\label{eq:sample_size_requirment1}
N_* \geq(p+t)\left(\frac{16 C \kappa_U^2 \sigma_x^2}{\kappa_L} \gamma\right)^2 \quad \text{and} \quad
\bar{n}^{|\eta|^2} \geq \frac{16 C \gamma \kappa_U \sigma_x^2 \sigma_{\varepsilon}^2\left\{\log \left(4|\mathcal{E}|\left|S^*\right|\right)+t\right\}}{\kappa_L \min _{j \in S^*}\left|\beta_j^*\right|^2}.
\end{equation}
Given the context of substantial missingness and imprecise imputation, in the rest of the proof, we assume $n_*^{\widehat{\tau}, |\eta|} \geq n_*^{\widehat{\tau}^2, |\eta|^2}$. 
\begin{align*}
	\mathcal{D}_{\mathrm{R}, \widehat{\mathrm{R}}_{\mathrm{Adj}}}(\boldsymbol{\beta}) + \gamma \mathcal{D}_{\mathrm{J}, \widehat{J}_{\mathrm{Adj}}}(\boldsymbol{\beta}) &\le C\gamma \left\{ \kappa_U^2 \sigma_x^2\left\|\boldsymbol{\beta}-\boldsymbol{\beta}^*\right\|_2^2\sqrt{\frac{t+p}{N_*}} \right.\\
	&\quad + \left.  \frac{\kappa_L}{16 C'\gamma} \left\|\boldsymbol{\beta}-\boldsymbol{\beta}^*\right\|_2^2 +  \frac{8C'\gamma}{\kappa_L}\kappa_U^3\sigma_x^4\sigma_{\varepsilon}^2\sigma_z^2  \left\{\frac{t+p}{n_*^{\widehat\tau,|\eta|}} + \sigma_x^2\left(\frac{p+\log(2|\mathcal{E}|)+t}{\overline{\sqrt{nN}}^{\widehat\tau,|\eta|}}\right)^2\right\} \right.\\
	&\quad+ \left. 6C'\kappa_U\sigma_x^2 \sigma_{\varepsilon}^2 \sigma_z^2 \frac{t+p}{n^{\widehat\tau^2,|\eta|^2}_*} +\frac{1}{6C'}\mathrm{~J}(\boldsymbol{\beta} ; \boldsymbol{\omega}) + \frac{16C'\gamma}{\kappa_L}\kappa_U^{3} \sigma_x^2 \sigma_{\varepsilon}^2 \sigma_z^2 \frac{t+p}{n^{\widehat\tau^2,|\eta|^2}_*}+\frac{\kappa_L}{16C'\gamma}\left\|\boldsymbol{\beta}-\boldsymbol{\beta}^{\star}\right\|_2^2 \right.\\
	&\quad+ \left. \left\|\boldsymbol{\beta}-\boldsymbol{\beta}^*\right\|_2^2\left\{\frac{\kappa_U \sigma_x^2 \sigma^2_{\varepsilon} \sigma_z^2}{\min _{j \in S^*}\left|\beta_j^*\right|^2} \frac{\log \left(4|\mathcal{E}|\left|S^*\right|\right)+t}{\bar{n}^{|\eta|^2}}\right\}+\kappa_U \sigma_x\sigma^2_{\varepsilon}\sigma_z\frac{t+p}{n^{\widehat\tau,|\eta|}_*}\right\}\\
	&\le C\gamma \left\{\frac{\kappa_L}{16 C'\gamma} \left\|\boldsymbol{\beta}-\boldsymbol{\beta}^*\right\|_2^2   \right.\\
	&\quad + \left.  \frac{\kappa_L}{16 C'\gamma} \left\|\boldsymbol{\beta}-\boldsymbol{\beta}^*\right\|_2^2 +  \frac{8C'\gamma}{\kappa_L}\kappa_U^3\sigma_x^4\sigma_{\varepsilon}^2\sigma_z^2  \left\{\frac{t+p}{n_*^{\widehat\tau,|\eta|}} + \sigma_x^2\left(\frac{p+\log(2|\mathcal{E}|)+t}{\overline{\sqrt{nN}}^{\widehat\tau,|\eta|}}\right)^2\right\} \right.\\
	&\quad+ \left. 6C'\kappa_U\sigma_x^2 \sigma_{\varepsilon}^2 \sigma_z^2 \frac{t+p}{n^{\widehat\tau^2,|\eta|^2}_*} +\frac{1}{6C'}\mathrm{~J}(\boldsymbol{\beta} ; \boldsymbol{\omega}) + \frac{16C'\gamma}{\kappa_L}\kappa_U^{3} \sigma_x^2 \sigma_{\varepsilon}^2 \sigma_z^2 \frac{t+p}{n^{\widehat\tau^2,|\eta|^2}_*}+\frac{\kappa_L}{16C'\gamma}\left\|\boldsymbol{\beta}-\boldsymbol{\beta}^{\star}\right\|_2^2 \right.\\
	&\quad+ \left. \frac{\kappa_L}{16 C'\gamma} \left\|\boldsymbol{\beta}-\boldsymbol{\beta}^*\right\|_2^2 +\kappa_U \sigma_x\sigma^2_{\varepsilon}\sigma_z\frac{t+p}{n^{\widehat\tau,|\eta|}_*}\right\}\\
	&\le C\gamma \left\{\frac{4\kappa_L}{16 C'\gamma} \left\|\boldsymbol{\beta}-\boldsymbol{\beta}^*\right\|_2^2 +\frac{1}{6C'}\mathrm{~J}(\boldsymbol{\beta} ; \boldsymbol{\omega})   \right.\\
	&\quad + \left.   \frac{8C'\gamma}{\kappa_L}\kappa_U^3\sigma_x^4\sigma_{\varepsilon}^2\sigma_z^2 \frac{t+p}{n_*^{\widehat\tau,|\eta|}}+ 6C'\kappa_U\sigma_x^2 \sigma_{\varepsilon}^2 \sigma_z^2 \frac{t+p}{n^{\widehat\tau^2,|\eta|^2}_*} + \frac{16C'\gamma}{\kappa_L}\kappa_U^{3} \sigma_x^2 \sigma_{\varepsilon}^2 \sigma_z^2 \frac{t+p}{n^{\widehat\tau^2,|\eta|^2}_*}+\kappa_U \sigma_x\sigma^2_{\varepsilon}\sigma_z\frac{t+p}{n^{\widehat\tau,|\eta|}_*} \right.\\
	&\quad+ \left. \frac{8C'\gamma}{\kappa_L}\kappa_U^3\sigma_x^6\sigma_{\varepsilon}^2\sigma_z^2  \left(\frac{p+\log(2|\mathcal{E}|)+t}{\overline{\sqrt{nN}}^{\widehat\tau,|\eta|}}\right)^2\right\}\\
	&\le  \frac{\kappa_L}{4} \left\|\boldsymbol{\beta}-\boldsymbol{\beta}^*\right\|_2^2 +\frac{\gamma}{6}\mathrm{~J}(\boldsymbol{\beta} ; \boldsymbol{\omega})  \\
	&\quad +C_1 \left\{\frac{\gamma^2}{\kappa_L}\kappa_U^3\sigma_x^4\sigma_{\varepsilon}^2\sigma_z^2 \frac{t+p}{n_*^{\widehat\tau^2,|\eta|^2}} +\frac{\gamma^2}{\kappa_L}\kappa_U^3\sigma_x^6\sigma_{\varepsilon}^2\sigma_z^2  \left(\frac{p+\log(2|\mathcal{E}|)+t}{\overline{\sqrt{nN}}^{\widehat\tau,|\eta|}}\right)^2\right\},
\end{align*}
for some universal constant $C_1$. Inserting the above inequality into equation (\ref{eq:decomQhatQ}), assuming $\gamma \ge 3\gamma^*$ and applying Proposition \ref{prop:lower_bound_Qbeta_Qbetastar}, we have 
\begin{align*}
	\widehat{\mathrm{Q}}_{\mathrm{Adj}}(\boldsymbol{\beta} ; \gamma, \boldsymbol{\omega})-\widehat{\mathrm{Q}}_{\mathrm{Adj}}\left(\boldsymbol{\beta}^* ; \gamma, \boldsymbol{\omega}\right)  &= - \mathcal{D}_{\mathrm{R}, \widehat{\mathrm{R}}_{\mathrm{Adj}}}(\boldsymbol{\beta}) - \gamma \mathcal{D}_{\mathrm{J}, \widehat{J}_{\mathrm{Adj}}}(\boldsymbol{\beta}) + \mathrm{Q}(\boldsymbol{\beta} ; \gamma, \boldsymbol{\omega}) -  \mathrm{Q}(\boldsymbol{\beta}^* ; \gamma, \boldsymbol{\omega})\\
	&\ge -\frac{\kappa_L}{4} \left\|\boldsymbol{\beta}-\boldsymbol{\beta}^*\right\|_2^2 - \frac{\gamma}{6}\mathrm{~J}(\boldsymbol{\beta} ; \boldsymbol{\omega})\\
	&\quad -C_1\gamma \left\{\frac{\gamma}{\kappa_L}\kappa_U^3\sigma_x^4\sigma_{\varepsilon}^2\sigma_z^2 \frac{t+p}{n_*^{\widehat\tau^2,|\eta|^2}} +\frac{\gamma}{\kappa_L}\kappa_U^3\sigma_x^6\sigma_{\varepsilon}^2\sigma_z^2  \left(\frac{p+\log(2|\mathcal{E}|)+t}{\overline{\sqrt{nN}}^{\widehat\tau,|\eta|}}\right)^2\right\}\\
	&\quad +  \frac{\kappa_L}{2}\left\|\boldsymbol{\beta}-\boldsymbol{\beta}^*\right\|_2^2+\frac{\gamma}{6} \kappa_L^2 \overline{\mathrm{d}}_{\operatorname{supp}(\boldsymbol{\beta})}+\frac{\gamma}{6} \mathrm{~J}(\boldsymbol{\beta} ; \boldsymbol{\omega})\\
	&= \frac{\kappa_L}{4} \left\|\boldsymbol{\beta}-\boldsymbol{\beta}^*\right\|_2^2+\frac{\gamma}{6} \kappa_L^2 \overline{\mathrm{d}}_{\operatorname{supp}(\boldsymbol{\beta})} \\
	&\quad -C_1 \left\{\frac{\gamma^2}{\kappa_L}\kappa_U^3\sigma_x^4\sigma_{\varepsilon}^2\sigma_z^2 \frac{t+p}{n_*^{\widehat\tau^2,|\eta|^2}} +\frac{\gamma^2}{\kappa_L}\kappa_U^3\sigma_x^6\sigma_{\varepsilon}^2\sigma_z^2  \left(\frac{p+\log(2|\mathcal{E}|)+t}{\overline{\sqrt{nN}}^{\widehat\tau,|\eta|}}\right)^2\right\}\\ 
\end{align*}
Next, note that if 
\begin{equation}\label{eq:sample_size_requirment2}
	\frac{n_*^{\widehat\tau^2,|\eta|^2}}{t+p} \geq \frac{18 C_1 \kappa_U^3 \sigma_x^4 \sigma_{\varepsilon}^2\sigma_z^2} {\left(\kappa_L^3 / \gamma\right) \mathbf{s}_{-}}\quad \text { and } \quad \frac{\overline{\sqrt{nN}}^{\widehat\tau,|\eta|}}{(p+\log(2|\mathcal{E}|)+t)}\geq \frac{\sqrt{18 C_1} \kappa_U^{3 / 2} \sigma_x^3 \sigma_{\varepsilon}\sigma_z}{\sqrt{\left(\kappa_L^3 / \gamma\right) \mathbf{s}_{-}}}, 
\end{equation}
then for any $\boldsymbol{\beta}$ with $\operatorname{supp}(\boldsymbol{\beta}) \cap G_{\boldsymbol{\omega}} \neq \emptyset$, we have the following inequality
\begin{align*}
	\widehat{\mathrm{Q}}_{\mathrm{Adj}}(\boldsymbol{\beta} ; \gamma, \boldsymbol{\omega})-\widehat{\mathrm{Q}}_{\mathrm{Adj}}\left(\boldsymbol{\beta}^* ; \gamma, \boldsymbol{\omega}\right)  &\ge \frac{\kappa_L}{4} \left\|\boldsymbol{\beta}-\boldsymbol{\beta}^*\right\|_2^2+\frac{\gamma}{6} \kappa_L^2 \overline{\mathrm{d}}_{\operatorname{supp}(\boldsymbol{\beta})} \\
	&\quad  \frac{-C_1\gamma^2}{\kappa_L}\kappa_U^3\sigma_x^4\sigma_{\varepsilon}^2\sigma_z^2 \frac{t+p}{n_*^{\widehat\tau^2,|\eta|^2}} +\frac{-C_1\gamma^2}{\kappa_L}\kappa_U^3\sigma_x^6\sigma_{\varepsilon}^2\sigma_z^2  \left(\frac{p+\log(2|\mathcal{E}|)+t}{\overline{\sqrt{nN}}^{\widehat\tau,|\eta|}}\right)^2\\ 
	&\ge \frac{\kappa_L}{4} \left\|\boldsymbol{\beta}-\boldsymbol{\beta}^*\right\|_2^2+\frac{\gamma}{6} \kappa_L^2 \overline{\mathrm{d}}_{\operatorname{supp}(\boldsymbol{\beta})} \\
	&\quad  \frac{-C_1\gamma^2}{\kappa_L}\kappa_U^3\sigma_x^4\sigma_{\varepsilon}^2\sigma_z^2 \frac {\left(\kappa_L^3 / \gamma\right) \mathbf{s}_{-}}{18 C_1 \kappa_U^3 \sigma_x^4 \sigma_{\varepsilon}^2\sigma_z^2}  +\frac{-C_1\gamma^2}{\kappa_L}\kappa_U^3\sigma_x^6\sigma_{\varepsilon}^2\sigma_z^2  \left( \frac{\sqrt{\left(\kappa_L^3 / \gamma\right) \mathbf{s}_{-}}}{\sqrt{18 C_1} \kappa_U^{3 / 2} \sigma_x^3 \sigma_{\varepsilon}\sigma_z}\right)^2\\
	&= \frac{\kappa_L}{4} \left\|\boldsymbol{\beta}-\boldsymbol{\beta}^*\right\|_2^2+\frac{\gamma}{6} \kappa_L^2 \overline{\mathrm{d}}_{\operatorname{supp}(\boldsymbol{\beta})} \\
	&\quad  \frac{-C_1\gamma^2}{\kappa_L}\kappa_U^3\sigma_x^4\sigma_{\varepsilon}^2\sigma_z^2 \frac {\left(\kappa_L^3 / \gamma\right) \mathbf{s}_{-}}{18 C_1 \kappa_U^3 \sigma_x^4 \sigma_{\varepsilon}^2\sigma_z^2}  +\frac{-C_1\gamma^2}{\kappa_L}\kappa_U^3\sigma_x^6\sigma_{\varepsilon}^2\sigma_z^2  \frac{\left(\kappa_L^3 / \gamma\right) \mathbf{s}_{-}}{18 C_1 \kappa_U^{3 } \sigma_x^6 \sigma_{\varepsilon}^2\sigma_z^2}\\
	&\stackrel{(a)}{\ge} \frac{\kappa_L}{4} \left\|\boldsymbol{\beta}-\boldsymbol{\beta}^*\right\|_2^2+\frac{\gamma}{18}  \kappa_L^2 \overline{\mathrm{d}}_{\operatorname{supp}(\boldsymbol{\beta})} \\
	&>0.
\end{align*}
Since $\operatorname{supp}(\boldsymbol{\beta}) \cap G_{\boldsymbol{\omega}} \neq \emptyset$, we have $\mathbf{s}_{-}=\min _{S \subseteq[p], S \cap G_\omega \neq \emptyset} \overline{\mathrm{d}}_S \le \overline{\mathrm{d}}_{\mathrm{supp}(\boldsymbol{\beta})}$, thus $-\mathbf{s}_{-} \ge -\overline{\mathrm{d}}_{\mathrm{supp}(\boldsymbol{\beta})}$, which implies $(a)$. Therefore, we have shown that with high probability, the support of the empirical risk minimizer $\widehat{\boldsymbol{\beta}}_{\mathrm{Adj}}$ must not overlap with the set $G_{\boldsymbol{\omega}}$, or the empirical risk minimizer must not contain any pooled linear spurious variable.  

Now we claim that $S^* \subseteq \operatorname{supp}(\widehat{\boldsymbol{\beta}}_{\mathrm{Adj}})$. Notice if we let 
\begin{equation}\label{eq:sample_size_requirment3}
	\frac{n_*^{\widehat{\tau}^2,|\eta|^2}}{t+p} \geq \frac{12 C_1 \gamma^2 \kappa_U^3 \sigma_x^4 \sigma_{\varepsilon}^2\sigma_z^2}{\kappa_L^2 \mathbf{s}_{+}}  \quad \text { and } \quad \frac{\overline{\sqrt{n N}}^{\widehat{\tau},|\eta|}}{(p+\log (2|\mathcal{E}|)+t)} \geq \frac{\sqrt{12 C_1} \gamma \kappa_U^{3 / 2} \sigma_x^3 \sigma_{\varepsilon}\sigma_z}{\sqrt{\kappa_L^2 \mathbf{s}_{+}}},
\end{equation}
then for any $\boldsymbol{\beta}$ with $S^* \nsubseteq \operatorname{supp}(\boldsymbol{\beta})$, we have the following
\begin{align*}
	\widehat{\mathrm{Q}}_{\mathrm{Adj}}(\boldsymbol{\beta} ; \gamma, \boldsymbol{\omega})-\widehat{\mathrm{Q}}_{\mathrm{Adj}}\left(\boldsymbol{\beta}^* ; \gamma, \boldsymbol{\omega}\right) &\ge \frac{\kappa_L}{4} \left\|\boldsymbol{\beta}-\boldsymbol{\beta}^*\right\|_2^2+\frac{\gamma}{6} \kappa_L^2 \overline{\mathrm{d}}_{\operatorname{supp}(\boldsymbol{\beta})} \\
	&\quad  \frac{-C_1\gamma^2}{\kappa_L}\kappa_U^3\sigma_x^4\sigma_{\varepsilon}^2\sigma_z^2 \frac{t+p}{n_*^{\widehat\tau^2,|\eta|^2}} +\frac{-C_1\gamma^2}{\kappa_L}\kappa_U^3\sigma_x^6\sigma_{\varepsilon}^2\sigma_z^2  \left(\frac{p+\log(2|\mathcal{E}|)+t}{\overline{\sqrt{nN}}^{\widehat\tau,|\eta|}}\right)^2\\ 
	&\ge \frac{\kappa_L}{4} \left\|\boldsymbol{\beta}-\boldsymbol{\beta}^*\right\|_2^2+\frac{\gamma}{6} \kappa_L^2 \overline{\mathrm{d}}_{\operatorname{supp}(\boldsymbol{\beta})} \\
	&\quad  \frac{-C_1\gamma^2}{\kappa_L}\kappa_U^3\sigma_x^4\sigma_{\varepsilon}^2\sigma_z^2  \frac{\kappa_L^2 \mathbf{s}_{+}}{12 C_1 \gamma^2 \kappa_U^3 \sigma_x^4 \sigma_{\varepsilon}^2\sigma_z^2}  +\frac{-C_1\gamma^2}{\kappa_L}\kappa_U^3\sigma_x^6\sigma_{\varepsilon}^2\sigma_z^2  \left(\frac{\sqrt{\kappa_L^2 \mathbf{s}_{+}}}{\sqrt{12 C_1} \gamma \kappa_U^{3 / 2} \sigma_x^3 \sigma_{\varepsilon}\sigma_z} \right)^2\\ 
	&\ge \frac{\kappa_L}{4} \left\|\boldsymbol{\beta}-\boldsymbol{\beta}^*\right\|_2^2+\frac{\gamma}{6} \kappa_L^2 \overline{\mathrm{d}}_{\operatorname{supp}(\boldsymbol{\beta})} \\
	&\quad  \frac{-C_1\gamma^2}{\kappa_L}\kappa_U^3\sigma_x^4\sigma_{\varepsilon}^2\sigma_z^2  \frac{\kappa_L^2 \mathbf{s}_{+}}{12 C_1 \gamma^2 \kappa_U^3 \sigma_x^4 \sigma_{\varepsilon}^2\sigma_z^2}  +\frac{-C_1\gamma^2}{\kappa_L}\kappa_U^3\sigma_x^6\sigma_{\varepsilon}^2\sigma_z^2  \frac{\kappa_L^2 \mathbf{s}_{+}}{12 C_1 \gamma^2 \kappa_U^{3 } \sigma_x^6 \sigma_{\varepsilon}^2\sigma_z^2} \\ 
	&\ge\frac{\kappa_L}{4} \left\|\boldsymbol{\beta}-\boldsymbol{\beta}^*\right\|_2^2+\frac{\gamma}{6} \kappa_L^2 \overline{\mathrm{d}}_{\operatorname{supp}(\boldsymbol{\beta})} -\frac{\kappa_L \mathbf{s}_{+}}{6}   \\ 
	&\ge\frac{\kappa_L}{4} \left\|\boldsymbol{\beta}-\boldsymbol{\beta}^*\right\|_2^2-\frac{\kappa_L\mathbf{s}_{+} }{6} \\
	&=\frac{\kappa_L}{4} \left\|\boldsymbol{\beta}-\boldsymbol{\beta}^*\right\|_2^2-\frac{\kappa_L}{6}\min _{j \in S^*}\left|\beta_j^*\right|^2 \\
	&\stackrel{(b)}{\ge}\frac{\kappa_L}{4} \min _{j \in S^*}\left|\beta_j^*\right|^2-\frac{\kappa_L }{6}\min _{j \in S^*}\left|\beta_j^*\right|^2 \\
	&=\frac{\kappa_L}{12} \min _{j \in S^*}\left|\beta_j^*\right|^2\\
	&>0,
\end{align*}
where $(b)$ holds by \ref{eq:I4_inequality} and the fact we are focusing on any $\boldsymbol{\beta}$ with $S^* \nsubseteq \operatorname{supp}(\boldsymbol{\beta})$, or $\left|S^* \backslash \operatorname{supp}(\boldsymbol{\beta})\right| >0$. Therefore, we have shown that the support of the empirical risk minimizer must contain the full set $S^*$. 

Combining the conditions \ref{eq:sample_size_requirment1}, \ref{eq:sample_size_requirment2}, \ref{eq:sample_size_requirment3} together, we have 
\begin{equation}
	\frac{\bar{n}^{|\eta|^2}}{(\log (4|\mathcal{E}|)+t+p)}\ge (18C_1\kappa_U^{3/2}\sigma_x^3\sigma_{\varepsilon}^2\sigma_z^2 \vee 16 C \kappa_U \sigma_x^2\sigma_{\varepsilon}^2) \frac{\gamma}{\kappa_L}\left(\frac{1}{\sqrt{\mathbf{s}_{+} \wedge (\gamma \kappa_L \mathbf{s}_{-} )}} + \frac{1}{\mathbf{s}_{+}}\right)
	\label{eq:sample_size_requirment4}
	\end{equation}
	and 
	\begin{equation}
	\frac{n_*^{\widehat{\tau}^2,|\eta|^2}}{t+p} \geq (18C_1 \kappa_U^3\sigma_x^4\sigma_{\varepsilon}^2\sigma_z^2 \vee 16C^2 \kappa_U^4 \sigma_x^4 ) \left(\frac{\gamma}{\kappa_L}\right)^2\left\{\frac{1}{(\gamma\kappa_L \mathbf{s}_{-}) \wedge \mathbf{s}_{+}} + 1   \right\}.
	\label{eq:sample_size_requirment5}
\end{equation}
Recall that $ \eta_{\max} = \max_{e\in\mathcal{E}} |\eta^{(e)}|$. Notice sufficient conditions of \ref{eq:sample_size_requirment4} and \ref{eq:sample_size_requirment5} are
\begin{equation}
	\frac{\bar{n}}{(\log (4|\mathcal{E}|)+t+p)}\ge \eta^2_{\max}  (18C_1\kappa_U^{3/2}\sigma_x^3\sigma_{\varepsilon}^2\sigma_z^2 \vee 16 C \kappa_U \sigma_x^2\sigma_{\varepsilon}^2) \frac{\gamma}{\kappa_L}\left(\frac{1}{\sqrt{\mathbf{s}_{+} \wedge (\gamma \kappa_L \mathbf{s}_{-} )}} + \frac{1}{\mathbf{s}_{+}}\right)
	\label{eq:sample_size_requirment6}
	\end{equation}
	and 
	\begin{equation}
	\frac{n_*^{\widehat{\tau}^2}}{t+p} \geq \eta^2_{\max}  (18C_1 \kappa_U^3\sigma_x^4\sigma_{\varepsilon}^2\sigma_z^2 \vee 16C^2 \kappa_U^4 \sigma_x^4 ) \left(\frac{\gamma}{\kappa_L}\right)^2\left\{\frac{1}{(\gamma\kappa_L \mathbf{s}_{-}) \wedge \mathbf{s}_{+}} + 1   \right\}.
	\label{eq:sample_size_requirment7}
\end{equation}
respectively. Here we see that $\eta_{\max} = o(\sqrt{\bar{n} \wedge n_*^{\widehat{\tau}^2}} )$ is necessary for \ref{eq:sample_size_requirment6} and \ref{eq:sample_size_requirment7} to hold. 

In summary, under event $\mathcal{A}_{1, t} \cap \mathcal{A}_{2, t}$, which happen with probability at least $1-37e^{-t}$ by Corollary \ref{corollary:R_four_cases} and \ref{corollary:oneside_boundJ_high_missing_imprecise_imp}, we have 
\begin{equation}\label{eq:proof_vsc_property_hold}
S^* \subseteq \operatorname{supp}\left(\widehat{\boldsymbol{\beta}}_{\mathrm{Adj}}\right) \subseteq G_{\boldsymbol{\omega}}^c.
\end{equation}

Next, we prove the variable selection property under the case of large missingness and accurate imputation. In this setting, $\widehat\tau_{\min} > 0.618$ and $\eta_{\max}< 1$. Under this setting, the event $\mathcal{A}_{1, t}$ correspond to the event $\mathcal{A}_{1, t, Case 4}$ in Corollary \ref{corollary:R_four_cases}, and the event $\mathcal{A}_{2, t}$ has the property listed in Corollary \ref{corollary:oneside_boundJ_high_missing_good_imp}. 

The structure of the proof is similar as the previous case $\eta_{\max} \ge 1$. Using the fact that $n_*^{\hat{\tau}} \le n_*^{\hat{\tau}^2}$, we can impose the following conditions on the sample size to ensure the desired variable selection consistency. 
\begin{equation}\label{eq:sample_size_requirment8}
\frac{\bar{n}}{(\log (4|\mathcal{E}|)+t+p)}\ge (18C_1\kappa_U^{3/2}\sigma_x^3\sigma_{\varepsilon}^2\sigma_z^2 \vee 16 C \kappa_U \sigma_x^2\sigma_{\varepsilon}^2) \frac{\gamma}{\kappa_L}\left(\frac{1}{\sqrt{\mathbf{s}_{+} \wedge (\gamma \kappa_L \mathbf{s}_{-} )}} + \frac{1}{\mathbf{s}_{+}}\right)
\end{equation}
and 
\begin{equation}\label{eq:sample_size_requirment9}
\frac{n_*^{\hat{\tau}}}{t+p} \geq (18C_1 \kappa_U^3\sigma_x^4\sigma_{\varepsilon}^2\sigma_z^2 \vee 16C^2 \kappa_U^4 \sigma_x^4 ) \left(\frac{\gamma}{\kappa_L}\right)^2\left\{\frac{1}{(\gamma\kappa_L \mathbf{s}_{-}) \wedge \mathbf{s}_{+}} + 1   \right\}.
\end{equation}
This completes the proof. 

\qed

\subsection{Proof of Theorem \ref{theorem:nonasy_l2_error}}

\subsubsection{Proof of the Rate (\ref{eq:large_eta_non-asymptotic_l2_error_bound_rate_1}) and (\ref{eq:small_eta_non-asymptotic_l2_error_bound_rate_1})}

Under high missingness setting, i.e. $\widehat\tau_{\min} > 0.618$, we first consider the case of having imprecise imputation, i.e. $\eta_{\max} \geq 1$, and prove the rate (\ref{eq:large_eta_non-asymptotic_l2_error_bound_rate_1}). Under this setting, the event $\mathcal{A}_{1, t}$ correspond to the event $\mathcal{A}_{1, t, Case 3}$ in Corollary \ref{corollary:R_four_cases}, and the event $\mathcal{A}_{2, t}$ has the property listed in Corollary \ref{corollary:oneside_boundJ_high_missing_imprecise_imp}. 

Conditional on $\mathcal{A}_{1, t} \cap \mathcal{A}_{2, t}$, the upper bound (\ref{eq:decomposition_C12345}) holds. Follow the upper bounds (\ref{eq:I2_ub}) for $I_2$  and (\ref{eq:I3_ub}) for  $I_3$, we now give a different bound on $I_4$ than (\ref{eq:I4_ub1}).
Let $S = \text{supp}(\boldsymbol{\beta})$, observe
\begin{align*}
I_4 &= \left|S^* \backslash S\right| \kappa_U \sigma_x^2 \sigma^2_{\varepsilon} \sigma_z^2\frac{\log \left(4|\mathcal{E}|\left|S^*\right|\right)+t}{\bar{n}^{|\eta|^2}}\\
&= \left\{\left|S^* \backslash S\right|\min _{j \in S^*}\left|\beta_j^*\right|\sqrt{\kappa_L / 16 C \gamma}\sqrt{1/|S^*|}\right\}\left\{\frac{\sqrt{|S^*|}\kappa_U \sigma_x^2 \sigma^2_{\varepsilon} \sigma_z^2}{\sqrt{\kappa_L / 16 C \gamma}\min _{j \in S^*}\left|\beta_j^*\right|} \frac{\log \left(4|\mathcal{E}|\left|S^*\right|\right)+t}{\bar{n}^{|\eta|^2}}\right\}\\
&\le\frac{\kappa_L}{16 C \gamma}\frac{\left|S^* \backslash S\right|}{|S^*|}\left|S^* \backslash S\right| \min _{j \in S^*}\left|\boldsymbol{\beta}_j^*\right|^2+ \frac{16 C \gamma \kappa_U^2 \sigma_x^4 \sigma_{\varepsilon}^4\sigma_z^4|S^*|\left\{\log \left(4|\mathcal{E}|\left|S^*\right|\right)+t\right\}^2}{\kappa_L\left(\min _{j \in S^*}\left|\beta_j^*\right|^2\right) (\bar{n}^{|\eta|^2})^2}\\
&\le\frac{\kappa_L}{16 C \gamma}\|\boldsymbol{\beta} - \boldsymbol{\beta}^*\|_2^2+ 16 C  \kappa_U^2 \sigma_x^4 \sigma_{\varepsilon}^4\sigma_z^4|S^*|\frac{\gamma/\kappa_L}{\min _{j \in S^*}\left|\beta_j^*\right|^2}\frac{\left\{\log \left(4|\mathcal{E}|\left|S^*\right|\right)+t\right\}^2}{ (\bar{n}^{|\eta|^2})^2}.
\end{align*}

Meanwhile, let $c_1 = (16 C \kappa_U^2 \sigma_x^2)^2$, and let $N_* \geq c_1(\gamma/\kappa_L)^2(p+t)$, then $I_1 \le \frac{\kappa_L}{16 C \gamma}\left\|\boldsymbol{\beta}-\boldsymbol{\beta}^*\right\|_2^2$. In addition, assume $\gamma>1+\kappa_L$ and $n_*^{\widehat{\tau}, |\eta|} \geq n_*^{\widehat{\tau}^2, |\eta|^2}$. For all the $\boldsymbol{\beta} \in \mathbb{R}^p$, 
\begin{align*}
	\mathcal{D}_{\mathrm{R}, \widehat{\mathrm{R}}_{\mathrm{Adj}}}(\boldsymbol{\beta}) + \gamma \mathcal{D}_{\mathrm{J}, \widehat{J}_{\mathrm{Adj}}}(\boldsymbol{\beta}) &\le C\gamma \left\{\kappa_U^2 \sigma_x^2\left\|\boldsymbol{\beta}-\boldsymbol{\beta}^*\right\|_2^2\sqrt{\frac{t+p}{N_*}}  \right.\notag\\
	&\quad + \left.  \kappa_U^{3 / 2}  \sigma_x^2 \sigma_{\varepsilon}\sigma_z \left\|\boldsymbol{\beta}-\boldsymbol{\beta}^*\right\|_2\left\{\sqrt{\frac{t+p}{n_*^{\hat\tau,|\eta|}}} +\sigma_x \frac{p+\log(2|\mathcal{E}|)+t}{\overline{\sqrt{nN}}^{\hat\tau,|\eta|}}\right\} \right.\notag\\
	&\quad+ \left. \kappa_U^{1 / 2} \sigma_x \sigma_{\varepsilon} \sigma_z \sqrt{\frac{t+p}{n^{\hat\tau^2,|\eta|^2}_*}} \times \sqrt{\sum_{e \in \mathcal{E}} \omega^{(e)}\left\|\mathbb{E}\left[x_S^{(e)} \varepsilon^{(e)}\right]\right\|_2^2} \right.\notag\\
	&\quad+ \left. \left|S^* \backslash S\right| \kappa_U \sigma_x^2 \sigma^2_{\varepsilon} \sigma_z^2\frac{\log \left(4|\mathcal{E}|\left|S^*\right|\right)+t}{\bar{n}^{|\eta|^2}}+ \kappa_U \sigma_x\sigma^2_{\varepsilon}\sigma_z\frac{t+p}{n^{\hat\tau,|\eta|}_*}\right\}\notag\\
	&\le C\gamma \left\{ \frac{\kappa_L}{16 C \gamma}\left\|\boldsymbol{\beta}-\boldsymbol{\beta}^*\right\|_2^2 \right.\\
	&\quad + \left.  \frac{\kappa_L}{16 C\gamma} \left\|\boldsymbol{\beta}-\boldsymbol{\beta}^*\right\|_2^2 +  \frac{8C\gamma}{\kappa_L}\kappa_U^3\sigma_x^4\sigma_{\varepsilon}^2\sigma_z^2  \left\{\frac{t+p}{n_*^{\hat\tau,|\eta|}} + \sigma_x^2\left(\frac{p+\log(2|\mathcal{E}|)+t}{\overline{\sqrt{nN}}^{\hat\tau,|\eta|}}\right)^2\right\} \right.\\
	&\quad+ \left. 6C\kappa_U\sigma_x^2 \sigma_{\varepsilon}^2 \sigma_z^2 \frac{t+p}{n^{\hat\tau^2,|\eta|^2}_*} +\frac{1}{6C}\mathrm{~J}(\boldsymbol{\beta} ; \boldsymbol{\omega}) + \frac{16C\gamma}{\kappa_L}\kappa_U^{3} \sigma_x^2 \sigma_{\varepsilon}^2 \sigma_z^2 \frac{t+p}{n^{\hat\tau^2,|\eta|^2}_*}+\frac{\kappa_L}{16C\gamma}\left\|\boldsymbol{\beta}-\boldsymbol{\beta}^{\star}\right\|_2^2 \right.\\
	&\quad+ \left. \frac{\kappa_L}{16 C \gamma}\|\boldsymbol{\beta} - \boldsymbol{\beta}^*\|_2^2+ 16 C  \kappa_U^2 \sigma_x^4 \sigma_{\varepsilon}^4\sigma_z^4|S^*|\frac{\gamma/\kappa_L}{\min _{j \in S^*}\left|\beta_j^*\right|^2}\frac{\left\{\log \left(4|\mathcal{E}|\left|S^*\right|\right)+t\right\}^2}{ (\bar{n}^{|\eta|^2})^2}+\kappa_U \sigma_x\sigma^2_{\varepsilon}\sigma_z\frac{t+p}{n^{\hat\tau,|\eta|}_*}\right\}\\
	&\le C\gamma \left\{ \frac{4\kappa_L}{16 C \gamma}\left\|\boldsymbol{\beta}-\boldsymbol{\beta}^*\right\|_2^2+\frac{1}{6C}\mathrm{~J}(\boldsymbol{\beta} ; \boldsymbol{\omega}) \right.\\
	&\quad  \left.   +  \frac{8C\gamma}{\kappa_L}\kappa_U^3\sigma_x^4\sigma_{\varepsilon}^2\sigma_z^2  \left\{\frac{t+p}{n_*^{\hat\tau,|\eta|}} + \sigma_x^2\left(\frac{p+\log(2|\mathcal{E}|)+t}{\overline{\sqrt{nN}}^{\hat\tau,|\eta|}}\right)^2\right\} \right.\\
	&\quad+ \left. 6C\kappa_U\sigma_x^2 \sigma_{\varepsilon}^2 \sigma_z^2 \frac{t+p}{n^{\hat\tau^2,|\eta|^2}_*}  + \frac{16C\gamma}{\kappa_L}\kappa_U^{3} \sigma_x^2 \sigma_{\varepsilon}^2 \sigma_z^2 \frac{t+p}{n^{\hat\tau^2,|\eta|^2}_*} +\kappa_U \sigma_x\sigma^2_{\varepsilon}\sigma_z\frac{t+p}{n^{\hat\tau,|\eta|}_*}\right.\\
	&\quad \left.+ 16 C  \kappa_U^2 \sigma_x^4 \sigma_{\varepsilon}^4\sigma_z^4|S^*|\frac{\gamma/\kappa_L}{\min _{j \in S^*}\left|\beta_j^*\right|^2}\frac{\left\{\log \left(4|\mathcal{E}|\left|S^*\right|\right)+t\right\}^2}{ (\bar{n}^{|\eta|^2})^2}\right\}\\
	&\le \left\{ \frac{\kappa_L}{4}\left\|\boldsymbol{\beta}-\boldsymbol{\beta}^*\right\|_2^2+\frac{\gamma}{6}\mathrm{~J}(\boldsymbol{\beta} ; \boldsymbol{\omega}) \right.\\
	&\quad  \left.   + C_1 \frac{\gamma^2}{\kappa_L}\kappa_U^3\sigma_x^4\sigma_{\varepsilon}^2\sigma_z^2  \left\{\frac{t+p}{n_*^{\hat\tau^2,|\eta|^2}} + \sigma_x^2\left(\frac{p+\log(2|\mathcal{E}|)+t}{\overline{\sqrt{nN}}^{\hat\tau,|\eta|}}\right)^2\right\} \right.\\
	&\quad \left.+ C_1\frac{\gamma^2}{\kappa_L}  \kappa_U^2 \sigma_x^4 \sigma_{\varepsilon}^4\sigma_z^4\frac{|S^*|}{\min _{j \in S^*}\left|\beta_j^*\right|^2}\left(\frac{\log \left(4|\mathcal{E}|\left|S^*\right|\right)+t}{ \bar{n}^{|\eta|^2}}\right)^2\right\}
\end{align*}
Next, observe 
\begin{align*}
\frac{\kappa_L}{2}\left\|\widehat{\boldsymbol{\beta}}_{\mathrm{Adj}}-\boldsymbol{\beta}^*\right\|_2^2+\frac{\gamma}{6} \mathrm{~J}(\widehat{\boldsymbol{\beta}}_{\mathrm{Adj}} ; \boldsymbol{\omega}) &\stackrel{(a)}{\le} \mathrm{Q}(\widehat{\boldsymbol{\beta}}_{\mathrm{Adj}} ; \gamma, \boldsymbol{\omega})-\mathrm{Q}\left(\boldsymbol{\beta}^* ; \gamma, \boldsymbol{\omega}\right) \\
&\stackrel{(b)}{=}\mathcal{D}_{\mathrm{R}, \widehat{\mathrm{R}}_{\mathrm{Adj}}}(\boldsymbol{\beta}) + \gamma \mathcal{D}_{\mathrm{J}, \widehat{J}_{\mathrm{Adj}}}(\boldsymbol{\beta})+\widehat{\mathrm{Q}}_{\mathrm{Adj}}({\widehat{\boldsymbol{\beta}}_{\mathrm{Adj}}} ; \gamma, \boldsymbol{\omega})-\widehat{\mathrm{Q}}_{\mathrm{Adj}}\left(\boldsymbol{\beta}^* ; \gamma, \boldsymbol{\omega}\right)\\
&\le \mathcal{D}_{\mathrm{R}, \widehat{\mathrm{R}}_{\mathrm{Adj}}}(\boldsymbol{\beta}) + \gamma \mathcal{D}_{\mathrm{J}, \widehat{J}_{\mathrm{Adj}}}(\boldsymbol{\beta})\\
&\le \frac{\kappa_L}{4}\left\|\widehat{\boldsymbol{\beta}}_{\mathrm{Adj}}-\boldsymbol{\beta}^*\right\|_2^2+\frac{\gamma}{6} \mathrm{~J}(\widehat{\boldsymbol{\beta}}_{\mathrm{Adj}} ; \boldsymbol{\omega}) \\
&\quad +C_1 \frac{\gamma^2}{\kappa_L} \kappa_U^3 \sigma_x^4 \sigma_{\varepsilon}^2 \sigma_z^2\left\{\frac{t+p}{n_*^{\hat{\tau}^2,|\eta|^2}}+\sigma_x^2\left(\frac{p+\log (2|\mathcal{E}|)+t}{\overline{\sqrt{nN}}^{\hat\tau,|\eta|}}\right)^2\right\} \\
&\quad +C_1 \frac{\gamma^2}{\kappa_L} \kappa_U^2 \sigma_x^4 \sigma_{\varepsilon}^4 \sigma_z^4 \frac{|S^*|}{\min _{j \in S^*}\left|\beta_j^*\right|^2}\left(\frac{\log \left(4|\mathcal{E}|\left|S^*\right|\right)+t}{\bar{n}^{|\eta|^2}}\right)^2,
\end{align*}
where $(a)$ holds by Proposition \ref{prop:lower_bound_Qbeta_Qbetastar} and $(b)$ holds by the decomposition (\ref{eq:decomQhatQ}). 
Therefore, we have 
\begin{align*}
\frac{\kappa_L}{2}\left\|\widehat{\boldsymbol{\beta}}_{\mathrm{Adj}}-\boldsymbol{\beta}^*\right\|_2^2+\frac{\gamma}{6} \mathrm{~J}(\widehat{\boldsymbol{\beta}}_{\mathrm{Adj}} ; \boldsymbol{\omega})  &\le \frac{\kappa_L}{4}\left\|\widehat{\boldsymbol{\beta}}_{\mathrm{Adj}}-\boldsymbol{\beta}^*\right\|_2^2+\frac{\gamma}{6} \mathrm{~J}(\widehat{\boldsymbol{\beta}}_{\mathrm{Adj}} ; \boldsymbol{\omega}) \\
&\quad +C_1 \frac{\gamma^2}{\kappa_L} \kappa_U^3 \sigma_x^4 \sigma_{\varepsilon}^2 \sigma_z^2\left\{\frac{t+p}{n_*^{\hat{\tau}^2,|\eta|^2}}+\sigma_x^2\left(\frac{p+\log (2|\mathcal{E}|)+t}{\overline{\sqrt{nN}}^{\hat\tau,|\eta|}}\right)^2\right\} \\
&\quad +C_1 \frac{\gamma^2}{\kappa_L} \kappa_U^2 \sigma_x^4 \sigma_{\varepsilon}^4 \sigma_z^4 \frac{|S^*|}{\min _{j \in S^*}\left|\beta_j^*\right|^2}\left(\frac{\log \left(4|\mathcal{E}|\left|S^*\right|\right)+t}{\bar{n}^{|\eta|^2}}\right)^2,
\end{align*}
which leads to 
\begin{align}
\left\|\widehat{\boldsymbol{\beta}}_{\mathrm{Adj}}-\boldsymbol{\beta}^*\right\|_2^2  &\le 4C_1\kappa_U^3 \sigma_x^6 \sigma_{\varepsilon}^2 \sigma_z^2 \frac{\gamma^2}{\kappa_L^2} \left\{\frac{t+p}{n_*^{\hat{\tau}^2,|\eta|^2}}+\left(\frac{p+\log (2|\mathcal{E}|)+t}{\overline{\sqrt{nN}}^{\hat\tau,|\eta|}}\right)^2\right\} \notag\\
&\quad +4C_1 \kappa_U^2 \sigma_x^4 \sigma_{\varepsilon}^4 \sigma_z^4 \frac{\gamma^2}{\kappa_L^2} \frac{ |S^*|}{\min _{j \in S^*}\left|\beta_j^*\right|^2}\left(\frac{\log \left(4|\mathcal{E}|\left|S^*\right|\right)+t}{\bar{n}^{|\eta|^2}}\right)^2\notag\\
&\le 4C_1\kappa_U^3 \sigma_x^6 \sigma_{\varepsilon}^2 \sigma_z^2 \frac{\gamma^2}{\kappa_L^2} \left\{\frac{t+p}{n_*^{\hat{\tau}^2,|\eta|^2}}+\left(\frac{p+\log (2|\mathcal{E}|)+t}{\bar{n}^{|\eta|^2}}\right)^2\right\}\notag \\
&\quad +4C_1 \kappa_U^2 \sigma_x^4 \sigma_{\varepsilon}^4 \sigma_z^4  \frac{\gamma^2}{\kappa_L^2}\frac{|S^*|}{\min _{j \in S^*}\left|\beta_j^*\right|^2}\left(\frac{\log \left(4|\mathcal{E}|\left|S^*\right|\right)+t}{\bar{n}^{|\eta|^2}}\right)^2\notag\\
&\le 4C_1\kappa_U^3 \sigma_x^6 \sigma_{\varepsilon}^4 \sigma_z^4 \frac{\gamma^2}{\kappa_L^2} \left(\frac{(t+p)\eta_{\max}^2}{n_*^{\hat{\tau}^2}}+\frac{\{p+\log (2|\mathcal{E}|)+t\}^2\eta_{\max}^4 }{\bar{n}^2}\right) \notag\\
&\quad +4C_1 \kappa_U^2 \sigma_x^4 \sigma_{\varepsilon}^4 \sigma_z^4 \frac{\gamma^2}{\kappa_L^2} \frac{ |S^*|}{\min _{j \in S^*}\left|\beta_j^*\right|^2}\frac{\{\log \left(4|\mathcal{E}|\left|S^*\right|\right)+t\}^2\eta_{\max}^4}{\bar{n}^2}.
\label{eq:sufficeint_l2_error_bound1}
\end{align}
Here we see that $\eta_{\max }=o \left(\sqrt{\bar{n} \wedge n_*^{\hat{\tau}^2}}\right)$ is necessary for the upper bound in (\ref{eq:sufficeint_l2_error_bound1}) to converge. 

Now take the square root of both sides, we get 
\begin{align*}
\left\|\widehat{\boldsymbol{\beta}}_{\mathrm{Adj}}-\boldsymbol{\beta}^*\right\|_2  &\le \sqrt{4C_1\kappa_U^3 \sigma_x^6 \sigma_{\varepsilon}^4 \sigma_z^4} \frac{\gamma}{\kappa_L}  \sqrt{\frac{(t+p)\eta_{\max}^2}{n_*^{\hat{\tau}^2}}+\frac{\{p+\log (2|\mathcal{E}|)+t\}^2\eta_{\max}^4 }{\bar{n}^2}} \\
&\quad +\sqrt{4C_1 \kappa_U^2 \sigma_x^4 \sigma_{\varepsilon}^4 \sigma_z^4} \frac{\gamma}{\kappa_L}\sqrt{ \frac{ |S^*|}{\min _{j \in S^*}\left|\beta_j^*\right|^2}\frac{\{\log \left(4|\mathcal{E}|\left|S^*\right|\right)+t\}^2\eta_{\max}^4}{\bar{n}^2}}\\
&\le \sqrt{4C_1\kappa_U^3 \sigma_x^6} \sigma_{\varepsilon}^2 \sigma_z^2 \frac{\gamma}{\kappa_L} \left( \sqrt{\frac{(t+p)\eta_{\max}^2}{n_*^{\hat{\tau}^2}}}+\frac{(p+\log (2|\mathcal{E}|)+t)\eta_{\max}^2 }{\bar{n}} \right)\\
&\quad +\sqrt{4C_1 \kappa_U^2 \sigma_x^4} \sigma_{\varepsilon}^2 \sigma_z^2 \frac{\gamma}{\kappa_L}\sqrt{ \frac{ |S^*|}{\min _{j \in S^*}\left|\beta_j^*\right|^2}}\frac{(\log \left(4|\mathcal{E}|\left|S^*\right|\right)+t)\eta_{\max}^2}{\bar{n}}.
\end{align*}
Setting $c_{2} = \sqrt{4C_1\kappa_U^3 \sigma_x^6}$ and $c_{3} =\sqrt{4C_1 \kappa_U^2 \sigma_x^4}$, we have 
\begin{align*}
\frac{\|\widehat{\boldsymbol{\beta}}_{\mathrm{Adj}}-\boldsymbol{\beta}^*\|_2}{\sigma_{\varepsilon}^2 \sigma_z^2 (\gamma/\kappa_L)}&\le c_{2}\left( \sqrt{\frac{(t+p)\eta_{\max}^2}{n_*^{\hat{\tau}^2}}}+\frac{(p+\log (2|\mathcal{E}|)+t)\eta_{\max}^2 }{\bar{n}} \right)\\
&\quad + c_{3}\frac{\sqrt{|S^*|}(\log \left(4|\mathcal{E}|\left|S^*\right|\right)+t)\eta_{\max}^2}{(\min _{j \in S^*}\left|\beta_j^*\right|)\bar{n}}.
\end{align*}

We now discuss the rate (\ref{eq:small_eta_non-asymptotic_l2_error_bound_rate_1}). The proof for the case precise imputation $\eta_{\max} < 1$ follows a similar methodology, with the event $\mathcal{A}_{1, t}$ correspond to the event $\mathcal{A}_{1, t, Case 4}$ in Corollary \ref{corollary:R_four_cases}, and the event $\mathcal{A}_{2, t}$ has the property listed in Corollary \ref{corollary:oneside_boundJ_high_missing_good_imp}. Moreover, we have $\eta_{\max} < 1$. Using the fact that $n_*^{\hat{\tau}} \le n_*^{\hat{\tau}^2}$. Together, we obtain the following upper bound:
\begin{equation}\label{eq:l2errorbound_precise_impu}
\frac{\|\widehat{\boldsymbol{\beta}}_{\mathrm{Adj}}-\boldsymbol{\beta}^*\|_2}{\sigma_{\varepsilon}^2 \sigma_z^2 (\gamma/\kappa_L)}\le c_{2}\left( \sqrt{\frac{t+p}{n_*^{\hat{\tau}}}}+\frac{p+\log (2|\mathcal{E}|)+t }{\bar{n}} \right)+ c_{3}\frac{\sqrt{|S^*|}(\log \left(4|\mathcal{E}|\left|S^*\right|\right)+t)}{(\min _{j \in S^*}\left|\beta_j^*\right|)\bar{n}}.
\end{equation}

\subsubsection{Proof of the Rate (\ref{eq:large_eta_non-asymptotic_l2_error_bound_rate_2}) and (\ref{eq:small_eta_non-asymptotic_l2_error_bound_rate_2})}

Now we in additional assume the additional conditions in Theorem \ref{theorem:nonasy_vsc} hold. We first prove the rate (\ref{eq:large_eta_non-asymptotic_l2_error_bound_rate_2}). We assume that conditions (\ref{eq:sample_size_requirment6}) and (\ref{eq:sample_size_requirment7}) are satisfied, so the variable selection property (\ref{eq:proof_vsc_property_hold}) holds. 

Define $p_0 = |(G_{\boldsymbol{\omega}})^c|\le p$ to be the dimension of covariates without any linear spurious variables, and we apply Corollary \ref{corollary:R_four_cases} and Corollary \ref{corollary:oneside_boundJ_high_missing_imprecise_imp} with $\boldsymbol{x}_{(G_{\boldsymbol{\omega}})^c}$. We define the event $\mathcal{A}_{1, t}$ correspond to the event $\mathcal{A}_{1, t, Case 3}$ in Corollary \ref{corollary:R_four_cases}, and the event $\mathcal{A}_{2, t}$ has the property listed in Corollary \ref{corollary:oneside_boundJ_high_missing_imprecise_imp}. 

When $\mathcal{A}_{1, t} \cap \mathcal{A}_{2, t} \cap \mathcal{A}_{1, t}^{p_0} \cap \mathcal{A}_{2, t}^{p_0}$ occurs, the decomposition (\ref{eq:decomposition_C12345})  with $p=p_0$ holds for $\boldsymbol{\beta} = \widehat{\boldsymbol{\beta}}_{\mathrm{Adj}}$. Notice the term $I_4=0$ because of the variable selection property (\ref{eq:proof_vsc_property_hold}). Similar as the proof of rate (\ref{eq:large_eta_non-asymptotic_l2_error_bound_rate_1}), we assume $N_* \geq c_1(\gamma/\kappa_L)^2(p+t)$. To that end we have 
\begin{align*}
	\mathcal{D}_{\mathrm{R}, \widehat{\mathrm{R}}_{\mathrm{Adj}}}(\boldsymbol{\beta}) + \gamma \mathcal{D}_{\mathrm{J}, \widehat{J}_{\mathrm{Adj}}}(\boldsymbol{\beta})&\le C\gamma \left\{\kappa_U^2 \sigma_x^2\left\|\widehat{\boldsymbol{\beta}}_{\mathrm{Adj}}-\boldsymbol{\beta}^*\right\|_2^2\sqrt{\frac{t+p}{N_*}}  \right.\notag\\
	&\quad + \left.  \kappa_U^{3 / 2}  \sigma_x^2 \sigma_{\varepsilon}\sigma_z \left\|\widehat{\boldsymbol{\beta}}_{\mathrm{Adj}}-\boldsymbol{\beta}^*\right\|_2\left\{\sqrt{\frac{t+p}{n_*^{\hat\tau,|\eta|}}} +\sigma_x \frac{p+\log(2|\mathcal{E}|)+t}{\overline{\sqrt{nN}}^{\hat\tau,|\eta|}}\right\} \right.\notag\\
	&\quad+ \left. \kappa_U^{1 / 2} \sigma_x \sigma_{\varepsilon} \sigma_z \sqrt{\frac{t+p}{n^{\hat\tau^2,|\eta|^2}_*}} \times \sqrt{\sum_{e \in \mathcal{E}} \omega^{(e)}\left\|\mathbb{E}\left[x_{\mathrm{supp}(\widehat{\boldsymbol{\beta}}_{\mathrm{Adj}})}^{(e)} \varepsilon^{(e)}\right]\right\|_2^2} \right.\notag\\
	&\quad+ \left.  \kappa_U \sigma_x\sigma^2_{\varepsilon}\sigma_z\frac{t+p}{n^{\hat\tau,|\eta|}_*}\right\}\notag\\
	&\le C\gamma \left\{\frac{\kappa_L}{16 C \gamma}\left\|\widehat{\boldsymbol{\beta}}_{\mathrm{Adj}}-\boldsymbol{\beta}^*\right\|_2^2\right.\\
	&\quad + \left.  \frac{\kappa_L}{16 C\gamma} \left\|\widehat{\boldsymbol{\beta}}_{\mathrm{Adj}}-\boldsymbol{\beta}^*\right\|_2^2 +  \frac{8C\gamma}{\kappa_L}\kappa_U^3\sigma_x^4\sigma_{\varepsilon}^2\sigma_z^2  \left\{\frac{t+p}{n_*^{\hat\tau,|\eta|}} + \sigma_x^2\left(\frac{p+\log(2|\mathcal{E}|)+t}{\overline{\sqrt{nN}}^{\hat\tau,|\eta|}}\right)^2\right\} \right.\\
	&\quad+ \left. 6C\kappa_U\sigma_x^2 \sigma_{\varepsilon}^2 \sigma_z^2 \frac{t+p}{n^{\hat\tau^2,|\eta|^2}_*} +\frac{1}{6C}\mathrm{~J}(\widehat{\boldsymbol{\beta}}_{\mathrm{Adj}} ; \boldsymbol{\omega}) + \frac{16C\gamma}{\kappa_L}\kappa_U^{3} \sigma_x^2 \sigma_{\varepsilon}^2 \sigma_z^2 \frac{t+p}{n^{\hat\tau^2,|\eta|^2}_*}+\frac{\kappa_L}{16C\gamma}\left\|\widehat{\boldsymbol{\beta}}_{\mathrm{Adj}}-\boldsymbol{\beta}^{\star}\right\|_2^2 \right.\\
	&\quad+ \left. \kappa_U \sigma_x\sigma^2_{\varepsilon}\sigma_z\frac{t+p}{n^{\hat\tau,|\eta|}_*}\right\}\\
	&\le C\gamma \left\{ \frac{3\kappa_L}{16 C \gamma}\left\|\widehat{\boldsymbol{\beta}}_{\mathrm{Adj}}-\boldsymbol{\beta}^*\right\|_2^2+\frac{1}{6C}\mathrm{~J}(\widehat{\boldsymbol{\beta}}_{\mathrm{Adj}} ; \boldsymbol{\omega})\right.\\
	&\quad  \left.   +  \frac{8C\gamma}{\kappa_L}\kappa_U^3\sigma_x^4\sigma_{\varepsilon}^2\sigma_z^2  \left\{\frac{t+p}{n_*^{\hat\tau,|\eta|}} + \sigma_x^2\left(\frac{p+\log(2|\mathcal{E}|)+t}{\overline{\sqrt{nN}}^{\hat\tau,|\eta|}}\right)^2\right\} \right.\\
	&\quad+ \left. 6C\kappa_U\sigma_x^2 \sigma_{\varepsilon}^2 \sigma_z^2 \frac{t+p}{n^{\hat\tau^2,|\eta|^2}_*}  + \frac{16C\gamma}{\kappa_L}\kappa_U^{3} \sigma_x^2 \sigma_{\varepsilon}^2 \sigma_z^2 \frac{t+p}{n^{\hat\tau^2,|\eta|^2}_*} +\kappa_U \sigma_x\sigma^2_{\varepsilon}\sigma_z\frac{t+p}{n^{\hat\tau,|\eta|}_*}\right\}\\
	&\le \left\{ \frac{\kappa_L}{4}\left\|\widehat{\boldsymbol{\beta}}_{\mathrm{Adj}}-\boldsymbol{\beta}^*\right\|_2^2+\frac{\gamma}{6}\mathrm{~J}(\widehat{\boldsymbol{\beta}}_{\mathrm{Adj}} ; \boldsymbol{\omega}) \right.\\
	&\quad \left.+ C_1 \frac{\gamma^2}{\kappa_L}\kappa_U^3\sigma_x^4\sigma_{\varepsilon}^2\sigma_z^2  \left\{\frac{t+p}{n_*^{\hat\tau^2,|\eta|^2}} + \sigma_x^2\left(\frac{p+\log(2|\mathcal{E}|)+t}{\overline{\sqrt{nN}}^{\hat\tau,|\eta|}}\right)^2\right\} \right\}\\
\end{align*}
	
Next observe
\begin{align*}
\frac{\kappa_L}{2}\left\|\widehat{\boldsymbol{\beta}}_{\mathrm{Adj}}-\boldsymbol{\beta}^*\right\|_2^2+\frac{\gamma}{6} \mathrm{~J}(\widehat{\boldsymbol{\beta}}_{\mathrm{Adj}} ; \boldsymbol{\omega}) &\le \mathrm{Q}(\widehat{\boldsymbol{\beta}}_{\mathrm{Adj}} ; \gamma, \boldsymbol{\omega})-\mathrm{Q}\left(\boldsymbol{\beta}^* ; \gamma, \boldsymbol{\omega}\right) \\
&= \mathcal{D}_{\mathrm{R}, \widehat{\mathrm{R}}_{\mathrm{Adj}}}(\boldsymbol{\beta}) + \gamma \mathcal{D}_{\mathrm{J}, \widehat{J}_{\mathrm{Adj}}}(\boldsymbol{\beta})+\widehat{\mathrm{Q}}{\mathrm{Adj}}({\widehat{\boldsymbol{\beta}}_{\mathrm{Adj}}} ; \gamma, \boldsymbol{\omega})-\widehat{\mathrm{Q}}_{\mathrm{Adj}}\left(\boldsymbol{\beta}^* ; \gamma, \boldsymbol{\omega}\right)\\
&\le\mathcal{D}_{\mathrm{R}, \widehat{\mathrm{R}}_{\mathrm{Adj}}}(\boldsymbol{\beta}) + \gamma \mathcal{D}_{\mathrm{J}, \widehat{J}_{\mathrm{Adj}}}(\boldsymbol{\beta})\\
&\le \frac{\kappa_L}{4}\left\|\widehat{\boldsymbol{\beta}}_{\mathrm{Adj}}-\boldsymbol{\beta}^*\right\|_2^2+\frac{\gamma}{6} \mathrm{~J}(\widehat{\boldsymbol{\beta}}_{\mathrm{Adj}} ; \boldsymbol{\omega}) \\
&\quad +C_1 \frac{\gamma^2}{\kappa_L} \kappa_U^3 \sigma_x^4 \sigma_{\varepsilon}^2 \sigma_z^2\left\{\frac{t+p}{n_*^{\hat{\tau}^2,|\eta|^2}}+\sigma_x^2\left(\frac{p+\log (2|\mathcal{E}|)+t}{\overline{\sqrt{nN}}^{\hat\tau,|\eta|}}\right)^2\right\} 
\end{align*}

Therefore, we have 
\begin{align*}
\frac{\kappa_L}{2}\left\|\widehat{\boldsymbol{\beta}}_{\mathrm{Adj}}-\boldsymbol{\beta}^*\right\|_2^2+\frac{\gamma}{6} \mathrm{~J}(\widehat{\boldsymbol{\beta}}_{\mathrm{Adj}} ; \boldsymbol{\omega}) &\le \frac{\kappa_L}{4}\left\|\widehat{\boldsymbol{\beta}}_{\mathrm{Adj}}-\boldsymbol{\beta}^*\right\|_2^2+\frac{\gamma}{6} \mathrm{~J}(\widehat{\boldsymbol{\beta}}_{\mathrm{Adj}} ; \boldsymbol{\omega}) \\
&\quad +C_1 \frac{\gamma^2}{\kappa_L} \kappa_U^3 \sigma_x^4 \sigma_{\varepsilon}^2 \sigma_z^2\left\{\frac{t+p}{n_*^{\hat{\tau}^2,|\eta|^2}}+\sigma_x^2\left(\frac{p+\log (2|\mathcal{E}|)+t}{\overline{\sqrt{nN}}^{\hat\tau,|\eta|}}\right)^2\right\},
\end{align*}
which leads to
\begin{align}
\left\|\widehat{\boldsymbol{\beta}}_{\mathrm{Adj}}-\boldsymbol{\beta}^*\right\|_2^2&\le 4C_1 \frac{\gamma^2}{\kappa_L^2} \kappa_U^3 \sigma_x^6 \sigma_{\varepsilon}^2 \sigma_z^2\left\{\frac{t+p}{n_*^{\hat{\tau}^2,|\eta|^2}}+\left(\frac{p+\log (2|\mathcal{E}|)+t}{\overline{\sqrt{nN}}^{\hat\tau,|\eta|}}\right)^2\right\}\notag\\
&\le 4C_1\kappa_U^3 \sigma_x^6 \sigma_{\varepsilon}^2 \sigma_z^2 \frac{\gamma^2}{\kappa_L^2} \left\{\frac{t+p}{n_*^{\hat{\tau}^2,|\eta|^2}}+\left(\frac{p+\log (2|\mathcal{E}|)+t}{\bar{n}^{|\eta|^2}}\right)^2\right\}\notag\\
&\le 4C_1\kappa_U^3 \sigma_x^6 \sigma_{\varepsilon}^2 \sigma_z^2 \frac{\gamma^2}{\kappa_L^2} \left(\frac{(t+p)\eta_{\max}^2}{n_*^{\hat{\tau}^2}}+\frac{\{p+\log (2|\mathcal{E}|)+t\}^2\eta_{\max}^4 }{\bar{n}^2}\right)
\label{eq:simple_rate_sufficeint_l2_error_bound1}
\end{align}
Here we see that $\eta_{\max }=o \left(\sqrt{\bar{n} \wedge n_*^{\hat{\tau}^2}}\right)$ is necessary for the upper bound in (\ref{eq:simple_rate_sufficeint_l2_error_bound1}) to converge. 

Take the square root of both sides, we get 
\begin{align*}
\left\|\widehat{\boldsymbol{\beta}}_{\mathrm{Adj}}-\boldsymbol{\beta}^*\right\|_2&\le \sqrt{4C_1\kappa_U^3 \sigma_x^6 \sigma_{\varepsilon}^2 \sigma_z^2 \frac{\gamma^2}{\kappa_L^2} }\sqrt{\frac{(t+p)\eta_{\max}^2}{n_*^{\hat{\tau}^2}}+\frac{\{p+\log (2|\mathcal{E}|)+t\}^2\eta_{\max}^4 }{\bar{n}^2}}\\
&= \sqrt{4C_1\kappa_U^3 \sigma_x^6} \sigma_{\varepsilon} \sigma_z \frac{\gamma}{\kappa_L} \sqrt{\frac{(t+p)\eta_{\max}^2}{n_*^{\hat{\tau}^2}}+\frac{\{p+\log (2|\mathcal{E}|)+t\}^2\eta_{\max}^4 }{\bar{n}^2}}\\
&\le \sqrt{4C_1\kappa_U^3 \sigma_x^6} \sigma_{\varepsilon} \sigma_z \frac{\gamma}{\kappa_L} \left(  \sqrt{ \frac{(t+p)\eta_{\max}^2}{n_*^{\hat{\tau}^2}}} + \frac{(p+\log (2|\mathcal{E}|)+t)\eta_{\max}^2 }{\bar{n}}  \right). 
\end{align*}
Recall $c_{2} = \sqrt{4C_1\kappa_U^3 \sigma_x^6}$, we have 
$$
\frac{\|\widehat{\boldsymbol{\beta}}_{\mathrm{Adj}}-\boldsymbol{\beta}^*\|_2}{\sigma_{\varepsilon} \sigma_z(\gamma/\kappa_L) }\le c_{2} \left(  \sqrt{ \frac{(t+p)\eta_{\max}^2}{n_*^{\hat{\tau}^2}}} + \frac{(p+\log (2|\mathcal{E}|)+t)\eta_{\max}^2 }{\bar{n}}  \right).
$$

Follow the similar structure, with the event $\mathcal{A}_{1, t}$ corresponds to the event $\mathcal{A}_{1, t, Case 4}$ in Corollary \ref{corollary:R_four_cases}, and the event $\mathcal{A}_{2, t}$ in Corollary \ref{corollary:oneside_boundJ_high_missing_good_imp}, we now show the rate (\ref{eq:small_eta_non-asymptotic_l2_error_bound_rate_2}). Using the fact that $n_*^{\hat{\tau}} \le n_*^{\hat{\tau}^2}$, we obtain the following upper bound:
\begin{equation}\label{eq:l2errorbound_condition_vsc_precise_impu}
\frac{\|\widehat{\boldsymbol{\beta}}_{\mathrm{Q}}^{\mathrm{PP}}-\boldsymbol{\beta}^*\|_2}{\sigma_{\varepsilon} \sigma_z(\gamma/\kappa_L) }\le c_{2} \left(  \sqrt{ \frac{t+p}{n_*^{\hat{\tau}}}} + \frac{p+\log (2|\mathcal{E}|)+t }{\bar{n}}  \right).
\end{equation}

This completes the proof. 
\qed 

\subsection{Additional Simulation Results}

\subsubsection{Performance on Variable Selection}\label{additional_simulation:variable_selection}

\begin{figure}[H]
    \centering
    \begin{subfigure}[t]{0.32\textwidth} 
        \centering
        \includegraphics[width=\textwidth]{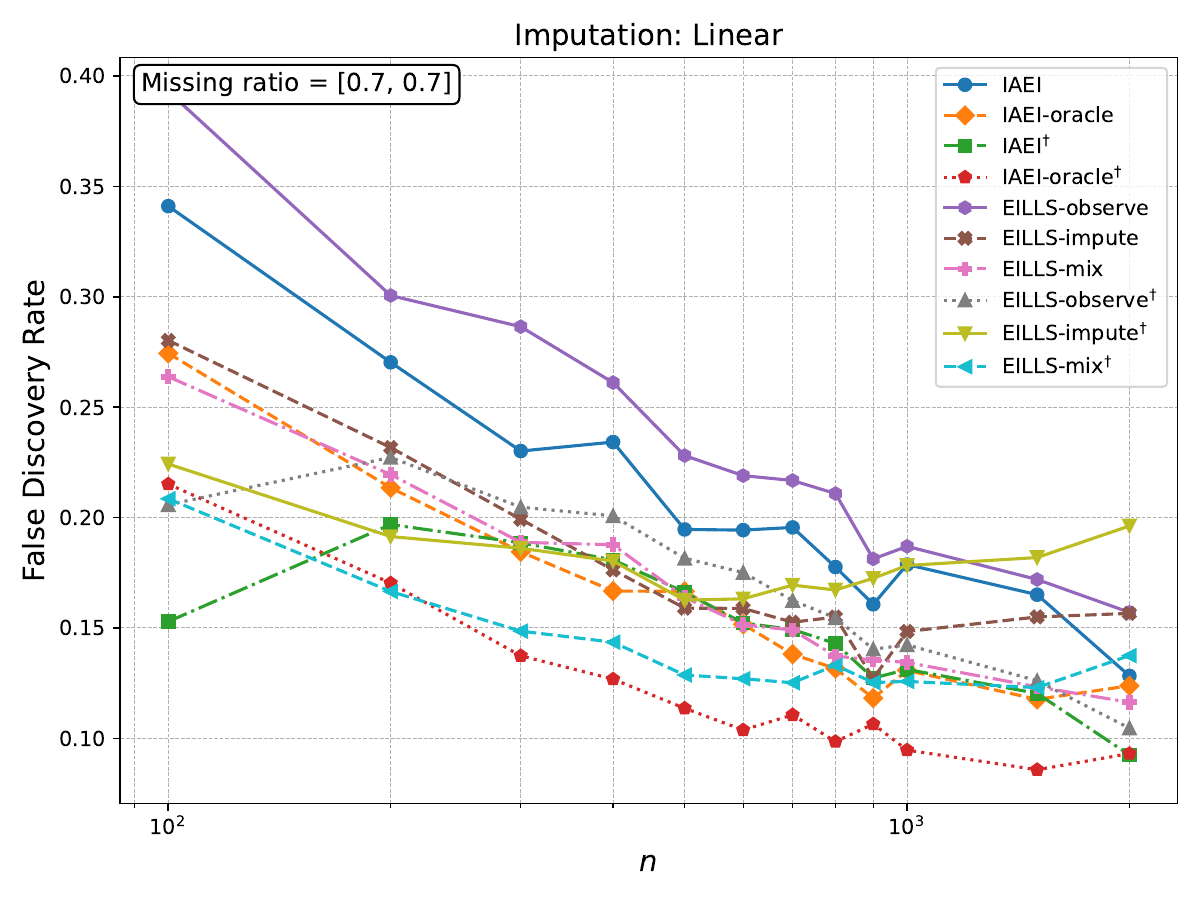} 
        \caption{FDR results under Model 1 with precise Linear imputation.}
        \label{fig:./figures/fig3b_and_fig3b1_and_fig3a1/precise_imputation/model1/fig3a1_precise_model1_model0_0.7_linear_all_fdr.pdf}
    \end{subfigure}
    \hfill 
    \begin{subfigure}[t]{0.32\textwidth} 
        \centering
        \includegraphics[width=\textwidth]{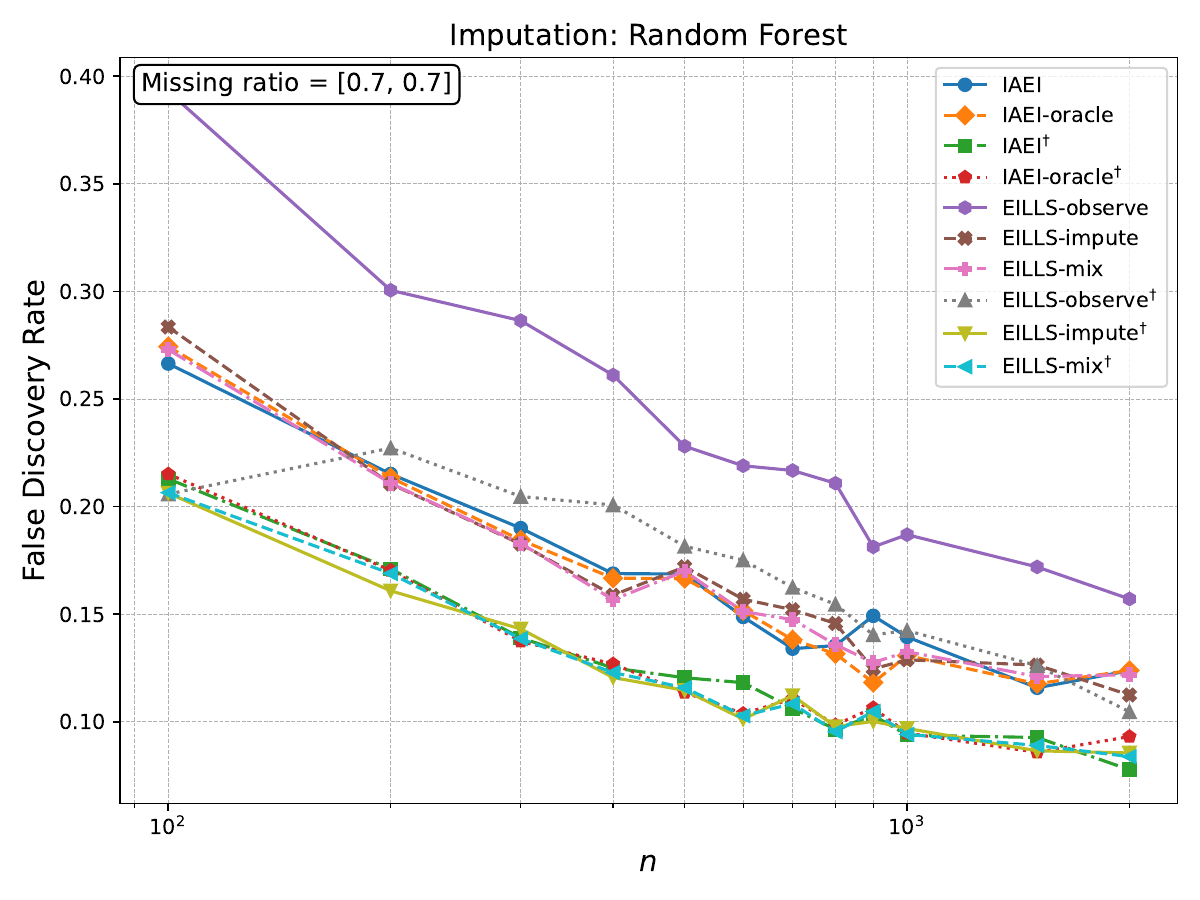} 
        \caption{FDR results under Model 1 with precise RandomForest imputation.}
        \label{fig:./figures/fig3b_and_fig3b1_and_fig3a1/precise_imputation/model1/fig3a1_precise_model1_model0_0.7_rf_all_fdr.pdf}
    \end{subfigure}
	\hfill 
    \begin{subfigure}[t]{0.32\textwidth} 
        \centering
        \includegraphics[width=\textwidth]{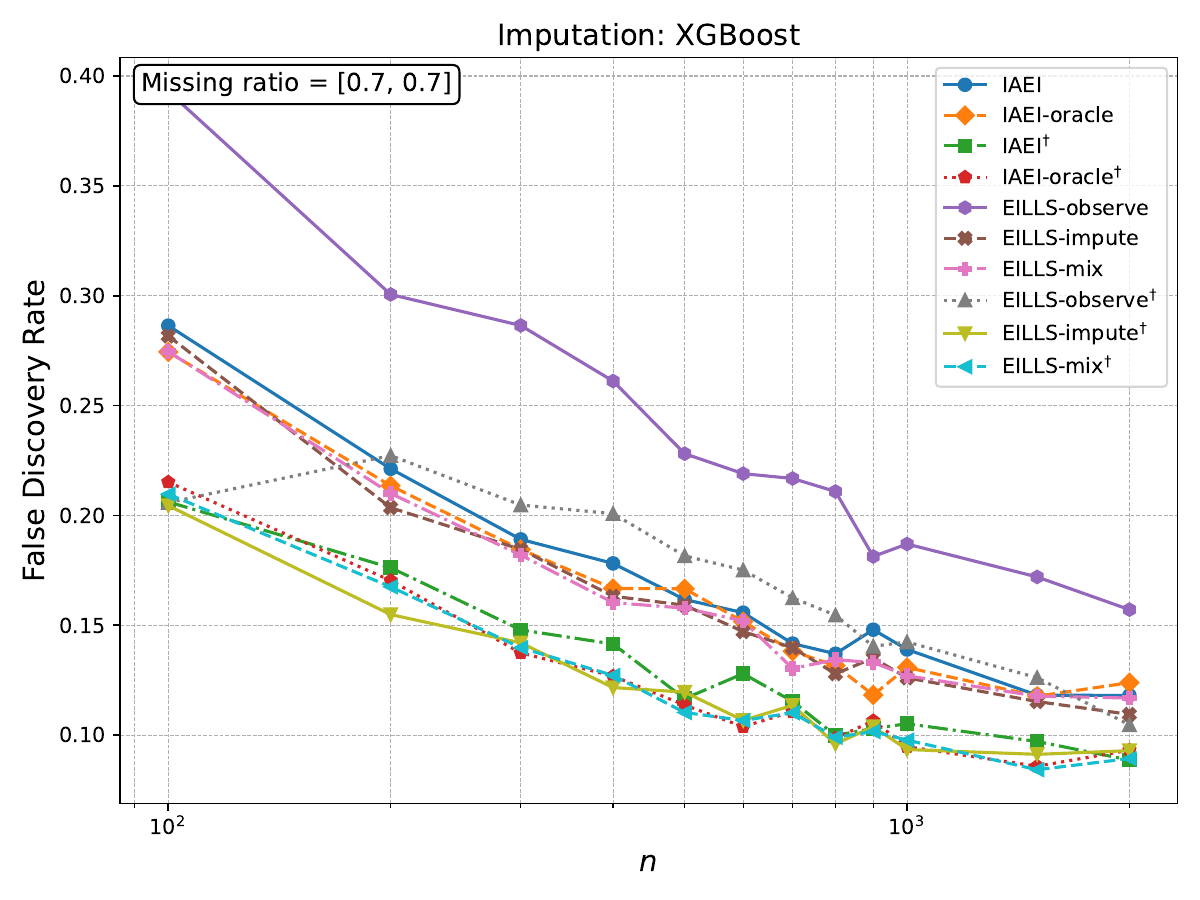} 
        \caption{FDR results under Model 1 with precise XGBoost imputation.}
        \label{fig:./figures/fig3b_and_fig3b1_and_fig3a1/precise_imputation/model1/fig3a1_precise_model1_model0_0.7_xgboost_all_fdr.pdf}
    \end{subfigure}
    \label{fig:./figures/fig3b_and_fig3b1_and_fig3a1/precise_imputation/model1/fig3a1_precise_model1_model0_0.7_linear_all_fdr.pdf./figures/fig3b_and_fig3b1_and_fig3a1/precise_imputation/model1/fig3a1_precise_model1_model0_0.7_rf_all_fdr.pdf./figures/fig3b_and_fig3b1_and_fig3a1/precise_imputation/model1/fig3a1_precise_model1_model0_0.7_xgboost_all_fdr.pdf}
\end{figure}

\begin{figure}[H]
    \centering
    \begin{subfigure}[t]{0.32\textwidth} 
        \centering
        \includegraphics[width=\textwidth]{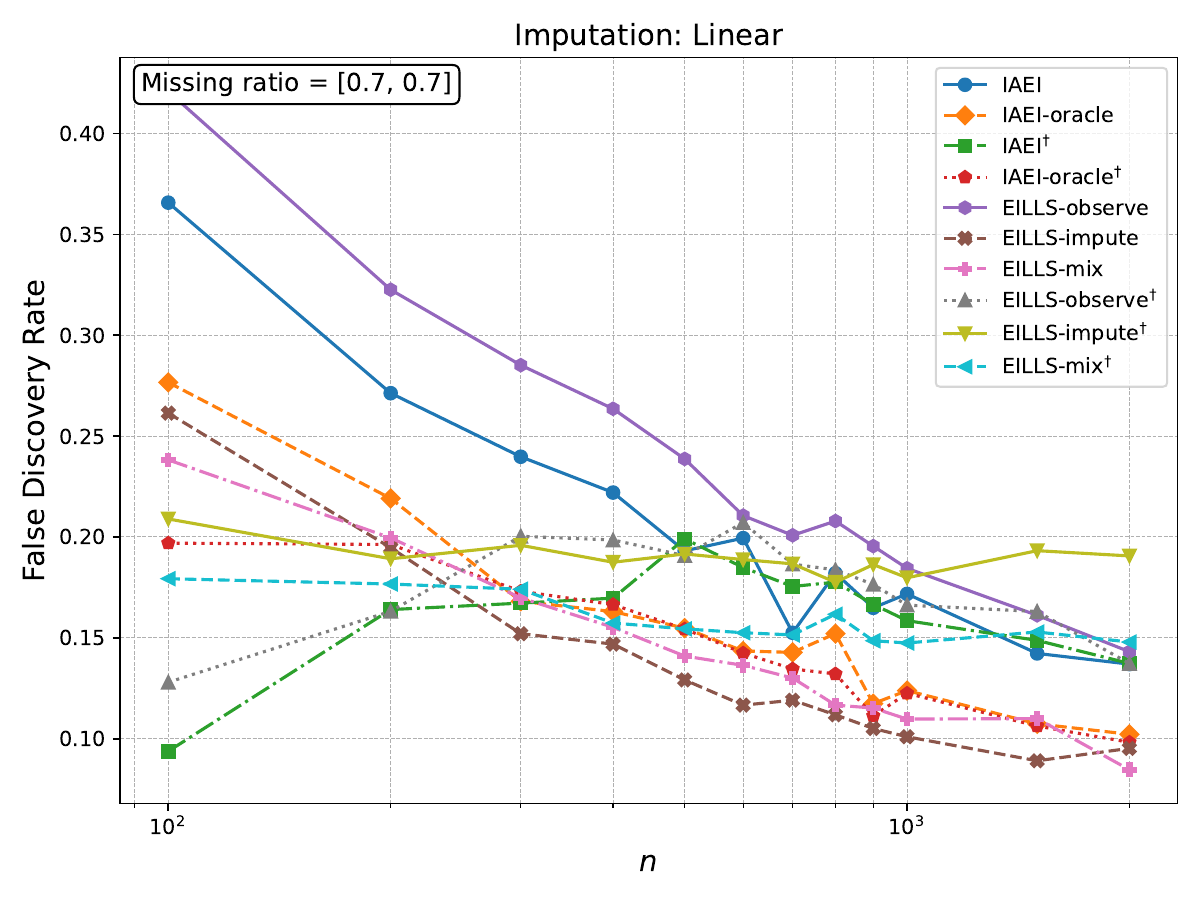} 
        \caption{FDR results under Model 2 with precise Linear imputation.}
        \label{fig:./figures/fig3b_and_fig3b1_and_fig3a1/precise_imputation/model2/fig3a1_precise_model2_model0_0.7_linear_all_fdr.pdf}
    \end{subfigure}
    \hfill 
    \begin{subfigure}[t]{0.32\textwidth} 
        \centering
        \includegraphics[width=\textwidth]{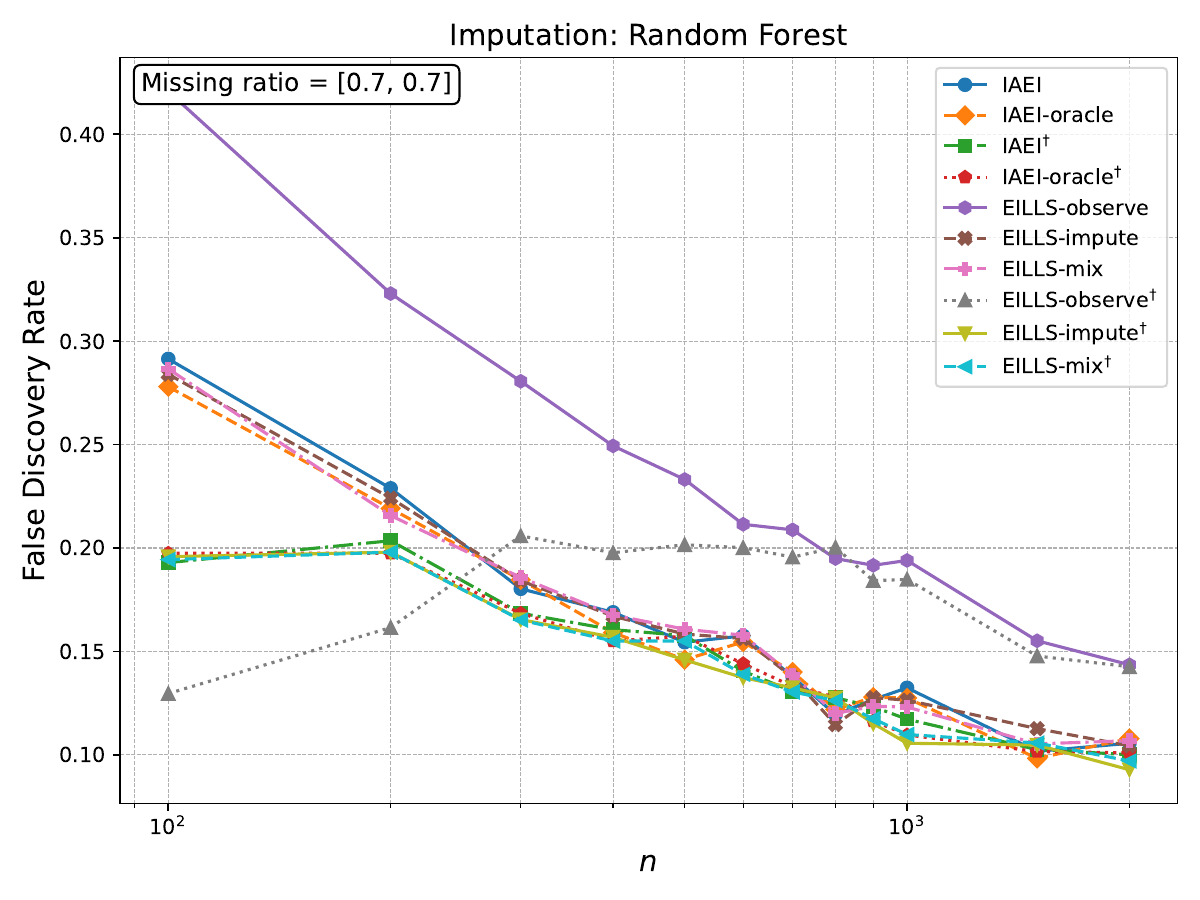} 
        \caption{FDR results under Model 2 with precise RandomForest imputation.}
        \label{fig:./figures/fig3b_and_fig3b1_and_fig3a1/precise_imputation/model2/fig3a1_precise_model2_model0_0.7_rf_all_fdr.pdf}
    \end{subfigure}
	\hfill 
    \begin{subfigure}[t]{0.32\textwidth} 
        \centering
        \includegraphics[width=\textwidth]{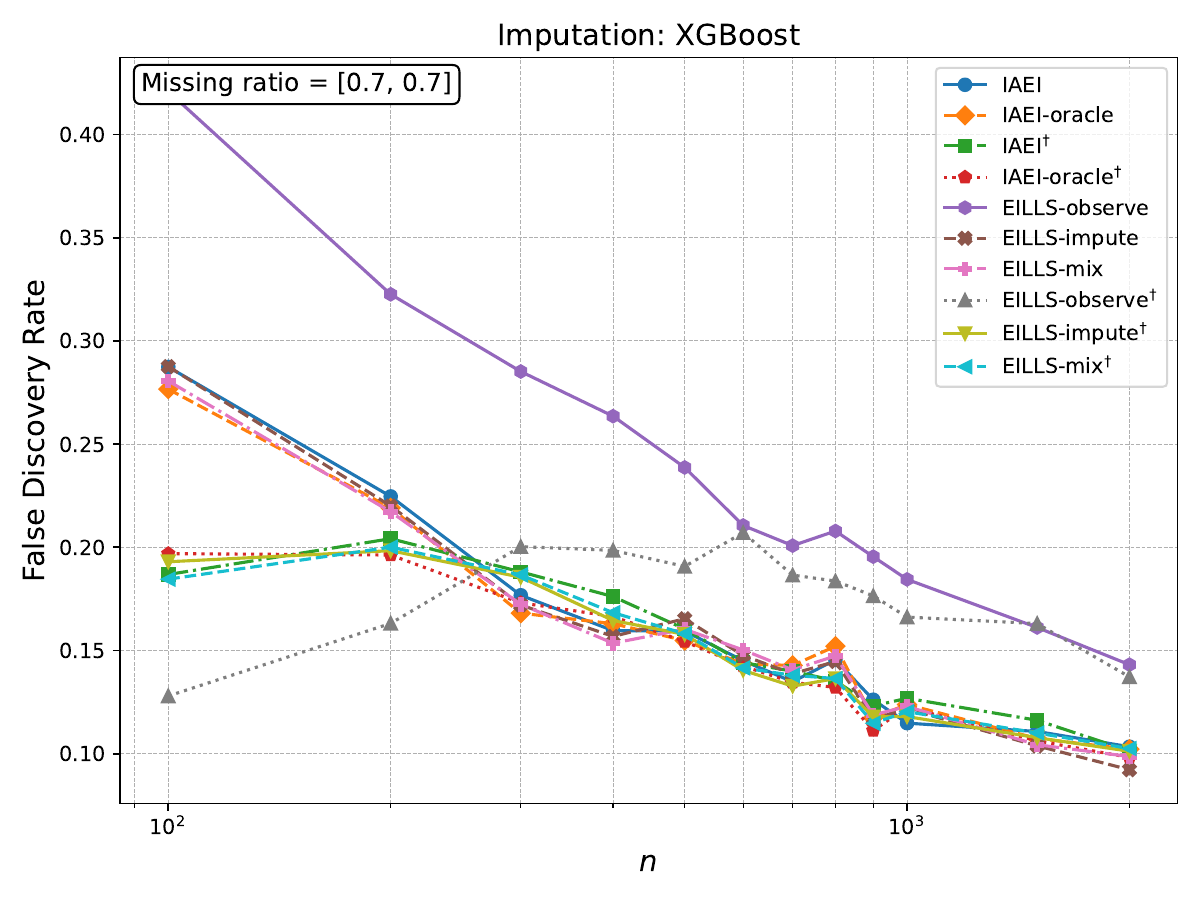} 
        \caption{FDR results under Model 2 with precise XGBoost imputation.}
        \label{fig:./figures/fig3b_and_fig3b1_and_fig3a1/precise_imputation/model2/fig3a1_precise_model2_model0_0.7_xgboost_all_fdr.pdf}
    \end{subfigure}
    \label{fig:./figures/fig3b_and_fig3b1_and_fig3a1/precise_imputation/model2/fig3a1_precise_model2_model0_0.7_linear_all_fdr.pdf./figures/fig3b_and_fig3b1_and_fig3a1/precise_imputation/model2/fig3a1_precise_model2_model0_0.7_rf_all_fdr.pdf./figures/fig3b_and_fig3b1_and_fig3a1/precise_imputation/model2/fig3a1_precise_model2_model0_0.7_xgboost_all_fdr.pdf}
\end{figure}

\begin{figure}[H]
    \centering
    \begin{subfigure}[t]{0.32\textwidth} 
        \centering
        \includegraphics[width=\textwidth]{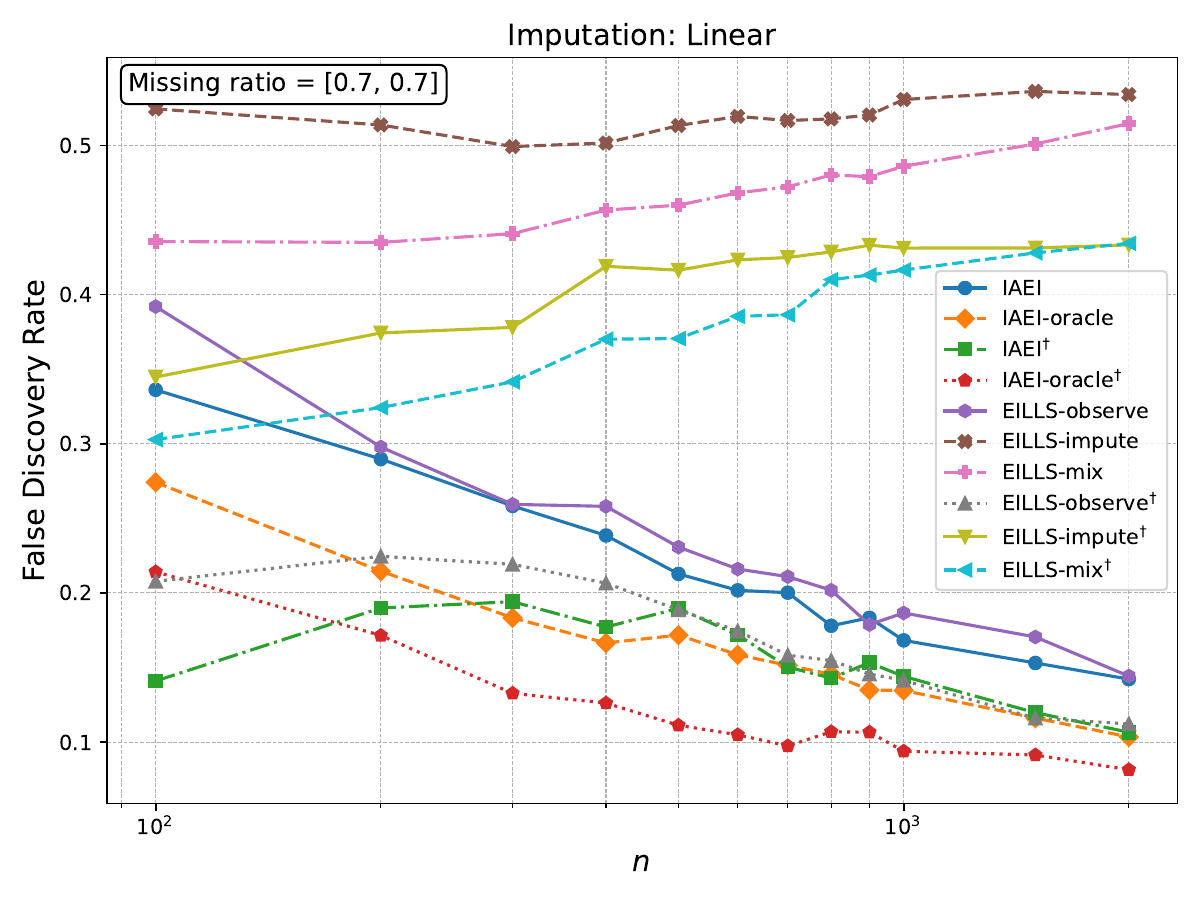} 
        \caption{FDR results under Model 1 with bias Linear imputation.}
        \label{fig:./figures/fig3b_and_fig3b1_and_fig3a1/bias_imputation/model1/fig3a1_bias_model1_model0_0.7_linear_all_fdr.pdf}
    \end{subfigure}
    \hfill 
    \begin{subfigure}[t]{0.32\textwidth} 
        \centering
        \includegraphics[width=\textwidth]{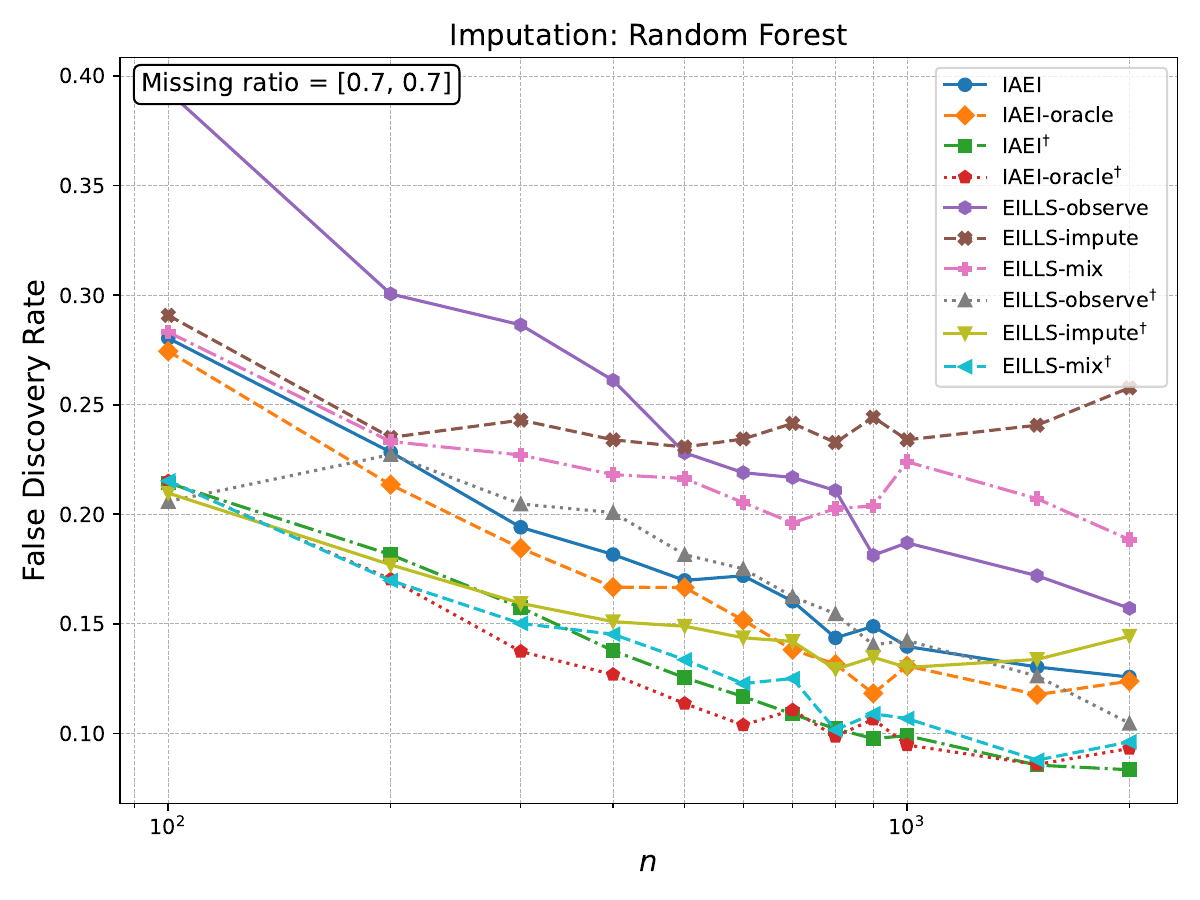} 
        \caption{FDR results under Model 1 with bias RandomForest imputation.}
        \label{fig:./figures/fig3b_and_fig3b1_and_fig3a1/bias_imputation/model1/fig3a1_bias_model1_model0_0.7_rf_all_fdr.pdf}
    \end{subfigure}
	\hfill 
    \begin{subfigure}[t]{0.32\textwidth} 
        \centering
        \includegraphics[width=\textwidth]{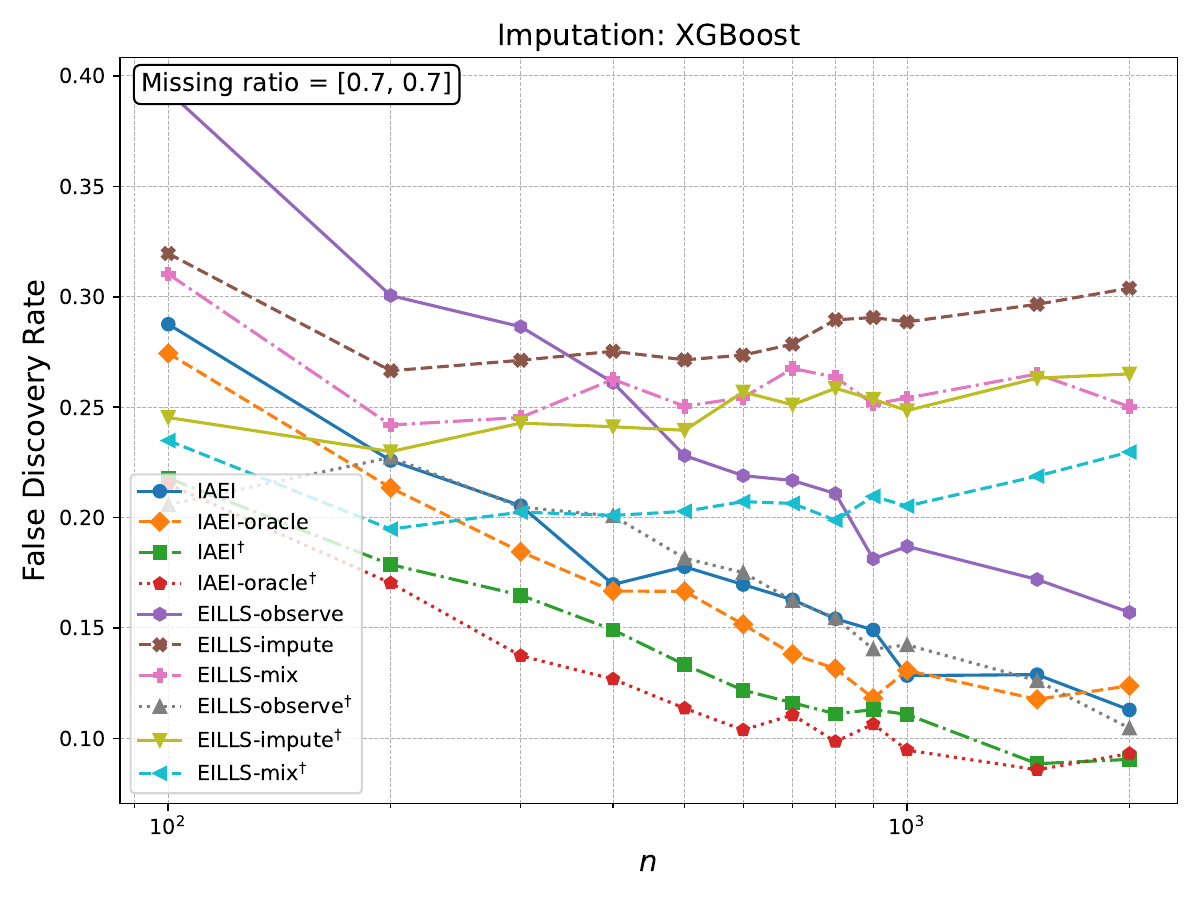} 
        \caption{FDR results under Model 1 with bias XGBoost imputation.}
        \label{fig:./figures/fig3b_and_fig3b1_and_fig3a1/bias_imputation/model1/fig3a1_bias_model1_model0_0.7_xgboost_all_fdr.pdf}
    \end{subfigure}
    \label{fig:./figures/fig3b_and_fig3b1_and_fig3a1/hbias_imputation/model1/fig3a1_hbias_model1_model0_0.7_linear_all_fdr.pdf./figures/fig3b_and_fig3b1_and_fig3a1/hbias_imputation/model1/fig3a1_hbias_model1_model0_0.7_rf_all_fdr.pdffig:./figures/fig3b_and_fig3b1_and_fig3a1/hbias_imputation/model1/fig3a1_hbias_model1_model0_0.7_xgboost_all_fdr.pdf}
\end{figure}

\begin{figure}[H]
    \centering
    \begin{subfigure}[t]{0.32\textwidth} 
        \centering
        \includegraphics[width=\textwidth]{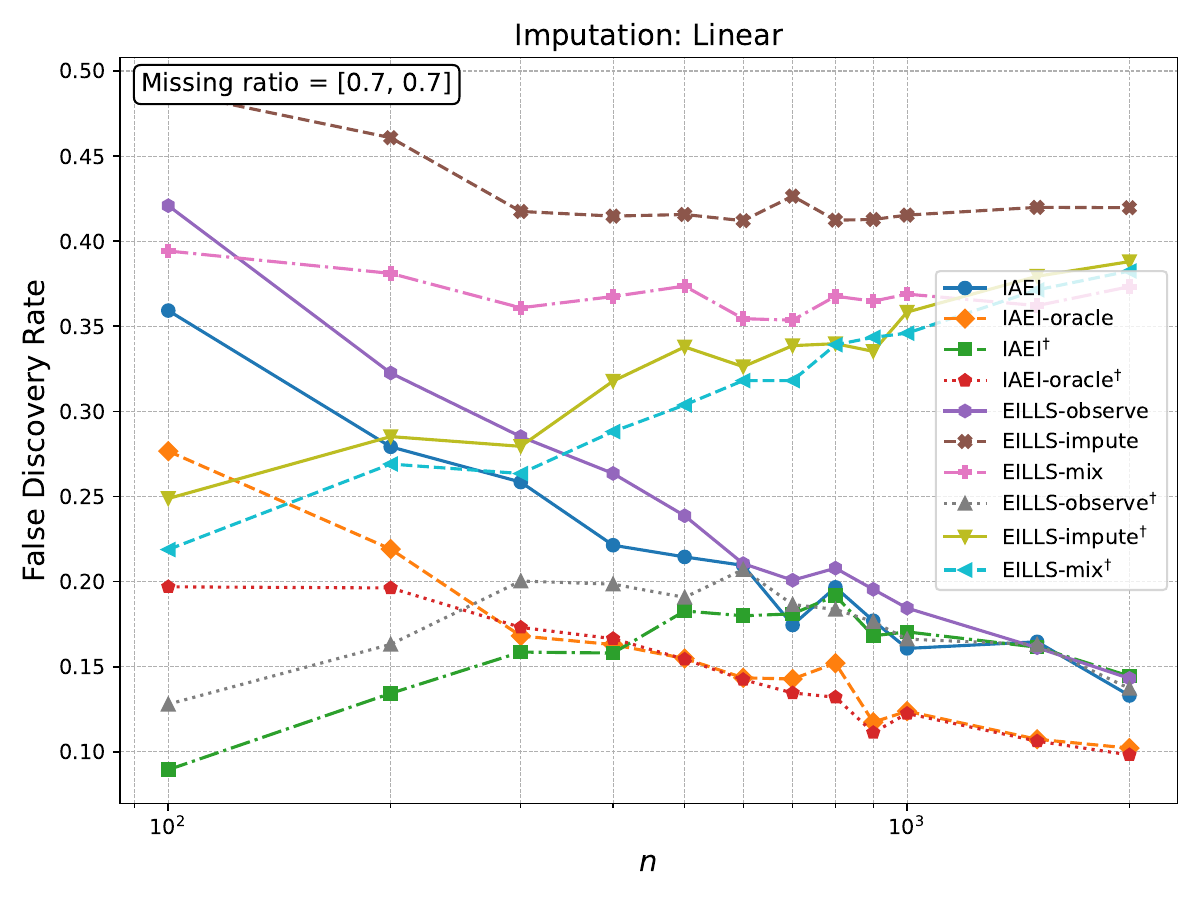} 
        \caption{FDR results under Model 2 with bias Linear imputation.}
        \label{fig:./figures/fig3b_and_fig3b1_and_fig3a1/bias_imputation/model2/fig3a1_bias_model2_model0_0.7_linear_all_fdr.pdf}
    \end{subfigure}
    \hfill 
    \begin{subfigure}[t]{0.32\textwidth} 
        \centering
        \includegraphics[width=\textwidth]{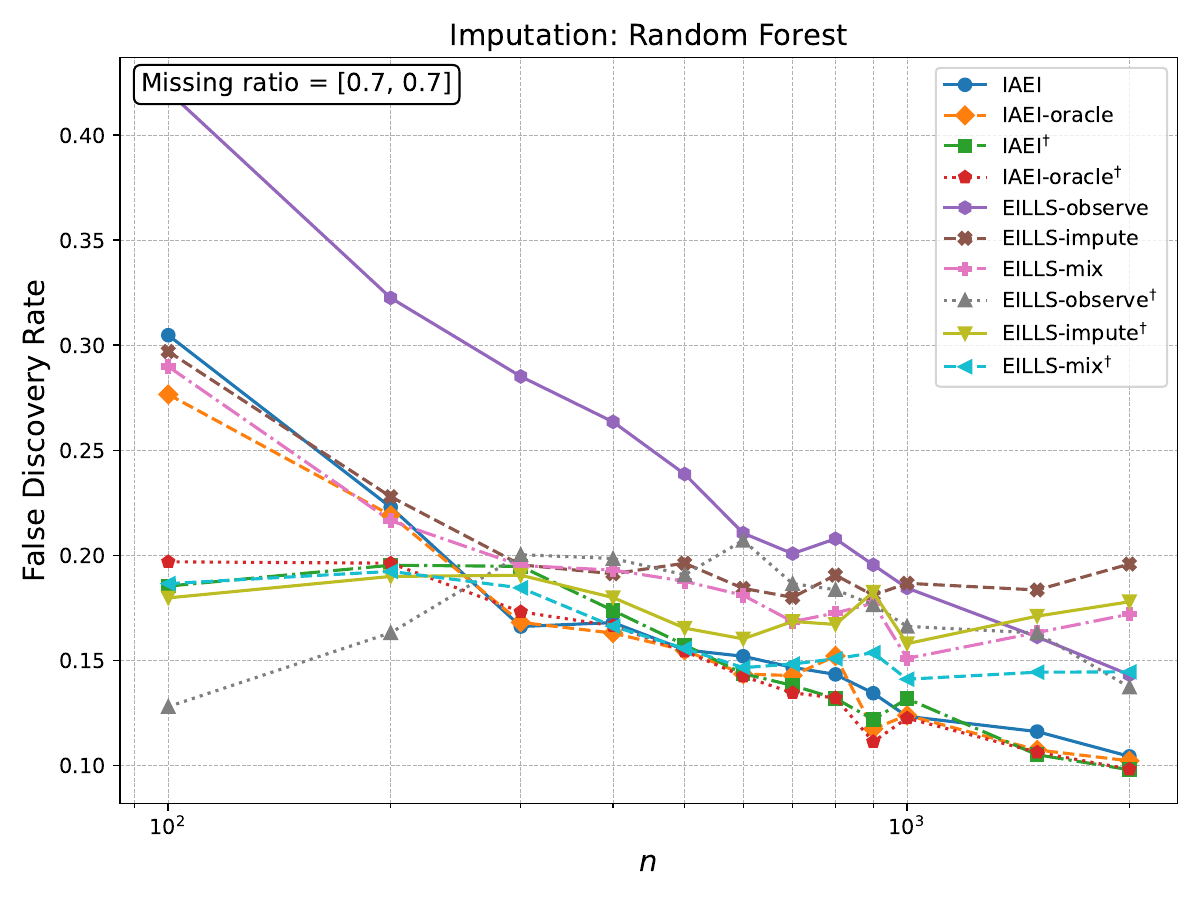} 
        \caption{FDR results under Model 2 with bias RandomForest imputation.}
        \label{fig:./figures/fig3b_and_fig3b1_and_fig3a1/bias_imputation/model2/fig3a1_bias_model2_model0_0.7_rf_all_fdr.pdf}
    \end{subfigure}
	\hfill 
    \begin{subfigure}[t]{0.32\textwidth} 
        \centering
        \includegraphics[width=\textwidth]{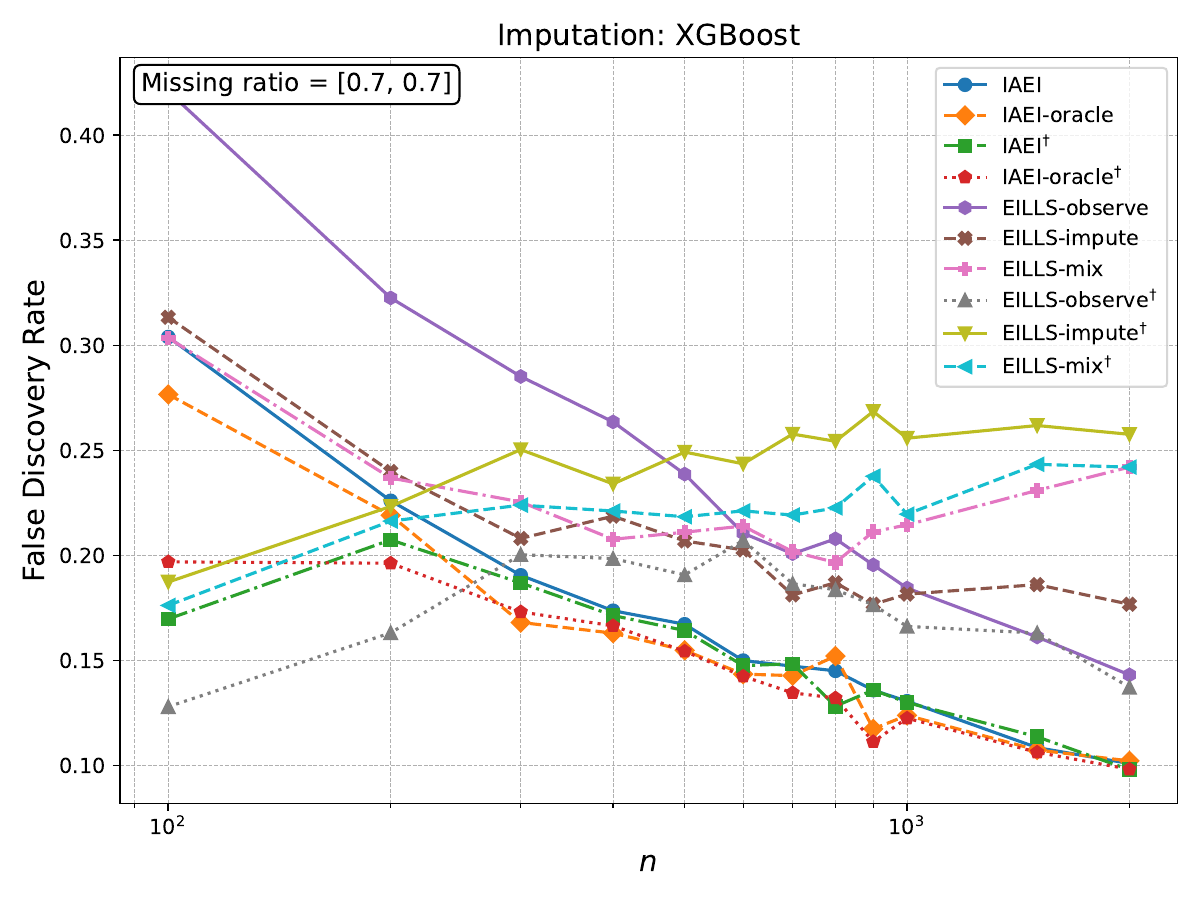} 
        \caption{FDR results under Model 2 with bias XGBoost imputation.}
        \label{fig:./figures/fig3b_and_fig3b1_and_fig3a1/bias_imputation/model2/fig3a1_bias_model2_model0_0.7_xgboost_all_fdr.pdf}
    \end{subfigure}
    \label{fig:./figures/fig3b_and_fig3b1_and_fig3a1/bias_imputation/model2/fig3a1_bias_model2_model0_0.7_linear_all_fdr.pdf./figures/fig3b_and_fig3b1_and_fig3a1/bias_imputation/model2/fig3a1_bias_model2_model0_0.7_rf_all_fdr.pdf./figures/fig3b_and_fig3b1_and_fig3a1/bias_imputation/model2/fig3a1_bias_model2_model0_0.7_xgboost_all_fdr.pdf}
\end{figure}

\begin{figure}[H]
    \centering
    \begin{subfigure}[t]{0.45\textwidth} 
        \centering
        \includegraphics[width=\textwidth]{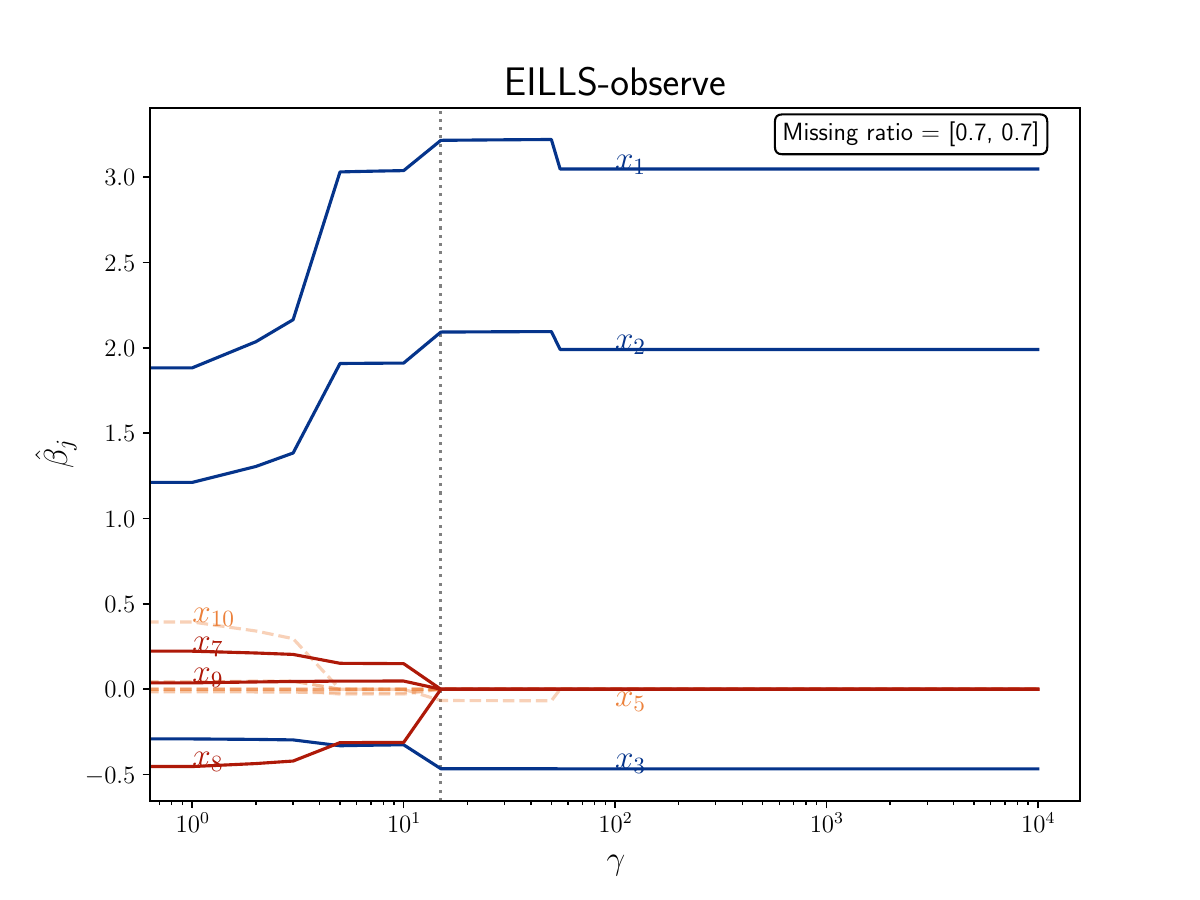} 
        \caption{EILLS-observe.}
        \label{fig:./figures/fig3a/precise_imputation/model0/fig3a_xgboost_eills_observe.pdf}
    \end{subfigure}
    \hfill
    \begin{subfigure}[t]{0.45\textwidth} 
        \centering
        \includegraphics[width=\textwidth]{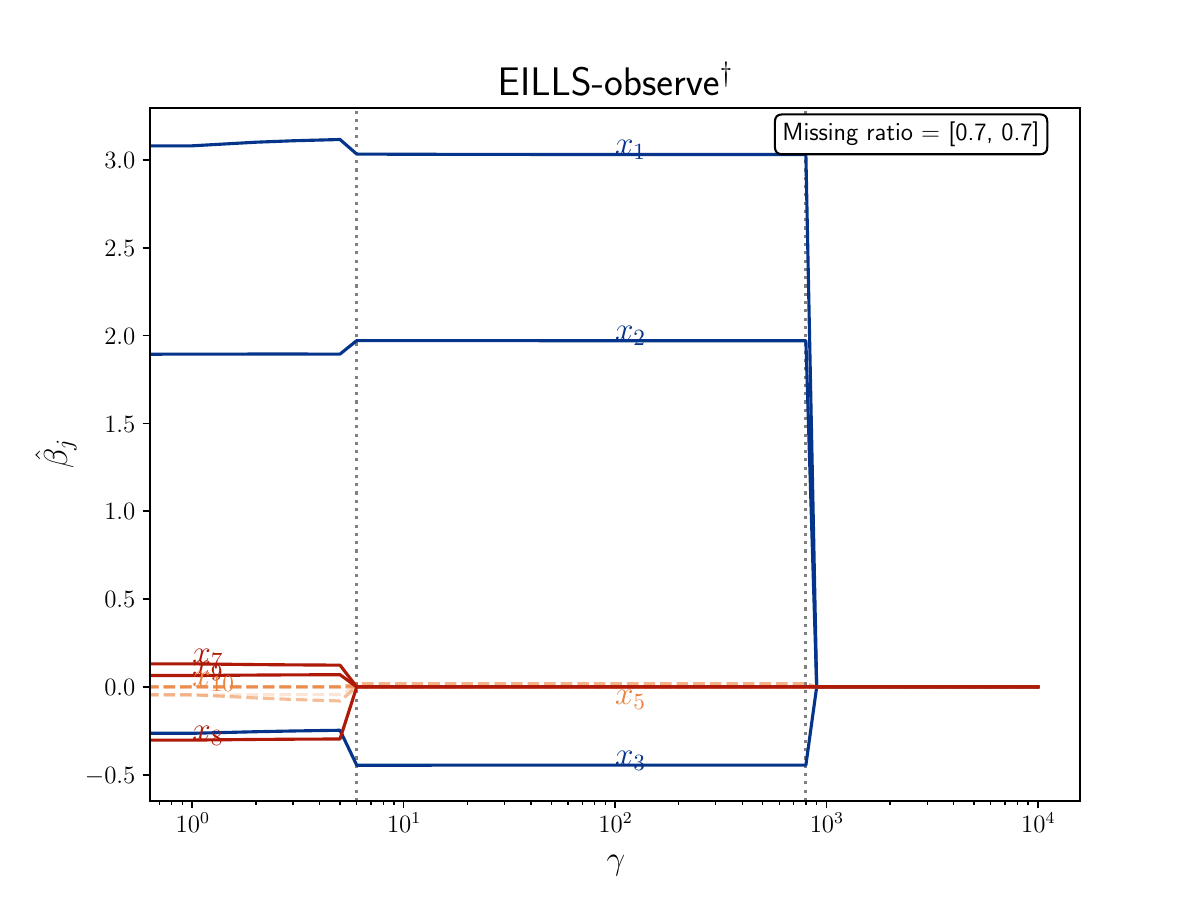} 
        \caption{EILLS-observe$^{\dagger}$.}
        \label{fig:./figures/fig3a/precise_imputation/model0/fig3a_xgboost_eills_observe_ce.pdf}
    \end{subfigure}

	\caption{Variable selection performance of EILLS-observe compared to EILLS-observe$^{\dagger}$ under Model 0. The left plot illustrates the performance using the original penalty, while the right plot highlights the improved stability and accuracy achieved with the enhanced penalty.}
    
    \label{fig:variable_selection_one_simulation1model0}
\end{figure}

\begin{figure}[H]
    \centering
    
    \begin{subfigure}[t]{0.45\textwidth} 
        \centering
        \includegraphics[width=\textwidth]{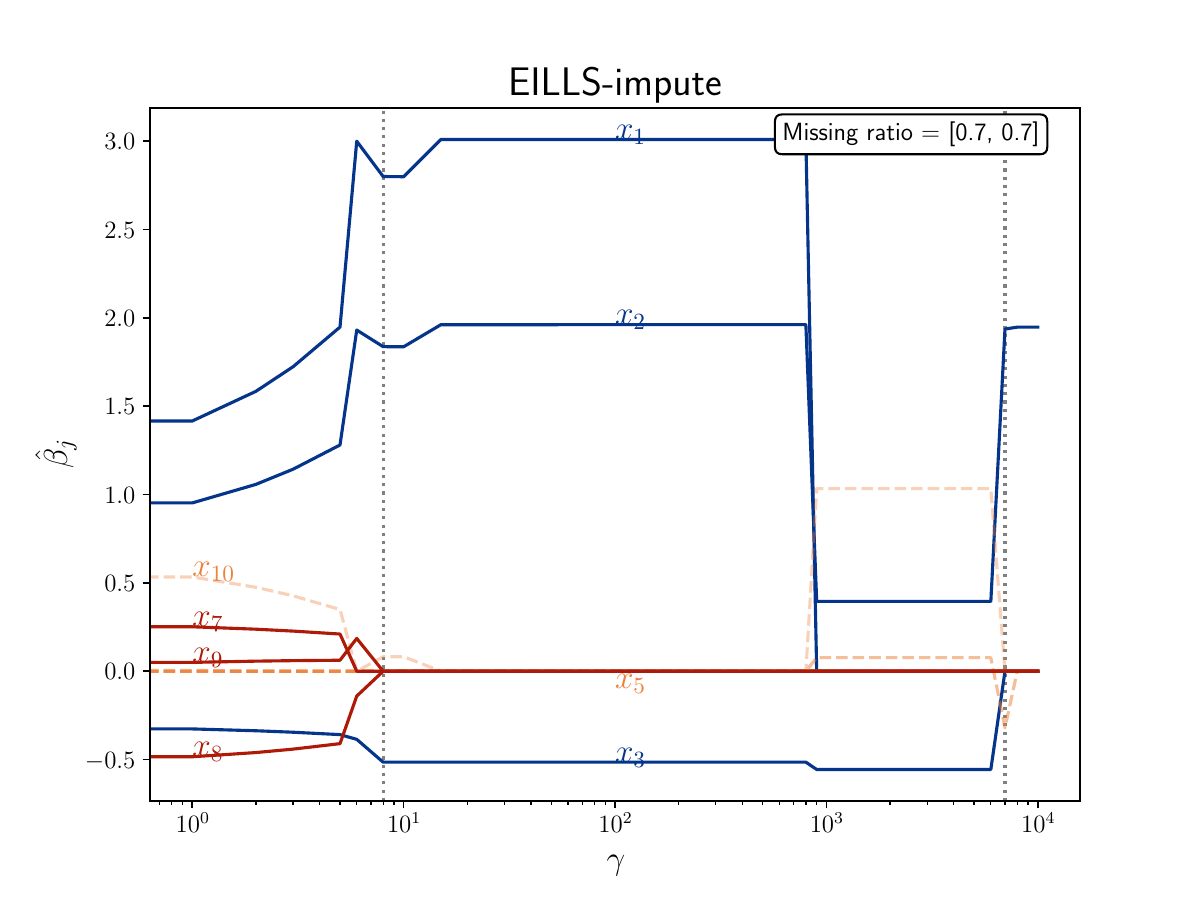} 
        \caption{EILLS-impute.}
        \label{fig:./figures/fig3a/precise_imputation/model0/fig3a_xgboost_eills_impute.pdf}
    \end{subfigure}
    \hfill
    \begin{subfigure}[t]{0.45\textwidth} 
        \centering
        \includegraphics[width=\textwidth]{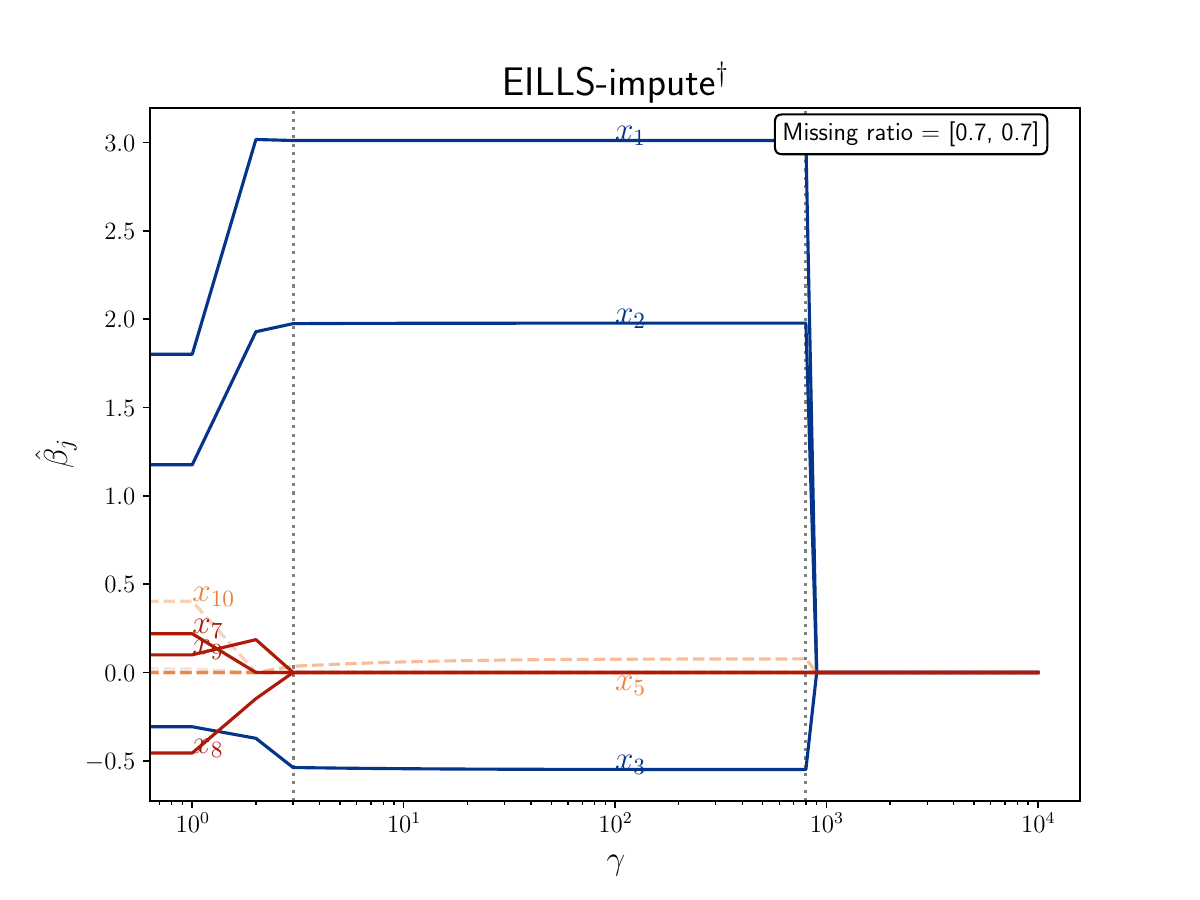} 
        \caption{EILLS-impute$^{\dagger}$.}
        \label{fig:./figures/fig3a/precise_imputation/model0/fig3a_xgboost_eills_impute_ce.pdf}
    \end{subfigure}

	\caption{Variable selection performance of EILLS-impute compared to EILLS-impute$^{\dagger}$ under Model 0 with precise imputation. The suboptimal performance in the left plot is not attributable to biased imputation, as this analysis focuses on precise imputation. The left plot illustrates the performance with the original penalty, while the right plot demonstrates the enhanced stability and accuracy achieved with the improved penalty.}
    
    \label{fig:variable_selection_one_simulation2model0}
\end{figure}

\begin{figure}[H]
    \centering

    \begin{subfigure}[t]{0.45\textwidth} 
        \centering
        \includegraphics[width=\textwidth]{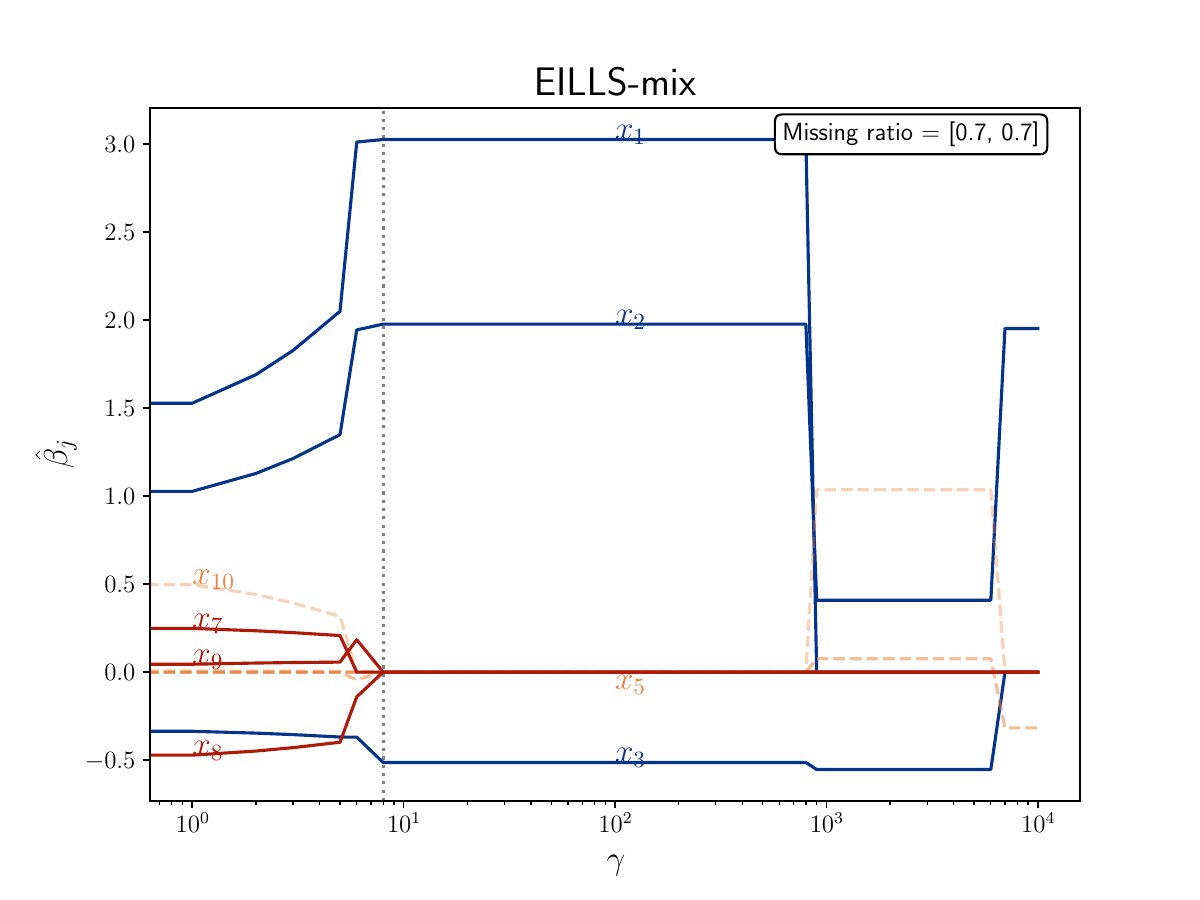} 
        \caption{EILLS-mix.}
        \label{fig:./figures/fig3a/precise_imputation/model0/fig3a_xgboost_eills_mix.pdf}
    \end{subfigure}
    \hfill
    \begin{subfigure}[t]{0.45\textwidth} 
        \centering
        \includegraphics[width=\textwidth]{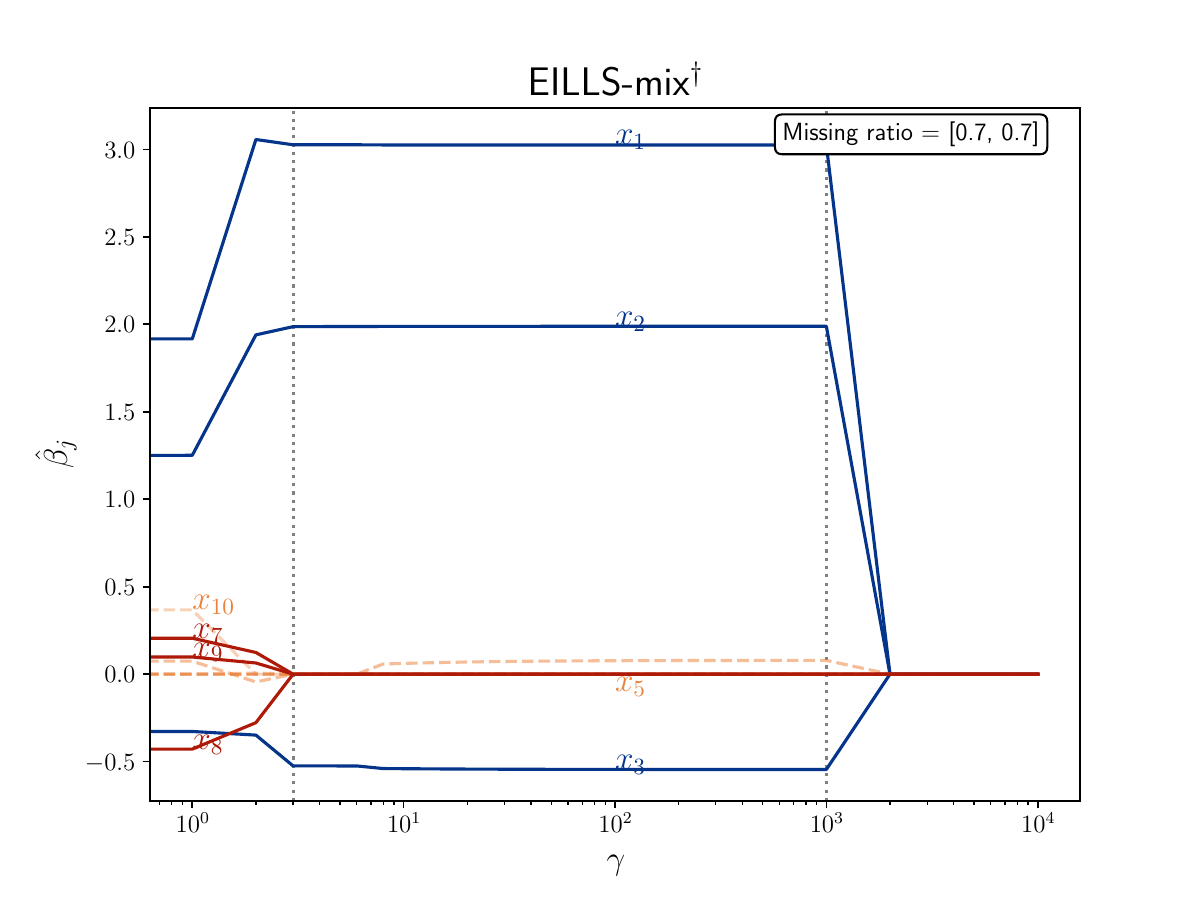} 
        \caption{EILLS-mix$^{\dagger}$.}
        \label{fig:./figures/fig3a/precise_imputation/model0/fig3a_xgboost_eills_mix_ce.pdf.}
    \end{subfigure}

	\caption{Variable selection performance of EILLS-mix vs. EILLS-mix$^{\dagger}$ under Model 0 with precise imputation, where the right plot demonstrates the enhanced penalty’s improved stability and accuracy.}
    
    \label{fig:variable_selection_one_simulation3model0}
\end{figure}

\begin{figure}[H]
    \centering

    \begin{subfigure}[t]{0.45\textwidth} 
        \centering
        \includegraphics[width=\textwidth]{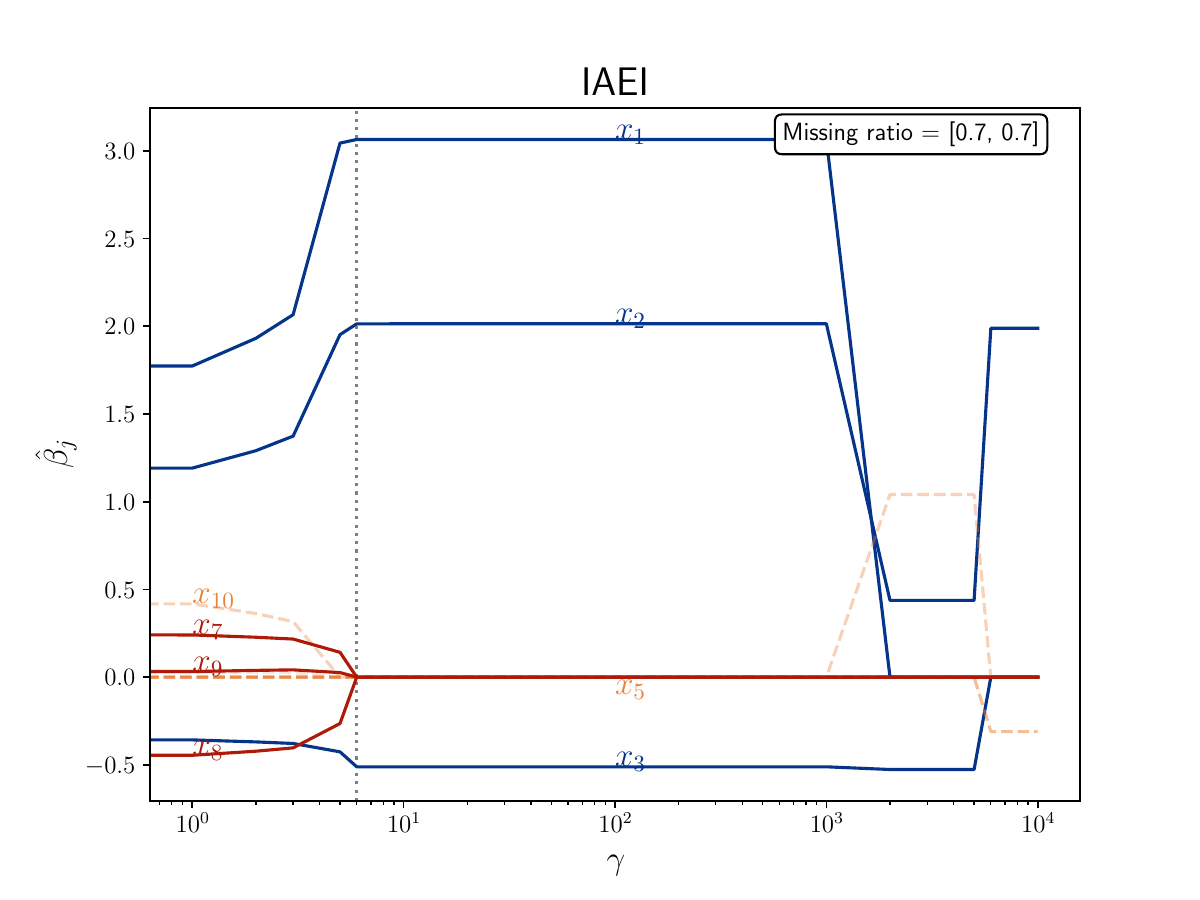} 
        \caption{IAEI.}
        \label{fig:./figures/fig3a/precise_imputation/model0/fig3a_xgboost_iaei.pdf}
    \end{subfigure}
    \hfill
    \begin{subfigure}[t]{0.45\textwidth} 
        \centering
        \includegraphics[width=\textwidth]{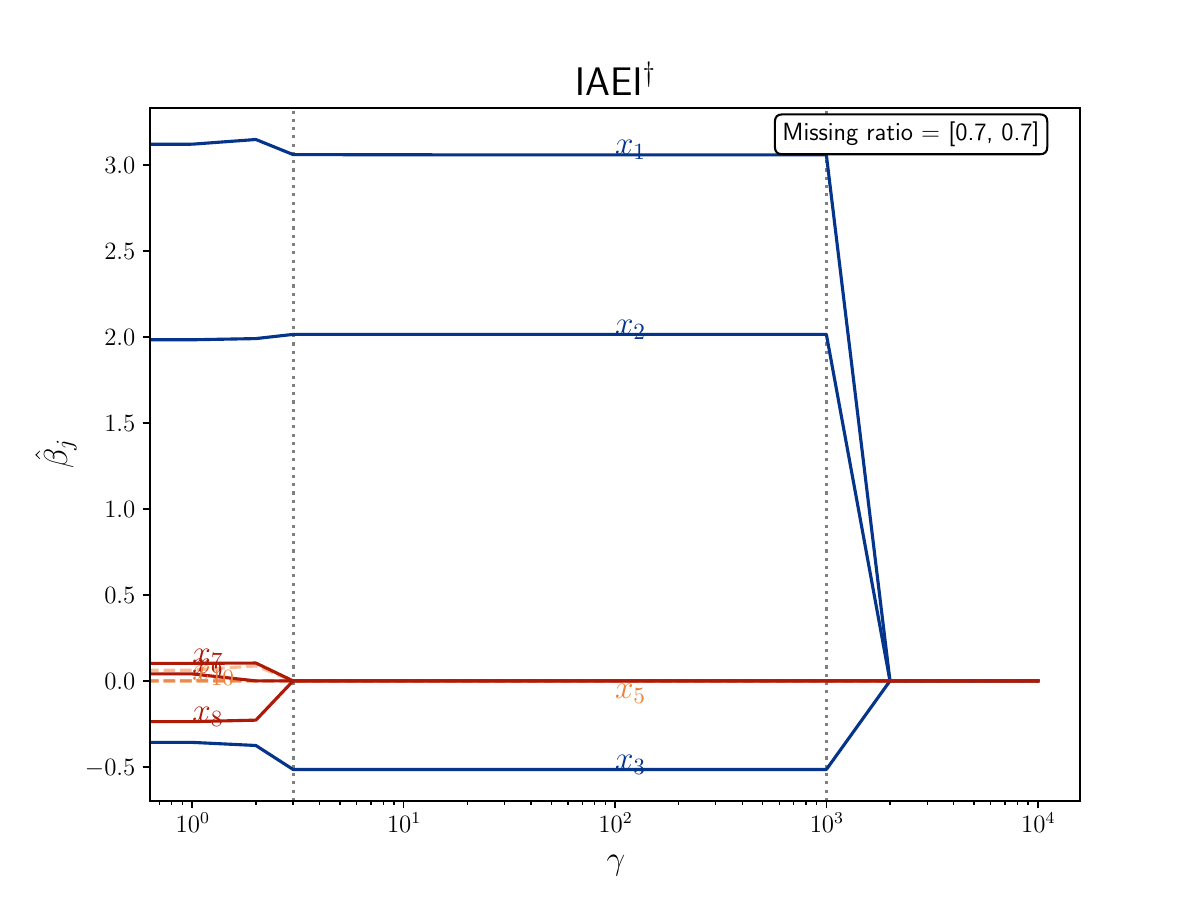} 
        \caption{IAEI$^{\dagger}$.}
        \label{fig:./figures/fig3a/precise_imputation/model0/fig3a_xgboost_iaei_ce.pdf}
    \end{subfigure}
    
    \caption{Variable selection performance of IAEI vs IAEI$^{\dagger}$ under Model 0. While the left plot shows the result with original penalty, the right plot demonstrates the improved stable performance using the enhanced penalty.}
    \label{fig:variable_selection_one_simulation4model0}
\end{figure}

\begin{figure}[H]
    \centering
    \begin{subfigure}[t]{0.45\textwidth} 
        \centering
        \includegraphics[width=\textwidth]{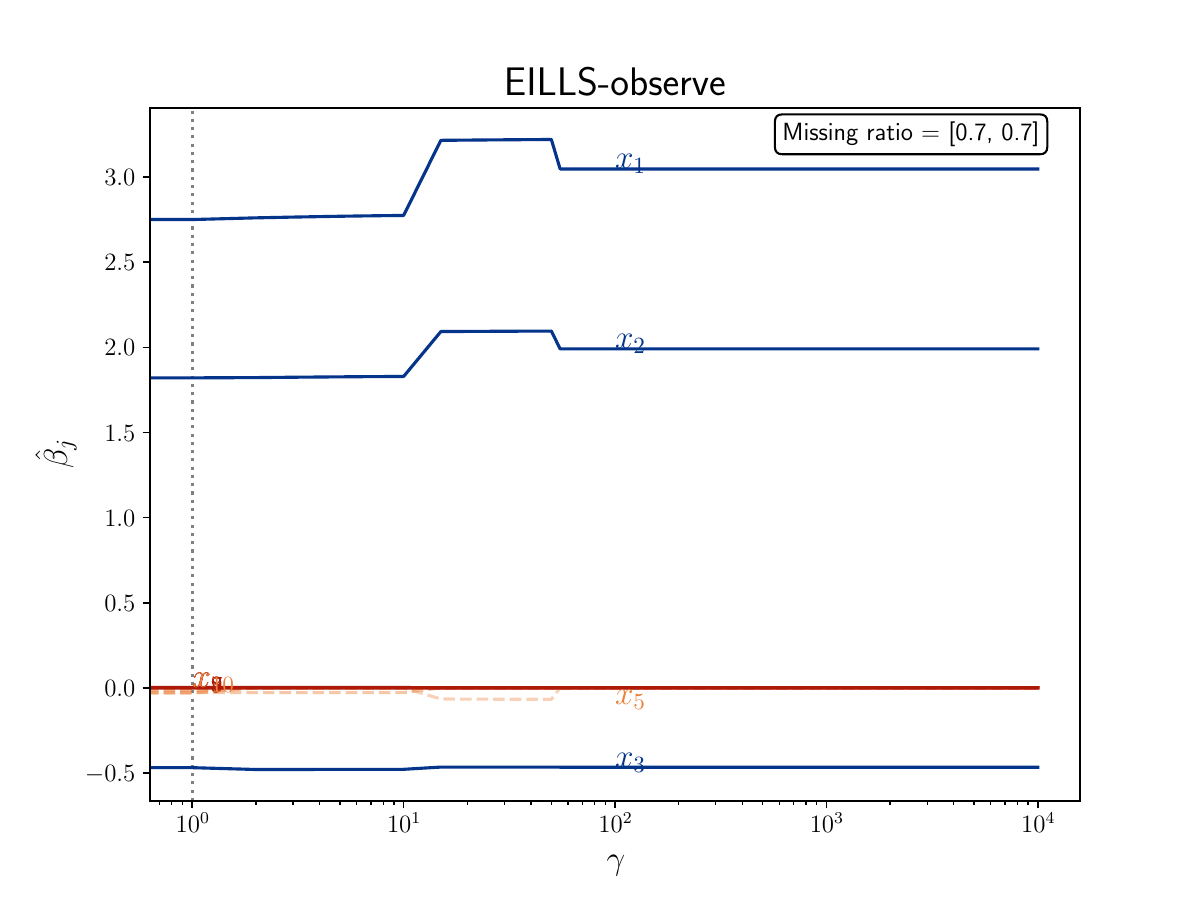} 
        \caption{EILLS-observe.}
        \label{fig:./figures/fig3a/precise_imputation/model1/fig3a_xgboost_eills_observe.pdf}
    \end{subfigure}
    \hfill
    \begin{subfigure}[t]{0.45\textwidth} 
        \centering
        \includegraphics[width=\textwidth]{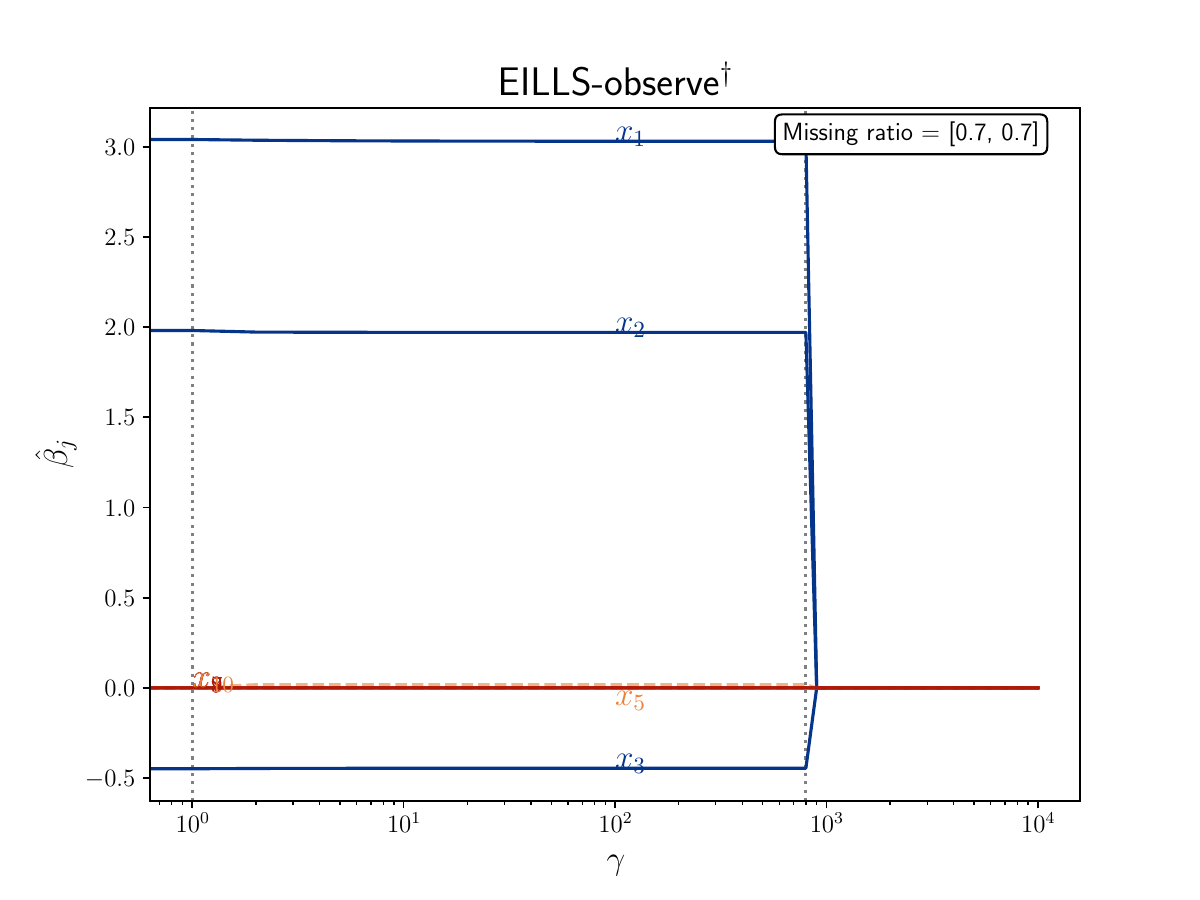} 
        \caption{EILLS-observe$^{\dagger}$.}
        \label{fig:./figures/fig3a/precise_imputation/model1/fig3a_xgboost_eills_observe_ce.pdf}
    \end{subfigure}

	\caption{Variable selection performance of EILLS-observe compared to EILLS-observe$^{\dagger}$ under Model 1. The left plot illustrates the performance using the original penalty, while the right plot highlights the improved stability and accuracy achieved with the enhanced penalty.}
    
    \label{fig:variable_selection_one_simulation1model1}
\end{figure}

\begin{figure}[H]
    \centering
    
    \begin{subfigure}[t]{0.45\textwidth} 
        \centering
        \includegraphics[width=\textwidth]{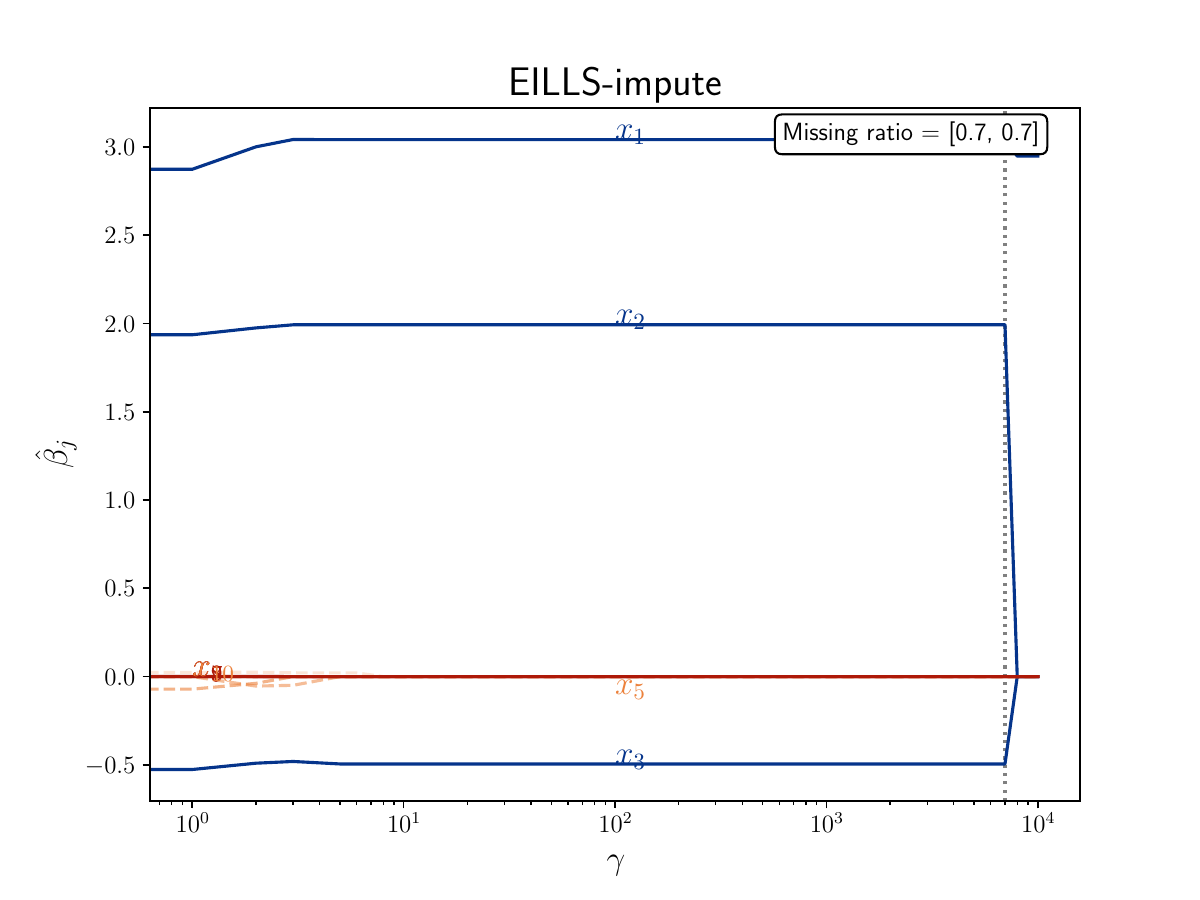} 
        \caption{EILLS-impute.}
        \label{fig:./figures/fig3a/precise_imputation/model1/fig3a_xgboost_eills_impute.pdf}
    \end{subfigure}
    \hfill
    \begin{subfigure}[t]{0.45\textwidth} 
        \centering
        \includegraphics[width=\textwidth]{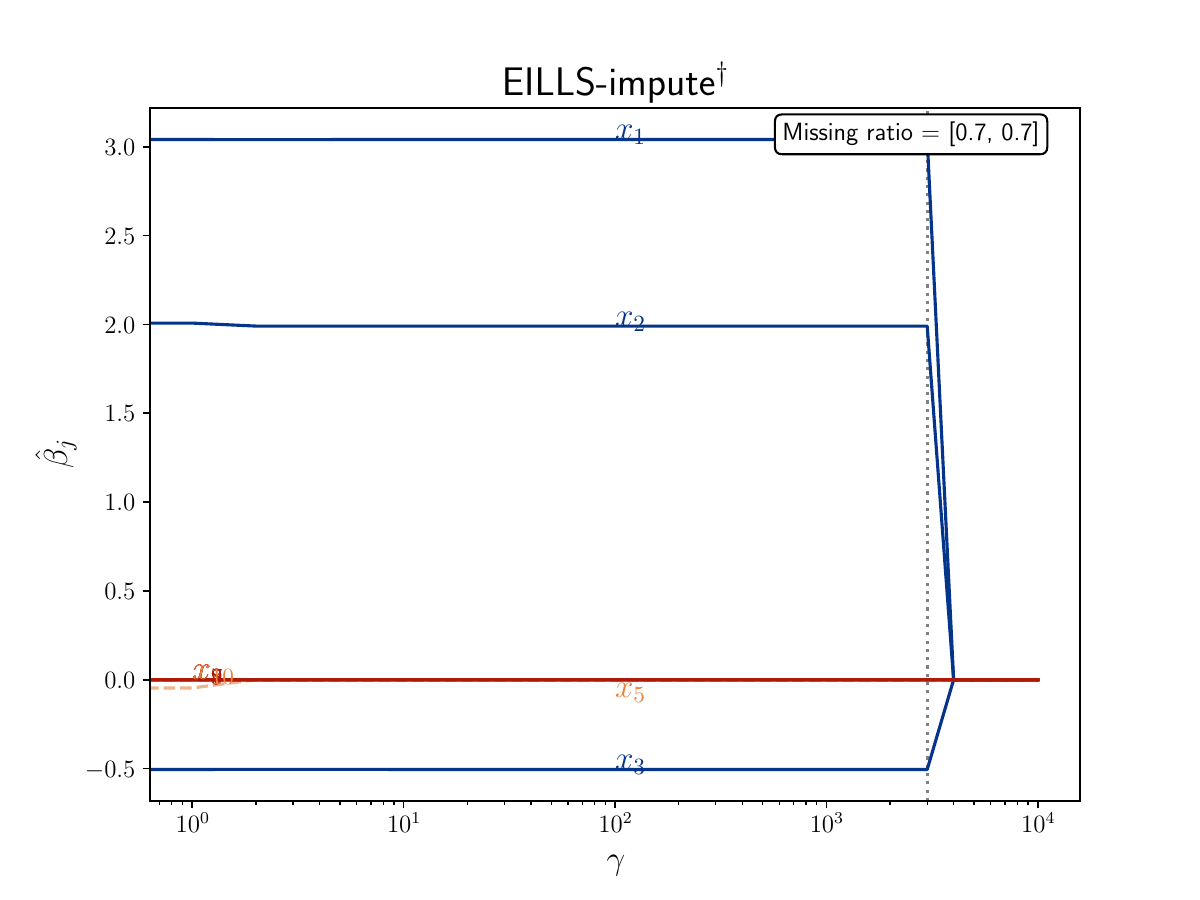} 
        \caption{EILLS-impute$^{\dagger}$.}
        \label{fig:./figures/fig3a/precise_imputation/model1/fig3a_xgboost_eills_impute_ce.pdf}
    \end{subfigure}

	\caption{Variable selection performance of EILLS-impute compared to EILLS-impute$^{\dagger}$ under Model 1 with precise imputation. The suboptimal performance in the left plot is not attributable to biased imputation, as this analysis focuses on precise imputation. The left plot illustrates the performance with the original penalty, while the right plot demonstrates the enhanced stability and accuracy achieved with the improved penalty.}
    
    \label{fig:variable_selection_one_simulation2model1}
\end{figure}

\begin{figure}[H]
    \centering

    \begin{subfigure}[t]{0.45\textwidth} 
        \centering
        \includegraphics[width=\textwidth]{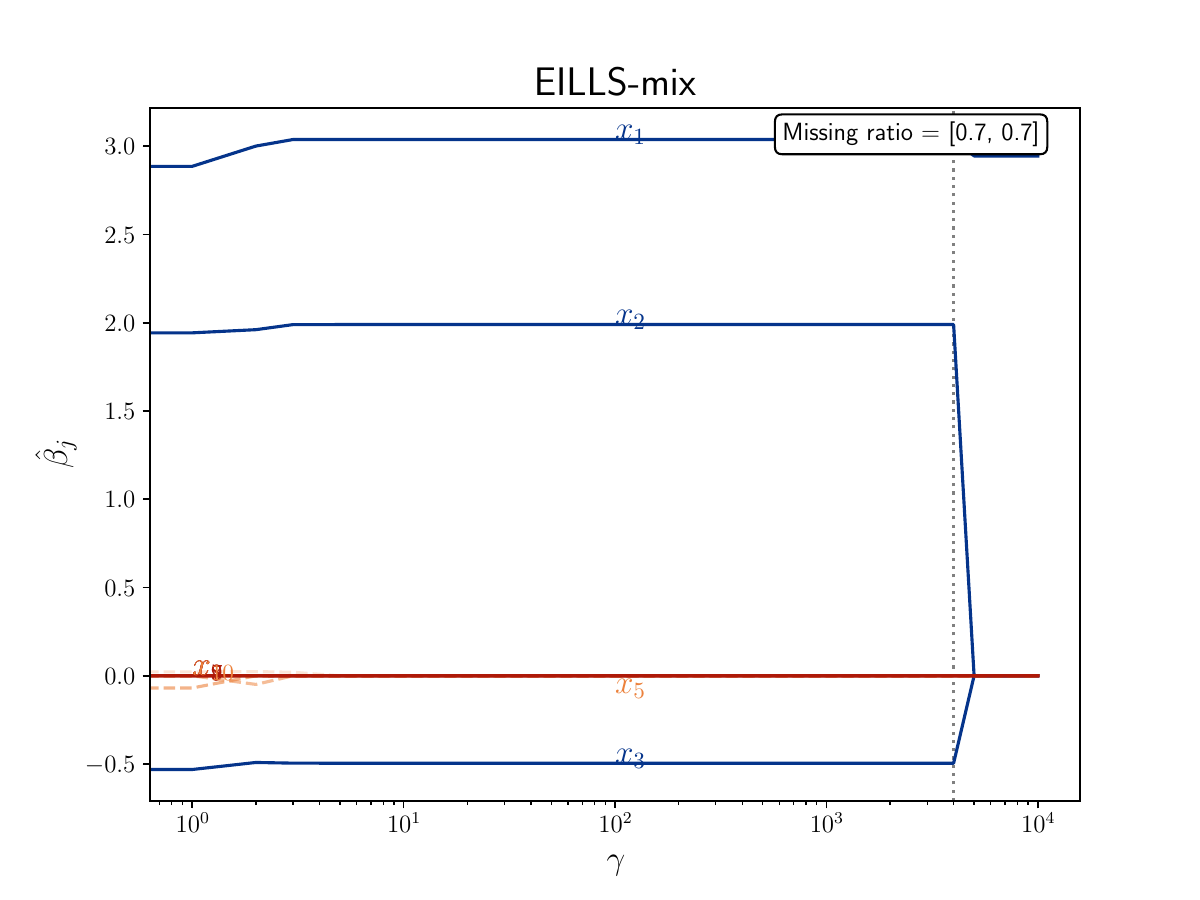} 
        \caption{EILLS-mix.}
        \label{fig:./figures/fig3a/precise_imputation/model1/fig3a_xgboost_eills_mix.pdf}
    \end{subfigure}
    \hfill
    \begin{subfigure}[t]{0.45\textwidth} 
        \centering
        \includegraphics[width=\textwidth]{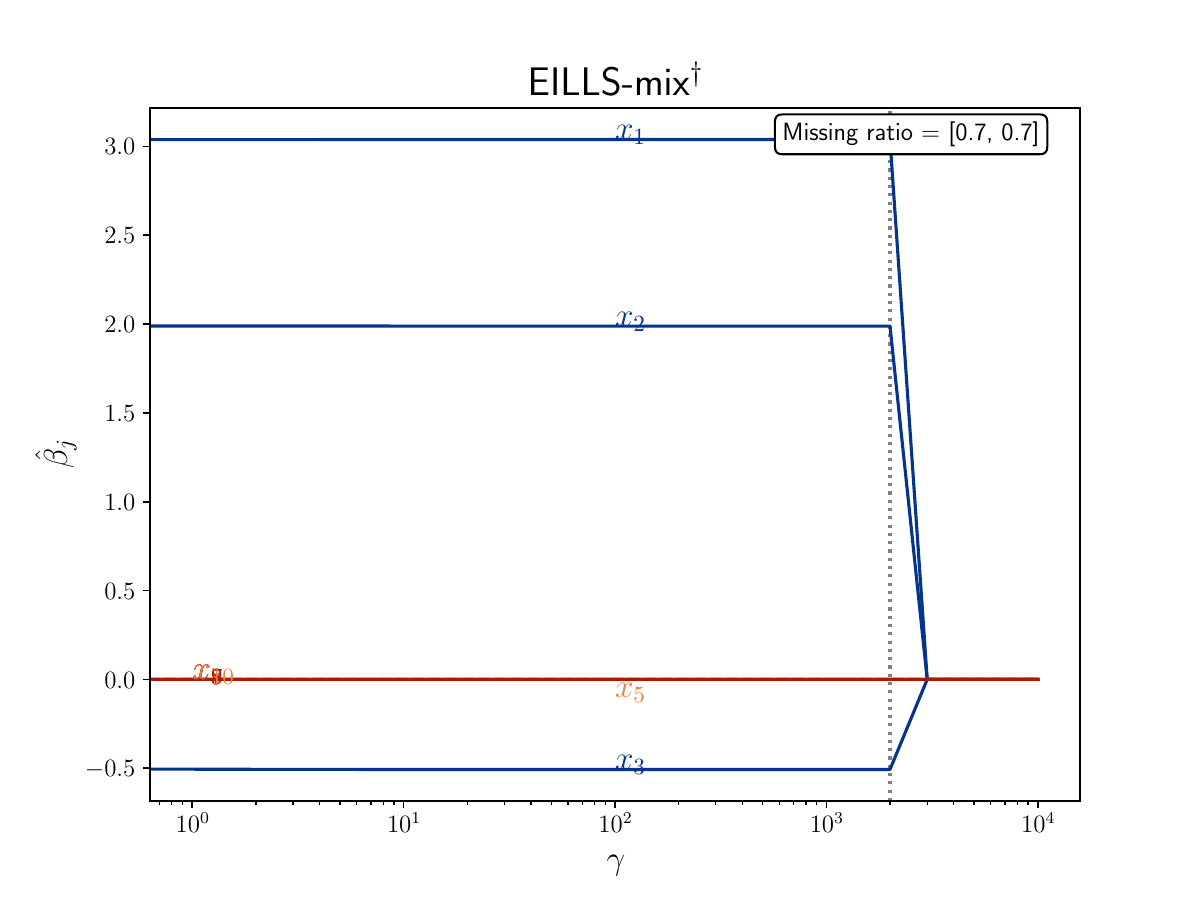} 
        \caption{EILLS-mix$^{\dagger}$.}
        \label{fig:./figures/fig3a/precise_imputation/model1/fig3a_xgboost_eills_mix_ce.pdf.}
    \end{subfigure}

	\caption{Variable selection performance of EILLS-mix vs. EILLS-mix$^{\dagger}$ under Model 1 with precise imputation, where the right plot demonstrates the enhanced penalty’s improved stability and accuracy.}
    
    \label{fig:variable_selection_one_simulation3model1}
\end{figure}

\begin{figure}[H]
    \centering

    \begin{subfigure}[t]{0.45\textwidth} 
        \centering
        \includegraphics[width=\textwidth]{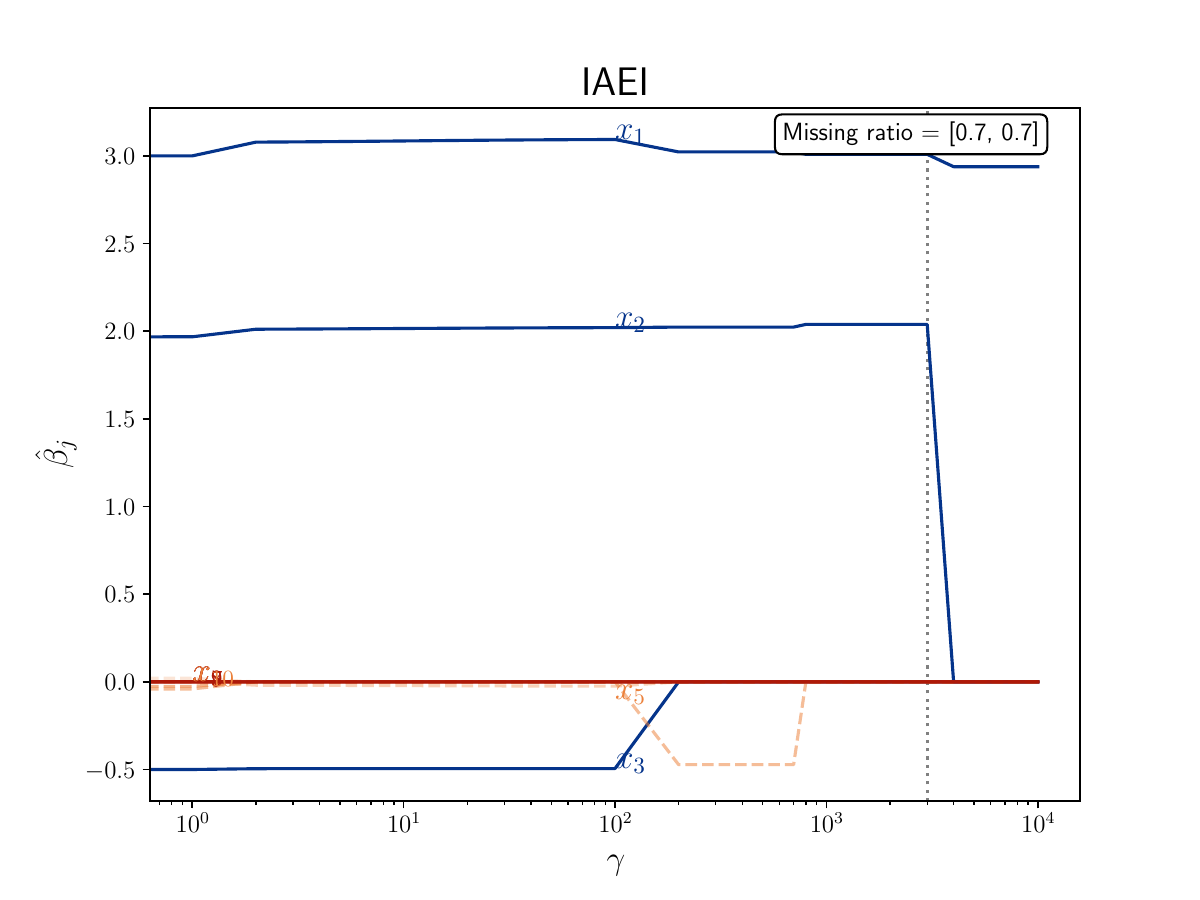} 
        \caption{IAEI.}
        \label{fig:./figures/fig3a/precise_imputation/model1/fig3a_xgboost_iaei.pdf}
    \end{subfigure}
    \hfill
    \begin{subfigure}[t]{0.45\textwidth} 
        \centering
        \includegraphics[width=\textwidth]{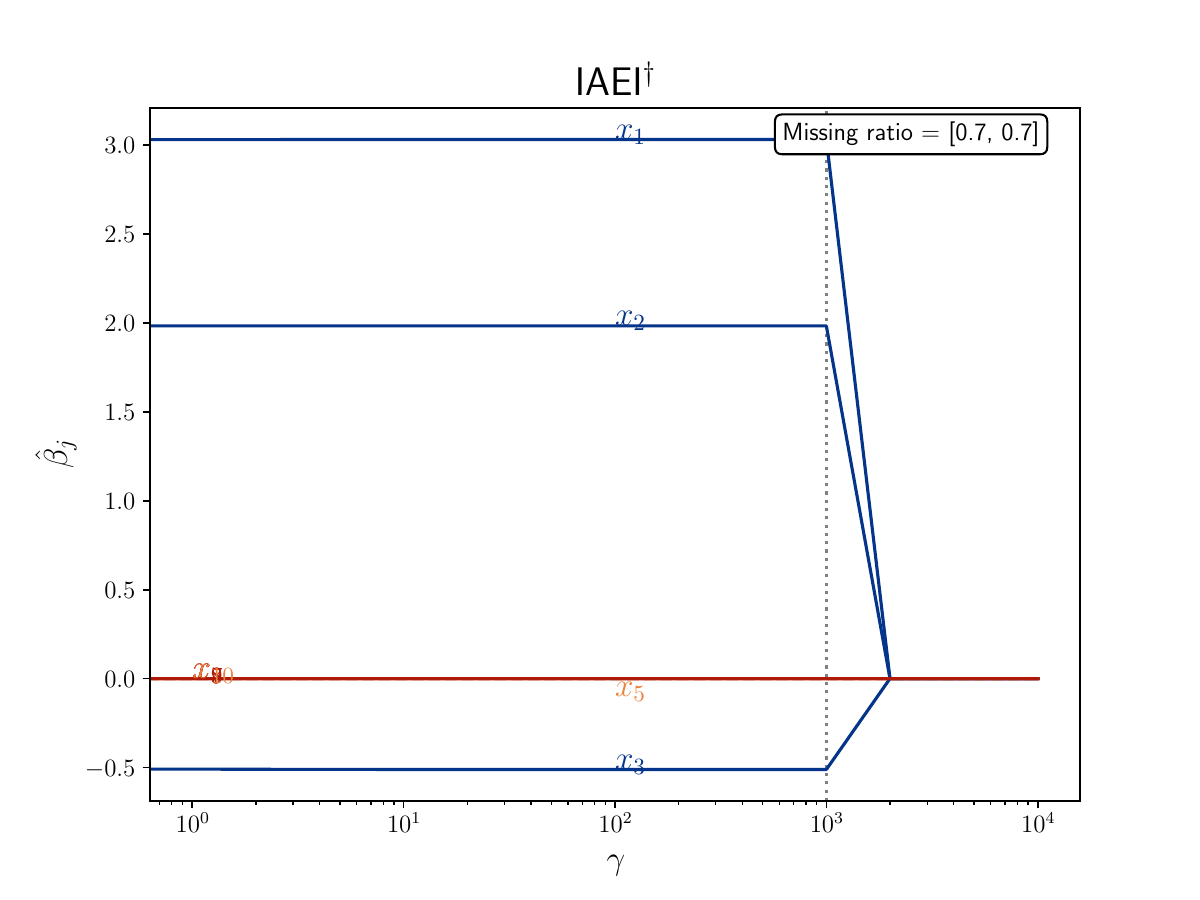} 
        \caption{IAEI$^{\dagger}$.}
        \label{fig:./figures/fig3a/precise_imputation/model1/fig3a_xgboost_iaei_ce.pdf}
    \end{subfigure}
    
    \caption{Variable selection performance of IAEI vs IAEI$^{\dagger}$ under Model 1. While the left plot shows the result with original penalty, the right plot demonstrates the improved stable performance using the enhanced penalty.}
    \label{fig:variable_selection_one_simulation4model1}
\end{figure}

\begin{figure}[H]
    \centering
    \begin{subfigure}[t]{0.45\textwidth} 
        \centering
        \includegraphics[width=\textwidth]{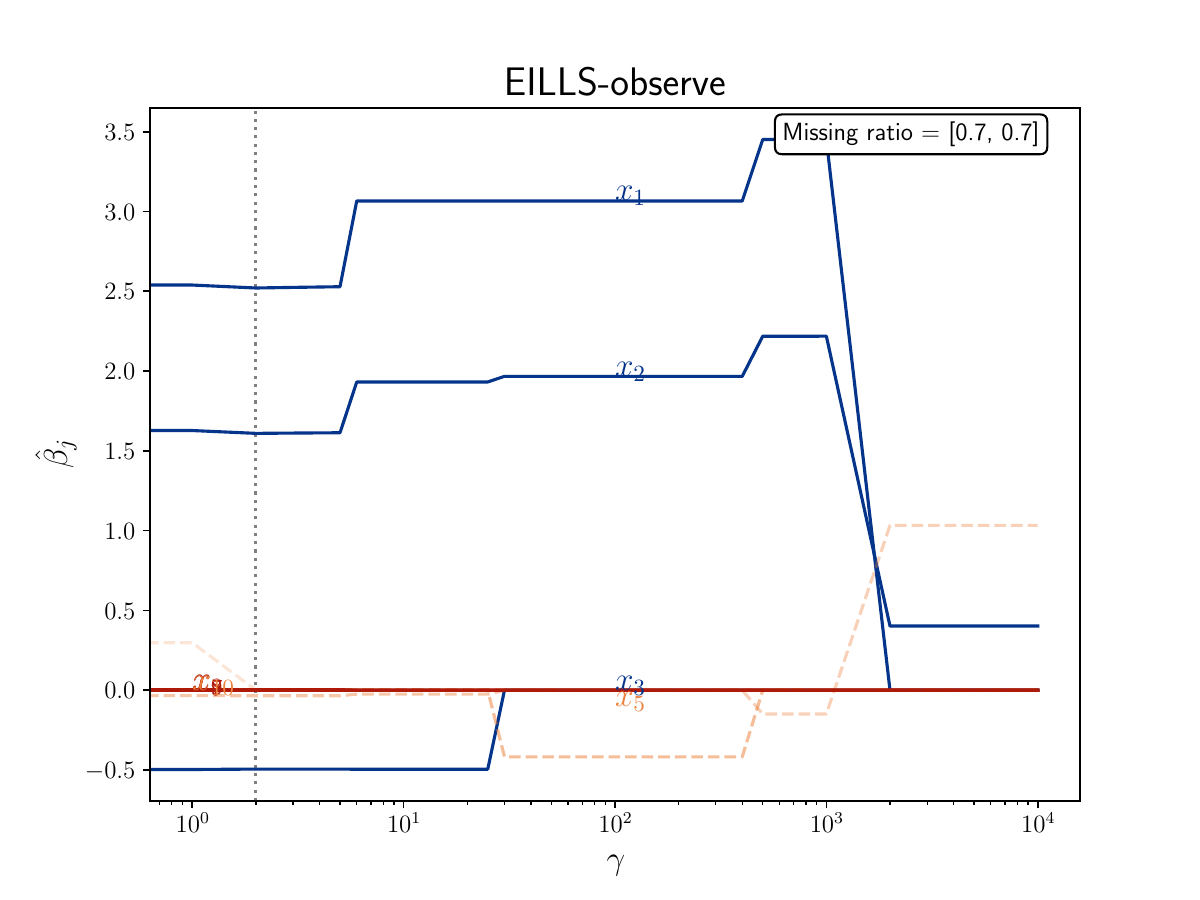} 
        \caption{EILLS-observe.}
        \label{fig:./figures/fig3a/precise_imputation/model2/fig3a_xgboost_eills_observe.pdf}
    \end{subfigure}
    \hfill
    \begin{subfigure}[t]{0.45\textwidth} 
        \centering
        \includegraphics[width=\textwidth]{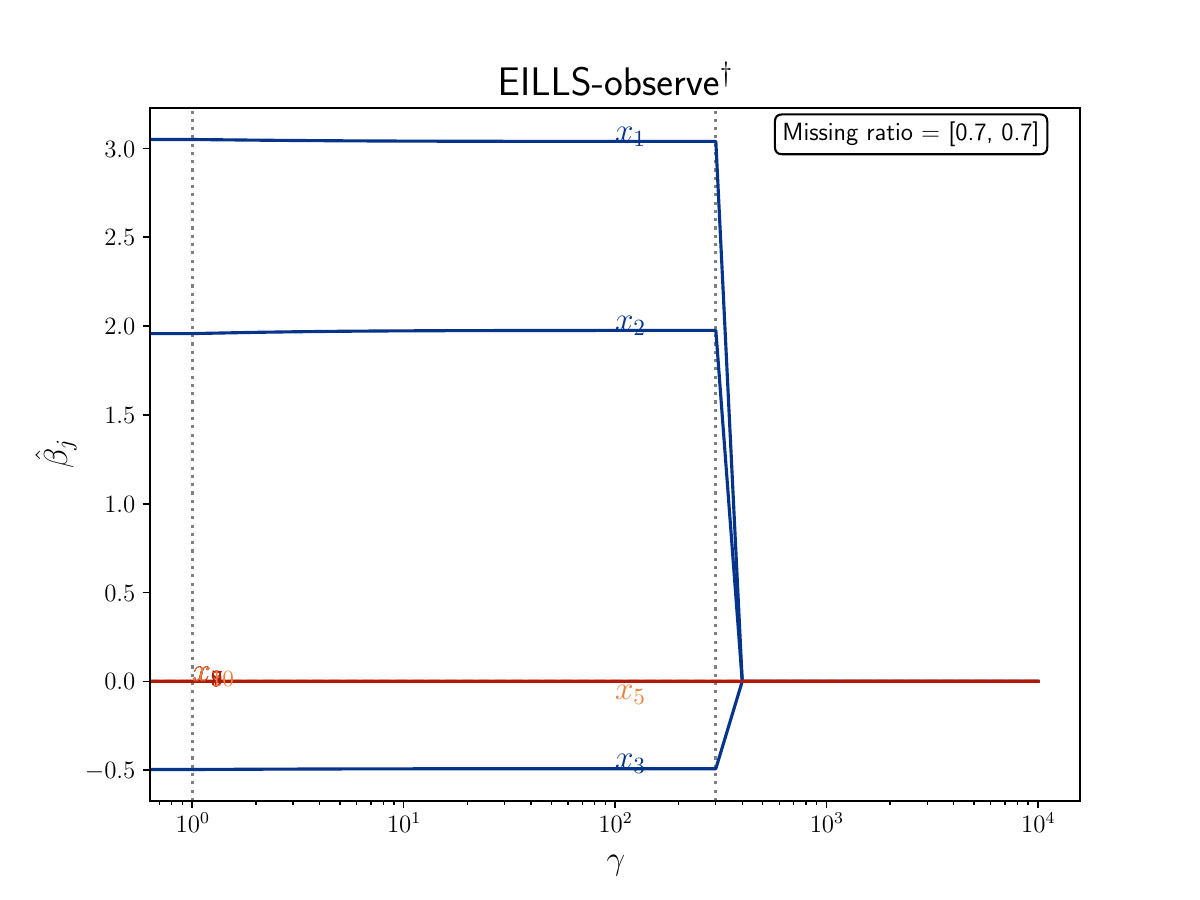} 
        \caption{EILLS-observe$^{\dagger}$.}
        \label{fig:./figures/fig3a/precise_imputation/model2/fig3a_xgboost_eills_observe_ce.pdf}
    \end{subfigure}

	\caption{Variable selection performance of EILLS-observe compared to EILLS-observe$^{\dagger}$ under Model 2. The left plot illustrates the performance using the original penalty, while the right plot highlights the improved stability and accuracy achieved with the enhanced penalty.}
    
    \label{fig:variable_selection_one_simulation1model2}
\end{figure}

\begin{figure}[H]
    \centering
    
    \begin{subfigure}[t]{0.45\textwidth} 
        \centering
        \includegraphics[width=\textwidth]{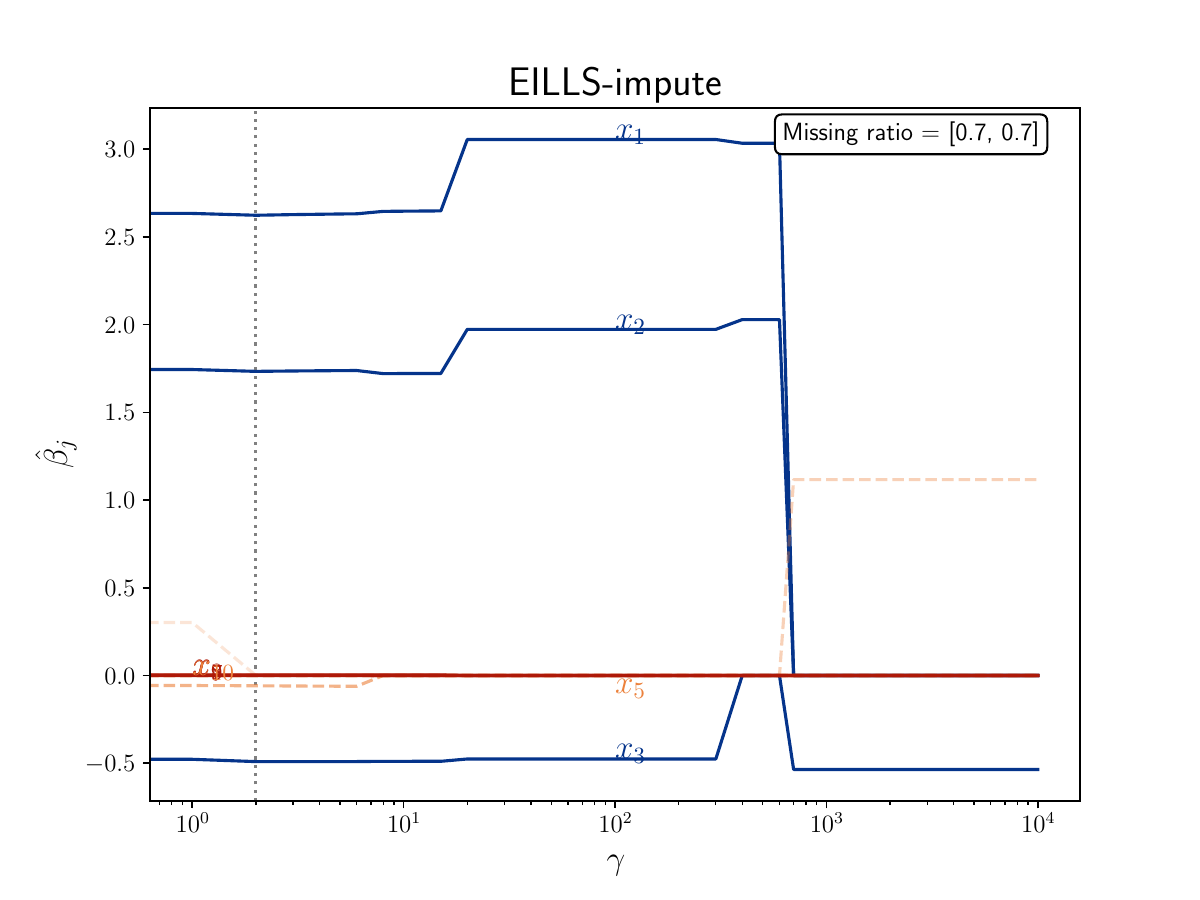} 
        \caption{EILLS-impute.}
        \label{fig:./figures/fig3a/precise_imputation/model2/fig3a_xgboost_eills_impute.pdf}
    \end{subfigure}
    \hfill
    \begin{subfigure}[t]{0.45\textwidth} 
        \centering
        \includegraphics[width=\textwidth]{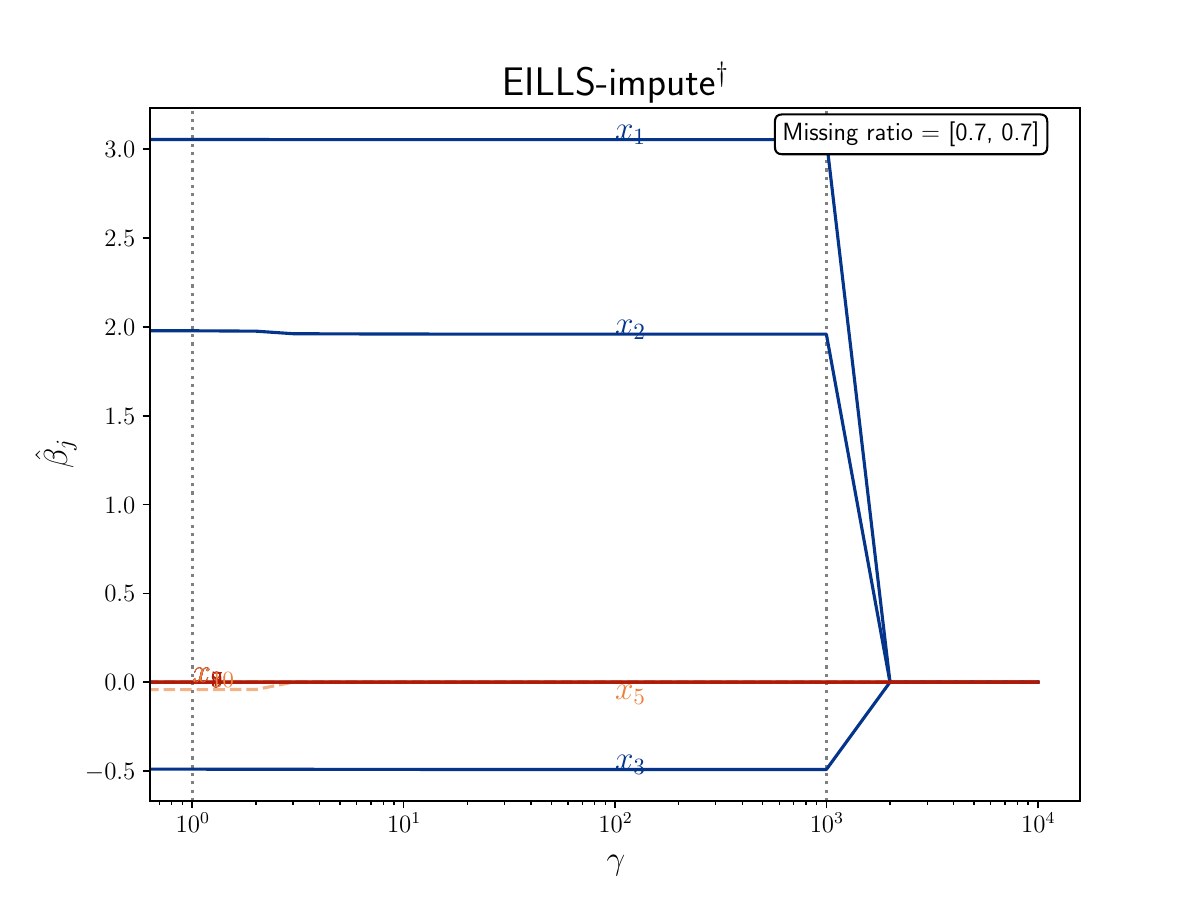} 
        \caption{EILLS-impute$^{\dagger}$.}
        \label{fig:./figures/fig3a/precise_imputation/model2/fig3a_xgboost_eills_impute_ce.pdf}
    \end{subfigure}

	\caption{Variable selection performance of EILLS-impute compared to EILLS-impute$^{\dagger}$ under Model 2 with precise imputation. The suboptimal performance in the left plot is not attributable to biased imputation, as this analysis focuses on precise imputation. The left plot illustrates the performance with the original penalty, while the right plot demonstrates the enhanced stability and accuracy achieved with the improved penalty.}
    
    \label{fig:variable_selection_one_simulation2model2}
\end{figure}

\begin{figure}[H]
    \centering

    \begin{subfigure}[t]{0.45\textwidth} 
        \centering
        \includegraphics[width=\textwidth]{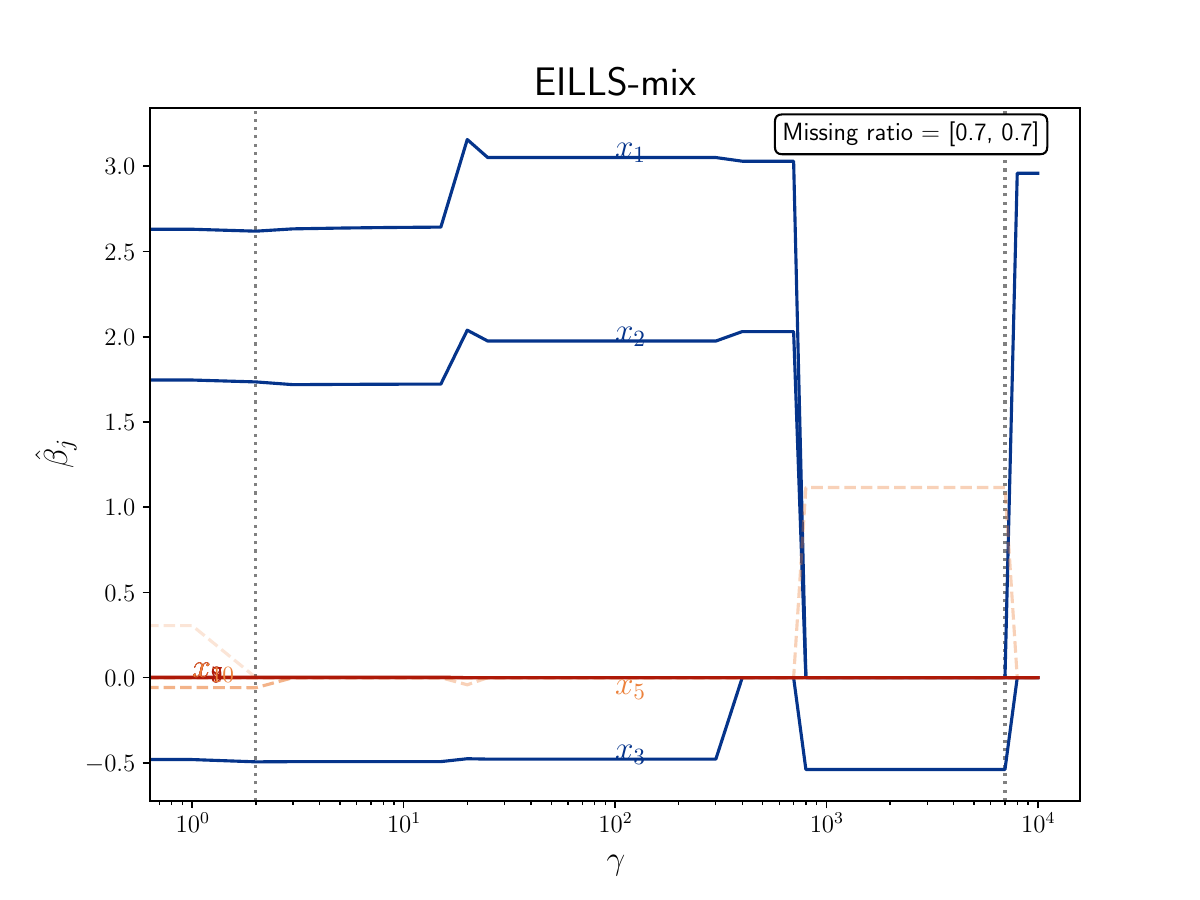} 
        \caption{EILLS-mix.}
        \label{fig:./figures/fig3a/precise_imputation/model2/fig3a_xgboost_eills_mix.pdf}
    \end{subfigure}
    \hfill
    \begin{subfigure}[t]{0.45\textwidth} 
        \centering
        \includegraphics[width=\textwidth]{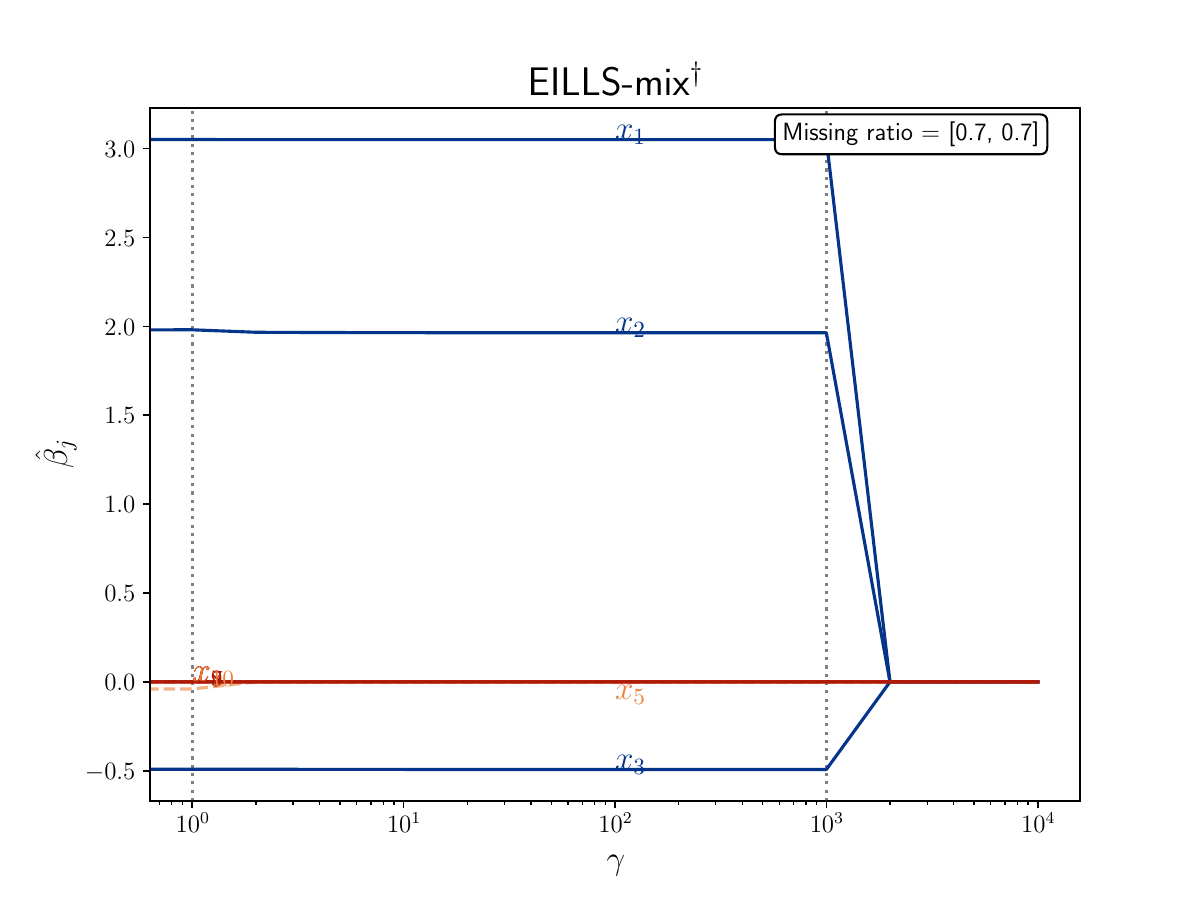} 
        \caption{EILLS-mix$^{\dagger}$.}
        \label{fig:./figures/fig3a/precise_imputation/model2/fig3a_xgboost_eills_mix_ce.pdf.}
    \end{subfigure}

	\caption{Variable selection performance of EILLS-mix vs. EILLS-mix$^{\dagger}$ under Model 2 with precise imputation, where the right plot demonstrates the enhanced penalty’s improved stability and accuracy.}
    
    \label{fig:variable_selection_one_simulation3model2}
\end{figure}

\begin{figure}[H]
    \centering

    \begin{subfigure}[t]{0.45\textwidth} 
        \centering
        \includegraphics[width=\textwidth]{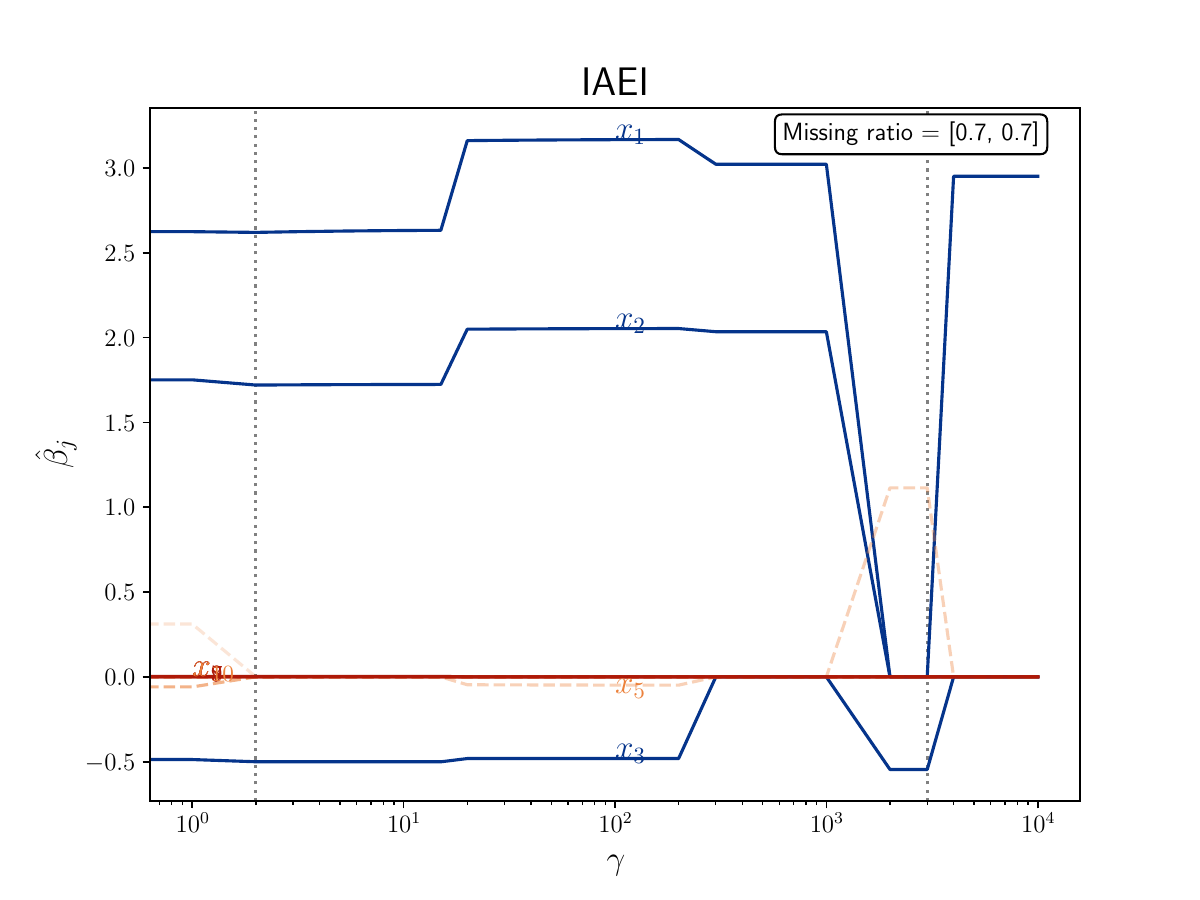} 
        \caption{IAEI.}
        \label{fig:./figures/fig3a/precise_imputation/model2/fig3a_xgboost_iaei.pdf}
    \end{subfigure}
    \hfill
    \begin{subfigure}[t]{0.45\textwidth} 
        \centering
        \includegraphics[width=\textwidth]{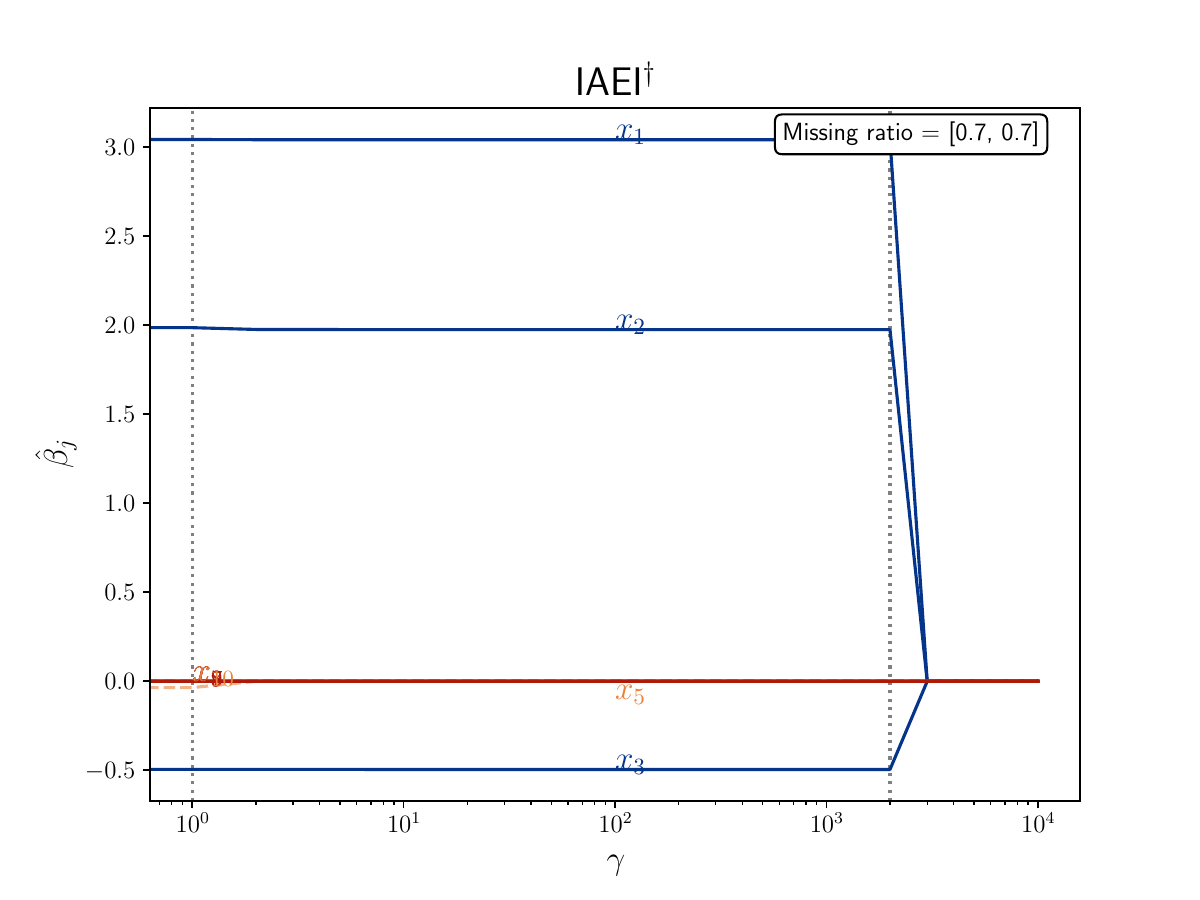} 
        \caption{IAEI$^{\dagger}$.}
        \label{fig:./figures/fig3a/precise_imputation/model2/fig3a_xgboost_iaei_ce.pdf}
    \end{subfigure}
    
    \caption{Variable selection performance of IAEI vs IAEI$^{\dagger}$ under Model 2. While the left plot shows the result with original penalty, the right plot demonstrates the improved stable performance using the enhanced penalty.}
    \label{fig:variable_selection_one_simulation4model2}
\end{figure}

\subsubsection{Performance on $\ell_2$ Error Convergence}\label{additional_simulation:l2_error_convergence}

\begin{figure}[H]
    \centering
    \begin{subfigure}[t]{0.45\textwidth}
        \centering
        \includegraphics[width=\textwidth]{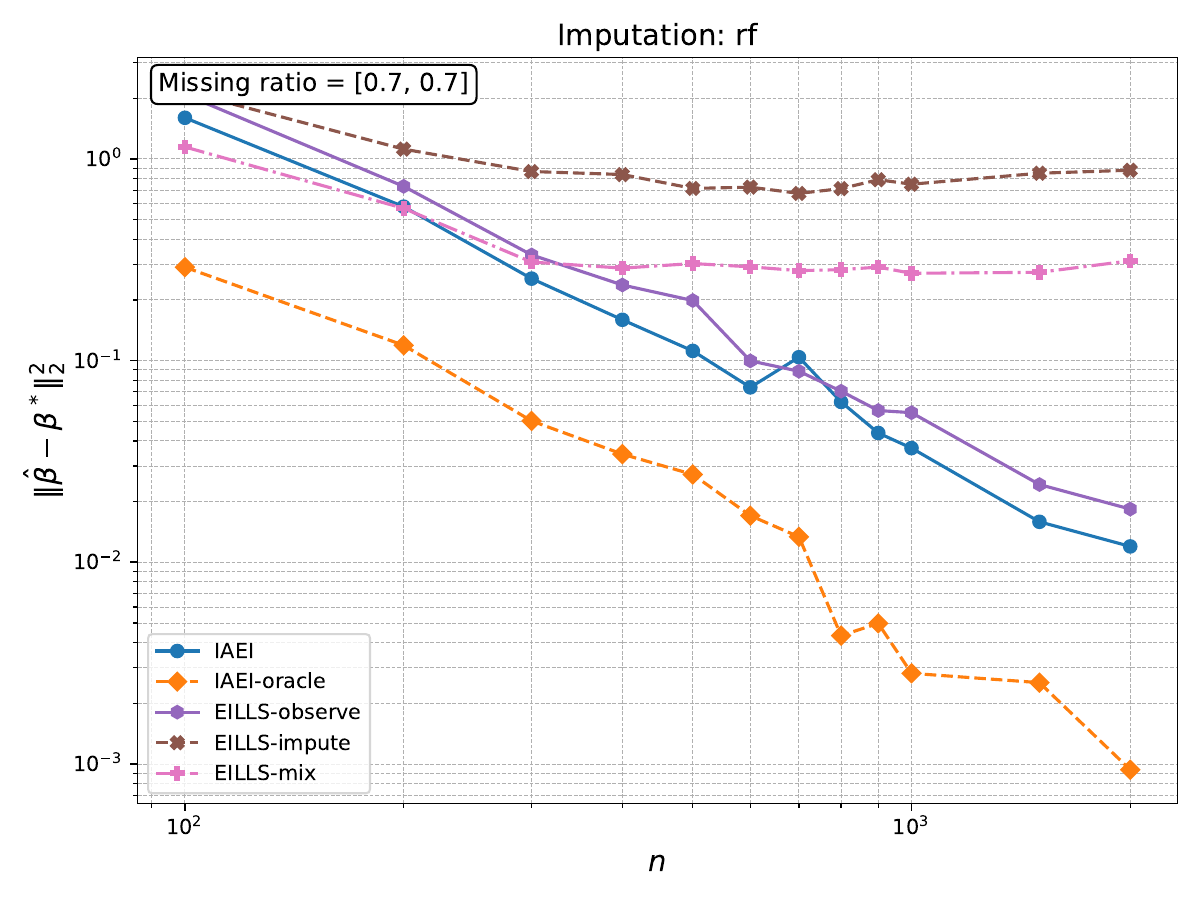}
        \caption{Compare methods in Model 0.}
        \label{fig:./figures/fig3b_and_fig3b1_and_fig3a1/bias_imputation/model0/fig3b_bias_model0_model0_0.7_rf_simple.pdf}
    \end{subfigure}
    \hfill
    \begin{subfigure}[t]{0.45\textwidth}
        \centering
        \includegraphics[width=\textwidth]{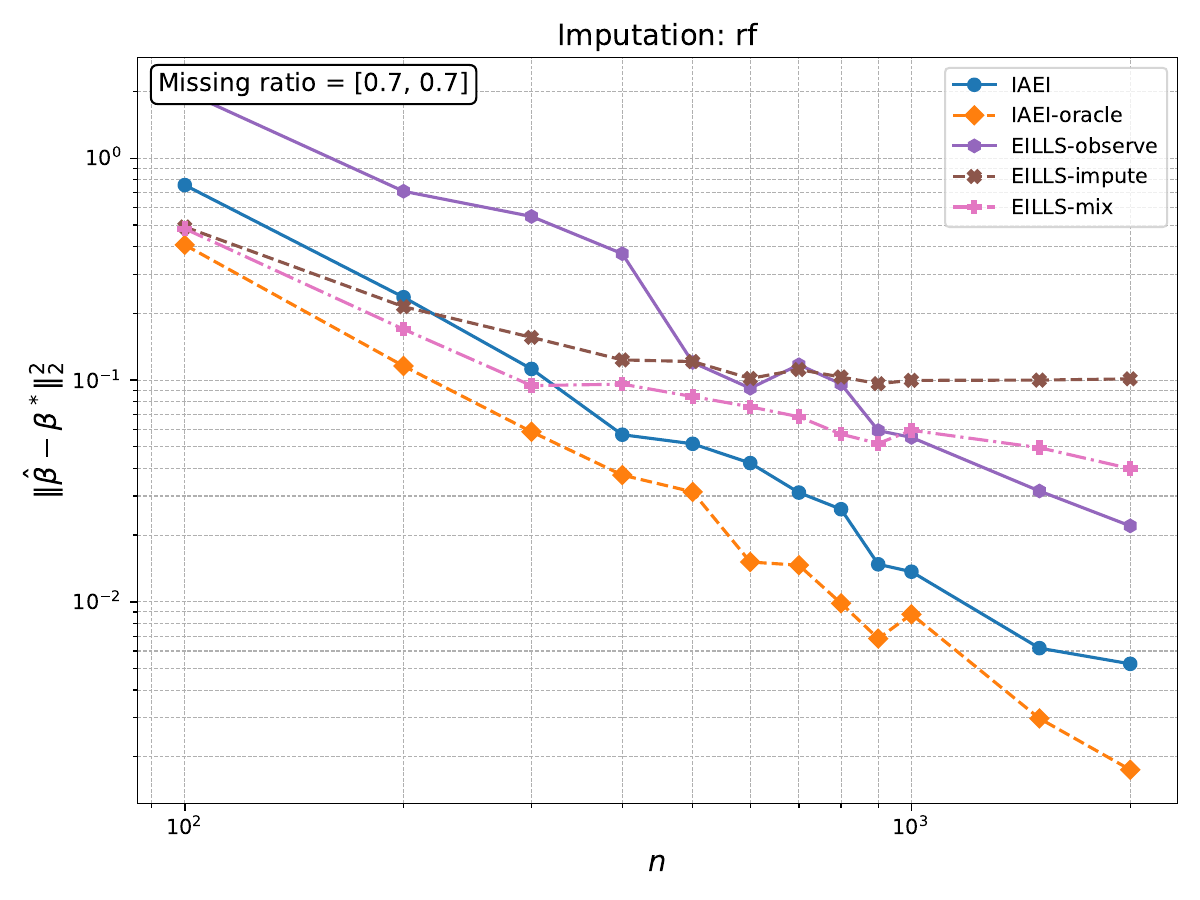}
        \caption{Compare methods in Model 1.}
        \label{fig:./figures/fig3b_and_fig3b1_and_fig3a1/bias_imputation/model1/fig3b_bias_model1_model0_0.7_rf_simple.pdf}
    \end{subfigure}

    \vspace{1em} 

    \begin{subfigure}[t]{0.45\textwidth}
        \centering
        \includegraphics[width=\textwidth]{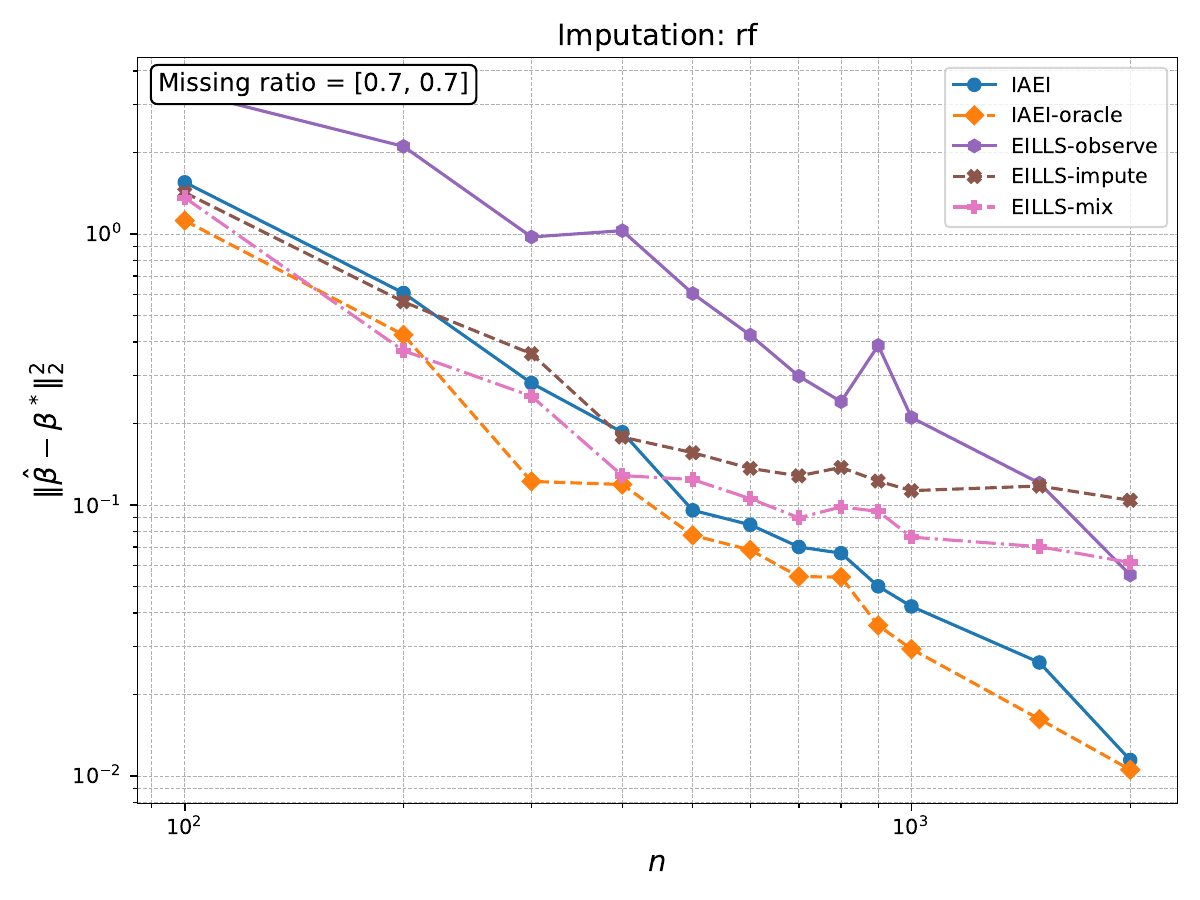}
        \caption{Compare methods in Model 2.}
        \label{fig:./figures/fig3b_and_fig3b1_and_fig3a1/bias_imputation/model2/fig3b_bias_model2_model0_0.7_rf_simple.pdf}
    \end{subfigure}
    \hfill
    \begin{subfigure}[t]{0.45\textwidth}
        \centering
        \includegraphics[width=\textwidth]{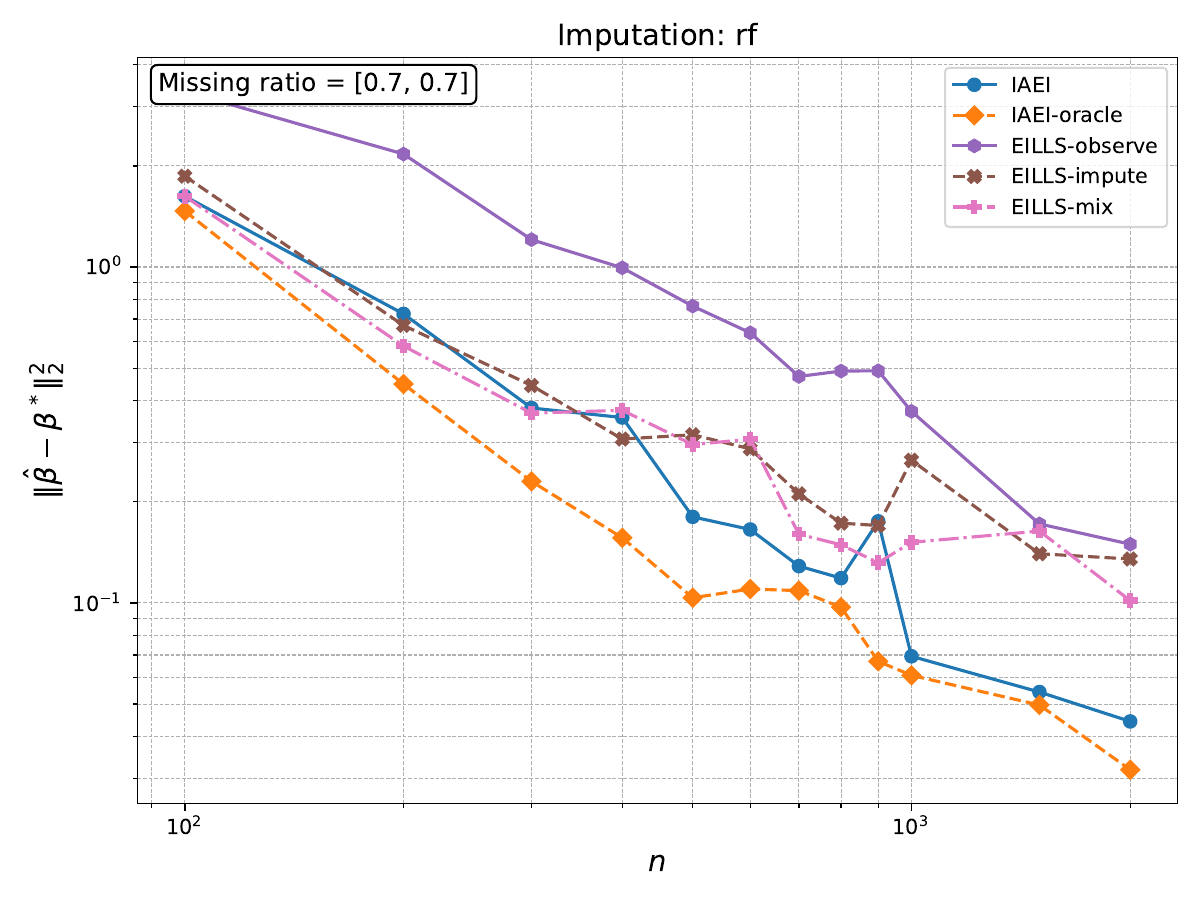}
        \caption{Compare methods in Model 3.}
        \label{fig:./figures/fig3b_and_fig3b1_and_fig3a1/bias_imputation/model3/fig3b_bias_model3_model0_0.7_rf_simple.pdf}
    \end{subfigure}
    \caption{In all Models 0–3, IAEI with RandomForest imputation consistently outperforms the other methods. Notably, in Models 1–3, its performance is significantly superior.}
    \label{fig:fig:./figures/fig3b_and_fig3b1_and_fig3a1/bias_imputation/model2/fig3b_bias_model2_model0_0.7_rf_simple.pdffig:./figures/fig3b_and_fig3b1_and_fig3a1/bias_imputation/model3/fig3b_bias_model3_model0_0.7_rf_simple.pdf}
\end{figure}

\end{document}